\documentclass[runningheads]{llncs}
\usepackage{amsmath}
\usepackage{cases}
\usepackage{graphicx}
\usepackage{caption}
\usepackage{subcaption}
\usepackage{afterpage}
\usepackage{amssymb}
\usepackage{mathtools}
\usepackage{textcomp, gensymb}
\DeclareMathOperator*{\argminA}{arg\,min} 
\newcommand{\textrb}[1]{\textrm{\textbf{#1}}}

\begin{document}
\title{Neural Radiance Fields: Past, Present, and Future}
\titlerunning{NeRFs: Past, Present, Future}
\author{Ansh Mittal\inst{1}\orcidID{0000-0001-6648-6064}}
\authorrunning{A. Mittal}
\institute{University of Southern California, Los Angeles CA 90007, USA\\
\url{https://www.usc.edu/} \and
\email{\{anshm\}@usc.edu}}
\maketitle
\begin{abstract}
The various aspects like modeling and interpreting 3D environments and surroundings have enticed humans to progress their research in 3D Computer Vision, Computer Graphics, and Machine Learning. An attempt made by Mildenhall et al in their paper about NeRFs (Neural Radiance Fields) led to a boom in Computer Graphics, Robotics, Computer Vision, and the possible scope of High-Resolution Low-Storage Augmented Reality and Virtual Reality-based 3D models have gained traction from res with more than 500 preprints related to NeRFs published. This paper serves as a bridge for people starting to study these fields by building on the basics of Mathematics, Geometry, Computer Vision, and Computer Graphics to the difficulties encountered in Implicit Representations at the intersection of all these disciplines. This survey provides the history of rendering, Implicit Learning, and NeRFs, the progression of research on NeRFs, and the potential applications and implications of NeRFs in today's world. In doing so, this survey categorizes all the NeRF-related research in terms of the datasets used, objective functions, applications solved, and evaluation criteria for these applications.
\keywords{NeRFs (Neural Radiance Fields) \and Rendering \and Computer Vision \and Deep Learning \and Computer Graphics \and Robotics \and Augmented Reality \and Virtual Reality \and Implicit Machine Learning}
\end{abstract}

\section{Introduction}
\label{intro}
Recently, implicit representation functions and machine learning have pervaded through different applications and problem statements unsolvable in traditional machine learning. And applications related to View Synthesis, 3D Rendering, and Modeling have also started to leverage Radiance Fields (also known as NeRFs). The preprints papers in this domain have been rising exponentially with applications in different disciplines like 3D Computer Vision, Astronomy, Climate change, and many other areas of study that require images/3D geometry of locations. Although the work in Radiance Fields started as a means to offer better View Synthesis, after the development in the literature and GPU technology, it has permeated other fields such as Image Reconstruction, Super Resolution, Pose Estimation, 3D Aware Image Synthesis, Depth Estimation, Image Generation, 3D Reconstruction, Neural Rendering, and other computer vision related problems. So, this survey aims to comprehensively categorize the models, extensions, and applications of NeRFs in terms of datasets, objective functions used, and problem statements tackled, evaluation measures adopted while discussing the techniques and key concepts associated with the field of NeRFs (Neural Radiance Fields) in the research literature. Despite the comprehensiveness of surveys related to Differentiable Rendering, Neural Radiance Fields, and Differential Geometry, they fail to cater to someone starting research in this field. So, this survey is able to bridge that gap while also discussing the history of how NeRF came to be.

This research paper make the following important contributions to the already existing research literature (such as Gao et al~\cite{gao2022nerf}).
\begin{enumerate}
    \item This paper can be used by a layman to understand basic concepts of mathematics, computer graphics, computer vision, geometry (differential geometry),  etc.
    \item It makes a distinct classification based on the Loss Functions, Evaluation metrics, Applications, and by the year it became online (on arxiv or other sources on the Internet)
    \item It comprehensively tries to discuss the development of Neural Radiance Fields and how Fourier Features can be used to model high-frequency function for rendering~\cite{tancik2020fourier,mildenhall2019local,sitzmann2020implicit}
    \item Further, based on the past and the papers published in 2020--2023, this paper tries to trace out a path of future development of Neural Radiance Fields and Neural Rendering
\end{enumerate}

The \S\ref{intro} presents a brief introduction that this paper aims to resolve i.e. to delineate a classification for NeRF models and discuss about its history and future. The remainder of this paper has been organized as follow. \S\ref{history} discusses the various neural architectures before the conceptions of NeRFs. It further discusses various surveys that have been put forward in this domain and the need for this survey. \S\ref{present:papers} discusses the chronological description of the models, extensions, and improvements over Neural Radiance Fields. \S\ref{present:datasets} briefly discusses the datasets that have been leveraged till now in the field of NeRFs. Further, these are also then taxonomized according to different objectives (discussed in \S\ref{present:obj_funct}), different evaluation metrics discussed in \S\ref{present:evalmetrics}), and different applications (discussed in \S\ref{present:diffexploredappns}). All sections from \S\ref{present:datasets}--\ref{present:diffexploredappns} discuss the research in the past couple of years in the field of NeRFs (limited till December 2022). Finally, the future prospects of this family of models is explored in \S\ref{future}. \S\ref{conclusion} concludes the paper and discusses some future prospects for this research.

\section{Basic Principles and Concepts}
\label{basic_concepts}
This section defines the concepts that should be helpful to the reader when considering the possible objective functions and applications that Radiance Fields tries to solve. This is by no means a comprehensive list of concepts and the reader is advised to search on his own. For instance, Voxel Grids, Octree, Triangular meshes, and Radiance Fields (from 2020 onwards) are some of the ways that are used to store and represent 3D geometries.\\
These concepts can be divided into the basics of Computer Graphics and Computer Vision. Since, NeRFs combine the fields of Computer Graphics, Computer Vision, and Signal Processing, it is important to understand all these concepts for better understanding the rest of the literature.

\subsection{Computer Graphics}
\label{cgraph}
This section briefly describes some terms in Computer Graphics and that are going to be helpful in reading the remainder of this article.
\begin{itemize}
\item \textbf{Pixel:} In computer graphics, a pixel (short for "picture element") is the smallest unit of a digital image. It represents a single point in a 2D grid of square or rectangular cells called a raster image. Each pixel contains information about its color and brightness, typically represented using binary values. The combination of millions of pixels arranged in a specific pattern creates a digital image that can be displayed on a screen or printed on paper.\\
In computer vision, a pixel is also the smallest unit of a digital image, but its meaning is slightly different. In this context, a pixel is a mathematical representation of an image that enables algorithms to analyze and process visual information. Each pixel is assigned a numerical value based on its intensity, which can range from 0 (black) to 255 (white) in grayscale images. In color images, each pixel is represented by three or four values that define its red, green, and blue color components, and optionally its alpha channel for transparency.\\
In both computer graphics and computer vision, pixels are fundamental to the creation and processing of digital images.

\item \textbf{Resel:} In computer graphics research, a resel (short for "resolution element") is a unit of measurement used to quantify the resolution of an image. It is a mathematical construct that represents the smallest distinguishable element in an image, based on the size of the image's point spread function (PSF).\\
The PSF describes how a point of light in the scene is blurred by the imaging system before being recorded by a camera or other sensor. The resel is defined as the area of the PSF that contains 50\% of the total energy of the point of light. In other words, it represents the smallest area in the image that can be resolved as a distinct feature.\\
The resel is a useful metric in computer graphics research because it can be used to compare the resolution of different imaging systems, or to evaluate the effectiveness of image processing algorithms in enhancing or preserving image detail. It is also used in fields such as microscopy and astronomy to quantify the resolution of optical systems.

\item \textbf{Voxel} is a commonplace term used in 3D Computer Graphics and 3D Computer Vision. It represents a discrete value on a grid in 3D space. It is analogous to Pixel (in 2D bitmap space). Voxels don't have their position information encoded but a voxel is rendered based on relative position to other voxels. A voxel has heavy use-case in GIS systems~\cite{chmielewski2017estimating}, Healthcare systems based on Computer Vision~\cite{gillmann2021visualizing,galbusera2019artificial,smiti2020machine,cruz2021deepcsr,mittal2019aicnns}, and Autonomous Vehicles~\cite{li2020deep,cui2021deep,cortinhal2020salsanext,chen2019pct}. These can be used to form another data structure which is discussed further.
\item \textbf{Tixel} or a tactile pixel is smallest measuring/transmitting element of a tactile element of a tactile matrix. This is one of the fundamental units for Haptic Technologies~\cite{kaltenbrunner2013ultra}. Here, the tactile matrix is a machine-readable system, which transmits information from surface to the receiver that is the exoskeleton interface for tactile capturing.
\item \textbf{Texel} or Texture element or Texture equivalent of a Pixel, is the most fundamental measuring unit of Texture. Textures have an array of texels (much like our computer images which have arrays of pixels). They are image regions regions obtained through simple procedures like Thresholding and the midpoint of centroids of two surrounding texels can be used to define the spatial relationship between two texels in a texture.
\item \textbf{Image Sharpening} refers to techniques that highlight the edges and finer details in an image. It is widely used as a preprocessing step for object classification. This is done to increase local contrast and sharpening image. In essence, it works by adding a high-pass filtered version of original image to the image again. It is one of the most commonly used process for sharpening one-dimensional signals and is called unsharp masking. Since, an image can be considered a 2D-signal, hence, this operation can also be used for images. This operation can also be represented by the eq~\ref{eqn:21}
\begin{equation}\label{eqn:21}
    S_{i,j} = \hat{x_{i,j}} + \lambda \cdot \textrm{F}(\hat{x_{i,j}})
\end{equation}

where $\hat{x_{i, j}}$ is the original pixel values for the image at the coordinates $(i,j)$, $\textrm{F}(\cdot)$ is the high-pass filter for the image, $\lambda$ is the tuning parameter greater than or equal to zero, whereas the RHS represents the sharpened pixel for the image at the coordinates $(i,j)$. $\lambda$ decides the desirable sharpness of the image. In case of RGB images, $x_{i,j}$, $\lambda$, and $S_{i,j}$ become vectors with 3D. 
\item \textbf{Data Storage techniques} These are some of the data storage techniques in 3D Computer Vision and Computer Graphics. This list might not be an exhaustive list but includes most of the formats referred during the future discussions in this paper.
\begin{itemize}
\item \textbf{Voxel Grids} represent the grid created by voxels, which are the extensions of ``Pixels'' into the third dimension. Each voxel represents a small cubic volume in space and can store various properties such as occupancy, color, or texture. Voxel grids are commonly used for tasks such as 3D reconstruction, scene understanding, and object recognition. They offer several advantages over other data structures, such as point clouds or meshes, including Regularty, Completeness, and Efficiency. Voxel grids can be constructed from various sources of 3D data, such as point clouds, depth maps, or mesh models. The size and resolution of the voxel grid can be adjusted to balance the trade-off between accuracy and computational efficiency. Once constructed, voxel grids can be used for a variety of tasks such as object segmentation, feature extraction, and shape analysis.

\item \textbf{Polygon (Triangular) Meshes} is a collection of faces, vertices, and edges leveraged to define the shape of the polyhedra object under consideration in 3D Computer Vision, Computer Graphics. These meshes might contain only triangular (Triangular Meshes), Quadrilaterals (Quads), simple Polygons (N-gons) as this can help in simplifying rendering. Occassionally, they might be composed of Polygon with Holes or Concave Polygons.
\item \textbf{QuadTree} is a tree data structure to partition a two-dimensional space by recursively dividing it into four quadrants or regions. Data at leaf node varies but the leaf generally represent a ``unit of interesting spatial information''. The subdivided regions may be square or rectangular, or can even have other arbitrary shapes. It was first coined in 1974 by Finkel et al~\cite{finkel1974quad}. Quadtrees and its variants generally tend to decompose the space into adaptable cells. Each of these cells have a maximum capacity and the cells can split when the maximum capacity is reached. A Tree-Pyramid is the spatial analog of the complete trees. Hence, each of its nodes has four child nodes except the leaf nodes. Further, all leave are on the same level. The data in the Tree-Pyramid can be stored compactly in a Binary heap~\cite{sonka2014image}.
\item \textbf{Octree} is the 3D equivalent of a Quadtree where each node has exactly eight children nodes. This data structure can be used to partition a 3D space by dividing it into eight octants and is often used in 3D graphics and game engines such as Unity and Unreal Engine. The term was pioneered by Donald Meagher at Rensselaer Polytechnic Institute~\cite{meagher1980octree}. Octree, much like quadtrees may be used as Point-Region Octree (PR-Octree; node stores an explicit 3D point, the center of subdivision of that node; point defines one corner for all eight children), Matrix-based Octree (MX-Octree; )
\item \textbf{Point Cloud} is a discrete representation of points in space. These points may represent a 3D object or a shape. 3D scanners or photogrammetry softwares are used for generating a Point Cloud. These types of representations are used for several purposes such as 3D computer-aided design (CAD) models for manufacturing parts, for metrology, quality inspections, animations, rendering, etc.
\end{itemize}
\item \textbf{Surface Normals} describe the orientation of a surface at a particular point. A surface normal is a vector that is perpendicular to the surface at the given point. Surface normals are often used to determine the shading or lighting of a 3D object, and are a key component in many rendering algorithms. The surface normal vector can be computed using the gradient of the surface at the given point. Eqn~\ref{eqn:21-0-1} defines the gradient for a smooth, continuous surface.
\begin{equation}
\label{eqn:21-0-1}
\Delta f(x,y,z) = (\frac{\partial f}{\partial x}, \frac{\partial f}{\partial y}, \frac{\partial f}{\partial z}) 
\end{equation}
where $f(x,y,z)$ is a scalar function that describes the surface. The surface normal vector can then be computed by normalizing the gradient vector as given by the eqn~\ref{eqn:21-0-2}.
\begin{equation}
    \label{eqn:21-0-2}
    \mathcal{N} = \frac{\Delta f(x,y,z)}{||\Delta f(x, y, z)||} 
\end{equation}
where $||v||$ denotes the magnitude of vector $v$. The resulting $\mathcal{N}$ vector is a unit vector that points in the direction of the surface normal.\\
Surface normals can be computed for various types of 3D data, such as point clouds, meshes, or voxel grids. In practice, surface normals are often computed using a local neighborhood of points or vertices, such as by averaging the normals of adjacent triangles in a mesh or by fitting a plane to a set of nearby points in a point cloud.

\item \textbf{Transmittance} refers to the fraction of light that passes through a medium as it travels from one point to another. It is often used in computer graphics to model the effects of atmospheric scattering on the appearance of objects in a scene. Transmittance is also used in machine learning to describe the ability of a material or substance to transmit light through it, which is important in fields such as optics and photonics.
\item \textbf{Reflectance} is the fraction of incident light that is reflected by a surface. It is often used in computer graphics and machine learning to model the appearance of materials and to infer properties of objects from images. Reflectance can be affected by factors such as surface roughness, surface orientation, and the wavelength of the incident light.
\item \textbf{Radiance} refers to the amount of light that is emitted, reflected, or transmitted by a surface in a given direction. It is typically measured in units of power per unit area per unit solid angle (e.g., watts per steradian per square meter). Radiance is important in computer graphics because it determines the color and brightness of objects in a scene as they are rendered from different viewpoints and under different lighting conditions. In machine learning, radiance is also used to describe the behavior of electromagnetic radiation in the context of remote sensing and other applications.

\item \textbf{Lighting}
\begin{itemize}
\item \textbf{Specural Lighting} is a type of lighting effect in computer graphics that simulates the way light reflects from the shiny surface (such as glass or metal). Its primarily used for creating highlights and glints on such surfaces that helps to make them appear more realistic. It is based on the Phong reflection model (discussed later when discussing Bidirectional Reflectance Distribution Function) that describes how light interacts with surface in terms of three components: Ambient, Specular, Diffuse. Specular component is responsible for simulating the light off shiny surfaces. 

The amount of specular lighting at a given point on a surface is determined by the angle between the incoming light ray and the normal vector of the surface at that point, as well as the angle between the outgoing reflection ray and the viewing direction. These angles are defined by the incident vector $\mathcal{L}$, the surface normal or the normal $\mathcal{N}$, and the viewing direction vector $\mathcal{V}$. Hence, the angle between incident light ray and surface normal is given by their dot product as depicted in the eqn~\ref{eqn:21-1}

\begin{equation}
    \label{eqn:21-1}
    \cos{\theta} = \mathcal{L} \cdot \mathcal{N}
\end{equation}

Reflection of the light ray from the surface is given by vector $\mathcal{R}$ that is the result of incident ray reflecting about the surface normal. This has been given in the eqn.~\ref{eqn:21-2}. 

\begin{equation}
    \label{eqn:21-2}
    \mathcal{R} = 2 (\mathcal{L} \cdot \mathcal{N})\mathcal{N} - \mathcal{L}
\end{equation}

and the angle between the reflected ray $\mathcal{R}$ and the viewing direction $\mathcal{V}$ is given by the eqn.~\ref{eqn:21-3}. 

\begin{equation}
    \label{eqn:21-3}
    \cos{\alpha} = \mathcal{R} \cdot \mathcal{V}
\end{equation}

The amount of specular lighting at a given point on the surface is then given by the eqn.~\ref{eqn:21-4}

\begin{equation}
    \label{eqn:21-4}
     \mathcal{I}_{spec}= \textrm{K}_s \times \mathcal{I}_{light} \times \cos^n{\alpha}
\end{equation}

where $\mathcal{I}_light$ is the intensity of the incoming light, $\textrm{K}_s$ is a material property that controls the amount of specular reflection, and $n$ is the shininess or roughness of the surface. The shinier the surface, the higher the value of $n$ and the more concentrated the specular reflection will be.\\
In summary, specular lighting is a key component of the Phong reflection model in computer graphics, and is used to simulate the reflection of light off of shiny, reflective surfaces. The amount of specular lighting is determined by the angle between the incoming light ray and the surface normal, as well as the angle between the reflection ray and the viewing direction, and is controlled by material properties such as $\textrm{K}_s$ and the shininess of the surface.

\item \textbf{Diffuse Lighting} simulates the way light scatters off of matte, non-reflective surfaces, such as wood or paper. It is used to create subtle shading and shadows on such surfaces, which can help to make them appear more realistic.

The amount of diffuse lighting, much like Specular Lighting, at a given point on a surface is determined by the angle between the incoming light ray and the normal vector of the surface at that point. This angle is defined by the incident vector $\mathcal{L}$ and the normal vector $\mathcal{N}$. The angle between the incident light ray and the surface normal is given by the dot product of the above mentioned values as given in the eqn.~\ref{eqn:21-5}:

\begin{equation}
    \label{eqn:21-5}
    \cos{\theta} = \mathcal{L} \cdot \mathcal{N}
\end{equation}

The amount of diffuse lighting at a given point on the surface is then given by the eqn~\ref{eqn:21-6} given below.

\begin{equation}
    \label{eqn:21-6}
     \mathcal{I}_{diff}= \textrm{K}_d \times \mathcal{I}_{light} \times \cos{\theta}
\end{equation}

where $\mathcal{I}_light$ is the intensity of the incoming light, $\textrm{K}_d$ is a material property that controls the amount of diffuse reflection, and $\cos{\theta}$ is the cosine of the angle between the incident light ray and the surface normal. The higher the value of $\cos{\theta}$, the more parallel the incoming light is to the surface, and the brighter the resulting diffuse reflection will be.

So, diffuse lighting is used to simulate the scattering of light off of non-reflective surfaces. The amount of diffuse lighting is determined by the angle between the incoming light ray and the surface normal, and is controlled by material properties such as $\textrm{K}_d$.

\item \textbf{Ambient Lighting} is a type of lighting effect in computer graphics that simulated the way that light is scattered and reflected by the objects in the environment. This is used to simulate the subtle, overall illumination of a scene and can help to create a more realistic and natural-looking image.

In the Phong reflection model, ambient lighting is the simplest component and represents the light that is scattered uniformly in all directions. It is not affected by the direction of the light source or the surface normal, but is instead a constant value that is added to the overall color of the object. The amount of ambient lighting is determined by a material property known as the ambient reflection coefficient, denoted as $\textrm{K}_a$.

The amount of ambient lighting at a given point on a surface is given by the eqn.~\ref{eqn:21-7}:

\begin{equation}
    \label{eqn:21-7}
     \mathcal{I}_{amb}= \textrm{K}_a \mathcal{I}_{light}
\end{equation}

where $\mathcal{I}_{light}$ is the intensity of the ambient light in the scene. The value of $\textrm{K}_a$ determines how much of this ambient light is reflected by the surface.

Hence, ambient lighting is a simple and uniform type of lighting effect that simulates the overall illumination of a scene. It is not affected by the direction of the light source or the surface normal, but is instead a constant value that is added to the overall color of the object. The amount of ambient lighting is controlled by a material property known as the ambient reflection coefficient.

\end{itemize}

\item \textbf{Albedo} is the property of a surface to be able to reflect sunlight from the sun (/heat from the sun). Hence, light colored surface have a high albedo as they reflect quite a lot of sun rays whereas dark surfaces absorb the rays from the sun. In more technical terms, it can be stated as the ratio of a measure of Diffuse Reflection of Solar radiation to the total amount of Solar radiation.\\
In computer graphics research, albedo refers to the proportion of incoming light that is reflected by a surface. It is an important parameter used in material models to determine the color and reflectance properties of objects in a scene. The albedo of a surface is typically represented as a value between 0 and 1, where a value of 0 indicates that the surface absorbs all of the incident light, and a value of 1 indicates that the surface reflects all of the incident light.\\
The albedo of a surface can be calculated using the eqn~\ref{eqn:22} where $\mathcal{A}$ is albedo, $\mathcal{L}$ is the total amount of light that falls on the surface (incident light), and reflected light $\mathcal{R}$ is amount of light that is reflected by the surface.\\
\begin{equation}\label{eqn:22}
    \mathcal{A} = \mathcal{R}/\mathcal{L}
\end{equation}

Albedo is often used along with other material properties (like specular reflectivity, roughness) in Computer Graphics for modeling appearance of surfaces in a scene. This dependence of albedo can be mathematically represented as given in the eqn~\ref{eqn:23}.

\begin{equation}
\label{eqn:23}
    \rho (\theta) = \frac{\rho_0 + \rho_1 \cos{\theta}}{1 + \rho_0 \cos{\theta}}
\end{equation}
where $\rho(\theta)$ represents directional hemisphere reflectance factor representing ratio of reflected radiation to incident radiation at $\theta$ incidence angle, $\rho_0, \rho_1$ are specular and diffuse reflectance coefficient for ratios of reflected to incidence rays for perfectly smooth (reflective) surface and perfectly rough (diffuse) surface. This means the surface albedo depends on both specular and diffuse reflectance factors in turn depending on the roughness, texture, reflectivity. 
 As the angle of incidence $\theta$ increases, the specular reflectance factor $\rho_0$ becomes more important, and the albedo of the surface increases. However, if the surface is surrounded by objects that absorb light, the amount of reflected light will be reduced, which will decrease the albedo of the surface. \\
 In addition to the surrounding environment, the albedo of a surface can also be affected by the presence of shadows or other sources of occlusion. If a surface is partially shaded, the amount of incident light will be reduced, which will decrease the albedo of the surface. Overall, the albedo of a surface is a complex phenomenon that depends on several factors. Albedo is an important concept in computer graphics research because it plays a crucial role in determining the overall appearance of objects in a scene, and it can be used to create realistic lighting and shading effects.

\item \textbf{Lambertian} of a surface is the property that defines an ideal ``matte'' or diffusely reflecting surface. In computer graphics, it is a type of surface that scatters incident light equally in all directions. It is also known as a diffuse surface, and it is an idealized model of many real-world surfaces such as paper, cloth, and matte painted surfaces.\\
The Lambertian model assumes that the amount of light scattered by a surface is proportional to the cosine of the angle between the incoming light direction and the surface normal. This can be expressed using an eqn.~\ref{eqn:24} where $\rho$ is reflectance factor constant value between 0 and 1.
\begin{equation}
    \label{eqn:24}
    \mathcal{LR} = \frac{\rho}{\pi}
\end{equation}
The Lambertian model is characterized by a constant reflectance factor, which is independent of the angle of incidence of the light. This means that a Lambertian surface appears equally bright from all directions, and it does not produce any specular highlights or reflections.\\
In computer graphics, the Lambertian model is often used as a simplifying assumption for simulating the lighting of diffuse surfaces, as it is a simple and efficient model that can produce realistic results in many cases. However, it is important to note that not all surfaces behave like Lambertian surfaces, and more complex models may be necessary for accurately simulating the lighting of some surfaces, such as metals or plastics.

\item \textbf{BRDF (Bidirectional Reflectance Distribution Function)} is a mthematical function to describe how light is reflected by surface in different directions. It is a fundamental concept in Computer Graphics and COmputer Vision. In Virtual Environment, it is used to model the reflection of light from surfaces. It is represented by 4 variables which are incoming light direction ($\omega_i$), outgoing light direction ($\omega_o$), surface normal ($\mathcal{N}$), and light wavelength ($\lambda$). It has been represented by the eqn.~\ref{eqn:25} where $\textrm{dL}(\omega_o), \textrm{dE}(\omega_i)$ are the differential amount of light reflected in the direction of $\omega_o$ and differential amount of incident energy in direction of $\omega_i$.
\begin{equation}
    \label{eqn:25}
    f(\omega_i, \omega_o, \mathcal{N}, \lambda) = \frac{\textrm{dL}(\omega_o)}{\textrm{dE}(\omega_i)},
\end{equation}

This function describes the amount of light reflected from surface in particular direction for certain incident light direction and surface normal. Many models have been proposed for the BRDF with different parameters and assumptions. One such common model is the Lambertian BRDF which integrates the eqns.~\ref{eqn:24}--\ref{eqn:25} to get a model that assumes that the surface is perfectly diffuse and reflects light equally in all directions and given by eqn.~\ref{eqn:26}
\begin{equation}
    \label{eqn:26}
     f(\omega_i, \omega_o, \mathcal{N}, \lambda) = \frac{\rho}{\pi}
\end{equation}
Another model is called Phong BRDF which models specular reflections from shiny surfaces and can be given by the eqn.~\ref{eqn:27}
\begin{equation}
    \label{eqn:27}
    f(\omega_i, \omega_o, \mathcal{N}, \lambda) = \frac{\textrm{K}_d}{\pi} + \textrm{K}_s \frac{n+2}{2\pi}\cos^n{\theta},
\end{equation}
where $n$ is the specular exponent, higher values for which result in sharper specular component, $\theta$ is the angle between the perfect specular reflective direction and outgoing direction with angles clamped to $\pi/2$ to avoid negative value, $\textrm{K}_d, \textrm{K}_s$ are the diffuse reflectivity (fraction of the incoming energy that is reflected diffusely) and specular reflectivity (fraction of the incoming energy that is reflected specularly). This equation can be written in the form of eqn.~\ref{eqn:28}

\begin{equation}
    \label{eqn:28}
    f(\omega_i, \omega_o, \mathcal{N}, \lambda) = \frac{\textrm{K}_d}{\pi} + \frac{\textrm{K}_s}{\pi} (\mathcal{N} \cdot h)^{\alpha},
\end{equation}
where $h$ is halfway vector between $\omega_o$ and $\omega_i$ and $\alpha$ is the specular component to control the specular highlight. Overall, the BRDF is an important concept in computer graphics and computer vision, as it allows us to model how light interacts with surfaces in virtual environments. By understanding the properties of different BRDF models, we can create more realistic and compelling virtual environments, and develop better algorithms for computer vision applications such as object recognition and tracking.
 
\item \textbf{Color Spaces} are mathematical models that define a range of possible colors and provide a mechanism for mapping those colors to numerical values. In computer vision and computer graphics, color spaces are used to represent and process colors in digital images and video.
Here are some of the most commonly used color spaces:

\begin{itemize}
    \item \textbf{RGB (Red-Green-Blue) color space} represents colors as combinations of red, green, and blue values. It is an additive color model, meaning that colors are created by adding together different amounts of red, green, and blue light. RGB color space is widely used in digital displays and image capture devices.
    \item \textbf{CMYK (Cyan-Magenta-Yellow-Black) color space} is used primarily in printing. It represents colors as combinations of cyan, magenta, yellow, and black values. CMYK is a subtractive color model, meaning that colors are created by subtracting different amounts of ink from a white substrate.
    \item \textbf{HSV (Hue-Saturation-Value)} color space represents colors using three parameters. Hue represents the color itself, saturation represents the intensity of the color, and value represents the brightness of the color. HSV color space is often used in image processing applications. This can be represented by the following eqns~\ref{eqn:28-1}--\ref{eqn:28-6}.
    \begin{equation}
        \label{eqn:28-1}
        \textrm{R}' = \textrm{R}/255; \textrm{G}' = \textrm{G}/255; \textrm{B}' = \textrm{B}/255; 
    \end{equation}
    \begin{equation}
        \label{eqn:28-2}
        \textrm{C}_{max} = \max{(\textrm{R}', \textrm{G}', \textrm{B}')}, \textrm{   }
        \textrm{C}_{min} = \min{(\textrm{R}', \textrm{G}', \textrm{B}')}
    \end{equation}
    
    \begin{equation}
        \label{eqn:28-3}
        \Delta =  \textrm{C}_{max} - \textrm{C}_{min}
    \end{equation}
    
    \begin{equation}
        \label{eqn:28-4}
        \textrm{H} = 
        \begin{cases}
         0\degree, \Delta = 0,\\
         60\degree \times \left( \frac{\textrm{G}' - \textrm{B}'}{\Delta} mod 6 \right),    \textrm{C}_{max} = \textrm{R}'\\
         60\degree \times \left( \frac{\textrm{B}' - \textrm{R}'}{\Delta} \textrm{       } + 2 \right),  \textrm{       }  \textrm{C}_{max} = \textrm{G}'\\
         60\degree \times \left( \frac{\textrm{R}' - \textrm{G}'}{\Delta} \textrm{       } + 4 \right),  \textrm{       }  \textrm{C}_{max} = \textrm{B}'\\
        \end{cases}
    \end{equation}
    \begin{equation}
        \label{eqn:28-5}
        \textrm{S} =         \begin{cases}
         0\degree, \textrm{                  }  \textrm{C}_{max} = 0,\\
         \frac{\Delta}{\textrm{C}_{max}},    \textrm{C}_{max} \not = 0 \\
        \end{cases}
    \end{equation}
    
    \begin{equation}
        \label{eqn:28-6}
        \textrm{V} = \textrm{C}_{max}
    \end{equation}
    \item \textbf{LAB color space} is a color space designed to separate color information from brightness. It consists of three components: L, which represents brightness, and a and b, which represent the color in the green-red and blue-yellow axes. LAB color space is commonly used in image processing applications such as color correction and color-based segmentation.
    \item \textbf{YUV color space} separates color information from brightness, similar to LAB. The Y component represents brightness, while the U and V components represent the color in the blue-yellow and green-red axes. YUV color space is commonly used in video processing and compression applications.
    \item \textbf{XYZ color space} is a standard color space used by the International Commission on Illumination (CIE). It is based on the human visual system and represents all colors visible to the human eye. XYZ color space is often used as an intermediate color space in color transformations.

\end{itemize}
\item \textbf{Artifacts} In computer graphics, artifacts refer to any unwanted or unintended visual distortions or anomalies that occur in an image or animation. Artifacts can be caused by various factors, such as limitations of hardware or software, errors in algorithms or data processing, or limitations in the human visual system.

There are many different types of artifacts that can occur in computer graphics, including:
\begin{enumerate}
    \item \textbf{Aliasing}: This occurs when an image or object appears jagged or pixelated, especially along edges or curves. Aliasing is caused by undersampling, where the resolution of the image or object is not high enough to accurately represent the details. In general, it refers to the unwanted effects that arise when high-frequency signals appear in computer graphics. An instance of this can be when downscaling any image from its original size so its quite common in NeRFs as they work with a downsampled image dataset.\\
    Aliasing is a common problem in computer graphics that occurs when the resolution of an image is not high enough to accurately represent a high-frequency signal, such as a diagonal line or a curve. This results in the appearance of jagged edges or stair-step patterns, which can make the image look pixelated or low-quality.\\
    Aliasing is caused by the sampling rate, or the frequency at which the image is sampled. The Nyquist-Shannon sampling theorem states that the sampling rate must be at least twice the highest frequency component of the signal in order to accurately represent it. If the sampling rate is too low, high-frequency components of the signal will be lost or undersampled, leading to aliasing.\\
    In computer graphics, aliasing can occur in several contexts. One common example is when rendering lines or curves using rasterization techniques. In this case, the line or curve is represented as a series of pixels, and if the sampling rate is too low, the line or curve will appear jagged or stair-stepped.\\
    To reduce aliasing in computer graphics, a variety of techniques have been developed, including anti-aliasing. Anti-aliasing works by oversampling the image, and then using filtering techniques to smooth out the jagged edges or stair-step patterns. This can be done by averaging the colors of neighboring pixels, or by using more advanced filtering techniques such as Gaussian filtering.\\
    The amount of aliasing in an image can be quantified using a measure known as the Nyquist frequency, which is defined as half the sampling rate. The Nyquist frequency represents the highest frequency component that can be accurately represented in the image. If a signal has a frequency component that is higher than the Nyquist frequency, it will be undersampled and will result in aliasing.\\
    In summary, aliasing is a common problem in computer graphics that occurs when the resolution of an image is not high enough to accurately represent a high-frequency signal. This results in jagged edges or stair-step patterns, which can make the image look pixelated or low-quality. The Nyquist-Shannon sampling theorem states that the sampling rate must be at least twice the highest frequency component of the signal in order to accurately represent it. Techniques such as anti-aliasing can be used to reduce aliasing in computer graphics.
    \item \textbf{Moiré patterns}: These are interference patterns that occur when two or more repetitive patterns overlap, causing a new pattern to emerge. Moiré patterns can be seen in printed images or screens, and they can also occur in computer graphics when rendering textures or patterns.
    \item \textbf{Shadow acne}: This occurs when shadows appear jagged or pixelated, especially on curved surfaces. Shadow acne is caused by numerical errors in the algorithms used to calculate shadows.
    \item \textbf{Banding}: This occurs when gradients appear as distinct bands of color, rather than a smooth transition. Banding can be caused by limitations in color depth or by compression algorithms that reduce the number of colors used in an image.
    \item \textbf{Blooming}: This occurs when bright areas of an image bleed into darker areas, causing a halo effect. Blooming is caused by limitations in the dynamic range of the display or by overexposure in the original image.
    \item \textbf{Ghosting}: This occurs when fast-moving objects or camera movements leave a trail or afterimage. Ghosting is caused by limitations in the refresh rate of the display or by motion blur.
    \item \textbf{Artifacts in 3D rendering}: These can include glitches or errors in geometry or texture mapping, flickering or popping of objects, or distortions caused by limitations in the rendering engine.
\end{enumerate}
To avoid or minimize artifacts in computer graphics, various techniques and technologies are used, such as anti-aliasing, high dynamic range rendering, advanced shadowing algorithms, and color correction. Additionally, careful calibration and testing can help identify and address potential sources of artifacts.\\
So it can be said that artifacts in computer graphics are any unwanted or unintended visual distortions or anomalies that occur in an image or animation. Artifacts can be caused by various factors, such as limitations of hardware or software, errors in algorithms or data processing, or limitations in the human visual system. There are many different types of artifacts, and various techniques and technologies are used to avoid or minimize them.
\item \textbf{Alpha Compositing} is a technique used in computer graphics to combine two or more images together based on their transparency values. This technique is commonly used in compositing and image editing applications.\\
In alpha compositing, each pixel of an image is represented by four values: RGB$\alpha$. The $\alpha$ value represents the transparency of the pixel, with a value of 0 indicating complete transparency and a value of 1 indicating complete opacity. When two images are combined using alpha compositing, the alpha values of each pixel are used to determine how much of each image should be visible in the final result. It is described by the following eqn~\ref{eqn:31} given below.

\begin{equation}
    \label{eqn:31}
    \mathcal{C} = ((1-\alpha)  \mathcal{C}_1) + (\alpha  \mathcal{C}_2)
\end{equation}

where $\mathcal{C}, \mathcal{C}_1, \mathcal{C}_2$ represent the color of final pixel, pixel in the first image, and pixel in the second image. This equation essentially blends the colors of the two images together based on their alpha values, with the second image contributing more to the final result if its alpha value is higher.\\
One common use of alpha compositing is to add an image with a transparent background to another image. For example, if an image of a person with a transparent background is overlaid onto a background image using alpha compositing, the person will appear to be seamlessly integrated into the background.\\
Another use of alpha compositing is to create effects such as fades, glows, and shadows. By manipulating the alpha values of an image, it is possible to create the illusion of transparency and depth, which can be used to add visual interest and realism to a scene.

\item \textbf{Baking} is the process of performing offline calculations extensively and caching results in a Vertex Attributes/Texture Maps. When coming across this term in future, it might be used in context of generating low-level models, lightmaps, or normal maps.\\
Baking in computer graphics refers to the process of precomputing and storing lighting and other surface properties of 3D models or scenes into textures or other forms of data that can be used to speed up rendering and improve performance. This process is commonly used in real-time applications such as video games, where real-time rendering performance is crucial.\\
Baking is done by simulating and rendering the lighting and other surface properties of a 3D model or scene at a high level of detail, and then storing the results in a texture or other form of data. This precomputed data can then be used during real-time rendering to quickly and efficiently simulate the lighting and surface properties of the 3D model or scene.\\
There are several types of data that can be baked, including lighting, shadows, ambient occlusion, and reflection maps. Each of these types of data can be computed and stored separately, and then combined during real-time rendering to create a more realistic and detailed image.\\
The process of baking can be broken down into several steps. First, the 3D model or scene is prepared for baking by optimizing the geometry, setting up the lighting and material properties, and defining the areas that need to be baked. Next, the baking process itself is performed, which involves rendering the scene from multiple angles and at different levels of detail to capture the necessary data. Finally, the baked data is stored in a texture or other form of data that can be used during real-time rendering.\\
The amount of data that needs to be stored for baking can be quite large, especially for complex scenes or models. To optimize performance, various compression and optimization techniques can be used to reduce the size of the data without compromising quality.\\
In summary, baking in computer graphics is the process of precomputing and storing lighting and other surface properties of 3D models or scenes into textures or other forms of data that can be used to speed up rendering and improve performance. This process involves simulating and rendering the lighting and other surface properties of a 3D model or scene at a high level of detail, and then storing the results in a texture or other form of data. Baking can be used to compute and store data for lighting, shadows, ambient occlusion, and reflection maps, among other things.

\item \textbf{Dynamic Range} refers to the range of brightness or luminance values that can be represented in an image or scene. It is a measure of the difference between the brightest and darkest parts of the image, and is typically expressed in terms of the ratio between the maximum and minimum luminance values. It can be categorized as follows.
\begin{itemize}
\item \textbf{HDR (High-Dynamic Range)} refers to the ability to capture, store, and display a wider range of brightness and color values than traditional displays. HDR technology is becoming increasingly popular in computer graphics and is commonly used in applications such as video games, virtual reality, and video production.\\
Traditional displays typically have a limited range of brightness and color values that they can display, typically around 8 bits per channel (or 24 bits total for RGB color), resulting in a maximum of 256 levels of brightness and color. In contrast, HDR displays can support a much wider range of brightness and color values, with some displays supporting up to 10 or 12 bits per channel, resulting in up to thousands of levels of brightness and color.\\
HDR is achieved through a combination of hardware and software technologies. In order to capture and store HDR images, cameras and sensors with higher dynamic range capabilities are used, along with file formats such as EXR or HDR that can store a wider range of brightness and color values. HDR images can be processed and manipulated using specialized software that can take advantage of the increased dynamic range, allowing for more realistic and visually stunning results.\\
In addition to capturing and storing HDR images, HDR displays are also required to properly display HDR content. This requires the use of specialized hardware and software that can handle the increased dynamic range and color gamut, and can map the wider range of brightness and color values to the limited range of values that can be displayed on the screen.\\
One common technique used in HDR is tone mapping, which is the process of mapping the wide range of brightness and color values in an HDR image to the limited range of values that can be displayed on a traditional display. This can be done using a variety of techniques, including logarithmic or linear mapping, or through more advanced algorithms such as tone mapping operators.\\
In summary, High Dynamic Range (HDR) in computer graphics refers to the ability to capture, store, and display a wider range of brightness and color values than traditional displays. HDR is achieved through the use of specialized hardware and software that can handle the increased dynamic range and color gamut, and can map the wider range of brightness and color values to the limited range of values that can be displayed on the screen. HDR technology is becoming increasingly popular in computer graphics and is used in applications such as video games, virtual reality, and video production.

\item \textbf{LDR (Low-Dynamic Range)} refers to images or video that have a limited range of brightness and color values. LDR images are typically created by capturing or rendering images with a fixed range of brightness and color values, and displaying them on traditional displays with a limited dynamic range.\\
Traditional displays typically have a limited range of brightness and color values that they can display, typically around 8 bits per channel (or 24 bits total for RGB color), resulting in a maximum of 256 levels of brightness and color. LDR images are captured, processed, and displayed within this limited range, resulting in images that may appear less vibrant or less realistic than HDR images.\\
In LDR images, the range of brightness and color values is limited by the range of the display device. For example, if an image is captured with a range of brightness values that exceeds the range of the display device, the brightness values will be compressed or clipped to fit within the display range. This can result in loss of detail in the highlights or shadows of the image.\\
LDR images are still commonly used in many applications, such as traditional video production and broadcast television, where the limited dynamic range of displays is sufficient for the intended use. LDR images can also be created from HDR images through a process called tone mapping, which maps the wider range of brightness and color values in an HDR image to the limited range of values that can be displayed on a traditional display.\\
In summary, Low Dynamic Range (LDR) in computer graphics refers to images or video that have a limited range of brightness and color values. LDR images are typically created by capturing or rendering images with a fixed range of brightness and color values, and displaying them on traditional displays with a limited dynamic range. LDR images are still commonly used in many applications, such as traditional video production and broadcast television. LDR images can also be created from HDR images through a process called tone mapping.
\end{itemize}

\item \textbf{Yaw, Pitch, Roll} refer to the three degrees of freedom that are associated with the movement of objects on their axes. There are six degrees of freedom for a rigid body. An object can move up-down, left-right, forward-backward, pitch, yaw, roll. This is better depicted by the fig.~\ref{fig:workflow-1} given below. It would become clear by looking at the following figure.

\begin{figure*}[!htp]
    \centering
    \includegraphics[width=0.5\textwidth]{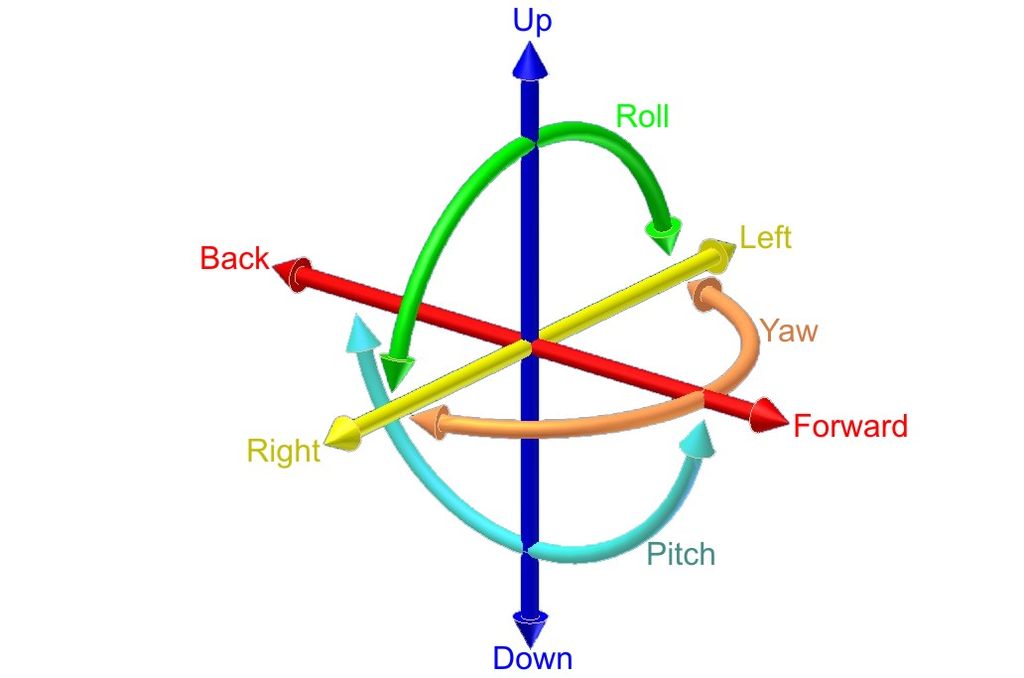}
    \caption{A schematic representation of six degrees of freedom}
    \label{fig:workflow-1}
\end{figure*}

\item \textbf{Gimbal Lock} is a phenomenon in 3D computer graphics and animation that makes the rotations of an object limited or ambiguous, that causes unexpected or incorrect behavior. It occurs when using Euler angles to represent the 3D space orientation of an object.\\
One of the common measures to represent 3D space orientation of an object is the usage of Euler angles. And these Euler angles are used to define three rotations around all three different axes. However, when the object is rotated to certain positions, the rotations can become aligned or parallel, resulting in a loss of a degree of freedom in the rotation. This leads to the condition called Gimbal Lock.\\
When visualizing Gimbal Lock, assume a set of three nested rings, each representing a rotation around a different axis (e.g., $x$, $y$, and $z$). When the rings are aligned, the rotations around the $y$ and $z$ axes become the same, resulting in a loss of 2-degrees of freedom in the rotation.\\
Assume having an object with three Euler angles representing its orientation: yaw/heading (rotation around the y-axis), pitch (rotation around the x-axis), and roll (rotation around the z-axis). Then this object orientation can be represented the as a rotation matrix $\textrb{R}$, which can be calculated as given in the eqn.~\ref{eqn:32} below.
\begin{equation}
    \label{eqn:32}
    \textrb{R} = \textrb{R}_x \times \textrb{R}_y \times \textrb{R}_z,
\end{equation}
where $\textrb{R}_x$, $\textrb{R}_y$, and $\textrb{R}_z$ are given by the following eqns.~\ref{eqn:33}.
\begin{equation}
    \label{eqn:33}
    \textrb{R}_z = \begin{pmatrix}
    \cos{\theta_{r}} & -\sin{\theta_{r}} & 0\\
    \sin{\theta_{r}} & \cos{\theta_{r}} & 0\\
    0 & 0 & 1
    \end{pmatrix},
    \textrb{R}_x = \begin{pmatrix}
    1 & 0 & 0 \\
     0 & \cos{\theta_{p}} & -\sin{\theta_{p}}\\
     0 & \sin{\theta_{p}} & \cos{\theta_{p}}\\
    \end{pmatrix},
    \textrb{R}_y = \begin{pmatrix}
     \cos{\theta_{y}} & 0 &  \sin{\theta_{y}}\\
     0 & 1 & 0 \\
     -\sin{\theta_{y}} & 0 &  \cos{\theta_{y}}
    \end{pmatrix},
\end{equation}

However, when the pitch angle approaches $\pm90\degree$, the rotation around the y-axis (yaw) and the rotation around the z-axis (roll) become aligned, resulting in a loss of a degree of freedom. This is gimbal lock, and it can cause unexpected or incorrect behavior in animations and simulations.

One solution to gimbal lock is to use quaternion rotations instead of Euler angles, as quaternions do not suffer from gimbal lock. Alternatively, other rotation representations, such as axis-angle or matrix representations, can also be used to avoid gimbal lock.

\subsection{Mathematics and Geometry}
\label{mathngeom}
This section deals with the basic terminology in the discipline of Mathematics and Geometry which is useful for the rest of the survey.
\item \textbf{Hadamard Product} is an operation that takes two matrices of the same size and multiplies their corresponding entries together to produce a new matrix of the same size. The Hadamard product is also known as the element-wise product, point-wise product, or Schur product.
The Hadamard product of two matrices A and B of the same size is denoted by the symbol $\odot$ in the eqn~\ref{eqn:34} given below.

\begin{equation}
    \label{eqn:34}
    (\mathcal{A} \odot \mathcal{B})_{ij} = \mathcal{A}_{ij} \odot \mathcal{B}_{ij}
\end{equation} 

where $i$ and $j$ represent the row and column indices of the matrices. Hence, the Hadamard product produces a new matrix where each element is the product of the corresponding elements in the original matrices. For instance, the Hadamard product of the following eqn~\ref{eqn:35} given below.

\begin{equation}
    \label{eqn:35}
    \mathcal{A} = \begin{pmatrix}
    1 & 2\\
    3 & 4
    \end{pmatrix}, \textrm{and   }
    \mathcal{B} = \begin{pmatrix}
    5 & 6\\
    7 & 8
    \end{pmatrix}, \textrm{then   }
    \mathcal{A} \odot \mathcal{B} = \begin{pmatrix}
    5 & 12\\
    21 & 32
    \end{pmatrix}
\end{equation}

This is a useful operation in various applications of linear algebra, such as image processing, computer vision, and machine learning. For example, in image processing, the Hadamard product can be used to perform point-wise operations on images, such as contrast enhancement or color manipulation. In machine learning, the Hadamard product can be used to perform element-wise multiplication of feature vectors, which can be useful for regularization or feature selection.

\item \textbf{Kronecker Product} is a mathematical operation that takes two matrices and produces a new matrix whose entries are all possible products of the entries of the original matrices. It is denoted by the symbol $\otimes$. Now, when given 2 matrices $\mathcal{A}_{(m \times n)}$ and $\mathcal{B}_{(p \times q)}$, the Kronecker product $\mathcal{A} \otimes \mathcal{B}$ is given by the eqn~\ref{eqn:36}
\begin{equation}
    \label{eqn:36}
    (\mathcal{A} \otimes \mathcal{B})_{ij} = \mathcal{A}_{ij} \times \mathcal{B}_{kl}
\end{equation}

where $i, j, k, l$ represent row, column indices of the matrices. So, Kronecker product produces a new matrix where each entry is product of the corresponding entries in original matrices, arranged in a matrix block structure. For instance, the Hadamard product of the following eqn~\ref{eqn:37} given below.

\begin{equation}
    \label{eqn:37}\begin{array}{r}
    \mathcal{A} = \begin{pmatrix}
    1 & 2\\
    3 & 4
    \end{pmatrix}, \textrm{and   }
    \mathcal{B} = \begin{pmatrix}
    5 & 6\\
    7 & 8
    \end{pmatrix}, \\
    \textrm{then   }
    \mathcal{A} \otimes \mathcal{B} = \begin{pmatrix}
    1 \times \mathcal{B} & 2 \times \mathcal{B}\\
    3 \times \mathcal{B} & 4 \times \mathcal{B}
    \end{pmatrix} \Rightarrow  \begin{pmatrix}
    5 & 6 & 10 & 12\\
    7 & 8 & 14 & 16\\
    15 & 18 & 20 & 24\\
    21 & 24 & 28 & 32
    \end{pmatrix}
    \end{array}
\end{equation}

The Kronecker product is a useful operation in various applications of linear algebra, such as signal processing, computer vision, and quantum mechanics. For example, in signal processing, the Kronecker product can be used to model the behavior of complex systems by combining simpler models. In computer vision, the Kronecker product can be used to construct feature descriptors for image analysis. In quantum mechanics, the Kronecker product is used to describe the composite state of multiple quantum systems.

\item \textbf{Signed Distance Function} describes the distance between a point in space and a geometrical object. It is often used in computer graphics, particularly in the field of 3D rendering, to create realistic images of 3D objects. This function returns a signed value, which indicates whether the point is inside or outside the object. If the point is inside the object, the value returned by the SDF is negative; if the point is outside the object, the value is positive. The general equation for an SDF function is given in eqn.~\ref{eqn:38}

\begin{equation}
\label{eqn:38}
f(p) = d(p, O)
\end{equation}
where $p$ is point in space, $O$ is a geometrical object, and $d$ represents the euclidean distance between $p$ and $O$. The different types of SDFs are given below.

\begin{itemize}
\item \textbf{Sphere Signed Distance Function} is used to describe a sphere and its equation is as given by eqn~\ref{eqn:39}
\begin{equation}
    \label{eqn:39}
    f(p) = \textrb{length} (p - o_{c}) - r
\end{equation}
where $p$ is point in space, $o_{c}$ is sphere center, $r$ is sphere radius, $\textrb{length}$ represents the Euclidean distance.
\item \textbf{Box Signed Distance Function} is used to describe a box and its equation is as given by eqn~\ref{eqn:40}

\begin{equation}
\label{eqn:40}
    \begin{multlined}
    f(p) = \max(|(p_x - center_x)| - size_x, 0.0)\\
    \begin{aligned}
    &+\max(|(p_y - center_y)| - size_y, 0.0)\\
    &+\max(|(p_z - center_z)| - size_z, 0.0)
    \end{aligned}
\end{multlined}
\end{equation}
where $p$ is point in space, $o_{c}$ is box center, $\textrb{size}$ is box size.
\item \textbf{Cylinder Signed Distance Function} is used to describe a cylinder and its equation is as given by eqn~\ref{eqn:41}
\begin{equation}
    \label{eqn:41}
    f(p) = \textrb{length} (p.xz - o_{c}.xz) - r
\end{equation}
where $p$ is point in space, $o_{c}$ is cylinder center, $r$ is cylinder radius, $\textrb{length}$ represents the Euclidean distance.
\item \textbf{Truncated Signed Distance Function} is a mathematical function that is used to represent a 3D shape in a volumetric way. It is commonly used in 3D computer vision and robotics applications, such as 3D reconstruction, SLAM, and object recognition. The TSDF function is an extension of the Signed Distance Function (SDF) that allows the representation of both the exterior and interior of an object. It does this by truncating the distance function at a certain threshold value, which creates a smooth transition between the interior and exterior regions of the object. TSDF can be represented by the eqn~\ref{eqn:42}

\begin{equation}
    \label{eqn:42}
    \mathcal{TSDF} (p) = \min{\left( \max{\left( \frac{f(p)} {\textrb{dis}_{trunc}}, -1 \right)}, 1 \right)}
\end{equation}
where $p$ is 3D point in space, $f(p)$ is SDF at $p$ and $\textrb{dis}_{trunc}$ is truncated threshold to determine distance beyond where function is truncated. Output of TSDF is scalar that indicates distance between $p$ and nearest surface of the object. The output sign indicates whether the point is inside or outside of object. There are differeent types of TSDF used in Computer Vision as given below.

\begin{itemize}
    \item \textbf{Volumetric TSDF} is used to represent a 3D object as a voxel grid. The distance values are stored in the voxels of the grid, and the grid is updated as new data is acquired.
    \item \textbf{Point-based TSDF} is used to represent a 3D object as a set of 3D points. The distance values are computed for each point, and the points are updated as new data is acquired.
    \item \textbf{Mesh-based TSDF} is used to represent a 3D object as a mesh. The distance values are computed for each vertex of the mesh, and the mesh is updated as new data is acquired.
\end{itemize}
\item \textbf{NSDF (Neural Signed Distance Function)} is a type of neural network architecture that is used to learn a continuous representation of a 3D object from a set of 3D points. It is commonly used in 3D shape reconstruction, object recognition, and robotics applications. The NSDF function takes as input a 3D point and outputs a signed distance value, which indicates the distance between the point and the nearest surface of the object. The sign of the distance value indicates whether the point is inside or outside the object. NSDF can be represented by eqn~\ref{eqn:43}--\ref{eqn:44} given below.

\begin{equation}
    \label{eqn:43}
    f(p) = \textrb{MLP}(p), \textrm{so   }
\end{equation}

\begin{equation}
    \label{eqn:44}
    \mathcal{NSDF}(p) = \frac{f(p)}{||f(p)||_2}
\end{equation}

where $p$ is 3D point, $\textrb{MLP}(p)$ is MLP output applied to $p$, and $||f(p)||_2$ is the L2 norm of the MLP output. The NSDF output is a unit vector that points towards the nearest surface of the object. There are different types of NSDF used in computer vision and robotics applications. Some of them are:

\begin{itemize}
    \item \textbf{PointNet++-based NSDF} is based on the PointNet++ architecture, which is a neural network architecture that is designed to work directly on point clouds. The NSDF is learned by training the PointNet++ network to predict the signed distance value for each point in the input point cloud.
    \item \textbf{Implicit Function-based NSDF} is based on an implicit function that defines the 3D object surface as the zero level set of the function. The NSDF is learned by training a neural network to approximate the implicit function using a set of 3D points.
    \item \textbf{Atlas-based NSDF} is based on an atlas of 3D shapes, where each shape is represented as an NSDF. The NSDF is learned by training a neural network to predict the correct NSDF for a given shape in the atlas.
    \item \textbf{Occupancy-based NSDF} is based on an occupancy function that defines the occupancy of a 3D space. The NSDF is learned by training a neural network to predict the occupancy value for each point in the input point cloud.
\end{itemize}
\item \textbf{Dual Contouring SDF} is a method for generating a 3D mesh from the SDF representation of an object. It uses a technique called dual contouring to extract the surface mesh from the SDF.
\item \textbf{Distance Field SDF} is a method for generating a voxel grid that represents the SDF of an object. It is useful for collision detection and path planning in robotics applications.
\item \textbf{Occupancy SDF} is similar to the Distance Field SDF, but instead of representing the SDF as a voxel grid, it represents it as an occupancy grid, where each grid cell is either occupied or unoccupied. It is useful for object recognition and scene understanding in computer vision applications.
\item \textbf{Fast Marching SDF} is a method for computing the SDF of an object using the Fast Marching Method, which is an algorithm for solving partial differential equations. It is useful for real-time applications that require fast computation of the SDF.
\item \textbf{Level Set SDF} is a method for representing the SDF as the zero level set of a higher-dimensional function. It is useful for modeling the motion of deformable objects and for tracking the evolution of interfaces in fluids and other physical systems.
\end{itemize}
\item \textbf{Plenoptic Function} is a mathematical representation of complete information of light rays travelling in the 3D space. It describes how light is emitted from the scene, travels through space, interacts with objects, and is finally captured by a sensor or observer. The plenoptic function is a powerful tool for modeling and simulating various optical effects, such as depth of field, motion blur, and light field imaging. It is a seven-dimensional function that takes into account the position of the light source, its wavelength, the direction of the light rays, and time. This function can be depicted by the eqn~\ref{eqn:45} given below.
\begin{equation}
    \label{eqn:45}
    \mathcal{PF}(x,y,z,\theta,\phi,t, \lambda) = \mathcal{I}
\end{equation}
where $x,y,z$ is 3D space position of the point, $\theta, \phi$ represent the direction of light rays (or sometimes the viewing direction, since light direction is dependant on the viewer direction), $t$ represents time, and $\mathcal{I}$ is the radiance of light intensity at the queried position, directions, and times. In computer graphics, the plenoptic function is often used to simulate the behavior of light in virtual scenes. One extension of the plenoptic function is the bidirectional reflectance distribution function (BRDF) (discussed). There are several extensions of the Plenoptic Function (not related to Computer Graphic), which include:

\begin{itemize}
\item \textbf{Plenoptic Sampling Function} represents the plenoptic function in a discrete form, by sampling the function at discrete positions and directions. The Plenoptic Sampling Function is used in the design of plenoptic cameras.
\item \textbf{Compressed Plenoptic Function} compresses the Plenoptic Function by exploiting the sparsity of the function. It is useful for efficient storage and transmission of plenoptic data.
\item \textbf{Multi-View Plenoptic Function} represents the plenoptic function from multiple viewpoints. It takes into account the position and orientation of multiple cameras or sensors and combines the plenoptic data from each viewpoint to create a single multi-view plenoptic function.
\item \textbf{Time-Varying Plenoptic Function} represent the plenoptic function over time. It takes into account the time-varying nature of light and captures the changes in the plenoptic function over time.
\item \textbf{Polarization Plenoptic Function} represents the polarization state of light in addition to the position, direction, and time. It captures the polarization properties of light, which are important for a range of applications including remote sensing and biomedical imaging.
\item \textbf{SPF (Surface Plenoptic Function)} is the 5-D functional representation of the intensity of light reflected from the surface at every point on the surface, in all directions, and at all times. It is a generalization of the Plenoptic Function that takes into account the surface geometry and the orientation of the surface normal. It is often depicted by the eqn~\ref{eqn:46} given below.
\begin{equation}
    \label{eqn:46}
    \mathcal{SPF} (x,y,z,\theta,\phi) = \mathcal{L}
\end{equation}
The SPF is useful for a range of applications in computer vision, graphics, and robotics, including surface reconstruction, object recognition, and image-based rendering. One of the main advantages of the SPF is that it provides a compact representation of the surface appearance that can be used to synthesize novel views of the surface from any viewpoint and under any lighting conditions. To estimate the SPF from images, several methods have been proposed in the past, including Shape-from-Polarization, Shape-from-Shading, and Shape-from-Gradient. These methods use different cues to estimate the surface geometry and the surface normal, which are then used to estimate the SPF.\\
The SPF is a powerful representation of surface appearance that can capture complex lighting effects and surface properties such as reflectance and transparency. However, it requires accurate estimation of the surface geometry and orientation, which can be challenging in some cases, particularly for complex or textured surfaces.

\item \textbf{BTF (Bidirectional Texture Function)} describes the variation of a surface's appearance with respect to viewing and illumination angles. BTF is a mathematical model that describes the appearance of a surface as it changes with respect to different viewing and illumination directions. It is a higher-dimensional extension of the BRDF, which describes the reflection properties of a surface. The BTF is typically defined by the eqn.~\ref{eqn:47}
\begin{equation}
\label{eqn:47}
\mathcal{BTF} (x, y,\omega_i, \omega_o, \omega_v)
\end{equation}

where $x$ and $y$ are the spatial coordinates of the surface, $\omega_i$ and $\omega_o$ are the incoming and outgoing light directions, and $\omega_v$ is the viewing direction. The BTF describes how the surface appearance changes as the illumination and viewing directions change.
The BTF can be measured using specialized imaging techniques that capture the surface appearance under a variety of viewing and illumination conditions. These measurements can be used to construct a BTF model that can be used for rendering and simulation of the surface appearance under arbitrary lighting and viewing conditions.\\
The BTF has many applications in computer graphics, including virtual reality, video games, and visual effects. It allows for realistic rendering of surface appearance under complex lighting conditions, such as indirect illumination and multiple light sources. The BTF is often used in conjunction with other reflectance models, such as the BRDF and the BSDF (bidirectional scattering distribution function), to create more accurate and realistic renderings of surfaces.
\end{itemize}
These extensions of the Plenoptic Function are important for a range of applications in computer vision, robotics, remote sensing, and other fields that involve capturing, analyzing, and synthesizing light in the 3D space.

\item \textbf{Spherical Gaussians} are a type of RBF (Radial Basis Function) that are commonly used in computer graphics and computer vision applications to model and represent 3D data. They are particularly useful for tasks such as surface reconstruction, shape analysis, and point cloud processing. The basic form of the spherical Gaussian function is given by the eqn~\ref{eqn:48}.
\begin{equation}
    \label{eqn:48}
    \mathcal{G}(x) = e^{-\frac{||x||^2}{\sigma^2}}
\end{equation}
where $x$ is 3D vector for point in space, $||x||$ is its Euclidean distance, and $\sigma$ scales and controls the size of Gaussiann function.

Spherical Gaaussians are often combined to give a weighted sum of multiple Gaussians, resulting in a more expressive and flexible function as described in eqn~\ref{eqn:49}
\begin{equation}
    \label{eqn:49}
    f(x) = \sum_{i=1}^{N} w_i \times \mathcal{G}(x - c_i)
\end{equation}

where $w_i, c_i$ are the weight factor and center position, and its taken over a total of $N$ centers.

The parameters of the spherical Gaussians, including the center positions, weight factors, and scaling factor, can be learned from data using various techniques such as least squares or maximum likelihood estimation. The resulting function can then be used to interpolate or extrapolate 3D data, such as point clouds or surface meshes.\\
Spherical Gaussians have several desirable properties, such as smoothness, isotropy, and rotational symmetry. These properties make them well-suited for modeling and analyzing 3D data, particularly when the data is noisy or incomplete. Spherical Gaussians are also efficient to compute, making them practical for real-time applications.

\item \textbf{Eikonal Equation} is a partial differential equation that describes the propagation of waves. It is used in fields such as computer vision and graphics, to model various phenomena, including surface deformation and motion of fluids. It can have several forms but one of the general form of the eikonal equation is given by the eqn.~\ref{eqn:50}

\begin{equation}
\label{eqn:50}
|\nabla u(x)| = f(x)
\end{equation}
In computer graphics and computer vision, the Eikonal equation is used to compute the distance function to a surface or object. Given an input point cloud or surface, the Eikonal equation is solved to compute the distance function to the surface at every point in space. This distance function can be used to create level sets, which are surfaces that represent a fixed distance away from the input surface.
Applications of the Eikonal Equation in computer graphics and computer vision include Ray Tracing, Shape from shading, Image segmentation, Medical imaging, level set computation, distance map computation, texture synthesis, path planning, collision avoidance. 

\item \textbf{Fourier Series} is a periodic representation of a function $f(x)$. It is defined as a sum of sine and cosine functions of different frequencies and is mathematically given by the following eqn~\ref{eqn:51} with a period of $2\pi$, and $a_0, a_n, b_n$ being coefficients depending on function f(x).

\begin{equation}
    \label{eqn:51}
    f(x) = \frac{a_0}{2} + \sum_{n=1}^{\infty} \left[ a_n cos(nx) + b_n sin(nx) \right]
\end{equation}
Now, these coefficients can be given based on the following formulae given in eqn~\ref{eqn:52}

\begin{equation}
    \label{eqn:52}
    \begin{array}{l}
    a_0 = \frac{1}{\pi} \int_{-\pi}^{\pi} f(x) \,dx \\ 
    a_n = \frac{1}{\pi} \int_{-\pi}^{\pi} f(x) cos(nx) \,dx\\
    b_b = \frac{1}{\pi} \int_{-\pi}^{\pi} f(x) sin(nx) \,dx\\
    \end{array}
\end{equation}

\item \textbf{Fourier Transform} helps to transform  function in the time domain to function in the frequency domain. It and it's inverse are given by the eqn~\ref{eqn:52} where $k$ is frequency variable, $i$ is the imaginary unit, $F(x)$ is the Fourier transform, and $f(x)$ is the inverse fourier transform.
\begin{equation}
    \label{eqn:53}
    \begin{array}{l}
    F(x) = \int_{-\pi}^{\pi} f(x) \exp^{-2 \pi \iota kx} \,dx\\
    
    f(x) = \int_{-\pi}^{\pi} F(k) \exp^{2 \pi \iota kx} \,dk
    \end{array}
\end{equation}
The Fourier transform allows us to analyze the frequency content of a function and is widely used in signal processing, image processing, and many other fields of engineering and science.

\item \textbf{Discrete Fourier Transform} is a mathematical technique that is a discrete variant of the Fourier Transform discussed above. It transforms a discrete-time sequence of N complex numbers $x[n]$ into a sequence of N complex numbers $X[k]$ representing the frequency content of the signal. It is defined by the eqn~\ref{eqn:54} where $k$ is frequency index ranging from $0$ to $N-1$.

\begin{equation}
\label{eqn:54}
X[k] = \sum_{n=0}^{N-1} \left( x[n] \exp^{\frac{-2 \pi \iota k n}{N}} \right)
\end{equation}

It can be calculated using matrix multiplication of the $N \times N$ DFT matrix $F$ and the $N-1$ column vectors. 

\item \textbf{FFT (Fast-Fourier Transform)} is used for computing the DFT of a sequence of N complex numbers in $O(n\log{}n)$ rather than $O(n^2)$ required by the naive DFT. The FFT algorithm exploits the symmetry and periodicity properties of the DFT to compute the transform more efficiently.

The FFT algorithm divides the sequence $x[n]$ into smaller sub-sequences, computes the DFT of each sub-sequence recursively, and combines the results to obtain the final DFT. The algorithm can be implemented using either a radix-2 or a radix-4 butterfly structure, which exploits the symmetry and periodicity properties of the DFT to reduce the number of multiplications and additions required. In computer vision, the FFT is widely used in image processing for operations such as image filtering, convolution, and correlation. These operations involve the computation of the Fourier transform of the image and the filter kernel, multiplication in the frequency domain, and the inverse Fourier transform to obtain the filtered image. The FFT algorithm allows these operations to be performed efficiently on large images in real-time applications such as video processing and computer vision systems.

\item \textbf{Positional Encodings} is used in Natural Language Processing and sequence modeling to inject the position information of each token in a sequence into the input representation. The idea is to provide the model with information about the position of each token, which is crucial for tasks such as language understanding and translation.

In Transformer-based models, the positional encoding is added to the input embeddings before being fed to the model. The positional encoding is calculated based on the position of each token in the sequence and a set of learned parameters, which encode the position information into a continuous vector space.

The equation for positional encoding is given in eqn~\ref{eqn:55} where $\text{PE}{(pos, 2i)}$ and $\text{PE}{(pos, 2i+1)}$ are the $2i$-th and $(2i+1)$-th elements of the positional encoding vector for the $pos$-th position in the sequence, $d_{\text{model}}$ is the dimensionality of the model, and $i$ is the index of the embedding dimension.

\begin{equation}
    \label{eqn:55}
    \begin{aligned}
    PE_{(pos,2i)} = \sin\left(\frac{pos}{10000^{2i/d_{model}}}\right)\\
    PE_{(pos,2i+1)} = \cos\left(\frac{pos}{10000^{2i/d_{model}}}\right)
    \end{aligned}
\end{equation}

The application of positional encoding is not limited to natural language processing. It can also be used in computer vision and other domains where sequential information is important. For example, in image processing, positional encoding can be used to encode the position information of each pixel in an image, which can be useful for tasks such as object detection, segmentation, and classification. In machine learning, positional encoding can be used to improve the performance of models that work with sequences, such as time series forecasting and speech recognition.

In NeRF, the input to the neural network is a sequence of 3D points along each camera ray, which are used to sample the radiance and volume density at each point. To encode the positional information of each point, NeRF uses a positional encoding scheme similar to the one used in Transformer-based models.

The positional encoding in NeRF is calculated by the eqn.~\ref{eqn:56} given below where  $\text{PE}_i$ is the $i$-th element of the positional encoding vector, $d$ is the dimensionality of the positional encoding, and $x$, $y$, and $z$ are the 3D coordinates of the point. The $\sin$ and $\cos$ terms allow the encoding to capture both periodic and non-periodic features in the spatial structure of the scene.

\begin{equation}
\label{eqn:56}
\begin{aligned}
\text{PE}_\text{ray}(i, 2k) &= \sin\left(\frac{i}{10000^{2k/D}}\right)\\
\text{PE}_\text{ray}(i, 2k+1) &= \cos\left(\frac{i}{10000^{2k/D}}\right)
\end{aligned}
\end{equation}

The use of positional encoding in NeRF allows the model to learn a continuous function that can map any point in the 3D space to its corresponding color and opacity. This enables the generation of photo-realistic 3D scenes from a set of 2D images, which can have various applications in fields such as computer graphics, virtual reality, and augmented reality.

\item \textbf{MHRE (Multi-Resolution Hash Encoding)} is used in Computer Science to represent images using a compact binary code. It is based on the idea of using multiple image resolutions to get local and global frequency information. Here, the image is divided into multiple sub-regions and computing a hash code for each sub-region at different resolutions. Hash code for each subregion is obtained by comparing the pixel values in the sub-region to a predefined threshold. If a pixel value is greater than the threshold, the corresponding bit in the hash code is set to 1, otherwise it is set to 0. The hash codes for all sub-regions at each resolution level are concatenated to form the final binary code for the image.

The idea behind MRHE encoding of an image $I$, $i^{\textrm{th}}$ sub-region hash code $h_i$, resolution level $r_i$ (with total $m$ resolutions) is given by the eqn.~\ref{eqn:57}

\begin{equation}
    \label{eqn:57}
    \mathcal{MHRE}(I) = [h_1^{r_1}, h_2^{r_1}, ..., h_n^{r_1}, h_1^{r_2}, h_2^{r_2}, ..., h_n^{r_2}, ..., h_n^{r_m}]
\end{equation}

The MRHE technique is particularly useful in applications such as image retrieval, object recognition, and image matching, where large-scale image databases need to be searched efficiently. The compact binary code generated by MRHE can be used to quickly compare images and retrieve similar images from a database. Additionally, the multiple resolutions used in MRHE capture both local and global information, making it robust to variations in lighting, texture, and scale.

\item \textbf{Spectral Bias (/F-principle)} is a phenomenon that can occur in computer vision tasks when the data used to train a model has a specific spectral structure that the model can exploit, leading to better performance than expected based on its ability to capture the underlying statistical structure of the data.\\
In computer vision, the Fourier Transform is commonly used to analyze the frequency domain of an image. The F-Principle states that the performance of a machine learning model can be improved if the Fourier Transform of the training data has a specific spectral structure, such as a concentrated or sparse distribution of frequencies. This spectral structure can be exploited by the model to learn more efficiently or accurately than if the distribution of frequencies were more uniformly distributed.\\
For example, consider a dataset of images of faces. If the images in the dataset have a similar lighting condition, such as all being taken in bright, natural light, then the spectral structure of the dataset will be biased towards certain frequencies related to that lighting condition. If a model is trained on this dataset, it may learn to exploit this spectral bias and perform well on images with similar lighting conditions, but may perform poorly on images with different lighting conditions.\\
To mitigate the effects of spectral bias, it is important to use diverse and representative datasets during training, and to evaluate the performance of the model on out-of-distribution data to ensure it is generalizing well to new situations. Additionally, techniques such as data augmentation, which introduces variations in the training data, can help to reduce the impact of spectral bias and improve the generalization performance of the model.

\item \textbf{NTKs (Neural Tangent Kernels)} are a mathematical tool used to analyze the training dynamics of neural networks. It allows us to understand how the weights of a neural network change during training and how this affects the network's ability to generalize to new data.

In the context of computer vision, NTK has been used to analyze the performance of neural networks on image classification tasks. Additionally, NTK has been applied to novel applications such as Neural Radiance Fields (NeRF) which is a method for synthesizing photo-realistic 3D scenes from 2D images.

The eqn~\ref{eqn:58} represents an NTK where $K_{NTK}(x, x')$ is the NTK between input vectors $x$ and $x'$, $f(x)$ is the output of the neural network for input $x$, and $w_i$ represents the weights of the neural network.
\begin{equation}
\label{eqn:58}
\textrm{K}_{NTK}(x, x') = \lim_{n \rightarrow \infty} \frac{1}{n} \sum_{i=1}^n \frac{\partial f(x)}{\partial w_i} \frac{\partial f(x')}{\partial w_i}
\end{equation}
The NTK is computed as the limit of the average product of the partial derivatives of the network output with respect to the weights over an infinite number of weights.

In the context of NeRF, NTK is used to analyze the performance of a neural network that maps 2D images to 3D scenes. By analyzing the NTK, researchers can gain insight into the relationship between the 2D images and the resulting 3D scene and design more effective neural network architectures. Specifically, NTK allows for the efficient computation of the Jacobian matrix, which is used to estimate the gradient of the loss function during training.

\item \textbf{Tracing and Casting} are two of the important techniques used in Computer Vision and Computer Graphics. In Vision, tracing refers to the process of tracking the movement of objects/points in sequence of images/frames. This can be done by using various algorithms and techniques, such as optical flow, feature tracking, and point matching. The goal of tracing in computer vision is to understand the motion and dynamics of objects in a scene and extract useful information, such as the velocity, acceleration, and trajectory of the objects. Whereas in Graphics, tracing refers to the process of generating photorealistic images by tracing the paths of light rays as they interact with virtual objects in a scene. This is typically done using ray tracing algorithms, which simulate the behavior of light as it interacts with objects in the scene. The goal of tracing in computer graphics is to generate realistic images that accurately simulate the lighting and materials of the virtual objects in the scene.\\
Casting, on the other hand, refers to the process of projecting an object or point onto a surface or plane in both computer vision and computer graphics. In computer vision, casting is used to estimate the 3D position of an object or point in the world from a 2D image or frame. This is typically done using methods such as perspective projection or homography, which involve modeling the relationship between the 3D world and the 2D image. The goal of casting in computer vision is to estimate the position and orientation of objects in the world from a 2D image or frame.\\
In computer graphics, casting is used to simulate the behavior of light as it interacts with objects in a scene. This is typically done using methods such as shadow casting and light mapping, which involve modeling the way that light interacts with objects in the scene to create realistic shadows and lighting effects. The goal of casting in computer graphics is to generate realistic images that accurately simulate the behavior of light in the scene.
\begin{itemize}
\item \textbf{Path Tracing} is a widely used algorithm in computer graphics for rendering realistic images of 3D scenes. It is a Monte Carlo method that simulates the behavior of light as it interacts with objects in the scene. The algorithm works by tracing a path of light rays from the camera through the scene, randomly sampling the possible paths of light rays as they reflect, refract, and scatter off of objects in the scene. Path tracing has been used in a wide range of applications, from film and animation to video game development. One of the seminal papers on path tracing is work done by Kajiya et al~\cite{kajiya1986rendering}, which introduced the concept of a physically-based model for rendering 3D scenes. Another influential work by Matt Pharr and Greg Humphreys~\cite{pharr2016physically} provides a comprehensive introduction to the theory and practice of physically-based rendering, including path tracing.

\item \textbf{Ray Marching} is used to sample multiple points from the 3D space rather than just one point where the ray intersects with the 3D surface. This method is used when analytical methods fail to provide reliable results. Its used in computer graphics to render 3D scenes using volumetric databy tracing a ray through a volume and sampling the density and color of the volume along the ray's path. Ray marching is commonly used to generate realistic images of clouds, smoke, and other volumetric phenomena. Seminal works on ray marching~\cite{levoy1988volume,drebin1988volume} introduced the concept of volume rendering and the ray marching algorithm. Another work by William E. Lorensen and Harvey E. Cline~\cite{parker1998interactive} proposed a variant of ray marching for generating isosurfaces of volumetric data.

\item \textbf{Ray Tracing} is a technique used for 3D offline rendering. It is achieved by recursively tracing the path of rays through a scene. It tracing is a technique used in graphics and vision to generate realistic images by tracing the path of light as it interacts with objects in a scene. It works by tracing rays of light from the camera through each pixel and calculating how the light interacts with objects in the scene, including reflection, refraction, and shadows. The first work to introduce ray tracing concept was by Turner Whitted~\cite{wallace1989ray} as it presented a recursive algorithm for computing reflections and refractions.

\item \textbf{Distributed Ray Tracing} is used in computer graphics and computer vision to generate realistic images by simulating the propagation of light in a scene using a distributed network of processors. It involves dividing the rendering process into smaller tasks, distributing them across multiple processors, and then combining the results to produce a final image. It is achieved by some modifications to the process of Ray Tracing where multiple rays are traced through each pixel to model soft phenomenon such as soft shadows, depth of fields. One of the first works to introduce this concept was presented by Cook et al~\cite{cook1984distributed} and it initiated the concept of parallelizing the ray tracing algorithm across multiple processors. 

\item \textbf{Beam Tracing} uses pyramid-shaped beams to address some of the shortcomings of the ray-tracing. These shortcomings include Aliasing effects and the creation of artifacts while rendering.

\item \textbf{Ray Casting} is generally a 2.5 D rendering method and is achieved by casting non-recursive rays from the camera into the scene.
\item \textbf{Sphere Tracing}, also known as raymarching or distance field raymarching, is an algorithm for rendering 3D objects using signed distance functions (SDFs). An SDF is a function that, for any point in space, returns the shortest distance to the surface of the object.

\item \textbf{Marching Cube} is a method for implicit surface triangulation. So, it is an algorithm for generating polygonal meshes from 3D scalar fields, such as voxel data or isosurfaces. The algorithm works by dividing the volume into small cubes and determining the surface intersection points within each cube.

The following steps constitute the Marching Cube algorithm.
\begin{enumerate}
    \item Divide the volume into small cubes.
    \item For each cube, evaluate the scalar field at its 8 vertices.
    \item Determine the surface intersections by comparing the scalar values at the vertices to a predefined isovalue.
    \item Create triangles to connect the intersection points within each cube.
    \item Combine the triangles to form the final polygonal mesh.
\end{enumerate}

The mathematical formulation for each  step is as follows.
\begin{enumerate}
\item Let V(x, y, z) be a scalar field representing the 3D volume.
\item For each cube vertex, evaluate the scalar field: v = V(x, y, z)
\item Create a bitmask by comparing v to the isovalue.
\item Use the bitmask to look up the corresponding edge intersections.
\item Calculate the intersection points: I = (p1 + p2) / 2, where p1 and p2 are the endpoints of the intersecting edge.
\item Generate triangles connecting the intersection points within the cube.
\end{enumerate}

Applications for Marching Cube include
\begin{itemize}
    \item Medical imaging and visualization (CT, MRI, etc.)
    \item Geological data visualization
    \item Fluid dynamics simulations
    \item Terrain generation in video games and virtual environments
\end{itemize}

\end{itemize}
\item \textbf{SLAM} It is a technique often used at the intersection of Robotics and Computer Vision. It further helps with several simple or complex Robotic operations. It generally requires the leverage of C++ to work with robots for manipulation, localization, path finding, etc.
\item \textbf{RGB-D Registration} is the use of Depth along with RGB features to register for different application such as Nuclear Magnetic Resonance Imaging (NMRI), CT scans, Geospatial imaging (when non-azimuthal orbit), etc. 
\item \textbf{DeepV2D (Deep Video to Depth)} is a deep learning-based method for estimating depth maps from a video sequence, leveraging differentiable Structure-from-Motion (SfM) to improve the overall performance. The key idea of DeepV2D is to integrate the geometric principles of SfM with a deep learning framework to jointly optimize for depth, camera motion, and 3D structure. This approach allows the model to learn depth estimation while enforcing geometric consistency across multiple frames in a video sequence.

The main components of the DeepV2D framework are:

\begin{enumerate}
\item CNN-based Depth Estimation: A convolutional neural network (CNN) is used to estimate the initial depth maps for each frame in the video sequence. This network is trained to predict depth maps from single images, but its predictions may not be geometrically consistent across frames.
\item Differentiable SfM: The differentiable SfM module enforces geometric consistency across frames by optimizing the camera motion (poses) and refining the depth estimates. The module consists of differentiable components for camera pose estimation, warping, and a photometric loss function. The differentiability of the SfM module allows the gradients to flow back through the entire pipeline during training, improving both depth estimation and camera motion estimation.
\item Joint Optimization: DeepV2D jointly optimizes depth maps, camera poses, and 3D structure by minimizing a combination of photometric and geometric loss functions. The photometric loss measures the difference between the input images and the images reconstructed by warping neighboring frames using the estimated depth and camera poses. The geometric loss enforces smoothness in the depth maps and penalizes large deviations from the initial CNN-based depth predictions.
\end{enumerate}
DeepV2D is trained in an end-to-end manner using backpropagation, allowing the model to learn depth estimation and camera motion jointly. The integration of differentiable SfM in the deep learning framework helps the model to enforce geometric consistency across video frames and improve the overall performance of depth estimation.
\item \textbf{COLMAP} is a library written in C++ for two of the vital operations in Computer Vision and Computer Graphics namely, MVS and SfM which have been discussed below.
\begin{itemize}
\item \textbf{MVS (Multi-View Stereo)} is a computer vision technique used to reconstruct 3D geometry from multiple 2D images taken from different viewpoints. It builds on the principles of stereo vision, which involves estimating the depth of a scene by comparing the slight differences in appearance between two or more images, known as parallax.

The main steps involved in Multi-View Stereo are:
\begin{enumerate}
    \item \textbf{Image Acquisition:} Capture a set of images of a scene or object from multiple viewpoints, ensuring sufficient overlap and coverage. Ideally, the camera's intrinsic (focal length, principal point, etc.) and extrinsic (position and orientation) parameters should be known or estimated.
    \item \textbf{Feature Extraction and Matching:} For each image, identify distinctive points or features and match them across the different images. This can be achieved using feature detectors and descriptors, such as SIFT, SURF, ORB, etc.
    \item \textbf{Pairwise Stereo Matching:} Compute the depth or disparity maps for each pair of images, based on the feature matches. This can be done using block-matching, dynamic programming, or graph-cut algorithms, for example.
    \item \textbf{3D Reconstruction:} Merge the depth or disparity maps from all image pairs into a single, consistent 3D model. This can involve techniques like depth map fusion, volumetric reconstruction, or point cloud merging.
    \item \textbf{Post-processing (optional):} Refine the 3D model by applying methods like meshing, texturing, and hole-filling to improve the visual quality and completeness of the result.
\end{enumerate}
Multi-View Stereo has various applications, including:
\begin{itemize}
\item 3D scanning and modeling of objects and environments
\item Photogrammetry
\item Augmented and virtual reality
\item SLAM (Robotics and autonomous navigation)
\item Cultural heritage preservation
\end{itemize}

In summary, Multi-View Stereo is a technique used to reconstruct 3D geometry from multiple 2D images taken from different viewpoints. It involves feature extraction and matching, pairwise stereo matching, and 3D reconstruction, and can be applied in diverse fields such as photogrammetry, augmented reality, and robotics.

\item \textbf{SfM (Structure from Motion)} is a computer vision technique used to estimate the 3D structure of a scene and the motion (camera positions and orientations) from a set of 2D images taken from different viewpoints. It combines elements of both feature-based image matching and 3D reconstruction to create a sparse 3D point cloud or a more detailed 3D model.

The main steps involved in Structure-from-Motion are:
\begin{itemize}
\item \textbf{Image Acquisition:} Capture a set of images of a scene or object from multiple viewpoints, ensuring sufficient overlap and coverage.
\item \textbf{Feature Extraction and Matching:} For each image, identify distinctive points or features and match them across the different images. This can be achieved using feature detectors and descriptors, such as SIFT, SURF, ORB, etc.
\item \textbf{Estimating Camera Poses and 3D Structure:} Use the feature matches to simultaneously estimate the camera's extrinsic parameters (position and orientation) and the 3D coordinates of the features in the scene. This can be done using algorithms like bundle adjustment, which minimizes the reprojection error between the observed 2D feature positions and the reprojected 3D points.
\item \textbf{Incremental Reconstruction:} Add images one at a time, updating the camera poses and 3D structure iteratively. This process is known as incremental SfM and helps maintain accuracy and consistency in the reconstruction.
\item \textbf{Dense Reconstruction (optional): }Perform a dense Multi-View Stereo (MVS) on the sparse 3D point cloud obtained from SfM to generate a more detailed 3D model.
\item \textbf{Post-processing (optional):} Refine the 3D model by applying methods like meshing, texturing, and hole-filling to improve the visual quality and completeness of the result.
\end{itemize}
In summary, Structure-from-Motion is a technique used to estimate the 3D structure of a scene and the motion (camera positions and orientations) from a set of 2D images taken from different viewpoints. It involves feature extraction and matching, camera pose and 3D structure estimation, and optional dense reconstruction and post-processing, and can be applied in diverse fields such as photogrammetry, augmented reality, and robotics.

\end{itemize}
\item \textbf{Structured Light Scanning and Rendering} is a 3D operation for getting RGB-D data using LASERs to scan the surface of the object which needs to be sampled for the data. The LiDAR technology works on these principles. LiDAR is often adjusted on cars for autonomous driving capabilities.
\item \textbf{Time-of-Flight Scanning} refers to the scanning carried out using planes or quadcopters for the purpose of understanding the terrain under the position where the plane/drone is flying.
\item \textbf{Jitter} refers to the noise while streaming a sequence of images using different medium. This can restrain the company from getting optimal sampled inp
\item \textbf{Image and Facial Features}
\begin{itemize}
\item \textbf{SIFT (Scale-Invariant Feature Transform)}~\cite{lowe2004distinctive} is a feature extraction method proposed by David Lowe in 1999. It is designed to be invariant to scale, rotation, and illumination changes. The main steps of the SIFT algorithm are:
\begin{enumerate}
    \item Scale-space extrema detection: Identify potential interest points by searching for local maxima and minima across multiple scales.
    \item Keypoint localization: Refine the candidate keypoints by eliminating low-contrast and poorly localized points.
    \item Orientation assignment: Assign a dominant orientation to each keypoint based on local gradient histograms.
    \item Keypoint descriptor computation: Create a feature descriptor for each keypoint using histograms of gradient orientations in the keypoint's local neighborhood.
\end{enumerate}
\item \textbf{SURF (Speeded-Up Robust Features)}, proposed by Herbert Bay et al. in 2006, is an image feature extraction method inspired by SIFT but designed to be faster and more efficient. The main steps of the SURF algorithm are:
\begin{enumerate}
    \item Interest point detection: Use the Hessian matrix to detect blob-like structures in the image at multiple scales.
    \item Keypoint localization: Identify the local maxima and minima of the determinant of the Hessian matrix as potential keypoints.
    \item Orientation assignment: Assign a dominant orientation to each keypoint based on the Haar wavelet responses in its local neighborhood.
    \item Keypoint descriptor computation: Create a feature descriptor for each keypoint using Haar wavelet responses in the keypoint's local neighborhood.
\end{enumerate}

\item \textbf{ORB (Oriented FAST and Rotated BRIEF)} proposed by Ethan Rublee et al. in 2011, is a fast binary descriptor that combines the strengths of the FAST keypoint detector and the BRIEF descriptor. The main steps of the ORB algorithm are:
\begin{enumerate}
    \item Interest point detection: Use the Hessian matrix to detect blob-like structures in the image at multiple scales.
    \item Keypoint detection: Use the FAST (Features from Accelerated Segment Test) corner detector to identify interest points in the image.
    \item Orientation assignment: Compute an orientation for each keypoint based on the intensity centroid of its local neighborhood.
    \item Keypoint descriptor computation: Create a feature descriptor for each keypoint using the BRIEF (Binary Robust Independent Elementary Features) descriptor, which is modified to be rotation-invariant.
\end{enumerate}

\item \textbf{3DMM feature} refer to 62D features for the face alignment (yaw, pitch, roll) that helps to understand the face of the person being considered for the sampled input. 3DMM, introduced by Volker Blanz and Thomas Vetter in 1999, is a statistical model that represents the 3D shape and texture of human faces. The 3DMM is generated by analyzing a dataset of 3D face scans, capturing the variations in shape and texture across individuals. The main components of a 3DMM are:
\begin{enumerate}
    \item Shape model: A principal component analysis (PCA) is applied to the 3D facial geometry data to create a linear shape model, which can represent different face shapes as a combination of the mean face shape and a set of principal components.
    \item Texture model: Similarly, PCA is applied to the texture data (e.g., color or reflectance) of the 3D face scans to create a linear texture model.
    \item Model fitting: Given a 2D image or a set of images, the 3DMM can be fit to the input data by estimating the shape and texture coefficients, as well as the camera parameters and illumination conditions, that minimize the discrepancy between the input
\end{enumerate}
\end{itemize}
\item \textbf{RANSAC (RANdom SAmple Consensus)} is an iterative algorithm used in the field of computer vision and other areas for robustly estimating model parameters from a dataset that contains outliers. The main idea behind RANSAC is to repeatedly select a random subset of the data points, fit the model to this subset, and evaluate the quality of the fit. The model with the highest consensus from the data points is chosen as the final result.
The RANSAC algorithm consists of the following steps:
\begin{itemize}
    \item Randomly select a minimal subset of data points: Choose the minimum number of data points required to fit the model (e.g., 2 points for a line, 3 points for a plane, etc.).
    \item Fit the model: Estimate the model parameters using the selected data points.
    \item Evaluate the model: Calculate the error between the model's predictions and the remaining data points. If the error is below a predefined threshold, consider the data point as an inlier, otherwise, as an outlier.
    \item Count the inliers: Determine the number of inliers that support the fitted model.
    \item Iterate: Repeat steps 1-4 a predefined number of times or until a stopping criterion is met (e.g., the proportion of inliers exceeds a threshold or the maximum number of iterations is reached).
    \item Refine the model: Fit the model again using all inliers from the best iteration to obtain the final model parameters.
\end{itemize}

RANSAC is particularly useful when dealing with data that contains a significant number of outliers, as it is resistant to their influence. The algorithm is widely used in computer vision tasks, such as feature matching, homography estimation, fundamental matrix estimation, and camera pose estimation.

Hence, RANSAC is an iterative algorithm for robustly estimating model parameters from a dataset with outliers. It works by repeatedly selecting a random subset of data points, fitting the model, and evaluating the consensus of the data points. The model with the highest consensus is chosen as the final result. RANSAC is widely used in computer vision and other fields where robust parameter estimation is required.

\item \textbf{Selective Search} is a region proposal algorithm used in the context of computer vision, specifically for object detection tasks. The main idea behind Selective Search is to generate potential bounding boxes or regions of interest that are likely to contain objects in an image. These regions are then used as input for a classifier or a more complex detection pipeline to identify and classify objects.

Selective Search combines the strengths of both exhaustive search and segmentation to create a diverse set of region proposals. The algorithm is based on the assumption that objects can be identified at different scales and are often characterized by distinct colors, textures, or intensity patterns.

The main steps of Selective Search are:
\begin{enumerate}
    \item Generate initial regions: Perform an initial over-segmentation of the image using a simple method like the Felzenszwalb and Huttenlocher's graph-based segmentation algorithm. This step creates many small, highly detailed segments.
    \item Extract features: For each segment, compute features like color, texture, and shape.
    \item Hierarchical grouping: Recursively merge the most similar segments based on a similarity measure computed from the extracted features. This step creates a hierarchical structure of regions, where each level represents a different granularity or scale.
    \item Generate region proposals: Collect regions from different levels of the hierarchy and use non-maximum suppression to remove redundant or overlapping proposals. This step produces a diverse set of potential object locations.
\end{enumerate}

Selective Search is particularly useful as a preprocessing step for object detection algorithms, as it reduces the number of candidate regions that need to be processed by the actual object detector. This can significantly speed up the detection process and improve the overall performance. Selective Search was widely used in combination with traditional object detectors, like those based on the Histogram of Oriented Gradients (HOG) features, and early deep learning-based detectors, like R-CNN.

\item \textbf{Bundle-Adjustment} is an optimization technique used in computer vision and photogrammetry for refining estimates of the 3D structure of a scene and the camera parameters (poses and intrinsics) from a set of overlapping images. It is often employed as a final step in Structure-from-Motion (SfM) and Multi-View Stereo (MVS) pipelines to improve the accuracy and consistency of the 3D reconstruction.

The main idea behind Bundle Adjustment is to minimize the reprojection error, which is the difference between the observed 2D positions of features in the images and the reprojected 3D points onto the images using the estimated camera parameters and 3D structure. Bundle Adjustment is a non-linear least squares optimization problem, as the relationship between the camera parameters, 3D structure, and image measurements is non-linear.

The main components of Bundle Adjustment are:
\begin{itemize}
    \item \textbf{Camera parameters:} The intrinsic (e.g., focal length, principal point, lens distortion) and extrinsic (e.g., rotation and translation) parameters of each camera in the scene.
    \item \textbf{3D structure:} The 3D coordinates of the scene points that are visible in the images.
    \item \textbf{Image measurements:} The 2D coordinates of the features in each image, usually obtained through feature matching.
    \item \textbf{Objective function:} The sum of the squared reprojection errors between the observed image measurements and the reprojected 3D points, given the current estimates of the camera parameters and 3D structure.
    \item \textbf{Optimization algorithm:} A non-linear least squares solver, such as the Levenberg-Marquardt algorithm, is used to iteratively minimize the objective function.
\end{itemize}

So, Bundle Adjustment is an optimization technique used to refine the estimates of the 3D structure of a scene and the camera parameters by minimizing the reprojection error. It is widely employed in computer vision and photogrammetry pipelines, such as Structure-from-Motion and Multi-View Stereo, to improve the accuracy and consistency of the 3D reconstruction.

\item \textbf{Shepard Metzler objects} are complex 3D shapes, originally introduced by psychologists Roger Shepard and Jacqueline Metzler in a 1971 study on mental rotation. In the context of computer vision, these objects have been adopted as a tool to assess and evaluate the performance of models in tasks related to 3D understanding, such as 3D object recognition, pose estimation, and unsupervised feature learning.

Shepard-Metzler objects are typically constructed by combining several simple 3D shapes, such as cubes or cylinders, into a single complex structure. The resulting objects exhibit a high degree of variability in terms of their geometry, making them well-suited for testing the ability of computer vision algorithms to recognize, analyze, and understand 3D structures.

Some common use cases of Shepard-Metzler objects in computer vision research include:
\begin{itemize}
    \item Benchmarking: Researchers use Shepard-Metzler objects as a benchmark dataset to evaluate the performance of their algorithms in tasks like 3D object recognition and pose estimation. By using these complex objects, they can test the robustness and generalization capabilities of their models.
    \item Unsupervised feature learning: Shepard-Metzler objects are used to study unsupervised feature learning in the context of 3D understanding. Researchers may train models to learn meaningful representations of these objects, which can then be used for downstream tasks such as classification, segmentation, or reconstruction.
    \item Mental rotation tasks: Inspired by the original psychological study, computer vision researchers can use Shepard-Metzler objects to explore how models perform mental rotation tasks, which involve determining whether two 3D objects are the same when they are presented at different orientations.
\end{itemize}

\item \textbf{Super-Resolution Tasks} aim to enhance the quality of an image or recover missing or degraded information. While super-resolution often refers to increasing the spatial resolution of an image, it also encompasses other image enhancement tasks like denoising, dehazing, deblurring, and inpainting. These tasks address different types of image degradation and improve the overall quality and visual appearance of an image.

\begin{itemize}
\item \textbf{Image Denoising} is the process of removing noise from an image while preserving its original details. Noise can be introduced during image acquisition (e.g., due to sensor limitations) or transmission (e.g., compression artifacts). Denoising techniques range from traditional filtering methods, such as Gaussian filters or non-local means, to more advanced deep learning-based approaches, like denoising autoencoders and convolutional neural networks.

\item \textbf{Image Dehazing} refers to the removal of haze or fog from an image to improve its visibility and contrast. Haze is caused by the scattering of light due to the presence of atmospheric particles, which can result in low contrast and color shift. Dehazing methods usually estimate the transmission map and atmospheric light to recover the haze-free image. Classical methods include Dark Channel Prior and Guided Filtering, while deep learning-based methods leverage convolutional neural networks for end-to-end dehazing.

\item \textbf{Image Deblurring} aims to recover a sharp image from a blurred one, typically caused by camera shake, object motion, or out-of-focus effects. Deblurring techniques estimate the blur kernel (also known as point spread function) and use it to restore the sharp image. Traditional methods include blind deconvolution and Wiener filtering, while deep learning-based methods utilize convolutional neural networks to either directly estimate the sharp image or the blur kernel.

\item \textbf{Image Inpainting} is the process of filling in missing or corrupted parts of an image in a visually plausible manner. This task requires not only filling in the missing regions but also preserving the structure and texture of the surrounding areas. Classical inpainting methods include diffusion-based techniques and patch-based approaches like exemplar-based inpainting. More recent deep learning-based methods leverage generative models, such as autoencoders and generative adversarial networks (GANs), to learn the underlying structure and texture patterns and generate plausible content for the missing regions.
\item \textbf{SISR (Single Image Super-Resolution)} aims to recover a high-resolution (HR) image from a single low-resolution (LR) input image. Traditional methods include interpolation-based techniques (e.g., bicubic interpolation, Lanczos resampling) and learning-based approaches (e.g., sparse coding). More recently, deep learning-based methods, such as super-resolution convolutional neural networks (SRCNN) and generative adversarial networks (GANs), have demonstrated significant improvements in reconstructing HR images with more accurate textures and details.
\item \textbf{Video super-resolution} aims to improve the spatial resolution of video sequences while maintaining temporal consistency. This task presents additional challenges due to the need to consider motion between frames and avoid flickering artifacts. Techniques for video super-resolution include multi-frame methods, which exploit the temporal information present in multiple frames, and deep learning-based methods, which use convolutional neural networks (CNNs) or recurrent neural networks (RNNs) to model the spatial and temporal relationships between frames.
\item \textbf{Image colorization} is the process of adding color to grayscale images while preserving their original structure and texture. Traditional methods often require user input in the form of color scribbles or reference images to guide the colorization process. Deep learning-based approaches, such as convolutional neural networks (CNNs) and generative adversarial networks (GANs), have shown the ability to automatically learn color patterns and generate plausible colorizations without user intervention.
\item \textbf{Image-to-Image translation} is a more general task that involves converting an input image from one domain or representation to another, such as converting a sketch to a photorealistic image or transforming a day-time scene to a night-time scene. Deep learning-based methods, such as conditional GANs (e.g., Pix2Pix) and cycle-consistent GANs (e.g., CycleGAN), have been successful in learning the mapping between different image domains without requiring paired training data.
\item \textbf{Depth estimation} is the task of recovering depth information from 2D images, which can be used for various applications like 3D reconstruction, autonomous navigation, or augmented reality. Traditional methods typically rely on stereo vision or structured light, while more recent deep learning-based approaches use convolutional neural networks (CNNs) to predict depth maps directly from single images or monocular video sequences.
\end{itemize}
\item \textbf{Image Warping}, also known as image transformation or image registration, is a process used in computer vision and graphics to manipulate an image's shape, size, and orientation by applying a transformation to its pixel coordinates. The goal of image warping is to change the appearance of an image while preserving its structure and content. It is commonly used in tasks such as image stitching, rectification, stabilization, and view synthesis.

Image warping can be performed using different types of transformations, including:
\begin{enumerate}
    \item Rigid Transformations: These transformations preserve the distances between points and include rotation, translation, and their combination. Rigid transformations are often used in image registration to align images taken from different viewpoints or at different times.
    \item Affine Transformations: Affine transformations extend rigid transformations by adding scaling and shearing effects. They preserve parallelism but not necessarily angles or distances. Affine transformations are commonly used in tasks like image rectification, where the goal is to remove perspective distortion and align parallel lines in the image.
    \item Homographies: Homographies are projective transformations that map points in one image plane to another while preserving straight lines. They can represent a wide range of transformations, including rotation, translation, scaling, shearing, and perspective distortion. Homographies are frequently used in tasks like image stitching, where multiple images of a scene are combined into a single panoramic image, and image rectification.
    \item Non-linear or Deformable Transformations: These transformations can model complex, non-linear changes in the image geometry, such as bending, stretching, or twisting. Non-linear transformations are often employed in tasks like image morphing, where one image is smoothly transformed into another, or in deformable registration, where images with non-rigid structures need to be aligned.
\end{enumerate}
To apply a transformation to an image, a mapping function is used to determine the correspondence between the source image's pixel coordinates and the destination image's pixel coordinates. Then, a resampling or interpolation method, such as nearest-neighbor, bilinear, or bicubic interpolation, is applied to estimate the pixel values in the destination image.

\item \textbf{Decal sheet or Texture Atlas} is a ``sticker'' picture that is applied onto a 3D mesh to represent a texture.
\item \textbf{UV-mapping} is the process of 3D surface projection to a 2D image for texture mapping. Here, $U$ and $V$ represents the axes of 2D texture because in Computer Graphics, $X, Y, Z$ are used to denote axes of 3D surfaces, points, objects and $W$ is used to denote calculation of Quarternions rotations. In \textbf{UV texture map}~\cite{mullen2011mastering}, an ordinary image can be used to paint a 3D surface with color (much like what is done in Unity). The UV mapping technique involves "programmatically" copying a triangle-shaped section of the image map and pasting it onto a triangle on the object to assign pixels in the picture to surface mappings on the polygon~\cite{murdock20083ds}.  UV texturing is an alternative to projection mapping; it simply maps into a texture space rather than the physical space of the object (e.g., utilizing any pair of the model's X, Y, and Z coordinates or any modification of the position). The UV texture coordinates are used in the rendering computation to specify how to paint the three-dimensional surface.\\

\begin{figure*}[!htp]
    \centering
    \includegraphics[width=\textwidth]{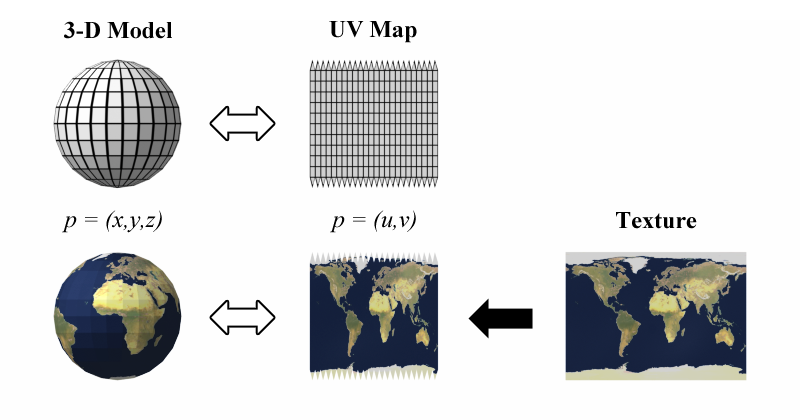}
    \caption{A schematic representation of an application of a texture in the UV space related to effect in 3D space}
    \label{fig:workflow}
\end{figure*}

UV coordinates can be generated for each vertex of the mesh for a model created as a polygon mesh. A way to achieve this is by unfolding the triangular meshes at the seams, hence, laying out triangles on flat surface automatically. A UV sphere might be transformed into an equirectangular projection. The textures can be painted on the triangle individually after the model is unwrapped and when rendering the 3D object, it can take up the appropriate texture from the texture atlas.\\
UV coordinates are applied to each face optionally~\cite{murdock20083ds} so a shared spatial vertex position can have different UV coordinates for each of its triangles so adjacent triangles may be cut apart and positioned on texture maps' different areas. For finding the UV coordinates of a point $\hat{A}$ on a sphere with $\hat{d}$ (distance of the point from the sphere's origin) when sphere's poles are aligned with the $Y$, then the eq.~\ref{eq:t}
\begin{equation}\label{eq:t}
    u = 0.5 + \frac{\textrm{arctan2}(d_{z}, d_{x})}{2 \pi}, 
    v = 0.5 + \frac{\textrm{arcsin}(d_{y})}{\pi}
\end{equation}
Hence, the UV-mapping process can be described in simple three stage approach where the mesh is first uwrapped, a texture is created and applied to the respective face of the polyon.

\item \textbf{Snell's Law}, also known as the \textbf{Law of Refraction}, describes the relationship between the angles of incidence and refraction for a light wave passing through the interface between two media with different refractive indices. It is named after Dutch mathematician Willebrord Snellius, who formulated the law in 1621.

Snell's Law states that the ratio of the sine of the angle of incidence ($\theta_1$) to the sine of the angle of refraction ($\theta_2$) is equal to the ratio of the refractive indices of the two media:
\begin{equation}
    \eta_1 * \sin{\theta_1} = \eta_2 * \sin{\theta_2}
\end{equation}
where $\eta_1, \eta_2$ are the refractive indices of the first and second media, respectively. The refractive index of a medium is a measure of how much the speed of light is reduced inside the medium compared to its speed in a vacuum. $\theta_1$ is the angle of incidence, which is the angle between the incident light ray and the normal (a line perpendicular to the interface) at the point where the light ray enters the second medium. $\theta_2$ is the angle of refraction, which is the angle between the refracted light ray and the normal at the point where the light ray enters the second medium.
Snell's Law is a fundamental principle in optics and is used to explain various phenomena such as refraction, total internal reflection, and the formation of images by lenses. It is also employed in the design and analysis of optical systems, including cameras, microscopes, and fiber optics.

\item \textbf{Fresnel's Law}, also known as Fresnel's Equations, describes the behavior of light when it encounters the interface between two different media, such as air and glass. It is named after French physicist Augustin-Jean Fresnel, who developed the equations in the early 19th century. Fresnel's Law helps explain how light is reflected and transmitted (refracted) at the interface between two media with different refractive indices. These equations are fundamental to understanding various optical phenomena and are used in the design and analysis of optical systems.

Fresnel's Law consists of two equations that describe the reflection (eqn~\ref{eqn:59}) and transmission (eqn~\ref{eqn:60}) coefficients for the parallel ($p$) and perpendicular ($s$) components of the electric field of an electromagnetic wave as described by the eqn~\ref{eqn:61}.
\begin{itemize}
    \item \textbf{Reflection coefficients:}
\begin{equation}
\label{eqn:59}
\begin{aligned}
R_s = \left( \frac{\eta_1  * \cos{\theta_1} - \eta_2  * \cos{\theta_2}}{\eta_1  * \cos{\theta_1} + \eta_2  * \cos{\theta_2}} \right)^2\\
R_p =  \left( \frac{\eta_1  * \cos{\theta_2} - \eta_2 * \cos{\theta_1}}{\eta_1  * \cos{\theta_2} + \eta_2 * \cos{\theta_1}} \right)^2
\end{aligned}
\end{equation}
\item \textbf{Transmission coefficients:}
\begin{equation}
\label{eqn:60}
\begin{aligned}
T_s = 1 - R_s\\
T_p = 1 - R_p
\end{aligned}
\end{equation}

\end{itemize}

The total reflection and transmission coefficients can be computed as the average of the parallel ($p$) and perpendicular ($s$) components:
\begin{equation}
    \label{eqn:61}
    \begin{aligned}
    R = (R_s + R_p) / 2\\
    T = (T_s + T_p) / 2
    \end{aligned}
\end{equation}
These coefficients represent the fraction of the incident light that is reflected and transmitted at the interface between the two media.

Fresnel's Law is essential for understanding various optical phenomena such as reflection, refraction, total internal reflection, and the behavior of polarized light. It is also used in the design and analysis of optical systems, including lenses, mirrors, coatings, and filters.
\end{itemize}
(Work in Progress...)

\section{History of Neural Rendering and NeRFs}
\label{history}
The inception of works in Novel View Synthesis can be traced back to early research in computer graphics and computer vision, where researchers started to explore the use of 3D models and rendering which were later adapted for synthesizing novel views of a scene. One such work for rendering ``Hierarchical geometric models for visible surface algorithms''~\cite{clark1976hierarchical} allowed for more efficient visibility determination and rendering of complex scenes. Although this paper focused on rendering and not on novel view synthesis, the concepts and techniques presented later had a significant impact on the development of model-based novel view synthesis. A work~\cite{gortler1996lumigraph} that was influence by this early work introduced an IBR (image-based rendering) technique that used a hierarchical data structure to represent the scene radiance as a function of position and direction efficiently. It leveraged the idea of hierarchical geometric models to render efficiently and determine visibility. It allowed for the synthesis of novel views from a set of input images without requiring explicit 3D geometry, making it an early powerful method for view synthesis and image-based rendering.

Some attempts of Model-free Novel View Synthesis from Images Rendering date back to 1990s during the time when the foundational book ``Digital Image Warping''~\cite{wolberg1990digital} came out, providing a comprehensive overview of image warping and morphing techniques. This book presented various image transformation techniques, including geometric transformations, spatial transformations, and image resampling. It covered the mathematical foundations and algorithms behind all these techniques, such as coordinate transformations, interpolation methods, and sampling theory. The concepts presented in this book have laid the groundwork for various novel view synthesis methods, as they require image warping to generate new views from input images. By offering both theoretical background and practical examples, "Digital Image Warping" has served as a valuable resource for researchers and practitioners in computer graphics, image processing, and computer vision, influencing the development of numerous techniques and algorithms.

The book ``Three-dimensional computer vision: a geometric viewpoint''~\cite{faugeras1993three} introduced the geometric aspects of 3D computer vision comprehensively by building on mathematical foundations of projective geometry, camera models, and estimation techniques essential for many computer vision tasks. It further presented various methods for recovering the 3D structure of a scene from images, such as stereo vision, SfM, and SfS (shape from shading). The geometric concepts and techniques it covered became the basis of many view synthesis algorithms. By providing a thorough understanding of the geometric principles underlying 3D computer vision, it enabled researchers to develop more advanced and accurate methods for tasks such as novel view synthesis, scene reconstruction, and object recognition. Both these books contributed vitally to the field of computer vision and novel view synthesis by providing foundational knowledge on image warping and 3D geometry. They influenced and progressed the development of numerous techniques and algorithms in computer vision, laying the groundwork for researchers to explore more advanced and sophisticated methods for novel view synthesis and related tasks.

The ``Digital Image Warping''~\cite{wolberg1990digital} delved into practical aspects and applications of image warping, such as image registration, image mosaicking, and image morphing. Later works like Feature-based Image Metamorphosis~\cite{beier1992feature}, View Interpolation for Image Synthesis~\cite{chen1993view}, Image metamorphosis using snakes and free-form deformations~\cite{lee1995image}, Representation of scenes from collections of images~\cite{kumar1995representation}, and View Morphing~\cite{seitz1996view}, were heavily inspired by this work. At that time, Feature-based Image Metamorphosis had been used to transform one image into another by smoothly interpolating the features of the source and target images. The metamorphosis process involves establishing correspondences between key features in the source and target images, interpolating the geometric and color information, and blending the two images over time to create a smooth transition. Although it wasn't a direct technique for novel view synthesis, it was applied to this task with minor modifications such as capturing multiple images of the scene, estimation of camera parameters (SfM or Bundle Adjustment can be used), Identifying feature correspondences (SIFT, SURF, or ORB), estimating Novel views (interpolating the camera parameters of the input images), warping of input images (warping process leverage techniques like piecewise affine warping or thin-plate splines), and finally blending warped images (weighted blending).

An image-based rendering technique called view interpolation was introduced in ``View Interpolation for Image Synthesis''~\cite{chen1993view} which allowed for the generation of novel views of a scene from a sparse set of input images. The method worked by estimating the depth information between the input images, followed by warping the images to the desired viewpoint, and finally combining the warped images using a blending technique. This approach reduced the need for explicit 3D models and enables the synthesis of novel views with relatively simple computations. Later, ``Representation of scenes from collections of images''~\cite{kumar1995representation} proposed a technique to construct a 3D representation of a scene from a collection of images taken from different viewpoints. It involved estimating the geometry of the scene using stereo vision techniques, constructing a 3D mesh based on the depth information, and using image warping to project the textures onto the mesh. The scene representation would then be used for various applications, such as view synthesis or object recognition. ``Image metamorphosis using snakes and free-form deformations''~\cite{lee1995image} shared some similarities with ``Digital Image Warping'' and presented an alternative method for image morphing to combines snakes (active contours) and free-form deformations to produce smooth and visually appealing transitions between images. It involved using snakes (instead of features) to establish correspondences between the source and target images and using free-form deformations to interpolate the features and warp the images. It addressed some of the limitations of traditional feature-based morphing techniques and improves the quality of the morphing results. ``View Morphing''~\cite{seitz1996view} is an image-based rendering technique that generates novel views of a scene by interpolating between two or more input images. The approach involves estimating the scene geometry, warping the input images to align them with the desired viewpoint, and blending the images to create a smooth transition between the viewpoints. The technique is able to handle occlusions and disocclusions in the scene and can be used for various applications, such as view synthesis, image mosaicking, and video summarization.

``3-D scene representation as a collection of images'' proposed to use a collection of images taken from different viewpoints, along with their camera parameters, to represent a 3D scene. They develop a methodology for registering these images in a common coordinate system and use this registration to synthesize novel views of the scene by interpolating between the input images. The approach is image-based, as it does not rely on explicit 3D geometry but rather uses the input images and their associated camera parameters to synthesize new views of the scene. By focusing on the input images and their registration, this method demonstrates the potential of image-based representations for 3D scene understanding and view synthesis tasks without the need for constructing detailed 3D models. The ``Light Field Rendering''~\cite{levoy1996light} introduced the concept of light field rendering, a technique for synthesizing novel views of a scene using a dense set of input images. These light fields represent the radiance of a scene as a function of position and direction and can be thought of as a 4D function. The paper described how to efficiently sample, reconstruct, and render the light field using a set of 2D images, allowing for the synthesis of new views without the need for explicit 3D geometry. Further, research such as ``Physically-Valid View Synthesis by Image Interpolation''~\cite{seitz1995physically} presented a view synthesis technique that uses image interpolation to create novel views of a scene. It was based on the idea of ``view morphing,'' which combined two or more input images to synthesize a new view using epipolar constraints. The paper discussed a method that took into account the geometric relationships between the input images, the camera positions, and the scene structure, ensuring that the synthesized views are physically valid. This research on Morphing has later also been used in different research constituting different surfaces and objects~\cite{gregory1999interactive}.

Proceedings such as ``Image-based Modeling and Rendering''~\cite{debevec1998image} have hence been influential contributors for theshift of focus from model-free novel view synthesis approaches to model-based approaches. The talk surveyed different image-based rendering techniques, which included both model-based and image-based methods for synthesizing novel views of a scene. The authors discussed the advantages and limitations of both approaches, highlighting the need for more accurate and efficient algorithms for handling 3D data in novel view synthesis. The traction gained from such keynote talks gained traction and led to the development in the field of model-based novel view synthesis. Further, it was also driven by the following advantages which accumulated over time.
\begin{enumerate}
\item \textbf{Availability of 3D data:} With advances in 3D scanning technologies and the increasing availability of 3D models, researchers have been able to leverage this data to create more accurate and realistic novel view synthesis algorithms.
\item \textbf{Advances in computer graphics and computational power:} The development of sophisticated rendering techniques and the increase in computational power have made it possible to work with complex 3D models and generate realistic novel views in real-time.
\item \textbf{Improved algorithms for 3D reconstruction:} Researchers have developed more accurate and efficient algorithms for recovering the 3D structure of a scene from images, making it possible to use model-based approaches for novel view synthesis.
\item \textbf{Handling occlusions:} Model-based approaches can better handle occlusions and disocclusions in the scene, as they have explicit 3D information about the scene geometry.
\end{enumerate}

Later attempts in Neural Rendering usually involved lengthy pipeline operations for creating virtual models of the real scenes. These operations start with a scanning process (using camera images) to capture the photometric properties, and raw scene geometry is captured using depth scanners or dense stereo matching. The Raw Scene Geometry capture is often noisy and provides incomplete point cloud data. This incomplete data can further be processed by applying surface reconstruction and meshing approaches. Particularly, with the context of 3D reconstruction with commodity RGB-D sensors, researchers have made significant progress enabling robust tracking~\cite{izadi2011kinectfusion,newcombe2011kinectfusion,choi2015robust,whelan2016elasticfusion,dai2017bundlefusion}, and large scale 3D scene capture~\cite{chen2013scalable,niessner2013real,zeng2013octree}. Then, the material and texture estimation operations are leveraged to determine the photometric properties from the obtained mesh of surface fragments and store them in different data structures like texture maps~\cite{blinn1976texture}, bump maps~\cite{blinn1978simulation}, view-dependent textures~\cite{debevec1998efficient}, and surface light fields~\cite{wood2000surface}. Basically, textures are alternative that mitigate the missing spatial resolution problem where finding the consistent UV-mapping becomes vital as geometry and color are decoupled. To compensate for wrong geometry and camera drift, some approaches are based on finding non-rigid warps~\cite{zhou2014color,huang20173dlite}.

Lastly, the computationally-heavy rendering operations (ray-tracing/radiance transfer estimation) generate photo-realistic views of the modeled scene. An advantage of non-rigid warps discussed earlier is that 3D reconstructed content can be visualized by standard rendering techniques and approach directly generalizes to 4D capture as long as it can be reliably tracked~\cite{newcombe2015dynamicfusion,dou2016fusion4d,innmann2016volumedeform}. The reconstructions discussed earlier can be used to synthesize images from novel viewpoints. This pipeline of operations has been constructed and polished by research in computer graphics society for several years. Despite that this pipeline generates highly realistic results under controlled settings, several operational stages of this pipeline and consequently the entire pipeline remains brittle until the addition of the domain of Machine Learning. This augmentation has hence encouraged people from Machine Learning, Computer Vision, and Computer Graphic society to come together and make this pipeline more robust. But, the imperfections in reconstruction directly appear when translating the embeddings to visual artifacts in renderings, that became the major hurdle in 2020s to make content creation from real-world accessible.

With the growth of Neural Networks and the inception of CNN-based architectures like AlexNet~\cite{krizhevsky2017imagenet}, VGG-Net~\cite{simonyan2014very}, and ResNets~\cite{he2016deep}, the focus shifted for Machine Learning techniques to permeate the domain of Computer Vision and Computer Graphics. This is evident by the following research literature which leverages CNNs exhaustively. ``View Synthesis by Appearance Flow''~\cite{zhou2016view} introduced an end-to-end trainable CNN model that learned to synthesize target views by predicting appearance flows, which were then used to warp input images, allowing for view synthesis without explicitly estimating 3D geometry. ``Deep View Morphing''~\cite{ji2017deep} proposed a deep learning-based approach to view morphing, leveraging CNNs to learn and predict dense correspondence maps between source and target images, enabling the generation of high-quality novel views. Habtegebrial et al., in ``Fast View Synthesis with Deep Stereo Vision''~\cite{habtegebrial2018fast}, developed a deep learning-based method for view synthesis that combined stereo vision techniques with a CNN, aiming to produce accurate novel views with reduced computational complexity compared to other learning-based approaches. Lastly, ``InLoc: Indoor Visual Localization with Dense Matching and View Synthesis''~\cite{taira2018inloc} by Taira et al. presented a large-scale indoor localization system that matched dense local features between a query image and a database of 3D point clouds~\cite{wijmans2017exploiting}, utilizing view synthesis to improve the matching performance and achieve accurate and robust localization in challenging indoor environments. This was an application of stereo vision models in indoor 3D scene reconstruction. 

An alternative direction is to fill in the missing content by leveraging coarse geometry proxy based on high-resolution 2D textures~\cite{huang20173dlite}. \textbf{PBG (Point-based Graphics)} approaches for rendering and re-rendering pipelines use a collection of points or unconnected disks (surfels) for geometry modeling instead of Surface Mesh Estimation~\cite{levoy1985use,grossman1998point,gross2002point,kobbelt2004survey}. Whereas, the IBR (Image-Based Rendering) techniques~\cite{mcmillan1995plenoptic,seitz1996view,gortler1996lumigraph,levoy1996light} warp the Camera Coordinate System using coarse approximations (low-frequency embeddings in case of functional representation) of scene geometry to obtain the photorealistic views, where the 3D geometry proxy is only used to select suitable views for cross-projection and view-dependant blending~\cite{heigl1999plenoptic,buehler2001unstructured,carranza2003free,zheng2009parallax,hedman2016scalable}. Finally, the deep neural rendering methods~\cite{isola2017image,nalbach2017deep,chen2018deep,bui2018point,hedman2018deep} replace the physics-based rendering with generative Neural Network to rectify some of the mistakes of rendering network. One such method that has gained traction from Computer Vision society is what is the primary focus of this survey--NeRF.


\textbf{LLFF (Local Light Field Fusion)} (example depicted in fig.~\ref{fig:fig2}) is one prior work done by Mildenhall et al~\cite{mildenhall2019local}, some of whose authors were also contributors for the work under focus in this survey. LLFFs leverage MPI (Multiplane Image) scene representations~\cite{zhou2018stereo} for view synthesis from an irregular grid of sampled views for expanding each sampled view into a local light fields~\cite{levoy1996light} and then renders novel views by fusing these fields. LLFFs achieve perceptual quality similar to Nyquist Rate Sampling (with $4000 \times$ less views) by creating a bound for real world scenes creation/rendering which is specifies the density of sample views for a given scene by extending the traditional plenoptic sampling~\cite{chai2000plenoptic} theory. Here, the LLFFs result in a quadratic decrease in the number of input views based on the number of predicted planes for each layered scene representation and still maintain the Nyquist level performance by specializing to the subset of natural scenes. Further, this work was corroborated with a smartphone app and web-cum-desktop application for capturing and viewing respectively.

\begin{figure*}[!htp]
    \centering
    \includegraphics[width=\textwidth]{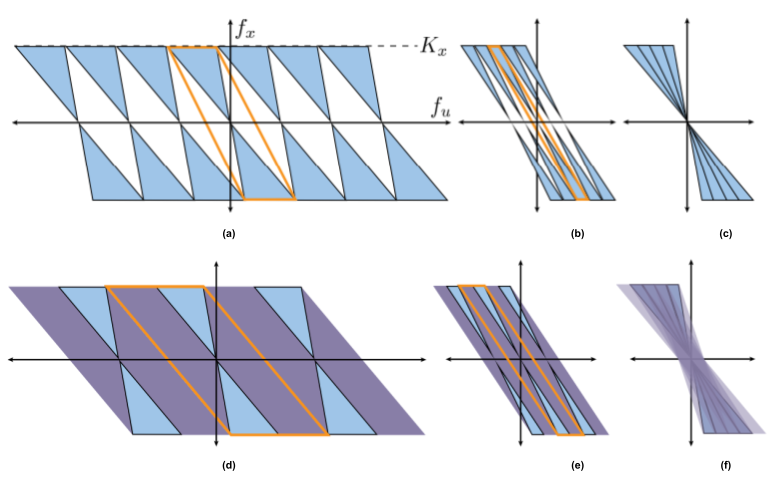}
    \caption{Traditional Plenoptic Sampling without occlusions~\cite{chai2000plenoptic} in (a) where the double-wedge (in blue) encloses the fourier support of a light field without occlusions. Here, the double-wedge width is used to calculate Nyquist Rate Sampling, that determines by minimum and maximum scene depths $[z_{min}, z_{max}]$ and $K_x$ (max spatial frequency). In (b), the light field is split into $D$ non-overlapping layers and Nyquist Sampling Rate decreases by a factor of $D$ and (c) represents that without occlussions, the full light field spectrum is sum of the spectra from each layer. In LLFF~\cite{mildenhall2019local} extends this to mitigate occlusions as continuous light fields are reconstructed from MPIs. In (d), occlusions expand the Fourier supports to ||gm (where blue represents support without occlusions and purple regions depict the occlusion expanding the fourier support) doubling the Nyquist Sampling Rate. Further, (e) depicts that separate reconstruction of the light field for $D$ layers decrease the Nyquist Sampling Rate by a factor of $D$ whereas (f) depicts the case that full light spectrum cannot be reconstructed by summation over individual layer spectra because their support union in smaller than support union in (a). Hence, the need for alpha compositing individual light layers from back to front in the primal domain to compute the full light field. Figures used from LLFF original paper~\cite{mildenhall2019local}}
    \label{fig:fig1}
\end{figure*}

\begin{figure*}[hbt!]
\begin{subfigure}{\textwidth}
  \centering
  \centerline{\includegraphics[width=\textwidth]{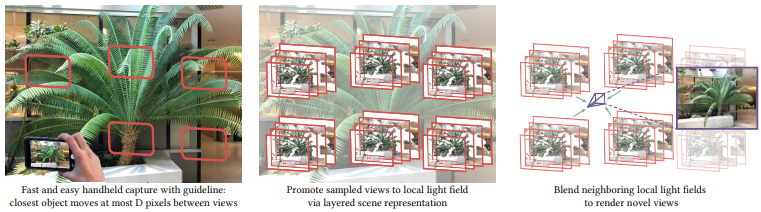}}  
  \caption{A step-by-step representation of Multi-Plane Image Compositing in Local Light Field Fusion by Mildenhall et al~\cite{mildenhall2019local}}
  \label{fig:fig2}
\end{subfigure}
\begin{subfigure}{\textwidth}
  \centering
  \centerline{\includegraphics[width=\textwidth]{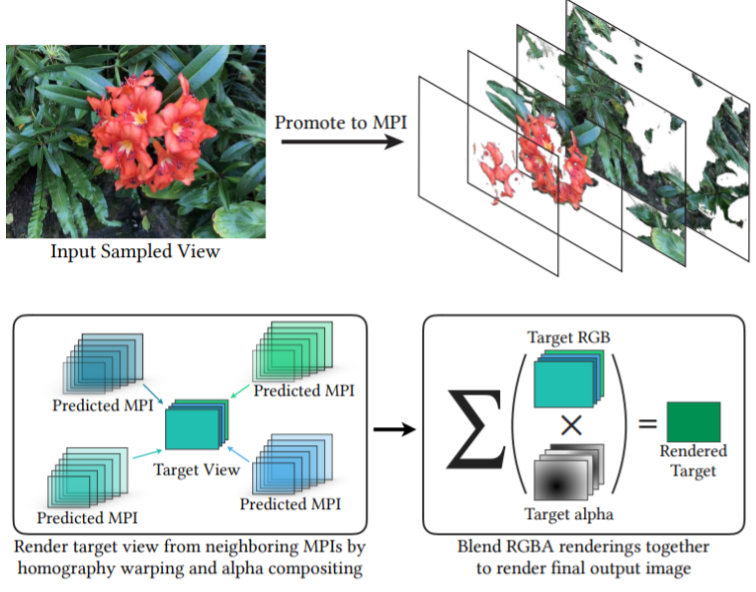}}  
  \caption{LLFF (Local Light Field Fusion)~\cite{mildenhall2019local}: Promotion to MPI Scene Representation for each input image with $D$ RGB$\alpha$ planes at regularly sampled depths in the input view's camera frustrum. Here, each MPI can be used to render continuously-valued novel views using alpha compositing of color value along rays within a local neighborhood within a novel view's (imaginary) camera.}
  \label{fig:fig2-1}
\end{subfigure}
\end{figure*}

This architecture is based on the idea that prescriptive sampling is vital for useful IBR-algorithms. Hence, it extends the prior plenoptic sampling theory to decrease the dense sampling requirements for traditional light field rendering by decomposing a scene into $D$ depth ranges and individual-range-based light field samples. This allows the camera sampling interval to be increased by a factor of $D$ and the light field spectrum emitted by a scene within each depth range to lie within a densely packed shape that is $D$ times tighter that the full scene’s double-wedge spectrum as depicted in fig.~\ref{fig:fig1}. This synthesis pipeline also leaves a possibility for it to be used in the future light field hardwares to reduce required cameras. As discussed by Chai et al~\cite{chai2000plenoptic}, the Light Field's Fourier support (without any occlusions and Lambertian effects~\cite{zhang2003spectral}) lies inside double-wedge shape within scene depths limit $[z_{min}, z_{max}]$ as given by the figure~\ref{fig:fig1}, whereas occlusion expands the limits of this shape in the light field as the occluder acts as convolver due to the kernel (lying on the line corresponding to the occluder's depths) with farther scene content. Further, Fourier support of the light field gets limited due to the closest occluder which convolves the furthest scene content's line resulting in a parallelogram-like-shape as depicted in fig~\ref{fig:fig1}. This can only be half packed when compared to double-wedge shape and hence required maximum camera sampling interval $\Delta_{u}$ for an occluded light field and is given be eqn.~\ref{eqn:0-1}

\begin{equation}
    \label{eqn:0-1}
    \Delta_{u} \leq \frac{1}{2 K_{x} f \left(\frac{1}{z_{min}} - \frac{1}{z_{max}}\right)}, 
\end{equation}

In the continuous light field $B_{x}$ and camera spatial resolution $\Delta_{x}$, $K_{x}$ is maximum spatial frequency in sampled light field. It is given by the eqn.~\ref{eqn:0-2}

\begin{equation}
\label{eqn:0-2}
    K_x = min \left(B_x, \frac{1}{2\Delta_x}\right)
\end{equation}

LLFFs render novel views from an MPI scene representation (that consist of a set of evenly disparity sampled fronto-parallel RGB$\alpha$ planes from view frustrum of reference camera) from local-neighborhood's continuous valued  camera poses by alpha compositing the color rays into the novel views with the ``over'' operation~\cite{porter1984compositing}. This is similar to the early work by Lacroute et al~\cite{lacroute1994fast} and can be considered an encoding of a local light field much like light field displays~\cite{wetzstein2011layered,wetzstein2012tensor}. Further, it alpha composite the depth range fields or shield fields~\cite{lanman2008shield} from back to front for computing the full scene light field to decrease the Sampling Rate for occlusions. These fields plays a vital role in handling occlusions as the Fourier Spectra Union have less degree of confidence when compared to the original occluded light fields as depicted in fig~\ref{fig:fig1}. Each layer-by-layer alpha compositing step increases the Fourier support by convolving previously accumulated light-field's spectrum with occluded depth layer spectrum. The width of Fourier Spectrum Support Parallelogram for the reconstructed light field for  each occluded depth range light field will enjoy full Fourier Support width and is given and illustrate by eqn.~\ref{eqn:0-3} and fig~\ref{fig:fig1}, respectively. 

\begin{equation}
\label{eqn:0-3}
 \frac{2 K_{x} f \left(\frac{1}{z_{min}} - \frac{1}{z_{max}}\right)}{D}, 
\end{equation}

Hence, LLFFs are able to extend the layered plenoptic sampling framework and handle occlusions uisng depth maps and increasing the required sampling interval by a factor of $D$ and is given by the eqn.~\ref{eqn:0-4}. 

\begin{equation}
    \label{eqn:0-4}
    \Delta_{u} \leq \frac{D}{2 K_{x} f \left(\frac{1}{z_{min}} - \frac{1}{z_{max}}\right)}, 
\end{equation}

In addition to this, LLFFs also introduced an additional caveat for the scene's bouding volume to be within a frustums of atleast two neighboring sampled views by introducing the concept of finite field of view instead of infinite field of view~\cite{chai2000plenoptic,zhang2003spectral}. This bound is given by the camera sampling interval~\ref{eqn:0-5} and the constrain given in the eqn~\ref{eqn:0-6}.

\begin{equation}
    \label{eqn:0-5}
    \Delta_{u} \leq \frac{W \Delta_x z_{min}}{2f}, 
\end{equation}

\begin{equation}
    \label{eqn:0-6}
    \Delta_{u} \leq min \left( \frac{D}{2 K_{x} f \left(\frac{1}{z_{min}} - \frac{1}{z_{max}} \right)}, \frac{W \Delta_x z_{min}}{2f}  \right)
\end{equation}

Now, if we set $z_{max} = \infty$ for the traditional assumption and set $K_x = \frac{1}{2 \Delta_x}$ for allowing maximum representable frequency, then we have eqn~\ref{eqn:0-7}.

\begin{equation}
    \label{eqn:0-7}
    \frac{\Delta_u}{\Delta_x z_{min}} = d_{max} \leq min \left( D, \frac{W}{2}  \right)
\end{equation}

where $d_{max}$ is the maximum pixel density of any scene between adjacent scenes. Hence, the maximum density i.e. $d_{max}$ must be less than $min(D, W/ 2)$ and it reduces to the Nyquist bound when $D=1$. Hence, this promotion to $D$ depth layers decreases the required sampling rate by a factor of $D$ until the required field of view overlaps for estimation of stereo geometry. Now, further as we know that any scene can be a function of two viewing directions hence this sampling rate compounds to $D^2$ sampling reduction. 

LLFFs practically synthesize new views from set of input images and camera poses by first using a CNN to promote each captured input imahe to an MPI and then reconstructing novel views by blending the rendered light field from the nearby MPIs as depicted in fig~\ref{fig:fig2} and fig~\ref{fig:fig2-1}. Firstly, 5 images of a scene are taken (1 reference image and 4 nearest neighbors in 3D space) and reprojected to $D$ depth planes and sampled linearly within reference view frustum to get 5 $H \times W \times 3$ PSVs (plane sweep volumes). LLFFs employ a 3D CNN for taking 5 PSVs as inputs to concatenate them along channel dimension which results in opacity $\alpha$ for each MPI pixel $(x, y, d)$ and 5 color selection weights sum to unity at each MPI. This parametrization is leveraged instead of fg-bg parameterization~\cite{zhou2018stereo} as it prevents an MPI from incorporating content occluded from the reference view but visible in other input views. 3D convolutions enables the architecture to predict MPIs with $D$ number of variable planes by involving convolutions across the height, width, and depth instead of fully 2D convolutions. This enables to jointly choose disparity and view sampling densities to satisfy the eqn.~\ref{eqn:0-7}

LLFFs reconstruct interpolated views as a weighted combination of renderings from multiple nearby MPIs. This process combines local light field approximations to a near-plane spanning light field  and a far-planed determines by the input views' field-of-view. This allows to render new biew path without constrains for 3D translation and rotation as in standard light field rendering. LLFFs consider the accumulated alpha values as vital for each MPI rendering when blending to allow  MPI rendering to ``fill in'' content occluded from camera views. LLFFs prediction network leverage RGB image sets $C_k$ with camera poses $p_k$ and produce MPI sets $M_k$ for each image. Each RGB$\alpha$ MPI plane was homographically warped into target pose $p_t$'s reference frame using a predicted MPI $M_k$ and then alpha composited together from back to front. All these steps lead to an RGB image ($C_{t,k}$) and an alpha image ($\alpha_{t,k}$) where the ouput is rendered at pose $p_t$ using MPI pose $p_k$. Further, the final RGB output $C_t$ is obtained by blending rendered images $C_{t,k}$ from multiple MPIs as one MPI does not contain all content visible from the new camera pose as illustrated in fig.~\ref{fig:fig2-1} and given by eqn~\ref{eqn:0-8}.

\begin{equation}
    \label{eqn:0-8}
    C_t = \frac{\sum_{k}w_{t,k} \alpha_{t,k} C_{t,k}}{\sum_{k}w_{t,k} \alpha_{t,k}},
\end{equation}
where the method leverages scalar blending weights $w_{t,k}$ (can be smooth filter) that are modulated by the corresponding accumulated alpha images ($\alpha_{t,k}$) and normalized so that $\alpha = 1$. For instance, modulating the blending weights by referring to eqn.~\ref{eqn:0-8} accumulated alpha values prevents artifacts in $C_t$. Bilinear Interpolation from the 4 nearest MPIs is leveraged for regular grid of data rather than the ideal sinc function interpolation for extremely limited number of sampled views and efficiency. All five MPIs are used for irregularly sampled data and $w_{t,k} \propto exp (-\gamma \ell(p_t, p_k))$. Here, $\ell$ is $L^2$ distance between translation vectors of poses $p_t$ and $p_k$, and $\gamma$ is given by eqn~\ref{eqn:0-9}.

\begin{equation}
    \label{eqn:0-9}
    \gamma = \frac{f}{Dz_{min}},
\end{equation}
 
where $f$ is the focal length and $z_{min}$ is the minimum distance to the scene. This strategy for MPI neighborhood blending is particularly efficient for non-Lambertian renderings. Virtual apparent depth of a specularity changes with viewpoint~\cite{swaminathan2002motion} for curved surfaces and hence, specularities can appear as curves in light fields' epipolar slices and lines can appear for diffuse points. Here, each of the MPI represents a specularity for a local range of views by placing the specularity at a single virtual depth. This approach leverages real world and synthetic dataset of natural scenes~\cite{song2017semantic,qiu2017unrealcv,schoenberger2016sfm,schoenberger2016mvs}. These results that quantitatively validate and demonstrate that LLFFs match the perceptual quality of Nyquist rate Sampling while leveraging $\approx 4000 \times$ fewer input images has been demonstrated for over 60 diverse real world scenes and this work paved way for future IBR algorithms.

 Further, there have been many countless works from 2015--2020 on representing scenes as unstructured or weakly structured feature representations~\cite{tatarchenko2015single,worrall2017interpretable,eslami2018neural} but the first work to implicit learn the function for scene as continuous representation was presented in \textbf{SRNs (Scene Representation Networks)}.\\

\begin{figure*}[!htp]
    \centering
    \includegraphics[width=\textwidth]{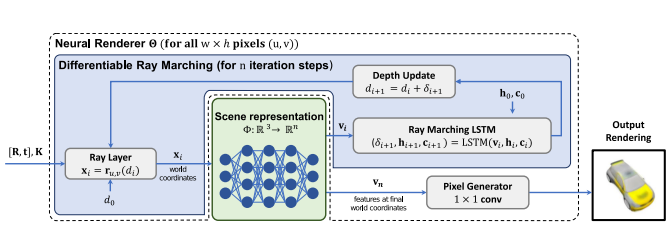}
    \caption{A model schematic representation of SRNs (Scene Representation Networks) by Sitzmann et al~\cite{sitzmann2019scene}}
    \label{fig:fig3}
\end{figure*}

\textbf{SRNs} (proposed by Sitzmann et al; depicted in fig~\ref{fig:fig3}) enapsulated both the geometry and the appearance of a scene by leveraging camera poses~\cite{sitzmann2019scene}. This training was performed end-to-end without explicit suprervision in 3D geometry, purely from a set of posed 2D images. These networks played an important role in constructing the basis for NeRFs and gaining better evaluations for scene reconstructions. All this was done using Differential Ray Marching using LSTM (Long-Short Term Memory networks). Here, the objective was to combine multiple different disciplines such as Geometric Deep Learning, Neural Scene Representation, and Neural Image Synthesis, and Differentiable Ray Marching to represent the scene as a continuous function $\phi$ responsible for matching a feature representation \textbf{v} of learned scene properties for a particular spatial location:

\begin{equation}\label{eqn:1}
    \phi = \mathbb{R}^{3} \rightarrow \mathbb{R}^{n},  \textrm{\textbf{x}} \mapsto \phi (\textrm{\textbf{x}}) = \textrm{\textbf{v}}
\end{equation}
where \textbf{v} may be a feature vector that encodes visual information related to the scene or higher order information such as signed distance of \textbf{x} to closest surface. Hence, this work became one of the first works to formulate Scenes as continuous formulation. Earlier works used to store information in a discretized manner such as Voxel Grids' discretization in $\mathbb{R}^{3}$~\cite{maturana2015voxnet,kar2017learning,nguyen2018rendernet,zhu2018visual,tung2019learning,sitzmann2019deepvoxels} and Point Clouds~\cite{qi2017pointnet,insafutdinov2018unsupervised,lin2018learning,meshry2019neural} may have sparsely discretized points anywhere in the space~$\mathbb{R}^{3}$. Here, the concepts of SRNs was the first one to encode all the information of a scene in a function which was represented by an MLP (Multi-layer Perceptron) which limited the scene information to the capacity of the MLP. Further, SRNs perform Neural Rendering $\theta$ by mapping scene representation $\phi$, intrinsic \textbf{K}, and extrinsic~\textbf{E} to an image~$\textit{I}$ as given by the following equation.
\begin{equation}\label{eqn:2}
    \theta : \chi \times \mathbb{R}^{3 \times 4} \times \mathbb{R}^{3 \times 3} \rightarrow \mathbb{R}^{H \times W \times 3},
    (\phi, \textrm{\textbf{E}}, \textrm{\textbf{K}}) \mapsto \theta (\phi, \textrm{\textbf{E}}, \textrm{\textbf{K}}) = \textit{I}
\end{equation}
where $\chi$ represents universal space for functions. Here, these networks faced a disadvantage at implicitly storing the geometry in the MLPs. Hence, a learned neural ray marching algorithms was used in this approach and a pixel generator network was proposed to map geometry/appearance features to colors. The differentiable Ray Marching Algorithm takes inspiration from Classic Sphere Tracing algorithm~\cite{hart1996sphere} where it solves the eqn~\ref{eqn:3} by starting at a distance close to the camera and learning the step distance and stepping such that ray intersects the geometry (where each step has signed distance length to the closest surface point of the scene)~\cite{van2016conditional}.
\begin{equation}\label{eqn:3}
    \begin{multlined}
    \textrm{arg min } d \\
\textrm{s.t.   \textbf{r}}_{u, v}(d) \in  \Omega,  d > 0
\end{multlined}
\end{equation}
where $\Omega$ refers to all the points on the surface of the scene. The step length is learned using a RM-LSTM (Ray Marching Long-Short-Term Memory)~\cite{hochreiter1997long}. Further, the authors behind SRN also leverage a pixel generator architecture which is based on $1 \times 1$ convolutions instead of 2D convolutions. This is because the 2D convolutions may hinder multi-view consistency in case if they are capturing the context of local area which is disrupted when bringing the camera closer to another adjacent local area of a scene but it also discounts the ability of these networks to work in a smaller memory budget.
SRNs further generalize across different scenes by jointly optimizing the following objectives by leveraging stochastic gradient descent.

\begin{equation}\label{eqn:4}
\begin{multlined}
    \argminA_{\{\theta,\Psi,\{\textrm{\textbf{z}}_j\}^{m}_{j=1}\}} \sum_{j=1}^{M}\sum_{i=1}^{N} \underbrace{||\Theta_{\theta}(\Phi_{\psi(\textrm{\textbf{z}}_{j})}, \textbf{E}_{i}^{j},
    \textbf{K}_{i}^{j})||_{2}^{2}}_{\mathcal{L}_{\textrm{img}}} \\
    \begin{aligned}
    & + \underbrace{\lambda_{depth} ||\textrm{min(\textbf{d}}_{i,final}^{j}, 0)||_{2}^{2}}_{\mathcal{L}_{\textrm{depth}}} \\
    & + \underbrace{\lambda_{latent}||\textrm{\textbf{z}}_j||_{2}^{2}}_{\mathcal{L}_{\textrm{latent}}}
     \end{aligned}
\end{multlined}
\end{equation}

 Here, the first loss was an $l_{2}$ loss which listed the similarity of the constructed or rendered image based on ground-truth of that particular path of area in a scene, whereas $\mathcal{L}_{\textrm{depth}}$ is the constraint in eq.~\ref{eqn:3}, and finally $\textbf{z}_{j}$ on Gaussian Distribution priors. This equation can also further be modified for the purpose of Few-Shot Reconstruction. The scope of experimentation is beyond the scope of this paper, so, the readers are further advised to study the paper in detail to get a better idea of the implementation details.\\

\begin{figure*}[!htp]
    \centering
    \includegraphics[width=\textwidth]{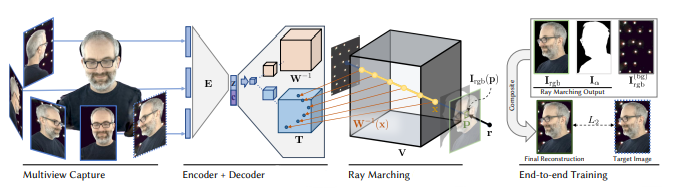}
    \caption{A block representation of NVs (Neural Volumes) by Lombardi et al~\cite{lombardi2019neural}}
    \label{fig:fig4}
\end{figure*}

Lombardi et al~\cite{lombardi2019neural} presented \textbf{NV (Neural Volumes)} as a way to learn both the geometry and appearance variations simultaneously that led to better generalization across viewpoints. Their work has been depicted in figure~\ref{fig:fig4}. To address the issue  of terminating in poor local minima when optimizing over a mesh-based representation using gradient-based optimization, the authors used a volumetric representation as a function of color and opacity at each position in 3D space. Hence, helping in mitigating problems such as accountability for reflectance variability or tracking topological evolution in dynamic media in motion capture systems~\cite{hawkins2005acquisition,roth2006specular,atcheson2008time,xu2014dynamic}. Further, during the optimization of this encoder-decoder model, semi-transparent representation of geometry disperses gradient information along the ray of integration, effectively widening the basin of convergence, enabling the discovery of good solutions. To overcome the limitation of using intensive memory for the sparse voxel grids, the authors worked with a warping technique that indirectly escapes the restrictions imposed by a regular voxel grid structure, allowing the learning algorithm to make the best use of available memory. Further, since the voxel grid structure isn't used directly there were less grid-based artifacts in this approach.

Much like SRNs, the NV method is based on two main parts of the architecture which consist of an encoder-decoder network that converges after mapping image to a 3D volume \textbf{V(x)}, and differentiable ray marching for rendering image from \textbf{V} with a given camera parameters. This is analogous to an Auto-Encoder with fixed function volume rendering operation for the final layer. Here, the model maps from 3D location space to the 3D positions, $\textrm{x} \in \mathbb{R}$ as stated in eqn.~\ref{eqn:5} given below.
\begin{equation}\label{eqn:5}
    \textrm{\textbf{V}}: \mathbb{R}^{3} \rightarrow \mathbb{R}^{4},
    \textrm{\textbf{V(x)}} = (\textrm{\textbf{V}}_\textrm{rgb}(\textbf{\textrm{x}}), \textrm{\textbf{V}}_\alpha(\textbf{\textrm{x}})),
\end{equation}
where $\textrm{\textbf{V}}_\textrm{rgb}(\textbf{\textrm{x}}) \in \mathbb{R}^3$ and $\textrm{\textbf{V}}_\alpha(\textbf{\textrm{x}})$ are the color and opacity (range $[0, \infty]$) at the 3D coordinate \textbf{x}. This enables the model to learn soft discrete volume representation that allowed gradients to flow backward easily and model fine and transparent structures (like hair and smoke).\\
NV also discusses the potential of leveraging only 3 images from 3 orthogonally placed cameras to achieve maximize convergence while downsampling an image by a factor of 8. Here, the Encoder is a Convolutional Variational encoder whereas the authors leverage an MLP for Decoder (which demands prohibitive memory requirement to produce reconstructions with higher quality). Hence, Lombardi et al~\cite{lombardi2019neural} model the volume function as a discrete 3D grid of voxels hence, gaining the output tensor in the form of discretized voxels. The authors here optimized the decoder as a function $S(\textrm{\textbf{x}; \textbf{Y}}): \mathbb{R}^{3} \rightarrow \mathbb{R}^{C}$ to samples from  the grid \textbf{Y}
by scaling continuous values in the range $[-1, 1]$ to $[1, D]$ in each dimension followed by trilinear interpolation. Hence, for  general case of 3D cube with center at $\textrm{\textbf{x}}_{0}$ with $W$ sides is given by the eqn.~\ref{eqn:6}.
\begin{equation}
    \label{eqn:6}
    \textrm{\textbf{V(x; z)}} = S \cdot \left(\frac{\textrm{x - x}_{0}}{W/2} ; \textrm{\textbf{g(z)}}\right),
\end{equation}

Note that here the $\textrm{\textbf{g(z)}}$ represents the MLP used for decoding of the Voxel Grid. The authors also discussed how they can translate a similar architecture with $1 \times 1 \times 1$ cube with 1024 channel for duplicating the function generated by the MLP. Later, a softplus activation to restrain the voxels to positive values rather than negative values that might occur after decoing. Further, a warping function is applied to the voxel decoder to mitigate the wasteful behavior of the Voxel Grids (also solved by Octree-based approaches~\cite{riegler2017octnet} but they are complex and require object distribution to be known a \textit{priori}). Hence, the authors used an inverse warping formulation as given in the equation~\ref{eqn:7} below.

\begin{equation}
\label{eqn:7}
    \textrm{W}^{-1} (\textrm{x}) \rightarrow \textrm{y,     where \textbf{x, y}} \in \mathbb{R}^{3},
\end{equation}

This helps to map points from a warp volume $\textrm{\textbf{W}}^{-1}(\textrm{\textbf{x}})$ to a template volume $\textrm{\textbf{T(x)}}$ where both volumes are decoded from the dynamic latent code \textbf{z}. Hence, in the eqn.~\ref{eqn:7}, \textbf{x} is a 3D coordinate in the output space and \textbf{y} is the corresponding 3D coordinate in RGB$\alpha$ template volume. Hence, the mapping can be depicted by using the eqn.~\ref{eqn:8}.

\begin{equation}
\label{eqn:8}
\textrb{V}_{\textrm{RGB$\alpha$}}(\textbf{x}) = \textrb{T}_{\textrm{RGB$\alpha$}} (\textrb{W}^{-1} (\textrb{x})).
\end{equation}

These warp fields were developed as a mixture of Affine Warps to produce an affine warp field without encountering any overfitting issues or any non-linear bending. This formulation has been depicted by the eqn.~\ref{eqn:9}--\ref{eqn:10} given below.

\begin{equation}
    \label{eqn:9}
     \textrb{W}^{-1} (\textrb{x}) = \sum_{i} \textrb{A}_{i}(\textrb{x}) \textrb{  } a_{i}(\textrb{x}),
\end{equation}

\begin{equation}
    \label{eqn:10}
    \textrm{with  } \textrb{A}_{i} (\textrb{x}) = \textrb{R}_{i} (\textrb{s}_{i  } \circ (\textrb{x - t}_{i})), \textrm{                   } a_{i} (\textrb{x}) = \frac{w_{i}(\textrb{A}_{i} (\textrb{x}))}{\sum_{j} w_{i}(\textrb{A}_{i} (\textrb{x}))},
\end{equation}
 
 where $ \textrb{A}_{i} (\textrb{x})$ is the $i^{th}$ affine transformation, $w_{i}$ is the weight of each warp, and $\circ$ is the element-wise multiplication. Further, the authors also made the RGB decoder constrained by viewpoint conditions by  given the normalized direction of the camera to the decoder along the encodings as an input. Further, the convolutions for the RGB and $\alpha$ component were disentangled as RGB values were conditioned on the viewpoint. Further, for the ray-tracing and End-to-End training this approach leverages Semi-Transparent Volume Rendering and a combination of Reconstruction Priors and Hyperparameter Tuning.\\
 Here, the ray marching accumulates not only color but also opacity to model the occlusions and when opacity hits the value of 1, then no color is accumulated on the ray. Hence, the rendering process for this approach is as given by the eqn.~\ref{eqn:11}--\ref{eqn:12}.
 
 \begin{equation}
     \label{eqn:11}
     \textrb{I}_\textrm{rgb}(\textrb{p}) = \int_{t_{min}}^{t_{max}} \textrb{V}_\textrm{rgb} (\textrb{r}_o + t\textrb{r}_d) \frac{\textrm{  d}\alpha(t)}{\textrm{d}t}\textrm{  d}t,
 \end{equation}

  \begin{equation}
     \label{eqn:12}
     \textrm{where    }\alpha(t) = min \left (\int_{t_{min}}^{t} \textrb{V}_\alpha (\textrb{r}_o + s\textrb{r}_d) \textrm{  d}s, 1 \right),
 \end{equation}
 
  \begin{equation}
     \label{eqn:13}
     \textrm{and                   }\textrb{r}_d = \frac{\textrb{P}^{-1}\textrb{p--r}_o}{||\textrb{P}^{-1}\textrb{p--r}_o||} \in \mathbb{R}^3,
 \end{equation}

where $\textrb{I}_\textrm{rgb} (\textrb{p})$ is the color at a pixel location in the camera coordinate system with center $\textrb{r}_o \in \mathbb{R}^3$, ray direction is given by the eqn.~\ref{eqn:13} for the Image Coordination System-to-World Coordinate System.

Color calibration based on per-camera and per-channel bias $b$ and gain $g$ ensures that the reconstructed images is accounted for the slight differences in overall intensity in the image. Further, a background estimation based on the eqn.~\ref{eqn:14}

  \begin{equation}
     \label{eqn:14}
     \widehat{ \textrb{I}}_\textrm{rgb} (\textrb{p}) =  \left( 1 - \textrb{I}_\alpha(\textrb{p}) \right) \textrb{I}_\textrm{rgb}^\textrm{(bg)}(\textrb{p}) + \left( g\textrb{I}_\textrm{rgb}(\textrb{p}) + b \right),
 \end{equation}
 
 where $\textrb{I}_\textrm{rgb}^\textrm{(bg)}$ and $\widehat{ \textrb{I}}_\textrm{rgb}$ are the background image extracted based on static or non-changing background, and final image, respectively. Further, NV introduces two priors in its architecture--total variation of the log voxel opacities, and beta distribution (Beta(0.5, 0.5)) on the final image opacity. These priors are demonstrated by the eqn.~\ref{eqn:15}--\ref{eqn:16}.

 \begin{equation}
     \label{eqn:15}
     \mathcal{P}_\textrm{var\_lvoxop}(\textrb{V}_\alpha) = \frac{1}{N} \sum_{x} \lambda_\textrm{var\_lvoxop}\left|\left|\frac{\partial}{\partial \textrb{x}} \textrm{log \textbf{V}}_\alpha\textrb{(x)} \right|\right| 
 \end{equation}

  \begin{equation}
     \label{eqn:16}
     \textrm{and       }\mathcal{P}_\beta (\textrb{I}_\alpha) = \frac{1}{P} \sum_{\textrb{p}} \lambda_\beta \left[ \textrm{log(\textbf{I}$_\alpha$(\textbf{p})) + log(1 -- \textbf{I}$_\alpha$(\textbf{p}))} \right]
 \end{equation}

where \textbf{p}, $P$, and $\mathcal{P}$ represent a pixel in image, number of pixels, and the priors used. Further, the training objective of NV is based on the equation~\ref{eqn:17} where $\textrb{I}_\textrm{rgb}^{*}$ is the ground truth image and $D_{KL}(\textrb{z}|| N (0,1))$ is the KL divergence between the latent encoding $\textrb{z}$ and Bell Distribution used in a VAE (Variational Autoencoder).

\begin{equation}
 \label{eqn:17}
  \begin{multlined}
     l(\theta) = \frac{1}{P} \sum_{\textrb{p}} \left|\left|   \widehat{ \textrb{I}}_\textrm{rgb}(\textrb{p}) \textbf{--} \textrb{I}_\textrm{rgb}^{*}(\textrb{p}) \right|\right|^{2} \\
     \begin{aligned} 
     & + \lambda_\textrm{KL} D_\textrm{KL} \left(\textrb{z} || \mathcal{N}(0, 1) \right) \\
     & + \mathcal{P}_\textrm{var\_lvoxop}(\textrb{V}_\alpha)\\
     & + \mathcal{P}_\beta (\textrb{I}_\alpha).
     \end{aligned}
\end{multlined}
\end{equation}

Various usecases and results for the same have been discussed in the paper. Further, we refrain from discussing the implementation, training, and hyperparameter details for this network architecture hence the reader is advised to read the paper for more comprehensive intuition about the author's approach.\\

Thies et al proposed \textbf{DNR (Deferred Neural Rendering)}~\cite{thies2019deferred} to leverage traditional graphics components and make those learnable using Neural Networks. The main idea behind this work is to use imperfect 3D content from photometric reconstructions with incomplete or otherwise noisy surface geometry to produce photo-realistic (re-renderings). The contribution here was the introduction of \textit{Neural Textures} that were learned as feature maps from the scene capture process. These are stored similar to texture maps on top of 3D mesh proxies. High Dimensional Feature Maps contain more information (when compared to traditional texture maps) that were encoded in high dimensional feature maps and used deferred neural pipeline for interpretation. In this approach, both the Deferred Neural Renderer and Neural Textures were trained end-to-end which enabled synthesizing photo-realistic images with imperfections in the original 3D content. This approach gave significant explicit control over generated output such as synthesizing temporally-consistent video re-renderings of recorded 3D scenes in 3D space. This allowed Neural Textures to coherently re-render and manipulate existing video content in real-time static and dynamic environments. Further, this approach has been leveraged for several applications such as Novel View Synthesis, Scene Editing, Facial Reenactments. Moreover, this rendering that relied on geometry proxy (needs to be reconstructed in pre-processing step) was much faster than traditional reenactments for high-resolution outputs. A drawback of this approach was the retraining of neural textures for every object fed to the DNR. This pipeline allowed generalization to happen on 10 objects used, which suggested applications in the domain of transfer learning, segmentation, scene illumination, surface reflectance, novel shading, and lighting capabilities with usage of this pipeline. Further, this work had the capability to be extended beyond the traditional 2D to volumetric grids. The Average Precision for the Ground Truth proposal has been given in the eqn~\ref{eqn:17-1}.

\begin{equation}
\label{eqn:17-1}
    AP@n = \frac{1}{GTP} \sum_k^n{P@k \times rel@k}
\end{equation}
\\
 
\begin{figure*}[!htp]
    \centering
    \includegraphics[width=\textwidth]{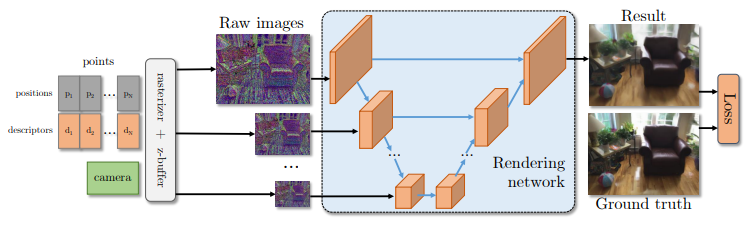}
    \caption{A block diagram representation of NPBGs (Neural Point Based Graphics) by Aliev et al~\cite{aliev2020neural}}
    \label{fig:fig5}
\end{figure*}

 Another Neural Rendering approach called \textbf{NPBG (Neural Point-Based Graphics)} (proposed by Aliev et al~\cite{aliev2020neural}) explores the idea of augmenting each point with a learnable neural descriptor that encodes local geometry and appearance. Authors claim that this approach results in compelling results for scenes scanned with standard RGB cameras and hand-held commodity RGB-D sensors even with the objects that are challenging for standard mesh-based rendering. The idea of NPBG combines the ideas behind IBR, NPG, and Neural Rendering discussed briefly earlier. It uses a Deep CNN (Convolutional Neural Network) to generate photorealistic scenes from novel viewpoints by leveraging Scene Geometry Raw Point-Clouds Representation. Latent Vector (here, Neural Descriptors) learning/estimation facilitates realistic rendering by describing both geometric and photometric properties of scenes in conjunction with learning of the rendering network. The fig.~\ref{fig:fig5} gives an intuition into how NPBG works and how the rendering algorithm is jointly optimized while learning the Neural Descriptors.\\
 This approach formulates the problem of Neural Rendering with the point clouds ($P = \{p_1, p_2, ..., p_N\}$) and M-dimensional Neural Descriptors ($\mathbb{D} = {d_1, d_2, ..., d_N}$) given and the requirement for the renderer to obtain a new view with a camera $\mathcal{C}$ (with extrinsic parameter and intrinsic parameters).
 
The authors suggest to outline the rendering process such that descriptors are used as pseudo-colors to project the point into the target view. Further, the rendering network is used to map the pseudo-colors to photorealistic RGB space. Formally, NPBG creates an $M-$channel raw-image of size $W \times H$, and for each point $p_i$ projecting to $(x, y)$ given by the eqn.~\ref{eqn:18}

\begin{equation}
    \label{eqn:18}
    S (\textrb{P, D,} C) [[x], [y]] = d_\textrm{i},
\end{equation}

where $[a]$ is nearest integer for $a \in \mathbb{R}$. Further, this approach proposes to leverage a z-buffer to remove occluded points as several points can project onto the same pixel but the lack of topological information in point clouds can result in the \textit{bleeding} problem (where occluded surfaces and background points could be seen through from front surface). This approach proposes to leverage multi-scale progressive rendering to avoid relying on choice of disk radius (important for \textit{splatting} which uses a 3D elliptic disk with an estimated radius to replace each point). Here, a sequence of images $S[1], S[2], ..., S[T]$, where the i-th image has the size of $\frac{W}{2^t} \times \frac{H}{2^t}$ is obtained by rendering a point cloud on a canvas pyramid of different resolution $T$ times by performing simple point-cloud projection. Hence, $S[1]$ has the maximum details and is prone to the most surface bleeding whereas $S[T]$ suffers from least surface bleeding but has the minimum geometric details. Every other image in sequence ($S[2], S[3], ..., S[T-1]$) has different detail-bleeding tradeoffs. Now, all the raw sequence of images are mapped to three channel RGB image $I$ using a U-Net~\cite{ronneberger2015u} (with Gated Convolutions~\cite{yu2019free} to handle sparse inputs) based rendering network ($\mathcal{R}_\theta$ as given in the eqn,~\ref{eqn:19} below.

\begin{equation}
    \label{eqn:19}
    I (\textrb{P, D,}C,\theta) = \mathcal{R}_\theta \left( S[1] (\textrb{P, D,}C), ...., S[T] (\textrb{P, D,}C)\right)
\end{equation}

The, the raw images $S[i]$ are concatenated to the U-Net first block with corresponding resolutions (this coarse-to-fine mechanism and mipmapping are similar). Hence, rendering network provides the mechanism for implicit detail selection in this case. Further, OpenGL is used for rasterization of raw image sequences $S[1], S[2], ..., S[T]$. The dimensionality of the descriptors is set to eight ($M=8$).\\
The fitting process in this system assumes that during the fitting operation, there are $K$ different scenes are available. Given point cloud $\textrb{P}^k$ and training ground truth set with $L_k$ RGB images $\textrb{I}^k - \{I^{k,1}, I^{k,2},..., I^{k, L_k}\}$ and known camera parameters $\{C^{k,1}, C^{k,2}, ..., C^{k, L_k}\}$ for $k-$th scene, helps formulate objective function $\mathcal{L}$ (mismatch between the rendered and ground truth RGB image) as given in the eqn.~\ref{eqn:20}.
\begin{equation}
    \label{eqn:20}
    \mathcal{L}(\theta, \textrb{D}^1, \textrb{D}^2, ..., \textrb{D}^K) = \sum_{k=1}^{K} \sum_{l=1}^{L_k} \left(\Delta \left( \mathcal{R_\theta} (\{S[i](\textrb{P}^k, \textrb{D}^k, C^{k,l})\}_{i=1}^{T}) \right), I^{k, l}\right). 
\end{equation}

Here, $\Delta$ represents the mismatch between ground truth and rendered images, $\textrb{D}^k$ are neural descriptors for point clouds of the $k$-th scene and a perceptual loss~\cite{dosovitskiy2016generating,johnson2016perceptual} helps compute the mismatch for activations of a pretrained VGG16 network~\cite{simonyan2014very}. Both the neural descriptors and Rendering network parameters in the training set of the scenes are used for the optimizing the loss in eqn.~\ref{eqn:20} by the ADAM optimizer~\cite{kingma2014adam}. This system can be used for transfer/incremental learning which suggests that the results for new viewpoints tend to be better when rendering network is fitted to multiple scenes of a similar kind when compared to a single scene. Hence, this process is two-stage training process where the network is first pretrain for the rendering network on a scene family, followed by fine-tuning the rendering network to a new scene. In the later stage, the learning process~\ref{eqn:20} starts with no neural  scenes descriptors and pretrained weights of the rendering network.

 

 The inspiration of NeRF architecture can be said to be derived from Mildenahall and Tancik's own work~\cite{mildenhall2019local,tancik2020fourier}, Sitzmann et al's \textbf{SIREN (Implicit Neural Representations with Periodic Activation Functions)}~\cite{sitzmann2020implicit} networks introduced in the same year. SIRENs~\cite{sitzmann2020implicit} introduced a novel approach for constructing implicit representations using periodic activation functions. This technique has shown to provide a more efficient and stable way to represent complex functions with deep neural networks. Whereas, ``Fourier features let networks learn high frequency functions in low dimensional domains''~\cite{tancik2020fourier} proposed a way to augment neural networks with Fourier features to allow them to learn high-frequency functions (real-life image representations) in low-dimensions. This approach enables neural networks to represent complex functions with fewer parameters and to learn more efficiently. Finally, the NeRFs build on the previous ideas to propose a method to represent 3D scenes using a continuous function for the plenoptic function that can be implicitly stored as Neural Representation. This approach allowed for high-quality view synthesis by rendering images from any viewpoint in a scene using much less storage when compared to polygonal meshes or pointclouds.
(Work in Progress for Surveys related to NeRFs...)

\section{Important Architectures, Models, and Extensions}
\label{present:papers}
This section briefly discusses the Radiance Fields architecture and discusses the improvements, extensions, and models for different applications. These research works have been briefly described according to the year they were published on Arxiv. The author wants to keep the source of the research related to the Radiance Field Literature so as to stay consistent in the source of information
\begin{itemize}
\item \textbf{Research in 2020:} This subsection briefly discusses the research articles in 2020. NeRFs were introduced in the March of this year as a preprint rolled out by Mildenhall et al~\cite{mildenhall2020nerf}.
\begin{itemize}
    \item \textbf{Original NeRF}~\cite{mildenhall2020nerf} is the model that uses fully connected deep neural network to represent a continuous 3D scene function, which takes a 3D point and a viewing direction as inputs and outputs a color and density function values.
    \item \textbf{Fourier Features Let Neural Networks Learn High Frequency Functions in Low Dimensional Domains}~\cite{tancik2020fourier} introduces Fourier features, which improve the ability of neural networks to learn high-frequency functions in low-dimensional domains. The authors propose mapping input coordinates to a higher-dimensional space using a random Fourier feature mapping, which significantly improves the network's ability to model high-frequency functions. This technique proves to be particularly useful for tasks such as geometric modeling and texture synthesis.
    \item \textbf{GRAFs (Generative Radiance Fields)}~\cite{schwarz2020graf} extend  NeRFs to the domain of generative modelling. While NeRF focuses on learning a continuous 3D representation of a scene from a set of 2D images with known camera parameters, GRAFs aim to generate novel 3D scenes and synthesize views of those scenes without relying on explicit 3D geometry. GRAFs combine the NeRFs representation with a generative model like VAE (Variational Auto-Encoder)  or a GAN (Generative Adversarial Networks). The generative model learns a latent space that encodes the underlying structure of the 3D scene with the conditional NeRF mapping the points in this latent space to 3D radiance fields. The training objective typically involves a combination of reconstruction and regularization loss. After training, GRAFs can sample latent vectors from the learned latent spaces and synthesize views of the scenes using conditional NeRF model. This makes GRAFs particularly useful for tasks such as view synthesis, 3D scene interpolation, and data augmentation, as they can generate diverse and realistic 3D scenes without relying on explicit 3D geometry.
    \item \textbf{NSVF (Neural Sparse Voxel Fields)}~\cite{liu2020neural} extends NeRF by introducing a sparse voxel field representation, which significantly reduces memory requirements and accelerates rendering. This method combines a sparse voxel octree with a neural network to model scene appearance, enabling efficient reconstruction and rendering of large scale scenes.
    \item \textbf{NeRF-W (NeRF in the Wild)}~\cite{martin2021nerf} is an extension of the original NeRF model to handle more complex, real-world scenes with uncontrolled illumination. NeRF-W introduces additional input features to represent the spatially-varying lighting conditions, improving the model's ability to generalize across diverse scenes.
    \item \textbf{Neural Reflectance Fields}~\cite{bi2020neural} presented a method to capture spatially-varying appearance and fine-geometric details of real-world objects using NeRF. It introduces a multi-scale reflectance field representation and a multi-view photometric stereo method to estimate per-point reflectance and normals, enabling high-fidelity reconstruction of complex objects.
    \item \textbf{Neural Face Reflectance Fields}~\cite{tewari2021monocular} proposed a method to reconstruct neural face reflectance fields from a single monocular input image. This method estimates per-pixel surface normals, albedo, lighting parameters and combines them with NeRF-based representation to synthesize high-quality novel views of the face.
    \item \textbf{GRF (Learning a General Radiance Fields for 3D Representation and Rendering)}~\cite{trevithick2021grf} presents a method for learning a 3D representation of a scene and rendering it using neural radiance fields. The authors introduce a multi-scale representation that allows the network to model both global structures and fine details of a scene. The resulting model can be used for various tasks, such as novel view synthesis, relighting, and geometric reconstruction.
    \item \textbf{NeRF++}~\cite{zhang2020nerf++} model builds upon the original NeRF by incorporating additional features such as spatially-varying reflectance and incorporating auxiliary tasks to improve training. The main goal of NeRF++ is to enhance the quality of the generated images while maintaining robustness and efficiency.
    \item \textbf{Neural Scene Graphs for Dynamic Scenes}~\cite{ost2021neural} focuses on representing and rendering dynamic 3D scenes using neural scene graphs. The authors propose a method that combines scene graph representations with neural radiance fields to enable efficient rendering of complex and dynamic scenes. The method can generate high-quality images of dynamic scenes with realistic lighting and reflections, as well as handle occlusions and disocclusions.
    \item \textbf{GIRAFFE (Compositional Generative Neural Feature Fields)}~\cite{niemeyer2021giraffe} presents a compositional generative model that represents scenes as neural feature fields. The method combines a scene graph representation with a conditional NeRF model to synthesize novel views of complex scenes with varying objects, poses, and appearances, enabling high-quality image synthesis and scene editing.
    \item \textbf{DeRF (Decomposed Radiance Fields)}~\cite{rebain2021derf} introduced a method to separate the global illumination, local shading, and surface color components of a scene, enabling efficient view synthesis and relighting. The method decomposes a neural radiance field into these components and learns a hierarchical representation that captures the spatial and directional dependencies in the scene.
    \item \textbf{NerFies (Deformable Neural Radiance Fields)}~\cite{park2021nerfies} extended NeRF to represent non-rigidly deforming objects, such as human faces and bodies. The method combines a neural radiance field with a parametric deformation model, enabling reconstruction and view synthesis of dynamic scenes from monocular video.
    \item \textbf{Space-Time Neural Irradiance Fields}~\cite{xian2021space} introduced a space-time representation that models the temporal dynamics of free-viewpoint video. The method extends NeRF to represent time-varying appearance and geometry, enabling high-quality view synthesis and interpolation of dynamic scenes.
    \item \textbf{Neural Scene Flow Fields}~\cite{li2021neural} presented a method to represent the scene flow of dynamic objects using NeRF. The method learns a neural scene flow field that models the temporal evolution of object appearance and geometry. By encoding both spatial and temporal information, it enables high-quality view synthesis of dynamic scenes from a limited number of input views.
    \item \textbf{D-NeRF (Neural Radiance Fields for Dynamic Scenes)}~\cite{pumarola2021d} extended NeRF to handle dynamic scenes by learning a temporally coherent neural radiance field from a set of multi-view images with known camera parameters. D-NeRF models the time-varying appearance and geometry by conditioning the neural radiance field on a latent code that varies with time. The method introduces an additional loss term to enforce temporal smoothness in the learned radiance field, allowing for high-quality view synthesis of dynamic scenes with temporal consistency. This approach enables the reconstruction of dynamic scenes from multi-view videos and synthesizes novel views while maintaining the appearance and motion continuity.
    \item \textbf{pixelNeRF}~\cite{yu2021pixelnerf} extends NeRF to generate high-quality novel views from one or few input images, instead of requiring multiple images with known camera parameters. It leverages 2D convolutions to efficiently aggregate information from input images and incorporates this information into the neural radiance field, allowing for improved view synthesis from limited input data.
    \item \textbf{Dynamic Neural Radiance Fields for Monocular 4D Facial Avatar Reconstruction}~\cite{gafni2021dynamic} proposed a method to reconstruct dynamic neural radiance fields for human faces from a single monocular video. The method captures the 4D facial geometry and appearance, allowing for high-quality facial avatar reconstruction and novel view synthesis with temporal consistency.
    \item \textbf{NeRD (Neural Reflectance Decomposition)}~\cite{boss2021nerd} presented a method for learning scene-specific neural reflectance models from image collections. The method involves decomposing input images into albedo, normal, and lighting components, allowing for more accurate and flexible representations of complex scenes. This approach can be used to synthesize novel images, re-light scenes, and perform geometry-aware image editing.
    \item \textbf{NeRV (Neural Reflectance and Visibility Fields)}~\cite{srinivasan2021nerv} introduced a method to represent both reflectance and visibility fields for relighting and view synthesis tasks. The method leverages a novel hierarchical representation that enables efficient and high-quality rendering of complex scenes under novel lighting conditions and viewpoints.
    \item \textbf{iNeRF (Inverting Neural Radiance Fields)}~\cite{yen2021inerf} presented a method to invert neural radiance fields to estimate the pose of an object or scene given a single image. By learning a mapping from 2D images to the latent space of the radiance field, iNeRF enables accurate pose estimation without explicit 3D geometry.
    \item \textbf{Portrait Neural Radiance Fields}~\cite{gao2020portrait} focused on generating portrait neural radiance fields from a single input image. The method leverages a combination of geometric priors, facial landmarks (such as 3DMM features), and a neural radiance field representation to reconstruct 3D facial geometry and appearance, enabling high-quality novel view synthesis. The priors are incorporated from a dataset of 3D facial scans. The resulting method can generate novel views and relight the face, even with limited input data.
    \item \textbf{Object-Centric Neural Scene Rendering}~\cite{guo2020object} worked on an object-centric representation for neural scene rendering. The method represents scenes as a collection of object-centric neural radiance fields and learns to compose them into a global scene representation, enabling high-quality view synthesis and intuitive scene editing.
    \item \textbf{NeRFlow (Neural Radiance Flow)}~\cite{du2021neural} extends NeRF to model the temporal dynamics of video sequences. The method learns a continuous 4D radiance field that represents both the spatial and temporal information in the video, enabling high-quality view synthesis and video processing tasks, such as interpolation and extrapolation.
    \item \textbf{Compositional Radiance Fields}~\cite{wang2021learning} presents a method to learn compositional radiance fields for dynamic human heads. The method combines a parametric human head model with a neural radiance field representation, allowing for high-quality view synthesis and editing of dynamic human head models.
    \item \textbf{Non-Rigid Neural Radiance Fields}~\cite{tretschk2021non} proposes a method to reconstruct and synthesize novel views of non-rigidly deforming scenes from monocular video. The method combines a neural radiance field with a non-rigid deformation model, enabling high-quality reconstruction and view synthesis of dynamic scenes.
    \item \textbf{Neural Body}~\cite{peng2021neural} paper proposes a method for representing and synthesizing novel views of dynamic humans using structured latent codes and implicit neural representations. The authors introduce a method that decomposes the input data into body shape, pose, and appearance components, enabling the network to learn a more accurate and flexible representation of the human body. The method can generate realistic and temporally coherent novel views of dynamic humans, even with sparse input data. Hence, the authors build upon the NeRF framework by introducing structured latent codes to represent dynamic humans. Specifically, they decompose the input data into body shape, pose, and appearance components. This decomposition allows the model to capture the complex and dynamic nature of human bodies more effectively than standard NeRF. (Work in Progress...)
\end{itemize}

\item \textbf{Research in 2021:} This subsection briefly discusses the research articles in the following year 2021. There were 110 research papers related to the field of Neural Radiance Fields (not including survey papers) in 2021.
\begin{itemize}
    \item \textbf{Non-line-of-sight Imaging via Neural Transient Fields}~\cite{shen2021non} presented a method for non-line-of-sight (NLOS) imaging using neural transient fields. The authors proposed a learning-based framework that leverages transient light measurements to reconstruct hidden objects. The key idea was to represent the volumetric distribution of light in a scene using a neural network, which learns the mapping from transient measurements to the 3D geometry and reflectance of the hidden objects. While not a direct extension of NeRF, this work shares similarities with NeRF in representing volumetric information using neural networks.
    \item \textbf{Pixel-Aligned Volumetric Avatars}~\cite{raj2021pixel} introduced a method for creating pixel-aligned volumetric avatars from multi-view video data. It proposed a learning-based approach that combines the benefits of traditional multi-view stereo and neural rendering methods. By using a fully convolutional neural network to predict a volumetric representation, the method generates high-quality avatars with consistent appearance across multiple views. Although this research does not directly extend NeRF, it shares the goal of creating high-quality 3D representations that can be efficiently rendered into 2D images.
    \item \textbf{A-NeRF (Articulated Neural Radiance Fields)}~\cite{su2021nerf} was an extension of NeRF that focused on learning human shape, appearance, and pose. It propose a method that represents humans as an articulated structure, where each body part has its own local neural radiance field. This decomposition allows A-NeRF to capture the complex geometry and appearance of humans more effectively than the original NeRF. The method can reconstruct detailed 3D models of humans and render novel views with realistic appearance and accurate poses.
    \item \textbf{NeRF-- (Neural Radiance Fields Without Known Camera Parameters)}~\cite{wang2021nerf} was an extension of NeRF that relaxes the requirement of having known camera parameters. The authors propose a method that jointly learns the neural radiance field and camera parameters from a collection of uncalibrated input images. By incorporating differentiable camera optimization, the method can estimate both the 3D scene representation and the camera parameters simultaneously. NeRF-- enables the use of NeRF for scenarios where camera parameters are not known or only partially available.
    \item \textbf{ShaRF (Shape-Conditioned Radiance Fields)}~\cite{rematas2021sharf} is an extension of NeRF that focuses on learning 3D scene representations from a single input view. It combines a shape-conditioned radiance field with a shape prior learned from a large dataset of 3D models. The method leverages this shape prior to guide the learning of the radiance field, which enables accurate 3D reconstructions and novel view synthesis even with limited input data. ShaRF demonstrates the potential of using NeRF in single-view scenarios by incorporating additional prior knowledge about the underlying 3D structure.
    \item \textbf{NeuTex}~\cite{xiang2021neutex} is an extension of NeRF that focuses on improving the efficiency of volumetric neural rendering by introducing neural texture mapping. The authors propose a two-stage approach that decouples geometry and appearance: first, a coarse geometry is reconstructed using a neural radiance field, and then, a neural texture map is learned to refine the appearance. The training paradigm involves a combination of a differentiable ray-marching algorithm and texture loss. The method allows for faster rendering and lower memory requirements compared to the original NeRF, while maintaining high-quality results for tasks like novel view synthesis and relighting.
    \item \textbf{Mixture of Volumetric Primitives for Efficient Neural Rendering}~\cite{lombardi2021mixture} introduces an improvement over NeRF by representing scenes as a mixture of volumetric primitives, such as spheres and cylinders, which reduces the complexity of the scene representation. The training paradigm involves optimizing the parameters of these primitives, their positions, and a neural radiance field for each primitive. The loss function is a combination of the rendering loss and a regularization term. The method enables more efficient rendering and storage while maintaining the quality of the generated images, making it suitable for real-time applications and rendering large scenes.
    \item \textbf{DONeRF (Depth Oracle Networks-based Compact Radiance Fields)}~\cite{neff2021donerf} is an extension of NeRF that aims to achieve real-time rendering of compact radiance fields. The authors propose using a depth oracle network to predict the depth of the scene, which reduces the number of samples required for rendering. The training paradigm involves a two-stage approach: first, the depth oracle network is trained using supervised depth data, and then, the neural radiance field is trained using the predicted depth. The loss function combines a rendering loss and a depth loss. DONeRF significantly accelerates the rendering process while maintaining high-quality results, making it suitable for real-time applications.
    \item \textbf{NeX}~\cite{wizadwongsa2021nex} is an improvement over NeRF that focuses on real-time view synthesis by using a neural basis expansion approach. The authors propose representing the scene as a linear combination of basis functions, which are learned using a neural network. The training paradigm involves optimizing the neural network to predict the basis coefficients and the radiance field. The loss function is a combination of the rendering loss and a regularization term. NeX enables real-time view synthesis with high-quality results and lower memory requirements compared to NeRF, making it suitable for real-time applications and virtual reality.
    \item \textbf{FastNeRF}~\cite{garbin2021fastnerf} is an improvement over NeRF that aims to achieve high-fidelity neural rendering at high frame rates. The authors propose a multi-scale hierarchical approach that accelerates rendering by leveraging a coarse-to-fine strategy. The training paradigm involves training multiple neural radiance fields at different levels of detail, and the loss function combines the rendering loss and a multi-scale consistency loss. FastNeRF significantly accelerates the rendering process while maintaining high-quality results, making it suitable for applications that require high frame rates, such as video games and virtual reality.
    \item \textbf{Neural Lumigraph Rendering}~\cite{kellnhofer2021neural} is an extension of NeRF that combines the benefits of traditional Lumigraph rendering and neural radiance fields. The authors propose a method that leverages a sparse set of input views to learn a multi-plane representation of the scene. This multi-plane representation is then used as input to a neural rendering network, which generates high-quality novel views. The training paradigm involves optimizing the neural rendering network using a rendering loss, which measures the difference between the ground truth and the predicted images. This method offers a more efficient and flexible rendering solution than NeRF and can be used for applications like view synthesis and relighting.
    \item \textbf{iMAP (Implicit Mapping and Positioning)}~\cite{sucar2021imap} is an application of NeRF that focuses on real-time simultaneous mapping and positioning in 3D environments. The authors propose a method that learns an implicit 3D map of the environment using a neural radiance field and estimates the camera pose simultaneously. The training paradigm involves optimizing the neural network to minimize the re-projection error and a depth consistency loss. iMAP demonstrates the potential of using NeRF for real-time 3D mapping and positioning tasks, such as robot navigation and augmented reality.
    \item \textbf{Mip-NeRF}~\cite{barron2021mip} is an improvement over NeRF that introduces a multiscale representation to reduce aliasing artifacts in the rendered images. The authors propose a mipmapping approach that learns a hierarchy of neural radiance fields at different levels of detail. The training paradigm involves optimizing the neural networks at each level of detail using a rendering loss and a consistency loss between adjacent levels. Mip-NeRF achieves higher-quality rendering with fewer aliasing artifacts, making it suitable for applications that require high-quality image synthesis, such as virtual reality and film production.
    \item \textbf{Kilo-NeRF}~\cite{reiser2021kilonerf} is an improvement over NeRF that aims to speed up rendering by using thousands of tiny multilayer perceptrons (MLPs) instead of a single large MLP. The authors propose dividing the scene into small regions and training a separate MLP for each region, which allows for more efficient rendering and parallelization. The training paradigm involves optimizing the MLPs using a rendering loss that measures the difference between the ground truth and the predicted images. Kilo-NeRF significantly accelerates the rendering process while maintaining high-quality results, making it suitable for applications that require real-time rendering, such as video games and virtual reality.
    \item \textbf{PlenOctrees}~\cite{yu2021plenoctrees} is an improvement over NeRF that focuses on real-time rendering by introducing a hierarchical data structure called PlenOctrees. The authors propose representing the scene using an octree, where each node stores a neural radiance field that models the local geometry and appearance. The training paradigm involves optimizing the neural radiance fields using a rendering loss and a multi-scale consistency loss. PlenOctrees enables real-time rendering of neural radiance fields with high-quality results and lower memory requirements compared to NeRF, making it suitable for real-time applications and virtual reality.
    \item \textbf{Baking Neural Radiance Fields}~\cite{hedman2021baking} is an improvement over NeRF that focuses on real-time view synthesis by converting learned neural radiance fields into a more efficient representation for rendering. The authors propose a method that bakes the neural radiance field into a set of layered depth images (LDIs) and then uses these LDIs for real-time rendering. The training paradigm involves optimizing the neural radiance field using a rendering loss that measures the difference between the ground truth and the predicted images. This method significantly accelerates the rendering process while maintaining high-quality results, making it suitable for real-time applications such as video games and virtual reality.
    \item \textbf{MINE (Multiview Image-based Neural Extensions)}~\cite{li2021mine} is an extension of NeRF that combines the benefits of multiplane images (MPIs) and NeRF for novel view synthesis. The authors propose a method that learns a continuous MPI representation using a neural network, which can be used for efficient rendering of novel views. The training paradigm involves optimizing the neural network using a rendering loss and a depth consistency loss. MINE demonstrates the potential of combining the strengths of MPIs and NeRF to achieve high-quality view synthesis with lower computational requirements, making it suitable for real-time applications and virtual reality.
    \item \textbf{MVSNeRF (Multi-view Stereo Neural Radiance Fields)}~\cite{chen2021mvsnerf} is designed to leverage the strengths of both multi-view stereo (MVS) and NeRF techniques. MVSNeRF combines a coarse MVS reconstruction with the NeRF framework to generate high-quality novel views. This hybrid approach takes advantage of the geometric information provided by MVS to improve the efficiency and quality.
    \item \textbf{GNeRF (Generative Adversarial Network-based Neural Radiance Field)}~\cite{meng2021gnerf} is an extension of NeRF that introduces a GAN-based framework for learning neural radiance fields without known camera poses. The authors propose a method that jointly learns the neural radiance field and camera poses using a combination of a rendering loss, an adversarial loss, and a cycle consistency loss. The training paradigm involves optimizing the neural radiance field, camera poses, and the discriminator network. GNeRF demonstrates the potential of using GANs to learn high-quality neural radiance fields in scenarios where camera poses are unknown, making it suitable for applications like novel view synthesis and 3D reconstruction.
    \item \textbf{In-Place Scene Labelling and Understanding with Implicit Scene Representation}~\cite{zhi2021place} presents an application of NeRF that focuses on in-place scene labeling and understanding using implicit scene representations. The authors propose a method that learns a joint representation of geometry, appearance, and semantic information using a neural radiance field. The training paradigm involves optimizing the neural network using a combination of rendering loss, depth consistency loss, and a semantic loss. This method demonstrates the potential of using NeRF for scene understanding tasks, such as semantic segmentation and object recognition, in addition to view synthesis.
    \item \textbf{CAMPARI (Camera-Aware Multiplane Radiance Fields)}~\cite{niemeyer2021campari} is an improvement over NeRF that introduces a camera-aware decomposed generative framework for learning neural radiance fields. The authors propose a method that decomposes the input images into depth layers and learns a separate neural radiance field for each layer. The training paradigm involves optimizing the neural networks using a combination of rendering loss, depth consistency loss, and a cycle consistency loss. CAMPARI enables more efficient rendering and better handling of occlusions, making it suitable for applications like novel view synthesis and relighting.
    \item \textbf{NeRF-VAE (Neural Radiance Fields - Variational Autoencoder)}~\cite{kosiorek2021nerf} is an extension of NeRF that combines the geometry-aware representation of NeRF with the generative capabilities of VAEs. The authors propose a method that learns a latent space for 3D scenes, which can be used for generating novel scenes and interpolating between existing ones. The training paradigm involves optimizing the neural network using a combination of rendering loss, KL-divergence loss, and a reconstruction loss. NeRF-VAE demonstrates the potential of using NeRF for generative modeling tasks, such as 3D scene generation and interpolation.
    \item \textbf{Unconstrained Scene Generation with Locally Conditioned Radiance Fields}~\cite{devries2021unconstrained} presents an extension of NeRF that focuses on unconstrained scene generation using locally conditioned radiance fields. The authors propose a method that conditions the neural radiance field on local spatial information, allowing it to generate complex and diverse scenes. The training paradigm involves optimizing the neural network using a combination of rendering loss and a perceptual loss. This method demonstrates the potential of using NeRF for generative modeling tasks, such as scene synthesis and object generation, in an unconstrained setting.
    \item \textbf{Putting NeRF on a Diet}~\cite{jain2021putting} presents an improvement over NeRF that focuses on semantically consistent few-shot view synthesis. The authors propose a method that leverages semantic information to guide the neural radiance field learning process, enabling it to synthesize novel views with as few as one input image. The training paradigm involves optimizing the neural network using a combination of rendering loss, semantic consistency loss, and a cycle consistency loss. This method demonstrates the potential of using NeRF for few-shot view synthesis tasks while maintaining semantic consistency, making it suitable for applications like virtual reality and film production.
    \item \textbf{Decomposing 3D Scenes into Objects via Unsupervised Volume Segmentation}~\cite{stelzner2021decomposing} presents an application of NeRF that focuses on decomposing 3D scenes into objects using unsupervised volume segmentation. The authors propose a method that learns to segment the scene into different objects by leveraging the geometric information provided by the neural radiance field. The training paradigm involves optimizing the neural network using a combination of rendering loss and a segmentation loss. This method demonstrates the potential of using NeRF for unsupervised object segmentation tasks, which can be useful for applications like 3D object recognition and scene understanding.
    \item \textbf{Convolutional Neural Opacity Radiance Fields}~\cite{luo2021convolutional} is an improvement over NeRF that introduces a convolutional architecture for learning neural radiance fields. The authors propose a method that replaces the fully connected layers in NeRF with convolutional layers, allowing the model to leverage local spatial information and achieve better performance. The training paradigm involves optimizing the neural network using a rendering loss that measures the difference between the ground truth and the predicted images. This method demonstrates the potential of using convolutional architectures in NeRF, leading to more efficient and accurate rendering, making it suitable for applications like novel view synthesis and relighting.
    \item \textbf{NARF (Neural Articulated Radiance Fields)}~\cite{noguchi2021neural} is an extension of NeRF that focuses on modeling articulated objects, such as humans or animals, by leveraging the articulated structure of the object. The authors propose a method that combines NeRF with a kinematic model to represent the geometry and appearance of articulated objects in a coherent and continuous manner. The training paradigm involves optimizing the neural network using a combination of rendering loss and a rigidity loss that encourages a consistent geometry across poses. NARF demonstrates the potential of using NeRF for modeling articulated objects, making it suitable for applications like character animation and motion capture.
    \item \textbf{Neural RGB-D Scene Reconstruction}~\cite{azinovic2022neural} is an application of NeRF that focuses on 3D scene reconstruction from RGB-D (color and depth) images. The authors propose a method that leverages the depth information available in RGB-D images to learn a more accurate and efficient neural radiance field. The training paradigm involves optimizing the neural network using a rendering loss that measures the difference between the ground truth and the predicted images. This method demonstrates the potential of using NeRF for 3D scene reconstruction tasks with RGB-D images, making it suitable for applications like robotics and augmented reality.
    \item \textbf{BARF (Bundle-Adjusting Neural Radiance Fields)}~\cite{lin2021barf} is an improvement over NeRF that introduces bundle adjustment, a technique commonly used in structure-from-motion, for refining camera poses and optimizing the neural radiance field simultaneously. The authors propose a method that jointly optimizes the camera poses and the neural radiance field, leading to more accurate and consistent scene reconstruction. The training paradigm involves optimizing the neural network and camera poses using a combination of rendering loss and a bundle adjustment loss. BARF demonstrates the potential of incorporating bundle adjustment in NeRF, making it suitable for applications like 3D reconstruction and novel view synthesis.
    \item \textbf{SRF (Stereo Radiance Fields)}~\cite{chibane2021stereo} is an extension of NeRF that focuses on learning view synthesis for sparse views of novel scenes. The authors propose a method that leverages stereo pairs of input images to learn a more accurate and efficient neural radiance field. The training paradigm involves optimizing the neural network using a combination of rendering loss, depth consistency loss, and a stereo consistency loss that encourages consistency between the stereo pairs. SRF demonstrates the potential of using NeRF for view synthesis tasks with sparse input views, making it suitable for applications like virtual reality and film production.
    \item \textbf{FiG-NeRF (Figure-Grounded Neural Radiance Fields)}~\cite{xie2021fig} is an extension of NeRF that focuses on 3D object category modeling by learning a figure-ground decomposition of the scene. The authors propose a method that learns separate neural radiance fields for the object and the background, enabling the model to focus on the object's geometry and appearance. The training paradigm involves optimizing the neural network using a combination of rendering loss and a figure-ground consistency loss that encourages a consistent decomposition of the scene. FiG-NeRF demonstrates the potential of using NeRF for 3D object category modeling tasks, making it suitable for applications like object recognition and scene understanding.
    \item \textbf{Shadow Neural Radiance Fields}~\cite{derksen2021shadow} is an application of NeRF that focuses on multi-view satellite photogrammetry, specifically on reconstructing 3D geometry and appearance of Earth's surface from satellite images. The authors propose a method that extends NeRF to account for the complex illumination effects caused by shadows and atmospheric scattering in satellite images. The training paradigm involves optimizing the neural network using a combination of rendering loss, depth consistency loss, and a shadow consistency loss. This method demonstrates the potential of using NeRF for satellite photogrammetry tasks, making it suitable for applications like Earth observation and remote sensing.
    \item \textbf{UNISURF (Unifying Neural Implicit Surfaces and Radiance Fields)}~\cite{oechsle2021unisurf} is an improvement over NeRF that aims to unify neural implicit surface representations and neural radiance fields for multi-view reconstruction. The authors propose a method that combines the advantages of both representations, enabling the model to learn geometry and appearance more accurately and efficiently. The training paradigm involves optimizing the neural network using a combination of rendering loss, depth consistency loss, and a surface consistency loss. UNISURF demonstrates the potential of combining neural implicit surfaces and radiance fields for multi-view reconstruction tasks, making it suitable for applications like 3D reconstruction and novel view synthesis.
    \item \textbf{Editable Free-Viewpoint Video using a Layered Neural Representation}~\cite{zhang2021editable} presents an application of NeRF that focuses on editable free-viewpoint video using a layered neural representation. The authors propose a method that decomposes the scene into a set of layered neural radiance fields, enabling users to edit the content of the video by modifying individual layers. The training paradigm involves optimizing the neural network using a combination of rendering loss, depth consistency loss, and a layer consistency loss. This method demonstrates the potential of using NeRF for editable free-viewpoint video tasks, making it suitable for applications like video editing and content creation.
    \item \textbf{Animatable Neural Radiance Fields}~\cite{peng2021animatable} is an extension of NeRF that focuses on modeling dynamic human bodies by incorporating a parametric human body model into the neural radiance field framework. The authors propose a method that learns a neural radiance field conditioned on the pose and shape parameters of the human body model, enabling the generation of realistic animations. The training paradigm involves optimizing the neural network using a combination of rendering loss, pose consistency loss, and a shape consistency loss. This method demonstrates the potential of using NeRF for animating human bodies, making it suitable for applications like character animation and motion capture.
    \item \textbf{Editing Conditional Radiance Fields}~\cite{liu2021editing} presents an extension of NeRF that focuses on editing conditional radiance fields, allowing users to edit the appearance and geometry of a scene. The authors propose a method that conditions the neural radiance field on an editing vector, enabling the model to generate edited versions of the scene by modifying the vector. The training paradigm involves optimizing the neural network using a combination of rendering loss, depth consistency loss, and an editing consistency loss. This method demonstrates the potential of using NeRF for editing tasks, making it suitable for applications like content creation, virtual reality, and film production.
    \item \textbf{Dynamic View Synthesis from Dynamic Monocular Videos}~\cite{gao2021dynamic} is an application of NeRF that focuses on dynamic view synthesis from dynamic monocular videos. The authors propose a method that extends NeRF to handle dynamic scenes by conditioning the neural radiance field on time, allowing the model to learn a time-varying geometry and appearance. The training paradigm involves optimizing the neural network using a combination of rendering loss, temporal consistency loss, and a geometric consistency loss. This method demonstrates the potential of using NeRF for dynamic view synthesis tasks, making it suitable for applications like video editing, virtual reality, and film production.
    \item \textbf{Recursive-NeRF}~\cite{yang2021recursive} is an improvement over NeRF that aims to make the model more efficient by using a dynamically growing neural radiance field. The authors propose a method that recursively refines the radiance field by training smaller MLPs (Multilayer Perceptrons) on top of the existing ones, allowing the model to adapt to new views with minimal additional computation. The training paradigm involves optimizing the neural network using a rendering loss that measures the difference between the ground truth and the predicted images. Recursive-NeRF demonstrates the potential of using a dynamically growing NeRF for efficient view synthesis, making it suitable for real-time applications.
    \item \textbf{Stylizing 3D Scene via Implicit Representation and HyperNetwork}~\cite{chiang2022stylizing} presents an extension of NeRF that focuses on stylizing 3D scenes using implicit representation and hypernetworks. The authors propose a method that combines NeRF with a hypernetwork, enabling the model to generate stylized versions of the scene by learning a style-specific radiance field. The training paradigm involves optimizing the neural network using a combination of rendering loss, style consistency loss, and a content consistency loss. This method demonstrates the potential of using NeRF for stylizing 3D scenes, making it suitable for applications like content creation, virtual reality, and film production.
    \item \textbf{NeRFactor}~\cite{zhang2021nerfactor} is an improvement over NeRF that focuses on factorizing shape and reflectance under an unknown illumination. The authors propose a method that extends NeRF to learn a separate neural representation for geometry, reflectance, and illumination, allowing the model to disentangle these factors and reconstruct the scene more accurately. The training paradigm involves optimizing the neural network using a combination of rendering loss, reflectance consistency loss, and a geometry consistency loss. NeRFactor demonstrates the potential of using NeRF for disentangling shape, reflectance, and illumination, making it suitable for applications like 3D reconstruction and relighting.
    \item \textbf{Neural Actor}~\cite{liu2021neural} is an extension of NeRF that focuses on neural free-view synthesis of human actors with pose control. The authors propose a method that conditions the neural radiance field on the pose of the actor, allowing the model to generate novel views of the actor in different poses. The training paradigm involves optimizing the neural network using a combination of rendering loss, pose consistency loss, and a geometric consistency loss. Neural Actor demonstrates the potential of using NeRF for synthesizing human actors with pose control, making it suitable for applications like character animation, virtual reality, and film production.
    \item \textbf{Light Field Networks}~\cite{sitzmann2021light} is an improvement over NeRF that aims to speed up rendering by using a single network evaluation. The authors propose a method that learns a global light field representation, which can be sampled to generate novel views without querying the network for each ray. The training paradigm involves optimizing the neural network using a rendering loss that measures the difference between the ground truth and the predicted images. Light Field Networks demonstrates the potential of using a global light field representation for efficient rendering, making it suitable for real-time applications.
    \item \textbf{NeRF in detail (Learning to Sample for View Synthesis)}~\cite{arandjelovic2021nerf} is an improvement over NeRF that focuses on learning to sample rays more effectively for view synthesis. The authors propose a method that combines NeRF with a sampling network, which learns to predict importance sampling weights along each ray to improve the efficiency of rendering. The training paradigm involves optimizing the neural network using a rendering loss that measures the difference between the ground truth and the predicted images. NeRF in detail demonstrates the potential of using importance sampling to enhance the efficiency of NeRF, making it suitable for real-time view synthesis tasks.
    \item \textbf{NeuS (Neural Surfaces)}~\cite{wang2021neus} is an improvement over NeRF that aims to learn neural implicit surfaces by volume rendering for multi-view reconstruction. The authors propose a method that combines volume rendering with implicit surface representations, enabling the model to learn geometry and appearance more accurately and efficiently. The training paradigm involves optimizing the neural network using a combination of rendering loss, depth consistency loss, and surface consistency loss. NeuS demonstrates the potential of using volume rendering with implicit surface representations for multi-view reconstruction tasks, making it suitable for applications like 3D reconstruction and novel view synthesis.
    \item \textbf{Volume Rendering of Neural Implicit Surfaces}~\cite{yariv2021volume} presents an extension of NeRF that focuses on volume rendering of neural implicit surfaces. The authors propose a method that combines NeRF with a voxel-based representation, enabling the model to learn a more efficient and accurate representation of the scene. The training paradigm involves optimizing the neural network using a rendering loss that measures the difference between the ground truth and the predicted images. Volume Rendering of Neural Implicit Surfaces demonstrates the potential of using voxel-based representations with NeRF, making it suitable for applications like 3D reconstruction and novel view synthesis.
    \item \textbf{HyperNeRF}~\cite{park2021hypernerf} is an improvement over NeRF that focuses on learning higher-dimensional representations for topologically varying neural radiance fields. The authors propose a method that extends NeRF to handle higher-dimensional input spaces, allowing the model to represent more complex and topologically varying scenes. The training paradigm involves optimizing the neural network using a rendering loss that measures the difference between the ground truth and the predicted images. HyperNeRF demonstrates the potential of using higher-dimensional representations to enhance the expressiveness of NeRF, making it suitable for modeling complex scenes and objects.
    \item \textbf{Animatable NeRFs from Monocular RGB Videos}~\cite{chen2021animatable} presents an application of NeRF that focuses on animatable neural radiance fields from monocular RGB videos. The authors propose a method that extends NeRF to learn a time-varying radiance field from monocular RGB videos, allowing the model to generate novel views of dynamic scenes. The training paradigm involves optimizing the neural network using a combination of rendering loss, temporal consistency loss, and a geometric consistency loss. This method demonstrates the potential of using NeRF for animating dynamic scenes, making it suitable for applications like video editing, virtual reality, and film production.
    \item \textbf{Fast Training of Neural Lumigraph Representation using Meta Learning}~\cite{bergman2021fast} presents an improvement over NeRF that focuses on fast training of neural lumigraph representation using meta-learning. The authors propose a method that combines NeRF with meta-learning techniques, enabling the model to learn a neural lumigraph representation more quickly. The training paradigm involves optimizing the neural network using a rendering loss that measures the difference between the ground truth and the predicted images, along with a meta-learning loss that measures the generalization performance of the model. This method demonstrates the potential of using meta-learning to speed up the training of NeRF, making it suitable for real-time applications.
    \item \textbf{Depth-Supervised NeRF}~\cite{deng2022depth} is an improvement over NeRF that focuses on using fewer views and faster training without sacrificing quality. The authors propose a method that incorporates depth supervision into the NeRF training process, allowing the model to learn more accurate and efficient representations with fewer input views. The training paradigm involves optimizing the neural network using a combination of rendering loss and depth consistency loss. Depth-Supervised NeRF demonstrates the potential of using depth supervision to enhance the efficiency of NeRF, making it suitable for applications with limited input views.
    \item \textbf{3D Neural Scene Representation for VisuoMotor Control}~\cite{li20223d} presents an application of NeRF that focuses on 3D neural scene representation for visuomotor control. The authors propose a method that extends NeRF to learn a compact and efficient 3D scene representation, which can be used for visuomotor control tasks like robotic manipulation and navigation. The training paradigm involves optimizing the neural network using a combination of rendering loss and a task-specific loss that measures the performance of the visuomotor control policy. This method demonstrates the potential of using NeRF for visuomotor control applications, making it suitable for robotics and autonomous systems.
    \item \textbf{Unsupervised Discovery of Object Radiance Fields}~\cite{yu2021unsupervised} presents an extension of NeRF that focuses on unsupervised discovery of object radiance fields. The authors propose a method that learns object-centric radiance fields without any supervision, allowing the model to discover and represent objects in a scene in an unsupervised manner. The training paradigm involves optimizing the neural network using a combination of rendering loss, object consistency loss, and a scene consistency loss. Unsupervised Discovery of Object Radiance Fields demonstrates the potential of using NeRF for unsupervised object discovery and representation, making it suitable for applications like 3D reconstruction and scene understanding.
    \item \textbf{Neural Rays for Occlusion-aware Image-based Rendering}~\cite{liu2022neural} is an improvement over NeRF that focuses on occlusion-aware image-based rendering. The authors propose a method called Neural Rays, which extends NeRF by explicitly modeling occlusion in the radiance field. This enables the model to generate more accurate and realistic novel views, especially in scenes with complex occlusions. The training paradigm involves optimizing the neural network using a combination of rendering loss and an occlusion-aware loss. Neural Rays demonstrate the potential of incorporating occlusion information into NeRF, making it suitable for applications like image-based rendering and 3D reconstruction.
    \item \textbf{Differentiable Surface Rendering via Non-Differentiable Sampling}~\cite{cole2021differentiable} presents an improvement over NeRF that focuses on differentiable surface rendering using non-differentiable sampling. The authors propose a method that combines NeRF with a differentiable rendering framework that can handle non-differentiable sampling operations, enabling the model to learn more accurate and efficient representations of surfaces. The training paradigm involves optimizing the neural network using a combination of rendering loss and a surface consistency loss. Differentiable Surface Rendering via Non-Differentiable Sampling demonstrates the potential of using differentiable rendering with NeRF, making it suitable for applications like 3D reconstruction and novel view synthesis.
    \item \textbf{FLAME-in-NeRF}~\cite{athar2021flame} is an application of NeRF that focuses on free-view face animation. The authors propose a method that combines NeRF with the FLAME model, a parametric face model that encodes facial shape, expression, and pose information. This enables the generation of novel views of animated faces with high-quality and realistic appearance. The training paradigm involves optimizing the neural network using a rendering loss that measures the difference between the ground truth and the predicted images. FLAME-in-NeRF demonstrates the potential of using NeRF for face animation, making it suitable for applications like virtual reality, video games, and film production.
    \item \textbf{iButter}~\cite{wang2021ibutter} is an application of NeRF that focuses on generating interactive bullet time effects for human free-viewpoint rendering. The authors propose a method that extends NeRF to handle dynamic human scenes, enabling the model to generate novel views of humans in motion with a bullet time effect. The training paradigm involves optimizing the neural network using a combination of rendering loss and a temporal consistency loss. iButter demonstrates the potential of using NeRF for interactive bullet time generation, making it suitable for applications like video editing, virtual reality, and film production.
    \item \textbf{Self-Calibrating Neural Radiance Fields}~\cite{jeong2021self} presents an improvement over NeRF that focuses on self-calibrating neural radiance fields. The authors propose a method that extends NeRF by learning to self-calibrate camera parameters, enabling the model to handle input data with noisy or incomplete camera parameters. The training paradigm involves optimizing the neural network using a combination of rendering loss and a calibration consistency loss. Self-Calibrating Neural Radiance Fields demonstrate the potential of incorporating self-calibration into NeRF, making it suitable for applications like 3D reconstruction and novel view synthesis, especially when working with imperfect input data.
    \item \textbf{NeRFingMVS}~\cite{wei2021nerfingmvs} presents an improvement over NeRF that focuses on indoor multi-view stereo. The authors propose a method called NeRFing MVS, which extends NeRF by incorporating guidance from a multi-view stereo reconstruction to optimize the neural radiance fields for indoor scenes. The training paradigm involves optimizing the neural network using a combination of rendering loss and a guidance loss from the multi-view stereo reconstruction. NeRFing MVS demonstrates the potential of incorporating multi-view stereo guidance into NeRF, making it suitable for applications like indoor scene reconstruction and novel view synthesis.
    \item \textbf{CodeNeRF}~\cite{jang2021codenerf} is an extension of NeRF that focuses on disentangling neural radiance fields for object categories. The authors propose a method that learns disentangled representations of shape and appearance in neural radiance fields, enabling the model to generate novel views of objects from different categories with consistent appearance properties. The training paradigm involves optimizing the neural network using a rendering loss that measures the difference between the ground truth and the predicted images, along with a disentanglement loss that encourages separation of shape and appearance features. CodeNeRF demonstrates the potential of using disentangled representations in NeRF, making it suitable for applications like object synthesis and 3D modeling.
    \item \textbf{Learning Object-Compositional Neural Radiance Fields for Editable Scene Rendering}~\cite{yang2021learning} presents an extension of NeRF that focuses on learning object-compositional neural radiance fields for editable scene rendering. The authors propose a method that learns neural radiance fields for individual objects and composes them into a scene, enabling the model to generate novel views of editable scenes. The training paradigm involves optimizing the neural network using a combination of rendering loss, object consistency loss, and a scene consistency loss. This method demonstrates the potential of using object-compositional radiance fields in NeRF, making it suitable for applications like scene editing and virtual reality.
    \item \textbf{Stochastic Neural Radiance Fields}~\cite{shen2021stochastic} is an extension of NeRF that focuses on quantifying uncertainty in implicit 3D representations. The authors propose a method that incorporates stochasticity into the neural radiance field, enabling the model to represent and quantify uncertainty in its predictions. The training paradigm involves optimizing the neural network using a combination of rendering loss and a stochastic consistency loss that encourages the model to capture uncertainty. Stochastic Neural Radiance Fields demonstrate the potential of incorporating uncertainty quantification into NeRF, making it suitable for applications like robust 3D reconstruction and probabilistic rendering.
    \item \textbf{Neural Human Performer}~\cite{kwon2021neural} is an application of NeRF that focuses on learning generalizable radiance fields for human performance rendering. The authors propose a method that extends NeRF to handle dynamic human performances, enabling the model to generate novel views of humans in motion with high-quality and realistic appearance. The training paradigm involves optimizing the neural network using a combination of rendering loss and a temporal consistency loss. Neural Human Performer demonstrates the potential of using NeRF for human performance rendering, making it suitable for applications like virtual reality, video games, and film production.
    \item \textbf{T{\"o}RF (Time-of-Flight Radiance Fields)}~\cite{attal2021torf} is an extension of NeRF that incorporates Time-of-Flight (ToF) information to enable view synthesis for dynamic scenes. The authors propose a method that combines neural radiance fields with ToF depth measurements, making it possible to generate novel views of dynamic scenes with moving objects. The training paradigm involves optimizing the neural network using a combination of rendering loss and a ToF depth consistency loss. TöRF demonstrates the potential of using ToF information in NeRF, making it suitable for applications like robotics, autonomous vehicles, and augmented reality.
    \item \textbf{Vision-Only Robot Navigation in a Neural Radiance World}~\cite{adamkiewicz2022vision} presents an application of NeRF for vision-only robot navigation. The authors propose a method that leverages neural radiance fields to build a 3D representation of the environment, allowing a robot to navigate using only visual information. The training paradigm involves optimizing the neural network using a combination of rendering loss and a navigation loss that encourages the model to learn a useful representation for navigation. This approach demonstrates the potential of using NeRF for vision-only robot navigation, making it suitable for applications like autonomous robots and drone navigation.
    \item \textbf{Neural Radiance Fields Approach to Deep Multi-View  Photometric Stereo}~\cite{kaya2022neural} presents an extension of NeRF that focuses on deep multi-view photometric stereo. The authors propose a method that combines neural radiance fields with multi-view photometric stereo to reconstruct the shape and appearance of objects in a scene. The training paradigm involves optimizing the neural network using a combination of rendering loss and a photometric consistency loss. This method demonstrates the potential of using NeRF for photometric stereo, making it suitable for applications like 3D reconstruction and virtual reality.
    \item \textbf{LENS (Localization Enhanced by NeRF Synthesis)}~\cite{moreau2022lens} is an application of NeRF that focuses on improving localization tasks using synthesized views. The authors propose a method that leverages NeRF to generate synthetic views that help in refining the localization estimates. The training paradigm involves optimizing the neural network using a combination of rendering loss and a localization loss that encourages the model to generate useful views for localization. LENS demonstrates the potential of using NeRF for enhancing localization tasks, making it suitable for applications like robotics, autonomous vehicles, and augmented reality.
    \item \textbf{NeRS (Neural Reflectance Surfaces)}~\cite{zhang2021ners} is an extension of NeRF that focuses on sparse-view 3D reconstruction in the wild. The authors propose a method that combines neural radiance fields with an additional reflectance component, enabling the model to handle challenging real-world scenes with sparse views. The training paradigm involves optimizing the neural network using a combination of rendering loss and a reflectance consistency loss. NeRS demonstrates the potential of using NeRF for sparse-view 3D reconstruction, making it suitable for applications like 3D modeling and virtual reality.
    \item \textbf{StyleNeRF}~\cite{gu2021stylenerf} is an extension of NeRF that incorporates style-based generators for high-resolution image synthesis. The authors propose a method that combines neural radiance fields with a style-based generator, enabling high-quality 3D-aware image synthesis. The training paradigm involves optimizing the neural network using a combination of rendering loss and a style consistency loss. StyleNeRF demonstrates the potential of using NeRF for high-resolution image synthesis, making it suitable for applications like virtual reality, 3D modeling, and artistic image generation.
    \item \textbf{CIPS-3D}~\cite{zhou2021cips} is an extension of NeRF that focuses on generating images using conditionally-independent pixel synthesis. The authors propose a method that combines neural radiance fields with a GAN-based generator, enabling 3D-aware image synthesis. The training paradigm involves optimizing the neural network using a combination of rendering loss and a GAN loss. CIPS-3D demonstrates the potential of using NeRF for 3D-aware image generation, making it suitable for applications like 3D modeling, virtual reality, and artistic image generation.
    \item \textbf{H-NeRF}~\cite{xu2021h} is an extension of NeRF that focuses on rendering and temporal reconstruction of humans in motion. The authors propose a method that combines neural radiance fields with temporal information, enabling the model to reconstruct and render dynamic human motions. The training paradigm involves optimizing the neural network using a combination of rendering loss and a temporal consistency loss. H-NeRF demonstrates the potential of using NeRF for human motion reconstruction and rendering, making it suitable for applications like animation, virtual reality, and video game development.
    \item \textbf{Dex-NeRF}~\cite{ichnowski2021dex} is an application of NeRF that focuses on grasping transparent objects. The authors propose a method that leverages NeRF to generate a 3D representation of transparent objects, allowing a robot to grasp them using visual information. The training paradigm involves optimizing the neural network using a combination of rendering loss and a grasping loss that encourages the model to learn a useful representation for grasping. Dex-NeRF demonstrates the potential of using NeRF for grasping transparent objects, making it suitable for applications like robotics, automation, and manufacturing.
    \item \textbf{Neural-PIL}~\cite{boss2021neural} is an extension of NeRF that focuses on reflectance decomposition using pre-integrated lighting. The authors propose a method that combines neural radiance fields with a pre-integrated lighting model, enabling the decomposition of reflectance properties of the scene. The training paradigm involves optimizing the neural network using a combination of rendering loss and a reflectance decomposition loss. Neural-PIL demonstrates the potential of using NeRF for reflectance decomposition, making it suitable for applications like 3D modeling, virtual reality, and computer graphics.
    \item \textbf{Template NeRF}~\cite{guo2021template} is an extension of NeRF that models dense shape correspondences from category-specific object images. The authors propose a method that leverages a template neural radiance field to establish dense shape correspondences across different instances of an object category. The training paradigm involves optimizing the neural network using a combination of rendering loss and a correspondence loss. Template NeRF is suitable for applications like 3D modeling, computer graphics, and object recognition.
    \item \textbf{DIVeR (Deterministic Integration for Volumetric Rendering)}~\cite{wu2022diver} is an improvement over NeRF, offering real-time and accurate neural radiance fields with deterministic integration for volume rendering. The authors propose a deterministic integration method that accelerates the rendering process while maintaining accuracy. The training paradigm involves optimizing the neural network using a rendering loss. DIVeR demonstrates the potential for real-time volume rendering applications, such as virtual reality and 3D modeling.
    \item \textbf{DVGO (Direct Voxel Grid Optimization)}~\cite{sun2022direct} is an improvement over NeRF, focusing on super-fast convergence for radiance field reconstruction. The authors propose a direct voxel grid optimization method that accelerates the convergence of the NeRF training process. The training paradigm involves optimizing the neural network using a rendering loss. DVGO is suitable for applications like 3D modeling, computer graphics, and real-time rendering.
    \item \textbf{Mip-NeRF 360}~\cite{barron2022mip} is an extension of NeRF, focusing on unbounded anti-aliased neural radiance fields. The authors propose a method that leverages mipmapping to reduce aliasing artifacts in NeRF-based renderings. The training paradigm involves optimizing the neural network using a rendering loss. Mip-NeRF 360 is suitable for applications like virtual reality, computer graphics, and 3D modeling, where high-quality renderings are essential.
    \item \textbf{VaxNeRF}~\cite{kondo2021vaxnerf} is an improvement over NeRF that introduces voxel acceleration for faster rendering. The authors propose a method that combines voxel-based acceleration with neural radiance fields, resulting in faster rendering times without sacrificing quality. The training paradigm involves optimizing the neural network using a rendering loss. VaxNeRF is suitable for real-time rendering applications, such as virtual reality and 3D modeling.
    \item \textbf{GeoNeRF}~\cite{johari2022geonerf} is an extension of NeRF that incorporates geometric priors for better scene representation. The authors propose a method that integrates geometric information into the NeRF framework, improving the quality and accuracy of the rendered images. The training paradigm involves optimizing the neural network using a combination of rendering loss and a geometric prior loss. GeoNeRF is suitable for applications like 3D modeling, computer graphics, and virtual reality.
    \item \textbf{NeRF in the Dark}~\cite{mildenhall2022nerf} is an extension of NeRF that focuses on high dynamic range (HDR) view synthesis from noisy raw images. The authors propose a method that leverages NeRF for rendering HDR images from raw, noisy input data. The training paradigm involves optimizing the neural network using a combination of rendering loss and a denoising loss. NeRF in the Dark is suitable for applications like photography, computer graphics, and virtual reality, where HDR rendering from noisy raw images is essential.
    \item \textbf{Deblur-NeRF}~\cite{ma2022deblur} is an extension of NeRF that focuses on reconstructing neural radiance fields from blurry images. The authors propose a method that leverages NeRF to render sharp images from input data that is blurred due to camera motion or other factors. The training paradigm involves optimizing the neural network using a combination of rendering loss and a deblurring loss. Deblur-NeRF is suitable for applications like photography, computer graphics, and virtual reality, where rendering sharp images from blurry inputs is important.
    \item \textbf{HDR-NeRF (High Dynamic Range Neural Radiance Fields)}~\cite{huang2022hdr} is an extension of NeRF, focusing on high dynamic range (HDR) neural radiance fields. The authors propose a method that enables NeRF to handle HDR images by modeling both radiance and color. The training paradigm involves optimizing the neural network using a rendering loss. HDR-NeRF is suitable for applications like photography, computer graphics, and virtual reality, where HDR rendering is essential.
    \item \textbf{iLabel (Interactive Neural Scene Labelling)}~\cite{zhi2021ilabel} is an application of NeRF that focuses on interactive neural scene labeling. The authors propose an interactive framework for labeling scenes with semantic information using a neural radiance field. The training paradigm involves optimizing the neural network using a combination of rendering loss and a semantic loss. iLabel is suitable for applications like 3D scene understanding, computer graphics, and virtual reality.
    \item \textbf{Urban Radiance Fields}~\cite{rematas2022urban} is an application of NeRF that focuses on modeling and rendering complex urban environments. The authors propose a method that leverages NeRF to capture and render detailed urban scenes. The training paradigm involves optimizing the neural network using a rendering loss. Urban Radiance Fields is suitable for applications like urban planning, virtual reality, and computer graphics.
    \item \textbf{NeRFReN (Neural Radiance Fields with ReflectioNs)}~\cite{guo2022nerfren} is an extension of NeRF that incorporates reflections into the rendering process. The authors propose a method that models reflections in neural radiance fields by incorporating a reflection term in the NeRF formulation. The training paradigm involves optimizing the neural network using a rendering loss. NeRFReN is suitable for applications like computer graphics, virtual reality, and 3D modeling, where realistic reflections are important.
    \item \textbf{Hallucinated Neural Radiance Fields in the Wild}~\cite{chen2022hallucinated} is an improvement over NeRF, focusing on rendering hallucinated neural radiance fields in the wild. The authors propose a method that leverages NeRF to generate plausible renderings of scenes with limited or no input data. The training paradigm involves optimizing the neural network using a rendering loss. Hallucinated-NeRF is suitable for applications like computer graphics, virtual reality, and 3D modeling, where input data may be scarce or unavailable.
    \item \textbf{NeuSample (Neural Sample Field for Efficient View Synthesis)}~\cite{fang2021neusample} is an improvement over NeRF that focuses on efficient view synthesis. The authors propose a method that learns a neural sample field to guide the sampling process, reducing the number of required samples for high-quality renderings. The training paradigm involves optimizing the neural network using a rendering loss. NeuSample is suitable for applications like virtual reality, computer graphics, and real-time rendering.
    \item \textbf{RegNeRF (Regularized Neural Radiance Fields)}~\cite{niemeyer2022regnerf} is an improvement over NeRF that focuses on view synthesis from sparse inputs. The authors propose a method that introduces regularization techniques to the NeRF framework to improve its performance when synthesizing views from a limited number of input images. The training paradigm involves optimizing the neural network using a combination of rendering loss and regularization losses (e.g., smoothness and sparsity losses). RegNeRF is suitable for applications like 3D modeling, computer graphics, and virtual reality, where view synthesis from sparse input data is a common challenge.
    \item \textbf{Zero-Shot Text-Guided Object Generation with Dream Fields (also known as Dream Fields)}~\cite{jain2022zero} is an application of NeRF for text-guided object generation. The authors propose a method that generates 3D objects conditioned on a given textual description. The training paradigm involves optimizing a neural network with a combination of rendering loss and textual conditioning loss. Applications include 3D object generation, virtual world creation, and content generation based on textual input. It extends NeRF by integrating textual guidance into the framework.
    \item \textbf{Efficient NeRF with learned depth-guided sampling}~\cite{lin2021efficient} is an improvement over NeRF that focuses on speeding up the rendering process by incorporating learned depth-guided sampling. The training paradigm involves optimizing the neural network using a combination of rendering loss and a depth-guided sampling loss. Applications include real-time rendering, computer graphics, and virtual reality. This method improves NeRF's efficiency by learning depth-guided sampling strategies.
    \item \textbf{Learning Neural Light Fields with Ray-Space Embedding Networks}~\cite{attal2022learning} is an extension of NeRF for learning neural light fields. The authors propose a method that learns a compact representation of light fields using a ray-space embedding network. The training paradigm involves optimizing the neural network using a rendering loss. Applications include light field rendering, computer graphics, and virtual reality. It extends NeRF by using a compact representation for light fields.
    \item \textbf{Neural Head Avatars}~\cite{grassal2022neural} is an application of NeRF that focuses on generating 3D head avatars from monocular RGB videos. The authors propose a method that reconstructs neural radiance fields for human heads from single-view RGB video input. The training paradigm involves optimizing the neural network using a combination of rendering loss and additional losses for facial details. Applications include virtual avatars, video conferencing, and virtual reality. It extends NeRF by applying it to human head reconstruction from single-view video input.
    \item \textbf{NeRF-SR}~\cite{wang2022nerf} is an improvement over NeRF that focuses on generating high-quality neural radiance fields using super-sampling. The authors propose a method that employs super-sampling during the training and rendering processes to improve the quality of the generated views. The training paradigm involves optimizing the neural network using a rendering loss. Applications include high-quality rendering, computer graphics, and virtual reality. It improves NeRF's rendering quality by incorporating super-sampling.
    \item \textbf{MoFaNeRF (Morphable Neural Radiance Fields)}~\cite{zhuang2022mofanerf} is an extension of NeRF that focuses on creating morphable neural radiance fields. The authors propose a method that learns a single, unified neural radiance field capable of representing multiple object instances within a category. The training paradigm involves optimizing the neural network using a combination of rendering loss and morphing loss. Applications include 3D object generation, animation, and computer graphics. It extends NeRF by learning a single representation capable of representing multiple object instances.
    \item \textbf{HumanNeRF (Generalizable Neural Human Radiance Field from Sparse inputs)}~\cite{zhao2021humannerf} is an extension of NeRF that focuses on reconstructing human shapes and appearance from sparse input views. The authors propose a method that leverages skeletal and semantic constraints to improve generalization and efficiency. The training paradigm involves optimizing the neural network using a combination of rendering loss and additional losses for skeletal and semantic constraints. Applications include human shape reconstruction, animation, and computer graphics. It extends NeRF by incorporating human-specific constraints.
    \item \textbf{Dense Depth Priors for Neural Radiance Fields from Sparse Input Views}~\cite{roessle2022dense} is an improvement over NeRF that incorporates dense depth priors to handle sparse input views. The authors propose a method that leverages dense depth information as a prior to improve the quality and stability of the reconstructed scene. The training paradigm involves optimizing the neural network using a combination of rendering loss and depth prior loss. Applications include scene reconstruction from sparse views and computer graphics. It improves NeRF by incorporating dense depth priors.
    \item \textbf{CG-NeRF (Conditional Generative Neural Radiance Fields)}~\cite{jo2021cg} is an extension of NeRF that introduces conditional generative neural radiance fields. The authors propose a method that conditions the neural radiance field on external information, such as object attributes or scene context. The training paradigm involves optimizing the neural network using a combination of rendering loss and conditioning loss. Applications include conditional scene generation and computer graphics. It extends NeRF by introducing conditional generation capabilities.
    \item \textbf{Ref-NeRF: Structured View-Dependent Appearance for Neural Radiance Fields}~\cite{verbin2022ref} is an extension of NeRF that focuses on modeling structured view-dependent appearance. The authors propose a method that explicitly models view-dependent reflections and refractions in the scene. The training paradigm involves optimizing the neural network using a rendering loss and additional losses for view-dependent appearance. Applications include rendering scenes with complex reflections and refractions, and computer graphics. It extends NeRF by modeling structured view-dependent appearance.
    \item \textbf{Geometry-Guided Progressive NeRF}~\cite{chen2021geometry} is an improvement over NeRF that focuses on efficient neural human rendering. The authors propose a method that leverages geometric priors and a progressive training strategy to improve rendering efficiency and generalization. The training paradigm involves optimizing the neural network using a rendering loss and additional losses for geometric constraints. Applications include human rendering, animation, and computer graphics. It improves NeRF by incorporating geometric priors and a progressive training strategy.
    \item \textbf{Deep Visual Constraints (Neural Implicit Models for Manipulation Planning from Visual Input)}~\cite{ha2022deep} is an application of NeRF that focuses on manipulation planning using visual input. The authors propose a method that learns an implicit model of the scene and integrates it into a planning framework. The training paradigm involves optimizing the neural network using a combination of rendering loss and planning loss. Applications include robotic manipulation planning, computer vision, and robotics. It extends NeRF by applying it to manipulation planning from visual input.
    \item \textbf{CLIP-NeRF}~\cite{wang2022clip} is an extension of NeRF that combines textual and visual information for scene manipulation. It leverages the CLIP (Contrastive Language-Image Pretraining) model to condition the neural radiance fields on both text and image inputs. The training paradigm involves optimizing the neural network using a combination of rendering loss and CLIP loss. Applications include text-and-image-based scene editing and computer graphics. It extends NeRF by introducing text-and-image-driven manipulation capabilities.
    \item \textbf{Neural Radiance Fields for Outdoor Scene Relighting}~\cite{rudnev2021neural} is an application of NeRF that focuses on relighting outdoor scenes. The authors propose a method that models outdoor lighting conditions and captures view-dependent appearance, allowing for relighting under various conditions. The training paradigm involves optimizing the neural network using a rendering loss and additional losses for lighting and appearance. Applications include outdoor scene rendering, virtual reality, and computer graphics. It applies NeRF to outdoor scene relighting.
    \item \textbf{CityNeRF}~\cite{xiangli2021citynerf} is an extension of NeRF that scales the representation to large-scale city scenes. The authors propose a method that leverages a hierarchical structure and efficient rendering techniques to handle city-scale data. The training paradigm involves optimizing the neural network using a rendering loss and additional losses for spatial coherence and efficiency. Applications include city-scale scene rendering, mapping, and urban planning. It extends NeRF by scaling the representation to city-scale scenes.
    \item \textbf{HeadNeRF (A Real-Time NeRF-based Parametric Head Model)}~\cite{hong2022headnerf} is an application of NeRF that focuses on real-time head modeling. The authors propose a method that combines a parametric head model with a neural radiance field, enabling real-time rendering and animation of head models. The training paradigm involves optimizing the neural network using a rendering loss and additional losses for parametric constraints. Applications include real-time head rendering, animation, and computer graphics. It applies NeRF to real-time head modeling.
    \item \textbf{GRAM (Generative Radiance Manifolds)}~\cite{deng2022gram} is an extension of NeRF that introduces generative radiance manifolds for 3D-aware image generation. The authors propose a method that leverages a manifold-based representation to model scene geometry and appearance, enabling high-quality image synthesis. The training paradigm involves optimizing the neural network using a combination of rendering loss and manifold loss. Applications include 3D-aware image generation and computer graphics. It extends NeRF by introducing generative radiance manifolds.
    \item \textbf{Solving Inverse Problems with NeRFGANs}~\cite{daras2021solving} is an extension of NeRF that combines the power of NeRF with GANs to solve inverse problems. The authors propose a method that uses NeRF as a generator and a discriminator to optimize the solution for various inverse problems. The training paradigm involves optimizing the neural network using a combination of rendering loss and adversarial loss. Applications include solving inverse problems in computer graphics and computer vision. It extends NeRF by integrating it with GANs to solve inverse problems.
    \item \textbf{HVTR (Hybrid Volumetric-Textural Rendering)} is an extension of NeRF that combines volumetric and textural rendering for human avatars. The authors propose a method that jointly learns geometric, appearance, and textural information, enabling more realistic rendering of human avatars. The training paradigm involves optimizing the neural network using a combination of rendering loss, texture loss, and geometric loss. Applications include human avatar rendering, animation, and virtual reality. It extends NeRF by introducing hybrid volumetric-textural rendering.
    \item \textbf{Mega-NeRF}~\cite{turki2022mega} is an extension of NeRF that scales the representation to large-scale scenes, suitable for virtual fly-throughs. The authors propose a method that leverages a hierarchical structure and efficient rendering techniques to handle large-scale data. The training paradigm involves optimizing the neural network using a rendering loss and additional losses for spatial coherence and efficiency. Applications include large-scale scene rendering, virtual reality, and computer graphics. It extends NeRF by scaling the representation to large-scale scenes.
    \item \textbf{3D-Aware Image Synthesis via Learning Structural and Textural Representations}~\cite{xu20223d} is an application of NeRF that focuses on 3D-aware image synthesis. The authors propose a method that learns structural and textural representations to synthesize high-quality images while considering 3D geometry. The training paradigm involves optimizing the neural network using a combination of rendering loss, structure loss, and texture loss. Applications include 3D-aware image synthesis, computer graphics, and computer vision. It applies NeRF to 3D-aware image synthesis.
    \item \textbf{Learning Implicit Body Representations from Double Diffusion Based Neural Radiance Fields}~\cite{yao2022learning} is an extension of NeRF that focuses on learning implicit body representations from double diffusion-based neural radiance fields. The authors propose a method that leverages a double diffusion process to capture fine-scale details and appearance information. The training paradigm involves optimizing the neural network using a combination of rendering loss, diffusion loss, and additional losses for geometry and appearance. Applications include human body modeling, animation, and virtual reality. It extends NeRF by introducing double diffusion-based neural radiance fields.
    \item \textbf{BANMo (Building Animatable 3D Neural Models)}~\cite{yang2022banmo} is an application of NeRF that focuses on building animatable 3D neural models from casual videos. The authors propose a method that leverages spatiotemporal information to learn animatable neural models that can be applied to a variety of video inputs. The training paradigm involves optimizing the neural network using a combination of rendering loss, temporal coherence loss, and additional losses for animation and appearance. Applications include video-based animation, computer graphics, and virtual reality. It applies NeRF to animatable 3D neural models.
    \item \textbf{InfoNeRF (Ray Entropy Minimization for Few-Shot Neural Volume Rendering)}~\cite{kim2022infonerf} is an improvement over NeRF that focuses on few-shot neural volume rendering. The authors propose a method that minimizes ray entropy to improve the quality of rendered images when trained with limited input views. The training paradigm involves optimizing the neural network using a combination of rendering loss and ray entropy loss. Applications include few-shot neural volume rendering, computer graphics, and computer vision. It improves NeRF by introducing ray entropy minimization for few-shot rendering. (Work in Progress...)
\end{itemize}

\item \textbf{Research in 2022:} This subsection briefly discusses the research articles in the last year.

\begin{itemize}    
    
                        
            
    \item \textbf{DFA-NeRF: Personalized Talking Head Generation via Disentangled Face Attributes Neural Rendering} focuses on generating high-quality, personalized talking head images. The authors tackled the challenge of synchronizing lip movement with audio while also generating personalized attributes like head movement and eye blink. They observed that while lip motion is closely related to audio, other personalized attributes like head poses and eye blinks are less correlated and vary from person to person. To address this, they proposed disentangling these attributes using a dynamic NeRF conditioned on the features generated from the input audio. The framework employs a Transformer-based VAE sampled from a Gaussian Process to learn plausible head poses and eye blinks. The lip movements are directly predicted from audio inputs for lip-synchronized generation. This approach allows for the creation of natural and high-fidelity talking head images, outperforming state-of-the-art methods in several benchmarks. The applications of this technique are significant in fields like film production and online meetings, where photo-realistic and expressive talking heads are essential.
    
    \item \textbf{Surface-Aligned Neural Radiance Fields for Controllable 3D Human Synthesis} authored by Tianhan Xu et al, introduced a novel method for reconstructing controllable implicit 3D human models from sparse multi-view RGB videos. Their approach defined the neural scene representation on the mesh surface points and signed distances from the surface of a human body mesh. They identified and addressed an indistinguishability issue in learning surface-aligned neural scene representation by proposing a projection of a point onto a mesh surface using barycentric interpolation with modified vertex normals. This method demonstrated superior quality in novel-view and novel-pose synthesis compared to existing methods and supported the control of body shape and clothes. The authors' contributions included an algorithm for injectively mapping a spatial point to a novel surface-aligned representation and novel surface-aligned neural radiance fields using this mapping, which showed better generalization performance on novel view and pose synthesis. Their approach, combining parametric 3D body models with NeRF, aimed to reconstruct a 3D human model that could be photorealistically rendered with fully controllable camera view, human pose, and body shape. The method was validated using the ZJU-MoCap and Human3.6M datasets, demonstrating its effectiveness in synthesizing images and its potential for real-world applications in areas like movies, games, and virtual reality, where free-viewpoint rendering of 3D human models is essential.
    
    \item \textbf{NeROIC: Neural Rendering of Objects from Online Image Collections}, developed by Zhengfei Kuang et al, is an innovative method for extracting high-quality geometric and material properties of objects from diverse online image collections. This approach, extending neural radiance fields (NeRF), operates in a multi-stage process. Initially, it infers object geometry and refines camera parameters using sparse images and coarse foreground masks. The method then progresses to deduce surface material properties and lighting conditions, employing a modular pipeline for efficient ray sampling and enhanced training. A unique aspect of their technique is the robust normal estimation, which mitigates geometric noise while capturing essential details. This method significantly outperforms existing techniques in rendering applications, showcasing its potential in fields like virtual reality and digital art, where accurate object rendering from unstructured image sources is vital.
    
    \item \textbf{VEOs (Virtual Elastic Objects)}, developed by Hsiao-yu Chen and colleagues, represents a significant advancement in the field of 3D reconstruction and virtual reality. The team created virtual objects that not only visually resemble their real-world counterparts but also mimic their physical behavior under novel interactions. This complex task involved capturing objects with a multi-view camera system while they deformed under external forces, such as a compressed air stream. The authors then reconstructed meshless geometry and deformation fields from these sequences using dynamic Neural Radiance Fields. A key innovation was the use of a differentiable, particle-based simulator to optimize the material parameters of the objects, enabling the simulation of new, plausible object configurations in response to different force fields or collision constraints. This method allowed for the re-rendering of the deformed state, integrating the simulation results with Neural Radiance Fields for realistic rendering. The versatility of VEOs was demonstrated through their ability to handle objects made of inhomogeneous materials, various shapes, and simulate interactions with other virtual objects. The authors' approach, detailed in their paper, opens new possibilities for realistic interactions with virtual objects in augmented and virtual reality applications.
    
    \item \textbf{Semantic-Aware Implicit Neural Audio-Driven Video Portrait Generation}, by Xian Liu et al, introduced a novel approach, Semantic-aware Speaking Portrait NeRF (SSP-NeRF), for creating high-fidelity audio-driven video portraits. This method uniquely combines the implicit 3D scene representation of NeRF with semantic awareness to handle the intricate dynamics and relationships between different parts of a speaking portrait. The authors developed two key modules: the Semantic-Aware Dynamic Ray Sampling module, which adjusts the number of rays sampled at each semantic region based on parsing difficulty, and the Torso Deformation module, which predicts displacements in the 3D scene to model non-rigid torso motions. These innovations allow SSP-NeRF to pay more attention to small but crucial areas like the lips, ensuring better lip-synced results and overall portrait quality. The method efficiently renders realistic video portraits with one unified set of NeRF, outperforming state-of-the-art methods in both objective evaluations and human studies. This technique has significant implications for digital human creation, film-making, and video dubbing, where generating lifelike, interactive virtual characters is essential.
    
    \item \textbf{Point-NeRF: Point-based Neural Radiance Fields}, a novel approach presented by Xu et al, revolutionizes volumetric neural rendering by combining the strengths of NeRF and deep multi-view stereo methods. This technique utilizes neural 3D point clouds with associated neural features to model a radiance field, enabling efficient rendering and reconstruction. Point-NeRF stands out by avoiding unnecessary sampling in empty scene spaces, leading to more accurate and faster reconstructions than existing methods. The authors introduced a learning-based framework for initializing and optimizing point-based radiance fields, leveraging deep multi-view stereo techniques for initial field generation. This process involves predicting depth with a cost-volume-based network, unprojecting to 3D space, and extracting 2D feature maps from input images to form a neural point cloud. The point cloud is then fine-tuned per scene, significantly reducing the time required for photorealistic rendering compared to NeRF. Additionally, the authors addressed common issues in 3D reconstruction methods, such as holes and outliers, by introducing novel point pruning and growing mechanisms during the optimization process. The technique's efficacy was demonstrated on various datasets, showing its superiority over existing methods in novel view synthesis. Point-NeRF's application extends to efficient and accurate 3D scene reconstruction and rendering, with potential real-world implications in virtual reality, augmented reality, and digital content creation.
    
    \item \textbf{From data to functa: Your data point is a function and you can treat it like one} by Emilien Dupont et al presents a transformative approach in deep learning by representing data as functions, or 'functa', using INRs. This method, which departs from traditional discrete array representations, involves first converting various data modalities like images and 3D shapes into functa and then utilizing these functa for deep learning tasks such as generative modeling, data imputation, and classification. The authors leverage Sinusoidal Representation Networks for INRs and introduce 'modulations' for a compact representation of functa. They also employ meta-learning for efficient creation of large functa datasets. The paper demonstrates the framework's versatility and efficiency across different applications, including image generation and novel view synthesis, highlighting its potential as a robust alternative for data representation in diverse deep learning contexts.
    
    \item \textbf{Differentiable Neural Radiosity}, introduced by Saeed Hadadan and Matthias Zwicker, marks a breakthrough in realistic rendering, as detailed in their research. This innovative method employs a neural network to represent the differential radiance field, crucial for computing derivatives of rendering processes relative to scene parameters like geometry and lighting. The technique outperforms existing methods like Automatic Differentiation and Radiative Backpropagation in efficiency and accuracy, particularly in handling indirect effects and visibility-related discontinuities in rendering equations. Its application significantly enhances inverse rendering problems, offering substantial benefits in fields requiring photo-realistic rendering such as virtual reality, game development, and film production, by adeptly managing complex lighting and material properties in scenes.
    
    \item \textbf{CLA-NeRF: Category-Level Articulated Neural Radiance Field}, developed by Wei-Cheng et al, represents a groundbreaking approach in the realm of robotics and computer vision for understanding and interacting with articulated objects. This method, trained solely on RGB images without the need for CAD models or depth information, is adept at view synthesis, part segmentation, and articulated pose estimation for various object categories. By extending the NeRF framework, traditionally used for static scenes, to handle articulated objects, the authors enabled CLA-NeRF to render objects and their parts in unseen articulated poses. The technique involves predicting color, density, and segmentation for each 3D location, followed by volume rendering for image synthesis. It uniquely estimates joint attributes from segmentation fields and applies deformation matrices during rendering, adapting to different poses. Tested on both synthetic and real-world data, CLA-NeRF demonstrated its capability in rendering realistic deformations and accurately estimating articulated poses, marking a significant advancement in robotic perception and interaction with articulated objects.
    
    \item \textbf{MedNeRF: Medical Neural Radiance Fields for Reconstructing 3D-aware CT-Projections from a Single X-ray} by Corona-Figueroa et al. introduces a deep learning model, MedNeRF, which reconstructs CT projections from a single-view X-ray. This innovative approach is based on neural radiance fields, specifically adapting Generative Radiance Fields (GRAF) for medical imaging. The authors generated Digitally Reconstructed Radiographs (DRRs) from chest and knee CT scans to train their model, avoiding additional radiation exposure to patients. They employed self-supervised learning and Data Augmentation Optimized for GAN (DAG) to enhance feature extraction and model learning. MedNeRF outperformed existing methods like GRAF and pixelNeRF in rendering detailed internal structures and estimating volumetric depth, as evidenced by quantitative metrics like PSNR, SSIM, FID, and KID. This technique holds significant potential for clinical applications, particularly in reducing radiation exposure during imaging processes, and could be beneficial in areas like osseous trauma, skeletal evaluation in dysplasia, and orthopedic pre-surgical planning.
    
    \item \textbf{PVSeRF: Joint Pixel-, Voxel- and Surface-Aligned Radiance Field for Single-Image Novel View Synthesis}, a novel framework introduced by Xianggang Yu and colleagues, tackled the challenge of reconstructing neural radiance fields from single-view RGB images for novel view synthesis. The authors identified a key limitation in existing solutions like pixelNeRF, which suffered from feature ambiguity due to reliance solely on pixel-aligned features. This led to difficulties in disentangling geometry and appearance, often resulting in implausible geometries and blurry results. To overcome this, PVSeRF uniquely combined pixel-aligned features with explicit geometry reasoning, incorporating both voxel-aligned features from a coarse volumetric grid and fine surface-aligned features extracted from a regressed point cloud. This approach significantly improved the disentanglement between appearance and geometry, leading to more accurate geometries and higher quality images of novel views. The authors demonstrated PVSeRF's superiority over state-of-the-art methods in extensive experiments on the ShapeNet benchmarks, highlighting its potential for practical applications in fields like gaming, movie production, and virtual/augmented reality. The framework's ability to synthesize high-quality novel views from a single image, while also providing cleaner implicit surface meshes, marked a significant advancement in the field of computer vision and graphics.
    
    \item \textbf{Block-NeRF (Scalable Large Scene Neural View Synthesis)}, introduced by Matthew Tancik and his team, revolutionized neural view synthesis for large-scale environments like cityscapes. Addressing the limitations of traditional Neural Radiance Fields (NeRF), Block-NeRF decomposes scenes into multiple, individually trained NeRFs, enabling scalable rendering and updates without retraining the entire model. This method incorporates architectural enhancements such as appearance embeddings, learned pose refinement, and exposure control, ensuring adaptability to diverse environmental conditions. Tested on a grand scale with 2.8 million images to recreate a San Francisco neighborhood, Block-NeRF demonstrated its potential in applications ranging from autonomous driving to aerial surveying. Its ability to handle varying lighting, weather conditions, transient objects, and visibility prediction marks it as a robust solution for dynamic, large-scale environment reconstruction and rendering.
    
    \item \textbf{NeuVV (Neural Volumetric Videos with Immersive Rendering and Editing)}, developed by Jiakai Zhang and his team, represents a groundbreaking approach in the realm of neural volumetric videos, offering immersive rendering and editing capabilities. This technique, designed to support real-time volumographic rendering, allows users to interact with and view content in a virtual 3D space, changing viewpoints by moving around freely. NeuVV utilizes a dynamic Neural Radiance Field (NeRF) to encode appearance, geometry, and motion from all viewpoints, employing factorization schemes like hyperspherical harmonics (HH) for modeling color variations and a learnable basis representation for abrupt density changes due to motion. This factorization is integrated into a Video Octree (VOctree) to significantly accelerate training while reducing memory overhead. NeuVV's real-time rendering capability enables a range of immersive content editing tools, allowing users to rearrange and repurpose content in both spatial and temporal dimensions. The system supports hybrid neural-rasterization rendering on consumer-level VR headsets, enabling immersive viewing and content editing in 3D virtual environments. NeuVV's innovative approach to neural volumetric video production marks a significant advancement in the field, offering high photo-realism and interactive editing capabilities for volumetric human performances.
    
    \item \textbf{Fourier PlenOctrees for Dynamic Radiance Field Rendering in Real-time} by Liao Wang and colleagues introduces the Fourier PlenOctree (FPO) technique, a novel method for efficient neural modeling and real-time rendering of dynamic scenes. This technique is a unique amalgamation of generalized NeRF, PlenOctree representation, volumetric fusion, and Fourier transform. The authors developed a coarse-to-fine fusion scheme using generalizable NeRF to generate the tree via spatial blending, addressing the challenge of dynamic scenes by tailoring the implicit network to model Fourier coefficients of time-varying density and color attributes. The FPO is constructed and trained directly on the leaves of a union PlenOctree structure of the dynamic sequence, resulting in compact memory overload and efficient fine-tuning capabilities. Extensive experiments demonstrated that FPO is significantly faster than the original NeRF and offers substantial acceleration over state-of-the-art techniques while maintaining high visual quality for free-viewpoint rendering of dynamic scenes. This advancement in neural representation and rendering technology marks a significant step forward in interactive and immersive applications like Telepresence and Virtual Reality.
    
    \item \textbf{Learning Multi-Object Dynamics with Compositional Neural Radiance Fields} by Danny Driess et al introduces a method for learning compositional multi-object dynamics models from image observations. This method uniquely combines implicit object encoders, NeRFs, and GNNs. Unlike traditional NeRF approaches that train on a single scene, this method employs a compositional, object-centric auto-encoder framework. It maps multiple views of a scene to a set of latent vectors, each representing an individual object. These latent vectors parameterize individual NeRFs for scene reconstruction. The authors then train a GNN dynamics model in the latent space, achieving compositionality in dynamics prediction. A key feature of this approach is that the latent vectors encode 3D information through the NeRF decoder, allowing the incorporation of structural priors in learning dynamics models. This results in more stable long-term predictions compared to several baselines. The method demonstrates its efficacy in modeling and learning the dynamics of compositional scenes, including both rigid and deformable objects, in simulated and real-world experiments. This novel approach significantly advances the field of dynamics modeling from visual observations, particularly in complex, multi-object environments.
    
    \item \textbf{Pix2NeRF (Unsupervised Conditional $\pi$-GAN for Single Image to Neural Radiance Fields Translation)} by Shengqu Cai et al introduces an innovative pipeline for generating NeRF of an object or scene from a single input image. This task is challenging as NeRF typically requires multiple views of the same scene with corresponding poses. The authors' method is based on $\pi$-GAN, a generative model for 3D-aware image synthesis, which maps random latent codes to radiance fields of a class of objects. They optimize the $\pi$-GAN objective for high-fidelity 3D-aware generation and a reconstruction objective, including an encoder coupled with the $\pi$-GAN generator to form an auto-encoder. This approach is unsupervised, capable of being trained with independent images without 3D, multi-view, or pose supervision. The applications of this pipeline are diverse, including 3D avatar generation, object-centric novel view synthesis from a single input image, and 3D-aware super-resolution. The Pix2NeRF model stands out as the first unsupervised single-shot NeRF model, learning scene radiance fields from images without extensive supervision. It represents a significant advancement in conditional GAN-based NeRF and NeRF-based GAN inversion, demonstrating superiority over naive GAN inversion methods and providing a strong baseline for future research in these areas.
    
    \item \textbf{ERF (Explicit Radiance Field Reconstruction From Scratch)} by Samir Aroudj and team, is a novel explicit dense 3D reconstruction approach, using hierarchical volumetric fields in a sparse voxel octree. They employed stochastic gradient descent (Adam) steered by an inverse differentiable renderer, enabling the reconstruction of high-quality models comparable to implicit methods. This method allows practical reconstruction of a variety of scenes, including challenging objects, without a controlled lab setup.
    
    \item \textbf{ICARUS (A Specialized Architecture for Neural Radiance Fields Rendering)} by Chaolin Rao and colleagues presents ICARUS, a specialized hardware architecture designed to accelerate NeRF rendering. The key challenge addressed by the authors is the low rendering speed of NeRF on even high-end GPUs. ICARUS overcomes this by using dedicated plenoptic cores (PLCore) consisting of a positional encoding unit (PEU), a MLP engine, and a volume rendering unit (VRU). This design allows ICARUS to execute the complete NeRF pipeline efficiently, rendering pixel colors without needing intermediate off-chip data storage and exchange. The authors transformed fully connected operations in the MLP to approximated reconfigurable multiple constant multiplications (MCMs), enhancing computational efficiency. A prototype of ICARUS was built using an FPGA-based system, and its power-performance-area (PPA) was evaluated using 40nm LP CMOS technology. The results showed that ICARUS significantly reduces time and power consumption compared to GPU and TPU implementations. This architecture opens new possibilities for real-time, high-fidelity NeRF rendering in various applications, including mobile and VR/AR devices, by providing a more efficient and specialized solution than general-purpose computing hardware. ICARUS represents a significant step forward in the practical deployment of NeRF in rendering applications, particularly for scenarios that require lightweight graphics architecture.
    
    \item \textbf{NeuroFluid (Fluid Dynamics Grounding with Particle-Driven Neural Radiance Fields)} by Shanyan Guan et al is a two-stage network for fluid dynamics grounding from visual observations. It combines a particle-driven neural renderer with a particle transition model, enabling unsupervised learning of particle-based fluid dynamics. This approach estimates fluid physics with varying properties, like viscosity and density, from sequential visual observations.

    \item \textbf{NeRF-Supervision (Learning Dense Object Descriptors from Neural Radiance Fields)} by Lin Yen-Chen et is a self-supervised pipeline using NeRF for learning object-centric dense descriptors from RGB images. They optimized a NeRF to extract dense correspondences, training a model for view-invariant object representation. This method significantly outperformed existing descriptors, enabling accurate robot manipulation of challenging objects like thin and reflective items.
    
    \item \textbf{Playable Environments: Video Manipulation in Space and Time} is a framework for interactive video generation and manipulation using a single image. This method allows 3D object movement and video generation through user-specified actions, learned in an unsupervised manner. It extends NeRF with style-based modulation, training on monocular videos for diverse applications like 3D video generation and scene stylization.
    
    \item Ziyu Wang et al introduced \textbf{NeReF (Neural Refractive Field for Fluid Surface Reconstruction and Implicit Representation)}, a neural refractive field for reconstructing transparent fluid surfaces. It estimates surface position and normal, overcoming refraction distortions in traditional methods. NeReF's global optimization and continuous representation enable robust reconstruction and view synthesis, especially for sparse multi-view acquisition.
    
    \item Yinhuai Wang et al proposed \textbf{NeRFocus (Neural Radiance Field for 3D Synthetic Defocus)}, a thin-lens-imaging-based NeRF framework for rendering 3D defocus effects. It models the circle of confusion for scene points, using frustum-based volume rendering and a probabilistic training strategy. NeRFocus achieves adjustable defocus effects with various camera settings, maintaining NeRF's performance without explicit 3D reasoning.
    
    \item Yichao Yan and team developed  \textbf{DialogueNeRF (Towards Realistic Avatar Face-to-Face Conversation Video Generation)}, a NeRF-based framework for generating face-to-face conversation videos between avatars. It models speakers driven by audio and listeners responding to visual and acoustic cues. The method, validated on a new human conversation dataset, produces realistic conversational interactions and maintains individual avatar styles.
    
    \item \textbf{Animatable Implicit Neural Representations for Creating Realistic Avatars from Videos} by Sida Peng et al addressed reconstructing animatable human models from multi-view videos. They introduced a pose-driven deformation field based on linear blend skinning, combining blend weight fields and 3D human skeletons. This approach allows explicit control of the canonical human model with input motions, outperforming recent human modeling methods.
    
    \item Roger Marí and team introduced \textbf{Sat-NeRF (Learning Multi-View Satellite Photogrammetry With Transient Objects and Shadow Modeling Using RPC Cameras)}, a model for learning multi-view satellite photogrammetry. It combines neural rendering with native satellite camera models and a shadow-aware irradiance model. Sat-NeRF handles appearance changes due to shadows and transient objects, achieving high-quality surface models and view synthesis.
    
    \item \textbf{Enhancement of Novel View Synthesis Using Omnidirectional Image Completion} by Takayuki Hara and Tatsuya Harada presents a method for synthesizing novel views from a single 360-degree RGB-D image using NeRF. They reprojected images to complete missing regions with a 360-degree image completion network, training NeRF with selected completed images for 3D consistency. This method synthesizes plausible novel views while preserving scene features.
    
    \item \textbf{ViewFormer (NeRF-free Neural Rendering from Few Images Using Transformers)} by Jonáš Kulhánek et al is a 2D-only method using transformers for novel view synthesis from sparse context views. It employs a codebook and transformer model, with a branching attention mechanism for efficient training. ViewFormer competes with NeRF-based methods, offering faster training without explicit 3D reasoning.
    
    \item Jianxiong Shen and colleagues introduced \textbf{Conditional-Flow NeRF (Accurate 3D Modelling with Reliable Uncertainty Quantification)}, a probabilistic framework for uncertainty quantification in NeRF-based approaches. CF-NeRF learns a distribution over radiance fields, providing reliable uncertainty estimation while preserving model expressivity. It achieves lower prediction errors and more reliable uncertainty values compared to state-of-the-art methods.

    \item \textbf{Sem2NeRF: Converting Single-View Semantic Masks to Neural Radiance Fields} introduced by Yuedong Chen and team is a framework for reconstructing 3D scenes modeled by NeRF from single-view semantic masks. It encodes semantic masks into latent codes controlling a pre-trained decoder, integrating a region-aware learning strategy. Sem2NeRF outperforms baselines on benchmark datasets, converting semantic masks to 3D-aware generative models.
    
    \item \textbf{NeRFusion (Fusing Radiance Fields for Large-Scale Scene Reconstruction)} is a method combining NeRF and TSDF-based fusion techniques for efficient large-scale reconstruction and photo-realistic rendering. They developed a novel recurrent neural network to incrementally reconstruct a global, sparse scene representation in real-time. This approach significantly outperforms NeRF in terms of reconstruction speed and quality, particularly in large-scale indoor scenes.
    
    \item \textbf{NeuMan (Neural Human Radiance Field from a Single Video)} is a framework for reconstructing humans and scenes from a single video for augmented reality applications. Using two NeRF models, one for the human and one for the scene, the method leverages rough geometry estimates for warping fields and achieves high-quality renderings of humans in novel poses and views, along with the background.
    
    \item \textbf{Continuous Dynamic-NeRF: Spline-NeRF} introduces an architecture (Spline-NeRF) based on Bezier splines ensuring C0 and C1 continuity for reconstructing dynamic scenes. This method, blending machine learning with classical animation techniques, provides a more coherent and smooth movement in dynamic scene reconstruction compared to previous NeRF-based methods
    
    \item \textbf{Structured Local Radiance Fields for Human Avatar Modeling} is a novel representation using structured local radiance fields anchored to nodes on a human body template. This method, focusing on animatable clothed human avatars, combines implicit representation with conditional generative latent space learning, enabling automatic construction of realistic human avatars with dynamic garments.
    
    \item \textbf{RGB-D Neural Radiance Fields: Local Sampling for Faster Training} by Arnab Dey and Andrew I. Comport proposes a method leveraging RGB-D data for efficient training of Neural Radiance Fields. By introducing a depth-guided local sampling strategy and a smaller network architecture, their approach achieves faster training times without compromising quality, addressing the limitations of previous NeRF-based methods.
    
    \item \textbf{DRaCoN (Differentiable Rasterization Conditioned Neural Radiance Fields for Articulated Avatars)} by Amit Raj et al is a framework for learning dynamic full-body avatars from RGB videos. Combining differentiable rasterization with a conditional neural 3D representation module, DRaCoN achieves high-quality, scalable avatar generation with fine 3D geometric details, outperforming state-of-the-art methods in visual quality and error metrics.
    
    \item \textbf{DDNeRF (Depth Distribution Neural Radiance Fields)} is a method enhancing sampling efficiency in Neural Radiance Fields by learning a more accurate representation of density distribution along rays. This approach uses fewer samples during training, reducing computational resources while achieving superior results for a given sampling budget.
    
    \item \textbf{MPS-NeRF (Generalizable 3D Human Rendering from Multiview Images)} by Xiangjun Gao et al addresses the challenge of rendering novel views and poses for unseen persons using multiview images. MPS-NeRF leverages a canonical NeRF and volume deformation scheme, using a parametric 3D human model for deformation. This method demonstrates efficacy in novel view synthesis and pose animation tasks on both real and synthetic data.
    
    \item \textbf{R2L (Distilling Neural Radiance Field to Neural Light Field for Efficient Novel View Synthesis)} by Huan Wang et al proposed a method to distill Neural Radiance Fields into Neural Light Fields, significantly improving rendering speed and quality. Their deep residual MLP network effectively learns the light field, enabling efficient novel view synthesis with a substantial reduction in computational requirements and runtime, while delivering better rendering quality than NeRF.
    
    \item \textbf{SinNeRF (Training Neural Radiance Fields on Complex Scenes from a Single Image)} is a semi-supervised framework for training neural radiance fields using a single view. The paper employes semantic and geometry regularizations, achieving photo-realistic novel-view synthesis on complex scenes, outperforming current NeRF baselines.
    
    \item \textbf{Neural Rendering of Humans in Novel View and Pose from Monocular Video} presented a method for generating photo-realistic humans under novel views and poses from monocular videos. The approach integrated observations across frames and encoded individual frame appearances, leveraging human pose and point clouds for improved rendering quality.
    
    \item \textbf{Unified Implicit Neural Stylization} by the authors proposes a Unified Implicit Neural Stylization framework, INS, for training stylized implicit representations in 2D and 3D scenarios. They decoupled the implicit function into style and content modules, enabling efficient stylization for various implicit neural representations.
    
    \item \textbf{IRON (Inverse Rendering by Optimizing Neural SDFs and Materials from Photometric Images)} is a neural inverse rendering pipeline, was introduced for recovering 3D content from photometric images. The method utilized neural representations for geometry and materials, featuring a hybrid optimization scheme and an edge sampling algorithm for neural SDFs, achieving superior inverse rendering quality.
    
    \item \textbf{SqueezeNeRF (Further factorized FastNeRF for memory-efficient inference)} is a further factorized version of FastNeRF, was proposed for memory-efficient inference. It rendered at high frame rates with significantly reduced cache size, making it suitable for embedded system applications.
    
    \item \textbf{AutoRF (Learning 3D Object Radiance Fields from Single View Observations)} focused on learning neural 3D object representations from single-view observations. The method encoded shape, appearance, and pose in a normalized, object-centric representation, enabling novel view synthesis and demonstrating generalizability across different real-world datasets.
    
    \item \textbf{NAN (Noise-Aware NeRFs for Burst-Denoising)} leveraged NeRFs for burst denoising in low-light scenes with high noise and large motion. The approach augmented NeRFs with inter-view and spatial awareness, achieving state-of-the-art results in burst denoising.
    
    \item \textbf{GARF (Gaussian Activated Radiance Fields for High Fidelity Reconstruction and Pose Estimation)} is a new positional embedding-free neural radiance field architecture using Gaussian activations, was presented. It outperformed current methods in high fidelity reconstruction and pose estimation, simplifying the approach without the need for frequency scheduling.
    
    \item \textbf{Modeling Indirect Illumination for Inverse Rendering} proposed an efficient approach to recover spatially-varying indirect illumination in inverse rendering. The method derived indirect illumination from the neural radiance field learned from input images, enabling realistic renderings under novel viewpoints and illumination.
    
    \item \textbf{Implicit Object Mapping With Noisy Data} evaluated the impact of noisy data on training NeRF for implicit object reconstruction. It introduced a pipeline for decomposing scenes into individual object-NeRFs using noisy object instance masks and bounding boxes, analyzing sensitivity to noisy poses and masks.
    
    \item  The authors developed \textbf{Generalizable Neural Performer(GNR): Learning Robust Radiance Fields for Human Novel View Synthesis}, a framework for synthesizing free-viewpoint images of human performers from sparse multi-view images. GNR utilized Implicit Geometric Body Embedding and Screen-Space Occlusion-Aware Appearance Blending to handle diverse human poses, shapes, and appearances. The method demonstrated robustness and generalizability across various human subjects and actions, outperforming state-of-the-art methods in cross-dataset evaluations.
    
    \item \textbf{AE-NeRF (Auto-Encoding Neural Radiance Fields for 3D-Aware Object Manipulation)} was proposed for 3D-aware object manipulation. This framework, structured as an auto-encoder, extracted disentangled 3D attributes (shape, appearance, camera pose) from images and rendered high-quality images through generative NeRF. AE-NeRF introduced global-local attribute consistency loss and swapped-attribute classification loss, along with a stage-wise training scheme, achieving superior performance in 3D object manipulation compared to recent methods.
    
    \item \textbf{NeurMiPs (Neural Mixture of Planar Experts for View Synthesis)} presented a novel planar-based scene representation for modeling geometry and appearance. Each planar expert in NeurMiPs consisted of local rectangular shapes and a neural radiance field, enabling efficient rendering through ray-plane intersections. This approach combined the efficiency of explicit mesh rendering with the flexibility of neural radiance fields, showing superior performance and speed in novel view synthesis compared to other 3D representations.
    
    \item \textbf{Sampling-free obstacle gradients and reactive planning in NeRF} explores NeRF for motion planning, specifically obstacle avoidance. They augmented NeRF with the capacity to infer occupancy, learning an approximation to a ESDF. This approach, integrated with Riemannian Motion Policies (RMP), enabled fast, sampling-free obstacle avoidance planning.

    \item \textbf{Unsupervised Discovery and Composition of Object Light Fields} introduces a method for unsupervised learning of object-centric neural scene representations using light fields. The authors propose a light field compositor module for reconstructing global light fields from object-centric light fields, significantly speeding up rendering and training compared to existing 3D approaches.
    
    \item \textbf{Panoptic Neural Fields (A Semantic Object-Aware Neural Scene Representation)} is a neural scene representation that decomposes a scene into objects and background. Each element is represented by an MLP, with the background MLP also outputting semantic labels. This method enables novel view synthesis, 2D panoptic segmentation, and 3D scene editing.
    
    \item \textbf{NeRF-Editing: Geometry Editing of Neural Radiance Fields} introduces a method for geometric editing of NeRF. The authors established a correspondence between explicit mesh representation and implicit neural representation, allowing for controllable shape deformation and rendering of edited scenes.
    
    \item \textbf{RapNeRF (Ray Priors through Reprojection: Improving Neural Radiance Fields for Novel View Extrapolation)} is a novel approach for improving NeRF in novel view extrapolation, was introduced. It uses a random ray casting policy and a ray atlas pre-computed from observed rays’ viewing directions, aiming to enhance rendering quality for extrapolated views while reducing view-dependent effects.
    
    \item \textbf{Fast Neural Network based Solving of Partial Differential Equations} introduces a method using Neural Networks for solving PDEs, building on NeRF advancements. This approach converges faster than traditional Physically Informed Neural Networks (PINNs) and is particularly effective for electromagnetic problems.
    
    \item \textbf{Mip-NeRF RGB-D (Depth Assisted Fast Neural Radiance Fields)} investigates incorporating depth information into NeRF using RGB-D data. The authors propose a method that uses depth for local sampling and geometric loss, improving accuracy in geometry and photometry while reducing training time.
    
    \item \textbf{StylizedNeRF (Consistent 3D Scene Stylization as Stylized NeRF via 2D-3D Mutual Learning)} is a method for 3D scene stylization using a mutual learning framework between NeRF and a 2D image stylization network. It ensures consistency in rendered images from different views and handles ambiguities in 2D stylization results.
    
    \item \textbf{PREF (Phasorial Embedding Fields for Compact Neural Representations)} is a frequency-based neural representation, was introduced, combining a shallow MLP with a phasor volume. This method captures high-frequency details efficiently and is more compact than previous frequency-based representations.
    
    \item \textbf{V4D (Voxel for 4D Novel View Synthesis)} is a method for 4D novel view synthesis using 3D voxel to model dynamic neural radiance fields, was proposed. It includes a conditional positional encoding module and a pixel-level refinement module, achieving state-of-the-art performance in synthesis and real scenes datasets.
    
    \item \textbf{Compressible-composable NeRF via Rank-residual Decomposition} presents a method for efficient manipulation of Neural Radiance Fields (NeRF) using a hybrid tensor rank decomposition. The approach allows dynamic adjustment of model size for different levels of detail and supports arbitrary transformation and composition of different models. The method achieves near-optimal low-rank approximation, enabling both compressibility and composability of NeRF models.
    
    \item \textbf{Neural Volumetric Object Selection} introduces a method for selecting objects in neural volumetric 3D representations like Multi-plane Images (MPI) and NeRF. Using 2D user scribbles, the method estimates a 3D segmentation of the desired object, which can be rendered into novel views. This approach incorporates neural volumetric 3D representation and multi-view image features, outperforming existing 2D and 3D segmentation methods.
    
    \item \textbf{Fast Dynamic Radiance Fields with Time-Aware Neural Voxels} introduces \textbf{TiNeuVox}, a framework for dynamic radiance fields using time-aware voxel features. It models temporal information and captures different scales of point motions, significantly accelerating the optimization of dynamic radiance fields while maintaining high rendering quality. The method is evaluated on both synthetic and real scenes, demonstrating its efficiency and effectiveness.
    
    \item \textbf{Decomposing NeRF for Editing via Feature Field Distillation} proposes distilled feature fields (DFFs) for query-based scene decomposition and local editing of NeRFs. It uses teacher-student distillation from pre-trained 2D image feature extractors to create a 3D feature field that semantically decomposes 3D space. This enables semantic selection and editing of regions in NeRF scenes, transferring recent progress in 2D vision and language foundation models to 3D scene representations.
    
    \item \textbf{Novel View Synthesis for High-fidelity Headshot Scenes} focuses on rendering high-quality human faces from arbitrary viewpoints. The authors propose a method combining NeRF and 3D Morpable Models (3DMM) to synthesize photorealistic scenes with faces, preserving skin details. They use a GAN to mix NeRF-synthesized images and 3DMM-rendered images, producing scenes with accurate geometry, background, and skin details.
    
    \item \textbf{DeVRF (Fast Deformable Voxel Radiance Fields for Dynamic Scenes)} is a novel representation for learning dynamic radiance fields rapidly. It models both the 3D canonical space and 4D deformation field of dynamic scenes using explicit voxel-based representations. The approach addresses overfitting issues through a static-dynamic learning paradigm and identifies strategies to improve efficiency and effectiveness, achieving significant speedup with high-fidelity results compared to state-of-the-art methods.
    
    \item \textbf{D$^2$NeRF (Self-Supervised Decoupling of Dynamic and Static Objects from a Monocular Video)} is a self-supervised approach that decouples moving objects from static backgrounds in 3D scenes using monocular videos. It employs separate neural radiance fields for dynamic and static components and introduces a shadow field network to detect and decouple dynamically moving shadows. The method outperforms state-of-the-art approaches in decoupling dynamic and static 3D objects, occlusion and shadow removal, and image segmentation for moving objects.
    
    \item \textbf{EfficientNeRF (Efficient Neural Radiance Fields)} is a method enhancing NeRF for 3D scene representation and novel-view image synthesis. They tackled the challenge of reducing both training and testing time without compromising accuracy. By analyzing density and weight distribution, they proposed valid and pivotal sampling at coarse and fine stages, respectively, to improve sampling efficiency. Additionally, a novel data structure, NerfTree, was designed for caching the entire scene during testing, accelerating rendering speed. This method significantly reduced training time by over 88\% and achieved a rendering speed of over 200 FPS, making NeRF more practical for real-world applications.
    
    \item \textbf{Points2NeRF (Generating Neural Radiance Fields from 3D point cloud)} is a novel approach for generating NeRFs from 3D point clouds. They addressed the challenge of NeRF's dependency on base images and their registration by using 3D point clouds, which often suffer from sparsity in under-sampled regions. Their solution involved an autoencoder-based architecture with a hypernetwork paradigm, transferring 3D points and associated color values through a lower-dimensional latent space to generate NeRF model weights. This method enabled the use of sparse 3D point clouds for high-quality view synthesis, offering a more robust and generalizable model for various 3D object classes.
    
    \item \textbf{Reinforcement Learning with Neural Radiance Fields} explored the use of NeRF for learning state representations in RL. The authors proposed NeRF-RL, where an encoder maps multiple image observations to a latent space describing objects in a scene, supervised by a latent-conditioned NeRF decoder. This approach showed improved RL performance in tasks involving robotic object manipulations, like hanging mugs on hooks, pushing objects, or opening doors. The use of NeRF as supervision led to a latent space better suited for downstream RL tasks, demonstrating the potential of 3D inductive biases in RL.
    
    \item \textbf{SNAKE (Shape-aware Neural 3D Keypoint Field)} is an unsupervised paradigm for detecting 3D keypoints from point clouds. It simultaneously predicts implicit shape indicators and keypoint saliency, entangling 3D keypoint detection with shape reconstruction. SNAKE outperformed existing methods in repeatability, especially with down-sampled input point clouds, and allowed accurate geometric registration in zero-shot settings. This approach generated semantically consistent 3D keypoints without explicit supervision, demonstrating the advantages of integrating shape awareness into keypoint detection.
    
    \item \textbf{ObPose (Leveraging Pose for Object-Centric Scene Inference and Generation in 3D)} is an unsupervised object-centric inference and generation model for learning 3D-structured latent representations from RGB-D scenes. OBPOSE uses a factorized latent space encoding object location and appearance, and introduces pose as an inductive bias. It employs a voxelised approximation to recover object shape from a neural radiance field (NeRF), modeling each scene as a composition of NeRFs. OBPOSE outperformed state-of-the-art methods in unsupervised scene segmentation and demonstrated capabilities in scene generation and editing, highlighting the benefits of disentangling location and appearance in 3D scene understanding.
    
    \item \textbf{SS-NeRF (Beyond RGB: Scene-Property Synthesis with Neural Radiance Fields)} is a framework extending NeRF for comprehensive 3D scene understanding. SS-NeRF synthesizes not only photo-realistic RGB images from novel viewpoints but also various accurate scene properties like appearance, geometry, and semantics. This approach enables tasks such as semantic segmentation, surface normal estimation, reshading, keypoint detection, and edge detection under a unified framework. SS-NeRF uses two branches to handle properties sensitive and insensitive to observation directions, allowing coherent handling of different scene properties. The framework facilitates multi-task learning and knowledge transfer, serving as a tool for generative and discriminative learning. SS-NeRF also acts as an auto-labeller for data creation and augments data for downstream discriminative tasks, showcasing its versatility in scene 
    
    \item \textbf{NeRF-In (Free-Form NeRF Inpainting with RGB-D Priors)} is a framework enabling free-form inpainting in Neural Radiance Fields (NeRF) using RGB-D priors. This approach allows users to remove unwanted objects from a 3D scene represented by a pre-trained NeRF. Users specify regions for removal by drawing masks on rendered views from the NeRF. The framework transfers these masks to other views, generates guiding color and depth images within masked regions, and then optimizes the NeRF model's parameters to inpaint the image content across multiple views. NeRF-In demonstrates the capability to produce visually plausible and consistent results across different views, simplifying the process of editing 3D scenes without requiring additional category-specific training or data.
    
    \item \textbf{TransNeRF (Generalizable Neural Radiance Fields for Novel View Synthesis with Transformer)} is a Transformer-based NeRF for novel view synthesis. Unlike traditional MLP-based NeRFs, TransNeRF can directly process an arbitrary number of observed views without auxiliary pooling operations, capturing complex relationships between source and target views. It utilizes the attention mechanism in a Transformer network to integrate deep associations of source views into a coordinate-based scene representation. TransNeRF enhances local consistency in both ray-cast and surrounding-view spaces, enabling more effective generalization to novel scenes without per-scene fine-tuning. This approach addresses the limitations of previous NeRF models in handling large differences between source and rendering views, demonstrating improved performance in scene-agnostic and per-scene fine-tuning scenarios.
    
    \item \textbf{AR-NeRF (Unsupervised Learning of Depth and Defocus Effects from Natural Images with Aperture Rendering Neural Radiance Fields)} is a novel extension of NeRF for unsupervised learning of depth and defocus effects from natural images. It utilizes an aperture rendering approach, representing each pixel with a collection of rays converging at the focus plane, allowing it to capture both viewpoint changes and defocus effects in a unified ray-tracing framework. The model optimizes an MLP to reflect these factors, enabling intuitive and continuous adjustment of defocus strength and focus distance. AR-NeRF's aperture randomized training disentangles defocus-aware and defocus-independent representations, learning to generate images while independently randomizing aperture size and latent codes. Applied to various natural image datasets, AR-NeRF demonstrates superior performance in depth prediction and defocus effect manipulation compared to baseline models, including AR-GAN and generative NeRFs.
    
    \item \textbf{SNeS (Learning Probably Symmetric Neural Surfaces from Incomplete Data)} is a method for 3D reconstruction of partly-symmetric objects using neural rendering techniques like NeRF. SNeS addresses the limitation of neural renderers in reconstructing unobserved parts of an object by utilizing structural priors such as symmetry. The method decomposes a neural renderer's color model into components like material albedo, reflectivity, and lighting, applying symmetry constraints to material-dependent components. SNeS is particularly effective in reconstructing vehicles, which often exhibit bilateral symmetry and highly-reflective materials. The approach allows for accurate reconstruction and high-quality novel view synthesis from limited and biased sets of views, overcoming challenges in realistic scenarios with restricted viewpoints.
    
    \item \textbf{RigNeRF (Fully Controllable Neural 3D Portraits)} is a system enabling full control of head pose and facial expressions in neural 3D portraits. RigNeRF uses a 3DMM to guide a deformable neural radiance field, allowing for the reanimation of a subject with arbitrary head poses and facial expressions. The method overcomes limitations of direct rendering 3DMMs by refining coarse deformation fields with residuals predicted by an MLP. This approach ensures rigidity during reanimation and allows for generalization to novel head poses and expressions not seen in the input video. RigNeRF can be trained on a short portrait video and enables free view synthesis of a portrait scene with explicit control over head pose and expression, capturing rich scene details and maintaining high fidelity in reanimated videos.
    
    \item \textbf{Physics Informed Neural Fields for Smoke Reconstruction with Sparse Data} is a method for high-fidelity reconstruction of dynamic fluids from sparse multiview RGB videos. The author's approach, leveraging physics-informed neural fields, overcomes challenges in fluid reconstruction due to complex lighting and occlusions. The method reconstructs fluid phenomena by integrating time-varying neural radiance fields with physics-informed deep learning, using neural networks to represent continuous spatio-temporal density, velocity, and radiance fields. It handles fluid scenes with unknown lighting and arbitrary obstacles, enabling fluid-obstacle interaction reconstruction without additional geometry input or human labeling. The method's robust optimization from sparse input views is achieved through a progressively growing model and a new regularization term, disentangling density-color ambiguity and avoiding overfitting. This approach demonstrates high-quality results in both synthetic and real flow captures, advancing fluid reconstruction in complex real-world scenes.
    
    \item \textbf{GRAM-HD (3D-Consistent Image Generation at High Resolution with Generative Radiance Manifolds)} a 3D-aware GAN method for generating high-resolution, 3D-consistent images. GRAM-HD leverages the Generative Radiance Manifold (GRAM) approach, applying 2D convolutions on 2D radiance manifolds for efficient super-resolution in 3D space. This method ensures multiview consistency by generating a high-resolution 3D representation for rendering while utilizing the computational efficiency of 2D CNNs. GRAM-HD can produce photorealistic images up to $1024\times1024$ resolution, significantly outperforming existing methods in terms of generation quality and speed. The approach represents a significant advancement in closing the gap between traditional 2D image generation and 3D-consistent free-view generation.
    
    \item \textbf{VoxGRAF (Fast 3D-Aware Image Synthesis with Sparse Voxel Grids)} is a method for 3D-aware image synthesis using sparse voxel grids. VoxGRAF addresses the computational inefficiency of querying an MLP for every sample along each ray in traditional 3D-aware generative models. By combining sparse voxel grids with progressive growing, free space pruning, and appropriate regularization, VoxGRAF replaces monolithic MLPs with 3D convolutions. The method disentangles the foreground object (modeled in 3D) from the background (modeled in 2D), allowing for scaling to higher voxel resolutions. Unlike previous approaches, VoxGRAF generates a full 3D scene in a single forward pass, enabling efficient rendering from arbitrary viewpoints while maintaining high visual fidelity and 3D consistency. This approach represents a significant advancement in 3D-aware generative modeling, offering a more efficient and consistent alternative to existing methods.
    
    \item \textbf{Neural Deformable Voxel Grid for Fast Optimization of Dynamic View Synthesis} proposes a method for synthesizing dynamic scenes using a fast deformable radiance field. The authors' approach, built on voxel-grid optimization, consists of two modules: a deformation grid for storing dynamic features and a light-weight MLP for decoding deformation, and a density and color grid for modeling scene geometry and appearance. The method explicitly models occlusion to improve rendering quality. Remarkably, it achieves comparable performance to existing dynamic NeRF methods but is over 70 times faster, requiring only 20 minutes for training. This efficiency makes it a practical solution for dynamic view synthesis, addressing the challenge of lengthy training procedures in dynamic NeRF methods.
    
    \item \textbf{Variable Bitrate Neural Fields} is a method for compressing feature grids in neural fields, significantly reducing memory consumption and enabling multiresolution representation for streaming. In this approach, the vector-quantized auto-decoder (VQ-AD), learns compressed feature-grids for signals without direct supervision, dynamically adapting to varying data complexity, bandwidth, and desired detail level. This method allows for progressive, variable bitrate streaming of data, achieving significant compression of neural radiance fields with minimal visual quality loss. The approach is particularly beneficial for graphics systems with tight memory, storage, and bandwidth constraints, offering a solution for efficiently handling high-resolution feature grids in neural field representations.
    
    \item \textbf{FWD (Real-time Novel View Synthesis with Forward Warping and Depth)} is a method for real-time NVS with sparse inputs. FWD uses explicit depth and differentiable rendering to achieve high-quality synthesis at real-time speeds. It can integrate sensor depth during training or inference to enhance image quality while maintaining speed. The method estimates depths for each input view to build a point cloud of latent features, then synthesizes novel views via a point cloud renderer. A view-dependent feature MLP and a Transformer-based fusion module are employed to effectively combine features from multiple inputs. FWD is trained end-to-end, learning depth and features optimized for synthesis quality. It outperforms existing methods in speed and quality, offering a practical solution for interactive applications requiring both real-time performance and high-quality NVS.
    
    \item \textbf{KiloNeuS (A Versatile Neural Implicit Surface Representation for Real-Time Rendering)} is a neural representation that reconstructs an implicit surface from multi-view images for real-time rendering. KiloNeuS partitions space into thousands of tiny MLPs, enabling fast inference. It learns the implicit surface locally using independent models, resulting in globally coherent geometry. The method is evaluated on a GPU-accelerated ray-caster with in-shader neural network inference, achieving an average of 46 FPS at high resolution. KiloNeuS outperforms its single-MLP counterpart in rendering quality and surface recovery. It also demonstrates versatility by integrating into an interactive path-tracer, taking full advantage of its surface normals. KiloNeuS represents a significant step towards real-time rendering of implicit neural representations under global illumination.
    
    \item \textbf{EventNeRF (Neural Radiance Fields from a Single Colour Event Camera)} is the first approach to create 3D-consistent, dense, and photorealistic novel view synthesis using only a single color event stream. EventNeRF is trained entirely in a self-supervised manner from events while preserving the original resolution of the colour event channels. The method includes a ray sampling strategy tailored to events for data-efficient training and produces RGB space results at unprecedented quality. EventNeRF outperforms existing methods in producing denser and more visually appealing renderings, especially in challenging scenarios with fast motion and low lighting conditions. The approach represents a significant advancement in utilizing event cameras for dense volumetric 3D reconstruction and novel view synthesis.
    
    \item \textbf{UNeRF (Time and Memory Conscious U-Shaped Network for Training Neural Radiance Fields)} is a method that reduces the high training times and memory requirements of NeRFs. UNeRF exploits the redundancy of NeRF's sample-based computations by partially sharing evaluations across neighboring sample points. Inspired by the UNet architecture, UNeRF reduces spatial resolution in the middle of the network and shares information between adjacent samples. This approach improves novel view synthesis, reduces memory footprint, and cuts down training time while maintaining or improving accuracy. UNeRF's design allows for improved resource utilization on various neural radiance fields tasks, including static scenes and dynamic human shape and motion.
    
    \item \textbf{Ev-NeRF (Event Based Neural Radiance Field)} is a NeRF derived from event data. Ev-NeRF effectively utilizes the high frame rate and sensitivity to brightness changes of event cameras, overcoming challenges in low lighting or extreme motion. It uses a self-supervised training approach from event measurements, creating an integrated neural volume that summarizes sparse data points. Ev-NeRF can produce intensity images from novel views with depth estimates, serving as high-quality input for various vision-based tasks. It demonstrates robust performance in challenging environments, offering solutions for high dynamic range imaging, noise reduction, depth estimation, and novel-view intensity image reconstruction. Ev-NeRF represents a significant advancement in combining NeRF formulation with raw event camera output, expanding the application area of event-based vision.
    
    \item \textbf{GNARF (Generative Neural Articulated Radiance Fields)} is a 3D GAN framework for generating editable radiance fields of human bodies and faces. GNARF learns to generate radiance fields in a canonical pose and warps them using an explicit deformation field into desired body poses or facial expressions. This approach enables high-quality generation of 3D human bodies and improves the quality of generated bodies or faces when editing poses or expressions. GNARF combines a tri-plane feature representation with an explicit feature volume deformation guided by a template shape. It demonstrates high-quality results for unconditional generation and animation of human bodies using the SURREAL and AIST++ datasets, and faces using the FFHQ dataset. GNARF represents a significant advancement in the generation of editable, multi-view-consistent human bodies and faces using 3D-aware GANs.
    
    \item \textbf{Regularization of NeRFs using differential geometry} introduces a method to regularize NeRFs using concepts from differential geometry. This approach provides elegant tools for robustly training NeRF-like models, modified to represent continuous and infinitely differentiable functions. The framework allows for regularizing different types of NeRF observations, improving performance in challenging conditions. It also encourages the regularity of surfaces through Gaussian or mean curvatures. This method outperforms previous state-of-the-art methods when training with limited input views, producing smoother and more accurate depth maps. The approach demonstrates the effectiveness of differential geometry in enhancing the robustness and quality of NeRF models.
    
    \item \textbf{Neural Rendering for Stereo 3D Reconstruction of Deformable Tissues in Robotic Surgery} discusses a novel framework for reconstructing deformable tissues in robotic surgery from endoscopic stereo videos. This framework uses dynamic neural radiance fields to represent deformable surgical scenes and optimizes shapes and deformations in a learning-based manner. To address challenges like tool occlusion and limited 3D clues from a single viewpoint, they introduced strategies such as tool mask-guided ray casting, stereo depth-cueing ray marching, and stereo depth-supervised optimization. Their method significantly outperforms current state-of-the-art reconstruction methods in handling complex non-rigid deformations. This is the first work leveraging neural rendering for surgical scene 3D reconstruction, demonstrating remarkable potential in medical applications.
    
    \item \textbf{Aug-NeRF (Training Stronger Neural Radiance Fields with Triple-Level Physically-Grounded Augmentations)} introduces robust data augmentations into NeRF training, targeting three levels: input coordinates, intermediate features, and pre-rendering output. These augmentations aim to simulate imprecise camera parameters, smoothen intrinsic feature manifolds, and account for potential degradation factors in multi-view image supervision. The method enhances NeRF's performance in novel view synthesis and geometry reconstruction, even in settings with heavily corrupted images.
    
    \item \textbf{LaTeRF (Label and Text Driven Object Radiance Fields)} extracts objects from scenes using 2D images, camera poses, textual descriptions, and point-labels. It extends NeRF by introducing an 'objectness' probability at each 3D point and leverages a pre-trained CLIP model for occlusion reasoning. This approach allows high-fidelity object extraction from complex scenes, outperforming previous methods in both synthetic and real-world datasets.
    
    \item \textbf{SNeRF (Stylized Neural Implicit Representations for 3D Scenes)} proposes a method for stylizing 3D scenes using NeRF and image-based neural style transfer. It addresses the memory constraints of NeRF by alternating between NeRF and stylization optimization steps, enabling high-resolution results on a single GPU. SNeRF produces high-quality stylized novel views with cross-view consistency for various scene types, including indoor, outdoor, and dynamic avatars.
    
    \item \textbf{VMRF (View Matching Neural Radiance Fields)} presents a novel view matching scheme for NeRF training without requiring prior knowledge of camera poses. It utilizes unbalanced optimal transport for feature mapping between rendered and real images, and a pose calibration technique for adjusting initially randomized camera poses. VMRF demonstrates superior performance in novel view synthesis across synthetic and real datasets.
    
    \item \textbf{A Learned Radiance-Field Representation for Complex Luminaires} introduces a method for rendering complex luminaires using a neural radiance field (NeRF) and a specialized loss function. It reduces the geometric complexity of luminaires while encoding the emitted light field, integrating the luminaire's NeRF into traditional rendering systems. The approach offers significant speed-ups in rendering scenes with complex luminaires.
    
    \item \textbf{Vision Transformer for NeRF-Based View Synthesis from a Single Input Image} presents a method combining vision transformers (ViT) and NeRF for synthesizing novel views from a single unposed image. It leverages global features from ViT and local features from a 2D CNN, trained with a multi-layer perceptron for volume rendering. This approach renders novel views with more accurate structure and finer details, outperforming existing methods on various datasets.
    
    \item \textbf{Neural apparent BRDF fields for multiview photometric stereo} extends NeRF for multiview photometric stereo by predicting surface normals and decomposing appearance into a neural BRDF and shadow prediction network. It balances learned components with physical image formation models, allowing extrapolation far from observed light and viewer directions. The method demonstrates competitive performance in shape and material estimation from multiview photometric stereo data.
    
    \item \textbf{NDF (Neural Deformable Fields for Dynamic Human Modelling)} is a new representation for dynamic human modeling from multi-view videos. NDF, wrapped around a parametric body model, captures dynamic changes and detailed geometry movements. This method significantly improves the realism in synthesizing digitized performers with novel views and poses, outperforming recent human synthesis methods.
    
    \item \textbf{AdaNeRF (Adaptive Sampling for Real-time Rendering of Neural Radiance Fields)} is a dual-network architecture for efficient rendering of NeRFs. The authors focused on reducing sample points through a novel training scheme that introduces sparsity, achieving real-time rendering with high quality at low sample counts. This approach offers a significant improvement in frame rate and quality over existing compact neural representations, particularly beneficial for applications in computer graphics and vision.
    
    \item \textbf{Generalizable Patch-Based Neural Rendering} proposes a novel paradigm for novel-view synthesis of unseen scenes, diverging from the reliance on deep convolutional features and NeRF-like models. This method predicts target ray colors directly from scene patches, using transformers and epipolar geometry. This approach, independent of the reference frame, outperforms state-of-the-art methods in novel view synthesis, even with less training data, demonstrating potential for broader applications in neural rendering.
    
    \item \textbf{Neural-Sim (Learning to Generate Training Data with NeRF)} is  a differentiable synthetic data pipeline using Neural Radiance Fields (NeRFs) for on-demand data generation. This method optimizes neural rendering parameters to maximize accuracy for target tasks, addressing the challenge of collecting diverse training data. Demonstrated on synthetic and real-world object detection tasks, Neural-Sim shows promise in improving model training efficiency and effectiveness, particularly in business scenarios for new product detection.
    
    \item \textbf{PS-NeRF (Neural Inverse Rendering for Multi-view Photometric Stereo)} is a method for multi-view photometric stereo that jointly estimates geometry, materials, and lighting of non-Lambertian objects. Utilizing a shadow-aware differentiable rendering layer, PS-NeRF optimizes surface normals and BRDFs, enabling novel-view rendering and relighting. This approach significantly enhances shape reconstruction accuracy compared to existing methods, showcasing its potential in advanced 3D reconstruction and rendering applications.
    
    \item \textbf{Learning Generalizable Light Field Networks from Few Images} explored a few-shot novel view synthesis using a neural light field network. This method, leveraging 3D ConvNets and implicit neural networks, efficiently generalizes across scenes and viewpoints. It outperforms existing convolutional methods in synthetic and real multi-view stereo data, offering a faster rendering solution for applications requiring rapid novel view generation.
    
    \item \textbf{Is Attention All That NeRF Needs?} by Varma T et al. introduced the Generalizable NeRF Transformer (GNT), a transformer-based architecture for reconstructing NeRFs and rendering novel views. GNT uses transformers for multi-view image feature fusion and sampling-based rendering integration, achieving state-of-the-art performance in complex scene generalization. This method demonstrates the versatility of transformers in graphical rendering, potentially revolutionizing view synthesis and 3D scene modeling.
    
    \item \textbf{Neural Radiance Transfer Fields for Relightable Novel-view Synthesis with Global Illumination} by Lyu et al combines Computer Vision and Graphics to propose a method for scene relighting under novel views. Their approach learns a neural precomputed radiance transfer function, handling global illumination effects and disentangling scene parameters. This method, trained on real images under a single unknown lighting condition, significantly advances in scene parameter recovery and synthesis quality, showing potential for applications in virtual reality and controllable image synthesis.
    
    \item \textbf{GAUDI (A Neural Architect for Immersive 3D Scene Generation)} is a generative model for creating complex 3D scenes, capable of rendering from moving cameras. GAUDI disentangles radiance fields and camera poses in its latent representation, enabling both unconditional and conditional generation of 3D scenes. This model shows state-of-the-art performance in generating scenes from sparse observations or text descriptions, promising advancements in fields like reinforcement learning, SLAM, and 3D content creation.
    
    \item \textbf{NeDDF (Neural Density-Distance Fields)} is a novel 3D representation that combines distance and density fields for visual localization tasks like SLAM. NeDDF extends the distance field formulation to shapes without explicit boundaries, such as fur or smoke, enabling explicit conversion from distance to density fields. This method achieves robust localization and high-quality registration, outperforming NeRF in localization tasks while providing comparable results in novel view synthesis.
    
    \item \textbf{NeRFA (End-to-end View Synthesis via NeRF Attention)} is a seq2seq formulation for view synthesis using ray points as input and outputting corresponding colors. NeRFA, inspired by NeRF, addresses limitations of standard attention in volumetric rendering and inefficiency in global attention. It performs multi-stage attention for computational efficiency and uses ray and pixel transformers for learning interactions, demonstrating superior performance over NeRF and NerFormer in various datasets.
    
    \item \textbf{Distilled Low Rank Neural Radiance Field with Quantization for Light Field Compression} proposes QDLR-NeRF for light field compression. This method learns an implicit scene representation as a NeRF with a Low-Rank (LR) constraint using Tensor Train (TT) decomposition. It further reduces model size through network distillation and optimized quantization of TT components. QDLR-NeRF shows better compression efficiency compared to state-of-the-art methods and allows high-quality synthesis of any light field view.
    
    \item \textbf{MobileNeRF (Exploiting the Polygon Rasterization Pipeline for Efficient Neural Field Rendering on Mobile Architectures)} is a NeRF representation based on textured polygons for efficient neural field rendering on mobile architectures. It uses traditional rendering with a z-buffer to yield an image with features at every pixel, interpreted by a small MLP in a fragment shader. MobileNeRF achieves interactive frame rates on mobile phones and other devices, offering a significant speed advantage over SNeRG while consuming less memory.
    
    \item \textbf{DoF-NeRF (Depth-of-Field Meets Neural Radiance Fields)} is a NeRF-based framework for simulating DoF effects in neural radiance fields. It extends NeRF to simulate lens aperture following geometric optics principles, allowing operation on views with different focus configurations and direct manipulation of DoF effects. DoF-NeRF can synthesize all-in-focus novel views from shallow DoF inputs and demonstrates an application in DoF rendering.
    
    \item \textbf{360Roam (Real-Time Indoor Roaming Using Geometry-Aware 360)} is a real-time indoor roaming system using geometry-aware 360° radiance fields. It employs adaptive-assigned neural perceptrons for rendering novel views on a single GPU and includes a floorplan output for enhanced immersive experience. 360Roam demonstrates superior performance in large-scale indoor scene roaming compared to baseline approaches.
    
    \item \textbf{Cascaded and Generalizable Neural Radiance Fields for Fast View Synthesis} introduces CG-NeRF, a method combining a coarse radiance fields predictor and a convolutional-based neural renderer for fast view synthesis. CG-NeRF infers consistent scene geometry based on implicit neural fields and renders new views efficiently on a single GPU. It outperforms state-of-the-art generalizable neural rendering methods on various datasets and can be fine-tuned quickly for specific scenes.
    
    \item \textbf{FDNeRF (Few-shot Dynamic Neural Radiance Fields for Face Reconstruction and Expression Editing)} proposes FDNeRF for 3D face reconstruction and expression editing from few-shot dynamic frames. FDNeRF uses a conditional feature warping module for expression-conditioned warping in 2D feature space, adapting to different identities and constrained by 3D radiance fields. It enables novel view synthesis, expression editing, and video-driven reenactment tasks, outperforming existing dynamic and few-shot NeRFs in both reconstruction and editing.
    
    \item \textbf{OmniVoxel (A Fast and Precise Reconstruction Method of Omnidirectional Neural Radiance Field)} is a method for reconstructing neural radiance fields using equirectangular omnidirectional images. The authors introduced feature voxels and spherical voxelization to balance reconstruction quality for inner and outer scenes. The method significantly reduced training time (20-40 minutes per scene) compared to traditional methods (15-20 hours per scene) and showed promising results on synthetic and real datasets.
    
    \item \textbf{Progressive Multi-scale Light Field Networks} discusses a progressive multi-scale light field network that encodes light fields at multiple levels of detail. This approach allows for progressive streaming, reducing rendering time and addressing aliasing at lower scales. The method supports per-pixel level of detail, enabling smooth transitions and foveated rendering, making it suitable for on-demand streaming applications.
    
    \item \textbf{Fast Learning Radiance Fields by Shooting Much Fewer Rays} introduced a strategy to accelerate the learning of radiance fields by shooting fewer rays in multi-view volume rendering. This approach utilized quadtree subdivision to dynamically focus on complex regions, significantly reducing training time while maintaining accuracy. This method demonstrated compatibility with various radiance field-based methods and showed efficiency in different scenes.
    
    \item \textbf{UPST-NeRF (Universal Photorealistic Style Transfer of Neural Radiance Fields for 3D Scene)} is a framework for photorealistic style transfer in 3D scenes using neural radiance fields. The method first pre-trains a 2D style transfer network, then optimizes a 3D scene using voxel features, and finally employs a hypernetwork for style transfer. This approach allows for consistent and photorealistic stylization of 3D scenes from various viewpoints.
    
    \item \textbf{DM-NeRF (3D Scene Geometry Decomposition and Manipulation from 2D Images)} is a method for decomposing and manipulating 3D scene geometry from 2D views. The approach uses an object field component to learn unique codes for individual objects in 3D space and an inverse query algorithm for object manipulation. DM-NeRF can reconstruct, decompose, manipulate, and render complex 3D scenes in a single pipeline.
    
    \item \textbf{PanoHDR-NeRF (Casual Indoor HDR Radiance Capture from Omnidirectional Images)} is a neural representation for capturing full HDR radiance fields of indoor scenes using casually captured LDR omnidirectional videos. The method involves uplifting LDR frames to HDR, training a tailored NeRF++ model, and rendering full HDR images from any scene location. PanoHDR-NeRF predicts plausible HDR radiance and synthesizes correct lighting effects for scene augmentation.
    
    \item \textbf{Neural Capture of Animatable 3D Human from Monocular Video} discusses a method for building an animatable 3D human representation from monocular video input. The approach uses a dynamic Neural Radiance Field (NeRF) rigged by a mesh-based parametric 3D human model. The method focuses on generalization to unseen poses and views, embedding input queries with local surface region relationships, and jointly optimizing NeRF and refined per-frame human mesh.
    
    \item \textbf{E-NeRF (Neural Radiance Fields from a Moving Event Camera)} is the first method to estimate a volumetric scene representation in the form of a NeRF from a fast-moving event camera. The approach can recover NeRFs during fast motion and high-dynamic-range conditions, rendering high-quality frames using only an event stream as input. E-NeRF combines events and frames to estimate NeRFs of higher quality than state-of-the-art approaches under severe motion blur.
    
    \item \textbf{PeRFception (Perception using Radiance Fields)} is a dataset for perception tasks using Plenoxels, a variant of NeRFs. This dataset, comprising object-centric and scene-centric scans, enables classification and segmentation tasks in both 2D and 3D formats. The authors demonstrated that this approach significantly compresses memory while retaining detailed information, thus facilitating efficient perception research.
    
    \item \textbf{Training and Tuning Generative Neural Radiance Fields for Attribute-Conditional 3D-Aware Face Generation} discusses conditional Generative Neural Radiance Field (GNeRF) model for 3D-aware face generation, focusing on attribute control and disentanglement. The authors introduced the TRIOT method, combining training and optimization techniques to enhance facial attribute editing. Their experiments showed that this model could produce high-quality, view-consistent edits while preserving non-target regions.
    
    \item \textbf{A Portable Multiscopic Camera for Novel View and Time Synthesis in Dynamic Scenes} presented a portable multiscopic camera system, designed for dynamic scene rendering from various viewpoints and times. The system, integrating a physical camera with a NeRF model, demonstrated superior performance in both real-world and synthetic datasets, outperforming existing solutions in novel view and time synthesis tasks.
    
    \item \textbf{Dual-Space NeRF (Learning Animatable Avatars and Scene Lighting in Separate Spaces)} is a method that separates the learning of scene lighting and human body animation using 2 MLPs in different spaces. They introduced barycentric mapping for pose-independent position representation, achieving better results in animating human avatars under various lighting conditions. Their approach showed effectiveness on datasets like Human3.6M and ZJU-MoCap.
    
    \item \textbf{Cross-Spectral Neural Radiance Fields} introduced X-NeRF, a method for learning Cross-Spectral scene representations using Neural Radiance Fields. X-NeRF optimizes camera poses across spectra and employs Normalized Cross-Device Coordinates (NXDC) for rendering images from different modalities. The method was validated on scenes featuring color, multi-spectral, and infrared images, demonstrating its effectiveness in modeling Cross-Spectral representations.
    
    \item \textbf{On Quantizing Implicit Neural Representations} explores the impact of quantization on implicit neural networks. The authors showed that non-uniform quantization, specifically clustered quantization, significantly improves performance at lower quantization levels. Their findings suggest that high-fidelity reconstructions are achievable even with binary neural networks, leading to substantial compression of neural radiance fields.
    
    \item \textbf{CLONeR (Camera-Lidar Fusion for Occupancy Grid-aided Neural Representations)} is a NeRF framework that integrates camera and LiDAR data for outdoor scenes. CLONeR uses separate MLPs for learning occupancy from LiDAR data and color from camera images. The method includes a differentiable 3D Occupancy Grid Map, enhancing the sampling efficiency for large outdoor scenes. Tested on the KITTI dataset, CLONeR outperformed existing NeRF models in novel view synthesis and depth prediction tasks.
    
    \item \textbf{Volume Rendering Digest (for NeRF)} provided a comprehensive overview of volume rendering techniques in the context of NeRF. The authors discussed the probabilistic interpretation of transmittance and opacity, and how these concepts are applied in volume rendering to calculate expected light emission from particles in a scene. The report also covered the handling of homogeneous media and
    
    \item \textbf{Neural Feature Fusion Fields (N3F) (3D Distillation of Self-Supervised 2D Image Representations)} is a method enhancing 2D image feature extractors for 3D scene analysis. N3F uses a self-supervised feature extractor as a teacher to train a 3D student network, distilling features into a 3D representation via neural rendering. This approach, applicable to various neural rendering formulations, improves semantic understanding in scene-specific neural fields and enhances tasks like object retrieval and segmentation, particularly in dynamic scenes like egocentric videos.
    
    \item \textbf{im2nerf (Image to Neural Radiance Field in the Wild)} is a learning framework for predicting a continuous neural object representation from a single image in the wild. The model encodes the input image into disentangled object representations (shape, appearance, camera pose) and conditions a NeRF on these representations. The framework uses volume rendering for novel views and employs an adversarial loss, object symmetry, and cycle camera pose consistency to address the under-constrained problem of single-view image-based 3D reconstruction.
    
    \item \textbf{Generative Deformable Radiance Fields for Disentangled Image Synthesis of Topology-Varying Objects} by Wang et al. discusses a generative model for synthesizing radiance fields of topology-varying objects with disentangled shape and appearance variations. The method generates deformable radiance fields, building dense correspondence between the density fields of objects and encoding their appearances in a shared template field. This unsupervised approach allows for high-quality disentanglement and effective image editing, showing potential in applications like VR and AR.
    
    \item \textbf{StructNeRF (Neural Radiance Fields for Indoor Scenes with Structural Hints)} is a solution for novel view synthesis in indoor scenes with sparse inputs. StructNeRF leverages structural hints in multi-view inputs, addressing the under-constrained geometry issue in NeRF. It uses a patch-based multi-view consistent photometric loss for textured regions and restricts non-textured regions to 3D consistent planes. This method improves both geometry and view synthesis performance of NeRF without additional training data.
    
    \item \textbf{3DMM-RF (Convolutional Radiance Fields for 3D Face Modeling)} is a facial 3D Morphable Model using convolutional radiance fields. The model, combining deep generative networks and neural radiance fields, accurately models a subject's identity, pose, and expression, and renders it in arbitrary illumination. The approach overcomes the rigidity and rendering speed limitations of traditional NeRF, offering potential in applications like avatar creation and virtual makeup.

    \item \textbf{Uncertainty Guided Policy for Active Robotic 3D Reconstruction using Neural Radiance Fields} tackles the problem of active robotic 3D reconstruction using NeRF-based object representation. They introduced a ray-based volumetric uncertainty estimator to infer the uncertainty of the underlying 3D geometry and guide the next-best-view selection policy. This approach, distinct from methods relying on explicit 3D geometric modeling, shows promise in robot vision applications.
    
    \item \textbf{LATITUDE (Robotic Global Localization with Truncated Dynamic Low-pass Filter in City-scale NeRF)} is a method for robotic global localization in city-scale NeRF. The two-stage mechanism includes a place recognition stage using a regressor trained on images from NeRFs, and a pose optimization stage using a Truncated Dynamic Low-pass Filter (TDLF) for coarse-to-fine pose registration. This method shows potential for high-precision navigation in large-scale city scenes.
    
    \item \textbf{ActiveNeRF (Learning where to See with Uncertainty Estimation)} is  a learning framework that incorporates uncertainty estimation into NeRF for efficient 3D scene modeling with a constrained input budget. ActiveNeRF selects new training samples based on an active learning scheme, evaluating the reduction of uncertainty and information gain. This approach improves the quality of novel view synthesis with minimal additional resources, showing effectiveness in both realistic and synthetic scenes.
    
    \item In \textbf{Density-aware NeRF Ensembles: Quantifying Predictive Uncertainty in Neural Radiance Fields}, the authors introduced a novel approach to quantify uncertainty in NeRFs by incorporating density-aware ensembles. They demonstrated that considering termination probabilities along rays, in addition to RGB variance, effectively identifies epistemic uncertainty in unobserved scene parts. This method outperformed existing techniques in uncertainty quantification benchmarks for NeRFs and was applied for next-best view selection and model refinement in robotics.
    
    \item \textbf{NeRF-SOS (Any-View Self-supervised Object Segmentation on Complex Scenes)} is a self-supervised framework for object segmentation in complex real-world scenes using Neural Radiance Fields (NeRF). The authors employed a collaborative contrastive loss at both appearance and geometry levels, leveraging pre-trained 2D visual features and NeRF's density fields. NeRF-SOS significantly surpassed existing methods in segmentation accuracy, demonstrating its effectiveness in both indoor and outdoor scenarios.
    
    \item \textbf{Loc-NeRF (Monte Carlo Localization using Neural Radiance Fields)} is a real-time vision-based robot localization method, combines Monte Carlo localization with NeRFs. It uses a pre-trained NeRF model as an environmental map and an RGB camera for localization, overcoming the limitations of existing NeRF-based localization methods which require good initial pose guesses and substantial computation. Loc-NeRF showed promising results in real-time global localization and tracking with real-world data.
    
    \item \textbf{NeRF-Loc (Transformer-Based Object Localization Within Neural Radiance Fields)} is a transformer-based framework for localizing objects within Neural Radiance Fields. It uses a pre-trained NeRF model and camera views to produce 3D bounding boxes of objects. The framework includes coarse and fine transformer encoder branches for encoding context and details, respectively. NeRF-Loc demonstrated superior performance over conventional RGB(-D) based methods in object localization tasks.
    
    \item \textbf{Enforcing safety for vision-based controllers via Control Barrier Functions and Neural Radiance Fields} tackled the challenge of ensuring safety in vision-based robotic controllers using Control Barrier Functions (CBFs) and NeRFs. The authors developed a novel CBF-based controller that uses NeRFs for single-step visual foresight, filtering out unsafe actions and maintaining safety in real-time simulation experiments.
    
    \item \textbf{Baking in the Feature: Accelerating Volumetric Segmentation by Rendering Feature Maps} proposed a method to accelerate 3D volumetric segmentation using NeRFs by rendering feature maps. The authors used features extracted from large datasets to improve segmentation performance, demonstrating higher accuracy with fewer semantic annotations compared to existing methods.
    
    \item \textbf{WaterNeRF (Neural Radiance Fields for Underwater Scenes)} is a method leveraging Neural Radiance Fields for underwater scene reconstruction and image restoration. It estimates parameters of a physics-based underwater image formation model and uses these for novel view synthesis, enabling image restoration and dense depth estimation. The method was evaluated on a real underwater dataset, showing promising results in restoring underwater imagery.
    
    \item \textbf{OmniNeRF (Hybriding Omnidirectional Distance and Radiance fields for Neural Surface Reconstruction)} addresses surface ambiguity in 3D reconstruction by proposing a hybrid implicit field combining Omnidirectional Distance Field (ODF) and NeRF. It focuses on reconstructing high-quality surfaces, especially at edges, by eliminating errors inherent in distance field training. OmniNeRF demonstrated improved accuracy in reconstructing 3D scene edges compared to existing methods.
    
    \item \textbf{360FusionNeRF (Panoramic Neural Radiance Fields with Joint Guidance)} is a semi-supervised learning framework for synthesizing novel views from a single 360° panorama image using NeRF. The authors incorporated geometric supervision and semantic consistency to guide the training process, improving the geometry of synthesized views. The method leverages depth supervision and semantic features from a pre-trained visual encoder like CLIP, enhancing realistic renderings of novel views. This approach showed state-of-the-art performance in various datasets, suggesting applications in VR photography and scene understanding.
    
    \item \textbf{SymmNeRF (Learning to Explore Symmetry Prior for Single-View View Synthesis)} is a NeRF-based framework that exploits symmetry priors for single-view view synthesis. By incorporating pixel-aligned image features and symmetric features into NeRF, the method enhances the recovery of occluded geometry and texture details. SymmNeRF, being scene-independent, generalizes well to unseen objects. This approach is particularly effective for synthesizing novel views with more details in man-made objects, demonstrating potential in 3D reconstruction and rendering applications. 
    
    \item \textbf{DreamFusion (Text-to-3D using 2D Diffusion)} is a method for text-to-3D synthesis using a pretrained 2D text-to-image diffusion model. The authors introduced a loss based on probability density distillation, optimizing a 3D model (NeRF) via gradient descent to achieve low loss in 2D renderings from random angles. This unsupervised approach requires no 3D training data or modifications to the image diffusion model, enabling the creation of 3D models from text prompts for diverse applications like simulators and digital media.
    
    \item \textbf{TT-NF (Tensor Train Neural Fields)} is a novel low-rank representation for learning neural fields on dense regular grids. The authors demonstrated the efficiency of TT-NF in tensor denoising and applied it to NeRF, discovering low-rank structures through learning. The method offers a compact and interpretable representation, showing potential in applications requiring efficient and scalable neural field representations.
    
    \item \textbf{Improving 3D-aware Image Synthesis with A Geometry-aware Discriminator} is a geometry-aware discriminator for improving 3D-aware GANs. GeoD enhances the generator's ability to produce accurate 3D shapes by providing geometry information from the discriminator. This method demonstrated superiority over state-of-the-art alternatives in various generator architectures and datasets, suggesting applications in photo-realistic 2D image synthesis and 3D shape generation.
    
    \item \textbf{Structure-Aware NeRF without Posed Camera via Epipolar Constraint} authored by Chen et al, presented an approach integrating pose extraction and view synthesis into a single end-to-end procedure for NeRF models. They utilized the epipolar constraint for camera pose optimization, enabling NeRF to be aware of the scene's structure without pre-known camera poses. This method improves the generalization performance of NeRF, showing effectiveness in various scenes and potential applications in 3D reconstruction and virtual reality.
    
    \item \textbf{NeRF (Neural Radiance Field in 3D Vision, A Comprehensive Review)} provides a comprehensive survey of NeRF papers, focusing on architecture and application-based taxonomies. It introduced the theory of NeRF and its training via differentiable volume rendering, benchmarking key models' performance and speed. This survey aims to introduce NeRF to new researchers, reference influential works, and motivate future research directions in fields like robotics, urban mapping, and virtual/augmented reality.
    
    \item \textbf{Unsupervised Multi-View Object Segmentation Using Radiance Field Propagation} discusses Radiance Field Propagation (RFP), an unsupervised approach for segmenting objects in 3D using multi-view images. RFP, based on NeRF, segments scenes into salient regions without supervision, annotations, or prior knowledge of object classes. The method enables individual 3D object editing operations and shows potential in applications requiring unsupervised image/scene segmentation and 3D scene editing.
    
    \item \textbf{IntrinsicNeRF (Learning Intrinsic Neural Radiance Fields for Editable Novel View Synthesis)} is a method that integrates intrinsic decomposition into NeRF for room-scale scenes. The authors proposed a novel distance-aware point sampling and adaptive reflectance iterative clustering optimization method. This approach allows for unsupervised training, resulting in consistent intrinsic decomposition and high-fidelity novel view synthesis, supporting applications like scene recoloring and illumination variation.
    
    \item \textbf{NARF22 (Neural Articulated Radiance Fields for Configuration-Aware Rendering)} is a pipeline using a configuration-parameterized NeRF for rendering articulated objects. The authors employed a two-stage parts-based training mechanism, enabling the model to generalize well across configurations with minimal training data. NARF22 demonstrates effectiveness in tasks like 6 DoF pose and configuration estimation from monocular videos, addressing scalability issues in robotic perception of articulated objects.
    
    \item \textbf{Neural Implicit Surface Reconstruction from Noisy Camera Observations} proposes a method for learning 3D surfaces from noisy camera parameters, leveraging neural implicit surfaces. In this approach approach, unlike traditional methods, can handle significant noise in camera parameters, achieving consistent intrinsic decomposition and high-quality 3D surface reconstruction. This method broadens the applications of neural implicit surface-based object reconstruction by reducing the need for precise camera calibration.
    
    \item \textbf{SelfNeRF (Fast Training NeRF for Human from Monocular Self-rotating Video)} introduces a method for synthesizing high-fidelity novel views of human performance using monocular self-rotating videos. By employing a surface-relative hash encoding and aggregating inter-frame information, SelfNeRF significantly speeds up the training process, making it practical for ordinary users. The method demonstrates effectiveness in dynamic neural radiance field reconstruction of humans in minutes.
    
    \item \textbf{Capturing and Animation of Body and Clothing from Monocular Video} is developed to combine a mesh-based body model with a NeRF to capture and animate clothed human avatars from monocular videos. SCARF enables detailed control over body poses, including facial expressions and hand articulations, and supports clothing transfer between avatars. This hybrid approach addresses the limitations of previous methods in capturing and animating detailed clothing and body movements.
    
    \item \textbf{ViewFool (Evaluating the Robustness of Visual Recognition to Adversarial Viewpoints)} is a method to generate adversarial viewpoints that mislead visual recognition models. By representing objects as NeRF, ViewFool identifies diverse adversarial viewpoints, revealing the vulnerability of common image classifiers to viewpoint changes. The authors also introduce ImageNet-V, a new dataset for benchmarking viewpoint robustness, highlighting the need for improved robustness in visual recognition models.
    
    \item \textbf{Towards Efficient Neural Scene Graphs by Learning Consistency Fields} proposed Consistency-Field-based NSG (CF-NSG), a method to improve the efficiency of NSG for dynamic scenes. CF-NSG leverages redundancy across video frames by reusing features and identifying consistent visual components. This approach significantly reduces computational overhead without compromising rendering quality, enabling more efficient scene manipulation and novel view synthesis.
    
    \item \textbf{Estimating Neural Reflectance Field from Radiance Field using Tree Structures} by Li et al discusses a method for estimating NReF from posed multi-view images under unknown lighting. They used NeRF as a proxy and employed tree structures for efficient computation. The method allows for high-quality free-view relighting and material editing, addressing challenges in extracting geometry information and resolving ambiguities in NReF estimation.
    
    \item \textbf{Robustifying the Multi-Scale Representation of Neural Radiance Fields} introduced a robust multi-scale NeRF representation, addressing the limitations of NeRF in handling multi-scale images and camera pose estimation errors. This method combines Mip-NeRF's multi-scale representation with a graph-neural network-based motion averaging for robust camera pose estimation. This approach significantly improves object representation from multi-view images, particularly beneficial for images captured with everyday cameras.
    
    \item \textbf{SiNeRF (Sinusoidal Neural Radiance Fields for Joint Pose Estimation and Scene Reconstruction)} is a sinusoidal NeRF model, to enhance NeRFmm's joint optimization for scene reconstruction and pose estimation. The authors introduced Mixed Region Sampling for efficient ray batch selection and sinusoidal activations for radiance mapping. SiNeRF outperforms NeRFmm in image synthesis quality and pose estimation accuracy, demonstrating its effectiveness in challenging scenes.
    
    \item \textbf{NeRF2Real (Sim2real Transfer of Vision-guided Bipedal Motion Skills using Neural Radiance Fields)} discusses a system for sim2real transfer of bipedal motion skills using NeRF. The authors combined NeRF-rendered static scenes with dynamic object simulations to train vision-based navigation and ball pushing policies for humanoid robots. This approach successfully transfers learned skills to real robots, demonstrating the potential of NeRF in robotics for simulating complex, visually-rich environments.
    
    \item \textbf{X-NeRF (Explicit Neural Radiance Field for Multi-Scene 360{\textdegree} Insufficient RGB-D Views)} is an explicit neural radiance field model for representing multiple scenes with insufficient 360{\textdegree} RGB-D views. X-NeRF transforms sparse RGB-D inputs into a complete radiance field using a 3D sparse generative CNN, enabling efficient volumetric rendering. This method outperforms implicit approaches in challenging settings, indicating its practicality for diverse scene representation with sparse inputs.
    
    \item \textbf{CoRFs (Controllable Radiance Fields for Dynamic Face Synthesis)} was developed for dynamic face synthesis. CoRFs integrate motion features within a style-based generator and incorporate face parsing, head regression, and identity encoding for consistent attribute representation. This method enables 3D-aware synthesis of facial dynamics with identity and motion control, offering advancements in digital content creation and VR applications.
    
    \item \textbf{Reconstructing Personalized Semantic Facial NeRF Models From Monocular Video} proposed a semantic model for human heads using neural radiance fields, capable of representing complex facial attributes. This method constructs personalized models from monocular videos, leveraging multi-level voxel fields for efficient training and rendering. It offers significant improvements in representing facial dynamics and detailed textures, applicable in AR/VR and digital entertainment.
    
    \item \textbf{GraspNeRF (Multiview-based 6-DoF Grasp Detection for Transparent and Specular Objects Using Generalizable NeRF)} is a multiview RGB-based 6-DoF grasp detection network for transparent and specular objects. Utilizing generalizable NeRF, GraspNeRF performs material-agnostic object grasping in cluttered environments. The method is trained on a large-scale synthetic dataset, demonstrating significant improvements in grasp success rates in real-world applications.
    
    \item \textbf{MonoNeRF (Learning Generalizable NeRFs from Monocular Videos without Camera Pose)} is a generalizable NeRF model trained on monocular videos without camera pose annotations. MonoNeRF uses an autoencoder-based architecture for depth and camera pose estimation, enabling applications in depth estimation, novel view synthesis, and camera pose estimation. This approach significantly improves performance in various indoor scene datasets, showcasing its versatility in learning neural radiance fields without camera ground truths.
    
    \item \textbf{IBL-NeRF (Image-Based Lighting Formulation of Neural Radiance Fields)} is a novel approach to decompose NeRF for large-scale indoor scenes into intrinsic components. This method extends the original NeRF formulation by capturing spatial variations in lighting within the scene volume, in addition to surface properties. The technique involves optimizing a prefiltered radiance field and estimating material reflectance properties, lighting information, and geometry from multi-view images. The method demonstrated superior visual quality and multi-view consistency in synthesized images, particularly in complex object layouts and light configurations.
    
    \item \textbf{SPIDR (SDF-based Neural Point Fields for Illumination and Deformation)}, a hybrid neural SDF representation, was developed to enable fine-grained geometry deformation in NeRFs and to handle illumination changes during object deformations. The method combines point cloud and neural implicit representations for better editability and uses a coordinate-based MLP for more accurate environment light estimation. The authors employed shadow mapping for efficient light visibility approximation post-deformation. SPIDR showed improved quality in reconstructed object surfaces and accurate estimations of environment light, enabling rendering of deformed scenes with updated illumination and shadowing effects.
    
    \item \textbf{Differentiable Physics Simulation of Dynamics-Augmented Neural Objects} presented a differentiable pipeline for simulating the motion of objects represented as continuous density fields, including NeRFs. The authors estimated dynamical properties like mass and inertia from the density field and introduced a differentiable contact model for collision forces. The method allows robots to build visually and dynamically accurate object models from images and videos. Demonstrations included learning friction coefficients from real videos and synthetic data, and optimizing robot arm trajectories for manipulating neural objects.
    
    \item \textbf{ARAH (Animatable Volume Rendering of Articulated Human SDFs)}, a method for creating animatable clothed human avatars with detailed geometry that generalizes well to unseen poses. By combining an articulated implicit surface representation with volume rendering, the method overcomes limitations in geometry detail and pose generalization. A novel joint root-finding algorithm was introduced for efficient point sampling and accurate point canonicalization. ARAH demonstrated state-of-the-art performance in geometry and appearance reconstruction, creating animatable avatars that adapt to a wide range of poses.
    
    \item \textbf{Parallel Inversion of Neural Radiance Fields for Robust Pose Estimation} introduced a parallelized optimization method based on NeRF for estimating 6-DoF camera pose from a single RGB image. The method integrates a momentum-based camera extrinsic optimization procedure into a fast NeRF implementation and employs parallel Monte Carlo sampling to overcome local minima. A robust pixel-based loss function was adopted to improve accuracy. The method demonstrated improved generalization and robustness in pose estimation on both synthetic and real-world benchmarks.
    
    \item \textbf{Coordinates Are NOT Lonely -- Codebook Prior Helps Implicit Neural 3D Representations} presented CoCo-INR, a novel coordinate-based model for implicit neural 3D representation, which incorporates codebook prior information. The method uses two attention modules: codebook attention to extract useful prototypes from the prior codebook and coordinate attention to enrich each coordinate's feature representation. This approach allows rendering 3D views with more photo-realistic appearance and geometries using fewer calibrated images, demonstrating robustness under sparse input views and fine detail-preserving capability.
    
    \item \textbf{HDHumans (A Hybrid Approach for High-fidelity Digital Humans)} proposed a method for high-resolution human character synthesis integrating a classical deforming character template with NeRF. This hybrid approach captures detailed deforming surfaces, including dynamic clothing, and synthesizes photo-realistic images of novel views and motions. It outperforms existing methods in synthesis quality, resolution, and 3D surface reconstruction.
    
    \item \textbf{One-Shot Neural Fields for 3D Object Understanding} presented a unified scene representation for robotics, where each object is depicted by a latent code capturing geometry and appearance. This representation, decoded from a single RGB image, facilitates tasks like novel view rendering, 3D reconstruction, collision checking, and stable grasp prediction. Their approach demonstrates robustness in robotic grasping using a compact representation pre-trained on synthetic data.
    
    \item \textbf{An Exploration of Neural Radiance Field Scene Reconstruction: Synthetic, Real-world and Dynamic Scenes} explored 3D scene reconstruction using NeRF approaches, focusing on synthetic and real-world scenes. The authors utilized neural graphic primitives multi-resolution hash encoding for static scenes and extended Dynamic NeRF for dynamic scenes. Their work demonstrated improved reconstruction detail and limitations in both static and dynamic settings, with a particular emphasis on real-world dynamic scenes.
    
    \item \textbf{Joint Rigid Motion Correction and Sparse-View CT via Self-Calibrating Neural Field} introduced a self-calibrating neural field for Sparse-View Computed Tomography (SVCT) reconstruction, addressing the challenge of rigid motion during CT acquisition. The authors proposed a method that parametrizes inaccurate projection poses as trainable variables, optimizing these alongside a MLP. Their model showed significant improvements over traditional NeRF-based methods in SVCT tasks with various levels of rigid motion.
    
    \item \textbf{Compressing Explicit Voxel Grid Representations: fast NeRFs become also small} proposed Re:NeRF, targeting the compressibility of Explicit Voxel Grid (EVG)-NeRFs to reduce memory storage while maintaining performance. The authors tested Re:NeRF across different EVG-NeRF architectures and datasets, demonstrating its effectiveness in compressing NeRF models for faster training and rendering, albeit with a trade-off in memory usage.
     
    \item \textbf{Learning Neural Radiance Fields from Multi-View Geometry} combined Multi-View Geometry algorithms with NeRF for image-based 3D reconstruction. The framework, MVG-NeRF, leveraged geometric priors from classical 3D reconstruction pipelines to guide NeRF optimization, improving the quality of estimated underlying surfaces and demonstrating effectiveness in real-world data.
    
    \item \textbf{NeRF-SLAM (Real-Time Dense Monocular SLAM with Neural Radiance Fields)} presented a real-time 3D mapping pipeline combining dense monocular SLAM with hierarchical volumetric neural radiance fields. Their approach, NeRF-SLAM, provided accurate pose estimates and depth maps with associated uncertainty, achieving better geometric and photometric accuracy compared to existing methods using only monocular images.
    
    \item \textbf{NeRFPlayer (A Streamable Dynamic Scene Representation with Decomposed Neural Radiance Fields)} is a framework for efficient reconstruction, compact modeling, and streamable rendering of dynamic scenes. The framework decomposed 4D spatiotemporal space into static, deforming, and new areas, each represented by a separate neural field. They demonstrated its effectiveness in handling dynamic scenes captured by single or multi-camera arrays, offering fast reconstruction and interactive rendering capabilities.
    
    \item \textbf{ProbNeRF (Uncertainty-Aware Inference of 3D Shapes from 2D Images)} is a model for learning probabilistic generative models of 3D objects and doing posterior inference from 2D images. ProbNeRF used Hamiltonian Monte Carlo for inference, allowing it to propose realistic and diverse hypotheses about unseen parts of objects. The model's key success factors included a deterministic rendering scheme, an annealed-HMC strategy, a hypernetwork-based decoder architecture, and inference over a full set of NeRF weights.
    
    \item \textbf{Mixed Reality Interface for Digital Twin of Plant Factory} proposed an immersive mixed reality interface for digital twin models in smart farming, enabling remote work and control of plant factory facilities. Utilizing deformable neural radiance fields for real-time scene processing from camera devices, the system offers an intuitive UI for monitoring and interaction. The architecture integrates with existing IoT infrastructures, allowing users to interact through HMD or 2D displays and send commands to IoT servers, enhancing the user experience in smart farming applications.
    
    \item \textbf{nerf2nerf (Pairwise Registration of Neural Radiance Fields)} introduced a novel method for 3D registration of NeRFs, focusing on aligning partially overlapping geometry from different scenes. The authors developed a concept of "surface fields" extracted from pre-trained NeRF models, which are invariant to illumination changes. This approach, termed nerf2nerf, involves robust optimization to align these surface fields, providing a new solution for registering scenes under varying lighting conditions. The method's effectiveness was demonstrated through evaluations on a dataset of synthetic and real-world scenes, showing its potential in computer vision and robotics applications, especially where traditional 3D reconstruction methods face challenges.
    
    \item \textbf{Learning-based Inverse Rendering of Complex Indoor Scenes with Differentiable Monte Carlo Raytracing} introduces an end-to-end learning-based framework for inverse rendering of complex indoor scenes, integrating differentiable Monte Carlo raytracing. This method takes a single image to recover geometry, spatially-varying lighting, and photorealistic materials. They introduced a physically-based differentiable rendering layer with screen-space ray tracing for more accurate specular reflections. Additionally, an out-of-view lighting network with uncertainty-aware refinement was designed, leveraging hypernetwork-based neural radiance fields. The framework was trained on a large-scale, photorealistic indoor scene dataset, demonstrating superior inverse rendering quality and enabling high-fidelity applications like complex object insertion and material editing.
    
    \item \textbf{A Quantum-Powered Photorealistic Rendering} introduced Quantum Radiance Fields (QRF), a novel approach combining quantum circuits, quantum activation functions, and quantum volume rendering for photorealistic rendering of real-world scenes. QRF addresses the challenges of extensive numerical integration in traditional rendering by leveraging the parallelism of quantum computing. This method significantly enhances the rendering speed and quality, outperforming classical neural networks in capturing fine signal details and high-frequency information. The study demonstrated QRF's effectiveness in 2D image regression and 3D scene reconstruction, showcasing its potential in applications requiring high-quality rendering, such as virtual reality.
    
    \item \textbf{ParticleNeRF (A Particle-Based Encoding for Online Neural Radiance Fields)} is a novel approach for dynamic NeRFs suitable for online use, adapting in real-time to changes in scene geometry. ParticleNeRF employs a particle-based parametric encoding, where features are coupled to particles in space, and their positions are updated based on photometric reconstruction loss. This method allows for dynamic adaptation to moving, articulated, or deforming objects, outperforming traditional NeRFs and other baselines in dynamic scenes. ParticleNeRF's physics system prevents particle collisions, ensuring efficient training and high-quality rendering of dynamic scenes, marking a significant advancement in online learning of dynamic environments.
    
    \item \textbf{3D-Aware Encoding for Style-based Neural Radiance Fields} highlights a two-stage encoder framework, NeRF-3DE, for the inversion of style-based NeRFs, like StyleNeRF. The first stage involves a base encoder that converts an input image to a latent code, trained with identity contrastive learning for view-invariance. The second stage features a refining encoder that adds finer details to the output image. This approach ensures 3D consistency in generated novel views while preserving the identity of the input image. The novelty lies in the base encoder producing the closest latent code on the latent manifold, making the refinement stage's output close to the NeRF manifold. The method demonstrates superior performance in image reconstruction and novel-view rendering compared to existing encoders for GAN inversion.
    
    \item \textbf{NeRFFaceEditing (Disentangled Face Editing in Neural Radiance Fields)} is a method for editing facial geometry and appearance in neural radiance fields. The authors achieved disentanglement using statistics of the tri-plane to represent appearance and a geometry decoder aligned with semantic masks. This approach allowed intuitive editing with decoupled control of geometry and appearance, outperforming existing solutions in both qualitative and quantitative evaluations.
    
    \item \textbf{CoNFies (Controllable Neural Face Avatars)} is a system for creating controllable neural face avatars using automated facial action recognition. This method mapped facial expressions to action units and intensities, enabling anatomically correct control without manual annotation. CoNFies excelled in synthesizing novel views and expressions, offering better visual and anatomical fidelity compared to previous methods.
    
    \item \textbf{AligNeRF (High-Fidelity Neural Radiance Fields via Alignment-Aware Training)} focuses on high-resolution NeRF. They combined MLPs with convolutional layers and introduced an alignment-aware training strategy to address misalignment in dynamic scenes. AligNeRF efficiently encoded high-frequency details and improved image quality in high-resolution settings, demonstrating superior performance on various datasets.
    
    \item \textbf{Magic3D (High-Resolution Text-to-3D Content Creation)} is a method for creating high-resolution 3D content from text prompts. Utilizing a two-stage optimization framework with diffusion priors and a differentiable renderer, Magic3D significantly reduced computation time and improved resolution compared to previous methods, offering new avenues for creative 3D applications.
    
    \item \textbf{DynIBaR (Neural Dynamic Image-Based Rendering)} is a novel approach for synthesizing views from monocular videos of dynamic scenes. This method aggregated features from nearby views in a scene motion-aware manner, overcoming limitations of previous dynamic NeRF methods. DynIBaR demonstrated significant improvements in rendering fidelity on dynamic scene datasets and in-the-wild videos with complex motion.
    
    \item \textbf{FLNeRF (3D Facial Landmarks Estimation in Neural Radiance Fields)} is the first model for directly predicting 3D facial landmarks on neural radiance fields. FLNeRF employed a coarse-to-fine approach with expression augmentation, enabling accurate landmark detection for a wide range of expressions. This method facilitated high-quality face editing and swapping, demonstrating its utility in various downstream tasks.
    
    \item \textbf{Next3D (Generative Neural Texture Rasterization for 3D-Aware Head Avatars)} is a 3D GAN framework for creating high-quality, 3D-consistent facial avatars from 2D images. This paper introduced Generative Texture-Rasterized Tri-planes, combining mesh-guided explicit deformation control with implicit volumetric representation flexibility. This method excelled in animatable portrait synthesis, including full-head rotations and facial expressions, and served as a strong 3D prior for applications like one-shot facial avatars and 3D-aware stylization.
    
    \item \textbf{SegNeRF (3D Part Segmentation with Neural Radiance Fields)} integrates a semantic field with radiance fields for 3D part segmentation from images. SegNeRF inherited NeRF's capabilities in novel view synthesis and 3D reconstruction, enabling 3D part segmentation with only image-level supervision. It achieved competitive performance against point-based methods on PartNet, demonstrating its effectiveness in predicting geometry, appearance, and semantic information.
    
    \item \textbf{Local-to-Global Registration for Bundle-Adjusting Neural Radiance Fields} introduced L2G-NeRF, a method combining local pixel-wise flexible alignment with global frame-wise constrained parametric alignment for bundle-adjusting Neural Radiance Fields. This approach optimized photometric reconstruction errors and used differentiable parameter estimation solvers for global transformation, outperforming state-of-the-art in high-fidelity reconstruction and camera pose misalignment resolution.
    
    \item \textbf{Shape, Pose, and Appearance from a Single Image via Bootstrapped Radiance Field Inversion} presented a framework for reconstructing shape, pose, and appearance from a single image using bootstrapped radiance field inversion. Their method leveraged an unconditional 3D-aware generator with a hybrid inversion scheme, producing accurate results in as few as 10 steps. This approach was effective for AR applications and demonstrated state-of-the-art results on real and synthetic benchmarks.
    
    \item \textbf{ESLAM (Efficient Dense SLAM System Based on Hybrid Representation of Signed Distance Fields)} is an efficient dense SLAM system using a hybrid representation of Signed Distance Fields. ESLAM incrementally reconstructed scenes and estimated camera positions using RGB-D frames. It utilized multi-scale axis-aligned perpendicular feature planes and shallow decoders, significantly improving accuracy and speed in 3D reconstruction and camera localization compared to existing methods.
    
    \item \textbf{SPARF (Neural Radiance Fields from Sparse and Noisy Poses)} is a method for novel-view synthesis using sparse input images with noisy camera poses. SPARF employed multi-view geometry constraints to jointly learn NeRF and refine camera poses, using pixel matches and a depth consistency loss. This approach set new standards in sparse-view scenarios on multiple challenging datasets, overcoming limitations of NeRF in real-world applications.
    
    \item \textbf{Level-S$^2$fM (Structure from Motion on Neural Level Set of Implicit Surfaces)} is a neural incremental SfM approach. It estimates camera poses and scene geometry from uncalibrated images by learning coordinate MLPs for implicit surfaces and radiance fields. The authors tackled challenges in the SfM pipeline, such as two-view geometry initialization and camera pose registration, by exploiting 2D correspondences and NIS. This method demonstrated promising results in camera pose estimation and scene geometry reconstruction, offering a new perspective for neural implicit rendering.
    
    \item \textbf{ONeRF (Unsupervised 3D Object Segmentation from Multiple Views)} is an unsupervised method for segmenting and reconstructing 3D objects from multi-view RGB images using separate NeRFs. It employs an iterative Expectation-Maximization algorithm to aggregate 2D visual features and corresponding 3D cues for joint segmentation and reconstruction. This approach can handle complex objects and supports various 3D scene editing tasks, offering a significant baseline for unsupervised learning in segmenting cluttered scenes into individual NeRFs.
    
    \item \textbf{DP-NeRF (Deblurred Neural Radiance Field with Physical Scene Priors)} addresses the challenge of reconstructing 3D scenes from blurred images using NeRF. It introduces two physical priors: a rigid blurring kernel for 3D consistency and an adaptive weight proposal for refining color composition. The method shows significant improvements in perceptual quality and 3D geometric and appearance consistency, particularly in scenarios with camera motion blur and defocus blur.
    
    \item \textbf{SPIn-NeRF (Multiview Segmentation and Perceptual Inpainting with Neural Radiance Fields)} proposes a novel 3D inpainting method for editing NeRF scenes, particularly for object removal. It uses sparse annotations in a single input image to obtain a 3D segmentation mask and then applies perceptual optimization-based approach for inpainting, ensuring view consistency. The method addresses the lack of diverse benchmarks for 3D scene inpainting by introducing a dataset with real-world scenes, demonstrating superior performance in both segmentation and inpainting tasks.
    
    \item \textbf{Exact-NeRF (An Exploration of a Precise Volumetric Parameterization for Neural Radiance Fields)} explores a precise volumetric parameterization for NeRF, addressing issues in frustum approximation in earlier NeRF work. It proposes a pyramid-based integral formulation for Integrated Positional Encoding (IPE), offering an exact solution for both bounded and unbounded scenes. This approach matches the performance of mip-NeRF and provides a natural extension to more challenging scenarios, contributing to the understanding of analytical solutions in future NeRF extensions.
    
    \item \textbf{Real-time Neural Radiance Talking Portrait Synthesis via Audio-spatial Decomposition} presents RAD-NeRF, a framework for real-time synthesis of talking portraits driven by audio. It decomposes the portrait representation into low-dimensional feature grids for efficient dynamic head modeling and introduces a lightweight Pseudo-3D Deformable Module for the torso. The method achieves significant speed improvements over previous works and supports various explicit controls of the talking portrait, such as head pose and eye blink, demonstrating its potential for applications like virtual video conferencing and digital human creation.
    
    \item \textbf{Dynamic Depth-Supervised NeRF for Multi-View RGB-D Operating Room Images} explored the use of NeRF for view synthesis in dynamic operating room scenes, emphasizing depth supervision from RGB-D sensor data. The authors optimized a dynamic depth-supervised NeRF with synchronized cameras, capturing surgical fields in various phases. They demonstrated that NeRF could generate geometrically consistent views from interpolated camera positions and time intervals, achieving high-quality rendering with minimal depth estimation error.
    
    \item \textbf{INSTA (Instant Volumetric Head Avatars)} presents a rapid method for reconstructing photo-realistic digital avatars from monocular RGB portrait videos. The technique models a dynamic neural radiance field around a parametric face model, allowing for interactive rendering of novel poses and expressions. INSTA significantly outperforms state-of-the-art methods in rendering quality and training time, leveraging the geometry prior of the face model for pose extrapolation in AR/VR applications.
    
    \item \textbf{PermutoSDF (Fast Multi-View Reconstruction with Implicit Surfaces using Permutohedral Lattices)} combines hash-based encodings and implicit surfaces for 3D reconstruction, introducing permutohedral lattice hashing for faster optimization. The method focuses on recovering high-frequency geometric details using RGB images for supervision and can render novel views in real-time. PermutoSDF's regularization scheme is crucial for capturing fine details like pores and wrinkles.
    
    \item \textbf{ActiveRMAP (Radiance Field for Active Mapping And Planning)} introduces an RGB-only active vision framework using radiance field representation for online 3D reconstruction and planning. The method formulates joint tasks as an iterative dual-stage optimization problem, alternating between optimizing the radiance field and path planning. The framework demonstrates competitive results in active reconstruction, outperforming existing methods using NeRFs.
    
    \item \textbf{PANeRF (Pseudo-view Augmentation for Improved Neural Radiance Fields Based on Few-shot Inputs)} addresses the challenge of rendering high-quality images with sparse input views by introducing a pseudo-view augmentation scheme. The method leverages saliency maps for training image division and proposes a two-step learning process with novel regularization methods. PANeRF synthesizes superior novel-view images compared to existing few-shot methods, demonstrating robustness across various conditions.
    
    \item \textbf{BAD-NeRF (Bundle Adjusted Deblur Neural Radiance Fields)} tackles the challenge of training NeRF with motion-blurred images and inaccurate camera poses. It integrates the physical image formation process of motion blur into NeRF training, jointly learning NeRF parameters and camera motion trajectories. BAD-NeRF shows superior performance in deblurring motion-blurred images and synthesizing high-quality novel view images, outperforming previous works.
    
    \item \textbf{ClimateNeRF (Extreme Weather Synthesis in Neural Radiance Field)} is a novel NeRF-editing procedure that fuses physical simulations with NeRF models of scenes. This method allows for the realistic rendering of weather effects like smog, snow, and floods in 3D scenes. The technique, which combines instant NGP (a variant of NeRF) and physical simulations, was shown to produce more realistic results than state-of-the-art 2D image editing and 3D NeRF stylization methods.
    
    \item \textbf{Immersive Neural Graphics Primitives} describes a NeRF-based framework for VR, integrating instant neural graphics primitives into Unity. This framework enables immersive VR experiences with photo-realistic neural graphics, supporting frame rates around 30 fps. The authors evaluated the framework's performance across different scene complexities and resolutions, demonstrating its potential for various VR applications.
    
    \item \textbf{ScanNeRF (A Scalable Benchmark for Neural Radiance Fields)} is a benchmark for evaluating NeRFs using a cost-effective scanning station. This station can capture thousands of images of an object, enabling the creation of a dataset for training and evaluating NeRFs. The authors demonstrated that their approach allows for the efficient generation of digital twins of real objects, highlighting its potential for future developments in neural rendering.
    
    \item \textbf{TPA-Net (Generate A Dataset for Text to Physics-based Animation)} is a method for generating high-resolution, physically realistic animations with descriptive texts. The auhors used advanced physical simulation frameworks like IPC and MPM to simulate a range of dynamics, producing a dataset beneficial for T2V, NeRF, and other research areas. This work represents a significant step towards fully automated Text-to-Video/Simulation.
    
    \item \textbf{Sampling Neural Radiance Fields for Refractive Objects} is a NeRF-based framework for rendering refractive objects. The authors extended stratified and hierarchical sampling techniques in NeRF for curved paths, improving rendering quality for scenes with refractive elements. Their method outperformed state-of-the-art techniques in both synthetic and real scenes, demonstrating its effectiveness in handling refraction and total reflection effects.
    
    \item \textbf{SuNeRF (Validation of a 3D Global Reconstruction of the Solar Corona Using Simulated EUV Images)} adapts NeRFs to the physical properties of the Sun. This model accurately reconstructs the 3D structure of the Sun from limited viewpoints, demonstrating potential for satellite observations and space-weather forecasting. SuNeRF's validation on simulated EUV images shows its capability to provide a consistent 3D reconstruction of the Sun.
    
    \item \textbf{High-fidelity Facial Avatar Reconstruction from Monocular Video with Generative Priors} introduced a method for reconstructing high-fidelity facial avatars from monocular videos, leveraging NeRF and 3D-aware generative priors. This approach significantly improved face reenactment and novel view synthesis by learning a personalized generative prior in the latent space of 3D-GAN. The method showed superior performance in various applications, including digital human creation, video conferencing, and AR/VR, compared to existing works.
    
    \item \textbf{One is All: Bridging the Gap Between Neural Radiance Fields Architectures with Progressive Volume Distillation} proposed Progressive Volume Distillation (PVD), a method enabling conversions between different NeRF architectures like MLP, sparse tensors, and hashtables. PVD facilitated the adaptation of neural representations for various applications post hoc, offering a systematic approach to distillation across architectures. The method demonstrated faster training and superior synthesis quality on datasets like NeRF-Synthetic, LLFF, and TanksAndTemples.
    
    
    \item \textbf{DINER (Depth-aware Image-based NEural Radiance fields)} is a method for constructing image-based NeRF guided by estimated dense depth maps. This approach allowed for rendering 3D objects under novel views with higher synthesis quality and completeness, especially useful in applications like video conferencing with holographic displays. DINER outperformed state-of-the-art methods in synthesizing novel views for human heads and general objects.
    
    \item \textbf{Mixed Neural Voxels for Fast Multi-view Video Synthesis} introduced MixVoxels, a method for representing dynamic scenes as a mixture of static and dynamic voxels for fast multi-view video synthesis. The approach efficiently handled high-dynamic movements and complex real-world environments, achieving better PSNR and faster rendering speeds than previous methods. MixVoxels found applications in free-viewpoint control for movies, sports events, and VR/AR.
    
    \item \textbf{ViewNeRF (Unsupervised Viewpoint Estimation Using Category-Level Neural Radiance Fields)} is an unsupervised method for category-level viewpoint estimation using NeRFs, was introduced in this paper. The method combined a conditional NeRF with a viewpoint predictor and scene encoder, enabling efficient and accurate viewpoint prediction in complex scenarios. It showed competitive results on synthetic and real datasets for both single scenes and multi-instance collections, addressing the limitations of previous NeRFs that required accurate camera poses.
    
    \item \textbf{QFF (Quantized Fourier Features for Neural Field Representations)} is a neural field representation technique for efficient and high-quality modeling of functions like Neural Image Representations, NeRFs, and SDF modeling. QFF offered a compact, continuous, and multiscale representation, leading to smaller model sizes, faster training, and improved output quality across various applications.
    
    \item \textbf{3D-TOGO (Towards Text-Guided Cross-Category 3D Object Generation)} is a model for generating 3D objects from textual descriptions across various categories. The authors combined a text-to-views generation module with a views-to-3D generation module, using pixelNeRF for the latter to create neural radiance fields. This approach, tested on the ABO dataset, showed superior performance in generating high-quality 3D objects with control over category, color, and shape, compared to existing methods like text-NeRF and Dreamfields.
    
    \item \textbf{RT-NeRF (Real-Time On-Device Neural Radiance Fields Towards Immersive AR/VR Rendering)} is an algorithm-hardware co-designed framework for accelerating NeRF rendering on AR/VR devices. Addressing inefficiencies in uniform point sampling and dense computations in NeRF, RT-NeRF integrates an efficient rendering pipeline and a coarse-grained view-dependent computing scheme. The hardware component features a hybrid encoding scheme and a dual-purpose adder \& search tree for efficient sparse decoding, achieving significant throughput improvements while maintaining rendering quality.

    \item \textbf{Surface Normal Clustering for Implicit Representation of Manhattan Scenes} proposed a method to leverage the geometric prior of Manhattan scenes for improving NeRF representations. The authors used surface normal clustering derived from rendered depths, focusing on scenes with unknown Manhattan frames. This approach, tested on diverse indoor datasets, demonstrated significant improvements over established baselines, offering a robust and practical solution for exploiting structured-scene knowledge in 3D neural radiance field representations.
    
    \item \textbf{Fast Non-Rigid Radiance Fields from Monocularized Data} discusses MoNeRF, a method for real-time volumetric novel view synthesis and temporal interpolation of dynamic scenes using monocularized data. MoNeRF employs an efficient deformation module and a static module representing the canonical scene as a fast hash-encoded NeRF. This approach, validated on both synthetic and real-world inward-facing scenes, showed faster convergence and higher visual accuracy for novel view generation compared to previous methods.
    
    \item \textbf{StegaNeRF (Embedding Invisible Information within Neural Radiance Fields)}, as introduced in this paper, is a novel framework for embedding steganographic information in NeRF renderings. The authors developed an optimization framework that embeds hidden information during NeRF training, allowing accurate extraction from rendered images while preserving visual quality. This method opens new possibilities for copyright protection and ownership identification in 3D content shared online, balancing rendering quality with high-fidelity transmission of concealed information.
    
    \item \textbf{MaRF (Representing Mars as Neural Radiance Fields)} is a framework designed to synthesize the Martian environment using images from rover cameras. It utilizes NeRFs and NGPs for efficient 3D scene reconstruction of Mars' surface. This approach addresses challenges in planetary geology and simulated navigation, offering a mixed-reality experience for scientists and engineers to virtually explore Mars. The framework demonstrates environments created from datasets captured by Curiosity, Perseverance, and Ingenuity, showcasing its potential for space applications and image compression.
    
    \item \textbf{Neural Fourier Filter Bank} introduces a novel method for efficient and detailed reconstructions, blending spatial and frequency decomposition inspired by wavelets. The authors propose a neural Fourier filter bank, leveraging a MLP with sine activations and Fourier features encodings. This approach outperforms existing methods in model compactness and convergence speed across tasks like 2D image fitting, 3D shape reconstruction, and NeRF.
    
    \item \textbf{Understanding Sinusoidal Neural Networks} delves into the structure and representation capacity of sinusoidal MLPs, crucial in computer graphics for representing images, signed distance functions, and radiance fields. The paper presents a harmonic sum expansion of sinusoidal MLPs, revealing their ability to generate a large number of new frequencies. This expansion aids in controlling the approximation and training of sinusoidal MLPs, with applications in image processing and neural field modeling.
    
    \item \textbf{GARF (Geometry-Aware Generalized Neural Radiance Field)}, a Geometry-Aware Generalized Neural Radiance Field, is introduced for real-time novel view rendering and unsupervised depth estimation on unseen scenes. The method employs a geometry-aware dynamic sampling strategy, an encoder-decoder structure, and a multi-view feature fusion module. It significantly reduces sampling points while improving rendering quality and 3D geometry estimation, applicable in indoor and outdoor scene analysis.
    
    \item \textbf{D-TensoRF (Tensorial Radiance Fields for Dynamic Scenes)}, a tensorial radiance field for dynamic scenes, is presented for novel view synthesis at specific times. The method models dynamic scenes as a 5D tensor, applying CP and MM decomposition for efficient computation. It incorporates smoothing regularization to capture temporal dependencies, offering compact model sizes, fast training, and competitive rendering results in dynamic scene modeling.
    
    \item \textbf{One-shot Implicit Animatable Avatars with Model-based Priors} proposes ELICIT, a method for creating animatable 3D human avatars from a single image. It leverages 3D geometry and visual semantic priors, using CLIP-based pretrained models and a skinned vertex-based template model (SMPL) for optimization. ELICIT enables text-conditioned unseen region generation and a segmentation-based sampling strategy, showing superior performance in avatar creation from single images.
    
    \item \textbf{Canonical Fields (Self-Supervised Learning of Pose-Canonicalized Neural Fields)}: CaFi-Net, a self-supervised method for canonicalizing the 3D pose of objects represented as neural fields, is introduced. It employs a Siamese network architecture with equivariant feature extraction layers, handling noisy radiance fields from NeRF. The method learns canonicalization without supervision, showing promising results in canonicalizing neural fields for various object categories.
    
    \item \textbf{SceneRF (Self-Supervised Monocular 3D Scene Reconstruction with Radiance Fields)}, introduced by Anh-Quan Cao and Raoul de Charette, is a novel self-supervised method for monocular 3D scene reconstruction using posed image sequences. It leverages the advancements in NeRF and introduces explicit depth optimization with a unique probabilistic sampling strategy for large scenes. This technique outperforms existing methods in synthesizing novel depth views and scene reconstruction, applicable in both indoor and outdoor settings.
    
    \item \textbf{SSDNeRF (Semantic Soft Decomposition of Neural Radiance Fields)}, presents a technique for semantic soft decomposition of neural radiance fields. It encodes semantic signals alongside radiance signals, enabling detailed 3D semantic scene representation. This method offers improved segmentation and reconstruction results, with applications in high-quality, temporally consistent video editing and re-compositing, particularly useful in casually captured selfie videos.
    
    \item \textbf{Non-uniform Sampling Strategies for NeRF on 360{\textdegree} images} proposed two novel non-uniform ray sampling schemes for NeRF tailored to 360° images, addressing the challenges posed by spatial distortion and wide viewing angles. The authors methods, distortion-aware and content-aware ray sampling, enhance the accuracy and efficiency of NeRF models on 360° images. This approach shows promise in improving performance for both synthetic and real-world scenes.
    
    \item \textbf{EditableNeRF (Editing Topologically Varying Neural Radiance Fields by Key Points)} is a method for editing dynamic scenes in NeRF, supporting topological changes. It uses surface key points for intuitive multi-dimensional editing, allowing users to generate novel scenes unseen during training. This approach, which includes a scene analysis method for key point detection and a weighted key points strategy, offers significant improvements in dynamic scene editing.
    
    \item \textbf{Few-View Object Reconstruction with Unknown Categories and Camera Poses} developed FORGE, a method for reconstructing objects from a few views without known camera poses or object categories. It predicts 3D features and relative camera poses, fusing these features into a neural radiance field for volume rendering. FORGE demonstrates strong generalization across object categories and significantly outperforms existing methods in pose estimation and reconstruction quality.
    
    \item \textbf{Neural Volume Super-Resolution} introduced a method for neural volume super-resolution, focusing on rendering high-resolution views from low-resolution captured scenes. The authors approach, which operates directly on the volumetric scene representation, ensures geometric consistency across viewing directions. They developed a novel 3D representation using 2D feature planes, achieving superior performance in rendering sharp, consistent high-resolution images from low-resolution inputs.
    
    \item \textbf{4K-NeRF (High Fidelity Neural Radiance Fields at Ultra High Resolutions} is a framework enhancing the NeRF methodology for ultra-high-resolution view synthesis. The authors utilized a 3D-aware encoder and decoder to model geometric information in lower resolutions and recover fine details in high resolutions. Their approach, incorporating ray correlation and patch-based sampling, significantly improved rendering quality in 4K scenarios, outperforming traditional NeRF methods.
    
    \item \textbf{GazeNeRF (3D-Aware Gaze Redirection with Neural Radiance Fields)}, a novel 3D-aware method for gaze redirection, was proposed by the authors. It leverages conditional image-based NeRFs with a two-stream architecture for separate volumetric feature prediction of the face and eye regions. The method, using rigid transformation of eye features, showed superior performance in gaze redirection accuracy and identity preservation compared to existing 2D and NeRF-based methods.
    
    \item \textbf{Rodin (A Generative Model for Sculpting 3D Digital Avatars Using Diffusion)}, a diffusion model for generating 3D digital avatars as neural radiance fields. Addressing the challenge of high memory and computational costs in 3D, Rodin uses a roll-out diffusion network with 3D-aware convolution and latent conditioning for global coherence. It supports generating detailed avatars from images or text prompts and enables text-guided semantic editing.
    
    \item \textbf{NoPe-NeRF (Optimising Neural Radiance Field with No Pose Prior)} tackled the challenge of training NeRF without precomputed camera poses, especially in scenarios with dramatic camera movement. The authors introduced NoPe-NeRF, which uses monocular depth priors corrected for scale and shift distortions during training. This approach, combined with novel loss functions, significantly improved pose estimation accuracy and novel view rendering quality in challenging camera trajectories.
    
    \item \textbf{VolRecon (Volume Rendering of Signed Ray Distance Functions for Generalizable Multi-View Reconstruction)}, a novel method for generalizable implicit reconstruction using Signed Ray Distance Function (SRDF), was proposed. It combines projection features from multi-view images and volume features from a global feature volume. The method demonstrated superior performance in sparse view reconstruction and generalization to large-scale scenes, outperforming existing methods like SparseNeuS and MVSNet.
    
    \item \textbf{NeRF-Art (Text-Driven Neural Radiance Fields Stylization)}, a text-guided NeRF stylization method, was introduced, enabling the transformation of appearance and geometry of a pre-trained NeRF model using text prompts. The method employs a global-local contrastive learning strategy and weight regularization to handle geometry and appearance transformations, demonstrating robustness in single-view stylization quality and cross-view consistency.
    
    \item \textbf{MEIL-NeRF (Memory-Efficient Incremental Learning of Neural Radiance Fields)} addresses the challenge of catastrophic forgetting in NeRF under incremental learning scenarios. The authors developed a Memory-Efficient Incremental Learning algorithm for NeRF, utilizing a self-distillation approach and a Ray Generator Network (RGN) to maintain performance without increasing memory requirements. This method is particularly beneficial for applications in large-scale scenes and edge devices with limited memory, where data is processed sequentially.
    
    \item \textbf{SteerNeRF (Accelerating NeRF Rendering via Smooth Viewpoint Trajectory)}, proposed by the authors, accelerates NeRF rendering by leveraging smooth viewpoint changes in interactive control. The method combines low-resolution volume rendering with high-resolution 2D neural rendering, significantly reducing rendering time while maintaining image fidelity. This approach is particularly effective for applications requiring real-time rendering, such as virtual reality and autonomous driving, achieving up to 30FPS at 1080P resolution with minimal memory overhead.
    
    \item \textbf{Masked Wavelet Representation for Compact Neural Radiance Fields} presents a novel approach for compact NeRF using wavelet transform on grid-based neural fields. The authors introduced a trainable masking technique to enhance sparsity in grid coefficients, leading to a more compact representation without sacrificing reconstruction quality. This method is particularly useful for applications requiring efficient storage and processing, such as real-time rendering and large-scale scene reconstruction.
    
    \item \textbf{SPARF (Large-Scale Learning of 3D Sparse Radiance Fields from Few Input Images)}, a large-scale dataset for novel view synthesis, was introduced alongside SuRFNet, a pipeline for generating sparse voxel radiance fields from few views. The authors leveraged 3D sparse convolutions and a specialized SRF loss, achieving state-of-the-art results in novel view synthesis with minimal input. This approach is significant for applications in 3D modeling and virtual reality, where high-quality rendering from limited views is essential.
    
    \item \textbf{StyleTRF (Stylizing Tensorial Radiance Fields)}, a technique for stylizing Tensorial Radiance Fields, was developed to generate stylized 3D scenes efficiently. The method utilizes a pre-optimized TensoRF scene representation and a stylization module to adapt style in a short time, enabling the generation of stylized novel views with traditional volumetric rendering techniques. This approach is particularly relevant for applications in augmented and virtual reality, where stylized content generation is desired.
    
    \item \textbf{Correspondence Distillation from NeRF-based GAN} is a method to establish dense correspondences across different NeRFs of the same category, leveraging a pre-trained NeRF-based GAN. The authors introduced a Dual Deformation Field (DDF) to facilitate learning without ground-truth correspondence annotations. This technique is crucial for applications like texture transfer, shape manipulation, and segmentation transfer, where establishing accurate and robust 3D correspondences is key.
    
    \item \textbf{PaletteNeRF (Palette-based Appearance Editing of Neural Radiance Fields)} is a method for efficient appearance editing of NeRF. They decomposed the appearance of 3D points into a linear combination of palette-based color bases, enabling intuitive and photorealistic recoloring across views. The method supports various editing applications, including illumination modification and style transfer, and outperforms baseline methods in editing complex real-world scenes.
    
    \item \textbf{Incremental Neural Implicit Representation with Uncertainty-Filtered Knowledge Distillation} presented a student-teacher framework to address the catastrophic forgetting problem in Neural Implicit Representations (NIRs) when learning from streaming data. The authors introduced an uncertainty-based filter and a random inquirer to select useful information from the teacher network. Their method showed significant improvements over existing methods in both 3D reconstruction and novel view synthesis tasks.
    
    \item \textbf{Removing Objects From Neural Radiance Fields} proposes a framework for removing objects from NeRFs using a combination of 2D image inpainting and a confidence-based view selection procedure. This approach allows for plausible removal of objects while maintaining multi-view coherence. They demonstrated the effectiveness of their method on a new dataset for NeRF inpainting, outperforming existing approaches in image quality and multi-view consistency.
    
    \item \textbf{PaletteNeRF (Palette-based Color Editing for NeRFs)} is an extension of vanilla NeRF, enabling efficient color editing on NeRF-represented scenes. The method approximates pixel colors as a sum of palette colors modulated by additive weights, allowing for efficient, view-consistent, and artifact-free color editing. The approach is compatible with recent NeRF models like KiloNeRF for real-time editing.
    
    \item \textbf{MonoNeRF (Learning a Generalizable Dynamic Radiance Field from Monocular Videos)} was proposed to learn a generalizable dynamic radiance field from monocular videos. The method learns point features and scene flows with trajectory and feature correspondence constraints, supporting applications like scene editing and unseen frame synthesis. MonoNeRF demonstrated the ability to render novel views from dynamic videos and adapt quickly to new scenes.
    
    \item \textbf{Dream3D (Zero-Shot Text-to-3D Synthesis Using 3D Shape Prior and Text-to-Image Diffusion Models)} is a method for zero-shot text-to-3D synthesis, utilizing 3D shape priors and text-to-image diffusion models. The approach generates high-quality 3D shapes from text inputs, bridging the gap between text and image modalities. Dream3D outperforms state-of-the-art methods in generating imaginative 3D content with superior visual quality and shape accuracy.
    
    \item \textbf{NeRF-Gaze: A Head-Eye Redirection Parametric Model for Gaze Estimation} proposes a novel approach to generate gaze data using a Neural Radiance Field (NeRF) based model. This model enables accurate control of gaze direction and facial attributes, including identity and illumination, independently for the face and eyes. The authors demonstrate that their method can synthesize high-fidelity head images with precise gaze direction, offering significant benefits for gaze estimation tasks in domain generalization and adaptation settings. Extensive experiments validate the model's effectiveness in generating diverse and accurate gaze datasets, which can be applied to improve gaze estimation models. The paper presents a significant contribution to gaze estimation research, offering a new avenue for creating more robust and generalizable gaze estimation models.
    
    \item \textbf{Equivariant Light Field Convolution and Transformer} presents a novel approach for 3D reconstruction from multiple views using equivariant shape priors. This method bridges the gap between optimization/NeRF-based single scene reconstruction and generalizable approaches that are not equivariant. The authors introduce a mathematical framework for light field convolutions, showing its equivalence to intra-view spherical convolutions, and then approximate these with SE(2)-convolutions. They also propose an equivariant transformer model over the ray space, leading to a significant advancement in the field of geometric computer vision and 3D reconstruction.
\end{itemize}

\item \textbf{Research in 2023:} This subsection briefly discusses the research articles in the year of 2023.

\begin{itemize}
    
    \item \textbf{Detachable Novel Views Synthesis of Dynamic Scenes Using Distribution-Driven Neural Radiance Fields} presents a novel approach for synthesizing new views in dynamic real-world scenes from monocular videos. The method, known as Distribution-Driven Neural Radiance Fields, enhances the quality of view synthesis and separates the background from dynamic scenes. It utilizes a background pipeline to model underlying distribution and a 6D-input NeRF for dynamic components. The approach effectively handles occlusion areas and static backgrounds, offering improved performance in synthesizing novel views of dynamic scenes.
    
    \item \textbf{Class-Continuous Conditional Generative Neural Radiance Field} introduces a novel model for 3D-aware conditional image generation, addressing the challenge of conditional and continuous feature manipulation in generative NeRF (Neural Radiance Field) models. The proposed method enables the generation of 3D-consistent images with fine details and control over features through continuous conditional label values. This approach marks a significant advancement in the field, as it allows for more precise and varied image generation, particularly in applications requiring detailed and accurate 3D image synthesis.
    
    \item \textbf{Towards Open World NeRF-Based SLAM} aims to improve the NICE-SLAM algorithm, a state-of-the-art NeRF-based SLAM (Simultaneous Localization and Mapping) method. The focus is on enhancing accuracy and robustness by incorporating depth measurement uncertainty and IMU (Inertial Measurement Unit) data. The authors also extend the algorithm to represent distant backgrounds, which are challenging for traditional NeRF models to capture. These improvements are critical for deploying NeRF-based SLAM in open-world environments, where expansive areas and varying conditions are encountered.
    
    \item \textbf{Benchmarking Robustness in Neural Radiance Fields} evaluates the robustness of Neural Radiance Field (NeRF)-based novel view synthesis algorithms against various types of corruptions. The authors find that NeRF-based models degrade significantly in the presence of corruption, demonstrating greater sensitivity to different corruptions than image recognition models. Additionally, the paper reveals that feature encoders in generalizable methods contribute marginally to robustness, and standard data augmentation techniques don't improve NeRF-based model robustness. The study aims to encourage further research on enhancing the robustness of NeRF-based approaches for real-world applications.
    
    \item \textbf{Neural Radiance Field Codebooks} introduces Neural Radiance Field Codebooks (NRC), a method for learning object-centric representations through novel view reconstruction. NRC creates a dictionary of object codes, decoded through a volumetric renderer, to identify reoccurring visual and geometric patterns across scenes. It outperforms other methods in object navigation and unsupervised segmentation for synthetic and real scenes. NRC addresses limitations of current 3D object-centric methods by ensuring constant rendering compute and scaling with data complexity, sharing semantic and geometric object information across scenes.
    
    \item \textbf{WIRE: Wavelet Implicit Neural Representations} introduces a new method for implicit neural representations (INRs) using a continuous complex Gabor wavelet activation function, termed Wavelet Implicit neural REpresentation (WIRE). This approach enhances INR accuracy, training time, and robustness, addressing issues like long training times for high-dimensional data and sensitivity to signal noise. WIRE’s effectiveness is demonstrated through various vision-related tasks such as image denoising, inpainting, super-resolution, and computed tomography reconstruction. Its robustness makes it particularly suitable for challenging vision inverse problems, offering advancements over other INRs in terms of signal representation and solving inverse problems.
    
    \item \textbf{A Large-Scale Outdoor Multi-modal Dataset and Benchmark for Novel View Synthesis and Implicit Scene Reconstruction} introduces the OMMO dataset, which is significant for its large-scale, real-world outdoor scenes, diverse in nature, including various geographical scales, camera trajectories, and lighting conditions. The authors focused on overcoming the challenges in Neural Radiance Fields (NeRF) research for outdoor scenes by creating a comprehensive dataset with calibrated images, point clouds, and prompt annotations. This dataset is designed to benchmark NeRF methods on tasks like novel view synthesis and surface reconstruction. The paper also outlines the data collection and annotation methods, emphasizing the real-world applicability of the dataset in advancing large-scale NeRF research.
    
    \item \textbf{Behind the Scenes: Density Fields for Single View Reconstruction} introduces a neural network architecture for inferring 3D geometric structures from single images. The authors developed a method that predicts an implicit density field, offering a more comprehensive approach than traditional depth map predictions. This method allows for effective volume rendering and novel view synthesis, even for occluded regions in the input image. The paper demonstrates the method's capabilities through experiments on various datasets, showing its potential in accurately capturing 3D scenes and facilitating depth prediction and novel view synthesis. This approach is significant for applications in computer vision, particularly in areas requiring detailed 3D scene understanding from limited visual data.
    
    \item \textbf{RecolorNeRF: Layer Decomposed Radiance Fields for Efficient Color Editing of 3D Scenes} paper presented a novel method for color editing in three-dimensional scenes using neural radiance fields. The authors introduced an innovative technique that decomposed the scene into layers, allowing for efficient and precise color adjustments. This method, distinct in its approach, leveraged neural networks to parse and manipulate the color properties of the scene without compromising the structural integrity and realism of the 3D models.
    
    The applications of this technique were many folds, extending beyond academic research into practical real-world scenarios. In industries such as virtual reality, game design, and film production, where realistic and customizable 3D environments are crucial, the authors' method offered significant improvements in both efficiency and flexibility. By enabling easier and more precise color adjustments, this technology promised to enhance the creative process, reduce production time, and open new possibilities for designers and artists in various fields.
    
    \item \textbf{NeRF in the Palm of Your Hand: Corrective Augmentation for Robotics via Novel-View Synthesis} authored by Allan Zhou and colleagues, presented SPARTN (Synthetic Perturbations for Augmenting Robot Trajectories via NeRF), a novel data augmentation technique for improving robot policies using eye-in-hand cameras. The authors focused on addressing the challenge of training visual robotic manipulation policies through imitation learning, which often suffers from compounding errors due to the need for extensive demonstrations or costly online expert supervision. SPARTN leverages neural radiance fields (NeRFs) to synthetically inject corrective noise into visual demonstrations, thereby generating perturbed viewpoints and calculating corrective actions without additional expert supervision or environment interaction. This method distills geometric information from NeRFs into a real-time, reactive RGB-only policy.
    
    The paper demonstrated SPARTN's effectiveness in both simulated and real-world 6-DoF robotic grasping experiments. In simulations, SPARTN showed a significant improvement in success rates over traditional imitation learning methods and even outperformed some methods requiring online supervision. In real-world experiments, the method improved absolute success rates by an average of 22.5\%, including on objects challenging for depth-based methods. The authors highlighted SPARTN's ability to close the gap between RGB-only and RGB-D success rates, eliminating the need for depth sensors. The technique's application in real-world scenarios, particularly in robotic grasping tasks, showcased its potential to enhance the performance of vision-based control systems in robotics. Despite its strengths, the authors acknowledged limitations, such as its applicability primarily to tasks with static scenes and the computational expense of training a neural radiance field for each demonstration. Future work could explore extending the method to dynamic scenes and reducing computational demands.

    \item \textbf{3D Reconstruction of Non-cooperative Resident Space Objects using Instant NGP-accelerated NeRF and D-NeRF} by Basilio Caruso and colleagues explored the application of neural radiance field (NeRF) algorithms, specifically Instant NeRF and Dynamic NeRF (D-NeRF), for 3D modeling of non-cooperative resident space objects (RSOs). These techniques were assessed for their potential in aiding on-orbit servicing (OOS) and active debris removal (ADR) in space. The authors conducted experiments using datasets of images of a spacecraft mock-up under varying lighting and motion conditions at the Orbital Robotic Interaction, On-Orbit Servicing, and Navigation Laboratory at Florida Institute of Technology.
    
    The study demonstrated that Instant NeRF could learn high-fidelity 3D models with significantly reduced computational costs, making it feasible for training on on-board computers in space. This was a notable advancement, considering the limitations of computational resources in space environments. The authors also evaluated D-NeRF, designed for dynamic 3D objects, but found it less suited for the specific dynamic objects in their experiments. The research focused on three key questions: the ability of NeRF-related algorithms to learn high-fidelity 3D models from 2D scenes of a real-life satellite mock-up, the computational efficiency of an accelerated version of NeRF, and the necessity of an algorithm for dynamic 3D objects.
    
    The results indicated that both NeRF and Instant NeRF performed well in reconstructing 3D models of the satellite, with Instant NeRF showing advantages in computational efficiency and perceived image quality. The study also highlighted the potential of these algorithms in generating 3D models of spinning objects, provided the background could be precisely removed. Future work was proposed to evaluate Instant NeRF in different inspection orbits and on NVIDIA Jetson hardware, aiming to enable fully autonomous on-orbit servicing experiments and swarm satellite operations around non-cooperative RSOs. This research contributes significantly to the field of space domain awareness, offering innovative solutions for the identification and servicing of RSOs in orbit.
    
    \item \textbf{Ultra-NeRF: Neural Radiance Fields for Ultrasound Imaging} introduces an innovative method for generating 3D ultrasound (US) images using Neural Radiance Fields (NeRF). The authors focus on enhancing ultrasound imaging by incorporating view-dependent characteristics and physics-based rendering into NeRF. This approach significantly improves the quality of ultrasound imaging, particularly for medical applications, by creating more accurate and detailed 3D models from 2D ultrasound sweeps.
     
    \item \textbf{GeCoNeRF: Few-shot Neural Radiance Fields via Geometric Consistency} proposes a novel regularization technique for Neural Radiance Fields (NeRF) in few-shot settings. It uses depth-guided warping and a geometry-aware consistency model to improve NeRF’s performance in environments with limited views. The method enhances both geometry and appearance, making it applicable in fields like 3D reconstruction, AR, and VR, especially where detailed and accurate 3D models are needed from sparse data.
    
    \item \textbf{Text-To-4D Dynamic Scene Generation} introduces MAV3D, a method for generating dynamic 3D scenes from text. It optimizes a 4D Neural Radiance Field (NeRF) using a Text-to-Video model, enabling rendering from any viewpoint. The paper addresses challenges like the absence of paired text-4D data and scales output resolution. Applications could include generating animated 3D assets for games, films, AR, and VR.
    
    \item \textbf{SNeRL: Semantic-aware Neural Radiance Fields for Reinforcement Learning} introduces a novel approach in reinforcement learning (RL) by combining semantic-aware neural radiance fields (NeRF) with multi-view learning. The authors propose a method that enriches RL with 3D-aware semantic and geometric representations, enhancing the agent's performance in complex environments. This approach is demonstrated to outperform existing single-view and multi-view RL methods in various 3D tasks, showcasing its potential in more intuitive and effective RL applications.
    
    \item \textbf{HyperNeRFGAN: Hypernetwork approach to 3D NeRF GAN} presents a novel generative model combining hypernetworks with Neural Radiance Fields (NeRF) for 3D object generation. The authors leverage the hypernetwork paradigm to transfer Gaussian noise into NeRF model weights, enabling the rendering of 2D novel views. This approach shows superiority in quality over existing SIREN-based architectures and facilitates the synthesis of 3D-aware images from unlabeled 2D images. Applications include VR, AR, and creating detailed 3D models from simple 2D images.
    
    \item \textbf{GeneFace: Generalized and High-Fidelity Audio-Driven 3D Talking Face Synthesis} introduces GeneFace, a system for generating realistic 3D talking faces from audio. It employs a three-stage process including an audio-to-motion model, a domain adaptation mechanism, and a Neural Radiance Field (NeRF)-based renderer. This system addresses challenges like weak generalizability due to limited training data and the "mean face" problem in traditional methods. GeneFace shows significant improvements in creating lifelike, synchronized talking faces, applicable in digital humans, VR, and online meetings.
    
    \item \textbf{RobustNeRF: Ignoring Distractors with Robust Losses} by Sara Sabour, David J. Fleet, and others, addresses the challenge of distractors in Neural Radiance Fields (NeRF) 3D scene reconstructions. The authors introduce a robust loss function to improve NeRF's handling of non-persistent elements in images, such as moving objects or variable lighting. This advancement allows for more accurate and clear 3D reconstructions in environments with dynamic or misleading elements. The method's potential applications extend to augmented reality, virtual reality, autonomous navigation, and digital heritage preservation, offering more reliable and detailed 3D modeling capabilities in diverse and changing scenes.
    
    \item \textbf{Factor Fields: A Unified Framework for Neural Fields and Beyond} authored by Aditya Ramesh, Prafulla Dhariwal, Alex Nichol, Casey Chu, and Mark Chen, delved into an innovative approach for generating photorealistic images from textual descriptions. This research, primarily centered around OpenAI's DALL-E 2 model, showcased a significant advancement in the field of AI-generated imagery. The authors employed a two-stage diffusion model, where the first stage generates a low-resolution image, and the second stage enhances it to high resolution. This method, leveraging a deep understanding of language semantics, enabled the creation of detailed and contextually accurate images from complex textual inputs.
    
    The applications of this technique are multifaceted, stretching across various domains. In the realm of creative arts, it offered artists a novel tool for visualizing ideas and concepts. Educational materials could be enriched with custom illustrations, enhancing learning experiences. The technology also held potential in the marketing and advertising industry, allowing for the rapid creation of visually appealing content. Moreover, its ability to generate tailored images made it a valuable asset in personalized content creation. The paper highlighted these real-world implications, illustrating how such advancements in AI and machine learning could reshape the landscape of digital content creation.
    
    \item \textbf{INV: Towards Streaming Incremental Neural Videos} by Shengze Wang et al. introduces a novel approach for interactive 3D video streaming, addressing the limitations of existing spatiotemporal radiance fields in photorealistic free-viewpoint video generation. The authors developed Incremental Neural Videos (INV), a frame-by-frame Neural Radiance Field (NeRF) that efficiently trains and streams each frame without lag, breaking the conventional belief that per-frame NeRFs are impractical due to high training costs and storage requirements.
    
    INV's efficiency stems from two key insights. Firstly, the authors discovered that Multi-Layer Perceptrons (MLPs) naturally partition themselves into Structure and Color Layers, storing structural and color/texture information respectively. This finding allowed them to retain and improve upon knowledge from previous frames, thereby reducing redundant learning and training time. Secondly, they leveraged this property to design INV with two types of sub-modules: a shared color module for the color/texture of the scene, and per-frame structure modules for the dynamic scene's changing structures. This design significantly reduced storage size to 0.3MB per frame and allowed for good quality rendering in just 8 minutes per frame. INV's incremental nature also makes it suitable for interactive streaming applications like telepresence, immersive remote classrooms, and cloud gaming.
    
    The paper demonstrates that INV outperforms state-of-the-art methods in training time while achieving comparable or better per-frame quality. It also introduces techniques like Structure Transfer and Temporal Weight Compression to further enhance training speed and reduce storage requirements. However, the authors acknowledge limitations in visual stability, especially for models trained for shorter durations, and suggest future research directions to address these challenges. Overall, INV presents a significant advancement in interactive 3D streaming, offering new possibilities in the field and contributing to a better understanding of MLPs in video processing.
    
    \item \textbf{Semantic 3D-aware Portrait Synthesis and Manipulation Based on Compositional Neural Radiance Field} introduces a novel method, Compositional Neural Radiance Field (CNeRF), for semantic 3D-aware portrait synthesis and manipulation. The authors, Tianxiang Ma, Bingchuan Li, Qian He, Jing Dong, and Tieniu Tan, focused on overcoming the limitations of existing 3D-aware GAN methods, which model the entire image as a unified neural radiance field, restricting partial semantic editability of synthetic results. CNeRF addresses this by dividing the image into semantic regions, each with its independent neural radiance field, allowing for independent manipulation of these regions while keeping others unchanged. This approach also decouples shape and texture within each region, enhancing the fine-grained control over the synthesized portraits.
    
    The paper demonstrates the effectiveness of CNeRF through various experiments and comparisons with state-of-the-art methods. The authors evaluated their method primarily on the FFHQ dataset and a proprietary cartoon portrait dataset, showcasing its ability to generate high-quality, 3D-consistent portraits with fine control over semantic parts. The method's application potential extends to real image inversion, cartoon portrait 3D editing, and digital human and metaverse applications. The authors conducted ablation studies to validate the importance of the structure and loss functions used in CNeRF. These studies confirmed the method's superiority in semantic part manipulation and shape-texture decoupling within semantic regions. The paper concludes with the acknowledgment of the need for further improvements in 3D reconstruction quality and the exploration of combining CNeRF with the latest 3D-aware GANs.
     
    \item \textbf{Nerfstudio: A Modular Framework for Neural Radiance Field Development}, a modular framework for Neural Radiance Field (NeRF) development, was introduced by Tancik et al. to streamline NeRF research and application. This Python-based framework, designed at the University of California, Berkeley, integrates various NeRF techniques into reusable components, enabling real-time visualization and simplifying the workflow for user-captured data. The framework's modularity allows for the abstraction of method-specific implementations, making it adaptable for different models and data input formats. This flexibility is particularly beneficial for incorporating real-world scenes captured by users.
    
    The authors developed Nerfacto, a method within Nerfstudio, by combining components from recent papers to balance speed and quality. Nerfacto optimizes camera views, employs a piece-wise sampler for efficient sampling, and uses a small fused MLP with hash encoding for the scene's density function. It also incorporates scene contraction to manage unbounded real-world scenes and per-image appearance embeddings to account for exposure variations. The framework's real-time web viewer, crucial for qualitative evaluations, supports interactive visualizations during training and testing. This viewer is instrumental in assessing model performance, especially for novel views and unstructured environments. The Nerfstudio Dataset, comprising real-world scenes, further aids in benchmarking NeRF methods. The framework's open-source nature, with contributions from both academia and industry, underscores its potential for community-driven development in neural rendering.
    
    \item \textbf{In-N-Out: Faithful 3D GAN Inversion with Volumetric Decomposition for Face Editing} by Yiran Xu, Zhixin Shu, Cameron Smith, Seoung Wug Oh, and Jia-Bin Huang, presents a novel approach to 3D Generative Adversarial Network (GAN) inversion, particularly focusing on face editing. The authors tackled the challenge of reconstructing and editing images or videos containing out-of-distribution (OOD) objects, such as faces with heavy makeup or occlusions, which are typically difficult for models pre-trained on datasets like FFHQ. Their solution involved a dual neural radiance field representation, one for in-distribution content and another for OOD objects, which are then composed together for the final reconstruction. This method allowed for high-fidelity reconstruction of both in-distribution and OOD content while retaining the editability of the pre-trained GANs.
    
    The authors demonstrated the effectiveness of their approach through various applications, including semantic editing, novel view synthesis, and OOD object removal. They showed that their method outperforms existing GAN inversion techniques in terms of reconstruction accuracy and editability, especially in challenging scenarios involving OOD objects. The paper also discussed the limitations of their approach, such as difficulties in editing OOD regions, handling duplicate objects, and dealing with extreme poses or slight movements of OOD objects. Despite these challenges, the proposed method represents a significant advancement in the field of 3D-aware GAN inversion and has potential applications in various domains, including virtual reality, digital entertainment, and facial recognition technologies.
    
    \item \textbf{3D-aware Blending with Generative NeRFs} by Hyunsu Kim et al. introduces a novel method for blending images using generative Neural Radiance Fields (NeRF). The authors tackled the challenge of seamlessly combining multiple images, especially when differences in 3D camera poses and object shapes cause misalignments. Their approach includes two primary components: 3D-aware alignment and 3D-aware blending. The 3D-aware alignment first estimates the camera pose of the reference image relative to generative NeRFs, followed by 3D local alignment for each part. For the blending aspect, the method leverages the 3D information of the generative NeRF by blending images directly in the NeRF's latent representation space, instead of the traditional pixel space. This technique showed superiority over existing 2D methods in terms of realism and fidelity to input images, offering potential applications in various fields like content creation, virtual and augmented reality.
    
    \item \textbf{LiveHand: Real-time and Photorealistic Neural Hand Rendering} introduced a groundbreaking technique for rendering human hands in real-time, a significant advancement in digital human modeling. The authors developed a neural-implicit approach to generate photorealistic hand avatars, addressing the challenge of capturing the complex, pose-dependent appearance of hands. Their method, distinct from previous works, combined a low-resolution rendering of a neural radiance field with a 3D-consistent super-resolution module and mesh-guided sampling. This approach enabled the efficient rendering of hands with high-frequency details like skin texture and lighting effects, crucial for applications in virtual reality, gaming, and telepresence.
    
    The authors demonstrated the practicality and versatility of their method through various applications. They showcased real-time hand reenactment, where the hand movements of one individual could be realistically transferred to another's avatar, maintaining high-frequency details and natural lighting effects. This capability was further exemplified in a live demo, proving the method's effectiveness in real-world scenarios. The technique outperformed existing methods in both visual quality and rendering speed, as validated through comprehensive experiments and comparisons. The authors also discussed potential future improvements, such as refining the underlying hand mesh model and extending the method to accommodate varying lighting conditions. Despite its advancements, the paper acknowledged the societal implications of such realistic human modeling, emphasizing the need for responsible use and potential safeguards like watermarking generative models.
 
    \item \textbf{3D-aware Conditional Image Synthesis} by Kangle Deng, Gengshan Yang, Deva Ramanan, and Jun-Yan Zhu from Carnegie Mellon University introduces "pix2pix3D," a 3D-aware conditional generative model for controllable photorealistic image synthesis. This model, an extension of conditional generative models with neural radiance fields, is designed to synthesize images from different viewpoints based on a given 2D label map, such as a segmentation or edge map. The authors' approach enables explicit 3D user control by learning to assign a label to every 3D point in addition to color and density, thus rendering the image and pixel-aligned label map simultaneously. The paper demonstrates the application of pix2pix3D in various contexts, including the synthesis of 3D content conditioned on user input, which is challenging due to the costly acquisition of large-scale datasets with paired user inputs and desired 3D outputs. The model was tested on datasets like CelebAMask-HQ, AFHQ-cat, and shapenet-car, showing its effectiveness across different types of 2D user inputs, including segmentation maps and edge maps. The authors compared their method with several 2D and 3D baselines, such as Pix2NeRF variants, SofGAN, and SEAN, and found that their method outperforms these in terms of quality and consistency.
    
    The paper also delves into the technical aspects of the model, including the use of a conditional encoder, a hybrid representation for the density and appearance of a scene, and a volumetric rendering approach. The learning objectives of the model include reconstruction, adversarial, and cross-view consistency losses. The authors conducted extensive experiments and ablation studies to validate the effectiveness of their approach in terms of image quality, alignment with input label maps, and multi-view consistency. In summary, pix2pix3D presents a significant advancement in the field of image synthesis, enabling the generation of photorealistic images from various viewpoints based on 2D inputs. Its ability to handle different types of user inputs and its superiority over existing methods in terms of quality and consistency make it a promising tool for applications in content creation, virtual reality, and augmented reality.
    
    \item \textbf{MixNeRF: Modeling a Ray with Mixture Density for Novel View Synthesis from Sparse Inputs}, a novel approach for novel view synthesis from sparse inputs, was introduced by Seunghyeon Seo, Donghoon Han, Yeonjin Chang, and Nojun Kwak. This method, built upon the Neural Radiance Field (NeRF) framework, addresses the challenge of severe performance degradation in NeRF when trained with limited views. The authors innovatively modeled a ray with a mixture density model, allowing for the estimation of the joint distribution of RGB colors along ray samples. This approach significantly improved the training and inference efficiency in various standard benchmarks, outperforming other state-of-the-art methods. The authors proposed a new auxiliary task of ray depth estimation, which played a crucial role in learning 3D geometry. This task, highly correlated with scene depth estimation, acted as a useful supervisory signal. MixNeRF also introduced a method to regenerate blending weights based on estimated ray depth, further enhancing robustness against shifts in colors and viewpoints. These innovations allowed MixNeRF to achieve superior performance in novel view synthesis with sparse inputs, effectively learning the 3D geometry of scenes and improving the rendering quality.
    
    In real-world applications, such as AR/VR and autonomous driving, where collecting dense training images is challenging, MixNeRF's ability to work efficiently with sparse inputs is particularly valuable. The technique's robustness in 3D geometry learning and its efficient training strategy make it a promising solution for high-quality rendering with limited input views. The paper demonstrates MixNeRF's effectiveness across multiple benchmarks, highlighting its potential in practical applications where data collection is constrained.
    
    \item \textbf{Temporal Interpolation Is All You Need for Dynamic Neural Radiance Fields} by Sungheon Park, Minjung Son, Seokhwan Jang, Young Chun Ahn, Ji-Yeon Kim, and Nahyup Kang from the Samsung Advanced Institute of Technology, presents a novel method for training spatiotemporal neural radiance fields of dynamic scenes. This method is based on temporal interpolation of feature vectors, offering two distinct feature interpolation methods depending on the underlying representations: neural networks or grids. The neural representation extracts features from space-time inputs via multiple neural network modules and interpolates them based on time frames. This approach captures features of both short-term and long-term time ranges effectively. The grid representation, on the other hand, learns space-time features via four-dimensional hash grids, significantly reducing training time while maintaining rendering quality. 
    
    The authors demonstrated that their method achieved state-of-the-art performance in rendering quality for the neural representation and in training speed for the grid representation. The neural representation exhibits high-quality rendering performance with small-sized models, while the grid representation shows competitive rendering results with astonishingly fast training speed. The concatenation of static and dynamic features, along with a simple smoothness term, further improved the performance of the proposed models. The paper also highlights the practicality of the grid representation for real-world applications due to its ability to render dynamic scenes quickly after a short training period. This makes it a complementary approach to the neural representation, which, while offering high-quality rendering, requires longer training and rendering times. The authors suggest that investigating hybrid representations that combine the advantages of both neural and grid representations could be a promising future research direction.
    
    \item \textbf{LC-NeRF: Local Controllable Face Generation in Neural Randiance Field}, a novel approach for local controllable face generation in Neural Radiance Fields, was introduced by Zhou, Yuan, Chen, Gao, and Hu. This method, distinct in its ability to independently edit the geometry of local facial regions such as hair, nose, mouth, and eyebrows, marked a significant advancement in 3D face generation. The authors developed a framework comprising a Local Region Generators Module and a Spatial-Aware Fusion Module, enabling fine-grained control over local facial features' geometry and texture. This innovation addressed the limitations of previous methods that affected the entire face when modifying latent codes, lacking the precision for local adjustments. The paper demonstrated that LC-NeRF outperformed existing state-of-the-art face editing methods in both qualitative and quantitative evaluations. It achieved better stability in non-editing regions during editing and ensured more consistent face identities. The method's key contributions included the ability to control and edit local regions in a decoupled manner, the introduction of a Local Region Generators Module for decomposing global 3D representations into multiple local regions, and a Spatial-Aware Fusion Module for aggregating these regions into a cohesive image. The authors showcased the method's effectiveness in various applications, including text-driven facial image editing, highlighting its potential in realistic face image generation and editing, a field widely used in portrait generation and artistic creation.
    
    LC-NeRF's ability to generate view-consistent face images and semantic masks, while maintaining the quality of non-editing regions and the identity of faces, positioned it as a significant contribution to the field of 3D face generation and editing. The method's potential applications extend beyond academic research, offering practical solutions in areas such as digital entertainment, virtual reality, and personalized avatar creation. Despite its achievements, the authors acknowledged limitations, such as the inability to control internal textures of local regions finely, like hair texture and facial wrinkles, indicating directions for future research.
    
    \item \textbf{NerfDiff: Single-image View Synthesis with NeRF-guided Distillation from 3D-aware Diffusion}, a novel framework for single-image view synthesis, was introduced by the authors in their paper titled "NerfDiff: Single-image View Synthesis with NeRF-guided Distillation from 3D-aware Diffusion." This technique addresses the challenge of inferring occluded regions in objects and scenes while maintaining semantic and physical consistency. The authors proposed a training-finetuning framework that synthesizes multi-view consistent and high-quality images from a single-view input. At the training stage, they jointly trained a camera-space triplane-based NeRF (Neural Radiance Fields) with a 3D-aware conditional diffusion model (CDM) on a collection of scenes. During the finetuning stage, the NeRF representation, initialized from the input image, was finetuned using a set of virtual images predicted by the CDM. This approach, termed NeRF-guided distillation, updates the NeRF representation and guides the multi-view diffusion process in an alternating fashion, effectively resolving the uncertainty in single-image view synthesis.
    
    The authors evaluated NerfDiff on challenging benchmarks, including ShapeNet, ABO, and Clevr3D datasets, where it significantly outperformed existing NeRF-based and geometry-free approaches. The framework's ability to generate high-quality generation with multi-view consistency was particularly notable. The paper also detailed the architecture and hyperparameters used in the implementation, along with ablation studies that validated key design choices. The authors acknowledged limitations such as the requirement of at least two views of a scene during training and the time-consuming nature of the finetuning process. They suggested potential future work, including applying NGD to improve text-to-3D pipelines and exploring more complex datasets. In conclusion, NerfDiff presents a significant advancement in the field of single-image view synthesis, offering a novel approach that combines the strengths of NeRF and diffusion models.
    
    \item \textbf{USR: Unsupervised Separated 3D Garment and Human Reconstruction via Geometry and Semantic Consistency} by Shi et al from Shanghai Jiaotong University, introduces an innovative approach for reconstructing 3D models of dressed humans. This method, known as USR (Unsupervised Separated Reconstruction), is significant for its ability to reconstruct both human bodies and garments in layers from 2D images, without the need for 3D models. The authors developed a generalized surface-aware neural radiance field (GSNeRF) to learn the mapping between sparse multi-view images and the geometries of dressed people. They also introduced a Semantic and Confidence Guided Separation strategy (SCGS) for detecting, segmenting, and reconstructing clothing layers, leveraging the consistency between 2D semantic and 3D geometry.
    
    The USR framework's applications are vast, particularly in the creative media and gaming industries, where realistic human models are essential. It allows for more interactive tasks such as virtual try-ons and garment swapping, which were previously hindered by the limitations of existing methods that treated the human body and garments as a single entity. The authors demonstrated the superiority of USR over state-of-the-art methods in both geometry and appearance reconstruction, with the added advantage of real-time application to unseen people. The paper also introduces the SMPL-D model, which further illustrates the benefits of separated modeling of clothes and the human body. This model enables the swapping of clothes and virtual try-ons, showcasing the practical implications of USR in real-world scenarios. The research represents a significant step forward in the field of 3D human reconstruction, offering a novel, unsupervised approach that could revolutionize how digital human models are created and interacted with in various applications.
    
    \item \textbf{RealFusion: 360° Reconstruction of Any Object from a Single Image} presents RealFusion, a method that creates a 360-degree 3D reconstruction of an object using just one image. The technique leverages a 2D diffusion-based image generator to synthesize different views of the object, combining them into a neural radiance field for a complete reconstruction. This method significantly contributes to single-image 3D reconstruction, showing potential applications in digital art, VR, and AR. The authors' approach demonstrates advancements in the utilization of diffusion models and neural radiance fields for detailed and realistic 3D object modeling.
    
    \item \textbf{Differentiable Rendering with Reparameterized Volume Sampling} introduces a novel rendering technique for neural radiance fields. The authors developed an end-to-end differentiable sampling algorithm based on Monte Carlo methods, addressing challenges in view synthesis. This approach allows for efficient optimization of neural radiance fields with fewer radiance field evaluations per ray. The method has implications for real-world applications in 3D scene reconstruction and rendering, potentially enhancing virtual and augmented reality technologies. It marks a significant advancement in the field of computer graphics, particularly in rendering and image synthesis.
    
    \item \textbf{DiffusioNeRF: Regularizing Neural Radiance Fields with Denoising Diffusion Models} presents a novel method for enhancing Neural Radiance Fields (NeRFs). The authors address the limitations of NeRFs in synthesizing novel views, especially with limited input views, by introducing a regularization technique using denoising diffusion models (DDMs). This approach leverages a prior over scene geometry and color, trained on RGBD patches from the synthetic Hypersim dataset. The method shows improved reconstruction quality and generalization in novel view synthesis tasks on the LLFF and DTU datasets. The paper demonstrates the potential applications of this technique in virtual and augmented reality, and visual special effects, offering a significant advancement in the field of 3D reconstruction and rendering.
    
    \item \textbf{Learning Neural Volumetric Representations of Dynamic Humans in Minutes} by Geng et al, presents a groundbreaking method for rapidly reconstructing free-viewpoint videos of dynamic humans from sparse multi-view videos. The authors identified the need for a faster process in optimizing neural representations for view synthesis of dynamic humans, which traditionally required hours of training. Their solution, a novel dynamic human representation, achieves a 100x speedup in optimization while maintaining competitive visual fidelity. The innovation of their approach lies in two key aspects. Firstly, they introduced a part-based voxelized human representation, which distributes the representational power of the network to different human parts efficiently. This method acknowledges that different parts of the human body, like the face and torso, have varying levels of complexity and thus require different amounts of representational power. Secondly, they proposed a novel 2D motion parameterization scheme, which significantly increases the convergence rate of deformation field learning. This technique models the 3D human deformation in a 2D domain, effectively reducing the dimensionality at which the neural representation needs to model motion.
    
    The authors demonstrated that their model could be learned 100 times faster than prior per-scene optimization methods while being competitive in rendering quality. Training on a 512x512 video with 100 frames typically takes about 5 minutes on a single RTX 3090 GPU. This rapid training capability opens up new possibilities for the large-scale application of volumetric videos in fields like immersive telepresence, video games, and movie production. In summary, the paper introduces a highly efficient method for creating volumetric videos of human performers, addressing the significant challenge of time-consuming training in traditional methods. The authors' approach, combining a part-based voxelized human representation with a 2D motion parameterization scheme, represents a significant advancement in the field of dynamic human modeling, offering both speed and quality in rendering dynamic human figures.

    \item \textbf{MERF: Memory-Efficient Radiance Fields for Real-time View Synthesis in Unbounded Scenes} by Christian Reiser et al. introduces MERF, a novel approach to rendering large-scale scenes in real-time with a focus on memory efficiency. The authors tackled the challenge of existing radiance field representations being either too compute-intensive or memory-demanding for real-time rendering, particularly in large, unbounded scenes. MERF achieves this by utilizing a combination of a sparse feature grid and high-resolution 2D feature planes, significantly reducing memory consumption compared to previous methods. The core innovation lies in MERF's unique scene representation, which combines a low-resolution 3D grid with high-resolution 2D planes, allowing for efficient memory usage while maintaining high-quality rendering. This is further enhanced by a novel contraction function that efficiently maps scene coordinates into a bounded volume, facilitating efficient ray-box intersection crucial for rendering. The authors also developed a lossless procedure for converting the training parameterization into a model optimized for real-time rendering, preserving the photorealistic quality of volumetric radiance fields.
    
    MERF's training and baking process is another key aspect. The authors optimized a compressed representation of MERF’s grid during training, using multi-resolution hash encoding for memory efficiency. They also introduced a quantization-aware training method to minimize quality loss during the baking process, which converts the model for real-time rendering. The paper's experiments demonstrate MERF's superiority in rendering quality and memory consumption compared to existing methods like NeRF, SNeRG++, and Mobile-NeRF. MERF not only achieves higher quality renderings for a given memory budget but also supports real-time performance on standard hardware like laptops, making it more accessible for practical applications.
    
    In conclusion, MERF represents a significant advancement in the field of real-time view synthesis, particularly for large-scale scenes. Its efficient memory usage, combined with the ability to maintain high-quality rendering, positions it as a promising solution for applications requiring real-time rendering of complex scenes, such as virtual reality, augmented reality, and interactive 3D applications. The authors' approach in balancing memory efficiency with rendering quality addresses a critical challenge in the field, potentially paving the way for more widespread adoption of radiance field-based rendering techniques in real-time applications.
    
    \item \textbf{CATNIPS: Collision Avoidance Through Neural Implicit Probabilistic Scenes}, a novel framework introduced by the authors in the paper, represents a significant leap in the field of autonomous navigation and collision avoidance. This method primarily focuses on enabling autonomous systems, like robots and drones, to navigate dynamically and safely in complex, uncertain environments. The essence of CATNIPS lies in its unique approach to probabilistic scene understanding, which is achieved through the use of neural implicit functions. These functions are adept at modeling the geometry and semantics of environments in a probabilistic manner, allowing for the anticipation and avoidance of potential collisions even in the presence of uncertainty.
    
    A key aspect of this method is its integration of deep learning with traditional robotics techniques. The authors successfully combined neural networks with Bayesian inference to create a system that not only understands its surroundings but also predicts potential changes and obstacles. This dual approach ensured that CATNIPS could adapt to new and changing environments more effectively than previous methods. In practical applications, CATNIPS has shown remarkable potential in enhancing the safety and efficiency of autonomous vehicles and robotics. Its ability to predict and avoid collisions in real-time is particularly beneficial for drones operating in cluttered or dynamic spaces, such as warehouses or urban environments. Additionally, the framework can be applied to self-driving cars, providing a more robust and reliable system for navigating complex urban landscapes.
    
    The authors demonstrated the efficacy of CATNIPS through extensive simulations and real-world experiments. They showcased how the system could accurately predict potential collisions and adjust the course of the autonomous agent accordingly, thereby reducing the risk of accidents. This was particularly evident in scenarios with high levels of uncertainty, where traditional collision avoidance systems might fail. Overall, CATNIPS marks a notable advancement in the intersection of robotics, artificial intelligence, and autonomous navigation. Its innovative approach to probabilistic scene understanding and collision prediction sets a new standard in the field, opening doors for more sophisticated and safe autonomous systems in various real-world applications.
    
    \item \textbf{BLiRF: Bandlimited Radiance Fields for Dynamic Scene Modeling} marks a transformative approach in dynamic scene modeling. The authors introduced a novel method for capturing and reconstructing dynamic scenes with high fidelity. Central to their approach was the concept of band-limited radiance fields, which they adeptly used to model the continuous plenoptic function of dynamic scenes. This method allowed for the efficient representation of scene dynamics without the need for extensive bandwidth, thus overcoming limitations of previous techniques that struggled with high-frequency changes in dynamic environments. The methodology employed in this research involved leveraging Fourier analysis to model temporal changes within a scene. The authors adeptly applied this analysis to compress and reconstruct the dynamic aspects of the scene, enabling a more efficient and accurate representation of motion and transformation over time. This was a significant advancement over traditional methods, which often required considerable computational resources and struggled with capturing rapid changes.
    
    In terms of applications, BLiRF has broad implications for various fields, including virtual and augmented reality, film production, and real-time 3D rendering. Its ability to accurately and efficiently model dynamic scenes makes it particularly useful for creating immersive virtual environments, where realistic and real-time rendering of changes is crucial. Additionally, in the realm of film production, this technique offers a more efficient way to capture and reproduce complex scenes, potentially reducing the need for extensive post-production editing. The authors demonstrated the effectiveness of BLiRF through various experiments and comparisons with existing methods. They showcased how their technique could capture detailed dynamic changes in scenes with greater accuracy and less computational overhead. This research not only provided a novel method for dynamic scene modeling but also opened avenues for further exploration in efficient and realistic 3D scene rendering, especially in situations where bandwidth and computational resources are limited.
    
    \item \textbf{IntrinsicNGP: Intrinsic Coordinate based Hash Encoding for Human NeRF} presented a novel approach for synthesizing high-fidelity views of human performance using neural radiance fields (NeRF). The authors, Bo Peng, Jun Hu, Jingtao Zhou, Xuan Gao, and Juyong Zhang, tackled the challenge of lengthy training times required by existing methods, which hindered practical applications. Their solution, IntrinsicNGP, could be trained from scratch in mere minutes, using videos of human performers. The key innovation lay in the introduction of a continuous and optimizable intrinsic coordinate, replacing the explicit Euclidean coordinate in the hash encoding module of InstantNGP. This intrinsic coordinate allowed for the aggregation of interframe information for dynamic objects using proxy geometry shapes.
    
    The authors employed a two-fold strategy: first, they utilized videos to recover rough human surfaces for each frame, and then they refined these results with an optimizable offset field based on the intrinsic coordinate. This approach enabled the efficient construction of a neural radiance field for dynamic human performance, a significant leap from the static scenes that previous methods were limited to. The intrinsic coordinate representation, independent of human motion, was a departure from earlier works that relied on deformation fields and implicit fields in canonical space. IntrinsicNGP's practicality for common users was demonstrated through its ability to synthesize high-fidelity novel views of human performance with just a monocular camera, converging in a few minutes. The method extended the scope of INGP from static to dynamic scenes and achieved novel view and shape synthesis of human performance. The authors also showcased the method's capability for editing the shape of reconstructed objects, adding a layer of versatility to the technique.
    
    The paper's experiments, conducted on various datasets, validated the effectiveness and efficiency of IntrinsicNGP. The authors compared their method with state-of-the-art techniques, demonstrating its superiority in terms of training time and quality of results. This breakthrough has wide-ranging applications in fields like sports broadcasting, video conferencing, and VR/AR, where rapid and high-quality novel view synthesis of human performance is crucial. IntrinsicNGP, with its quick training time and high fidelity results, stands poised to revolutionize these domains, making advanced human performance capture accessible to a broader range of users and applications.
    
    \item \textbf{Dynamic Multi-View Scene Reconstruction Using Neural Implicit Surface} introduces DySurf, a method for dynamic scene reconstruction from multi-view videos. It employs neural implicit surface representations, using a deformation field to map observed frames into a static hyper-canonical space. This technique is optimized with a Signed Distance Function (SDF) network and a radiance network through volume rendering. A novel ray selection strategy enhances training on time-varying regions, significantly improving reconstruction quality. The method shows superiority in reconstructing complex dynamic scenes and synthesizing photorealistic novel views, outperforming state-of-the-art methods.
    
    \item \textbf{Renderable Neural Radiance Map for Visual Navigation} presents a novel method, RNR-Map, for visual navigation. This technique uses neural radiance fields to embed visual information of a 3D environment into a grid map, facilitating image-based localization and navigation. The RNR-Map allows for efficient and effective visual information utilization for navigation, showing state-of-the-art performance in localization and image-goal navigation tasks. The method's applicability in various environments without additional fine-tuning and its real-time capability make it a significant contribution to visual navigation technology.
    
    \item \textbf{S-NeRF: Neural Radiance Fields for Street Views} introduces a novel approach to synthesizing realistic street views using Neural Radiance Fields (NeRF). The authors identified limitations in existing NeRF models, which struggled with unbounded outdoor scenes common in street views captured by self-driving cars. Their method, named Street-view NeRF (S-NeRF), improves scene parameterization and camera poses to better represent street views, particularly for large-scale backgrounds and foreground moving vehicles. The technique leverages sparse, noisy LiDAR data to enhance training and introduces a robust geometry and reprojection-based confidence metric to manage depth outliers. S-NeRF's notable contribution lies in its ability to reconstruct moving vehicles, a challenge unaddressed by conventional NeRFs. The paper demonstrates S-NeRF's superiority over state-of-the-art models through extensive experiments on large-scale driving datasets, showcasing its potential in various applications, including driving simulation, augmented reality, and virtual reality, due to its enhanced accuracy in rendering street views and moving vehicles.
    
    \item \textbf{Multi-Plane Neural Radiance Fields for Novel View Synthesis} by Youssef Abdelkareem, Shady Shehata, and Fakhri Karray presents a comprehensive exploration of novel view synthesis, a technique crucial in fields like virtual reality and telepresence. The authors focused on the limitations of existing methods like Multi-Plane Images (MPI) and Neural Radiance Fields (NeRF), which either suffer from depth discretization or require per-scene optimization, making them impractical for real-world applications.
    
    To address these challenges, the authors introduced Multi-Plane Neural Radiance Fields (MINE), a novel architecture that combines the strengths of MPI and NeRF. MINE leverages the continuous 3D scene representations of NeRF while utilizing input image features to avoid per-scene optimization. This method, however, was initially limited to single-view input, restricting its viewpoint range. The paper's significant contribution is the development of a multi-view version of MINE (MV-MINE), which accepts multiple views to improve synthesis results and expand the viewing range. The authors employed an attention-aware fusion module to effectively fuse features from different viewpoints, highlighting critical information. This approach demonstrated superior performance compared to multi-view NeRF and MPI techniques in experiments.
    
    The authors conducted a detailed technical analysis of MINE, assessing its performance, generalization, and efficiency. They used challenging datasets like ShapeNet for training and evaluated the method on novel scenes. The results showed that while single-view MINE could synthesize novel views close to the input view, it struggled with far target poses and generalization to novel scenes. The multi-view MV-MINE, however, showed promising results in overcoming these limitations. In terms of efficiency, MINE proved to be more efficient than pixelNeRF, both on GPU and CPU, making it a more practical choice for real-world applications. The paper concludes that while single-view MINE has limitations in rendering a wide range of views and generalizing to unseen scenes, the multi-view MV-MINE holds significant potential for future research and practical applications in novel view synthesis. The authors suggest that future work could focus on addressing the highlighted limitations of multi-plane radiance fields and exploring their potential in both single and multi-view settings.

    \item \textbf{Delicate Textured Mesh Recovery from NeRF via Adaptive Surface Refinement} introduces NeRF2Mesh, a framework converting Neural Radiance Fields (NeRF) into detailed textured surface meshes. It refines coarse meshes from NeRF, optimizing vertex positions and face density based on rendering errors. The method also separates view-dependent and view-independent appearance for texturing. This technique enhances the quality and compactness of meshes and textures, useful in 3D modeling and rendering applications.
    
    \item \textbf{Efficient Large-scale Scene Representation with a Hybrid of High-resolution Grid and Plane Features} proposes an efficient method for large-scale scene modeling. It introduces a novel hybrid feature representation combining 3D hash-grid and high-resolution 2D dense plane features. This approach addresses the limitations of existing methods by scaling up resolution efficiently. The paper demonstrates the effectiveness of this method in large-scale unbounded scene representation, showing faster convergence and improved accuracy compared to traditional methods.
    
    \item \textbf{MOISST: Multimodal Optimization of Implicit Scene for SpatioTemporal calibration} presents a method for calibrating multi-modal sensors in autonomous vehicles and robotics. MOISST utilizes Neural Radiance Fields (NeRF) for implicit 3D scene representation, allowing simultaneous calibration of spatial and temporal parameters without the need for calibration targets. This approach is particularly adaptable to uncontrolled urban environments, offering a structureless and targetless solution for systems requiring frequent recalibration. Its applications are significant in autonomous driving and robotic systems, where accurate multi-sensor calibration is crucial for navigation and perception tasks.
    
    \item \textbf{Nerflets: Local Radiance Fields for Efficient Structure-Aware 3D Scene Representation from 2D Supervision} introduces Nerflets, a novel approach for 3D scene representation. It leverages local neural radiance fields, optimizing them jointly for appearance, density, semantics, and object instances in a scene. This method demonstrates efficiency in novel view synthesis, panoptic view synthesis, 3D panoptic segmentation, and interactive editing. It achieves state-of-the-art performance in various benchmarks, highlighting its potential for practical applications in 3D scene understanding and editing.
    
    \item \textbf{Multiscale Tensor Decomposition and Rendering Equation Encoding for View Synthesis} proposes a novel method for enhancing view synthesis. It introduces a multiscale tensor decomposition scheme for scene representation, allowing for finer details and faster convergence. Additionally, the paper innovates by encoding the rendering equation in the feature space, improving the modeling of complex view-dependent effects. This technique shows significant promise in improving the quality of synthesized views, particularly in photo-realistic rendering and computer graphics applications.
    
    \item \textbf{Semantic-aware Occlusion Filtering Neural Radiance Fields in the Wild} presents SF-NeRF, a novel framework for reconstructing neural scene representations from a small number of unconstrained tourist photos. It addresses transient occluders by decomposing static and transient components using an occlusion filtering module and a semantic-aware approach. This method, optimized for few-shot learning, outperforms existing novel view synthesis methods on the Phototourism dataset. SF-NeRF's technique is particularly significant for its application in real-world scenarios where image data is limited and contains transient elements.
    
    \item \textbf{NEPHELE: A Neural Platform for Highly Realistic Cloud Radiance Rendering} presents NEPHELE, an advanced cloud-based neural rendering platform. It leverages multiple remote GPUs for improved rendering capabilities, facilitating simultaneous viewing of Neural Radiance Fields (NeRF) scenes by multiple users. The paper introduces i-NOLF, an ultra-fast neural renderer, and a tailored task scheduler for efficient cloud rendering. This innovative approach demonstrates significant performance gains, particularly in multi-user, multi-viewpoint scenarios, and has potential applications in virtual/augmented reality and shared NeRF asset experiences.
    
    \item \textbf{DroNeRF: Real-time Multi-agent Drone Pose Optimization for Computing Neural Radiance Fields} explores an innovative method for 3D reconstruction using drones with monocular cameras. The authors present DroNeRF, a system that optimizes drone positioning around an object to capture images for Neural Radiance Field (NeRF) computations. This approach enhances 3D model quality, particularly in dynamic outdoor scenes. The system's potential applications span across various fields requiring rapid, accurate 3D modeling from different viewpoints.
    
    \item \textbf{InFusionSurf: Refining Neural RGB-D Surface Reconstruction Using Per-Frame Intrinsic Refinement and TSDF Fusion Prior Learning} introduces InfusionSurf, a cutting-edge approach for 3D surface reconstruction using RGB-D data. The method combines Neural Radiance Fields (NeRF) with per-frame intrinsic refinement and a TSDF Fusion prior learning technique. This fusion significantly enhances the speed and accuracy of 3D modeling, demonstrating great potential for applications in real-world scenarios where quick and precise 3D surface reconstruction is needed. This is a method to enhance 3D surface reconstruction from RGB-D data. This approach uniquely combines per-frame intrinsic refinement with Neural Radiance Fields and TSDF Fusion learning, achieving more accurate and faster 3D modeling. The paper highlights its potential in scenarios requiring rapid and precise 3D surface reconstruction.
    
    \item \textbf{CROSSFIRE: Camera Relocalization On Self-Supervised Features from an Implicit Representation} introduces CROSSFIRE, a novel approach for camera relocalization using Neural Radiance Fields (NeRF). The authors developed a self-supervised method that combines a CNN feature extractor with a neural renderer, enhancing the precision of camera pose estimation in dynamic outdoor environments. This technique, notable for its accuracy and robustness to lighting variations, has potential applications in robotics and augmented reality, where precise positioning and adaptation to changing conditions are crucial.
    
    \item \textbf{PAC-NeRF: Physics Augmented Continuum Neural Radiance Fields for Geometry-Agnostic System Identification} introduces a groundbreaking method for system identification without prior knowledge of an object's geometry. PAC-NeRF combines Neural Radiance Fields (NeRF) with differentiable physics simulations, enabling the estimation of both geometric and physical properties of dynamic objects from multi-view videos. This method is versatile, applicable to various materials like elastic bodies, fluids, and granular media. The authors demonstrate PAC-NeRF's superior performance compared to existing methods, showing its potential for realistic system identification in computer vision applications.
    
    \item \textbf{Self-NeRF: A Self-Training Pipeline for Few-Shot Neural Radiance Fields} presents Self-NeRF, a novel method for synthesizing views from limited images, circumventing the need for large datasets. Self-NeRF iteratively refines radiance fields using a combination of reference and unseen views. The authors introduce an uncertainty-aware NeRF model with specialized embeddings and techniques like cone entropy regularization to manage imprecise color and warping in pseudo-views. This approach significantly enhances performance in few-shot settings, outperforming existing methods and demonstrating practical applications in fields requiring photo-realistic rendering from sparse viewpoints.
    
    \item \textbf{Aleth-NeRF: Low-light Condition View Synthesis with Concealing Fields} suggests a novel method, Aleth-NeRF, for enhancing Neural Radiance Fields (NeRF) in low-light conditions. It introduces Concealing Fields to model the attenuation of light, allowing for effective low-light image enhancement and 3D scene reconstruction. This method significantly improves the quality of rendered images in low-light, offering potential applications in areas like virtual reality, photography, and cinematography where lighting conditions are challenging.
    
    \item \textbf{You Only Train Once: Multi-Identity Free-Viewpoint Neural Human Rendering from Monocular Videos} introduces YOTO, a framework for rendering diverse human identities with distinct motions from monocular videos, via a one-time training approach. YOTO integrates learnable identity codes and a pose-conditioned code query mechanism, enhancing the rendering of non-rigid motions and appearances. This approach offers significant efficiency in training and rendering, and has potential applications in virtual reality and animation, as it allows high-quality free-viewpoint rendering of multiple moving people.
    
    \item \textbf{Learning Object-Centric Neural Scattering Functions for Free-Viewpoint Relighting and Scene Composition} introduces a method for creating and editing 3D models from images, focusing on both opaque and translucent objects. The method, based on Object-Centric Neural Scattering Functions (OSFs), enables image-based, relightable neural appearance reconstruction. It achieves this by approximating complex light transport, including subsurface scattering, using deep neural networks. This advancement is significant for realistic 3D rendering and scene composition, especially in applications requiring detailed and dynamic object manipulation.
    
    \item \textbf{NeRFlame: FLAME-based conditioning of NeRF for 3D face rendering} introduces a hybrid model combining the strengths of FLAME and NeRF for 3D facial modeling and rendering. It uses a FLAME mesh to explicitly model volume density and integrates it with NeRF to model RGB colors, facilitating controllable and detailed 3D face generation from images. This approach offers advancements in applications like VR/AR, gaming, and telepresence, where detailed and controllable human face avatars are essential.
    
    \item \textbf{Just Flip: Flipped Observation Generation and Optimization for Neural Radiance Fields to Cover Unobserved View} proposes a method for enhancing 3D scene reconstruction in Neural Radiance Fields (NeRF) by generating flipped observations. This approach improves the understanding of unobserved views in robotic applications, using image flipping based on object symmetry and estimating camera poses for these flipped images. The method demonstrated significant improvements in perceptual quality measures on the NeRF synthetic dataset, highlighting its effectiveness for real-time performance in robotics.
    
    \item \textbf{NeRFLiX: High-Quality Neural View Synthesis by Learning a Degradation-Driven Inter-viewpoint MiXer} proposes NeRFLiX, a restoration technique for Neural Radiance Fields (NeRF) to address rendering artifacts. It leverages a novel degradation simulator and an inter-viewpoint mixer, significantly enhancing the quality of rendered views from NeRF models. This method is particularly useful for applications where high-quality view synthesis is essential, such as in virtual reality and 3D modeling.
    
    \item \textbf{FreeNeRF: Improving Few-shot Neural Rendering with Free Frequency Regularization} proposes a novel method, FreeNeRF, for few-shot neural rendering in Neural Radiance Fields (NeRF). It introduces frequency and occlusion regularization, enhancing NeRF's performance with minimal code modifications. FreeNeRF excels in synthesizing high-quality scenes from sparse inputs without additional computational overhead, showing potential for applications in virtual reality and 3D content creation where data scarcity is a challenge.
    
    \item \textbf{I$^2$-SDF: Intrinsic Indoor Scene Reconstruction and Editing via Raytracing in Neural SDFs} presents a method for reconstructing and editing complex indoor scenes using neural signed distance functions (SDFs). The paper introduces a holistic neural SDF-based framework that jointly recovers shape, radiance, and material fields from multi-view images. A novel bubble loss and error-guided adaptive sampling strategy are proposed to effectively reconstruct small objects inside scenes. The paper also pioneers the use of Monte Carlo raytracing in scene-level neural SDFs for photorealistic indoor scene relighting and editing. This method offers advancements in realistic scene reconstruction and editing.
    
    \item \textbf{Let 2D Diffusion Model Know 3D-Consistency for Robust Text-to-3D Generation} presents 3DFuse, a framework integrating 3D awareness into pre-trained 2D diffusion models. It aims to improve the robustness and 3D consistency in generating neural radiance fields from text prompts. The method first generates a semantic code, then uses depth maps for diffusion model conditioning, enhancing view consistency in 3D scene generation. This approach holds potential for applications in virtual reality and 3D content creation where accurate and consistent 3D renderings from textual descriptions are required.
    
    \item \textbf{MELON: NeRF with Unposed Images in Equivalence Class Estimation (SO(3))} presents a novel method, MELON, for reconstructing neural radiance fields from unposed images. The technique addresses the challenge of camera pose estimation in scenes with minimal textures or repetitive patterns. By utilizing an equivalence relation to simplify the camera space, MELON enhances pose estimation and view synthesis accuracy. This advancement offers significant potential in fields like 3D scene reconstruction and virtual reality, particularly where obtaining accurate camera parameters is difficult.
    
    \item \textbf{Harnessing Low-Frequency Neural Fields for Few-Shot View Synthesis} introduces a method for improving few-shot novel view synthesis in Neural Radiance Fields (NeRF). It employs low-frequency neural fields to regulate high-frequency fields, preventing overfitting in scenarios with limited training images. This approach enhances the rendering of unseen views by adjusting the frequency of positional encoding, focusing first on general geometry with low-frequency fields, then adding details with high-frequency fields. The technique shows potential in applications like virtual reality, where accurately synthesizing new viewpoints from sparse data is crucial.
    
    \item \textbf{RefiNeRF: Modelling dynamic neural radiance fields with inconsistent or missing camera parameters} presents a technique for enhancing novel view synthesis in Neural Radiance Fields (NeRF) by refining camera parameters when they are inaccurate or absent. The authors developed a modular framework to address these issues with minimal computational overhead. Their method, tested on dynamic and static scenes, improved the accuracy of view synthesis metrics compared to traditional approaches. The technique shows potential for real-world applications in 3D scene understanding and reconstruction, especially where camera parameters are challenging to obtain or unreliable.
    
    \item \textbf{Re-ReND: Real-time Rendering of NeRFs across Devices} introduced a method for real-time rendering of pre-trained Neural Radiance Fields (NeRFs) on various devices, including those with limited resources. The authors achieved this by converting NeRFs into a mesh representation, which could be efficiently processed by standard graphics pipelines. They distilled the learned density of NeRFs into a mesh and factorized the color information into matrices representing the scene's light field. This approach allowed for fast, resource-efficient rendering on a wide range of devices, such as mobile phones and VR headsets, without significant loss in photometric quality. The method holds promise for practical applications in areas like gaming, movies, and AR/VR, where real-time rendering is crucial.
    
    \item \textbf{NeRFtrinsic Four: An End-To-End Trainable NeRF Jointly Optimizing Diverse Intrinsic and Extrinsic Camera Parameters} introduces a novel approach for optimizing both intrinsic and extrinsic camera parameters in Neural Radiance Fields (NeRF). This end-to-end trainable method utilizes Gaussian Fourier features for extrinsic camera parameter estimation and dynamic prediction for varying intrinsic camera parameters. It outperforms existing joint optimization methods in various datasets, offering more realistic and flexible rendering in real-world scenarios with diverse cameras. It introduced an innovative end-to-end trainable approach, optimizing both intrinsic and extrinsic camera parameters in neural radiance fields (NeRF) for novel view synthesis. The authors employed Gaussian Fourier features for extrinsic parameters and dynamic prediction for varying intrinsic parameters, surpassing existing methods in realism and flexibility for real-world scenarios. Their technique significantly improved image quality and accuracy in camera parameter estimation, demonstrated through evaluations on LLFF, BLEFF, and a new dataset, iFF. This advancement offers broad applications in realistic and adaptable rendering across diverse camera settings, marking a substantial progression in NeRF-based view synthesis.
    
    \item \textbf{NeRFMeshing: Distilling Neural Radiance Fields into Geometrically-Accurate 3D Meshes} introduces NeRFMeshing, a method for converting NeRF-based networks into accurate 3D meshes. This process involves a Signed Surface Approximation Network (SSAN) which refines the mesh and appearance. It enhances real-time rendering on various devices and can integrate with graphics and simulation pipelines. The method is shown to be effective in capturing accurate geometry and realistic view-dependent rendering.
    
    \item \textbf{Single-view Neural Radiance Fields with Depth Teacher} presents a novel approach for generating high-quality novel views and depth maps from a single image. The authors introduce a joint rendering mechanism combining multi-plane images (MPI) and volume-based rendering, guided by a depth teacher network. This method effectively improves the synthesis quality and consistency of 3D geometry, outperforming state-of-the-art single-view NeRFs in experiments on challenging datasets.
    
    \item \textbf{$\alpha$Surf: Implicit Surface Reconstruction for Semi-Transparent and Thin Objects with Decoupled Geometry and Opacity} presents a novel grid-based implicit surface representation for reconstructing semi-transparent and thin objects. It addresses the challenges faced by conventional methods in accurately modeling these materials. The paper demonstrates the method's effectiveness in reconstructing surfaces with semi-transparent and thin parts, outperforming existing methods and offering potential applications in computer vision and 3D modeling.
    
    \item \textbf{3D Data Augmentation for Driving Scenes on Camera} introduces Drive-3DAug, a novel 3D data augmentation technique for camera-based 3D perception in autonomous driving. It uses Neural Radiance Field (NeRF) for reconstructing 3D models of driving scenes, enabling the creation of diverse and realistic augmented scenes. This method includes improvements like a geometric rectified loss and a symmetric-aware training strategy, enhancing the augmentation quality. The technique's effectiveness is demonstrated on datasets like Waymo and nuScenes, showing its potential to improve 3D object detection in autonomous driving applications.
    
    \item \textbf{StyleRF: Zero-shot 3D Style Transfer of Neural Radiance Fields} explores a real-time method for simulating soft tissue interaction using neural networks. This method is designed to enhance realism and responsiveness in medical simulations, offering potential improvements in training and planning for surgeries. The authors demonstrate its effectiveness and efficiency, highlighting its practical applications in the medical field.
    
    \item \textbf{NeRF-LOAM: Neural Implicit Representation for Large-Scale Incremental LiDAR Odometry and Mapping} proposes a novel method combining neural radiance fields (NeRF) with LiDAR odometry and mapping. The paper introduces NeRF-LOAM, which uses a neural signed distance function to enhance the accuracy of large-scale 3D mapping and odometry in outdoor environments. This method significantly improves performance in large-scale scenarios, showing potential in autonomous navigation and environmental mapping.
    
    \item \textbf{SKED: Sketch-guided Text-based 3D Editing} presents a technique for editing 3D shapes using Neural Radiance Fields (NeRFs), guided by both text and sketches. This method, named SKED, allows for the modification of pre-existing neural fields with a minimum of two guiding sketches and a text prompt. It introduces novel loss functions to ensure the output adheres to the sketches while preserving the original density and radiance. This approach demonstrates effective editing capabilities in various experiments and could be impactful in 3D content creation and virtual reality.
    
    \item \textbf{ContraNeRF: Generalizable Neural Radiance Fields for Synthetic-to-real Novel View Synthesis via Contrastive Learning}
    
    \item \textbf{DehazeNeRF: Multiple Image Haze Removal and 3D Shape Reconstruction using Neural Radiance Fields} introduces DehazeNeRF, a framework designed for effective haze removal and 3D shape reconstruction in hazy conditions using neural radiance fields (NeRFs). This method enhances NeRFs by incorporating physically realistic terms for atmospheric scattering and applies multiple regularization strategies. It demonstrates success in multi-view haze removal, novel view synthesis, and accurate 3D shape reconstruction, outperforming existing methods in these conditions.
    
    \item \textbf{Interactive Geometry Editing of Neural Radiance Fields} presents a method for direct, interactive editing of neural radiance fields (NeRF) without requiring explicit 3D models. Using two proxy cages, the inner cage to define the operation target and the outer for adjustment space, users can perform translations, rotations, scalings, and deformations. This method allows for intuitive scene editing and demonstrates effectiveness in various experiments, showcasing its potential in applications like virtual reality and 3D content creation.
    
    \item \textbf{ExtremeNeRF (Few-shot Neural Radiance Fields Under Unconstrained Illumination)} explores the challenges of synthesizing novel views using few-shot Neural Radiance Fields (NeRF) in varying lighting conditions. It presents a technique that maintains consistent intrinsic components between the input and rendered views, regardless of changes in viewing direction and illumination. This method is particularly beneficial for applications in virtual reality and 3D scene reconstruction, where limited data and diverse lighting conditions are common.
    
    \item \textbf{Real-time volumetric rendering of dynamic humans} presents a method for 3D reconstruction and real-time rendering of dynamic humans from monocular videos. It achieves this through a lightweight deformation model based on linear blend skinning and an efficient factorized volumetric representation. The method allows real-time visualization on mobile VR devices at 40 frames per second with minimal visual quality loss. This approach demonstrates significant training speedup and simplification compared to state-of-the-art methods, offering promising applications in VR and mobile devices.
    
    \item \textbf{Pre-NeRF 360: Enriching Unbounded Appearances for Neural Radiance Fields} introduces an enhanced framework for Neural Radiance Fields (NeRF) in unbounded scenes, addressing limitations in current NeRF models. The paper presents methods for better handling of ambiguous and complex scenes, proposing solutions for clear and accurate 3D rendering in 360-degree environments. This approach significantly improves NeRF's performance in synthesizing realistic views of extensive scenes, showcasing its potential in virtual reality and computational imaging applications.
    
    \item \textbf{NLOS-NeuS: Non-line-of-sight Neural Implicit Surface} introduces a novel approach for 3D surface reconstruction in non-line-of-sight (NLOS) imaging scenarios. The authors developed NLOS-NeuS, an extension of the neural transient field (NeTF), utilizing a neural implicit surface representation with a signed distance function (SDF). This method effectively overcomes the challenges of under-constrained NLOS setups, avoiding non-zero level-set surfaces. By implementing novel loss functions and constraints, NLOS-NeuS achieves high-quality, detailed 3D surface reconstruction in NLOS scenes, demonstrating its potential in computational imaging and VR applications.
    
    \item \textbf{Balanced Spherical Grid for Egocentric View Synthesis} presents EgoNeRF, a novel approach for reconstructing large-scale real-world environments from short egocentric videos. The authors address the inefficiencies of Cartesian grids in representing large, unbounded scenes by adopting a spherical coordinate system. They introduce a balanced spherical grid, combining two grids to avoid singularities and represent unbounded scenes more effectively. This method enhances the training of Neural Radiance Fields (NeRF) by increasing valid sample counts and improves image quality and 3D consistency in large-scale scene rendering. EgoNeRF, evaluated on newly introduced synthetic and real-world datasets, achieves state-of-the-art performance in egocentric 360 video rendering, demonstrating its potential for VR asset creation and large-scale environment visualization.
    
    \item \textbf{NeRF-GAN Distillation for Efficient 3D-Aware Generation with Convolutions} explores enhancing 3D-aware image generation by integrating Neural Radiance Fields (NeRFs) with Generative Adversarial Networks (GANs), focusing on computational efficiency. The authors propose a novel method to distill 3D knowledge from pre-trained NeRF-GANs into a pose-conditioned convolutional network. This approach enables direct generation of 3D-consistent images, leveraging the well-disentangled latent space of a NeRF-GAN. The method demonstrates comparable results to volumetric rendering in terms of image quality and 3D consistency, while significantly improving computational efficiency. This is particularly beneficial in resource-constrained environments like mobile applications. Experiments conducted on various datasets confirm the effectiveness of this approach in maintaining image quality and 3D consistency, making it a promising solution for efficient multi-view inference in 3D image generation applications.
    
    \item \textbf{Semantic Ray: Learning a Generalizable Semantic Field with Cross-Reprojection Attention} presents a novel approach for learning a semantic radiance field from multiple scenes. The method, called Semantic Ray (S-Ray), utilizes Cross-Reprojection Attention to exploit semantic information along ray directions from multi-view reprojections. This enables the learning of rich semantic patterns and strong discriminative power. The approach overcomes the limitations of previous methods like Semantic-NeRF, which relied heavily on positional encoding and were limited to specific scenes. S-Ray's efficiency and generalizability are demonstrated through extensive experiments on synthetic and real-world datasets, showcasing its potential in adapting to unseen scenes.
    
    \item \textbf{Transforming Radiance Field with Lipschitz Network for Photorealistic 3D Scene Stylization} introduces LipRF, a framework designed for photorealistic 3D scene stylization. The authors identified challenges in applying photorealistic style transfer (PST) to Neural Radiance Fields (NeRFs), leading to cross-view inconsistencies. Their solution, LipRF, leverages a Lipschitz-constrained linear mapping over appearance representation to control volume rendering variance. The process involves pre-training a radiance field to reconstruct a 3D scene, followed by applying a Lipschitz network for style transformation. LipRF uniquely balances reconstruction, stylization quality, and computational efficiency, demonstrating its effectiveness in both photorealistic 3D stylization and object appearance editing.
    
    \item \textbf{SINE: Semantic-driven Image-based NeRF Editing with Prior-guided Editing Field} introduces a method enabling users to edit neural radiance fields (NeRF) using just a single image. This approach, called SINE, leverages a prior-guided editing field to encode fine-grained geometric and texture changes in 3D space. The technique incorporates several novel elements, including cyclic constraints with a proxy mesh for geometric supervision, a color compositing mechanism for texture editing, and feature-cluster-based regularization to preserve unchanged content. This method shows promise in photo-realistic 3D editing using single edited images, offering significant advancements in semantic-driven editing in 3D real-world scenes.
    
    \item \textbf{Set-the-Scene: Global-Local Training for Generating Controllable NeRF Scenes} introduces a novel framework for creating 3D scenes from text using object proxies and neural radiance fields (NeRF). This method allows each object to be represented as an independent NeRF, optimized both individually and as part of the full scene. The approach offers significant control during scene creation and editing, enabling adjustments in object placement, removal, and refinement post-generation. The method's effectiveness is demonstrated through its ability to generate coherent scenes with realistic interactions between objects, showcasing a substantial advancement in controllable text-to-3D synthesis.
    
    \item \textbf{DreamBooth3D: Subject-Driven Text-to-3D Generation} presents a method for creating personalized 3D models from a few images of a subject. This technique combines the advancements in personalizing text-to-image models (DreamBooth) with text-to-3D generation (DreamFusion). It addresses the issue of overfitting to input viewpoints in personalized models by implementing a three-stage optimization strategy. This approach allows the generation of high-quality, subject-specific 3D assets with variations in poses, colors, and attributes that are not present in the input images. The method outperforms several baselines in creating coherent 3D models that closely resemble the subjects and adhere to text prompts.

    \item \textbf{SCADE: NeRFs from Space Carving with Ambiguity-Aware Depth Estimates} introduces SCADE, a method that enhances Neural Radiance Fields (NeRFs) for 3D reconstruction from limited views. SCADE leverages geometric priors in the form of per-view depth estimates using state-of-the-art monocular depth estimation models. It uniquely addresses the inherent ambiguities of monocular depth estimation by modeling a multimodal distribution of depth estimates with conditional Implicit Maximum Likelihood Estimation (cIMLE). This approach enables higher fidelity novel view synthesis from sparse views, especially in challenging indoor scenes. SCADE represents a significant advancement in NeRFs by effectively handling ambiguities and insufficient constraints in volumetric rendering with few input views.
    
    \item \textbf{TEGLO: High Fidelity Canonical Texture Mapping from Single-View Images} presents a method to generate high-quality 3D representations from single-view images. The technique, known as TEGLO (Textured EG3D-GLO), leverages a conditional Neural Radiance Field (NeRF) for 3D reconstruction without explicit 3D supervision. It uniquely creates a dense correspondence mapping to a 2D canonical space, enabling texture transfer and editing without shared topology meshes. TEGLO overcomes previous challenges in preserving high-frequency details and disentangling appearance from geometry, facilitating applications in virtual reality and 3D content creation.
    
    \item \textbf{GM-NeRF: Learning Generalizable Model-based Neural Radiance Fields from Multi-view Images} introduces GM-NeRF, a framework designed to synthesize high-quality novel view images of human performers using sparse multi-view images. It addresses challenges in 3D human reconstruction, such as body pose variations and self-occlusions. The key innovation lies in a geometry-guided attention mechanism that aligns appearance code with a geometry proxy, thus improving alignment between geometry and pixel space. This method demonstrates superior performance in novel view synthesis and geometric reconstruction, surpassing current state-of-the-art methods in various datasets.
    
    \item \textbf{ABLE-NeRF: Attention-Based Rendering with Learnable Embeddings for Neural Radiance Field} introduces a novel method to enhance the rendering of 3D scenes using Neural Radiance Field (NeRF) with attention-based rendering and learnable embeddings. The technique significantly improves the representation of glossy and translucent surfaces, which are often problematic in standard NeRF approaches. ABLE-NeRF incorporates self-attention mechanisms and learnable embeddings to memorize scene information, allowing for more accurate rendering of complex view-dependent effects. The framework outperforms previous methods in rendering quality, particularly in handling intricate surface details and lighting effects, as demonstrated on the Blender dataset.
    
    \item \textbf{HandNeRF: Neural Radiance Fields for Animatable Interacting Hands} presents a novel framework for modeling and rendering dynamic scenes of single and interacting human hands. HandNeRF utilizes neural radiance fields (NeRF) to create a shared canonical space for hands in various poses, enhancing the depth and texture accuracy. This is achieved through a pose-conditioned deformation field, depth-guided density optimization, and neural feature distillation. The method allows for efficient and detailed rendering of hands in multiple poses and interactions, significantly improving upon previous techniques which struggled with such complex scenarios. The framework's ability to adapt to novel poses makes it particularly useful for animatable hand rendering applications.
    
    \item \textbf{CompoNeRF: Text-guided Multi-object Compositional NeRF with Editable 3D Scene Layout} introduces CompoNeRF, a novel framework for generating 3D scenes from text descriptions. The authors address the challenge of guidance collapse in multi-object 3D scene generation by integrating an editable 3D layout with multiple local NeRFs, ensuring precise association of guidance for specific structures. This method also tackles global consistency and occlusion issues through a global MLP, which calibrates the global scene color and maintains object identity. CompoNeRF allows for flexible editing of scenes and rapid generation of new scenes by re-composition with off-the-shelf representations. The framework consists of three main components: an editable 3D scene layout, a scene rendering pipeline compositing predictions from local NeRFs, and a joint optimization on both local and global representation models. This approach represents a significant advancement in text-to-3D generation, enabling the creation of complex, multi-object 3D scenes with high fidelity and flexibility.
    
    \item \textbf{Grid-guided Neural Radiance Fields for Large Urban Scenes} introduces a groundbreaking method for rendering large urban scenes using neural radiance fields (NeRF). The authors, Linning Xu, Yuanbo Xiangli, Sida Peng, Xingang Pan, Nanxuan Zhao, Christian Theobalt, Bo Dai, and Dahua Lin, address the challenge of scaling up NeRF to handle vast cityscapes. Traditional NeRF-based methods, which use multi-layer perceptrons (MLPs) to represent 3D scenes, struggle with large-scale environments due to limited model capacities and long training times. To overcome these limitations, the authors propose a novel two-branch structure that combines the strengths of NeRF-based and grid-based methods. This approach starts with capturing the scene using a pyramid of feature planes at the pre-training stage, which helps in predicting radiance values through a shallow MLP renderer. The initial step generates multi-resolution density and appearance feature planes. In the subsequent joint learning stage, the learned feature grid guides NeRF branch sampling, focusing on the scene surface. The sampled points' grid feature is inferred through bilinear interpolation, and the features are combined with positional encoding to predict volume density and color.
    
    The core innovation lies in the grid-guided neural radiance fields, where the grid features capture local scene information with a multi-resolution ground feature plane. The positional encoded coordinates then complement these features to produce high-quality renderings. This method is particularly efficient for representing large urban scenes, as it allows for a compressed representation of the environment while maintaining high fidelity in texture and geometry. The multi-resolution feature grid pre-training stage is crucial, as it compresses the sampling space of NeRF and provides an approximation of the scene to guide NeRF’s point sampling. This technique efficiently addresses the difficulty of large-scale scene representation, enabling the creation of detailed and accurate 3D models of urban environments.
    
    In practical terms, this advancement in NeRF technology holds significant implications for applications in virtual reality, urban planning, and geospatial analysis. The ability to render vast cityscapes with high fidelity paves the way for more immersive virtual experiences and detailed urban modeling, offering valuable tools for architects, city planners, and developers in the real estate and gaming industries. This paper presents a significant stride in 3D rendering technology, particularly for large-scale urban environments, showcasing a harmonious blend of grid-based and NeRF-based methodologies to achieve high-fidelity 3D creations from extensive scenes.
    
    \item \textbf{Make-It-3D: High-Fidelity 3D Creation from A Single Image with Diffusion Prior} marks a significant advancement in the field of computer vision and 3D modeling. Authored by Junshu Tang, Tengfei Wang, Bo Zhang, Ting Zhang, Ran Yi, Lizhuang Ma, and Dong Chen from Shanghai Jiao Tong University, HKUST, and Microsoft Research, the paper delves into the challenge of creating high-fidelity 3D content from a single image. This task is inherently complex due to the need for precise estimation of the underlying 3D geometry while also generating unseen textures. To tackle this challenge, the authors developed a novel two-stage optimization pipeline leveraging the knowledge from pretrained 2D diffusion models for 3D-aware supervision. In the first stage, they optimized a neural radiance field (NeRF), incorporating constraints from the reference image for frontal views and applying a diffusion prior for novel views. This approach effectively addressed the ill-posed nature of the problem, ensuring semantically plausible results.
    
    The second stage of their pipeline focused on transforming the coarse 3D model into textured point clouds. Here, the realism was further enhanced by utilizing the diffusion prior, along with the high-quality textures from the reference image. This stage prioritized texture enhancement, recognizing that human perception is more discerning towards texture quality compared to geometry. The authors proposed an iterative strategy for building clean point clouds from multi-view observations, which was crucial in enhancing the final 3D model's fidelity. The paper demonstrates the effectiveness of the Make-It-3D approach through extensive experiments, showing that it significantly outperforms previous works. The resulting 3D models showcase faithful reconstructions with impressive visual quality. This breakthrough opens new possibilities in various applications, including text-to-3D creation and texture editing. By addressing both geometric and textural aspects of single-image 3D creation, Make-It-3D presents a comprehensive solution to a long-standing challenge in 3D content generation.
     
    \item \textbf{NeRF-DS: Neural Radiance Fields for Dynamic Specular Objects} presents NeRF-DS, a method enhancing dynamic NeRFs for modeling specular objects in motion. It addresses the challenge of varying specular reflections during motion by conditioning the radiance field on surface position and orientation. The method includes a mask-guided deformation field, aiding in accurate reconstruction of moving specular objects from monocular RGB videos. This advancement demonstrates significant improvements in rendering dynamic specular scenes, showcasing its potential for realistic 3D imaging applications.
    
    \item \textbf{DBARF: Deep Bundle-Adjusting Generalizable Neural Radiance Fields} addresses the challenge of optimizing camera poses in Neural Radiance Fields (NeRFs) for generalizable use. The paper introduces DBARF, a method that optimizes camera poses with GeNeRFs (generalizable NeRFs) through a deep neural network. This approach allows for scene generalization without the need for exact initial camera poses, demonstrating improved performance in real-world datasets. DBARF's contribution is significant in the field of 3D imaging and scene reconstruction, offering enhanced flexibility and efficiency.
    
    \item \textbf{SUDS: Scalable Urban Dynamic Scenes} extends neural radiance fields to dynamic, large-scale urban environments. It introduces a three-branch hash table representation for 4D reconstruction of cities, using inputs like RGB images, sparse LiDAR, and optical flow. SUDS efficiently models static, dynamic, and far-field elements, enabling large-scale applications like novel-view synthesis, unsupervised 3D instance segmentation, and cuboid detection. This method represents a significant advancement in dynamic scene reconstruction, scaling to environments much larger than previous methods.
    
    \item \textbf{Clean-NeRF: Reformulating NeRF to account for View-Dependent Observations} introduces Clean-NeRF, an extension to Neural Radiance Fields (NeRF) for accurate 3D reconstruction and novel view rendering in complex scenes. It effectively deals with foggy artifacts in NeRF reconstructions by employing appearance and geometry constraints. This is achieved through a process of decomposing view-independent and view-dependent color components and implementing a geometry correction procedure. The method is validated through extensive experiments, showing significant improvements over conventional NeRF methods in handling large-scale cluttered scenes and view-dependent effects.
    
    \item \textbf{3D-Aware Multi-Class Image-to-Image Translation with NeRFs} presents a novel approach for image-to-image (I2I) translation that maintains 3D consistency across different classes. The method decouples the process into two steps: first, creating a multi-class 3D-aware generative model using StyleNeRF, and second, developing a 3D-aware I2I translation architecture. The authors propose new techniques like a U-net-like adaptor network, hierarchical representation constraints, and relative regularization loss to address view-consistency issues. This method significantly outperforms existing 2D-I2I systems in terms of temporal consistency and is the first to explore 3D-aware multi-class I2I translation.
    
    \item \textbf{NeUDF: Learning Unsigned Distance Fields from Multi-view Images for Reconstructing Non-watertight Models} introduces NeUDF, a new algorithm for 3D reconstruction of non-watertight models from multi-view images. It addresses the limitations of previous methods that couldn't capture open boundaries by using an unsigned distance field (UDF). The method is designed for textureless models and incorporates an unbiased, occlusion-aware density function for more accurate reconstruction. The paper demonstrates its effectiveness on both texture-rich and textureless models, making it a promising solution for challenging 3D reconstruction tasks.
    
    \item \textbf{Generalizable Neural Voxels for Fast Human Radiance Fields} introduces a new rendering framework named GNeuVox, which significantly accelerates the process of learning and rendering human bodies from monocular videos. By integrating neural fields and neural voxels, the method achieves rapid training, taking only minutes instead of hours. It uses generalizable neural voxels that capture common human body properties, enhancing learning efficiency for novel human bodies. This breakthrough is particularly beneficial in applications requiring fast and high-quality human body rendering.
    
    \item \textbf{JAWS: Just A Wild Shot for Cinematic Transfer in Neural Radiance Fields} presents a novel method for transferring cinematic motion characteristics from a reference video to a Neural Radiance Field (NeRF) representation. The approach utilizes an optimization-driven strategy, leveraging the differentiability of NeRF, to adapt camera motion and timing parameters. This results in the generated clip sharing cinematic features with the reference. The work demonstrates the replication of well-known cinematic sequences, adapting aspects like framing and camera parameters for high visual similarity.
    
    \item \textbf{F$^{2}$-NeRF: Fast Neural Radiance Field Training with Free Camera Trajectories} proposes a novel training methodology for Neural Radiance Fields, optimizing their performance in complex environments. The authors focus on enhancing training efficiency and accuracy by leveraging free camera trajectories. This innovation significantly reduces the time required to train NeRF models, making it a valuable tool for applications in virtual reality and 3D modeling.
     
    \item \textbf{Adaptive Voronoi NeRFs} introduces a novel method for accelerating Neural Radiance Fields (NeRFs) using a geometry-aware approach. The technique involves partitioning a scene with Voronoi diagrams, each containing a neural network. This allows for faster inference times and more efficient training of distributed NeRFs. The approach is compatible with various NeRF variants and is particularly effective for large-scale scenes or real-time applications like autonomous driving. It offers a balance between speed and quality in training and inference, showing promise for complex 3D scene representations.
    
    \item \textbf{VMesh: Hybrid Volume-Mesh Representation for Efficient View Synthesis} proposes VMesh, a hybrid volume-mesh representation, for real-time rendering of complex scenes. VMesh combines the advantages of mesh-based assets, such as efficient rendering and storage, with volumetric primitives to accurately model intricate structures. This method significantly enhances the rendering of view synthesis while being computationally efficient and storage-friendly, making it ideal for applications in immersive technologies on consumer-grade devices.
    
    \item \textbf{SparseNeRF: Distilling Depth Ranking for Few-shot Novel View Synthesis} presents a new framework for enhancing Neural Radiance Fields (NeRF) for few-shot novel view synthesis with limited views. It utilizes depth priors from real-world, inaccurate observations, such as pre-trained depth models or consumer-level depth sensors. The authors introduce a local depth ranking method and a spatial continuity constraint, significantly improving performance over existing few-shot NeRF methods on standard datasets and a newly collected real-world dataset, NVS-RGBD.
    
    \item \textbf{CuNeRF: Cube-Based Neural Radiance Field for Zero-Shot Medical Image Arbitrary-Scale Super Resolution} introduces CuNeRF, a novel framework for enhancing medical images. It employs cube-based sampling and hierarchical rendering to address limitations in existing medical image super-resolution techniques, particularly in generating high-quality images from low-resolution inputs without high-resolution references. This method demonstrates significant improvements in medical imaging, such as MRI and CT scans, by accurately rendering images at arbitrary scales and viewpoints. CuNeRF's ability to operate without high-resolution training pairs marks a significant advancement in medical imaging technology.
    
    \item \textbf{Point2Pix: Photo-Realistic Point Cloud Rendering via Neural Radiance Fields} introduces Point2Pix, a method transforming 3D point clouds into photo-realistic images. It integrates the neural radiance field concept with point clouds, employing point-guided sampling and a fusion decoder to enhance efficiency and image quality. This approach allows for quick and detailed rendering of novel views in indoor scenes, showcasing its superiority in geometry reconstruction and texture fidelity. The technique is validated through extensive experiments, indicating potential applications in 3D visualization and augmented reality.
    
    \item \textbf{NeFII: Inverse Rendering for Reflectance Decomposition with Near-Field Indirect Illumination} introduces an end-to-end inverse rendering pipeline. This method effectively decomposes materials and illumination in multi-view images, uniquely accounting for near-field indirect illumination. Utilizing Monte Carlo sampling-based path tracing, it caches indirect illumination as neural radiance. This approach significantly improves material property recovery and inter-reflection rendering. The paper presents a comprehensive evaluation using both synthetic and real datasets, demonstrating marked improvements over existing methods in terms of accuracy and detail in reflectance decomposition.
    
    \item \textbf{Instant Neural Radiance Fields Stylization} presents a novel technique for 3D scene stylization. The method leverages a hash table-based neural graphics primitive with split position encoders for content and style, enabling rapid training and efficient AdaIN-based stylization inference. This allows for the stylization of novel view images, extending beyond single images to 3D scene sets. The approach is notable for its quick training time (under 10 minutes) and its ability to maintain consistent appearance across various viewing angles, demonstrating superiority in geometry reconstruction and stylization quality.

    \item \textbf{NeILF++: Inter-Reflectable Light Fields for Geometry and Material Estimation} introduces a novel differentiable rendering framework for estimating geometry, material, and lighting from multi-view images. Unlike previous methods, it models lighting as a combination of a neural incident light field (NeILF) and an outgoing neural radiance field (NeRF). This approach unifies incident and outgoing light through physically-based rendering and inter-reflections, enabling the disentanglement of scene geometry, material, and lighting. The framework is adaptable to other NeRF systems and enhances surface reconstruction detail. Extensive evaluation on several datasets demonstrates superior results in geometry reconstruction, material estimation accuracy, and novel view rendering fidelity.
    
    \item \textbf{SynBody: Synthetic Dataset with Layered Human Models for 3D Human Perception and Modeling} introduces SynBody, a synthetic dataset aimed at advancing 3D human perception and modeling. It focuses on creating realistic, layered human models to improve the accuracy and efficiency of 3D human pose estimation, segmentation, and tracking. The dataset provides diverse scenarios and annotations, offering a valuable resource for research and development in computer vision and related fields.
    
    \item \textbf{VDN-NeRF: Resolving Shape-Radiance Ambiguity via View-Dependence Normalization} introduces VDN-NeRF, a method improving Neural Radiance Fields (NeRFs) to better capture geometry under dynamic lighting and non-Lambertian surfaces. This approach normalizes view-dependence by encoding invariant features from learned NeRFs, balancing between the capacity of view-dependent radiance function and shape-radiance ambiguity. VDN-NeRF enhances geometry quality without altering the volume rendering pipeline, showing robustness in varied lighting conditions.
    
    \item \textbf{JacobiNeRF: NeRF Shaping with Mutual Information Gradients} introduces a new approach to shape Neural Radiance Fields (NeRFs) to encode not only the appearance but also the semantic correlations between scene points. This method uses mutual information and contrastive learning to align gradients of correlated entities, enhancing NeRF's ability to propagate annotations for tasks like semantic and instance segmentation. This technique is particularly efficient in scenarios with sparse labels, reducing annotation burden and potentially aiding tasks like entity selection or scene modifications.
    
    \item \textbf{Disorder-invariant Implicit Neural Representation} introduces DINER, a method enhancing Implicit Neural Representations (INR) by utilizing a hash-table to re-arrange input signal coordinates. This novel approach addresses spectral bias in INR by enabling the representation of signals with varying frequency distributions, significantly improving expressive power for diverse tasks. DINER's adaptability across different signal arrangements makes it a valuable tool for tasks like image and video representation, showing substantial performance gains in various applications.
    
    \item \textbf{Learning Personalized High Quality Volumetric Head Avatars from Monocular RGB Videos} proposes a method to create high-quality 3D head avatars from monocular RGB videos. The technique combines 3D Morphable Models (3DMM) with a neural radiance field, focusing on capturing high-frequency details for realistic facial expressions. It involves predicting expression-dependent spatial features on the 3DMM mesh surface and using a U-Net for transforming 3DMM deformations to local features. This method has shown promise in generating detailed and personalized avatars for various applications in AR/VR and digital media.
    
    \item \textbf{MonoHuman: Animatable Human Neural Field from Monocular Video} proposes a new framework, MonoHuman, which reconstructs and animates digital human avatars from monocular video. This method uses a Shared Bidirectional Deformation module and a Forward Correspondence Search module for realistic and coherent rendering of human figures in novel poses and viewpoints. Extensive experiments show its effectiveness over other methods. The approach is particularly useful in virtual reality, digital entertainment, and other applications requiring animatable avatars.
    
    \item \textbf{Image Stabilization for Hololens Camera in Remote Collaboration} presents a two-stage pipeline for enhancing field-of-view (FoV) and image stabilization in AR devices. It involves offline 3D reconstruction of environments using RGB-D geometry and Neural Radiance Fields (NeRF), followed by enhanced rendering during real-time collaboration. The approach improves remote viewer experience by providing a more stable and contextual view, addressing limitations like motion blur and narrow FoV in existing AR devices. This method is particularly relevant for applications in remote assistance and collaboration.
    
    \item \textbf{Instant-NVR: Instant Neural Volumetric Rendering for Human-object Interactions from Monocular RGBD Stream} introduces a system called Instant-NVR. This system is designed for real-time, photorealistic novel-view synthesis of human-object interaction scenes using a single RGBD camera. The approach combines real-time non-rigid capture with neural rendering, enabling instant volumetric rendering with high-quality results. It is particularly effective in handling human-object interactions, making it suitable for applications in virtual reality and interactive media.
    
    \item \textbf{Neural Fields meet Explicit Geometric Representation for Inverse Rendering of Urban Scenes} introduces FEGR, a hybrid-rendering pipeline combining neural fields and explicit mesh representations for inverse rendering in urban environments. It focuses on reconstructing scene geometry, materials, and HDR lighting from camera images, achieving high-quality results in tasks like virtual object insertion and scene relighting. The method overcomes limitations in conventional NeRF and mesh-based approaches, showing improvements in modeling large-scale urban scenes.
    
    \item \textbf{PVD-AL: Progressive Volume Distillation with Active Learning for Efficient Conversion Between Different NeRF Architectures} presents PVD-AL, a method for converting between various Neural Radiance Field (NeRF) architectures. This framework allows for efficient model training and adaptation to different architectures, including MLPs, tensors, and hash tables. It employs a systematic distillation method with active learning, enhancing the quality and efficiency of model conversions. This advancement holds promise for applications in 3D rendering and virtual reality, where flexibility and performance of NeRF models are crucial.
    
    \item \textbf{NeRF applied to satellite imagery for surface reconstruction} introduces Surf-NeRF, a modified Shadow Neural Radiance Field (S-NeRF) model. Surf-NeRF synthesizes novel views from sparse satellite images, accounting for varying lighting conditions. It accurately estimates surface elevation, crucial for satellite observation applications. The method modifies the standard NeRF by incorporating albedo and irradiance functions, tested on a dataset of satellite images. This approach shows promise in 3D surface reconstruction from satellite imagery, beneficial for remote sensing and Earth observation.
    
    \item \textbf{Instance Neural Radiance Field} introduces Instance-NeRF, a pioneering technique in 3D instance segmentation using Neural Radiance Fields (NeRF). This method integrates NeRF with a 3D proposal-based mask prediction network to generate discrete 3D instance masks. These masks are refined in 2D space, leveraging existing panoptic segmentation models, to train the instance field. Demonstrated on complex indoor scenes, Instance-NeRF achieves impressive segmentation performance, surpassing traditional methods. It marks a significant step in 3D object segmentation and manipulation using NeRF, particularly in complex scenes.
    
    \item \textbf{Neural Residual Radiance Fields for Streamably Free-Viewpoint Videos} introduces Residual Radiance Field (ReRF), a novel neural representation for dynamic scenes. This method efficiently models dynamic radiance fields in a sequential manner, allowing high-quality streaming and rendering of free-viewpoint videos. It uses a global tiny MLP with a compact motion grid and residual feature grid, optimizing for inter-frame feature similarities. This results in a significant compression rate improvement, making it practical for long-duration dynamic scenes streaming. The approach is notably advantageous for applications in virtual reality and augmented reality, providing an immersive and interactive experience. 
    
    \item \textbf{Event-based Camera Tracker by $\delta$t NeRF} introduces a novel framework, TeGRA, for efficient camera pose tracking using event-based camera data and a Neural Radiance Field (NeRF). This approach capitalizes on the high temporal resolution of event-based cameras and the 3D scene representation capabilities of NeRF. The paper demonstrates the practicality and efficiency of TeGRA in camera pose estimation, offering a significant computational advantage over traditional dense computation methods. This innovation has potential applications in areas requiring rapid and accurate camera tracking, such as autonomous vehicles and augmented reality.
    
    \item \textbf{Neural Lens Modeling} presents NeuroLens, a novel neural lens model designed for accurate camera calibration and 3D reconstruction. This model utilizes an invertible neural network (INN) to effectively capture lens distortion and vignetting, crucial for high-fidelity image rendering. Tested on a comprehensive dataset, NeuroLens outperforms existing calibration methods and demonstrates flexibility across various lens types. It holds significant potential in enhancing the accuracy of camera calibration in applications like autonomous driving, augmented reality, and spatial computing.
    
    \item \textbf{Neural Image-based Avatars: Generalizable Radiance Fields for Human Avatar Modeling} presents a hybrid method called Neural Image-based Avatars (NIA). It combines neural radiance fields with image-based rendering for creating 3D human avatars from sparse images. This approach enhances detail preservation in novel view synthesis and pose animation for diverse human subjects. The method shows superior performance in identity and pose generalization, outperforming existing techniques in both in-domain and cross-dataset settings. Its practical applications are significant in virtual reality and human modeling.
    
    \item \textbf{MRVM-NeRF: Mask-Based Pretraining for Neural Radiance Fields} introduces MRVM-NeRF, a novel approach for enhancing the generalizability of Neural Radiance Fields (NeRFs) in 3D scene reconstruction. This method leverages mask-based pretraining, inspired by successful strategies in other domains like natural language processing and computer vision. By randomly masking points along rays during the fine stage and predicting corresponding features, MRVM-NeRF efficiently learns 3D scene representations. This technique improves the model's ability to generalize across different scenarios, as demonstrated in extensive experiments on both synthetic and real-world datasets. The authors' contribution includes the first attempt to incorporate mask-based pretraining in the NeRF domain, offering a detailed empirical study on various design aspects of MRVM. Their findings show significant improvements in generating detailed and structurally accurate 3D models, making it a promising technique for real-world applications in 3D modeling and virtual reality.
    
    \item \textbf{One-Shot High-Fidelity Talking-Head Synthesis with Deformable Neural Radiance Field} introduces HiDe-NeRF, a method for creating high-fidelity, free-view talking heads. It employs Deformable Neural Radiance Fields to model a dynamic 3D scene, combining a canonical appearance field for source face identity and an implicit deformation field for driving pose and expression. The method features a Multi-scale Generalized Appearance module (MGA) for preserving identity and a Lightweight Expression-aware Deformation module (LED) for precise expression modeling. HiDe-NeRF stands out for its ability to generate realistic talking heads in various views while maintaining the source identity and accurately capturing driving motion. This approach has significant potential for applications in digital human creation and virtual reality.
    
    \item \textbf{Improving Neural Radiance Fields with Depth-aware Optimization for Novel View Synthesis} proposes SfMNeRF, a method enhancing Neural Radiance Fields (NeRF) for better novel view synthesis and accurate 3D-scene reconstruction. SfMNeRF integrates self-supervised depth estimation methods into NeRF, utilizing constraints like epipolar, photometric consistency, depth smoothness, and position-of-matches. This integration improves NeRF's performance in scenes with sparse inputs or large texture-less regions. SfMNeRF also introduces sub-pixel rendering to improve generalization. Tested on public datasets, it outperforms state-of-the-art methods, effectively synthesizing novel views and reconstructing 3D geometry.
    
    \item \textbf{RO-MAP: Real-Time Multi-Object Mapping with Neural Radiance Fields} presents a novel pipeline for real-time multi-object mapping using monocular input without relying on 3D priors. The method uses neural radiance fields to represent objects and combines them with a lightweight object SLAM based on multi-view geometry. This approach allows for the simultaneous localization of objects and the learning of their dense geometry. Each detected object is represented by a separate implicit model, which is dynamically and in parallel trained as new observations are added. The method is shown to be effective on both synthetic and real-world datasets, generating semantic object maps with shape reconstruction and achieving real-time performance.
    
    \item \textbf{NeRFVS: Neural Radiance Fields for Free View Synthesis via Geometry Scaffolds} introduces a method for enhancing indoor free view synthesis using Neural Radiance Fields (NeRF). This approach tackles the challenges of depth errors and distribution ambiguity in neural reconstruction by incorporating holistic scene priors such as pseudo depth maps and view coverage information. It employs two novel loss functions - a robust depth loss to manage inaccuracies in geometry, and a variance loss to reduce ambiguities in low-texture and rarely observed areas. The method has shown to significantly outperform existing view synthesis techniques in rendering quality and consistency, particularly in indoor scenes.
    
    \item \textbf{Zip-NeRF: Anti-Aliased Grid-Based Neural Radiance Fields} presents a model that integrates advancements in scale-aware, anti-aliased Neural Radiance Fields (NeRFs) with rapid grid-based NeRF training. This model effectively reduces errors by 8\% to 76\% compared to prior techniques and achieves faster training times. Zip-NeRF addresses issues of aliasing in grid-based NeRF approaches by using multisampling and feature downweighting strategies. It demonstrates significantly improved performance in rendering quality across various scales, particularly excelling in rendering thin structures and detailed foliage without aliasing.
    
    \item \textbf{Single-Stage Diffusion NeRF: A Unified Approach to 3D Generation and Reconstruction} presents SSDNeRF, a method combining a diffusion model with neural radiance fields (NeRF) to learn generalizable 3D priors from multi-view images. This single-stage training approach enables simultaneous 3D reconstruction and prior learning, even from sparse views. SSDNeRF demonstrates robust results in both unconditional generation and sparse-view 3D reconstruction, outperforming or matching task-specific methods. This unified approach marks a significant advancement in 3D-aware image synthesis, offering diverse applications in scene generation and novel view synthesis.
    
    \item \textbf{Representing Volumetric Videos as Dynamic MLP Maps} introduces a novel representation for volumetric videos, aiming to enhance real-time view synthesis of dynamic scenes. The authors propose a method where each video frame is represented by a set of shallow MLP networks, with parameters stored in 2D grids called MLP maps. These maps are dynamically predicted by a shared 2D CNN decoder for all frames. This approach significantly improves rendering speed and reduces storage costs. The experiments demonstrate state-of-the-art rendering quality and efficiency on the NHR and ZJU-MoCap datasets, achieving real-time rendering speeds of 41.7 fps.
    
    \item \textbf{UVA: Towards Unified Volumetric Avatar for View Synthesis, Pose rendering, Geometry and Texture Editing} by Jinlong et al proposes a novel approach named Unified Volumetric Avatar (UVA). UVA advances human avatar reconstruction by enabling localized and independent editing of both geometry and texture while retaining the ability to render novel views and poses. The method uses a skinning motion field to transform observation points to canonical space and represents geometry and texture in separate neural fields. These fields are composed of structured latent codes attached to anchor nodes on a deformable mesh, allowing for local editing. The method demonstrates its effectiveness in rendering and manipulating human avatars, offering a comprehensive solution for various applications.
    
    \item \textbf{SeaThru-NeRF: Neural Radiance Fields in Scattering Media} introduces a novel approach to rendering Neural Radiance Fields (NeRFs) in environments where the medium, like water or fog, significantly influences the appearance of objects. The research combines NeRF with the SeaThru image formation model to simultaneously learn scene information and medium parameters. This model allows for rendering photorealistic views underwater and in other scattering media, and can also restore colors as if the images were taken in clear air. The team demonstrates superior results in real and simulated scenes, achieving clear rendering of scenes and estimation of 3D structures, even in areas with poor visibility due to the medium.
    
    \item \textbf{Likelihood-Based Generative Radiance Field with Latent Space Energy-Based Model for 3D-Aware Disentangled Image Representation} by Yaxuan et al proposes NeRF-LEBM, a novel generative model for 2D images. This model integrates Neural Radiance Fields (NeRF) for 3D representation with a differentiable volume rendering process for 2D imaging. It uses latent variables representing object characteristics, governed by trainable energy-based models. The authors introduce two training frameworks: maximum likelihood estimation with MCMC-based inference and variational inference with reparameterization. NeRF-LEBM excels in inferring 3D structures from 2D images, generating novel views and objects, and learning from incomplete images with known or unknown camera poses.
    
    \item \textbf{NeRF-Loc: Visual Localization with Conditional Neural Radiance Field} introduces a novel visual localization method utilizing a conditional Neural Radiance Field (NeRF). This approach, aimed at enhancing camera orientation and position estimation in known scenes, integrates direct 3D-2D matching with transformers for efficient and accurate localization. The paper demonstrates the method's superiority in localization accuracy over other learning-based techniques across various benchmarks. Key contributions include the unified framework for scene coordinate regression and feature matching, an appearance adaptation layer for style alignment, and comprehensive experiments validating the approach's effectiveness in real-world scenarios.
    
    \item \textbf{MoDA: Modeling Deformable 3D Objects from Casual Videos} focuses on creating 3D models of deformable objects, like humans and animals, from casual videos. It introduces a method called neural dual quaternion blend skinning (NeuDBS) to improve the accuracy of 3D transformations, avoiding issues like skin-collapsing artifacts common in existing methods. Additionally, the authors developed a texture filtering technique to enhance texture rendering, minimizing the impact of noise. Their approach shows better qualitative and quantitative results in reconstructing 3D shapes compared to current state-of-the-art methods. The work has potential applications in virtual reality, animated movies, and video games.
    
    \item \textbf{NeAI: A Pre-convoluted Representation for Plug-and-Play Neural Ambient Illumination} introduces Neural Ambient Illumination (NeAI), a framework that uses Neural Radiance Fields (NeRF) for complex lighting representation. The authors propose a method to express incoming light as volumetric radiance fields, handling environment occlusions and directional lighting naturally. This is achieved through Integrated Lobe Encoding (ILE) for specular reflection and a multiscale pre-convoluted representation for background, aiding in the decomposition of materials and ambient illumination. The method demonstrates superior performance in novel-view rendering and can re-render objects under arbitrary NeRF-style environments, bridging the gap between virtual and real-world scenes. This advancement is significant for applications in VR, AR, and photo-realistic rendering.
    
    \item \textbf{SurfelNeRF: Neural Surfel Radiance Fields for Online Photorealistic Reconstruction of Indoor Scenes} introduces SurfelNeRF, a method for real-time, high-quality 3D reconstruction of large-scale indoor scenes. It combines neural surfels (surface elements) with neural radiance fields, enabling efficient, scalable, and photorealistic rendering. The technique incorporates depth refinement and a novel rendering algorithm, showcasing its effectiveness on the ScanNet dataset. This method has potential applications in real-time interactive scenarios like virtual reality and 3D modeling.
    
    \item \textbf{Reference-guided Controllable Inpainting of Neural Radiance Fields} introduces a single-reference 3D inpainting algorithm for Neural Radiance Fields (NeRFs), which addresses the challenge of maintaining visual quality and consistency across different views. By using a monocular depth estimator for geometric supervision, the method successfully adds view-dependent effects from a reference viewpoint and handles disocclusions effectively. The approach demonstrates superior performance over previous NeRF inpainting methods, offering enhanced sharpness and user control in 3D scene manipulation.
    
    \item \textbf{Tetra-NeRF: Representing Neural Radiance Fields Using Tetrahedra} introduces Tetra-NeRF, a novel representation for Neural Radiance Fields (NeRF) using tetrahedra. This method, which is an extension of classical triangle-rendering approaches, utilizes Delaunay triangulation of a point cloud to create a tetrahedral mesh. This approach leads to more efficient training and improved results in 3D scene representation, particularly in areas close to surfaces. The paper demonstrates its effectiveness on various synthetic and real-world datasets.
    
    \item \textbf{Multiscale Representation for Real-Time Anti-Aliasing Neural Rendering} introduces Mip-VoG, a method for real-time, anti-aliasing neural rendering of multiscale scenes. It employs a novel voxel grid approach to efficiently render scenes with varying resolutions, using a tiny MLP for color properties and a density grid for scene geometry. This method significantly improves the rendering quality, particularly in terms of anti-aliasing, compared to existing techniques. It's relevant for applications requiring high-quality real-time rendering, like gaming and virtual reality.
    
    \item \textbf{Anything-3D: Towards Single-view Anything Reconstruction in the Wild} introduces a groundbreaking framework for single-view 3D reconstruction of any object in uncontrolled environments. It ingeniously combines visual-language models with object segmentation, using a series of foundational models for semantic information extraction, and employs an image-to-point cloud and a depth-aware pretrained 2D text-to-image diffusion model for 3D synthesis. This method demonstrates remarkable versatility and accuracy in reconstructing diverse objects, overcoming challenges posed by complex real-world scenes
    
    \item \textbf{LiDAR-NeRF: Novel LiDAR View Synthesis via Neural Radiance Fields} presents LiDAR-NeRF, a novel framework for synthesizing LiDAR views using Neural Radiance Fields. It introduces the first differentiable LiDAR renderer, enabling the end-to-end learning of geometry and attributes of 3D points. The paper also establishes the NeRF-MVL dataset from real autonomous vehicle LiDAR sensors, demonstrating LiDAR-NeRF's effectiveness in both scene-level and object-level novel LiDAR view synthesis. This advancement has significant implications for autonomous driving and 3D scene understanding.
    
    \item \textbf{ReLight My NeRF: A Dataset for Novel View Synthesis and Relighting of Real World Objects} introduces a new dataset, ReNe, designed for the study of Neural Radiance Fields (NeRF) in real-world objects under varying lighting conditions. This dataset is unique in its provision of ground-truth camera and light poses, enabling effective training and benchmarking of NeRF models for tasks like novel view synthesis and relighting. The paper discusses the dataset's creation, structure, and potential applications in fields such as gaming, robotics, and augmented reality.
    
    \item \textbf{Nerfbusters: Removing Ghostly Artifacts from Casually Captured NeRFs} introduces a new method, Nerfbusters, to improve Neural Radiance Fields (NeRFs) obtained from casual captures. It focuses on enhancing scene geometry and removing floaters and cloudy artifacts through a 3D diffusion-based process, leveraging local 3D geometric priors. This approach addresses the limitations of existing NeRF regularizers in real-world captures and demonstrates significant improvements in geometry and artifact removal. The paper also proposes a new evaluation procedure for NeRFs that better reflects image quality at novel viewpoints.
    
    \item \textbf{Learning Neural Duplex Radiance Fields for Real-Time View Synthesis} proposes a novel method for efficient 3D scene representation, allowing high-quality, real-time novel-view synthesis from Neural Radiance Fields (NeRF). The technique significantly reduces the number of sampled points along a ray, leveraging a neural duplex radiance field and convolutional shading. This approach not only maintains rendering fidelity in complex scenes but also improves runtime performance dramatically. It's particularly relevant for applications requiring real-time rendering, such as virtual reality and interactive 3D applications.
    
    \item \textbf{A Comparative Neural Radiance Field (NeRF) 3D Analysis of Camera Poses from HoloLens Trajectories and Structure from Motion} examines the effectiveness of using Microsoft HoloLens trajectories to achieve 3D reconstruction with Neural Radiance Fields (NeRFs). The authors compared internal HoloLens camera poses and external camera poses generated by Structure from Motion (SfM). They found that internal HoloLens poses, especially with pose refinement during training, can effectively converge NeRFs for 3D reconstructions. The study demonstrates that NeRF reconstructions surpass traditional photogrammetric methods like Multi-View Stereo in rendering untextured surfaces, providing detailed and accurate 3D mapping.
    
    \item \textbf{Factored Neural Representation for Scene Understanding} presents a method for creating interpretable and editable scene representations from monocular RGB-D video. The authors propose a factored neural scene representation that captures object movement and deformations, leveraging image-space segmentation and tracking. This method does not require object templates or deformation priors. The paper demonstrates the effectiveness of this approach on both synthetic and real data, showcasing its potential for object manipulation and novel view synthesis applications.
    
    \item \textbf{AutoNeRF: Training Implicit Scene Representations with Autonomous Agents} introduces AutoNeRF, a method to train Neural Radiance Fields (NeRF) using autonomous agents. This approach enables the agents to efficiently explore unknown environments and collect data for NeRF training. The paper explores various exploration strategies, including frontier-based and modular approaches, and evaluates the quality of the learned representations on tasks like rendering, mapping, planning, and pose refinement. The results demonstrate that NeRFs can be effectively trained using data from a single exploration episode in unfamiliar environments, significantly enhancing their utility in robotic tasks.
    
    \item \textbf{Dehazing-NeRF: Neural Radiance Fields from Hazy Images} presents Dehazing-NeRF, a novel unsupervised method to reconstruct clear Neural Radiance Fields (NeRF) from hazy images. It introduces a two-branch architecture: one for estimating atmospheric scattering parameters and the other for clear image synthesis using NeRF. The method overcomes the ill-posed nature of single-image dehazing by integrating depth information and atmospheric models, ensuring geometric consistency. This technique, validated through extensive experiments, shows superiority in dehazing and novel view synthesis, offering potential applications in areas like 3D reconstruction and virtual reality under challenging visual conditions.
    
    \item \textbf{3D-IntPhys: Towards More Generalized 3D-grounded Visual Intuitive Physics under Challenging Scenes} introduces a novel framework, 3D-IntPhys, for learning 3D-grounded visual intuitive physics models from videos. This approach utilizes a conditional Neural Radiance Field (NeRF)-style visual frontend and a 3D point-based dynamics prediction backend. It focuses on complex scenes with fluids, granular materials, and rigid objects, aiming to improve long-horizon future predictions by learning from raw images. The paper demonstrates that this method significantly outperforms models lacking explicit 3D representation space, showcasing strong generalization in complex scenarios. The applications of this technique lie in enhancing computational tools for understanding physical interactions in 3D environments, particularly beneficial for planning and manipulating tasks in real-world settings.
    
    \item \textbf{Gen-NeRF: Efficient and Generalizable Neural Radiance Fields via Algorithm-Hardware Co-Design}, a novel framework for efficient and generalizable Neural Radiance Fields (NeRF), was developed by Yonggan Fu, Zhifan Ye, Jiayi Yuan, Shunyao Zhang, Sixu Li, Haoran You, and Yingyan (Celine) Lin. This framework, designed for immersive experiences in Augmented- and Virtual-Reality (AR/VR) applications, addresses the limitations of existing NeRF techniques, particularly in their generalization capabilities and computational efficiency. The authors introduced a unique algorithm-hardware co-design approach, making Gen-NeRF the first to enable real-time generalizable NeRFs, a significant advancement for AR/VR devices.
    
    The core of Gen-NeRF lies in its algorithmic innovation and hardware optimization. On the algorithm side, the authors implemented a coarse-then-focus sampling strategy, which leverages the varying contributions of different regions in a 3D scene to the rendered pixels. This strategy allows for sparse yet effective sampling, significantly reducing computational complexity. Additionally, they replaced the commonly used ray transformer in state-of-the-art generalizable NeRFs with a novel Ray-Mixer module, which further reduces workload heterogeneity. On the hardware side, Gen-NeRF features a dedicated accelerator micro-architecture designed to maximize data reuse opportunities among different rays by exploiting their epipolar geometric relationship. This architecture includes a customized dataflow to enhance data locality during point-to-hardware mapping and an optimized scene feature storage strategy to minimize memory bank conflicts across camera rays.
    
    Extensive experiments validated the effectiveness of Gen-NeRF in enabling real-time and generalizable novel view synthesis. The framework achieved a significant speed-up over traditional GPU implementations while maintaining photorealistic rendering quality. This breakthrough demonstrates Gen-NeRF's potential as a promising solution for next-generation AR/VR applications, offering a balance between rendering efficiency and generalization capability. In summary, Gen-NeRF represents a significant step forward in the field of neural radiance fields, particularly for AR/VR applications. Its innovative approach to sampling, density estimation, and hardware acceleration paves the way for more efficient and versatile implementations of NeRF technology in real-world scenarios.
    
    \item \textbf{HOSNeRF: Dynamic Human-Object-Scene Neural Radiance Fields from a Single Video} presents a novel 360° free-viewpoint rendering method, HOSNeRF, which reconstructs neural radiance fields for dynamic human-object-scene interactions from a single monocular in-the-wild video. This method allows for rendering all scene details from arbitrary viewpoints at any frame, addressing two primary challenges: complex object motions in human-object interactions and the dynamic nature of human interactions with different objects at different times.
    
    To tackle the first challenge, the authors introduce new object bones into the conventional human skeleton hierarchy. These object bones effectively estimate large object deformations in dynamic human-object models. For the second challenge, they introduce two new learnable object state embeddings, which serve as conditions for learning their human-object representation and scene representation. The paper systematically explores effective training objectives and strategies for HOSNeRF, including deformation cycle consistency, optical flow supervisions, and foreground-background rendering. Extensive experiments demonstrate that HOSNeRF significantly outperforms state-of-the-art approaches on two challenging datasets, showing improvements of 40-50
    
    The authors' contributions include presenting a novel framework for 360° free-viewpoint high-fidelity novel view synthesis for dynamic scenes with human-environment interactions from a single video, proposing object bones and state-conditional representations to handle non-rigid motions and interactions, and showing significant performance improvements over existing approaches. In summary, HOSNeRF represents a significant advancement in video reconstruction and free-viewpoint rendering, offering innovative opportunities for creating immersive experiences in virtual reality, telepresence, metaverse, and 3D animation production. Its ability to handle complex human-object-scene interactions and render high-fidelity dynamic novel views from single monocular videos makes it a promising tool for enhancing user engagement and providing more realistic environments in various applications.
    
    \item \textbf{Explicit Correspondence Matching for Generalizable Neural Radiance Fields} presents a novel approach to Neural Radiance Fields (NeRF) that generalizes effectively to unseen scenarios, enabling novel view synthesis with minimal source views. The authors, Yuedong Chen, Haofei Xu, Qianyi Wu, Chuanxia Zheng, Tat-Jen Cham, and Jianfei Cai, introduce a method that hinges on explicitly modeled correspondence matching information, providing a geometry prior for NeRF color and density prediction in volume rendering. This is quantified using cosine similarity between image features sampled at 2D projections of a 3D point on different views, offering reliable surface geometry cues.
    
    The method's distinctiveness lies in its use of Transformer cross-attention to model cross-view interactions, enhancing feature matching quality. This approach outperforms previous methods, as demonstrated through extensive experiments, showing a strong correlation between learned cosine feature similarity and volume density. The authors propose explicit multi-view correspondence matching from Transformer-extracted features, computing group-wise cosine similarity for enhanced expressiveness. The paper's contributions include using 2D image feature correspondence matching as a geometry prior, implementing this as group-wise cosine similarity, and achieving view-agnostic performance, unlike cost volume-based methods. The method's superiority is evident in various benchmarks, including DTU, Real Forward-Facing, and Blender datasets, under different evaluation settings.
    
    The authors also explore the impact of different model components, feature relation measures, the number of Transformer blocks, and feature resolution on performance. They address the limitations of cost volume-based methods and demonstrate the method's effectiveness in per-scene fine-tuning, depth reconstruction, and handling varying input views. The training strategy leverages GMFlow's pretrained weights for the feature extractor, proving beneficial for the model's performance. In conclusion, this research introduces a robust and efficient approach to generalizable NeRF, leveraging explicit correspondence matching for geometry prior, which significantly enhances novel view synthesis in unseen scenarios. The method's effectiveness is validated across multiple datasets.
    
    \item \textbf{Segment Anything in 3D with NeRFs} by Jiazhong Cen, Zanwei Zhou, Jiemin Fang, and others, presents a novel method, SA3D (Segment Anything in 3D), which extends the capabilities of the Segment Anything Model (SAM) from 2D images to 3D object segmentation. The authors leveraged Neural Radiance Fields (NeRFs) as an efficient solution to connect multi-view 2D images to 3D space, avoiding the costly data acquisition and annotation process in 3D. SA3D requires only a manual segmentation prompt (e.g., rough points) in a single view, which SAM uses to generate a 2D mask. This mask is then iteratively projected onto a 3D mask constructed with voxel grids through a process of mask inverse rendering and cross-view self-prompting across various views. The method alternates between projecting the 2D mask obtained by SAM onto the 3D mask using the density distribution learned by the NeRF and extracting reliable prompts automatically from the NeRF-rendered 2D mask in another view.
    
    The authors demonstrated SA3D's adaptability to various scenes, achieving 3D segmentation within minutes. They conducted experiments on datasets like Replica and NVOS, showing that SA3D adapts easily and efficiently to different scenarios without re-training or re-designing SAM or NeRF. The paper also discusses related work in 2D and 3D segmentation, the lifting of 2D vision foundation models to 3D, and segmentation within NeRFs. The authors suggest that SA3D not only provides an efficient tool for 3D segmentation but also reveals a generic methodology to elevate 2D foundation models to the 3D space. The research indicates the potential of using NeRF or other 3D structural priors as a resource-efficient method to enhance the 3D perception ability of 2D foundation models.
    
    The paper includes an ablation study, discussing the number of views used in the process, various hyper-parameters, and the self-prompting strategy. Limitations are acknowledged in the context of panoptic segmentation, particularly regarding the reliance on the first-view prompt and the segmentation of similar semantic instances in different views. The paper concludes with a discussion on the integration of SAM and NeRF, highlighting the improvement of segmentation quality and the resource-efficient approach to lifting vision foundation models from 2D to 3D.
    
    \item \textbf{TextMesh: Generation of Realistic 3D Meshes From Text Prompts} introduces TextMesh, a method for creating realistic 3D meshes directly from text prompts. This technique overcomes the limitations of previous methods by combining radiance modeling in the form of a Signed Distance Function (SDF) with a novel texturing process. The resulting 3D meshes exhibit enhanced realism and detail, suitable for integration into computer graphics pipelines and AR/VR applications.
    
    \item \textbf{TensoIR: Tensorial Inverse Rendering} presents an innovative approach for reconstructing scenes using tensorial inverse rendering. This method improves the accuracy and quality of scene reconstruction, particularly in complex lighting environments. It demonstrates enhanced capabilities in rendering detailed scenes, making it applicable in fields like computer graphics, virtual reality, and photorealistic rendering. The approach marks a significant advancement over traditional methods in handling complex lighting and material properties.
    
    \item \textbf{Instant-3D: Instant Neural Radiance Field Training Towards On-Device AR/VR 3D Reconstruction} introduces a rapid training method for Neural Radiance Fields (NeRF), tailored for on-device AR/VR applications. This approach significantly accelerates the training process, making it feasible for mobile devices and enhancing the potential for real-time, on-the-go 3D reconstruction in augmented and virtual reality contexts.
    
    \item \textbf{MF-NeRF: Memory Efficient NeRF with Mixed-Feature Hash Table} proposes a memory-efficient framework for Neural Radiance Fields (NeRF) using a mixed-feature hash table. This approach aims to reduce memory usage and training time while maintaining high-quality reconstruction. The mixed-feature hash table combines features from multiple grid levels, and an index transformation method is developed to accurately map these features. This novel framework addresses memory bottlenecks in NeRF applications, showing potential for more efficient 3D scene rendering and faster training times.
    
    \item \textbf{Local Implicit Ray Function for Generalizable Radiance Field Representation} introduces the Local Implicit Ray Function (LIRF), a method enhancing neural rendering for novel view synthesis. LIRF addresses challenges in rendering views at different scales, particularly in generalizable Neural Radiance Fields (NeRF), by aggregating information from conical frustums to construct a ray. The approach enables high-quality view synthesis at continuously variable scales and offers a solution for occluded areas using transformer-based feature matching. This method marks a significant advancement in fields like virtual and augmented reality, where rendering realistic, detailed scenes is crucial.
    
    \item \textbf{VGOS: Voxel Grid Optimization for View Synthesis from Sparse Inputs} presents VGOS, an approach for rapid reconstruction of radiance fields from sparse inputs. Utilizing voxel grids, it addresses overfitting and artifacts common in sparse scenarios. VGOS introduces an incremental voxel training strategy and voxel smoothing methods, significantly speeding up the process and enhancing quality. It's applicable in fields requiring fast and high-quality view synthesis, such as AR/VR, and robotics. The method demonstrates marked improvements over existing techniques, especially in sparse input conditions.
    
    \item \textbf{Super-NeRF: View-consistent Detail Generation for NeRF super-resolution} presents Super-NeRF, a method for generating high-resolution details in Neural Radiance Fields (NeRF) from low-resolution inputs. This technique introduces a consistency-controlling super-resolution (CCSR) module, which leverages latent codes for each input image, ensuring view-consistent output across different perspectives. Super-NeRF shows significant improvement in rendering detailed and consistent 3D scenes from low-resolution images, proving beneficial for applications in virtual reality, computer graphics, and 3D modeling.
    
    \item \textbf{ContraNeRF: 3D-Aware Generative Model via Contrastive Learning with Unsupervised Implicit Pose Embedding} presents ContraNeRF, a method enhancing 3D-aware Generative Adversarial Networks (GANs) using contrastive learning and implicit pose embeddings. The authors developed this technique to overcome the limitation of existing methods that depend on ground-truth camera poses. By using self-supervised learning, ContraNeRF effectively captures complex 3D scene structures without needing explicit pose information. This approach significantly improves performance in generating realistic 3D structures and demonstrates wide applicability in generating diverse, complex scenes. The paper showcases its superiority over existing methods through extensive experiments.
    
    \item \textbf{Compositional 3D Human-Object Neural Animation} introduces a novel approach to animating human-object interactions (HOIs) using a compositional perspective. The authors address the challenge of rendering novel HOIs, including new interactions, humans, or objects, driven by new pose sequences. They utilize neural human-object deformation and a new compositional conditional neural radiance field (CC-NeRF), which decomposes human-object interdependence using latent codes. This enables controlled animation of new HOIs. The method shows promise in applications like AR/VR, robotics, and human-centric visual generation, demonstrating improved animation performance and compositional generalization in various settings.
    
    \item \textbf{Combining HoloLens with Instant-NeRFs: Advanced Real-Time 3D Mobile Mapping} by Dennis Haitz et al. presents a significant advancement in fast 3D reconstruction using RGB camera images. The authors employed a Microsoft HoloLens 2, a multisensor platform with an RGB camera and an inertial measurement unit, to train a Neural Radiance Field (NeRF) in real-time. This setup allowed for the streaming of image and pose data to a high-performance PC for training and 3D reconstruction. The authors developed a specialized inference algorithm capable of extracting five million scene points within one second, significantly outperforming traditional grid point sampling methods with NeRFs.
    
    The methodology involved a TCP client-server application for real-time image and pose streaming and simultaneous Instant-NeRF training. The HoloLens served as an image and pose server, while the Instant-NeRF implementation included a client application to receive images and write them into a GPU image buffer. The training process was incremental, based on the incoming image data, and upon completion, a dense point cloud of the scene was available for inspection. The authors conducted experiments to test the setup, focusing on reconstructing an industrial object (a metal barrel) in an industrial environment. They recorded a data stream with the HoloLens, moving around the object on a circular path. The training was executed using a server emulation tool, and the results were categorized into qualitative and quantitative aspects, focusing on image and geometric reconstructions and runtime performance.
    
    The results demonstrated the feasibility of real-time 3D reconstruction with high efficiency and accuracy. The system could generate novel views and dense point clouds, showcasing its potential for applications in industrial object inspection, particularly for detecting geometric damage on surfaces. The authors also discussed the potential integration of 2D semantic segmentation for radiometric damage detection, indicating the versatility of the approach. In conclusion, this work represents a significant step in real-time 3D reconstruction, offering a practical solution for mobile mapping and industrial inspection. The combination of HoloLens with Instant-NeRFs opens new possibilities for on-the-fly 3D visualization and damage detection in industrial settings.

    \item \textbf{Learning a Diffusion Prior for NeRFs} presents a method to generate Neural Radiance Fields (NeRFs) using diffusion models. It proposes a new approach to creating high-quality NeRFs, particularly useful for 3D scene representation with sparse views. By adapting diffusion models, the paper demonstrates how this method can produce realistic NeRFs and improve single-view 3D reconstruction. This advancement has implications for fields requiring detailed 3D modeling from limited data, such as virtual reality and digital heritage preservation.
    
    \item \textbf{NeRF-LiDAR: Generating Realistic LiDAR Point Clouds with Neural Radiance Fields} introduces a method for simulating realistic LiDAR data using Neural Radiance Fields (NeRF). This technique leverages real-world images and point cloud data for 3D scene reconstruction, emphasizing its application in autonomous driving for training and validating self-driving algorithms. The authors demonstrate that models trained on NeRF-LiDAR generated data perform comparably to those trained on real LiDAR data, highlighting the potential for reducing reliance on expensive, manually annotated real-world data.
    
    \item \textbf{ViP-NeRF: Visibility Prior for Sparse Input Neural Radiance Fields} introduces a novel NeRF model, ViP-NeRF, which addresses the challenge of training neural radiance fields with sparse input views. The researchers develop a visibility prior using plane sweep volumes, avoiding the need for pre-training on large datasets. This method significantly improves scene representation and depth estimation, especially useful in virtual and augmented reality, where limited view data is common. The paper details the efficiency and effectiveness of ViP-NeRF in rendering high-quality 3D scenes.
    
    \item \textbf{Neural Radiance Fields (NeRFs): A Review and Some Recent Developments} reviews the NeRF framework and its advancements. It covers the original NeRF's use of a fully connected neural network for 3D scene representation and novel view synthesis, emphasizing photorealistic image rendering. The paper also discusses recent NeRF developments like PixelNeRF and RegNeRF, which address challenges in fewer-image calibration and view synthesis. Applications extend to 3D rendering in fields like cinematography and virtual reality.
    
    \item \textbf{GeneFace++: Generalized and Stable Real-Time Audio-Driven 3D Talking Face Generation} proposes GeneFace++, a system for generating realistic 3D talking faces in real-time, driven by audio input. The method integrates pitch-aware audio-to-motion mapping and landmark locally linear embedding to enhance temporal consistency and naturalness in facial landmark prediction. It also employs an efficient dynamic NeRF named Instant Motion-to-Video module for real-time rendering. This advancement in 3D talking face generation is especially relevant for applications in virtual reality, digital communication, and entertainment.
    
    \item \textbf{Federated Neural Radiance Fields} presents FedNeRF, a novel federated learning algorithm for NeRF. It enables distributed and parallel training of NeRF across multiple nodes without pooling data, reducing bandwidth and preserving privacy. Applied in scenarios like multi-agent sensing, it demonstrates efficiency in 3D modeling and collaborative learning. The method incorporates low-rank decomposition to compress model updates, offering bandwidth reduction and fast convergence without significantly impacting accuracy.
    
    \item \textbf{Inverse Global Illumination using a Neural Radiometric Prior}
    
    \item \textbf{Real-Time Radiance Fields for Single-Image Portrait View Synthesis} presents a one-shot method for converting single unposed images into photorealistic 3D representations in real time. This method leverages a Vision Transformer-based triplane encoder and is trained solely on synthetic data. It's particularly effective for applications like AR/VR and 3D telepresence, offering fast performance (24 fps on consumer hardware) and high-quality results without the need for test-time optimization. The technique demonstrates robust handling of challenging real-world input images, including occlusions and side views.
    
    \item \textbf{Shap-E: Generating Conditional 3D Implicit Functions} explores generating 3D objects using text prompts, enhancing Neural Radiance Fields (NeRF) and Signed Distance Functions (SDF). The authors present a two-step method involving an encoder for implicit function parameters and diffusion models for latent representations. This technique offers improved 3D asset creation for virtual reality, gaming, and digital content development, demonstrating significant advancements in conditional 3D object generation from descriptive texts.
    
    \item \textbf{Semantic-aware Generation of Multi-view Portrait Drawings} introduces SAGE, a method to synthesize multi-view portrait drawings using semantic-aware techniques. The researchers propose a two-stage generation and training strategy, focusing on semantic maps to guide synthesis, providing structural consistency. This method is particularly adept in artistic style applications, outperforming existing 3D-aware image synthesis methods in generating multi-view portraits across a wide range of viewpoints. The approach has potential applications in virtual reality, gaming, and digital media, where accurate and diverse portrait rendering is essential.
    
    \item \textbf{NeRSemble: Multi-view Radiance Field Reconstruction of Human Heads} presents a novel technique for creating high-fidelity radiance fields of human heads, enabling the synthesis of realistic novel views of dynamic human faces. This method, developed by researchers at the Technical University of Munich, utilizes a new multi-view capture setup with 16 calibrated machine vision cameras, capturing over 4700 high-resolution sequences of more than 220 human heads. By combining a deformation field and an ensemble of 3D multi-resolution hash encodings, the authors effectively model the dynamics of facial expressions and head movements. This approach outperforms existing dynamic radiance field methods, offering significant advancements in photo-realistic rendering for applications in areas like VR, gaming, and digital communication. The authors also introduce a new benchmark for human head reconstruction to facilitate further research.
    
    \item \textbf{Single-Shot Implicit Morphable Faces with Consistent Texture Parameterization} introduces a hybrid morphable face model that merges the benefits of implicit 3D representations and explicit UV texture maps. The authors, affiliated with Stanford University and NVIDIA, developed a single-shot inversion framework to create animatable 3D avatars from single images, leveraging a deep learning approach. This model significantly enhances photo-realism, geometry, and expression accuracy in 3D face reconstruction, holding potential applications in virtual reality, gaming, and digital telepresence. Their method demonstrates superior performance compared to existing models, especially in texture and geometry reconstruction for avatars.
    
    \item \textbf{NeuralEditor: Editing Neural Radiance Fields via Manipulating Point Clouds} introduces NeuralEditor, a tool for editing neural radiance fields (NeRFs) by manipulating point clouds. The authors, from the University of Illinois and Peking University, combine the precision of point clouds with the rendering capabilities of NeRFs. NeuralEditor uses K-D tree-guided density-adaptive voxels for efficient rendering and optimizes the shape editing process through precise point clouds. This method shows potential for applications in virtual reality, gaming, and digital media, offering enhanced flexibility and accuracy in 3D object and scene editing.
        
    \item \textbf{NeRF-QA: Neural Radiance Fields Quality Assessment Database}, by Pedro Martin, António Rodrigues, João Ascenso, and Maria Paula Queluz, presents a comprehensive database, NeRF-QA, comprising 48 videos synthesized using seven different NeRF-based methods. This database, which includes both real and synthetic 360-degree scenes, is accompanied by perceived quality scores derived from subjective assessment tests. The authors' primary objective is to evaluate the effectiveness of existing objective quality metrics for NeRF-based synthesized views and to facilitate the development of new, more specific quality metrics.
    
    The paper details the framework used for synthesizing views and conducting subjective quality assessments. This framework includes processes like video preprocessing, frame selection, pose estimation, NeRF training, view synthesis, and subjective assessment. The subjective quality assessment employs the Double Stimulus Continuous Quality Scale (DSCQS) method, ensuring a comprehensive evaluation of the perceptual quality of synthesized views by human viewers.
    
    The authors selected a subset of popular NeRF methods for their study, including DVGO, Instant-NGP, Mip-NeRF 360, NeRF++, Nerfacto, Plenoxels, and TensoRF. These methods were chosen based on their prominence in the scientific community and their suitability for synthesizing views of both real and synthetic scenes. The paper also discusses the experimental setup for the subjective assessment study, which involved synthesizing views from two popular datasets, Tanks and Temples and Realistic Synthetic 360°. The synthesized videos were then evaluated by participants using the DSCQS method, and the results were processed to obtain Differential Mean Opinion Score (DMOS) values for each video.
    
    The study's findings indicate a trade-off between performance and synthesis quality, with methods that took longer for training and rendering achieving better scores. Interestingly, for synthetic scenes, some synthesized videos received better subjective evaluations than the reference videos. This comprehensive analysis provides valuable insights into the perceptual quality of NeRF-based view synthesis and underscores the need for specialized quality metrics in this domain. The NeRF-QA database, with its extensive collection of synthesized videos and corresponding quality scores, stands as a significant contribution to the field, offering a robust platform for evaluating and enhancing the quality of NeRF-based synthesized views.
    
    \item \textbf{General Neural Gauge Fields}~\cite{zhan2023general} introduces a novel concept in the realm of neural fields, particularly focusing on neural radiance fields, to enhance the computation efficiency and rendering quality of 3D scenes. The authors address a critical aspect of neural field modeling: the transformation of the 3D coordinate system into another measuring system, such as 2D manifolds or hash tables, a process they refer to as gauge transformation. Traditionally, gauge transformations are pre-defined mapping functions, but the authors propose a groundbreaking approach where the gauge transformation is learned alongside the neural field in an end-to-end manner. This method addresses the limitations of pre-defined transformations, which can be sub-optimal for modeling neural fields.
    
    The paper extends the concept of gauge transformation into a general paradigm, encompassing both discrete and continuous cases, and develops a learning framework to jointly optimize gauge transformations and neural fields. A significant challenge in this approach is the potential for the learning of gauge transformations to collapse easily. To overcome this, the authors derive a general regularization mechanism based on the principle of information conservation during the gauge transformation. They also propose an information-invariant gauge transformation that inherently preserves scene information, thereby circumventing the high computation cost typically associated with gauge learning and regularization. This innovative approach is expected to yield superior performance in various applications, particularly in enhancing the efficiency and quality of 3D scene representation in computer vision and graphics.
    
    \item \textbf{Multi-Space Neural Radiance Fields}~\cite{yin2023multi} by Ze-Xin Yin et al. introduces a novel approach to enhance Neural Radiance Fields (NeRF) methods, particularly in scenes with reflective objects. Traditional NeRF methods struggle with rendering such scenes, often resulting in blurry or distorted images. The authors propose a multi-space neural radiance field (MS-NeRF) that represents a scene using a group of feature fields in parallel sub-spaces. This technique allows for a better understanding and handling of reflective and refractive objects, significantly improving the rendering quality of scenes with complex light paths, such as those involving mirrors.
    
    The MS-NeRF approach works as an enhancement to existing NeRF methods, requiring only minor computational overheads for training and inferring extra-space outputs. The authors demonstrate the superiority and compatibility of their approach using three representative NeRF-based models: NeRF, Mip-NeRF, and Mip-NeRF 360. They perform comparisons on a novel dataset consisting of 25 synthetic scenes and 7 real captured scenes, all featuring 360-degree viewpoints and complex reflections and refractions. The results show that MS-NeRF significantly outperforms existing single-space NeRF methods in rendering high-quality scenes with mirror-like objects.
    
    In their method, the authors use a multi-space module that modifies the output part of the vanilla NeRF. This module replaces the neural radiance field with neural feature fields, integrating features along the ray in each sub-space to collect feature maps. These maps encode color information and visibility from specific viewpoints. The MS module applies softmax function to the weights of each sub-space to form the final render results. This approach allows for automatic handling of mirror-like objects in 360-degree high-fidelity scene rendering, achieving significant improvements over existing methods both quantitatively and qualitatively.
    
    The paper also introduces a new dataset specifically designed for evaluating 360-degree high-fidelity rendering of scenes containing complex reflections and refractions. This dataset includes both synthetic and real-world scenes, providing a comprehensive benchmark for validating the ability to synthesize novel views with complex light paths. The authors' experiments demonstrate that the MS-NeRF method can effectively model reflection and refraction with small computational overheads, making it a valuable enhancement for NeRF-based methods in various applications, particularly in scenarios involving complex lighting conditions and reflective surfaces.
    
    \item \textbf{HashCC: Lightweight Method to Improve the Quality of the Camera-less NeRF Scene Generation}~\cite{olszewski2023hashcc} presents HashCC, an innovative approach for enhancing neural radiance fields (NeRF) image quality without requiring known camera positions. The authors, from the University of Warsaw, integrate Hash Encoding and Color Correction networks into NeRF architectures, addressing the challenges of blurry results in camera-less regimes. This advancement, with its simple implementation and minimal computational overhead, has potential applications in 3D scene reconstruction and virtual reality, offering improved color prediction and scene detail.
    
    \item \textbf{NerfAcc: Efficient Sampling Accelerates NeRFs}~\cite{li2023nerfacc} focuses on improving the efficiency of Neural Radiance Field (NeRF) rendering by enhancing sampling methods. The authors from UC Berkeley introduce NerfAcc, a Python toolbox that integrates advanced sampling techniques into NeRF-related methods, offering 1.5× to 20× speedups in training time with minimal codebase changes. This advancement is significant for applications in virtual reality and 3D rendering, demonstrating the importance of sampling in volumetric rendering and providing a unified view of various sampling strategies.
    
    \item \textbf{Instant-NeRF: Instant On-Device Neural Radiance Field Training via Algorithm-Accelerator Co-Designed Near-Memory Processing}~\cite{zhao2023instant}, by Zhao et al, addressed the challenge of reducing the training time for Neural Radiance Fields (NeRFs) on edge devices. Their profiling analysis identified memory-bound inefficiencies in NeRF training, particularly in computing embedding vectors and executing multi-layer perceptrons (MLPs). To overcome these bottlenecks, the authors proposed Instant-NeRF, a near-memory processing (NMP) framework specifically designed for on-device NeRF training. This framework tackled the unique workloads of NeRFs, including random hash table lookup, random point processing sequence, and heterogeneous bottleneck steps.
    
    Instant-NeRF's algorithm-accelerator co-design effectively utilized NMP architectures to alleviate memory-bound bottlenecks in NeRF training. The authors demonstrated the framework's effectiveness across eight datasets, showing significant improvements in training efficiency. This advancement in NeRF training technology is particularly beneficial for applications in Augmented and Virtual Reality (AR/VR), where rapid 3D scene reconstruction is crucial. Instant-NeRF's ability to enable instant on-device NeRF training opens new possibilities for immersive AR/VR experiences, making it a significant contribution to the field of computer vision and AR/VR technology.
    
    \item \textbf{NeRF2: Neural Radio-Frequency Radiance Fields}, by Zhao et al, introduced NeRF2, a novel approach for modeling the propagation of RF signals in complex environments. This method extends the concept of neural radiance fields from optics to electromagnetism, enabling precise predictions of RF signal behavior in scenarios where traditional models struggle. NeRF2 represents scenes as continuous volumetric functions, optimized using sparse signal measurements. It predicts the reception of RF signals at any position, given the transmitter's location, by incorporating both amplitude and phase information in a complex-valued multilayer perceptron (MLP).
    
    The authors addressed challenges unique to RF signals, such as their lower frequencies compared to visible light and the limitations of RF receivers. They developed two training approaches for single-antenna and array-antenna receivers, respectively. NeRF2's potential was demonstrated in applications like indoor localization and 5G MIMO, showing significant improvements in performance. The concept of turbo-learning was introduced, leveraging NeRF2's physical model to generate synthetic datasets that enhance the training of application-layer artificial neural networks (ANNs). This method significantly reduced the need for extensive real-world data collection while maintaining high accuracy. The field studies conducted validated NeRF2's effectiveness in practical RF applications, marking a significant advancement in the domain of wireless communication and signal processing.
    
    \item \textbf{HumanRF: High-Fidelity Neural Radiance Fields for Humans in Motion}, developed by Işık et al., represents a significant advancement in the field of neural radiance fields for capturing human motion. This method, designed for high-fidelity reconstruction of human actors from multi-view video input, enables playback from novel, unseen viewpoints. The key innovation in HumanRF lies in its dynamic video encoding, which captures fine details at high compression rates through a temporal matrix-vector decomposition. This approach allows for temporally coherent reconstructions of human actors in long sequences, addressing the challenge of representing high-resolution details in complex motion.
    
    The authors introduced ActorsHQ, a novel multi-view dataset providing 12MP footage from 160 cameras, to demonstrate the challenges of using high-resolution data and the effectiveness of HumanRF in leveraging this data. HumanRF significantly improves over existing state-of-the-art methods, marking a step towards production-level quality in novel view synthesis. The method's ability to handle resolutions of 12MP and its efficient use of a spatio-temporal decomposition based on a low-rank decomposition make it particularly suitable for applications in film production, computer games, and videoconferencing, where high-fidelity human representations are crucial.
    
    \item \textbf{SparseGNV: Generating Novel Views of Indoor Scenes with Sparse Input Views}, as presented by Cheng, Cao, and Shan in their paper, is a learning framework designed to generate novel views of indoor scenes from sparse input views. The framework combines 3D structures with image generative models to achieve photorealism and view consistency. SparseGNV consists of three modules: a neural geometry module that builds a neural point cloud for underlying geometry, a view generator module utilizing a transformer-based network for mapping scene context into a shared latent space and autoregressively decoding the target view, and an image converter module for reconstructing discrete image tokens into the target view image.
    
    The authors trained SparseGNV on a large indoor scene dataset, enabling it to learn generalizable priors and efficiently generate novel views of unseen indoor scenes in a feed-forward manner. The framework was evaluated on both real-world and synthetic indoor scenes, demonstrating superior performance over state-of-the-art methods based on either neural radiance fields or conditional image generation. The effectiveness of SparseGNV lies in its ability to synthesize consistent novel views with limited visual clues while maintaining scene consistency, making it a significant advancement in the field of computer vision and novel view synthesis.
    
    \item \textbf{Curvature-Aware Training for Coordinate Networks} by Saratchandran et al. presented a novel approach to training coordinate networks using second-order optimization methods, specifically L-BFGS, to significantly reduce training times while maintaining compressibility. The authors demonstrated the effectiveness of this method across various signal modalities, including audio, images, videos, shape reconstruction, and neural radiance fields. They identified that traditional first-order optimizers like Adam were slow for training coordinate networks, which was a barrier to their use in real-time applications. To address this, they proposed using second-order optimizers, which leverage the curvature information of the loss landscape for faster convergence.
    
    The paper provided a theoretical analysis showing that networks activated by sine or Gaussian functions guaranteed superlinear convergence with L-BFGS, unlike those with ReLU activations. This was empirically validated through experiments on image reconstruction tasks. The authors also introduced a patch-based decomposition strategy for large-scale datasets, which significantly accelerated training times compared to Adam. This strategy was particularly effective for modeling larger size signals, demonstrating up to 14 times faster training.
    
    In practical applications, the authors showcased the superiority of L-BFGS over Adam in tasks like 2D image reconstruction and novel view synthesis using neural radiance fields (NeRF). They introduced "KiloImage," a patch-based strategy for optimizing gigapixel images, and "KiloNeRF," for efficient training of large-scale NeRF models. These applications highlighted the real-world potential of the proposed method, offering significant improvements in training efficiency and quality of results for complex signal reconstruction tasks.
    
    \item \textbf{MV-Map: Offboard HD-Map Generation with Multi-view Consistency} a paper by Ziyang Xie, Ziqi Pang, and Yu-Xiong Wang, addresses the challenge of generating high-definition maps (HD Maps) for autonomous vehicles. The authors observed that existing bird’s-eye-view (BEV) perception models, while useful, often produce unreliable and inconsistent HD Maps due to their onboard computational limitations and inability to reason across multiple views. To overcome these limitations, they proposed an offboard HD Map generation pipeline named MV-Map, which leverages multi-view consistency and can handle an arbitrary number of frames.
    
    MV-Map operates in a "region-centric" framework, aggregating onboard predictions from all frames, weighted by confidence scores assigned by an "uncertainty network." This network is augmented with global 3D structure information optimized by a voxelized neural radiance field (Voxel-NeRF), enhancing multi-view consistency. The authors demonstrated that this approach significantly improves the quality of HD Maps, as evidenced by extensive experiments on the nuScenes dataset.
    
    The paper details the methodology behind MV-Map, including the onboard model for generating BEV features and semantic maps, the uncertainty network for assessing the reliability of single-frame information, and the integration of Voxel-NeRF for consistent 3D structure reconstruction. The authors also introduced a total-variance loss to focus the NeRF optimization on near-ground geometry, which is crucial for HD map generation. The training and inference procedures are described, highlighting the scalability and efficiency of MV-Map in handling large volumes of offboard data.
    
    In summary, MV-Map represents a significant advancement in offboard HD map generation, offering a practical and efficient solution for autonomous driving applications. The authors' approach effectively combines multi-view image data and 3D structure information to produce high-quality, consistent HD Maps, demonstrating the potential of offboard methods in this domain.
    
    \item \textbf{NerfBridge: Bringing Real-time, Online Neural Radiance Field Training to Robotics}, a paper presented at the ICRA 2023 Workshop on Unconventional Spatial Representations, introduces NerfBridge, an innovative tool designed to bridge the Robot Operating System (ROS) and Nerfstudio for real-time, online training of Neural Radiance Fields (NeRFs) from image streams. The authors, hailing from Stanford University, developed this open-source software to facilitate rapid research in applying NeRFs in robotics, offering an extensible interface to efficient training pipelines and model libraries provided by Nerfstudio. They demonstrated the utility of NerfBridge through a hardware setup involving a camera-mounted quadrotor, capable of training a NeRF in both indoor and outdoor environments.
    
    The paper delves into the background of NeRFs, highlighting their potential in robotics due to their ability to model complex, multi-scale real-world environments. However, traditional NeRF training libraries, while fast, are designed for offline use and necessitate a slow pose optimization pre-computation step. NerfBridge addresses this limitation by enabling real-time, online training of NeRFs, thus opening new avenues for robotics applications.
    
    The authors outlined the basic functionality of NerfBridge, explaining how it integrates streaming images with real-time NeRF training. The process involves continuously updating a NeRF using pixels from a pool of images, with the training proceeding on a static dataset until convergence. They emphasized the importance of real-time pose estimation for this process, which is often overlooked in NeRF literature. In their implementation, they used the ORBSLAM3 package for visual odometry to estimate camera poses.
    
    The practical applications of NerfBridge were demonstrated through indoor and outdoor mapping case studies using a quadrotor. These experiments showcased the ability of NerfBridge to construct detailed NeRFs of environments, capturing intricate details like building facades and room interiors. The authors concluded by highlighting the potential of NerfBridge in streamlining the integration of neural implicit maps in robotics and its role in fostering research at the intersection of robotics and neural implicit scene representations. They also pointed towards future work, including combining NeRF navigation algorithms with online training for robot trajectory optimization and mapping.
    
    \item \textbf{OR-NeRF: Object Removing from 3D Scenes Guided by Multiview Segmentation with Neural Radiance Fields}, developed by Youtan Yin, Zhoujie Fu, Fan Yang, and Guosheng Lin, presents a novel pipeline for efficiently removing objects from 3D scenes using Neural Radiance Fields (NeRF). This method addresses the challenges of time-consuming object labeling, limited capability to remove specific targets, and compromised rendering quality after removal. OR-NeRF allows for object removal with user-given points or text prompts on a single view, significantly reducing the time required for multiview segmentation and improving performance compared to previous methods. The authors' approach involves spreading user annotations to all views through 3D geometry and sparse correspondence, ensuring 3D consistency with less processing burden. They utilized the 2D segmentation model Segment-Anything (SAM) to predict masks and a 2D inpainting model to generate color supervision. The algorithm applies depth supervision and perceptual loss to maintain consistency in geometry and appearance after object removal. Experimental results demonstrated that OR-NeRF achieves better editing quality in less time than previous works, considering both quality and quantity. This method offers a practical solution for 3D scene editing, particularly in applications requiring efficient and accurate object removal.
    
    \item \textbf{MultiPlaneNeRF: Neural Radiance Field with Non-Trainable Representation}, introduced by Dominik Zimny, Artur Kasymov, Adam Kania, Jacek Tabor, Maciej Zieba, and Przemyslaw Spurek, presents an innovative approach to representing 3D objects from 2D images using Neural Radiance Fields (NeRF). This model addresses the limitations of vanilla NeRF, such as the requirement for separate training on each object and poor generalization to unseen data. MultiPlaneNeRF operates directly on 2D images, using a projection of 3D points onto these images to create non-trainable representations. This process does not involve parametrization, and a shallow decoder efficiently processes the representation. The authors demonstrated that MultiPlaneNeRF could be trained on a large dataset, enabling the implicit decoder to generalize across various objects. Consequently, simply replacing the 2D images allows for the generation of a new NeRF representation of an object without additional training. The experimental results showed that MultiPlaneNeRF achieves comparable performance to state-of-the-art models in synthesizing new views and possesses generalization capabilities. Furthermore, the MultiPlane decoder can be integrated into larger generative models like GANs, showcasing its versatility and potential for broad application in fields requiring efficient and accurate 3D scene reconstruction from limited 2D data.
    
    \item \textbf{ConsistentNeRF (Enhancing Neural Radiance Fields with 3D Consistency for Sparse View Synthesis)}, developed by Shoukang Hu, Kaichen Zhou, Kaiyu Li, Longhui Yu, Lanqing Hong, Tianyang Hu, Zhenguo Li, Gim Hee Lee, and Ziwei Liu, addresses the challenge of sparse view synthesis in Neural Radiance Fields (NeRF). The authors observed that learning the 3D consistency of pixels among different views is crucial for enhancing reconstruction quality in sparse view settings. ConsistentNeRF employs depth-derived geometry information and a depth-invariant loss to focus on pixels exhibiting 3D correspondence and maintaining consistent depth relationships. This method significantly improves model performance in sparse view conditions, achieving remarkable improvements in PSNR, SSIM, and LPIPS metrics across various benchmarks. The authors introduced a solution that integrates both multi-view and single-view 3D consistency to optimize NeRF in sparse view scenarios. They utilized depth-invariant loss to extract 3D consistency information from nearby views using the DPT Large pre-trained model. This approach led to state-of-the-art results compared to existing methods across diverse datasets. The significant improvements demonstrated by ConsistentNeRF highlight its effectiveness in enhancing 3D consistency in Neural Radiance Fields, showcasing its potential applicability in real-world situations where sparse view data is common.
    
    \item \textbf{Text2NeRF (Text-Driven 3D Scene Generation with Neural Radiance Fields)}, a novel framework for text-driven 3D scene generation, was introduced by Jingbo Zhang, Xiaoyu Li, Ziyu Wan, Can Wang, and Jing Liao. This method, capable of creating complex and realistic 3D scenes from textual descriptions, leverages Neural Radiance Fields (NeRF) and a pre-trained text-to-image diffusion model. The authors employed the diffusion model to generate content and geometric priors, which were then used to update the NeRF model, ensuring textured and geometric consistency across different views. A key innovation in Text2NeRF is the Progressive Inpainting and Updating (PIU) strategy, which allows for novel view synthesis by progressively expanding and updating the scene. This approach overcomes limitations of previous methods, which were constrained to simple geometries and styles, by enabling the generation of diverse, high-fidelity, and view-consistent 3D scenes. The authors demonstrated Text2NeRF's superiority over existing methods through extensive experiments, showcasing its potential in applications like video gaming, film industry, and metaverse development. The method's generality allows for the creation of a wide range of 3D scenes, including indoor, outdoor, and artistic settings, without the need for additional training data or multi-view constraints.
    
    \item \textbf{Registering Neural Radiance Fields as 3D Density Images} by Han Jiang, Ruoxuan Li, Haosen Sun, Yu-Wing Tai, and Chi-Keung Tang presents a novel method for registering Neural Radiance Fields (NeRFs) by converting them into 3D density images. This approach allows for the alignment of partially overlapping 3D scenes represented by NeRFs using a rigid transformation. The authors developed a framework that generalizes the traditional registration pipeline, which includes key point detection and point set registration, to operate on 3D density fields. They proposed the use of universal pre-trained descriptor-generating neural networks to describe corner points as key points in 3D. These networks can be trained and tested on different scenes using a contrastive learning strategy. The method involves discretizing the continuous neural field into 3D density images, detecting corners in these images, and then using a neural network to generate rotation-invariant descriptors from 3D density image patches. The descriptors are matched to obtain correspondences, and RANSAC is used to compute the rigid transformation between the two corner point sets. This approach effectively registers NeRF models, enabling the construction of large-scale NeRFs by registering smaller, overlapping NeRFs captured individually. The paper's contributions include the first 3D density image-based NeRF registration framework and a universal neural 3D corner descriptor, coupled with a strategy to train this network with contrastive learning. This advancement has significant implications for applications such as virtual reality and panoramic indoor and outdoor maps, where accurate and efficient registration of 3D scenes is crucial.
    
    \item \textbf{NeRFuser (Large-Scale Scene Representation by NeRF Fusion)} by Jiading Fang, Shengjie Lin, Igor Vasiljevic, Vitor Guizilini, Rares Ambrus, Adrien Gaidon, Gregory Shakhnarovich, and Matthew R. Walter, introduces a novel architecture for the registration and blending of pre-trained Neural Radiance Fields (NeRFs). The authors developed a method that operates on pre-generated NeRFs without needing the large sets of images typically used to generate them. Their approach, registration from re-rendering, infers the transformation between NeRFs based on images synthesized from individual NeRFs. For blending, they propose sample-based inverse distance weighting to blend visual information at the ray-sample level. NeRFuser significantly enhances the utility of NeRFs as a data representation, enabling the fusion of multiple NeRFs into a coherent scene representation. This method is particularly beneficial for large-scale scene reconstruction, where it is impractical to capture the entire scene in a single NeRF model. By fusing multiple NeRFs, NeRFuser can handle larger scenes with improved performance and reduced memory consumption. The paper demonstrates the robustness of NeRFuser on public benchmarks and a self-collected object-centric indoor dataset. This advancement in NeRF technology has broad implications for applications requiring efficient and high-quality rendering of large-scale 3D scenes, such as virtual reality, augmented reality, and digital content creation. NeRFuser's ability to register and blend multiple NeRFs without access to the original image datasets marks a significant step forward in the field of 3D scene reconstruction and rendering.
    
    \item \textbf{Weakly Supervised 3D Open-vocabulary Segmentation} by Kunhao Liu, Fangneng Zhan, Jiahui Zhang, Muyu Xu, Yingchen Yu, Abdulmotaleb El Saddik, Christian Theobalt, Eric Xing, and Shijian Lu presents a novel approach for 3D open-vocabulary segmentation of scenes. The authors addressed the challenge of the lack of large-scale and diverse 3D segmentation datasets by exploiting pre-trained foundation models CLIP and DINO in a weakly supervised manner. Their method utilizes open-vocabulary text descriptions of objects in a scene to distill multimodal knowledge and object reasoning capability into a Neural Radiance Field (NeRF), effectively lifting 2D features into view-consistent 3D segmentation without requiring manual segmentation annotations. The approach involves constructing a hierarchical set of image patches to extract pixel-level features from image-level CLIP features and designing a 3D Selection Volume to identify the appropriate hierarchical level for each 3D point. Additionally, they introduced a Relevancy-Distribution Alignment (RDA) loss to address CLIP feature ambiguities and a Feature-Distribution Alignment (FDA) loss to distill object boundary information from DINO features. This method enables precise and annotation-free 3D open-vocabulary segmentation, effectively handling text labels with long-tail distributions. The authors demonstrated that their method surpasses fully supervised models trained with segmentation annotations in certain scenes, highlighting the potential of learning 3D open-vocabulary segmentation from 2D images and text-image pairs. This advancement has significant implications for applications such as robot navigation, object localization, autonomous driving, 3D scene editing, and augmented/virtual reality, where accurate and versatile 3D scene understanding is crucial.
    
    \item \textbf{OD-NeRF (Efficient Training of On-the-Fly Dynamic Neural Radiance Fields)} by Zhiwen Yan, Chen Li, and Gim Hee Lee introduces OD-NeRF, a novel approach for efficiently training and rendering dynamic Neural Radiance Fields (NeRFs) on-the-fly. This method is designed to stream dynamic scenes as they unfold, in contrast to traditional dynamic NeRFs that require complete video sequences for training. The authors tackled the challenge of utilizing radiance fields estimated from previous frames effectively by proposing a NeRF model conditioned on multi-view projected colors and a transition and update algorithm for the occupancy grid. OD-NeRF leverages the projected colors from training views to implicitly track correspondence between current and previous frames, enabling fast convergence when training radiance fields of consecutive frames. The transition and update to the occupancy grid, used for efficient point sampling, further enhance the model's efficiency. This approach allows for interactive speed training and rendering, achieving 6 frames per second (FPS) on synthetic dynamic scenes and significant speed-up on real-world dynamic scenes. The paper's contributions include the introduction of on-the-fly training for dynamic NeRF, a projected color-guided dynamic NeRF model, and a transiting occupancy grid for efficient training. These advancements enable applications in social media, VR/AR, and gaming industries, where streaming dynamic scenes in real-time is crucial. OD-NeRF's ability to train and render dynamic scenes on-the-fly marks a significant step forward in the field of 3D scene reconstruction and rendering.
    
    \item \textbf{InpaintNeRF360 (Text-Guided 3D Inpainting on Unbounded Neural Radiance Fields)} by Dongqing Wang, Tong Zhang, Alaa Abboud, and Sabine Süsstrunk introduces a groundbreaking method for editing 3D scenes represented by Neural Radiance Fields (NeRF) using natural language instructions. The authors developed InpaintNeRF360, a framework that leverages a promptable segmentation model, Segment Anything Model (SAM), to perform multiview segmentation of objects in a scene based on encoded text. This approach addresses the challenge of editing NeRF scenes, particularly removing objects while maintaining geometric and photometric consistency, which is difficult due to NeRF's implicit scene representation. InpaintNeRF360 employs depth-space warping to ensure viewing consistency in the segmentations and refines the inpainted NeRF model using perceptual priors for visual plausibility. The method can simultaneously remove multiple objects or modify object appearance based on text instructions, synthesizing 3D viewing-consistent and photo-realistic inpainting. The authors demonstrated the effectiveness of InpaintNeRF360 on both unbounded and frontal-facing scenes trained through NeRF, showcasing its potential to enhance the editability of implicit radiance fields. This advancement in NeRF technology has significant implications for virtual reality, augmented reality, and digital content creation, where realistic and editable 3D scene representations are crucial. InpaintNeRF360's ability to interpret and execute complex editing instructions through natural language makes it a powerful tool for interactive 3D scene manipulation.
    
    \item \textbf{Interactive Segment Anything NeRF with Feature Imitation} by Xiaokang Chen, Jiaxiang Tang, Diwen Wan, Jingbo Wang, and Gang Zeng presents a novel approach to enhance Neural Radiance Fields (NeRF) with semantic understanding for interactive applications. The authors identified the lack of semantics in NeRF as a limitation for interaction in complex scenes and proposed a framework that integrates semantic feature imitation with NeRF. This method allows for zero-shot semantic segmentation in 3D space, leveraging the backbone features of off-the-shelf perception models to optimize NeRF parameters. Their framework reformulates the segmentation process by rendering semantic features directly and applying only the decoder from perception models, thus avoiding the need for expensive backbones and enhancing 3D consistency. The authors demonstrated that their method could project learned semantics onto extracted mesh surfaces, enabling real-time interaction and applications like texture editing and model composition. With the Segment Anything Model (SAM), their framework achieved a 16x acceleration in segmentation with comparable mask quality. This advancement in NeRF technology has significant implications for virtual reality, digital creation, and other applications requiring complex scene interactions. By enabling efficient and accurate semantic segmentation in 3D space, this method opens up new possibilities for interactive content creation and manipulation in virtual environments.
    
    \item \textbf{ZeroAvatar (Zero-shot 3D Avatar Generation from a Single Image)} by Zhenzhen Weng, Zeyu Wang, and Serena Yeung introduces ZeroAvatar, a method that significantly advances the field of zero-shot 3D avatar generation from a single image. The authors addressed the challenge of preserving the geometry of complex shapes, particularly human bodies, which existing methods often struggle with. ZeroAvatar incorporates an explicit 3D human body prior into the optimization process, using a parametric human body model to estimate and refine the parameters from a single image. This model serves as an additional geometry constraint to regularize the diffusion model and the underlying density field. The method also includes a UV-guided texture regularization term to guide the completion of texture on invisible body parts. ZeroAvatar leverages depth information from the posed body model as conditioning in addition to text (image caption) for the Stable Diffusion model, enhancing the fidelity and realism of the generated humans. The authors demonstrated that ZeroAvatar outperforms existing zero-shot image-to-3D methods, offering significant improvements in both geometry and appearance of the generated avatars. This advancement has broad implications for applications such as virtual character generation, animation, and augmented reality, where realistic and detailed 3D human avatars are essential. ZeroAvatar's ability to handle a wide range of human forms, from real-world humans to virtual avatars like cartoon characters, expands its potential in creative and practical applications.
    
    \item \textbf{Joint Optimization of Triangle Mesh, Material, and Light from Neural Fields with Neural Radiance Cache} by Jiakai Sun, Zhanjie Zhang, Guangyuan Li, Tianyi Chu, Lei Zhao, and Wei Xing introduces an innovative framework, JOC, for inverse rendering. This framework effectively combines the strengths of neural fields and modern graphics pipelines. The authors tackled the challenge of integrating neural field-based methods, which excel in 3D reconstruction but struggle with indirect illumination, with traditional graphics pipelines that are efficient but limited by the complexity of differentiable rendering. JOC operates in two stages: initially, it trains a Neural Radiance Cache (NRC) using a pre-trained neural field, capturing global illumination with a single-bounce differentiable Monte Carlo rendering. This approach allows for the joint optimization of geometry, material, and light through back-propagation. The framework uses Signed Distance Fields (SDF) for geometry representation, Disney's Physically-Based Rendering (PBR) BRDF model for materials, and High Dynamic Range (HDR) light probes for lighting. The paper demonstrates that JOC can effectively prevent indirect illumination effects from being baked into materials, resulting in high-quality reconstruction of triangle mesh, PBR material, and HDR light probe. This advancement is significant for applications in digital content creation and game engines, where accurate and efficient rendering of complex scenes is crucial. JOC's ability to integrate with modern graphics pipelines while leveraging the reconstruction capabilities of neural fields marks a notable step forward in the field of computer graphics and vision.

    \item \textbf{PlaNeRF (SVD Unsupervised 3D Plane Regularization for NeRF Large-Scale Scene Reconstruction)} by Fusang Wang, Arnaud Louys, Nathan Piasco, Moussab Bennehar, Luis Roldão, and Dzmitry Tsishkou introduces a novel method to enhance the geometry reconstruction of Neural Radiance Fields (NeRF) in large-scale outdoor urban scenes. The authors identified a key limitation in NeRF's ability to accurately reconstruct geometry, particularly in low-texture areas common in driving scenarios. To address this, they developed PlaNeRF, an unsupervised regularization method that does not rely on explicit 3D supervision. PlaNeRF employs a novel plane regularization technique based on Singular Value Decomposition (SVD) and leverages the Structural Similarity Index Measure (SSIM) in a patch-based loss design to initialize the volumetric representation of NeRF effectively. The authors demonstrated that PlaNeRF significantly improves NeRF's 3D structure using only RGB images and semantic maps, leading to more accurate geometry reconstruction in large outdoor scenes. This approach outperforms existing regularization methods in geometric accuracy and achieves comparable rendering quality to state-of-the-art methods on the KITTI-360 Novel View Synthesis benchmark. PlaNeRF's ability to improve geometry without relying on explicit or prior 3D information makes it a valuable tool for applications in autonomous driving, virtual reality, HD mapping, simulation, and scene editing, where accurate geometry for surface reconstruction is crucial.
    
    \item \textbf{Volume Feature Rendering for Fast Neural Radiance Field Reconstruction} by Kang Han Wei, Xiang Lu, and Yu Lu introduces a novel approach to Neural Radiance Fields (NeRFs) that significantly enhances rendering speed and quality. The authors proposed a method called Volume Feature Rendering (VFR), which fundamentally alters the standard volume rendering framework used in NeRFs. Instead of rendering the yielded color after neural network evaluation, VFR first renders the queried feature vectors of a ray and then transforms the rendered feature vector into the final pixel color using a neural network. This change allows for only a single neural network evaluation per pixel, drastically reducing the high computational complexity typically associated with multiple neural network evaluations in standard NeRF implementations. The VFR method leverages the advantages of volume representation in modeling complex geometry while also benefiting from the strength of single neural network evaluation, akin to the Signed Distance Function (SDF) approach. The authors also introduced a pilot network to assist in finding coarse geometry during the early training stage, addressing potential convergence issues. This approach enables the use of larger neural networks for better rendering quality while maintaining comparable training and rendering time costs. The paper demonstrates that VFR achieves state-of-the-art rendering quality on both synthetic and real-world datasets with a training time of just several minutes, marking a significant advancement in the field of 3D scene reconstruction and rendering. This method has broad implications for applications requiring efficient and high-quality rendering of complex 3D scenes, such as virtual reality, augmented reality, and digital content creation.
    
    \item \textbf{Towards a Robust Framework for NeRF Evaluation} by Adrian Azzarelli, Nantheera Anantrasirichai, and David R Bull, addresses the critical need for reliable methods to evaluate Neural Radiance Fields (NeRFs). Recognizing the limitations of conventional image quality assessment methods and analytical metrics, the authors proposed a new test framework that isolates the neural rendering network from the NeRF pipeline. This framework performs a parametric evaluation by training and evaluating the NeRF on an explicit radiance field representation. The authors introduced a configurable approach for generating representations specifically for evaluation purposes, employing ray-casting to transform mesh models into explicit NeRF samples and to "shade" these representations. This novel framework allows for the evaluation of different "tasks" (scenes with various visual effects or learning strategies) and types of networks (NeRFs and depth-wise implicit neural representations). The authors also proposed a new metric to measure task complexity, considering visual parameters and the distribution of spatial data. Their approach offers the potential to create a comparative objective evaluation framework for NeRF methods, addressing the challenges in benchmarking the performance of NeRFs, especially as state-of-the-art methods become closer in performance. This framework is a significant step towards robust and consistent objective parametric evaluation of NeRFs, providing a basis for evaluating the performance of different networks on the same scene with various material effects.
    
    \item \textbf{Compact Real-time Radiance Fields with Neural Codebook} by Lingzhi Li, Zhongshu Wang, Zhen Shen, Li Shen, and Ping Tan introduces a novel framework for creating compact radiance fields, addressing the storage and transmission overhead typically associated with grid-based neural radiance field representations like Plenoxels. The authors developed a non-uniform compression stem to significantly reduce model complexity and introduced the Neural Codebook, a parameterized module for encoding high-frequency details specific to per-scene models through fast optimization. This approach achieved over 40× reduction in grid model storage while maintaining competitive rendering quality and real-time rendering speed of 180 fps. The framework exploits intrinsic properties in grid models, using a non-uniform compression strategy to compress full-scale grid models and a position embedding modulated component, the Neural Codebook, to encode high-frequency information effectively. The authors demonstrated that their method could achieve substantial storage reduction compared to original grid models with minimal quality loss on benchmark datasets. They also employed several strategies to improve inference efficiency without relying on octree-based structures, achieving a smaller storage cost than modern real-time rendering methods and presenting the best tradeoff between storage cost and rendering speed. This advancement in neural radiance field technology has significant implications for applications requiring efficient storage and transmission of high-quality 3D scene representations, such as virtual reality and digital content creation.
    
    \item \textbf{HiFA (High-fidelity Text-to-3D Generation with Advanced Diffusion Guidance)} by Junzhe Zhu, Peiye Zhuang, and Sanmi Koyejo, presents a groundbreaking approach to text-to-3D generation, aiming to overcome the limitations of existing methods that often result in artifacts and inconsistencies across different views. The authors introduced a single-stage optimization process that utilizes both latent and image spaces of a pre-trained text-to-image diffusion model for enhanced supervision. A novel aspect of their method is the timestep annealing approach, which progressively reduces the sampled timestep throughout optimization, addressing the issue of divergence in denoised images and ensuring finer detail capture in later training stages. To tackle the challenges of generating detailed 3D assets in a single-stage optimization, the authors proposed two key techniques: variance regularization for z-coordinates along NeRF rays and a kernel smoothing technique for importance sampling. The variance regularization effectively mitigates the cloudiness issue in NeRFs, enabling the representation of crisp geometrical surfaces. The kernel smoothing technique addresses texture flickering issues in NeRFs by refining importance sampling weights, ensuring accurate and comprehensive sampling in high-density regions. The paper demonstrates that these holistic modifications significantly enhance the quality of 3D synthesis, enabling the generation of highly detailed and view-consistent 3D assets through a single-stage training process. This advancement has broad implications for applications in digital content generation, filmmaking, and virtual reality, where high-quality and realistic 3D models are essential. The authors' work marks a significant step forward in the field of 3D asset generation, offering a more efficient and effective method for creating detailed and accurate 3D models from textual descriptions.
    
    \item \textbf{Template-free Articulated Neural Point Clouds for Reposable View Synthesis} by Lukas Uzolas, Elmar Eisemann, and Petr Kellnhofer introduces a novel method for synthesizing novel views of dynamic 3D scenes using Neural Radiance Fields (NeRFs). The authors address the limitations of backward deformation fields in reanimating captured object poses, which often result in low visual fidelity, long reconstruction times, or applicability to only narrow domains. Their method utilizes a point-based representation and Linear Blend Skinning (LBS) to learn a Dynamic NeRF and an associated skeletal model from sparse multi-view video. This forward-warping approach significantly reduces learning time and achieves state-of-the-art visual fidelity in novel view and pose synthesis. The authors demonstrate the versatility of their representation on various articulated objects from common datasets, achieving reposable 3D reconstructions without needing object-specific skeletal templates. Their method stands out for its ability to handle a wide range of objects and poses with high image-synthesis quality and efficient training times, making it a valuable tool for applications in virtual/augmented reality, video games, and movie productions. The technique's class-agnostic nature and lack of reliance on templates or pose annotations mark a significant advancement in the field of dynamic 3D scene reconstruction and manipulation.
    
    \item \textbf{DaRF (Boosting Radiance Fields from Sparse Inputs with Monocular Depth Adaptation)} presents a novel framework, DäRF, which significantly enhances the performance of Neural Radiance Fields (NeRF) in scenarios with limited viewpoint data. The authors tackled the challenge of NeRF's performance degradation when known viewpoints are drastically reduced, a common issue in real-world applications. Their approach integrates Monocular Depth Estimation (MDE) networks, known for their strong generalization capability, with NeRF. This integration, however, posed new challenges due to the ambiguity in monocular depths. To address these, the authors introduced a method for online complementary training, which combines the strengths of NeRF and MDE. This method involves patch-wise scale-shift fitting and geometry distillation, adapting the MDE network to produce depths aligned with NeRF geometry. The framework imposes MDE's geometry prior on NeRF at both seen and unseen viewpoints, enhancing robustness and coherence. The authors demonstrated that DäRF achieves state-of-the-art results in both indoor and outdoor real-world datasets, showcasing its effectiveness in robust NeRF reconstruction with a limited number of real-world images. This advancement has significant implications for applications requiring 3D reconstruction and novel view synthesis, such as virtual reality, augmented reality, and 3D modeling, where obtaining densely well-calibrated input images is challenging.
    
    \item \textbf{FlowCam (Training Generalizable 3D Radiance Fields without Camera Poses via Pixel-Aligned Scene Flow)} by Cameron Smith, Yilun Du, Ayush Tewari, and Vincent Sitzmann introduces a groundbreaking method for reconstructing 3D neural fields from unposed images. This technique addresses the challenge of deploying 3D scene learners on large-scale video data, which traditionally depends on precise camera poses obtained from structure-from-motion. The authors propose a novel approach that jointly reconstructs camera poses and 3D neural scene representations online in a single forward pass. The method estimates poses by first converting frame-to-frame optical flow to 3D scene flow using differentiable rendering. This process maintains the locality and shift-equivariance of the image processing backbone. Camera poses are then estimated through a weighted least-squares fit to the scene flow field. This formulation enables end-to-end, self-supervised training on real-world video datasets, bypassing the need for ground-truth camera poses or depth maps. The authors validate their model on various real-world datasets, demonstrating robust performance across diverse scenes, including those challenging for optimization-based pose estimation techniques. This research has significant implications for 3D reconstruction and representation learning, offering a scalable and efficient solution for processing large-scale video data. The technique's ability to operate without precise camera poses opens up new possibilities for applications in virtual reality, augmented reality, and other fields requiring dynamic 3D scene understanding.
    
    \item \textbf{AvatarStudio (Text-driven Editing of 3D Dynamic Human Head Avatars)} by Mohit Mendiratta and colleagues introduces a novel method for editing dynamic 3D human head avatars using text prompts. The technique leverages neural radiance fields (NeRF) and text-to-image diffusion models to achieve high-quality, personalized edits that are consistent in both 3D and time domains. The authors developed an optimization strategy to incorporate multiple keyframes representing different camera viewpoints and time stamps into a single diffusion model. This approach, combined with their view-and-time-aware Score Distillation Sampling (VT-SDS), allows for coherent editing across various views and timeframes. The method's uniqueness lies in its ability to maintain the integrity of the input identity while allowing a wide range of text-based edits. The authors evaluated their approach through user studies, demonstrating its superiority over existing methods in generating temporally coherent results. This research has significant implications for virtual character creation in extended reality and media production, where dynamic and realistic human head avatars are essential. The technique's ability to edit avatars based on simple text inputs makes it accessible and user-friendly, potentially revolutionizing how digital faces are controlled and manipulated in various applications.
    
    \item \textbf{Analyzing the Internals of Neural Radiance Fields} by Lukas Radl, Andreas Kurz, and Markus Steinberger, explores the potential of Neural Radiance Fields (NeRFs) in learning mappings from position to volumetric density. The authors focus on the analysis of large, trained Rectified Linear Unit Multi-Layer Perceptrons (ReLU-MLPs) used in coarse-to-fine sampling NeRFs. They discover that trained NeRFs and Mip-NeRFs map samples with high density to local minima in activation feature space. This insight leads to a novel method for accelerating these large MLPs by transforming intermediate activations into a weight estimate, achieving up to 50\% reduction in computational requirements with minimal impact on rendering quality. The authors present a method for visualizing and analyzing intermediate activations of coordinate-based ReLU-MLPs. They also propose an approach for extracting a density estimate from early layers of coarse NeRFs, significantly reducing inference time. This method is validated on various architectures and datasets, demonstrating its applicability in different settings. The paper's contributions extend to real-world applications, particularly in rendering and visualizing complex scenes efficiently. By leveraging the intermediate activations of NeRFs, the authors provide a pathway to more efficient rendering processes, which is crucial in applications like virtual reality, augmented reality, and 3D modeling, where computational efficiency and rendering quality are paramount. This research not only enhances the understanding of NeRFs but also opens up new possibilities for their application in real-time rendering and complex scene visualization.
    
    \item \textbf{PanoGRF (Generalizable Spherical Radiance Fields for Wide-baseline Panoramas)}, authored by Zheng Chen, Yan-Pei Cao, Yuan-Chen Guo, Chen Wang, Ying Shan, and Song-Hai Zhang, introduces a novel approach to synthesizing novel views from wide-baseline panoramas, a key challenge in virtual reality (VR) applications. The authors developed PanoGRF, a method that constructs spherical radiance fields incorporating 360° scene priors, avoiding the information loss from panorama-to-perspective conversion. PanoGRF directly aggregates geometry and appearance features of 3D sample points from each panoramic view based on spherical projection. To improve geometry features, the authors incorporated 360° monocular depth priors into spherical depth estimation. This method significantly outperforms state-of-the-art generalizable view synthesis methods for wide-baseline panoramas. PanoGRF addresses the overfitting problem common in spherical radiance fields and enhances the accuracy of depth estimation in 360° multi-view stereo. The authors demonstrated the effectiveness of PanoGRF on multiple panoramic datasets, showcasing its superiority in rendering performance and its potential for immersive VR experiences. This method is particularly relevant for applications requiring high-quality view synthesis from wide-baseline panoramas, such as commercial VR platforms and virtual tours.
    
    \item \textbf{ZIGNeRF (Zero-shot 3D Scene Representation with Invertible Generative Neural Radiance Fields)} introduces a method for generating 3D images from single 2D images using a zero-shot approach. It employs a combination of 3D-aware GAN inversion and generative NeRF, enabling the generation of multi-view images from out-of-domain images without additional training. The technique can disentangle objects from backgrounds and perform 3D operations like rotation. This innovation shows promise for applications in virtual reality and 3D modeling.
    
    \item \textbf{Instruct-Video2Avatar (Video-to-Avatar Generation with Instructions)} by Shaoxu Li proposes a method for creating edited, photo-realistic, animatable 3D neural head avatars using text instructions and a short monocular RGB video. The method employs an image-conditioned diffusion model for editing one head image and a video stylization method for editing other head images in the sequence. This iterative process, involving three or more training cycles, synthesizes edited avatars with a deformable neural radiance field. The authors addressed the challenge of synthesizing 3D stylized avatars, which traditionally required manual modeling or user-friendly automatic creation methods. Their approach leverages implicit 3D representation for higher render quality and introduces a novel way to manipulate NeRFs for user-friendly interaction. The method combines InstructPix2Pix for image editing, EbSynth for video stylization, and INSTA for avatar optimization, resulting in a process that takes less than 20 minutes per training cycle. This technique outperformed previous state-of-the-art methods in both qualitative and quantitative studies, offering a significant advancement in the field of digital avatar creation. The method's ability to integrate human identity with desired stylization instructions, such as changing expressions or transforming into fictional characters, demonstrated its potential for applications in virtual reality, digital content creation, and personalized avatar generation.
    
    \item \textbf{BeyondPixels (A Comprehensive Review of the Evolution of Neural Radiance Fields)} by A.K.M. Shahariar Azad Rabby and Chengcui Zhang provides an extensive survey of the advancements in Neural Radiance Fields (NeRF). NeRF, a technique that synthesizes 3D objects from 2D images using AI algorithms, has revolutionized the field of neural rendering by offering a volumetric representation capable of generating color and density for every point in a 3D space. The paper categorizes the evolution of NeRF into various aspects, including quality and scalability improvements, handling of large datasets, representation of articulated objects, and scene editing capabilities. The authors discuss various methods developed to overcome NeRF's limitations, such as Mip-NeRF for anti-aliasing, Point-NeRF for point cloud representation, NeRFusion for large-scale scene reconstruction, and DRF-Cages for deforming radiance fields. These methods address issues like sampling and aliasing problems, slow training and rendering speeds, and the need for large, high-quality datasets. The paper also highlights NeRF's application in fields like audio, due to similarities in sound propagation and ray tracing. The survey stands out by focusing exclusively on NeRF-related papers, offering detailed summaries and comparisons, and proposing a new classification system based on the challenges addressed by these models. This comprehensive review serves as an up-to-date resource for researchers and practitioners interested in the rapidly evolving field of NeRF and its applications in computer vision, graphics, virtual reality, and more.
    
    \item \textbf{H2-Mapping (Real-time Dense Mapping Using Hierarchical Hybrid Representation)} by Chenxing Jiang, Hanwen Zhang, Peize Liu, Zehuan Yu, Hui Cheng, Boyu Zhou, and Shaojie Shen, introduces a novel NeRF-based mapping method for real-time, high-quality dense mapping, particularly suitable for robotics, AR/VR, and digital twins applications. The method employs a hierarchical hybrid representation, combining implicit multiresolution hash encoding with explicit octree SDF priors, to efficiently describe scenes at varying levels of detail. This representation allows for rapid scene geometry initialization and simplifies the learning of scene geometry. Additionally, the authors propose a coverage-maximizing keyframe selection strategy to address the forgetting issue in online mapping tasks, enhancing mapping quality, especially in marginal areas. The method is the first to achieve high-quality NeRF-based mapping on edge computers of handheld devices and quadrotors in real-time. Extensive experiments demonstrate superior performance over existing NeRF-based mapping methods in terms of geometry accuracy, texture realism, and time efficiency. This approach is groundbreaking for real-time applications in robotics and AR/VR, where dynamic and detailed mapping is essential.
    
    \item \textbf{BAA-NGP (Bundle-Adjusting Accelerated Neural Graphics Primitives)} by Sainan Liu, Shan Lin, Jingpei Lu, Shreya Saha, Alexey Supikov, and Michael Yip, introduces a novel framework for reconstructing 3D scenes from 2D images using implicit neural representation (INR). The method, Bundle-Adjusting Accelerated Graphics Primitives (BAA-NGP), addresses the challenge of camera pose estimation in real-world video sequences where cameras are not tracked. BAA-NGP leverages accelerated sampling and hash encoding to expedite both pose refinement/estimation and 3D scene reconstruction. The framework combines pose estimation with fast occupancy sampling and multiresolution hash encoding through a new curriculum learning strategy. BAA-NGP demonstrates a significant speed improvement in novel view synthesis compared to other bundle-adjusting neural radiance field methods, without sacrificing pose estimation quality. This approach is particularly beneficial for applications in virtual and augmented reality, robotics, and automation, where fast and accurate 3D reconstruction from unstructured image capture is crucial. The authors' work shows that BAA-NGP is either comparable or outperforms state-of-the-art techniques like BARF and COLMAP-based methods, making it a promising solution for learning INRs in real-world scenarios.
    
    \item \textbf{Variable Radiance Field for Real-Life Category-Specifc Reconstruction from Single Image} by Kun Wang, Zhiqiang Yan, Zhenyu Zhang, Xiang Li, Jun Li, and Jian Yang, presents a novel framework, Variable Radiance Field (VRF), for reconstructing category-specific objects from a single image without known camera parameters. This method overcomes the limitations of existing methods that rely on local feature retrieval and are slow and prone to distortion. VRF uses a multi-scale global feature extractor to parameterize the geometry and appearance of the object, a contrastive learning-based pretraining strategy to improve feature extraction, and hypernetworks to generate small neural networks for efficient rendering. The framework also includes a dynamic ray sampling module that learns an instance-specific 3D similarity transformation for semantic-consistent learning. VRF demonstrates superior performance in terms of quality and speed on the CO3D dataset and is applicable to shape interpolation and object placement tasks. This method offers a significant advancement in single-image 3D reconstruction, particularly for real-life images where camera parameters are unknown, making it highly relevant for applications in virtual reality, augmented reality, and e-commerce.
    
    \item \textbf{Enhance-NeRF (Multiple Performance Evaluation for Neural Radiance Fields)} by Qianqiu Tan, Tao Liu, Yinling Xie, Shuwan Yu, and Baohua Zhang, introduces a novel approach to improve Neural Radiance Fields (NeRF) for 3D reconstruction, particularly in outdoor scenes with complex lighting conditions. The Enhance-NeRF method addresses the challenges of color "fog" noise and instabilities in reconstructing unbounded scenes, common in traditional NeRF models. The authors propose a joint color approach to balance the display of objects with varying reflectivity and utilize a decoding architecture with prior knowledge for improved recognition. They also employ multi-layer performance evaluation mechanisms to enhance learning capacity. This approach allows for the efficient reconstruction of outdoor scenes within an hour on a single GPU, partially enhancing fitness capability and supporting outdoor scene reconstruction. Enhance-NeRF can be used as a plug-and-play component, easily integrating with other NeRF-based models. The method shows promise in applications such as virtual reality, augmented reality, and autonomous driving, where accurate and efficient 3D reconstruction of complex scenes is essential. The authors demonstrate that Enhance-NeRF can effectively suppress color haze interference and improve the quality of synthesized perspectives, making it a valuable contribution to the field of 3D reconstruction.
    
    \item \textbf{LU-NeRF (Scene and Pose Estimation by Synchronizing Local Unposed NeRFs)} by Zezhou Cheng, Carlos Esteves, Varun Jampani, Abhishek Kar, Subhransu Maji, and Ameesh Makadia, presents a novel approach for jointly estimating camera poses and neural radiance fields from unposed images. The method, LU-NeRF, operates in a local-to-global manner, first optimizing over local subsets of data called "mini-scenes" to estimate local pose and geometry. These mini-scenes are then synchronized into a global reference frame, followed by a final global optimization of pose and scene. This approach overcomes the limitations of existing unposed NeRF methods, which often rely on prior pose distributions or coarse pose initialization. LU-NeRF is designed to handle the unique challenges of few-shot local unposed NeRF, such as reconciling mirror-symmetric configurations. The method demonstrates superior performance compared to previous attempts at unposed NeRF, operating effectively in the general SE(3) pose setting. This makes LU-NeRF a promising solution for scenarios where accurate camera poses are unknown or hard to estimate, such as in low-texture or low-resolution settings, and can be complementary to feature-based Structure-from-Motion (SfM) pipelines. The technique has potential applications in areas like virtual reality, augmented reality, and robotics, where accurate 3D scene reconstruction from limited or unstructured data is crucial.
    
    \item \textbf{RePaint-NeRF (NeRF Editting via Semantic Masks and Diffusion Models)} by Xingchen Zhou, Ying He, F Richard Yu, Jianqiang Li, and You Li, introduces a novel framework for editing 3D content in Neural Radiance Fields (NeRF) using text prompts. This method, named RePaint-NeRF, leverages pre-trained diffusion models to guide changes in designated 3D content, significantly enhancing the editability and diversity of NeRF. The process involves two stages: first, a semantic feature module is used to select target objects for editing, and then, under the guidance of a diffusion model, the selected area is modified according to text prompts. This approach allows for practical semantic-masked object editing, enabling users to guide editing in continuous NeRF scenes. The method demonstrates effectiveness in changing content in various scenes under different text prompts, validated on both real-world and synthetic datasets. RePaint-NeRF's contributions include the capability to edit 3D content in NeRF through text prompts, expanding the application scope of NeRF models, and enabling semantic-masked object editing guided by diffusion models. This advancement opens up new possibilities for creative editing in 3D scenes, making it applicable to a wide range of scenarios where dynamic content modification is required.
    
    \item \textbf{GANeRF (Leveraging Discriminators to Optimize Neural Radiance Fields)} by Barbara Roessle, Norman Müller, Lorenzo Porzi, Samuel Rota Bulò, Peter Kontschieder, and Matthias Nießner, presents a groundbreaking approach to enhancing Neural Radiance Fields (NeRF) using Generative Adversarial Networks (GANs). The authors introduce GANeRF, a method that integrates adversarial loss into NeRF optimization, leveraging a 2D patch discriminator to guide the NeRF reconstruction process. This approach significantly improves the realism of rendered patches, particularly in under-constrained regions, by enforcing the radiance field representation to closely follow the distribution of real-world image patches. GANeRF also includes a 2D generator that refines NeRF renderings at multiple scales, further enhancing the output quality. The method demonstrates substantial improvements in rendering quality, including nearly halving LPIPS scores and increasing PSNR by 1.4dB on advanced indoor scenes compared to existing methods. GANeRF's contributions include a novel adversarial formulation for optimizing a 3D-consistent radiance field representation and a 2D generator for further refining rendering output. This approach is particularly effective for novel view synthesis in challenging, large-scale scenes, showcasing its potential in applications like virtual reality, computational photography, and 3D scene reconstruction.
    
    \item \textbf{HyP-NeRF (Learning Improved NeRF Priors using a HyperNetwork)}  by Bipasha Sen, Gaurav Singh, Aditya Agarwal, Rohith Agaram, K Madhava Krishna, and Srinath Sridhar, introduces HyP-NeRF, a novel method for learning high-quality, generalizable Neural Radiance Field (NeRF) priors using hypernetworks. This approach significantly enhances the quality of NeRFs by generating both the multi-resolution hash encodings and the weights of a NeRF model. The authors propose a two-step process: first, using a hypernetwork to predict NeRF parameters, and second, employing a denoising network to improve the quality of rendered views from the estimated NeRF. This denoising and finetuning strategy not only enhances the visual quality but also retains multiview consistency. HyP-NeRF demonstrates its versatility in various applications, including NeRF reconstruction from single-view or cluttered scenes and text-to-NeRF conversions. The method outperforms existing techniques in terms of generalization, compression, and retrieval, showcasing its potential in rendering high-quality, consistent NeRFs for diverse objects and scenes. This advancement holds significant promise for applications in 3D modeling, virtual reality, and augmented reality, where accurate and detailed 3D representations are essential.
    
    \item \textbf{NeRFool (Uncovering the Vulnerability of Generalizable Neural Radiance Fields against Adversarial Perturbations)} by Yonggan Fu, Ye Yuan, Souvik Kundu, Shang Wu, Shunyao Zhang, and Yingyan (Celine) Lin, explores the adversarial robustness of Generalizable Neural Radiance Fields (GNeRF). The authors introduce NeRFool and NeRFool+, the first frameworks to systematically analyze and exploit the vulnerabilities of GNeRF to adversarial perturbations. NeRFool reveals that increased conditionality on source views can heighten GNeRF's vulnerability, and that perturbations on density have a stronger destructive ability than those on color, particularly in scenes with complex geometry. Building on these insights, NeRFool+ integrates novel target view sampling and geometric error maximization techniques, effectively attacking GNeRF across a wide range of target views. The study also examines the impact of per-scene finetuning on robustness and the transferability of perturbations across different views. This research is pivotal for understanding the security implications of deploying GNeRF in real-world applications, such as autonomous driving and robot navigation, where robustness against adversarial attacks is crucial. The findings from NeRFool and NeRFool+ lay the groundwork for future innovations in developing robust, real-world GNeRF solutions.
    
    \item \textbf{From NeRFLiX to NeRFLiX++: A General NeRF-Agnostic Restorer Paradigm} by Kun Zhou, Wenbo Li, Nianjuan Jiang, Xiaoguang Han, and Jiangbo Lu, presents an innovative approach to enhancing Neural Radiance Fields (NeRF) rendered views. The authors introduce NeRFLiX, a NeRF-agnostic restorer paradigm, which employs a degradation-driven inter-viewpoint mixer to improve the quality of NeRF-rendered images. NeRFLiX utilizes a NeRF-style degradation modeling approach and constructs large-scale training data, enabling effective removal of NeRF-native rendering artifacts. The authors further develop NeRFLiX++ with a two-stage NeRF degradation simulator and a faster inter-viewpoint mixer, achieving superior performance with improved computational efficiency. NeRFLiX++ is capable of restoring ultra-high-resolution outputs from noisy, low-resolution NeRF-rendered views, demonstrating remarkable restoration ability on various novel view synthesis benchmarks. This paradigm significantly advances the field of photorealistic 3D rendering, offering a solution to the persistent challenge of rendering artifacts in NeRF models. The applications of this technique are vast, potentially benefiting industries such as virtual reality, gaming, and film production, where high-quality 3D rendering is crucial.
    
    \item \textbf{ATT3D (Amortized Text-to-3D Object Synthesis)} by Jonathan Lorraine, Kevin Xie, Xiaohui Zeng, Chen-Hsuan Lin, Towaki Takikawa, Nicholas Sharp, Tsung-Yi Lin, Ming-Yu Liu, Sanja Fidler, and James Lucas introduces a novel approach to synthesizing 3D objects from text prompts. This method, known as Amortized Text-to-3D (ATT3D), significantly accelerates the process by training a unified model on multiple prompts simultaneously, sharing computation across a prompt set. Unlike existing methods that require lengthy, per-prompt optimization, ATT3D enables the generation of accurate 3D objects in less than a second using a single GPU. The framework leverages Neural Radiance Fields (NeRFs) and text-to-image generative models, allowing for the creation of high-quality 3D models from text descriptions. ATT3D not only reduces overall training time but also generalizes to new prompts, interpolates between prompts, and amortizes over settings other than text prompts. This method offers a cost-effective solution for 3D content creation, making it more accessible and efficient, particularly beneficial for industries like entertainment, education, and marketing. The ability to interpolate between prompts opens up possibilities for novel asset generation and simple animations, enhancing the creative process in 3D modeling.
    
    \item \textbf{Parametric Implicit Face Representation for Audio-Driven Facial Reenactment} by Ricong Huang, Peiwen Lai, Yipeng Qin, and Guanbin Li introduces a groundbreaking audio-driven facial reenactment framework that effectively balances interpretability and expressive power. This work addresses the limitations of existing explicit and implicit methods in facial reenactment by proposing a novel parametric implicit representation (PIR). PIR combines the interpretability of 3D Morphable Models (3DMM) with the expressive power of implicit representations, enabling controllable and high-quality synthesis of talking heads. The framework consists of three components: contextual audio to expression encoding using a transformer-based architecture, implicit representation parameterization through conditional image synthesis, and rendering with PIR formulated as a conditional image inpainting problem. A unique aspect of this approach is the use of a tri-plane based generator for efficient learning of the implicit representation. The authors also introduce a data augmentation technique to enhance model generalizability, addressing the challenge of lip movement synchronization with unseen input audio. Extensive experiments demonstrate that this method outperforms state-of-the-art techniques, producing more realistic results with greater fidelity to the identities and talking styles of speakers. This research has significant implications for applications in filmmaking, virtual avatars, and video conferencing, where accurate and controllable facial reenactment is essential.
    
    \item \textbf{NeuS-PIR (Learning Relightable Neural Surface using Pre-Integrated Rendering)} by Shi Mao, Chenming Wu, Zhelun Shen, and Liangjun Zhang introduces a novel method for learning relightable neural implicit surfaces using pre-integrated rendering. This approach, built upon the high-quality neural implicit method NeuS, interprets the scene as a Signed Distance Function (SDF) over the learning space and enhances it by decomposing material and illumination properties from the radiance field. The method, NeuS-PIR, factorizes the radiance field into a spatially varying material field and a differentiable environment cubemap, jointly learning it with geometry represented by a neural surface. This collaborative optimization of geometry, material, and illumination leads to consistent improvements and high-quality relighting in novel conditions. The authors demonstrate that NeuS-PIR outperforms state-of-the-art methods in both synthetic and real datasets. The key contributions include a new method for factorizing object geometry, material, and illumination, a jointly optimizing scheme that allows these aspects to benefit each other during training, and a regularization method for material encoding that encourages sparse and consistent material factorization. This work has potential applications in AR/VR, autonomous driving simulation, and films, where accurate 3D reconstruction with physical properties is crucial.
    
    \item \textbf{Neural Relighting with Subsurface Scattering by Learning the Radiance Transfer Gradient} by Shizhan Zhu et al. presents a novel framework for relighting objects with subsurface scattering effects, a challenge that has been difficult for existing neural rendering methods. The authors' approach combines volume rendering and neural radiance transfer fields to optimize both shape and radiance transfer, extending relighting capabilities to a wider range of materials, including those with strong subsurface scattering effects. This method is fully data-driven and does not rely on specific material representations like BRDF or BSSRDF, enabling it to render high-quality appearances under varying lighting conditions and viewpoints. The framework is evaluated using a novel light stage dataset of objects with subsurface scattering effects, showing significant improvements over current state-of-the-art methods. The authors' contributions include a new approach to learning radiance transfer fields using volume rendering, optimizing appearance and geometry end-to-end, and the creation of a high-quality dataset for training and evaluation. This work has potential applications in virtual reality, gaming, visual effects, and architecture, where accurate simulation of lighting conditions is crucial.
    
    \item \textbf{DreamHuman (Animatable 3D Avatars from Text)} by Nikos Kolotouros et al. introduces a method for generating realistic, animatable 3D human avatar models solely from textual descriptions. The method, DreamHuman, leverages advancements in text-to-image synthesis models, neural radiance fields, and statistical human body models within a novel modeling and optimization framework. This approach enables the generation of dynamic 3D human avatars with high-quality textures and learned, instance-specific surface deformations. DreamHuman addresses the limitations of previous text-to-3D methods by incorporating 3D human body priors, which are essential for regularizing the generation and re-posing of the resulting avatar. The authors demonstrate that their method can generate a wide variety of animatable, realistic 3D human models from text, with diverse appearances, clothing, skin tones, and body shapes. These models significantly outperform both generic text-to-3D approaches and previous text-based 3D avatar generators in visual fidelity. DreamHuman has potential applications in industries such as gaming, special effects, film, and content creation, where it could automate complex parts of the design process.
    
    \item \textbf{Edit-DiffNeRF (Editing 3D Neural Radiance Fields using 2D Diffusion Model)} by Lu Yu et al. presents a novel framework, Edit-DiffNeRF, for editing 3D Neural Radiance Fields (NeRFs) using 2D diffusion models. This approach addresses the challenge of cross-view inconsistency and degradation in stylized view syntheses when simply coupling NeRF with diffusion models. The Edit-DiffNeRF framework comprises a frozen diffusion model, a delta module for editing the latent semantic space of the diffusion model, and a NeRF. The key innovation lies in editing the latent semantic space in frozen pretrained diffusion models, enabling fine-grained modifications to rendered views and effectively consolidating these instructions in a 3D scene via NeRF training. The authors introduce a multi-view semantic consistency loss to ensure semantic consistency across different viewpoints. This method demonstrates a 25\% improvement in aligning 3D edits with text instructions compared to prior work, effectively editing real-world 3D scenes. The Edit-DiffNeRF framework is particularly significant for applications requiring high-fidelity text-to-3D editing with pretrained NeRFs, offering enhanced control and accuracy in the optimization process of underlying scenes.
    
    \item \textbf{MA-NeRF (Motion-Assisted Neural Radiance Fields for Face Synthesis from Sparse Images)} by Weichen Zhang et al. introduces a novel framework for photorealistic 3D face avatar synthesis from sparse images. The authors address the limitations of existing parametric models and NeRF-based avatar methods by proposing MA-NeRF, which leverages parametric 3DMM models to reconstruct high-fidelity drivable face avatars capable of handling unseen expressions. The core of MA-NeRF includes a structured displacement feature and a semantic-aware learning module. The structured displacement feature introduces motion prior as an additional constraint, constructing a displacement volume to improve performance for unseen expressions. The semantic-aware learning incorporates multi-level priors, such as semantic embedding and learnable latent codes, to enhance the model's capabilities. Extensive experiments on public datasets demonstrate significant improvements over previous state-of-the-art methods, with MA-NeRF achieving impressive performance under both full and sparse input settings. This method holds potential for applications in VR/AR and the metaverse, offering a solution for creating detailed and expressive 3D face avatars from limited data.
    
    \item \textbf{Instruct-NeuralTalker (Editing Audio-Driven Talking Radiance Fields with Instructions)} by Yuqi Sun et al. presents a novel interactive framework for editing talking radiance fields (TRFs) using human instructions to achieve personalized talking face generation. This method, a first of its kind, allows for the editing of dynamic TRFs with text instructions, enabling highly personalized talking face generation. The authors build an efficient talking radiance field from a short speech video and apply a conditional diffusion model for image editing with human instructions during optimization. A key feature of their approach is the progressive dataset updating strategy, which ensures audio-lip synchronization while preventing distortion of lip shapes during editing. Additionally, they introduce a lightweight refinement network for enhancing image details and achieving controllable detail generation in the final rendered image. This method is significant for applications in digital humans, VR/AR, 3D telepresence, and virtual video conferencing, as it offers real-time rendering up to 30FPS on consumer hardware and significantly improves rendering quality compared to state-of-the-art methods. The authors demonstrate that their approach can generate high-quality talking faces that meet editing targets while maintaining audio-lip synchronization, showcasing the potential of instruction-based control in NeRF editing for dynamic scenarios.
    
    \item \textbf{NeRF synthesis with shading guidance} by Chenbin Li et al. introduces a novel approach to constructing large radiance fields from smaller NeRF (Neural Radiance Field) exemplars, focusing on maintaining continuity in geometry and appearance while controlling lighting effects. The authors propose a two-phase method for NeRF synthesis, which includes geometry-based synthesis and appearance-based synthesis, ensuring both geometric and appearance continuity in the synthesized results. They also introduce a boundary-constrained synthesis method to prevent artifacts at the boundaries when synthesizing large scenes. This method is particularly effective in handling complex lighting scenarios, as it uses a shading map and shading map guider to control the lighting of the synthesized scene, rather than decoupling the scene. The technique is applicable to various real-world scenarios, such as virtual reality, augmented reality, and video games, where accurate representation of complex real-world scenes is crucial. The authors demonstrate that their method can generate high-quality results with consistent geometry and appearance, even in scenes with complex lighting, enhancing the practicality of NeRF synthesis in real-world applications.
    
    \item \textbf{DreamTime (An Improved Optimization Strategy for Text-to-3D Content Creation)} by Yukun Huang, Jianan Wang, Yukai Shi, Xianbiao Qi, Zheng-Jun Zha, and Lei Zhang, presents a novel approach to text-driven 3D content generation using Neural Radiance Fields (NeRF) optimized with score distillation. The authors identify limitations in existing text-to-3D methods, such as quality concerns and low diversity, and attribute these to the conflict between NeRF optimization and uniform timestep sampling in score distillation. To address this, they propose a time-prioritized score distillation sampling (TP-SDS) strategy, which aligns NeRF optimization with the sampling process of diffusion models. TP-SDS prioritizes timestep sampling with monotonically non-increasing functions, focusing on global structure at the beginning of optimization and gradually shifting to finer details. The method is validated through extensive experiments, demonstrating significant improvements in both quality and diversity of text-to-3D generation. This work contributes to the field by revealing the conflict in existing methods, introducing DreamTime as an improved optimization strategy, and conducting comprehensive experiments to showcase the effectiveness of TP-SDS. The approach has potential applications in areas requiring high-fidelity and diverse 3D models generated from textual descriptions, such as virtual reality, gaming, and digital content creation.
    
    \item \textbf{CARVER (Benchmarking and Analyzing 3D-aware Image Synthesis with a Modularized Codebase)} by Qiuyu Wang, Zifan Shi, Kecheng Zheng, Yinghao Xu, Sida Peng, Yujun Shen, and their team, presents a comprehensive framework for 3D-aware image synthesis. The study introduces Carver, a modularized codebase that reformulates the generation process into distinct modules, including pose sampler, stochasticity mapper, point sampler, point embedder, feature decoder, volume renderer, and upsampler. This design enables independent development and replacement of each module, facilitating fair comparison and recognition of contributions from various approaches. The authors re-implement a range of classic algorithms using Carver, demonstrating its effectiveness and versatility. In-depth analyses are conducted on different modules, such as the comparison of point features, the necessity of the tailing upsampler in the generator, and the reliance on camera pose priors. These analyses deepen understanding of existing methods and suggest future research directions. The study also benchmarks 3D-aware image synthesis on datasets like FFHQ, ShapeNet Cars, and Cats, employing metrics like FID, reprojection error, and identity consistency. The results indicate that Carver can precisely replicate official results, confirming its reliability and utility in the field. This work significantly contributes to the advancement of 3D-aware image synthesis, offering a structured platform for exploring and integrating diverse methods and techniques.
    
    \item \textbf{Neural Spectro-polarimetric Fields} by Youngchan Kim, Wonjoon Jin, Sunghyun Cho, and Seung-Hwan Baek, introduces Neural Spectro-polarimetric Fields (NeSpoF), a neural representation designed to model physically-valid spectro-polarimetric fields. This approach captures the Stokes vector of a light ray at any given position, direction, and wavelength, addressing the limitations of tensor representations in handling high-dimensional light properties. NeSpoF is particularly efficient in managing low signal-to-noise ratio (SNR) measurements and showcases a compact memory footprint. The authors developed a hyperspectral-polarimetric imaging system and released the first multi-view hyperspectral-polarimetric image dataset, comprising both synthetic and real-world scenes. The main contributions include capturing, modeling, and rendering spectro-polarimetric fields, developing the NeSpoF representation, proposing a spatially-varying spectro-polarimetric calibration method, and releasing the first spectro-polarimetric multi-view dataset. This work demonstrates the synthesis capability of NeSpoF, taking into account view, spectrum, and polarization, and has potential applications in various fields requiring multi-dimensional visual analysis.
    
    \item \textbf{Local 3D Editing via 3D Distillation of CLIP Knowledge} by Junha Hyung, Sungwon Hwang, Daejin Kim, Hyunji Lee, and Jaegul Choo, introduces Local Editing NeRF (LENeRF), a framework for fine-grained and localized manipulation of 3D content using text inputs. LENErf leverages a pretrained language-image model to steer synthesis towards user-provided text prompts, along with a 3D MLP model initialized on an existing NeRF scene. The framework allows local editing by localizing a 3D ROI box in the input scene and blends the content synthesized inside the ROI with the existing scene using a novel volumetric blending technique. To achieve natural-looking and view-consistent results, the authors use existing and new geometric priors and 3D augmentations. LENErf has been tested both qualitatively and quantitatively on a variety of real 3D scenes and text prompts, demonstrating realistic, multi-view consistent results with flexibility and diversity compared to baselines. The framework is applicable for various 3D editing applications, including adding new objects to a scene, removing/replacing/altering existing objects, and texture conversion. This advancement in NeRF technology offers significant potential for applications in virtual reality, augmented reality, and digital content creation, where editing and blending objects seamlessly into 3D scenes are crucial. LENErf's ability to edit and integrate objects into existing scenes based on textual descriptions marks a significant step forward in the field of 3D scene manipulation and visualization.
    
    \item \textbf{Blended-NeRF (Zero-Shot Object Generation and Blending in Existing Neural Radiance Fields)} by Ori Gordon, Omri Avrahami, and Dani Lischinski, introduces Blended-NeRF, a robust and flexible framework for editing specific regions of interest in an existing Neural Radiance Field (NeRF) scene. This method, guided by text prompts and a 3D Region of Interest (ROI) box, leverages a pretrained language-image model to steer the synthesis towards the user-provided text prompt. It employs a 3D MLP model, initialized on an existing NeRF scene, to generate the object and blend it into the specified region in the original scene. The framework enables local editing by localizing a 3D ROI box in the input scene and blends the content synthesized inside the ROI with the existing scene using a novel volumetric blending technique. To achieve natural-looking and view-consistent results, the authors use existing and new geometric priors and 3D augmentations. Blended-NeRF has been tested both qualitatively and quantitatively on a variety of real 3D scenes and text prompts, demonstrating realistic, multi-view consistent results with flexibility and diversity compared to baselines. The framework is applicable for various 3D editing applications, including adding new objects to a scene, removing/replacing/altering existing objects, and texture conversion. This advancement in NeRF technology offers significant potential for applications in virtual reality, augmented reality, and digital content creation, where editing and blending objects seamlessly into 3D scenes are crucial. Blended-NeRF's ability to edit and integrate objects into existing scenes based on textual descriptions marks a significant step forward in the field of 3D scene manipulation and visualization.
    
    \item \textbf{TaiChi Action Capture and Performance Analysis with Multi-view RGB Cameras} by Jianwei Li, Siyu Mo, and Yanfei Shen, presents a novel framework for capturing and analyzing TaiChi performance using multi-view geometry and artificial intelligence technology. This study addresses the gap in vision-based motion capture and intelligent analysis for professional TaiChi movements. The authors developed a multi-camera system specifically for TaiChi motion capture and collected multi-view TaiChi data. The framework combines traditional visual methods with an implicit neural radiance field to achieve sparse 3D skeleton fusion and dense 3D surface reconstruction. Additionally, the team carried out normalization modeling of movement sequences based on motion transfer, enabling TaiChi performance analysis for different groups. Evaluation experiments demonstrated the efficiency of their method, showing significant advancements in the field of intelligent sports performance analysis. This research contributes to the understanding and improvement of TaiChi techniques and training, offering a sophisticated tool for coaches and athletes. The application of this technology in sports science represents a significant step forward in the non-invasive and detailed analysis of complex human movements, particularly in disciplines like TaiChi where precision and fluidity are essential.
    
    \item \textbf{Toward a Spectral Foundation Model: An Attention-Based Approach with Domain-Inspired Fine-Tuning and Wavelength Parameterization} by Tomasz Różański, Yuan-Sen Ting, Maja Jabłońska, and their colleagues, introduces an innovative approach to enhance spectral fitting techniques in astrophysical explorations. The study proposes three major enhancements to transcend the limitations of current spectral emulation models. Firstly, an attention-based emulator is implemented, capable of unveiling long-range information between wavelength pixels. Secondly, a domain-specific fine-tuning strategy is employed, where the model is pre-trained on spectra with fixed stellar parameters and variable elemental abundances, followed by fine-tuning on the entire domain. Lastly, by treating wavelength as an autonomous model parameter, akin to neural radiance fields, the model can generate spectra on any wavelength grid. This approach significantly outperforms current leading methods in spectral emulation accuracy and sample efficiency. The proposed solution is particularly relevant for large-scale stellar spectroscopy surveys, where the burgeoning volume of high-quality spectra necessitates a paradigm shift in analytic methods. The attention-based emulator, combined with domain-inspired fine-tuning and wavelength parameterization, offers a more accurate and efficient way of modeling stellar spectra, crucial for understanding celestial bodies' properties. This advancement in astrophysical research tools has the potential to significantly impact fields such as stellar physics, galactic archaeology, and exoplanet studies, where precise spectral analysis is key.
    
    \item \textbf{Envisioning a Next Generation Extended Reality Conferencing System with Efficient Photorealistic Human Rendering} by Chuanyue Shen, Letian Zhang, Zhangsihao Yang, Masood Mortazavi, Xiyun Song, Liang Peng, Heather Yu, and their team, proposes a pipeline for an extended reality metaverse conferencing system that leverages monocular video acquisition and free-viewpoint synthesis. This system aims to create an immersive online meeting experience by efficiently rendering photorealistic human dynamics. The core challenge addressed is the slow rendering speed of Neural Radiance Fields (NeRF), which hinders real-time conferencing. The authors explore an accelerated NeRF-based free-viewpoint synthesis algorithm to render human dynamics more efficiently. Their algorithm achieves comparable rendering quality while performing training and inference significantly faster than state-of-the-art methods. This exploration lays the groundwork for constructing metaverse conferencing systems capable of handling dynamic scene relighting with customized themes and multi-user conferencing that integrates real-world people into an extended world. The envisioned system has the potential to revolutionize industries such as business, education, and entertainment, offering more efficient meetings, facilitating remote learning, and providing diverse and engaging environments. This work represents a significant step towards realizing immersive and photorealistic conferencing experiences in the metaverse, enhancing the realism and interaction between participants in virtual meetings.
    
    \item \textbf{One-2-3-45 (Any Single Image to 3D Mesh in 45 Seconds without Per-Shape Optimization)} by Minghua Liu, Chao Xu, Haian Jin, Linghao Chen, Mukund Varma T, Zexiang Xu, and Hao Su, presents a groundbreaking method for converting any single image into a full 360-degree 3D textured mesh in a single feed-forward pass. This approach addresses the challenges of single image 3D reconstruction, such as lengthy optimization time, 3D inconsistency, and poor geometry. The method begins with a view-conditioned 2D diffusion model, Zero123, to generate multi-view images from the input view, and then lifts them to 3D space. Traditional reconstruction methods often struggle with inconsistent multi-view predictions, so the authors build their 3D reconstruction module upon an SDF-based generalizable neural surface reconstruction method, incorporating several critical training strategies to enable the reconstruction of 360-degree meshes. This method significantly reduces reconstruction time compared to existing methods, produces better geometry, more 3D consistent results, and adheres more closely to the input image. Evaluated on both synthetic data and in-the-wild images, the approach demonstrates superiority in mesh quality and runtime. Additionally, it can seamlessly support text-to-3D tasks by integrating with off-the-shelf text-to-image diffusion models. This advancement has broad applications in robotic object manipulation, navigation, 3D content creation, and AR/VR, offering a fast and effective solution for generating high-quality 3D models from single images. The ability to create detailed and accurate 3D meshes from a single image in a short time frame marks a significant step forward in the field of 3D reconstruction and computer vision.
    
    \item \textbf{FlipNeRF (Flipped Reflection Rays for Few-shot Novel View Synthesis)} by Seunghyeon Seo, Yeonjin Chang, and Nojun Kwak, introduces a novel regularization method for few-shot novel view synthesis in Neural Radiance Fields (NeRF). FlipNeRF utilizes flipped reflection rays, derived from input ray directions and estimated normal vectors, to effectively train with additional rays and accurately estimate surface normals and 3D geometry. This approach addresses the challenge of requiring a dense set of multi-view images, a significant bottleneck in practical applications of NeRF. FlipNeRF also incorporates an Uncertainty-aware Emptiness Loss and Bottleneck Feature Consistency Loss, enhancing the reliability of outputs and reducing floating artifacts across different scene structures. These innovations enable FlipNeRF to achieve state-of-the-art performance on multiple benchmarks, significantly improving rendering quality compared to other methods, especially in sparse view scenarios. The accurate surface normal estimation leads to more precise depth estimation, which is crucial for few-shot novel view synthesis. FlipNeRF's advancements in NeRF technology have broad implications for applications in virtual reality, augmented reality, and 3D content creation, where generating photorealistic images from sparse inputs is essential. This method offers a more efficient and effective solution for synthesizing high-quality images from novel viewpoints, marking a significant step forward in the field of 3D rendering and visualization.
    
    \item \textbf{Magic123 (One Image to High-Quality 3D Object Generation Using Both 2D and 3D Diffusion Priors)} by Guocheng Qian, Jinjie Mai, Abdullah Hamdi, Jian Ren, Aliaksandr Siarohin, Bing Li, Hsin-Ying Lee, Ivan Skorokhodov, Peter Wonka, Sergey Tulyakov, and Bernard Ghanem, introduces a novel two-stage coarse-to-fine approach for generating high-quality, textured 3D meshes from a single unposed image. The first stage involves optimizing a neural radiance field to produce coarse geometry, while the second stage adopts a memory-efficient differentiable mesh representation to yield a high-resolution mesh with visually appealing texture. The 3D content is learned through reference view supervision and novel views, guided by a combination of 2D and 3D diffusion priors. A single trade-off parameter is introduced to control the balance between exploration and exploitation in the generated geometry. The method also employs textual inversion and monocular depth regularization to ensure consistent appearances across views and prevent degenerate solutions. Magic123 significantly improves over previous image-to-3D techniques, as demonstrated through extensive experiments on synthetic benchmarks and diverse real-world images. This approach has broad applications in virtual reality, augmented reality, and digital content creation, where converting single images into detailed 3D models is crucial. Magic123's ability to create high-fidelity 3D content from unposed images in the wild marks a substantial advancement in the field of 3D reconstruction and rendering.
    
    \item \textbf{RGB-D Mapping and Tracking in a Plenoxel Radiance Field} by Andreas L. Teigen, Yeonsoo Park, Annette Stahl, and Rudolf Mester, presents a significant advancement in the field of 3D reconstruction and tracking using Neural Radiance Fields (NeRFs) and RGB-D data. The authors highlight the limitations of NeRFs when used with RGB sensors due to inherent ambiguities in visual data, which often result in inaccurate 3D models despite producing visually convincing images. To address this, they emphasize the importance of depth sensors for modeling accurate geometry in outward-facing scenes. The paper introduces an extension to the Plenoxel radiance field model, incorporating an analytical differential approach for dense mapping and tracking based on RGB-D data, without relying on a neural network. This method achieves state-of-the-art results in both mapping and tracking tasks, offering faster performance compared to competing neural network-based approaches. The proposed approach has significant implications for practical applications in robotics and extended reality (XR), where accurate and efficient dense 3D reconstruction is crucial. By leveraging RGB-D data, the authors demonstrate the potential of radiance fields for creating more reliable and detailed 3D maps, enhancing tasks such as path planning, collision avoidance, and interaction between real-world geometry and digital objects. This work marks a notable contribution to the field, particularly in improving the utility of NeRF models for real-world applications requiring precise 3D information.
    
    \item \textbf{NOFA (NeRF-based One-shot Facial Avatar Reconstruction)} by Wangbo Yu, Yanbo Fan, Yong Zhang, Xuan Wang, Fei Yin, Yunpeng Bai, Yan-Pei Cao, Ying Shan, Yang Wu, Zhongqian Sun, and Baoyuan Wu, introduces a groundbreaking one-shot 3D facial avatar reconstruction framework that requires only a single source image. This work addresses the limitations of existing NeRF-based facial avatars, which typically need multi-shot images for training and lack generalization to new identities. The authors leverage the generative prior of 3D GANs and develop an efficient encoder-decoder network to reconstruct the canonical neural volume of the source image. Additionally, a compensation network is proposed to complement facial details. For fine-grained control over facial dynamics, a deformation field is introduced to warp the canonical volume into driven expressions. The NOFA framework demonstrates superior synthesis results compared to several state-of-the-art methods through extensive experimental comparisons. This advancement in facial avatar reconstruction has significant implications for applications in virtual reality, augmented reality, the movie industry, and teleconferencing, where high-fidelity face reconstruction and fine-grained face reenactment are crucial. NOFA's ability to create realistic and dynamically controllable facial avatars from a single image marks a significant step forward in the fields of computer graphics and computer vision.
    
    \item \textbf{SAR-NeRF (Neural Radiance Fields for Synthetic Aperture Radar Multi-View Representation)} by Zhengxin Lei, Feng Xu, Jiangtao Wei, Feng Cai, Feng Wang, and Ya-Qiu Jin, introduces a novel NeRF model tailored for Synthetic Aperture Radar (SAR) image generation. SAR images, known for their sensitivity to observation configurations and significant variations across different viewing angles, pose challenges for deep learning methods in terms of generalization. The SAR-NeRF model addresses these challenges by combining SAR imaging mechanisms with neural networks. It models a set of SAR images implicitly as a function of attenuation coefficients and scattering intensities in the 3D imaging space through a differentiable rendering equation. The model learns the distribution of these coefficients and intensities of voxels, using a vectorized form of the 3D voxel SAR rendering equation and the sampling relationship between 3D space voxels and 2D view ray grids. Quantitative experiments on various datasets demonstrate SAR-NeRF's superior multi-view representation and generalization capabilities. Additionally, SAR-NeRF augmented datasets significantly improve SAR target classification performance under few-shot learning setups. This advancement in SAR image processing and representation is particularly impactful for applications in earth remote sensing, where accurate and diverse image interpretation is crucial. SAR-NeRF's ability to effectively represent and learn anisotropic features of SAR images enhances the performance of deep learning methods in SAR image classification and target recognition, offering a promising direction for overcoming the challenges posed by SAR imagery's complexity and sensitivity to observation conditions.
    
    \item \textbf{Neural Free-Viewpoint Relighting for Glossy Indirect Illumination} by Nithin Raghavan, Yan Xiao, Kai-En Lin, Tiancheng Sun, Sai Bi, Zexiang Xu, Tzu-Mao Li, and Ravi Ramamoorthi introduces a hybrid neural-wavelet Precomputed Radiance Transfer (PRT) solution for high-frequency indirect illumination, including glossy reflection, for relighting with changing view. This method represents the light transport function in the Haar wavelet basis and learns the wavelet transport using a small multi-layer perceptron (MLP) applied to a feature field as a function of spatial location and wavelet index. The MLP inputs include reflected direction and material parameters. The feature field, compactly represented by a tensor decomposition, and MLP parameters are optimized/learned from multiple images of the scene under different lighting and viewing conditions. This approach enables real-time precomputed rendering of challenging scenes involving view-dependent reflections and even caustics. The authors' work overcomes the limitations of conventional PRT methods, which are usually restricted to low-frequency spherical harmonic lighting, by leveraging modern MLP-based function approximators and recent rendering advances. This technique has significant applications in interactive rendering of scenes with complex global illumination effects, such as in virtual reality, digital entertainment, and computer graphics, where realistic and dynamic lighting plays a crucial role in enhancing the user experience and visual realism.
    
    \item \textbf{CeRF (Convolutional Neural Radiance Fields for New View Synthesis with Derivatives of Ray Modeling)} by Xiaoyan Yang, Dingbo Lu, Yang Li, Chenhui Li, and Changbo Wang, presents a novel approach to novel view synthesis by introducing Convolutional Neural Radiance Fields (CeRF). This method addresses the limitations of conventional multi-layer perceptron-based scene embedding and the geometric blurring in light field models. CeRF models the derivatives of radiance along rays using 1D convolutional operations, effectively extracting potential ray representations through a structured neural network architecture. Additionally, a recurrent module is employed to resolve geometric ambiguity in the fully neural rendering process. The authors' approach overcomes the multi-solution ambiguity inherent in volume rendering strategies and light field-based methods, where different density distributions or geometric configurations can yield the same integral color. CeRF demonstrates promising results compared to existing state-of-the-art methods, offering a significant advancement in the field of novel view synthesis. This technique has broad applications in areas such as virtual reality, augmented reality, and computer graphics, where high-quality image generation from different camera poses is essential. CeRF's ability to provide accurate and high-fidelity images from novel viewpoints makes it a valuable tool for creating immersive and realistic virtual environments.
    
    \item \textbf{Improving NeRF with Height Data for Utilization of GIS Data} by Hinata Aoki and Takao Yamanaka proposes a method to enhance Neural Radiance Fields (NeRF) for large-scale scene reconstruction by effectively utilizing height data obtained from Geographic Information Systems (GIS). This approach divides the scene space into multiple objects and a background based on height data, representing them with separate neural networks. Additionally, an adaptive sampling method is introduced, which uses height data to set sampling points densely at the borders of objects. This method significantly improves the accuracy of image rendering and speeds up the training process. The proposed model was tested with scenes containing multiple objects with height data, demonstrating better qualitative and quantitative results than previous methods. This paper's contributions include a novel volume rendering model for scenes with height data, effective use of height data to divide scene space and set sampling points, and improved rendered image quality. The technique has potential applications in virtual reality, augmented reality, and urban planning, where accurate and efficient rendering of large-scale outdoor scenes is crucial. The integration of GIS data into the NeRF framework offers a promising direction for enhancing 3D scene reconstruction and visualization.
    
    \item \textbf{Cross-Ray Neural Radiance Fields for Novel-view Synthesis from Unconstrained Image Collections} by Yifan Yang, Shuhai Zhang, Zixiong Huang, Yubing Zhang, and Mingkui Tan, proposes Cross-Ray NeRF (CR-NeRF), a novel approach to address the challenges of novel-view synthesis from unconstrained image collections. Traditional NeRF methods struggle with dynamic changes in appearance due to different capturing times and camera settings, as well as occlusions from transient objects like humans and cars. CR-NeRF overcomes these issues by leveraging interactive information across multiple rays, mimicking human perception which utilizes global information across pixels. The method introduces a cross-ray feature to represent multiple rays and recovers appearance by fusing global statistics, such as feature covariance of the rays and image appearance. Additionally, a transient objects handler and a grid sampling strategy are employed to mask out transient objects, avoiding occlusion artifacts. CR-NeRF's ability to capture more global information by leveraging correlation across multiple rays is theoretically demonstrated and empirically validated through extensive experiments on large real-world datasets. This advancement in NeRF technology is particularly beneficial for applications in virtual reality and digital humans, offering a robust solution for synthesizing occlusion-free novel views with consistent appearances from diverse and dynamic image collections.
    
    \item \textbf{PixelHuman (Animatable Neural Radiance Fields from Few Images)} by Gyumin Shim, Jaeseong Lee, Junha Hyung, and Jaegul Choo, introduces a novel human rendering model capable of generating animatable human scenes from just a few images of a person with unseen identity, views, and poses. Unlike previous methods that require a large number of images and extensive training time, PixelHuman can generalize to any input image for animatable human synthesis. The method synthesizes each target scene using a neural radiance field conditioned on a canonical representation and pose-aware pixel-aligned features, obtained through deformation fields learned in a data-driven manner. The key innovation is the introduction of a weight field table, which computes unique blend weight fields tailored to each human body shape. PixelHuman demonstrates state-of-the-art performance in multiview and novel pose synthesis from few-shot images. This advancement in neural radiance fields and human rendering has significant implications for applications in virtual reality, metaverse content creation, and computer graphics, where creating realistic, animatable human avatars from limited data is essential. PixelHuman offers an efficient and effective solution for generating high-quality 3D human avatars, enhancing the capabilities for personalized and interactive digital human representations.
    
    \item \textbf{OPHAvatars (One-shot Photo-realistic Head Avatars)} by Shaoxu Li introduces a method for synthesizing photo-realistic digital avatars from a single portrait reference. This approach first synthesizes a coarse talking head video using driving keypoints features from the portrait. Then, it creates a coarse talking head avatar using a deforming neural radiance field. The rendered images of this coarse avatar are updated with a blind face restoration model to enhance image quality. After several iterations, the method produces a photo-realistic, animatable 3D neural head avatar. The key innovation lies in using a deformable neural radiance field to eliminate unnatural distortion typically caused by image-to-video methods. This method outperforms state-of-the-art techniques in both quantitative and qualitative studies across various subjects. OPHAvatars represents a significant advancement in avatar generation, offering a practical and efficient solution for creating high-quality, animatable avatars from a single image. This technique has broad application prospects in fields such as virtual reality, digital entertainment, and telepresence, where realistic and personalized digital representations are essential.
    
    \item \textbf{Efficient Region-Aware Neural Radiance Fields for High-Fidelity Talking Portrait Synthesis} by Jiahe Li, Jiawei Zhang, Xiao Bai, Jun Zhou, and Lin Gu, presents ER-NeRF, a novel conditional Neural Radiance Fields (NeRF) based architecture for talking portrait synthesis. ER-NeRF achieves fast convergence, real-time rendering, and state-of-the-art performance with a small model size by explicitly exploiting the unequal contribution of spatial regions in talking portrait modeling. The approach introduces a compact and expressive NeRF-based Tri-Plane Hash Representation, which prunes empty spatial regions using three planar hash encoders to improve the accuracy of dynamic head reconstruction. For speech audio, a Region Attention Module generates region-aware condition features via an attention mechanism, establishing an explicit connection between audio features and spatial regions to capture local motion priors. Additionally, an Adaptive Pose Encoding is introduced to optimize the head-torso separation problem by mapping complex head pose transformations into spatial coordinates. Extensive experiments demonstrate that ER-NeRF renders high-fidelity, audio-lips synchronized talking portrait videos with realistic details and high efficiency compared to previous methods. This method has significant implications for applications in digital humans, virtual avatars, filmmaking, and video conferencing, offering an efficient and accurate solution for creating lifelike talking portraits driven by audio input.
    
    \item \textbf{Transient Neural Radiance Fields for LIDAR View Synthesis and 3D Reconstruction} by Anagh Malik, Parsa Mirdehghan, Sotiris Nousias, Kiriakos N. Kutulakos, and David B. Lindell, introduces a novel method for rendering transient Neural Radiance Fields (NeRFs) using raw, time-resolved photon count histograms measured by a single-photon lidar system. This approach differs from conventional NeRFs as it relies on a time-resolved version of the volume rendering equation to render lidar measurements and capture transient light transport phenomena at picosecond timescales. The authors evaluated their method on a unique dataset of simulated and captured transient multiview scans from a prototype single-photon lidar. This work expands the capabilities of NeRFs to a new dimension of imaging at transient timescales, enabling the rendering of transient imagery from novel views. Additionally, the approach recovers improved geometry and conventional appearance compared to point cloud-based supervision when trained on few input viewpoints. Transient NeRFs are particularly useful for applications that require simulation of raw lidar measurements for downstream tasks in autonomous driving, robotics, and remote sensing. This advancement in NeRF technology opens up new possibilities for accurately modeling and reconstructing complex scenes using lidar data, significantly benefiting fields that rely on high-precision 3D imaging and analysis.
    
    \item \textbf{Magic NeRF Lens (Interactive Fusion of Neural Radiance Fields for Virtual Facility Inspection)} by Ke Li, Susanne Schmidt, Tim Rolff, Reinhard Bacher, Wim Leemans, and Frank Steinicke, introduces an interactive framework for facility inspection in immersive virtual reality (VR) using Neural Radiance Fields (NeRF) and volumetric rendering. The authors developed a novel data fusion approach that combines the strengths of volumetric rendering and geometric rasterization, allowing a NeRF model to be merged with other conventional 3D data, such as computer-aided design (CAD) models. This method includes two innovative 3D magic lens effects to optimize NeRF rendering by exploiting properties of human vision and context-aware visualization. The framework was validated through a technical benchmark, a visual search user study, and expert reviews, demonstrating high usability. This approach is particularly crucial for large industrial facilities like particle accelerators and nuclear power plants, which are often inaccessible to humans due to safety hazards. The Magic NeRF Lens provides a VR system that can quickly replicate real-world remote environments, offering users a high level of spatial and situational awareness crucial for facility maintenance planning. The authors' work represents a significant advancement in the application of NeRF in VR for complex and critical infrastructure inspection, enhancing the capabilities for maintenance planning and operational safety in such environments.
    
    \item \textbf{Lighting up NeRF via Unsupervised Decomposition and Enhancement} by Haoyuan Wang, Xiaogang Xu, Ke Xu, and Rynson W.H. Lau, presents a novel approach named Low-Light NeRF (LLNeRF) for enhancing scene representation and synthesizing normal-light novel views directly from sRGB low-light images in an unsupervised manner. The core of LLNeRF is a decomposition of the radiance field learning, which concurrently enhances illumination, reduces noise, and corrects distorted colors during the NeRF optimization process. This method addresses the challenge of using low-light images, which typically suffer from low pixel intensities, heavy noise, and color distortion, to train a NeRF model. LLNeRF overcomes the limitations of existing low-light image enhancement methods and NeRF methods, which struggle with view inconsistency due to individual 2D enhancement processes. The authors' approach produces novel view images with proper lighting and vivid colors and details from a collection of camera-finished low dynamic range images of a low-light scene. Experiments demonstrate that LLNeRF outperforms existing low-light enhancement methods and NeRF methods, marking a significant advancement in the field of neural radiance fields and their application in low-light conditions. This technique has substantial implications for various real-world applications, including virtual reality, augmented reality, and digital content creation, where high-quality image synthesis from challenging lighting conditions is essential.
    
    \item \textbf{Urban Radiance Field Representation with Deformable Neural Mesh Primitives} by Fan Lu, Yan Xu, Guang Chen, Hongsheng Li, Kwan-Yee Lin, and Changjun Jiang, introduces a novel approach for efficiently constructing urban-level radiance fields. The authors developed Deformable Neural Mesh Primitive (DNMP), a flexible and compact neural variant of the classic mesh representation. DNMP combines the efficiency of rasterization-based rendering with the neural representation capability for photo-realistic image synthesis. Each DNMP consists of connected deformable mesh vertices with paired vertex features to parameterize the geometry and radiance information of a local area. To optimize and reduce storage requirements, the shape of each primitive is decoded from a low-dimensional latent space, and rendering colors are decoded from vertex features by a view-dependent MLP. This method achieves high-quality rendering and low computational costs, enabling fast rendering and low peak memory usage. A lightweight version of the method runs significantly faster than vanilla NeRFs and is comparable to highly-optimized Instant-NGP. This approach offers a new paradigm for urban-level scene representation, combining high-quality rendering with efficiency. The DNMP method is particularly beneficial for applications in urban planning, virtual reality, and augmented reality, where efficient and realistic rendering of large-scale urban environments is crucial.
    
    \item \textbf{EndoSurf (Neural Surface Reconstruction of Deformable Tissues with Stereo Endoscope Videos)} by Ruyi Zha, Xuelian Cheng, Hongdong Li, Mehrtash Harandi, and Zongyuan Ge, introduces a groundbreaking neural-field-based method for reconstructing soft tissues from stereo endoscope videos. EndoSurf effectively learns to represent a deforming surface from an RGBD sequence by modeling surface dynamics, shape, and texture with three neural fields. The process involves transforming 3D points from the observed space to the canonical space using a deformation field. Subsequently, a Signed Distance Function (SDF) field and a radiance field predict their SDFs and colors, respectively. These predictions enable the synthesis of RGBD images through differentiable volume rendering. The authors implemented multiple regularization strategies to constrain the learned shape and disentangle geometry and appearance. Experiments on public endoscope datasets showed that EndoSurf significantly surpasses existing solutions, especially in reconstructing high-fidelity shapes. This method has profound implications for medical applications, particularly in Robotic-Assisted Minimally Invasive Surgery (RAMIS), where accurate 3D reconstruction of observed tissues is crucial. EndoSurf's ability to provide detailed and accurate 3D models from stereo endoscope videos represents a significant advancement in surgical scene reconstruction, offering enhanced capabilities for surgical planning, navigation, and post-operative analysis.
    
    \item \textbf{Tri-MipRF (Tri-Mip Representation for Efficient Anti-Aliasing Neural Radiance Fields)} by Wenbo Hu, Yuling Wang, Lin Ma, Bangbang Yang, Lin Gao, Xiao Liu, and Yuewen Ma, presents a novel approach to address the quality-efficiency trade-off in Neural Radiance Fields (NeRF). The authors introduced Tri-Mip encoding, an innovative technique that enables instant reconstruction and anti-aliased high-fidelity rendering for NeRF. This method factorizes pre-filtered 3D feature spaces into three orthogonal mipmaps, allowing efficient 3D area sampling by utilizing 2D pre-filtered feature maps. This significantly enhances rendering quality without compromising efficiency. To complement the Tri-Mip representation, a cone-casting rendering technique was proposed for efficiently sampling anti-aliased 3D features, considering both pixel imaging and observing distance. Extensive experiments on synthetic and real-world datasets demonstrated that Tri-MipRF achieves state-of-the-art rendering quality and reconstruction speed while maintaining a compact representation. This method reduces the model size by 25\% compared to InstantNGP, marking a significant advancement in the field of 3D rendering and modeling. Tri-MipRF's ability to balance high-quality rendering with efficient reconstruction makes it a valuable tool for applications in virtual reality, augmented reality, and digital content creation, where both visual fidelity and computational efficiency are crucial.
    
    \item \textbf{FaceCLIPNeRF (Text-driven 3D Face Manipulation using Deformable Neural Radiance Fields)} by Sungwon Hwang, Junha Hyung, Daejin Kim, Min-Jung Kim, and Jaegul Choo, introduces a groundbreaking approach to 3D face manipulation using Neural Radiance Fields (NeRF). The authors developed a method that requires only a single text input to manipulate a face reconstructed with NeRF. This process involves training a scene manipulator, a latent code-conditional deformable NeRF, over a dynamic scene. This manipulator controls face deformation using the latent code. However, representing scene deformation with a single latent code can be limiting for local deformations. To address this, the team proposed the Position-conditional Anchor Compositor (PAC), which learns to represent a manipulated scene with spatially varying latent codes. The renderings from the scene manipulator are optimized to achieve high cosine similarity to a target text in the CLIP embedding space, enabling text-driven manipulation. This innovative approach is the first of its kind to facilitate text-driven manipulation of faces reconstructed with NeRF. The extensive results, comparisons, and ablation studies conducted by the authors demonstrate the effectiveness of FaceCLIPNeRF. This method has significant implications for 3D digital human content creation, offering an easy and intuitive way for both experts and non-experts to manipulate 3D face representations.
    
    \item \textbf{CopyRNeRF (Protecting the CopyRight of Neural Radiance Fields)}  by Ziyuan Luo, Qing Guo, Ka Chun Cheung, Simon See, and Renjie Wan, addresses the critical issue of copyright protection for Neural Radiance Fields (NeRF) models. The authors propose a unique solution that embeds a watermarked color representation into the NeRF model, replacing the original color representation. This method ensures robust message extraction in 2D renderings of NeRF while maintaining high rendering quality and bit accuracy. The key innovation lies in designing a distortion-resistant rendering scheme that embeds copyright messages directly into the NeRF model, rather than just the rendered samples. This approach effectively secures the intellectual property of NeRF models against unauthorized use or theft. The authors demonstrate that CopyRNeRF can claim model ownership by transmitting embedded copyright messages from the models to the rendering samples. This method is particularly relevant in the context of digital media, where the protection of intellectual property is paramount. By embedding copyright messages into the 3D structure implicitly encoded in the weights of the NeRF model, CopyRNeRF offers a novel and effective solution for safeguarding the rights of creators and developers in the rapidly evolving field of 3D modeling and rendering.
    
    \item \textbf{TransHuman (A Transformer-based Human Representation for Generalizable Neural Human Rendering)} by Xiao Pan, Zongxin Yang, Jianxin Ma, Chang Zhou, and Yi Yang, introduces a novel framework for generalizable neural human rendering, which trains conditional Neural Radiance Fields (NeRF) from multi-view videos of different characters. Addressing the limitations of previous methods that used a SparseConvNet (SPC)-based human representation, TransHuman employs transformers to capture global relationships between human parts, crucial for handling incomplete painted SMPL (Skinned Multi-Person Linear model). The framework consists of three main components: Transformer-based Human Encoding (TransHE), Deformable Partial Radiance Fields (DPaRF), and Fine-grained Detail Integration (FDI). TransHE processes the painted SMPL under canonical space using transformers, DPaRF binds each output token with a deformable radiance field for encoding the query point under the observation space, and FDI integrates fine-grained information from reference images. Extensive experiments on ZJU-MoCap and H36M datasets demonstrate that TransHuman significantly outperforms existing methods in terms of efficiency and state-of-the-art performance. This advancement in neural human rendering is particularly impactful for applications in mixed reality, gaming, and telepresence, offering a more efficient and accurate method for rendering free-viewpoint videos of dynamic human performers.
    
    \item \textbf{CarPatch (A Synthetic Benchmark for Radiance Field Evaluation on Vehicle Components)} by Davide Di Nucci, Alessandro Simoni, Matteo Tomei, Luca Ciuffreda, Roberto Vezzani, and Rita Cucchiara, introduces a novel synthetic benchmark specifically designed for evaluating Neural Radiance Fields (NeRFs) in the context of vehicle inspection. Recognizing the challenges posed by scenarios like vehicle inspection, where data limitations and difficult elements such as reflections can significantly impact the accuracy of 3D reconstructions, the authors developed CarPatch. This benchmark comprises a set of images annotated with intrinsic and extrinsic camera parameters, along with corresponding depth maps and semantic segmentation masks for each view. The team established global and part-based metrics to evaluate, compare, and better characterize some state-of-the-art NeRF techniques. CarPatch is publicly released and serves as a valuable resource for researchers and practitioners in the field. It provides a comprehensive evaluation guide and a baseline for future work in the challenging area of vehicle component analysis and 3D reconstruction, offering significant contributions to the advancement of NeRF applications in real-world scenarios like automotive inspection and quality control.
    
    \item \textbf{Dyn-E (Local Appearance Editing of Dynamic Neural Radiance Fields)} by Shangzhan Zhang, Sida Peng, Yinji Shentu, Qing Shuai, Tianrun Chen, Kaicheng Yu, Hujun Bao, and Xiaowei Zhou, presents a novel framework for editing the local appearance of dynamic Neural Radiance Fields (NeRFs). This approach enables users to manipulate pixels in a single frame of a training video to edit dynamic NeRFs. The key innovation lies in introducing a local surface representation of the edited region, which can be seamlessly integrated into the original NeRF. This representation is rendered alongside the original NeRF and can be warped to arbitrary other frames through a learned invertible motion representation network. This method allows users, even without professional expertise, to easily add desired content to the appearance of a dynamic scene. The authors extensively evaluated their approach on various scenes, demonstrating that it achieves spatially and temporally consistent editing results. Notably, this approach is versatile and applicable to different variants of dynamic NeRF representations. This advancement in NeRF technology has significant implications for 3D content editing, offering a user-friendly and efficient solution for adding or modifying content in dynamic scenes, with potential applications in virtual reality, augmented reality, and digital media production.
    
    \item \textbf{Strivec (Sparse Tri-Vector Radiance Fields)} by Quankai Gao, Qiangeng Xu, Hao Su, Ulrich Neumann, and Zexiang Xu, introduces a new neural representation for modeling 3D scenes as radiance fields with sparsely distributed and compactly factorized local tensor feature grids. Strivec leverages tensor decomposition, specifically the classic CANDECOMP/PARAFAC (CP) decomposition, to factorize each tensor into triple vectors. These vectors express local feature distributions along spatial axes and compactly encode a local neural field. The method also employs multi-scale tensor grids to capture geometry and appearance commonalities at multiple local scales. The final radiance field properties are determined by aggregating neural features from multiple local tensors across all scales. A key aspect of Strivec is its sparse distribution of tri-vector tensors around the actual scene surface, identified through a fast coarse reconstruction process. This approach capitalizes on the inherent sparsity of 3D scenes. The authors demonstrate that Strivec achieves superior rendering quality with significantly fewer parameters compared to previous methods like TensoRF and Instant-NGP. This advancement in neural scene representation is particularly beneficial for applications requiring high-quality rendering and compact model sizes, such as virtual reality, augmented reality, and 3D graphics.
    
    \item \textbf{Car-Studio (Learning Car Radiance Fields from Single-View and Endless In-the-wild Images)} by Tianyu Liu, Hao Zhao, Yang Yu, Guyue Zhou, and Ming Liu, introduces a novel approach for learning General Car-NeRF (Neural Radiance Fields) from an unlimited number of car images available online. The method involves extracting car patches from internet images using a 2D object detector, preprocessing these patches with rough camera intrinsics, object instance masks, and 3D detection bounding boxes, and then training the Car-NeRF model. This model is capable of outputting high-quality car images from any 3D viewpoint based on a single-view image. The authors' approach addresses the challenge of maintaining clarity and sharp contours in vehicle images when changing perspectives in an editable autonomous driving simulator. By using datasets built from in-the-wild images, their method enables controllable appearance editing, demonstrating competitive performance compared to existing baselines. This research contributes significantly to the fields of computer vision and autonomous driving, offering a practical solution for generating realistic car images for various applications, including simulators and virtual environments. The decision to release the dataset and code will further facilitate research and development in this area.
    
    \item \textbf{MapNeRF (Incorporating Map Priors into Neural Radiance Fields for Driving View Simulation)} by Chenming Wu, Jiadai Sun, Zhelun Shen, and Liangjun Zhang, presents a novel approach to enhance the capabilities of neural radiance fields (NeRF) in autonomous driving simulations. The authors focused on synthesizing out-of-trajectory driving views with semantic road consistency by incorporating map priors into NeRF. Their key innovation lies in using map information as a guide during the training of radiance fields, addressing the uncertainty in scene complexity. The method involves supervising the density field with coarse ground surface information from maps and warping depth with uncertainty from unknown camera poses to maintain multi-view consistency. This approach allows for the production of semantically consistent deviated views for vehicle camera simulation, a crucial aspect for the safety and reliability of autonomous driving systems. The authors' work addresses a significant challenge in autonomous driving simulations, where ensuring semantic consistency in synthesized images from deviated views is essential. By integrating map data into the NeRF framework, they have opened new possibilities for more robust and realistic simulations in autonomous driving, contributing to the advancement of safe and efficient autonomous vehicle technologies.
    
    \item \textbf{MARS (An Instance-aware, Modular and Realistic Simulator for Autonomous Driving)} by Zirui Wu, Tianyu Liu, Liyi Luo, and their team, introduces a groundbreaking autonomous driving simulator based on neural radiance fields (NeRFs). This simulator is distinguished by three key features: instance-awareness, modularity, and realism. The instance-aware approach allows the simulator to model foreground instances and background environments separately with independent networks, enabling separate control over static properties (like size and appearance) and dynamic properties (such as trajectory) of instances. Its modular design facilitates flexible switching between various modern NeRF-related backbones, sampling strategies, and input modalities, aiming to accelerate both academic progress and industrial deployment of NeRF-based autonomous driving simulation. Additionally, the simulator sets new standards in photorealism, outperforming existing counterparts. The authors have made a significant contribution to the field of autonomous driving by addressing the critical need for realistic sensor simulation in handling corner cases on the road, thereby enhancing the safety and reliability of autonomous vehicles. The decision to open-source the simulator further underscores its potential impact on the development and testing of autonomous driving technologies.
    
    \item \textbf{Seal-3D (Interactive Pixel-Level Editing for Neural Radiance Fields)} by Xiangyu Wang, Jingsen Zhu, Qi Ye, and their team, introduces Seal-3D, a pioneering interactive tool for pixel-level editing of Neural Radiance Fields (NeRF). This tool addresses the need for flexible, high-quality, and speedy editing methods for implicit 3D models, particularly in tasks like post-processing reconstructed scenes and 3D content creation. Seal-3D stands out for its ability to offer direct editing response and instant preview, a significant challenge in the realm of NeRF editing. The authors achieved this by developing a proxy function that maps editing instructions to the original space of NeRF models in the teacher model, coupled with a two-stage training strategy for the student model that includes local pretraining and global finetuning. The system built around Seal-3D showcases various editing types and can achieve compelling editing effects with an interactive speed of about 1 second. This innovation in NeRF technology is a substantial leap forward, offering practical applications in 3D reconstruction, novel view synthesis, and virtual reality, with the potential to significantly enhance workflows in 3D content creation and scene editing.
    
    \item \textbf{Dynamic PlenOctree for Adaptive Sampling Refinement in Explicit NeRF} by Haotian Bai, Yiqi Lin, Yize Chen, and Lin Wang, presents a significant advancement in the field of neural radiance fields (NeRF) with their development of the dynamic PlenOctree (DOT). This method addresses the limitations of the fixed structure in PlenOctree (POT) for direct optimization, which becomes sub-optimal as scene complexity evolves. DOT introduces a hierarchical feature fusion strategy during the iterative rendering process, focusing on adaptively refining the sample distribution to match changing scene complexities. The process involves identifying regions of interest through training signals for adaptive refinement and introducing sampling and pruning operations for octrees to aggregate features, thereby facilitating rapid parameter learning. The authors demonstrated that DOT not only enhances visual quality but also significantly reduces computational resources, with over 55.15/68.84\% fewer parameters and 1.7/1.9 times faster FPS for NeRF-synthetic and Tanks \& Temples datasets, respectively, compared to POT. This innovative approach in NeRF technology has profound implications for virtual reality and gaming, offering more efficient and high-quality rendering capabilities.
    
    \item \textbf{Robust Single-view Cone-beam X-ray Pose Estimation with Neural Tuned Tomography (NeTT) and Masked Neural Radiance Fields (mNeRF)} by Chaochao Zhou, Syed Hasib Akhter Faruqui, Abhinav Patel, and their colleagues, introduces innovative methods for pose estimation of radiolucent objects using X-ray projections, a critical task in image-guided, minimally invasive medical procedures. The team developed an algorithm, DiffDRR, within TensorFlow, which efficiently computes Digitally Reconstructed Radiographs (DRRs) and leverages automatic differentiation for pose estimation through iterative gradient descent. They proposed two novel methods for high-fidelity view synthesis: Neural Tuned Tomography (NeTT) and masked Neural Radiance Fields (mNeRF). Both methods utilize classic Cone-Beam Computerized Tomography (CBCT), with NeTT directly optimizing CBCT densities and mNeRF constrained by a 3D mask of the anatomic region segmented from CBCT. Their results showed that both NeTT and mNeRF significantly improve pose estimation, achieving success rates over 93\% with a 3D angle error of less than 3°. Notably, NeTT demonstrated a lower computational cost than mNeRF in both training and pose estimation, and a NeTT trained for a single subject could generalize to synthesize high-fidelity DRRs for other subjects. This research suggests that NeTT is a particularly attractive option for robust pose estimation in medical applications, offering efficient and accurate solutions for challenging tasks in image-guided procedures.
    
    \item \textbf{Context-Aware Talking-Head Video Editing}, a research paper presented at the 31st ACM International Conference on Multimedia, introduces a novel framework for editing talking-head videos through text transcript editing. The authors focused on efficiently inserting, deleting, and substituting words in pre-recorded videos while maintaining accurate lip synchronization and motion smoothness. The framework comprises two main components: an animation prediction module and a motion-conditioned rendering module. The animation prediction module, using a non-autoregressive network, efficiently obtains smooth and lip-sync motion sequences conditioned on the driven speech. It learns a speech-animation mapping prior from a multi-identity video dataset for better generalization to novel speech. The neural rendering module then synthesizes photo-realistic and full-head video frames based on the predicted motion sequence. This module leverages a pre-trained head topology and requires only a few frames for efficient fine-tuning to obtain a person-specific rendering model. Extensive experiments demonstrated that this method achieves smoother editing results with higher image quality and lip accuracy using less data than previous methods. This advancement in video editing technology holds significant potential for applications in media production, virtual reality, and digital communication, offering more efficient and realistic editing of talking-head videos.
    
    \item \textbf{High-Fidelity Eye Animatable Neural Radiance Fields for Human Face} by Hengfei Wang, Zhongqun Zhang, Yihua Cheng, and Hyung Jin Chang, presented a novel approach to face rendering using Neural Radiance Fields (NeRF) with a focus on accurately modeling eyeball rotation. The team tackled two primary challenges: capturing eyeball rotation effectively for training and constructing a manifold for representing this rotation. Their method involved fitting the FLAME parametric face model to multi-view images, ensuring multi-view consistency. They then introduced the Dynamic Eye-aware NeRF (DeNeRF), which transforms 3D points from different views into a canonical space to learn a unified face NeRF model. This model includes both rigid transformations, like eyeball rotation, and non-rigid transformations. Tested on the ETH-XGaze dataset, their model demonstrated the ability to generate high-fidelity images with accurate eyeball rotation and periocular deformation, even under novel viewing angles. This advancement in face rendering technology has significant implications for applications in virtual reality, digital human creation, and computer graphics, particularly in enhancing gaze estimation performance and creating more lifelike digital human representations.
    
    \item \textbf{NeRFs: Incorporating Season and Solar Specificity into Renderings made by a NeRF Architecture using Satellite Images} by Michael Gableman and Avinash Kak advanced the capabilities of Neural Radiance Fields (NeRFs) for rendering scenes from novel viewpoints using satellite imagery. Building on previous works like Shadow NeRF and Sat-NeRF, their main contribution was the creation of a NeRF that could render seasonal features independently of viewing and solar angles while accurately rendering shadows. They introduced an additional input variable, the time of the year, to teach the network to render seasonal features. To overcome the challenges posed by small training datasets typical in satellite imagery, they added terms to the loss function to prevent the network from confusing seasonal features with shadows. The performance of their network was demonstrated on eight Areas of Interest using images from the Maxar WorldView-3 satellite, showing its ability to accurately render novel views, generate height maps, predict shadows, and specify seasonal features independently from shadows. This work has significant implications for applications requiring accurate, season-specific renderings from satellite data, such as environmental monitoring, urban planning, and agricultural analysis.
    
    \item \textbf{NeRFs: The Search for the Best 3D Representation} by Ravi Ramamoorthi presented a comprehensive overview of Neural Radiance Fields (NeRFs) and their transformative role in 3D scene representation and view synthesis. Diverging from traditional 3D representation methods like meshes or voxel grids, NeRFs utilize a continuous volume representation, with view-dependent radiance and volume density parameters obtained from querying a neural network. The core of NeRFs is a fully-connected neural network, specifically a multi-layer perceptron (MLP), which differs from the commonly used convolutional neural networks (CNNs). This MLP-based approach enables simpler, more efficient algorithms that are easier to train and more suited to the task at hand. The practical applications of NeRFs are extensive and impactful, particularly in immersive virtual and augmented reality, digital twins, and the metaverse. Notably, NeRFs have been applied in Google's Maps and StreetView for creating immersive renderings from photographs, and in the Luma AI mobile app for generating 3D fly-throughs from a few images. Ramamoorthi's paper not only highlights the significance of NeRFs in advancing technology but also offers insights into the future of 3D representations.
    
    \item In \textbf{Learning Unified Decompositional and Compositional NeRF for Editable Novel View Synthesis}, the authors introduced a novel framework for joint scene novel view synthesis and editing using implicit neural scene representations. This approach, based on Neural Radiance Field (NeRF), effectively performs joint scene decomposition and composition, modeling real-world scenes. The method involves a two-stage NeRF framework: the first stage predicts a global radiance field for point sampling, and the second stage, more fine-grained, performs scene decomposition using a unique one-hot object radiance field regularization module and pseudo supervision via inpainting for ambiguous background regions. This decomposition and composition technique allows for disentangled 3D representations of different objects and the background, facilitating scene editing and novel view synthesis. The authors demonstrated that their method outperforms state-of-the-art methods in both novel-view synthesis and editing tasks, showing potential applications in scene understanding, content generation, and robotic manipulation.
    
    \item In \textbf{Where and How: Mitigating Confusion in Neural Radiance Fields from Sparse Inputs}, the authors addressed the challenges in Neural Radiance Fields from Sparse inputs (NeRF-S), particularly focusing on the issues of over-fitting and foggy surfaces in rendered results due to sparse inputs. They identified two fundamental questions: "WHERE to Sample?" and "HOW to Predict?". To tackle these, they introduced WaH-NeRF, a novel learning framework. This framework includes a Deformable Sampling strategy and a Weight-based Mutual Information Loss to resolve sample-position confusion, and a Semi-Supervised NeRF learning Paradigm with pose perturbation and a Pixel-Patch Correspondence Loss to alleviate prediction confusion. These innovations in WaH-NeRF significantly enhanced performance under the NeRF-S setting, outperforming previous methods. The applications of this technique are broad, particularly in areas where collecting dense viewing inputs is challenging, such as robotic navigation and autonomous driving, offering more reliable and accurate 3D scene modeling and rendering with sparse data.
    
    \item In \textbf{Mirror-NeRF: Learning Neural Radiance Fields for Mirrors with Whitted-Style Ray Tracing},  the authors propose a novel neural rendering framework, Mirror-NeRF, to address the challenge of synthesizing novel views in scenes with mirrors. Traditional Neural Radiance Fields (NeRF) models struggle with accurately rendering mirror reflections, often leading to incorrect geometry reconstruction and multi-view inconsistencies. Mirror-NeRF overcomes these limitations by introducing a reflection probability parameterized by a Multi-Layer Perceptron (MLP) and employing a ray tracing approach inspired by Whitted Ray Tracing. This method allows for physically accurate rendering of reflections, ensuring high-fidelity novel view synthesis with accurate geometry and reflection of mirrors. The paper also introduces techniques for learning both accurate geometry and reflection of mirrors, including surface normal parametrization for smooth distribution, plane consistency, forward-facing normal constraints, and a progressive training strategy for stability. These innovations enable Mirror-NeRF to support various scene manipulation applications, such as object placement, mirror roughness control, and reflection substitution. The authors demonstrate the effectiveness of Mirror-NeRF through extensive experiments on real and synthetic datasets, showcasing its ability to achieve photo-realistic novel view synthesis and physical correctness in scene manipulation.
    
    \item \textbf{Improved Neural Radiance Fields Using Pseudo-depth and Fusion} by Jingliang Li, Qiang Zhou, Chaohui Yu, Zhengda Lu, Jun Xiao, Zhibin Wang, and Fan Wang, focuses on enhancing Neural Radiance Fields (NeRF) for novel view synthesis and dense geometry modeling. The authors address the limitations of existing NeRF methods, particularly in scenes with diverse objects and structures. They propose an end-to-end method that utilizes multi-scale radiance fields, an auxiliary depth prediction head, and feature fusion. This approach involves constructing pyramid-structured encoding volumes to provide geometric information at various scales, catering to both large and small objects in a scene. The method also includes depth prediction and radiance field reconstruction, using the predicted depth map to supervise the rendered depth, narrow the depth range, and guide point sampling. Additionally, the authors suggest enhancing point volume features through depth-guided neighbor feature fusion to overcome inaccuracies caused by occlusion and lighting. The paper demonstrates the superior performance of this method in both novel view synthesis and dense geometry modeling without per-scene optimization.
    
    \item \textbf{3D Gaussian Splatting for Real-Time Radiance Field Rendering} by Bernhard Kerbl, Georgios Kopanas, Thomas Leimkühler, and George Drettakis introduces a novel approach to real-time rendering of radiance fields with high visual quality. The authors developed a 3D Gaussian scene representation coupled with a real-time differentiable renderer. This method significantly speeds up both scene optimization and novel view synthesis. The key innovation lies in representing the scene with 3D Gaussians, starting from sparse points produced during camera calibration. This preserves the desirable properties of continuous volumetric radiance fields for scene optimization while avoiding unnecessary computation in empty space. Additionally, the paper introduces an interleaved optimization/density control of the 3D Gaussians, optimizing anisotropic covariance for accurate scene representation. The authors also developed a fast visibility-aware rendering algorithm that supports anisotropic splatting, which accelerates training and enables real-time rendering. The method demonstrates state-of-the-art visual quality and real-time rendering on several established datasets, achieving a balance between high-quality rendering and competitive training times.
    
    \item In \textbf{Digging into Depth Priors for Outdoor Neural Radiance Fields} by Chen Wang et al., the authors delve into the integration of depth priors into outdoor Neural Radiance Fields (NeRF) training to address the shape-radiance ambiguity, particularly in sparse viewpoints settings. The study is comprehensive, evaluating common depth sensing technologies and various application methods. The authors conducted extensive experiments using two representative NeRF methods, four commonly-used depth priors, and different depth usages on two widely used outdoor datasets. Their findings offer valuable insights for practitioners and researchers in enhancing NeRF models with depth priors, highlighting the significant improvement in view synthesis, even with monocular depth estimation, in terms of fewer floaters and better preservation of object shapes. This research contributes to the field of computer vision and multimedia, particularly in novel view synthesis, by exploring the potential of depth priors in improving the performance of NeRF in outdoor environments.
    
    \item In \textbf{A General Implicit Framework for Fast NeRF Composition and Rendering}, the authors, Xinyu Gao, Ziyi Yang, Yunlu Zhao, Yuxiang Sun, Xiaogang Jin, and Changqing Zou, propose a novel approach to enhance the efficiency of Neural Radiance Fields (NeRF) composition and rendering. They introduce a new surface representation called Neural Depth Fields (NeDF) which allows for quick determination of spatial relationships between objects, enabling direct intersection computation between rays and implicit surfaces. This method significantly accelerates the process by avoiding reliance on explicit spatial structures and allows for the casting of dynamic shadows within or between objects using analytical light sources. The framework supports seamless placement and rendering of multiple NeRF objects with arbitrary rigid transformations, offering a low storage cost solution for interactive previewing of NeRF compositions. This advancement addresses the challenges of real-time composition in various NeRF applications, making it a valuable contribution to the field of image-based rendering and virtual world development.
    
    \item In \textbf{WaveNeRF (Wavelet-based Generalizable Neural Radiance Fields)}, the authors introduce WaveNeRF, a novel approach to neural radiance fields that leverages wavelet transforms for improved rendering of high-frequency features in multi-view stereo (MVS) environments. Unlike traditional methods that use Fourier Transform, WaveNeRF employs Wavelet Transform, which is coordinate invariant and preserves the relative spatial positions of pixels. This property is particularly advantageous in MVS, allowing for the warping of multiple input views in the direction of a reference view to form sweeping planes in both the spatial and frequency domains within the same coordinate system. The paper also introduces a Frequency-guided Sampling Strategy (FSS) that focuses on regions with larger high-frequency coefficients, enhancing rendering quality, especially around object surfaces. The authors' contributions include designing a WMVS module that effectively preserves high-frequency information, a Hybrid Neural Renderer (HNR) that merges features from both spatial and frequency domains, and developing FSS for improved rendering quality. This approach has potential applications in 3D scene reconstruction, novel view synthesis, and other computer vision tasks, particularly where high-quality rendering of detailed environments is crucial.
    
    \item In \textbf{VERF (Runtime Monitoring of Pose Estimation with Neural Radiance Fields)}, the authors introduce a novel approach for assessing the accuracy of camera pose estimates in real-time, particularly focusing on the position estimate's quality. Their method involves comparing optical flow fields between three images with co-linear camera origins and no rotation, allowing for the determination of the order of camera centers and the correctness of the pose estimate. This process is enhanced with an estimate of optical flow error and outlier rejection using geometric constraints, leading to a measure of confidence rather than a binary decision. The paper emphasizes the application of this technique in challenging real-world conditions, showcasing results from experiments conducted with the LLFF dataset, an A1 quadruped, and onboard Blue Origin’s sub-orbital New Shepard rocket. The authors' approach, which runs in less than half a second on a 3090 GPU, demonstrates significant potential for real-time monitoring in various demanding environments.
    
    \item In \textbf{S3IM (Stochastic Structural SIMilarity and Its Unreasonable Effectiveness for Neural Fields)}, , the authors introduce a novel approach to enhance the training of neural field methods, such as Neural Radiance Fields (NeRF) and neural surface representation. They propose the Stochastic Structural SIMilarity (S3IM) index, which measures the similarity between groups of pixels, capturing nonlocal structural information from stochastically sampled pixels. This approach is distinct from conventional methods that focus on point-wise Mean Squared Error (MSE) loss, which often overlooks structural information crucial for human visual perception. The S3IM index is model-agnostic and can be applied to various neural field methods, offering significant improvements in quality metrics, especially in challenging tasks and scenarios with sparse inputs, corrupted images, and dynamic scenes. The paper demonstrates the effectiveness of this nonlocal multiplex training paradigm in improving the performance of neural field methods.
    
    \item \textbf{Neural radiance fields in the industrial and robotics domain: applications, research opportunities and use cases} by Eugen Slapak, Enric Pardo, Matúš Dopiriak, Taras Maksymyuk, and Juraj Gazda, published in the Journal of LaTeX Class Files in August 2021, explores the application of Neural Radiance Fields (NeRFs) in industrial contexts. The authors discuss the increasing demand for high-quality three-dimensional (3D) graphical representations in industries such as computer-aided design (CAD), finite element analysis (FEA), scanning, and robotics. They identify the limitations of current 3D representation methods, such as high implementation costs and reliance on manual human input. The paper presents NeRFs as a promising solution for learning 3D scene representations from 2D images, addressing these challenges. The authors conduct proof-of-concept experiments, including NeRF-based video compression and using NeRFs for 3D motion estimation in collision avoidance scenarios. The results demonstrate significant compression savings and high accuracy in motion estimation, showcasing the potential of NeRFs in industrial applications. The paper also provides directions for future research in this area, highlighting the unexplored potential of NeRFs in various industrial subdomains.
    
    \item In \textbf{Ref-DVGO (Reflection-Aware Direct Voxel Grid Optimization for an Improved Quality-Efficiency Trade-Off in Reflective Scene Reconstruction)}, the authors, Georgios Kouros, Minye Wu, Shubham Shrivastava, Sushruth Nagesh, Punarjay Chakravarty, and Tinne Tuytelaars, address the challenge of rendering reflective objects in Neural Radiance Fields (NeRFs). They propose an approach that balances efficiency and quality by enhancing reconstruction quality and accelerating training and rendering processes. This is achieved through an implicit-explicit approach based on conventional volume rendering, utilizing an efficient density-based grid representation and reparameterizing reflected radiance. Their reflection-aware method demonstrates a competitive quality-efficiency trade-off compared to existing methods. The paper also discusses hypotheses regarding factors influencing the results of density-based methods in reconstructing reflective objects. The source code for their work is made available for further exploration and development.
    
    \item In \textbf{Language-enhanced RNR-Map (Querying Renderable Neural Radiance Field maps with natural language)}, the authors, Francesco Taioli, Federico Cunico, Federico Girella, Riccardo Bologna, Alessandro Farinelli, and Marco Cristani, introduce Le-RNR-Map, a novel approach for visual navigation using natural language queries. This technique enhances the Renderable Neural Radiance Map (RNR-Map) with CLIP-based embedding latent codes, allowing for natural language search without additional label data. The authors demonstrate the effectiveness of this map in single and multi-object searches and explore its compatibility with a Large Language Model as an "affordance query resolver." The Le-RNR-Map is visually descriptive due to Neural Radiance Field embeddings, generalizable with off-the-shelf models requiring no training, and queryable with both text and images. This work represents a significant advancement in embodied AI, particularly in tasks like Embodied Question Answering, Image-Goal Navigation, and Visual-Language Navigation, by creating a map representation that simultaneously addresses multiple distinct tasks.
    
    \item \textbf{Watch Your Steps: Local Image and Scene Editing by Text Instructions} by Ashkan Mirzaei et al. introduces a method for localizing the desired edit region in an image or scene based on text instructions. The authors leverage InstructPix2Pix (IP2P) to identify discrepancies between IP2P predictions with and without the given instruction, creating what they call a "relevance map." This map indicates the importance of changing each pixel to achieve the desired edits, ensuring that irrelevant pixels remain unchanged. The technique is also applied to 3D scenes in the form of neural radiance fields (NeRFs), where a field is trained on relevance maps of training views, defining the 3D region for modifications. The paper demonstrates state-of-the-art performance in both image and NeRF editing tasks, highlighting the potential for automated, localized modifications in various applications such as social media, marketing, and education.
    
    \item In \textbf{DReg-NeRF (Deep Registration for Neural Radiance Fields)}, the authors address the challenge of registering Neural Radiance Fields (NeRFs) without relying on human annotations or initializations. The study is motivated by the limitations of existing works that focus mostly on point cloud registration and the need for human-annotated keypoints in NeRF registration, which is impractical in many real-world scenarios. The authors propose a novel approach that involves using an occupancy grid alongside each NeRF model to extract a voxel grid. This grid is then processed through a 3D FPNs and a transformer module, which includes a self-attention layer for intra-feature enhancement and a cross-attention layer for learning inter-feature relations. The method decodes source and target features into correspondences and confidence scores, offering a solution that does not depend on pre-computed overlapping scores or human intervention. The paper's contributions include a dataset for registering multiple NeRF blocks, a network for NeRF block registration without human intervention, and extensive experiments demonstrating the accuracy and generalization ability of the method. This approach has potential applications in areas where accurate 3D scene reconstruction and registration are crucial, such as virtual and augmented reality, robotics, and autonomous navigation.
    
    \item In \textbf{MonoNeRD (NeRF-like Representations for Monocular 3D Object Detection)}, the authors, Junkai Xu, Liang Peng, Haoran Cheng, Hao Li, Wei Qian, Ke Li, Wenxiao Wang, and Deng Cai, introduce a novel framework for enhancing monocular 3D detection (M3D) in computer vision. They address the limitations of existing methods that rely on explicit scene geometric clues, such as depth map estimation and back-projection into 3D space, which often result in sparse 3D representations and substantial information loss, particularly for distant and occluded objects. The proposed solution, MonoNeRD, infers dense 3D geometry and occupancy by modeling scenes with Signed Distance Functions (SDF). This approach treats the representations as Neural Radiance Fields (NeRF) and employs volume rendering to recover RGB images and depth maps. This method marks the first introduction of volume rendering for M3D and demonstrates the potential of implicit reconstruction for image-based 3D perception. The effectiveness of MonoNeRD is validated through extensive experiments on the KITTI-3D benchmark and Waymo Open Dataset. The paper's contribution lies in its innovative approach to generating dense 3D representations, which could significantly impact applications like autonomous driving and robotic navigation.
    
    \item In \textbf{Semantic-Human: Neural Rendering of Humans from Monocular Video with Human Parsing}, , the authors introduce Semantic-Human, a novel approach for neural rendering of humans from monocular video. This method simultaneously reconstructs the neural radiance field and the neural segmentation field, enabling accurate and consistent human parsing while maintaining photorealistic details. The key innovation lies in the integration of silhouette constraints and a surface loss, which regularizes the motion field from the SMPL prior and the recovered volumetric geometry. Additionally, the authors incorporate a multi-class cross-entropy loss for human parsing, even in scenarios with sparse and noisy semantic labels. The paper demonstrates the effectiveness of this approach on the ZJU-MoCap dataset, achieving superior results in photorealistic rendering and robust human parsing with viewpoint consistency. This technique has significant implications for enhancing the realism and accuracy of human-centric applications in virtual reality, gaming, and digital content creation.
    
    \item \textbf{AltNeRF (Learning Robust Neural Radiance Field via Alternating Depth-Pose Optimization)} by Kun Wang et al. addresses the challenges faced by Neural Radiance Fields (NeRF) in generating realistic novel views from sparse scene images, particularly due to the lack of explicit 3D supervision and imprecise camera poses. The authors propose AltNeRF, a novel framework that utilizes self-supervised monocular depth estimation (SMDE) from monocular videos, eliminating the need for known camera poses. This approach leverages depth and pose priors to regulate NeRF training, enriching its capacity for precise scene geometry depiction and providing a robust starting point for pose refinement. An alternating algorithm is introduced, which integrates NeRF outputs into SMDE through a consistency-driven mechanism, enhancing the integrity of depth priors and progressively refining NeRF representations. This results in the synthesis of realistic novel views. The paper also presents a unique dataset of indoor videos captured via mobile devices and demonstrates AltNeRF's effectiveness in generating high-fidelity and robust novel views through extensive experiments.
    
    \item In \textbf{HollowNeRF (Pruning Hashgrid-Based NeRFs with Trainable Collision Mitigation)}, , the authors introduce HollowNeRF, a novel method for compressing Neural Radiance Fields (NeRF) using trainable hash collision mitigation. This approach aims to enhance rendering accuracy while using fewer parameters compared to existing NeRF methods. HollowNeRF operates on a hash-based pipeline, similar to Instant-NGP, but with a focus on prioritizing important features of visible voxels and pruning unnecessary features of invisible voxels. This results in a more efficient distribution of hash collision probability across the 3D volume. The key innovation of HollowNeRF lies in its ability to steer shared features in a hash bucket towards more important voxels, reducing interference with other features. This is achieved through a trainable saliency weight that scales features based on voxel visibility, and a 3D saliency grid that captures the visibility of coarse grid regions. Unlike other methods that require prior knowledge of surface geometries, HollowNeRF learns to prioritize important features through training. The paper also introduces a soft zero-skipping gate and a pruner to further enhance the model's efficiency. The authors demonstrate that HollowNeRF achieves better accuracy (measured in PSNR and LPIPS) than state-of-the-art methods while requiring significantly fewer parameters. This balance between cost and accuracy makes HollowNeRF a promising solution for efficient and accurate rendering in various applications, particularly in scenarios where computational resources are limited.
    
    \item \textbf{Strata-NeRF (Neural Radiance Fields for Stratified Scenes)} by Ankit Dhiman et al. introduces Strata-NeRF, a novel approach to model complex, multi-level scenes using a single neural radiance field. Traditional Neural Radiance Field (NeRF) methods excel in generating photo-realistic views of single objects or single-level scenes. However, they struggle with layered captures common in real-world scenarios, such as a tourist photographing both the exterior and interior of a monument. Strata-NeRF addresses this by conditioning the NeRFs on Vector Quantized (VQ) latent representations, enabling the model to handle sudden changes in scene structure. This technique is evaluated on multi-layered synthetic datasets and further validated on the RealEstate10K real-world dataset. The results demonstrate that Strata-NeRF effectively captures stratified scenes, reduces artifacts, and synthesizes high-fidelity views, outperforming existing methods. This advancement has significant implications for enhancing immersive experiences in applications like augmented and virtual reality.
    
    \item In \textbf{CamP (Camera Preconditioning for Neural Radiance Fields)}, , the authors address the challenge of camera parameterization in the context of Neural Radiance Fields (NeRFs). They observe that the choice of camera parameterization significantly impacts the initialization and accuracy of camera poses in NeRFs. The paper critiques existing approaches like BARF and SCNeRF for their representation of camera pose and intrinsic parameters, highlighting the need for better parameterization to improve scene reconstruction accuracy. The authors propose a novel technique called Camera Preconditioning (CamP), which aims to normalize the effects of each camera parameter with respect to the projection of points in a scene and to decorrelate each parameter’s effects from others. This is achieved through a preconditioner, a matrix applied to the camera parameters before they are passed to the NeRF model. The technique is evaluated using the state-of-the-art ZipNeRF method, and the experiments demonstrate its effectiveness across various camera parameterizations and challenging real-world scenarios. The paper contributes to the field of 3D reconstruction and scene understanding by offering a solution to improve the conditioning of camera parameters in NeRFs, which is crucial for accurate scene reconstruction from images. This advancement has potential applications in areas like virtual reality, augmented reality, and robotics, where precise 3D scene reconstruction is essential.
    
    \item In \textbf{Efficient View Synthesis with Neural Radiance Distribution Field}, the authors introduce a novel implicit representation called Neural Radiance Distribution Field (NeRDF). This method is designed to efficiently synthesize views of 3D scenes. NeRDF is unique in its approach to learning the radiance distribution along a given ray, which allows for the perception of spatial 3D geometry information. It combines the strengths of Neural Radiance Fields (NeRF) and Neural Light Fields (NeLF), requiring only a single network forwarding per ray, similar to NeLF, but with a smaller network size akin to NeRF. The key innovation lies in directly predicting the radiance distribution from the ray input, which is parameterized using trigonometric functions. The final pixel color is then re-synthesized via volume rendering. The authors also propose a knowledge distillation framework for training NeRDF, which includes novel designs like an input ray encoding method, an online view sampling strategy, and a volume density constraint loss. Their method demonstrates a significant improvement in rendering speed and efficiency, achieving high visual quality with lower memory costs, making it a promising technique for real-world applications in 3D scene rendering and view synthesis.
    
    \item In \textbf{SAMSNeRF (Segment Anything Model (SAM) Guides Dynamic Surgical Scene Reconstruction by NeRF)}, the authors introduce SAMSNeRF, a novel approach for reconstructing surgical scenes. This method combines a Segment Anything Model (SAM) with Neural Radiance Fields (NeRF) to achieve high-quality scene reconstruction. The core of SAMSNeRF is an 8-layer Multilayer Perceptron (MLP) neural network that represents a canonical radiance field and a time-dependent displacement field. These fields map 3D coordinates and viewpoint directions to RGB colors and space occupancy. The SAM, extensively trained on over 1 billion masks from 11 million images, excels in image segmentation tasks, including zero-shot segmentation with various prompts. This capability is particularly useful in medical or surgical images where additional fine-tuning is often not feasible. The authors also address depth refinement for both background tissues and foreground surgical tools, considering the challenges posed by specular reflection and fuzzy pixels in depth maps. This refinement process involves calculating the probabilistic distribution over residual maps and replacing certain depth pixels with smoother alternatives to enhance the depth accuracy for both the background and the surgical tools. The paper demonstrates the effectiveness of SAMSNeRF in reconstructing surgical scenes with high fidelity, leveraging the strengths of SAM in segmentation and NeRF in radiance field representation.
    
    \item In \textbf{Pose Modulated Avatars from Video}, , the authors introduce a novel approach to enhance the rendering of human figures in dynamic scenes using Neural Radiance Fields (NeRF). They focus on addressing the challenge of synthesizing fine geometric details, such as wrinkles, while avoiding artifacts in smoother regions. The key innovation lies in their frequency modulation technique, which leverages Sine functions as activations and adapts to pose dependencies. This is achieved through a graph neural network (GNN) that processes input poses to extract correlations between skeleton joints, encoding pose context. The method involves a two-branch neural network, with one branch explicitly operating in the frequency space. This allows for high-fidelity functional neural representations of human videos, with a particular emphasis on synthesizing high-frequency details in sharp regions and reducing artifacts near overlapping joints. The paper's contributions are significant in the realm of neural field applications for human avatar modeling, offering improvements over existing methods in rendering dynamic scenes with human figures.
    
    \item In \textbf{Blending-NeRF: Text-Driven Localized Editing in Neural Radiance Fields}, the authors introduce a novel approach for editing 3D objects using text descriptions. Their method, Blending-NeRF, is based on a layered Neural Radiance Field (NeRF) architecture that combines a pretrained NeRF with an editable NeRF. This setup allows for natural and precise editing of specific regions of 3D objects while maintaining their original appearance. The authors employ a pretrained vision-language method like CLIPSeg to specify the area to be modified in the text input workflow. The key contributions of this paper include the introduction of new blending operations for density addition, removal, and color alteration, enabling targeted and localized editing of 3D objects. The authors demonstrate the superiority of Blending-NeRF over previous methods through various experiments involving text-guided 3D object editing, such as shape and color modifications. This technique has potential applications in fields like virtual reality, game development, and digital art, where customized and high-fidelity 3D object manipulation is essential.
    
    \item In \textbf{ARF-Plus: Controlling Perceptual Factors in Artistic Radiance Fields for 3D Scene Stylization}, the authors introduce ARF-Plus, a framework designed to enhance the manipulation of perceptual factors in 3D stylization of radiance fields. This framework allows for the control of four key perceptual aspects: color preservation, stroke scale, selective spatial area, and depth-aware perception. The paper emphasizes the importance of these factors in achieving diverse and personalized 3D radiance fields style transfer, catering to various application scenarios. The authors build upon the Artistic Radiance Fields (ARF) approach, which effectively captures style characteristics while maintaining multi-view consistency. ARF-Plus extends this by offering user-level controllability, enabling individuals to customize controlled stylization according to their preferences. This includes retaining original colors while applying transferred patterns and textures, adjusting pattern sizes or brushstrokes, and selectively applying effects to specific areas or objects. The methodology involves reconstructing photo-realistic radiance fields from photographs of real-world scenes and then stylizing these fields with given style images. The process is formulated as an optimization problem, with images rendered from the radiance fields across different viewpoints. Various loss functions and gradient updating strategies are employed based on the selected perceptual controls. The result is a stylized 3D scene with free-viewpoint and consistent stylized renderings. This paper's contributions are significant in the field of 3D style transfer, particularly in enhancing user control over the stylization process, which opens up new possibilities for creative and practical applications in art, design, and virtual reality.
    
    \item \textbf{Relighting Neural Radiance Fields with Shadow and Highlight Hints} presents a novel neural implicit radiance representation for free viewpoint relighting of objects and scenes from a small set of unstructured photographs. The authors, Chong Zeng, Guojun Chen, Yue Dong, Pieter Peers, Hongzhi Wu, and Xin Tong, focus on digitally reproducing the appearance of real-world objects by considering complex light transport interactions. Unlike traditional methods that attempt to disentangle different light transport components, this approach does not separate the appearance into different components. Instead, it uses a data-driven approach with a multi-layer perceptron model to represent both local and global light transport. The model also incorporates shadow and highlight hints to aid in modeling high-frequency light transport effects. This technique is validated on synthetic and real scenes, demonstrating its effectiveness in a variety of shapes, material properties, and global illumination light transport scenarios. The paper contributes to the field of computer graphics and computer vision by offering a new method for relighting neural radiance fields, enhancing the realism and accuracy of digital object and scene representations.
    
    \item In \textbf{InsertNeRF: Instilling Generalizability into NeRF with HyperNet Modules}, the authors introduce a novel approach to enhance the generalizability of Neural Radiance Fields (NeRF) systems, including vanilla NeRF, mip-NeRF, and NeRF++ frameworks. The key innovation, named InsertNeRF, involves inserting multiple HyperNet modules into the NeRF framework. These modules are designed to leverage scene features effectively, enabling the system to adapt to various scenes without needing scene-specific retraining. InsertNeRF's architecture includes a multi-layer dynamic-static aggregation strategy, which not only utilizes global features for inherent completion capabilities but also models occlusion through dynamic-static weights. This approach significantly improves depth estimation and scene understanding. The paper demonstrates that InsertNeRF achieves state-of-the-art performance and extensive generalization capabilities across different NeRF-like systems and tasks with sparse inputs. The authors' contributions include the development of InsertNeRF with plug-and-play HyperNet modules, tailored HyperNet module structures for different NeRF attributes, and a novel aggregation strategy for sampling-aware scene features.
    
    \item In \textbf{Sparse3D: Distilling Multiview-Consistent Diffusion for Object Reconstruction from Sparse Views}, the authors introduce Sparse3D, a novel approach for reconstructing high-fidelity 3D objects from sparse and posed input views. This method is built upon two key components: a diffusion model that ensures multiview consistency and fidelity to user-provided input images, and a category-score distillation sampling (C-SDS) strategy. The diffusion model leverages the generalization capabilities of Stable Diffusion, and an epipolar controller guides it to generate novel-view images that are 3D consistent with the content of input images. The C-SDS strategy, inspired by VSD, addresses the issue of blurry and oversaturated outputs in NeRF reconstructions. The authors evaluate Sparse3D on the Common Object in 3D (CO3DV2) dataset, demonstrating its superiority over existing techniques in terms of the quality of synthesized novel views and reconstructed geometry, especially for object categories not present in the training domain. This approach opens new possibilities in 3D reconstruction, particularly in scenarios with limited view availability.
    
    \item \textbf{Unaligned 2D to 3D Translation with Conditional Vector-Quantized Code Diffusion using Transformers} by Abril Corona-Figueroa et al. addresses the challenging problem of generating 3D images from 2D views, particularly in scenarios with domain and geometric misalignments. Traditional methods like Generative Adversarial Networks and Neural Radiance Fields have limitations in this context. The authors propose a novel approach based on conditional diffusion with vector-quantized codes, which operates in a code space rich in information, enabling high-resolution 3D synthesis with full-coverage attention across views. This method generates 3D codes (e.g., for CT images) conditional on previously generated 3D codes and the entire codebook of two 2D views (e.g., 2D X-rays). The approach demonstrates state-of-the-art performance over specialized methods in varied evaluation criteria, including fidelity metrics such as density, coverage, and distortion metrics for complex volumetric imagery datasets in real-world scenarios. This advancement has significant implications for fields like clinical applications, virtual/augmented reality, self-driving vehicles, and security, where seamless translation between 2D and 3D imaging domains can overcome limitations of existing imaging devices.
    
    \item In \textbf{Multi-Modal Neural Radiance Field for Monocular Dense SLAM with a Light-Weight ToF Sensor}, the authors, Xinyang Liu, Yijin Li, Yanbin Teng, Hujun Bao, Guofeng Zhang, Yinda Zhang, and Zhaopeng Cui, introduce a novel dense SLAM (Simultaneous Localization and Mapping) system. This system uniquely combines a monocular camera with a lightweight Time-of-Flight (ToF) sensor, a cost-effective and compact solution commonly used in mobile devices for tasks like autofocus and obstacle detection. The authors propose a multi-modal implicit scene representation that can render signals from both the RGB camera and the lightweight ToF sensor. This approach optimizes the system by comparing with raw sensor inputs. The paper addresses the challenges posed by the sparse and noisy depth measurements of lightweight ToF sensors by employing a predicted depth as intermediate supervision and developing a coarse-to-fine optimization strategy. Additionally, the system explicitly utilizes temporal information to enhance the accuracy and robustness against the noisy signals from the ToF sensors. The proposed system demonstrates its effectiveness in camera tracking and dense scene reconstruction, showing promise for applications in augmented reality, indoor robotics, and other related fields.
    
    \item \textbf{Flexible Techniques for Differentiable Rendering with 3D Gaussians} by Leonid Keselman and Martial Hebert from Carnegie Mellon University explores advancements in shape reconstruction for computer vision applications. The authors focus on 3D Gaussians as an alternative to Neural Radiance Fields (NeRF) for photorealistic scene generation, addressing the high computational demands of NeRF. They extend previous work on differentiable rendering with 3D Gaussians, integrating features like differentiable optical flow, watertight mesh exporting, and per-ray normal rendering. The paper highlights the interoperability of two recent methods, Fuzzy Metaballs and 3D Gaussian Splatting, demonstrating their effectiveness in quick and robust reconstructions on both GPU and CPU. This research offers significant contributions to real-world applications in computer vision, particularly in areas requiring fast and realistic scene reconstruction.
    
    \item \textbf{CLNeRF: Continual Learning Meets NeRF} introduces an innovative approach to novel view synthesis, focusing on scenes that evolve over time in terms of appearance and geometry. The authors, Zhipeng Cai and Matthias Müller from Intel Labs, address the challenge of efficiently incorporating continuous changes in scenes, a task not covered by standard NeRF benchmarks. They propose a new dataset, World Across Time (WAT), featuring scenes with time-based changes, and develop CLNeRF, a method that integrates continual learning (CL) with Neural Radiance Fields (NeRFs). CLNeRF employs generative replay and the Instant Neural Graphics Primitives (NGP) architecture to prevent catastrophic forgetting and efficiently update the model with new data. The system also includes trainable appearance and geometry embeddings within NGP, enabling a single compact model to manage complex scene changes. Remarkably, CLNeRF achieves performance comparable to an upper bound model trained on all scans simultaneously, without the need to store historical images. This method outperforms other CL baselines across standard benchmarks and WAT, offering a significant advancement in the field of dynamic scene rendering.
    
    \item \textbf{Pose-Free Neural Radiance Fields via Implicit Pose Regularization} presents an innovative approach to training Neural Radiance Fields (NeRF) with unposed multi-view images. Traditional methods rely heavily on accurate camera poses, which are often challenging to obtain. This paper introduces IR-NeRF, a pose-free NeRF that enhances the robustness of pose estimation for real images through implicit pose regularization. The authors construct a scene codebook that stores scene features and captures the scene-specific pose distribution as priors. This method allows for more accurate reconstruction of real images, as it ensures that the estimated pose aligns with the actual pose distribution of the scene. The proposed approach shows significant improvements in novel view synthesis and outperforms existing methods across various synthetic and real datasets. This advancement has potential applications in 3D synthesis tasks where accurate camera poses are not readily available, offering a more flexible and robust solution for generating high-quality 3D scene representations from 2D images.
    
    \item \textbf{Canonical Factors for Hybrid Neural Fields} by Brent Yi, Weijia Zeng, Sam Buchanan, and Yi Ma, explores the biases introduced by factored feature volumes in building neural fields. The authors identify that these architectures, while offering compactness and efficiency, can be biased towards axis-aligned signals, affecting radiance field reconstruction. To address this, they propose learning canonicalizing transformations to remove these biases, resulting in more robust and efficient representations. Their approach, termed TILTED, is validated across various reconstruction tasks, demonstrating improvements in quality, robustness, compactness, and runtime. The paper highlights the potential of TILTED to match the capabilities of larger baselines while also pointing out weaknesses in current neural field evaluation methods.

    \item \textbf{Efficient Ray Sampling for Radiance Fields Reconstruction} by Shilei Sun, Ming Liu, Zhongyi Fan, Yuxue Liu, Chengwei Lv, Liquan Dong, and Lingqin Kong, focuses on enhancing the training efficiency of neural radiance fields (NeRF). The authors propose a novel ray sampling approach that targets improving training efficiency while maintaining high-quality photorealistic rendering. Their method analyzes the relationship between pixel loss distribution of sampled rays and rendering quality, identifying redundancy in NeRF's uniform ray sampling. By leveraging pixel regions and depth boundaries, the proposed sampling method reduces the number of rays in training views, ensuring each ray is more informative for scene fitting. This approach increases sampling probability in areas with significant color and depth variation, effectively reducing wasteful rays from other regions without compromising precision. The result is an accelerated network convergence and a more accurate perception of a scene's spatial geometry, particularly enhancing rendering outputs in texture-complex regions. The paper demonstrates that this method significantly outperforms existing techniques in public benchmarks. This advancement has practical implications in fields requiring efficient and high-quality 3D scene reconstruction, such as virtual reality, augmented reality, and digital content creation.
    
    \item \textbf{Drone-NeRF: Efficient NeRF Based 3D Scene Reconstruction for Large-Scale Drone Survey} introduces the Drone-NeRF framework, a novel approach to efficiently reconstruct large-scale, unbounded 3D scenes using Neural Radiance Fields (NeRF) tailored for drone oblique photography. The authors segmented the scene into uniform sub-blocks based on camera position and depth visibility, training sub-scenes in parallel with NeRF before merging them for a complete scene representation. This method optimizes camera poses and employs a uniform sampler to guide NeRF, integrating selected samples to enhance accuracy. A hash-coded fusion MLP is used to accelerate density representation, producing RGB and Depth outputs. The framework addresses challenges in scene complexity, rendering efficiency, and accuracy, particularly in drone-obtained imagery. This technique holds significant potential for applications in fields requiring extensive aerial surveys and 3D reconstructions, such as geography, environmental studies, and urban planning.
    
    \item \textbf{From Pixels to Portraits: A Comprehensive Survey of Talking Head Generation Techniques and Applications} by Shreyank N Gowda, Dheeraj Pandey, and Shashank Narayana Gowda provides an extensive overview of the latest methods in talking head generation. The authors categorize these methods into four primary approaches: image-driven, audio-driven, video-driven, and others, which include neural radiance fields (NeRF) and 3D-based methods. Each method is analyzed in detail, highlighting its unique contributions, strengths, and limitations. The paper also includes a comparative evaluation of publicly available models, focusing on aspects like inference time and human-rated quality of generated outputs. This survey aims to offer a clear and concise view of the current state of talking head generation, exploring the connections between different approaches and suggesting potential avenues for future research. This comprehensive survey serves as a valuable resource for researchers and practitioners in this rapidly advancing field.
    
    \item \textbf{Improving NeRF Quality by Progressive Camera Placement for Unrestricted Navigation in Complex Environments}, authored by Georgios Kopanas and George Drettakis, addresses the challenge of capturing more general scenes, such as rooms or buildings, for free-viewpoint navigation, particularly when there is no single central object of interest. The authors developed a metric to evaluate uniformity in space and angle, which is fast to evaluate. They proposed an algorithm using this metric to select the next best camera placement, aiming for a distribution closer to uniform in positions and angles. The method was evaluated on synthetic data and compared to baselines and previous work, showing promising results. Additionally, the algorithm was tested on a real dataset as a proof of concept. This approach can be beneficial for automated capture using robotics or drones, and it works with any NeRF model without changes to the optimization loop, introducing only a small performance overhead to the training. The solution is particularly advantageous in scenarios with a limited budget of cameras, achieving the best quality against multiple baselines and other algorithms.
    
    \item \textbf{SparseSat-NeRF: Dense Depth Supervised Neural Radiance Fields for Sparse Satellite Images} by Lulin Zhang and Ewelina Rupnik presents an innovative approach to digital surface model generation using Neural Radiance Fields (NeRF) adapted for sparse satellite views, named SparseSat-NeRF (SpS-NeRF). Traditional multi-view stereo matching (MVS) struggles with non-Lambertian surfaces, asynchronous acquisitions, or at discontinuities, whereas NeRF offers a continuous volumetric representation for reconstructing surface geometries. However, NeRF typically requires many views, which is uncommon in earth observation satellite imaging. SpS-NeRF overcomes this by employing dense depth supervision guided by cross-correlation similarity metrics provided by traditional semi-global MVS matching. The authors demonstrate the effectiveness of SpS-NeRF on stereo and tri-stereo Pléiades 1B/WorldView-3 images, showcasing its superiority in rendering sharper novel views and more reliable 3D geometries compared to NeRF and Sat-NeRF. This method holds significant potential for applications in urban planning, environmental monitoring, geology, and disaster rapid mapping, where high-quality digital surface models are crucial.
    
    \item \textbf{Adv3D: Generating 3D Adversarial Examples in Driving Scenarios with NeRF} by Leheng Li, Qing Lian, and Ying-Cong Chen explores the vulnerability of Deep Neural Networks (DNNs) to adversarial examples, particularly in the context of autonomous driving stacks like 3D object detection. The authors introduce Adv3D, a novel approach that models adversarial examples as Neural Radiance Fields (NeRFs). This method leverages NeRF's capabilities for photorealistic appearance and accurate 3D generation to create more realistic and physically realizable adversarial examples. Adv3D trains adversarial NeRF by minimizing the confidence of surrounding objects predicted by 3D detectors. The effectiveness of Adv3D is demonstrated through its ability to significantly reduce performance when rendering NeRF in various poses. The paper also introduces techniques like primitive-aware sampling and semantic-guided regularization for generating 3D patch attacks with camouflaged adversarial textures. These adversarial examples generalize well across different poses, scenes, and 3D detectors. Additionally, the authors propose a defense method involving adversarial training through data augmentation, addressing the safety-critical nature of self-driving cars. The research presents a significant step in understanding and improving the robustness of 3D detectors against adversarial attacks in autonomous driving scenarios.
    
    \item \textbf{Instant Continual Learning of Neural Radiance Fields} addresses the challenge of catastrophic forgetting in neural radiance fields (NeRFs) when trained in a continual learning setting. The authors propose a novel approach that leverages replay-based techniques, acknowledging that a trained NeRF is a compressed representation of all previously observed 2D views. By freezing a copy of the scene representation after each task's training, the method provides access to pseudo ground truth RGB values for all previously seen data. The paper introduces a hybrid implicit-explicit representation, replacing the frequency encoding in NeRF with a multi-resolution hash encoding, which significantly reduces the size of the decoder multilayer perceptron (MLP) and minimizes the effects of catastrophic forgetting. This approach not only preserves the quality of 3D scene reconstruction but also offers a faster alternative to previous methods, enabling quick learning of additional 3D scene content from new input views. The paper contributes to the fields of 3D scene reconstruction and continual learning, offering a solution that maintains high-quality scene representation while addressing the limitations of NeRFs in sequential data acquisition scenarios.
    
    \item \textbf{ResFields: Residual Neural Fields for Spatiotemporal Signals} by Marko Mihajlovic et al. introduces a novel approach to enhance the representation of complex temporal signals in neural fields. The authors address the limitations of Multi-layer Perceptrons (MLPs) in modeling large and intricate temporal signals by incorporating temporal residual layers into neural fields, creating a new class of networks called ResFields. This innovation allows for effective representation of dynamic signals and reduces the number of trainable parameters, thereby improving generalization capabilities. The paper demonstrates the utility of ResFields across various challenging tasks, including 2D video approximation, dynamic shape modeling via temporal Signed Distance Functions (SDFs), and dynamic Neural Radiance Fields (NeRF) reconstruction. The authors also showcase the practical application of ResFields in capturing dynamic 3D scenes using sparse RGBD cameras from a lightweight capture system. This research represents a significant advancement in the field of neural networks, particularly in the context of spatiotemporal signal processing.
    
    \item \textbf{Bayes' Rays: Uncertainty Quantification for Neural Radiance Fields} introduces BayesRays, a novel post-hoc algorithm designed to estimate the spatial uncertainty of any pre-trained Neural Radiance Field (NeRF) without requiring additional training or changes to the existing architecture. Developed by a team from the University of Toronto, Simon Fraser University, Google DeepMind, and Adobe Research, this method addresses the inherent uncertainties in learning from multiview images, a common challenge in applications like view synthesis and depth estimation. BayesRays operates by establishing a volumetric uncertainty field using spatial perturbations and a Bayesian Laplace approximation. The authors statistically derive their algorithm and demonstrate its superior performance in key metrics and applications, offering a significant advancement in the field of NeRF and its practical applications in areas such as autonomous driving and outlier detection.
    
    \item \textbf{Text2Control3D: Controllable 3D Avatar Generation in Neural Radiance Fields using Geometry-Guided Text-to-Image Diffusion Model} introduces a novel method for generating 3D facial avatars conditioned by text descriptions and geometric control factors extracted from monocular video captures. The authors focus on controlling facial expressions and shapes, key aspects of creating faithful virtual agents. Their approach involves generating viewpoint-aware images of an avatar using a geometry-conditional text-to-image diffusion model, then constructing the avatar in Neural Radiance Fields (NeRF). They address challenges such as maintaining consistent appearance and expression across images, solving the texture-sticking problem, and constructing 3D avatars from viewpoint-aware images with slight geometric inconsistencies. The paper contributes to the field by being the first to offer controllable text-to-3D avatar generation, proposing a zero-shot method for generating consistent avatar images, ameliorating texture-sticking issues, and reconstructing high-fidelity 3D avatars from generated images. This work has significant implications for enhancing virtual interactions and creating more personalized and expressive digital representations in various real-world applications.
    
    \item \textbf{SimNP: Learning Self-Similarity Priors Between Neural Points} presents a novel approach to 3D reconstruction that combines the strengths of category-level data priors and local representation with test-time optimization. The authors introduce SimNP, a method that learns how information can be shared between local object regions, enabling the propagation of information from visible to invisible parts during inference. This is achieved by learning characteristic self-similarity patterns from training data. The key innovation lies in using neural point representations to model relationships between local regions of objects, which are well-suited for capturing high-frequency patterns and allowing explicit formulations of similarities. The paper's contributions include presenting the first generalizable neural point radiance field for object category representation, a mechanism for learning general similarities between local object regions, and demonstrating improved reconstruction of unobserved regions from a single image. This method outperforms existing two-view methods and is more efficient in training and rendering. The approach addresses the challenge of combining detailed reconstruction of visible regions with a model that can infer unseen regions, leveraging structured self-similarities and symmetries in objects.
    
    \item \textbf{BluNF: Blueprint Neural Field} introduces a novel approach for generating a semantic editable blueprint from multiple viewpoints of a scene. This is achieved by leveraging prior semantic and depth information through the Blueprint Neural Field module, BluNF. BluNF generates a 2D semantic-aware editable blueprint that is robust to noise and sparse observations, enabling user-friendly editing of NeRF (Neural Radiance Fields) representations. This is the first instance of using an implicit neural field to construct a 2D semantic-aware blueprint for editing purposes, diverging from traditional methods that rely on CNNs or transformers. The key contributions of this paper include BluNF, which generates a blueprint of a scene through an implicit neural field without explicit geometric constraints or supervision; an underlying representation that is robust to sparse and noisy observations, outperforming multi-view-based or classical NeRF methods; and the combination of BluNF with NeRF representations of the same scene, enabling intuitive user manipulations on the generated 2D blueprint. These manipulations include masking, appearance changes, and object removal, which can be directly incorporated into NeRF view rendering. This approach offers a novel and intuitive method for editing NeRF representations, enhancing the usability and flexibility of 3D scene manipulation.
    
    \item \textbf{SimpleNeRF: Regularizing Sparse Input Neural Radiance Fields with Simpler Solutions} authored by Nagabhushan Somraj, Adithyan Karanayil, and Rajiv Soundararajan, addresses the challenge of training Neural Radiance Fields (NeRFs) with a sparse set of input images. The authors focus on novel regularizations for effective training in such scenarios. They classify prior work on sparse input NeRFs into generalized models and regularization-based models, noting the limitations of each approach. Their work aims to regularize NeRFs by learning augmented models for depth supervision in tandem with NeRF training. They simplify the capability of augmented NeRF models concerning positional encoding and view-dependent radiance, aiming to mitigate common distortions like floaters, shape-radiance ambiguity, and blurred renders in few-shot settings. The depth estimated by these simpler models is used to supervise the depth estimated by the NeRF model. The authors' approach can be seen as a semi-supervised learning model, using sparse depth from a Structure from Motion module and imposing consistency loss between depths estimated by different components of the NeRF. This research has potential applications in 3D imaging and reconstruction, particularly in scenarios with limited input data, offering a more efficient and reliable approach to synthesizing novel views of a scene.
    
    \item \textbf{DeformToon3D: Deformable 3D Toonification from Neural Radiance Fields} introduces an innovative approach for 3D-aware artistic toonification, which is the process of applying the style of artistic domains onto a target 3D face with stylized geometry and texture. This technique is particularly useful in various applications like comics, animation, virtual reality, and augmented reality, where it can facilitate quick, high-quality 3D avatar creation based on photographs in the style of designated artwork. The authors propose a novel method called DEFORMTOON3D, which decomposes the stylization of geometry and texture into more manageable subproblems. This method introduces a StyleField on top of a pre-trained 3D generator to deform each point in the style space to the pre-trained real space, guided by an instance code. This allows for easy extension to multiple styles with a single stylization field. For texture stylization, the method employs adaptive style mixing, injecting artistic domain information into the network for effective texture toonification. The approach is trained using synthetic paired data, which includes realistic faces generated by a pre-trained 3D GAN and corresponding stylized data generated by a 2D toonification model. Their method achieves high-quality geometry and texture toonification over a variety of styles while preserving the original GAN latent space. This preservation enables compatibility with existing tools built on the real face space GAN, including inversion, editing, and animation. Additionally, the method significantly reduces the storage footprint by requiring only a small stylization field with a set of AdaIN parameters for artistic domain stylization. The authors' contributions include the novel StyleField for efficient 3D shape modeling, an approach for multi-style toonification with a single model, and a full synthetic data-driven training pipeline.
    
    \item \textbf{Dynamic Mesh-Aware Radiance Fields} presents a novel approach to integrating polygonal mesh assets within photorealistic Neural Radiance Fields (NeRF) volumes. This integration allows for realistic rendering and simulation of dynamics in a physically consistent manner with the NeRF. The authors review light transport equations for both mesh and NeRF, distilling them into an efficient algorithm for updating radiance and throughput along a ray with multiple bounces. They address the discrepancy between linear and sRGB color spaces by training NeRF with High Dynamic Range (HDR) images and propose a strategy for estimating light sources and casting shadows on the NeRF. The paper also discusses the integration of this hybrid surface-volumetric formulation with a high-performance physics simulator supporting cloth, rigid, and soft bodies. This system, capable of running on a GPU at interactive rates, offers enhanced visual realism for mesh insertion in dynamic scenes, particularly affecting the appearance of reflective/refractive surfaces and the illumination of diffuse surfaces. The authors envision applications in VR/AR, interactive gaming, virtual tourism, education, training, and computer animation.
    
    \item \textbf{Mirror-Aware Neural Humans} introduces a novel approach to human motion capture using a single camera and a mirror, overcoming the limitations of multi-camera systems and single-view inputs. The authors, Daniel Ajisafe, James Tang, Shih-Yang Su, Bastian Wandt, and Helge Rhodin, propose a test-time optimization method for reconstructing a generative body model, including pose, shape, and appearance. This method is unique in its use of mirrors to provide a second view for accurate reconstruction, eliminating the need for multi-camera recording and synchronization. The paper demonstrates the practicality of this approach, highlighting the widespread availability of mirrors in urban environments and the ease of annotating diverse training images. The authors' contributions extend articulated neural radiance fields to include a notion of a mirror, making it sample-efficient over potential occlusion regions. This results in a consumer-level 3D motion capture system that starts from off-the-shelf 2D poses, automatically calibrates the camera, estimates mirror orientation, and lifts 2D keypoint detections to 3D skeleton pose. The system is shown to be more accurate than existing methods in challenging mirror scenes, both in terms of 3D pose metrics and image quality. The project is available at \url{https://danielajisafe.github.io/mirror-aware-neural-humans/}.
    
    \item \textbf{VeRi3D: Generative Vertex-based Radiance Fields for 3D Controllable Human Image Synthesis} introduces VeRi3D, a generative human vertex-based radiance field parameterized by vertices of the parametric human template, SMPL. The authors, Xinya Chen, Jiaxin Huang, Yanrui Bin, Lu Yu, and Yiyi Liao, focus on addressing the limitations in generalization ability and controllability of unsupervised learning of 3D-aware generative adversarial networks, particularly in human generative models. VeRi3D maps each 3D point to a local coordinate system defined on its neighboring vertices, using the corresponding vertex feature and local coordinates to map it to color and density values. This approach enables the generation of photorealistic human images with free control over camera pose, human pose, shape, and part-level editing. The paper demonstrates the effectiveness of VeRi3D in creating diverse, realistic renderings of clothed humans, which has significant applications in various fields. The project page for further details is available at XDimlab.
    
    \item \textbf{SC-NeRF: Self-Correcting Neural Radiance Field with Sparse Views} introduces a novel end-to-end network for synthesizing realistic images from sparse input views. The authors propose a geometry correction module based on multi-head attention to address black artifacts in rendered views caused by inconsistencies in scale and structure between training and testing scenes. Additionally, an appearance correction module is designed to alleviate boundary blank and ghosting artifacts in rendered views due to large viewpoint changes. The paper validates the model's effectiveness on four datasets: Blender, LLFF, DTU, and Spaces, with notable performance on outdoor scenes in the Spaces dataset, outperforming MVSNeRF and IBRNet in terms of PSNR. The approach is completely differentiable and can be trained end-to-end from sparse view inputs. The authors also discuss related work in Novel View Synthesis via NeRF, Multi-View Stereo, and the use of Transformer in NeRF, highlighting their method's unique use of transformers to correct geometric and appearance features for better generalization to different scenes.
    
    \item \textbf{Federated Learning for Large-Scale Scene Modeling with Neural Radiance Fields} presents a novel approach to large-scale scene modeling using Neural Radiance Fields (NeRF) in a federated learning setting. The authors propose a method that caches the outputs of each local model as a voxel grid, enabling local updates of the global model. This approach addresses the challenges of maintainability and computational costs associated with existing large-scale modeling methods. The paper also introduces a global pose alignment technique to alleviate sensor noise in clients' poses in the global coordinate. This alignment minimizes the difference between images rendered by global and local models. The effectiveness of the training pipeline is assessed on the Mill19 dataset, which contains thousands of high-definition images collected from drone footage over a large area near an industrial complex. The paper discusses the application of NeRF to robotics, particularly in SLAM systems, and highlights the advantages of federated learning, such as privacy preservation and leveraging computational resources of many clients. The authors' approach aims to improve communication efficiency and mitigate the challenges of asynchronous federated learning, making it suitable for large-scale applications like self-driving cars and delivery drones.
    
    \item \textbf{DisEntangLed avaTAr (DELTA): Disentangled Avatar with Hybrid 3D Representation} introduces a novel approach for creating photorealistic human avatars by combining explicit and implicit 3D representations. The authors, Yao Feng, Weiyang Liu, Timo Bolkart, Jinlong Yang, Marc Pollefeys, and Michael J. Black, propose DELTA, which models the face and body using explicit triangular meshes, while hair and clothing are represented with an implicit neural radiance field (NeRF). This hybrid approach leverages the strengths of both representations: meshes are efficient for faces and minimally clothed bodies, while NeRF excels in rendering complex structures like hair and clothing. DELTA is distinctive for its mesh-integrated volumetric renderer, enabling end-to-end differentiable learning from monocular videos without 3D supervision. The method shows promise in creating realistic avatars that can be easily reposed and used in various environments, highlighting the potential of hybrid 3D representation in avatar modeling.
    
    \item \textbf{Dynamic NeRFs for Soccer Scenes} explores the application of dynamic Neural Radiance Fields (NeRFs) for synthesizing soccer replays. The authors focus on the challenges of reconstructing soccer scenes using arrays of distant static cameras, considering the large static environment of a stadium and the small dynamic elements like players and the ball. They build upon modern deep learning-based approaches to computer vision problems, specifically dynamic NeRFs, which are neural models designed to reconstruct spatiotemporal content. The paper makes three key assumptions: using a static multi-camera setup similar to existing proprietary systems, limiting the study to synthetic soccer datasets, and considering general dynamic NeRFs without domain-specific knowledge. The authors compare recent state-of-the-art general dynamic NeRF models in three synthetic soccer environments of increasing complexity. Their contributions include a study of the performance of these models in different environments and the release of their code, dataset, and experimental settings to encourage further research in this area. The paper is organized into sections covering preliminaries about NeRFs, experimental setup, results, and a higher-level discussion on the feasibility and potential improvements of using these methods.

    \item \textbf{TECA (Text-Guided Generation and Editing of Compositional 3D Avatars)} introduces a novel approach for creating and editing realistic 3D face avatars with accessories and hair using text descriptions. The authors adopt a compositional method, leveraging neural and mesh-based 3D content creation techniques. They model avatars as a combination of face/body and non-face/body regions, using the SMPL-X body model for the head and shoulders, and Neural Radiance Fields (NeRF) for non-face components like hair and accessories. The process involves generating an image from a text description using a stable diffusion model, optimizing the shape parameters of the SMPL-X body model, and employing a sequential, compositional approach for additional style components. The method allows for the transfer of non-face components between avatars and uses segmentation to focus NeRF on specific regions like hair. The paper demonstrates the realism of the generated avatars through extensive qualitative analysis, highlighting the advantages of using meshes for face and body and NeRF for hair and clothing. This approach enables diverse applications, including virtual try-on, by supporting the editing of individual components and the seamless transfer of features between avatars.

    \item \textbf{Indoor Scene Reconstruction with Fine-Grained Details Using Hybrid Representation and Normal Prior Enhancement} addresses the challenge of reconstructing indoor scenes, which often contain both flat, low-frequency areas and high-frequency, fine-grained regions. Traditional methods using a single Multilayer Perceptron (MLP) struggle to express these complex structures effectively. To overcome this, the authors propose a novel hybrid representation that combines an MLP network for outlining the scene and a tri-plane feature branch with a shallow decoder for representing fine details. This approach is inspired by the success of tri-plane representation in encoding human faces, offering a memory-efficient alternative to voxel grids. The paper also introduces an image enhancement technique and an uncertainty estimation module to improve the quality of predicted geometric priors. These priors are crucial for guiding the network in leveraging more accurate information for reconstruction. The uncertainty module predicts pixel-wise uncertainty maps of the estimated normal priors, reducing the impact of inaccuracies on reconstruction quality. The authors' contributions include the hybrid implicit SDF architecture, the image enhancement technique, and the uncertainty estimation module. Their approach shows significant improvements over state-of-the-art methods in both qualitative and quantitative experiments, demonstrating effective generalization to real-world indoor scenarios.

    \item \textbf{CoRF: Colorizing Radiance Fields using Knowledge Distillation} introduces a novel approach, CoRF, for colorizing radiance field networks to produce 3D consistent colorized novel views from input grey-scale multi-view images. The authors propose a distillation-based method that leverages existing deep image colorization methods without incurring additional training costs for a separate colorization module. The training process is divided into two stages: training a radiance field network on grey-scale multi-view images, followed by distilling knowledge from a teacher colorization network into the trained radiance field network. Additionally, a multi-scale self-regularization technique is employed to mitigate spatial color inconsistencies. The effectiveness of this approach is demonstrated on various grey-scale image sequences from existing datasets, such as LLFF and Shiny, and on two real-world applications: colorizing multi-view IR images and grey-scale legacy content. The paper contributes to the field by offering a method for 3D consistent colorization of radiance fields, applicable in diverse real-world scenarios.
    
    \item \textbf{DT-NeRF: Decomposed Triplane-Hash Neural Radiance Fields for High-Fidelity Talking Portrait Synthesis} by Yaoyu Su, Shaohui Wang, and Haoqian Wang introduces a novel framework called DT-NeRF, which significantly enhances the photorealistic rendering of talking faces. The authors, affiliated with the Shenzhen International Graduate School, Tsinghua University, focus on overcoming challenges in audio-driven facial synthesis, particularly in 3D visual settings. Their approach involves decomposing the facial region into two specialized triplanes, one for the mouth and the other for broader facial features, and integrating audio features as query vectors through an audio-mouth-face transformer. This method leverages Neural Radiance Fields (NeRF) to enrich the volumetric representation of the entire face. The paper demonstrates the effectiveness of DT-NeRF in achieving state-of-the-art results on key evaluation datasets, addressing the challenges of enhanced mouth and facial representation and effective coupling of audio signals with facial dynamics.

    \item \textbf{MC-NeRF: Muti-Camera Neural Radiance Fields for Muti-Camera Image Acquisition Systems} by Yu Gao et al. introduces MC-NeRF, a method designed to optimize both intrinsic and extrinsic parameters for bundle-adjusting Neural Radiance Fields. This approach addresses challenges in multi-camera systems, such as varying intrinsic parameters and frequent pose changes. The authors provide a theoretical analysis to tackle degenerate cases and coupling issues arising from joint optimization of these parameters. They also propose an efficient calibration image acquisition scheme for multi-camera systems, including the design of a calibration object. The paper presents a global end-to-end network with a training sequence that enables the regression of intrinsic and extrinsic parameters, along with the rendering network. This method is particularly effective in scenarios where each image corresponds to different camera parameters, allowing for 3D scene representation without initial poses. The research is significant for applications in autonomous driving, multi-robot navigation, and security surveillance, where understanding the environment through visual 3D reconstruction is crucial.

    \item \textbf{Gradient based Grasp Pose Optimization on a NeRF that Approximates Grasp Success} by Gergely Sóti, Björn Hein, and Christian Wurll introduces a novel approach for robotic grasping using Neural Radiance Fields (NeRFs). The authors, affiliated with Hochschule Karlsruhe – University of Applied Sciences and Karlsruhe Institute of Technology, Germany, propose a method that directly maps gripper poses to corresponding grasp success values without considering objectness. Their technique leverages a NeRF architecture to learn a scene representation and train a grasp success estimator. This estimator maps each pose in the robot's task space to a grasp success value, which is then optimized using gradient-based optimization to obtain successful grasp poses. This approach uniquely avoids the need for rendering and the limitations of discretization, common in other NeRF-based methods. Demonstrated on four simulated 3DoF robotic grasping tasks, the method shows promising results, including the ability to generalize to novel objects and achieve an average translation error of 3mm from valid grasp poses. This research, funded by the German Federal Ministry of Education and Research under the KI5GRob project and opens new possibilities for applying this approach to higher DoF grasps and real-world scenarios.
    
    \item \textbf{Deformable Neural Radiance Fields using RGB and Event Cameras} by Qi Ma, Danda Pani Paudel, Ajad Chhatkuli, and Luc Van Gool presents a novel approach to model fast-moving deformable objects using neural radiance fields (NeRFs) from visual data. The authors address the challenge of high deformation and low acquisition rates by incorporating event cameras, which capture visual changes rapidly and asynchronously. Their method synergizes asynchronous event streams with calibrated sparse RGB frames, optimizing camera poses and the radiance field jointly. This efficient process leverages event collections actively during learning. The method's effectiveness is demonstrated through experiments on both realistically rendered graphics and real-world datasets, showing significant advantages over existing methods and baselines. This research opens new avenues for modeling dynamic scenes with deformable neural radiance fields in real-world applications. The authors have made their code available at: \url{https://qimaqi.github.io/DE-NeRF.github.io/}.
    
    \item \textbf{Breathing New Life into 3D Assets with Generative Repainting} by Tianfu Wang, Menelaos Kanakis, Konrad Schindler, Luc Van Gool, and Anton Obukhov introduces a novel pipeline for text-guided painting of legacy 3D geometry. The authors leverage pretrained generative 2D diffusion models and neural radiance fields (NeRF) to refresh existing 3D assets while ensuring 3D consistency and addressing issues in legacy representations. This approach accepts any legacy renderable geometry, orchestrates the interaction between 2D generative refinement and 3D consistency enforcement tools, and outputs a painted input geometry in various formats. The pipeline's modularity allows for easy partial upgrades, a crucial feature in the fast-paced domain of digital media. The authors demonstrate the advantages of their method through a large-scale study on a wide range of objects and categories from the ShapeNetSem dataset, showcasing both qualitative and quantitative improvements. This work represents a significant advancement in digital media and offers artists new possibilities for creating high-quality 3D assets based on textual descriptions.
    
    \item \textbf{Robust e-NeRF: NeRF from Sparse \& Noisy Events under Non-Uniform Motion} by Weng Fei Low and Gim Hee Lee presents a novel method for reconstructing Neural Radiance Fields (NeRF) from moving event cameras, particularly under challenging conditions of sparse and noisy events generated by non-uniform motion. The authors address the limitations of previous works that relied on dense, low-noise event streams and uniform camera motion. Their approach, Robust e-NeRF, effectively handles various real-world scenarios by incorporating a realistic event generation model that accounts for intrinsic parameters and non-idealities, such as time-independent, asymmetric threshold, and refractory period, as well as pixel-to-pixel threshold variation. Additionally, the method utilizes a pair of normalized reconstruction losses that generalize to arbitrary speed profiles and intrinsic parameter values without requiring prior knowledge. This work demonstrates significant advancements in NeRF reconstruction from event cameras, verified through experiments on real and realistically simulated sequences. The authors also contribute their code, synthetic dataset, and improved event simulator to the public domain, enhancing the potential for applications in areas like augmented reality and autonomous navigation.
    
    \item \textbf{DynaMoN: Motion-Aware Fast And Robust Camera Localization for Dynamic NeRF} by Mert Asim Karaoglu, Hannah Schieber, Nicolas Schischka, Melih Gorgulu, Florian Grötzer, Alexander Ladikos, Daniel Roth, Nassir Navab, and Benjamin Busam, introduces a novel approach to camera localization in dynamic scenes using Neural Radiance Fields (NeRF). The authors present DynaMoN, a system that combines dynamic camera localization with a dynamic NeRF for high-quality novel view synthesis. This approach leverages motion and semantic segmentation for robust camera tracking in dynamic environments. The system significantly outperforms existing methods in terms of speed and robustness, particularly in dynamic scenes. DynaMoN's effectiveness is demonstrated through extensive validation on challenging datasets like TUM RGB-D, BONN RGB-D Dynamic, and DyCheck’s iPhone dataset. This work represents a significant advancement in dynamic scene understanding and visualization, with potential applications in augmented reality, robotics, and autonomous navigation.
    
    \item \textbf{NeRF-VINS: A Real-time Neural Radiance Field Map-based Visual-Inertial Navigation System} by Saimouli Katragadda, Woosik Lee, Yuxiang Peng, Patrick Geneva, Chuchu Chen, Chao Guo, Mingyang Li, and Guoquan Huang, presents a pioneering approach to enhancing localization in robotics and computer vision. The paper introduces NeRF-VINS, a real-time, tightly-coupled system that integrates Neural Radiance Fields (NeRF) with a visual-inertial navigation system (VINS). This integration leverages NeRF's ability to synthesize novel views, crucial for overcoming limited viewpoints, and optimally fuses IMU and monocular image measurements with synthetically rendered images within an efficient filter-based framework. The result is a 3D motion tracking system with bounded error. The authors demonstrate NeRF-VINS's superior performance compared to state-of-the-art methods using prior map information, achieving real-time estimation at 15 Hz on a Jetson AGX Orin embedded platform with impressive accuracy. This work marks a significant advancement in achieving accurate, efficient, and consistent localization within an a priori environment map, with potential applications in AR/VR, consumer drones, and autonomous navigation.
    
    \item \textbf{Instant Photorealistic Style Transfer: A Lightweight and Adaptive Approach} by Rong Liu, Enyu Zhao, Zhiyuan Liu, Andrew Feng, and Scott John Easley from the University of Southern California and USC Institute for Creative Technologies, introduces a groundbreaking method for instant photorealistic style transfer, particularly suited for high-resolution inputs. This approach, named Instant Photorealistic Style Transfer (IPST), is designed to perform style transfer from a style image to a content image without the need for pre-training on pair-wise datasets or imposing extra constraints. The authors utilize a lightweight StyleNet that preserves non-color information during the transfer process. Additionally, they introduce an instance-adaptive optimization strategy to prioritize photorealism and accelerate convergence, enabling rapid training completion within seconds. The paper highlights IPST's capability to maintain temporal and multi-view consistency in multi-frame inputs like video and Neural Radiance Field (NeRF). The experimental results show that IPST requires less GPU memory, offers faster multi-frame transfer speed, and produces photorealistic outputs, making it a promising solution for various photorealistic transfer applications in fields like cinema and extended reality.
    
    \item \textbf{Steganography for Neural Radiance Fields by Backdooring} by Weina Dong, Jia Liu, Yan Ke, Lifeng Chen, Wenquan Sun, and Xiaozhong Pan, introduces a novel steganography approach using Neural Radiance Fields (NeRF). This method leverages the viewpoint synthesis capabilities of NeRF to embed secret messages. The authors propose using a secret viewpoint as a key, where the NeRF model generates a secret viewpoint image that acts as a backdoor. A message extractor is then trained through overfitting to create a one-to-one mapping between the secret message and the secret viewpoint image. The trained NeRF model and message extractor are delivered to the receiver over an open channel. The receiver, using the shared key, can obtain the rendered image in the secret view from the NeRF model and extract the secret message through the message extractor. This technique ensures security due to the complexity of viewpoint information, making it difficult for attackers to accurately steal the secret message. The paper demonstrates high-capacity steganography with fast performance and 100
    
    \item \textbf{Locally Stylized Neural Radiance Fields} by Hong-Wing Pang, Binh-Son Hua, and Sai-Kit Yeung from Hong Kong University of Science and Technology, Trinity College Dublin, and VinAI Research, Vietnam, presents a novel approach to applying stylization to 3D scenes using neural radiance fields (NeRF). This work addresses the challenge of transferring patterns from a style image onto different parts of a NeRF scene, ensuring appearance consistency across novel views. The authors propose a framework based on local style transfer, utilizing a hash-grid encoding to learn the embedding of appearance and geometry components. The hash table mapping allows for controlled stylization by optimizing the appearance branch while keeping the geometry branch fixed. A new loss function is introduced, employing a segmentation network and bipartite matching to establish region correspondences between the style image and content images from volume rendering. The paper demonstrates that this method achieves plausible stylization results with novel view synthesis, offering flexible controllability through manipulation and customization of region correspondences. This advancement in 3D style transfer has significant implications for applications in gaming, movies, and extended reality.
    
    \item \textbf{SpikingNeRF: Making Bio-inspired Neural Networks See through the Real World} by Xingting Yao, Qinghao Hu, Tielong Liu, Zitao Mo, Zeyu Zhu, Zhengyang Zhuge, and Jian Cheng from the Institute of Automation, Chinese Academy of Sciences, and the School of Future Technology, University of Chinese Academy of Sciences, introduces SpikingNeRF, a novel approach that integrates Spiking Neural Networks (SNNs) with Neural Radiance Fields (NeRF) for energy-efficient 3D scene reconstruction. This paper addresses the high energy consumption of traditional NeRF rendering by proposing a bio-inspired method that aligns the radiance ray with the temporal dimension of SNNs, enabling spike-based, multiplication-free computation. The method involves matching each sampled point on the ray to a specific time step and representing it in a hybrid manner with voxel grids. The voxel grids help determine whether sampled points should be masked for improved training and inference. To handle the irregular temporal length caused by this masking, the authors propose a temporal padding strategy for maintaining regular tensor lengths and a temporal condensing strategy for denser data structures, facilitating hardware-friendly computation. The paper demonstrates significant energy savings (70.79\% on average) while maintaining comparable synthesis quality to the ANN baseline, showcasing SpikingNeRF's potential in real-world 3D reconstruction tasks.

    \item \textbf{Controllable Dynamic Appearance for Neural 3D Portraits} by ShahRukh Athar, Zhixin Shu, Zexiang Xu, Fujun Luan, Sai Bi, Kalyan Sunkavalli, and Dimitris Samaras from Stony Brook University and Adobe Research, introduces CoDyNeRF, a system for creating fully controllable 3D portraits in real-world capture conditions. This work addresses the challenge of maintaining photometric consistency in dynamic portrait scenes reconstructed and reanimated using Neural Radiance Fields (NeRFs), where traditional models struggle with artifacts due to uneven lighting during head-pose and facial expression changes. CoDyNeRF overcomes this by learning to approximate illumination-dependent effects through a dynamic appearance model in the canonical space, conditioned on predicted surface normals and deformations from facial expressions and head-pose. The method utilizes 3DMM normals as a coarse prior for the normals of the human head, aiding in the prediction of surface normals that are otherwise difficult due to rigid and non-rigid deformations. The paper demonstrates the effectiveness of CoDyNeRF in free view synthesis of portrait scenes with explicit head pose and expression controls, and realistic lighting effects, using only a short smartphone-captured video for training. This advancement in 3D portrait generation holds significant potential for applications in virtual reality, gaming, and digital content creation.
    
    \item \textbf{Language-driven Object Fusion into Neural Radiance Fields with Pose-Conditioned Dataset Updates} by Ka Chun Shum, Jaeyeon Kim, Binh-Son Hua, Duc Thanh Nguyen, and Sai-Kit Yeung from Hong Kong University of Science and Technology, Trinity College Dublin, and Deakin University, presents a pioneering approach to editing 3D scenes using neural radiance fields (NeRFs). The paper addresses the challenge of inserting or removing objects in 3D scenes, a task traditionally reliant on manual methods in computer graphics. The authors introduce a language-driven method that leverages text-to-image diffusion models to synthesize new images combining a given object and background. These images are then used to refine the background radiance field, learning new object geometry and appearance. A key innovation is the dataset updates strategy, which controls the refinement process to gradually introduce the foreground object, starting from a randomly selected view and prioritizing views close to already-trained camera views. This method ensures view-consistent rendering and reduces artifacts, significantly enhancing the integration of objects into 3D scenes. The paper demonstrates the method's effectiveness in generating photorealistic images of edited scenes, outperforming existing methods in 3D reconstruction and neural radiance field blending.
    
    \item \textbf{Rendering stable features improves sampling-based localisation with Neural radiance fields} by Boxuan Zhang, Lindsay Kleeman, and Michael Burke from Monash University, Australia, addresses the computational challenges in object and scene-based localization using Neural Radiance Fields (NeRFs). The paper focuses on the inefficiency of sampling-based or Monte-Carlo localization schemes, which are computationally expensive due to multiple network forward passes required for inferring camera or object pose. The authors propose a solution that involves keypoint recognition techniques from classical computer vision, leading to a significant reduction in the number of forward passes needed. Their systematic empirical comparison of various approaches reveals that rendering stable features, as opposed to conventional feature matching approaches, can result in a tenfold reduction in forward passes, thereby greatly improving speed. This work contributes to the field of robotics, particularly in enhancing the efficiency of localization techniques using NeRFs.
    
    \item \textbf{MarkNerf: Watermarking for Neural Radiance Field} presents a novel watermarking algorithm designed to protect the copyright of implicit 3D models created using Neural Radiance Fields (NeRF). The authors, Lifeng Chen, Jia Liu, Yan Ke, Wenquan Sun, Weina Dong, and Xiaozhong Pan, propose a method that embeds watermarks into images in the training set through an embedding network. The NeRF model then uses these images for 3D modeling. A unique aspect of this approach is the use of a copyright verifier that generates a backdoor image by providing a secret perspective as input to the neural radiation field. This process allows a watermark extractor, developed using the hyperparameterization method of the neural network, to extract the embedded watermark from that perspective. In scenarios where there is suspicion of unauthorized use of the 3D model, the verifier can extract watermarks from a secret perspective to verify network copyright. The algorithm demonstrates effectiveness in safeguarding the copyright of 3D models, with the extracted watermarks showing robust resistance against various types of noise attacks. This paper, submitted to the Journal of LaTeX Class Files in August 2015, offers a significant contribution to the field of copyright protection for 3D models generated by NeRF.
    
    \item \textbf{Fast Satellite Tensorial Radiance Field for Multi-date Satellite Imagery of Large Size} introduces SatensoRF, a significant advancement in Neural Radiance Field (NeRF) models for satellite images, particularly addressing the challenges of slow processing speeds, the need for solar information as input, and limitations in handling large satellite images. The authors, Tongtong Zhang and Yuanxiang Li from Shanghai Jiao Tong University, propose a multiscale tensor decomposition approach for color, volume density, and auxiliary variables to model the lightfield with specular color. SatensoRF employs fewer parameters and accelerates the entire process for large-size satellite imagery. It also addresses the prevalent assumption of Lambertian surfaces in neural radiance fields, which falls short for vegetative and aquatic elements. The method incorporates total variation loss to rectify inconsistencies in multi-date imagery and treats the problem as a denoising task. SatensoRF was assessed using subsets from the spacenet multi-view dataset and demonstrated superior performance in novel view synthesis compared to the state-of-the-art Sat-NeRF series, requiring fewer parameters, resulting in faster training and inference speeds, and reduced computational demands.
    
    \item \textbf{NeuralLabeling: A versatile toolset for labeling vision datasets using Neural Radiance Fields} presents an innovative approach for annotating scenes using either bounding boxes or meshes to generate various types of data, including segmentation masks, affordance maps, 2D and 3D bounding boxes, 6DOF object poses, depth maps, and object meshes. The authors, Floris Erich, Naoya Chiba, Yusuke Yoshiyasu, Noriaki Ando, Ryo Hanai, and Yukiyasu Domae, utilize Neural Radiance Fields (NeRF) as a renderer, allowing for labeling with 3D spatial tools while incorporating geometric clues such as occlusions, using only images captured from multiple viewpoints. To demonstrate its practical application in robotics, they created the Dishwasher30k dataset, adding ground truth depth maps to 30,000 frames of transparent object RGB and noisy depth maps of glasses in a dishwasher. Training a simple deep neural network with the annotated depth maps yielded higher reconstruction performance than previous weakly supervised approaches. Submitted to arXiv on September 21, 2023, NeuralLabeling offers a time-efficient and low-cost solution for creating large computer vision datasets, leveraging the power of NeRF for photorealistic rendering and geometric understanding.
    
    \item \textbf{RHINO: Regularizing the Hash-based Implicit Neural Representation} introduces a novel approach to enhance Implicit Neural Representations (INR) through a hash-table, specifically addressing the issue of insufficient regularization in current state-of-the-art methods. The authors, Hao Zhu, Fengyi Liu, Qi Zhang, Xun Cao, and Zhan Ma from Nanjing University and Tencent AI Lab, identify the root of unreliable and noisy interpolations in the broken gradient flow between input coordinates and indexed hash-keys. To resolve this, they propose RHINO, which incorporates a continuous analytical function to facilitate regularization by connecting the input coordinate and the network without altering the architecture of current hash-based INRs. This connection ensures seamless backpropagation of gradients from the network’s output to the input coordinates, thereby enhancing regularization. RHINO demonstrates broadened regularization capability across different hash-based INRs like DINER and Instant NGP, and in various tasks such as image fitting, representation of signed distance functions, and optimization of 5D static / 6D dynamic neural radiance fields. Notably, RHINO outperforms current state-of-the-art techniques in both quality and speed. The paper was supported by the National Key Research and Development Project of China and the NSFC.
    
    \item \textbf{NeRRF: 3D Reconstruction and View Synthesis for Transparent and Specular Objects with Neural Refractive-Reflective Fields} presents a groundbreaking approach to synthesizing views and reconstructing 3D models of transparent and specular objects, which are common in real-world robotics and AR/VR applications. The authors, Xiaoxue Chen, Junchen Liu, Hao Zhao, Guyue Zhou, and Ya-Qin Zhang from Tsinghua University and Beihang University, address the limitations of Neural Radiance Fields (NeRF) in handling complex light path changes due to refraction and reflection. Their method, NeRRF, takes object silhouettes as input and utilizes marching tetrahedra with progressive encoding to reconstruct the geometry of non-Lambertian objects. It models refraction and reflection effects in a unified framework using Fresnel terms and introduces a virtual cone supersampling technique for efficient anti-aliasing. The method is benchmarked on various shapes, backgrounds, and Fresnel terms on both real-world and synthetic datasets. It also demonstrates its utility in editing applications like material editing, object replacement/insertion, and environment illumination estimation. The paper contributing significantly to the fields of novel view synthesis, neural rendering, and shape reconstruction.
    
    \item \textbf{MM-NeRF: Multimodal-Guided 3D Multi-Style Transfer of Neural Radiance Field} introduces a novel approach for 3D style transfer, addressing the challenges of high-quality stylization with texture details and multimodal guidance. The authors, Zijiang Yang, Zhongwei Qiu, Chang Xu, and Dongmei Fu from the University of Science and Technology Beijing and The University of Sydney, propose MM-NeRF, a multimodal-guided method for 3D multi-style transfer of Neural Radiance Fields (NeRF). MM-NeRF tackles the issue of inconsistent object states in different views caused by common training methods, which lead to a loss of texture details and low-quality rendering. The solution involves projecting multimodal guidance into a unified space to maintain consistency of multimodal styles, extracting multimodal features for guiding 3D stylization, and implementing a multi-head learning scheme to ease the learning of multi-style transfer. Additionally, a multi-view style consistent loss is introduced to address the inconsistency of multi-view supervision data, and an incremental learning mechanism is proposed for generalizing MM-NeRF to new styles with minimal effort. Extensive experiments on real-world datasets demonstrate MM-NeRF's ability to achieve high-quality 3D multi-style stylization with multimodal guidance while maintaining multi-view and style consistency.
    
    \item \textbf{Variational Inference for Scalable 3D Object-centric Learning} presents a novel approach to scalable unsupervised object-centric representation learning in 3D scenes. The authors, Tianyu Wang, Kee Siong Ng, and Miaomiao Liu from the School of Computing at The Australian National University, address the limitations of existing object-centric representation learning methods, which struggle to generalize to larger scenes due to their reliance on a fixed global coordinate system. Their method proposes learning view-invariant 3D object representations in localized object coordinate systems, separately estimating object pose and appearance representation, and explicitly mapping object representations across views while maintaining object identities. The framework includes an amortized variational inference pipeline for processing sequential input and updating object latent distributions online. To handle large-scale scenes with varying numbers of objects, the authors introduce a Cognitive Map for registering and querying objects on a per-scene global map, enabling scalable representation learning. The method leverages object-centric neural radiance fields (NeRF) within this unsupervised learning framework. Experimental results on synthetic and real datasets demonstrate the method's ability to infer and maintain object-centric representations of 3D scenes, outperforming previous models. This work significantly advances the field of 3D unsupervised object-centric learning, particularly in large-scale scene understanding.
    
    \item \textbf{Tiled Multiplane Images for Practical 3D Photography} introduces an innovative approach for synthesizing novel views from a single image, with applications in virtual reality and mobile computing. The authors, Numair Khan, Douglas Lanman, and Lei Xiao from Reality Labs Research at Meta, propose a method using Tiled Multiplane Images (TMPIs). This technique addresses the limitations of traditional Multiplane Images (MPIs), which require many depth layers and are highly redundant. TMPIs divide an MPI into small, tiled regions, each with only a few depth planes, reducing computational overhead while maintaining quality. The process involves using a single RGB image and an estimated monocular depth map to recreate the scene as a grid of MPIs. Novel views are rendered by warping and compositing each MPI tile into the target camera's frustum. This method offers a more efficient solution for single-view 3D photography in everyday settings, producing results comparable to state-of-the-art single-view MPI methods.
    
    \item \textbf{NAS-NeRF: Generative Neural Architecture Search for Neural Radiance Fields} introduces an innovative approach to optimize Neural Radiance Fields (NeRFs) for novel view synthesis, particularly addressing their high computational complexity which limits deployability. The authors, Saeejith Nair, Yuhao Chen, Mohammad Javad Shafiee, and Alexander Wong from the University of Waterloo and DarwinAI Corp., propose NAS-NeRF, a generative neural architecture search strategy that creates compact, scene-specialized NeRF architectures. This method dynamically balances architecture complexity and target synthesis quality metrics, tailoring architectures to each scene's specific requirements. NAS-NeRF incorporates constraints on target metrics and budgets to guide the search, resulting in architectures that are significantly smaller, require fewer FLOPs, and are faster on a GPU than baseline NeRFs, without compromising on synthesis quality. The experiments on the Blender synthetic dataset demonstrate the effectiveness of NAS-NeRF in generating efficient architectures suitable for various scenes, making it a significant advancement for deploying NeRFs on resource-constrained platforms.
    
    \item \textbf{3D Density-Gradient based Edge Detection on Neural Radiance Fields (NeRFs) for Geometric Reconstruction} introduces an innovative method for generating geometric 3D reconstructions from Neural Radiance Fields (NeRFs). The authors, Miriam Jäger and Boris Jutzi from the Institute of Photogrammetry and Remote Sensing at the Karlsruhe Institute of Technology in Germany, focus on overcoming the challenges associated with accurate and complete reconstructions based on density values. Their approach utilizes density gradients, derived from relative values, under the assumption that density increases from non-object to object areas. By processing the voxelized 3D density field, they apply 3D edge detection filters of the first and second derivatives, namely Sobel, Canny, and Laplacian of Gaussian. This method allows for the extraction of edges across a wide density range, independent of absolute magnitudes, leading to high geometric accuracy on object surfaces and remarkable object completeness. Notably, the Canny filter effectively eliminates gaps and delivers a uniform point density, achieving a favorable balance between correctness and completeness across various scenes. The paper addresses the field of photogrammetry, offering a new perspective on 3D reconstruction from NeRFs.
    
    \item \textbf{3D Reconstruction with Generalizable Neural Fields using Scene Priors} introduces a novel approach for scalable and efficient 3D scene reconstruction using Neural Fields incorporating Scene Priors (NFPs). The authors, Yang Fu, Shalini De Mello, Xueting Li, Amey Kulkarni, Jan Kautz, Xiaolong Wang, and Sifei Liu from UC San Diego and NVIDIA, address the challenge of training a separate network for each scene, which is inefficient and ineffective with limited views. Their NFP network maps single-view RGB-D images into signed distance and radiance values, enabling complete scene reconstruction by merging individual frames in the volumetric space without a fusion module. This method offers better flexibility and can be trained on large-scale datasets for fast adaptation to new scenes with fewer views. NFP demonstrates state-of-the-art performance in scene reconstruction and efficiency, and uniquely supports single-image novel-view synthesis, an underexplored area in neural fields. The approach is significant for applications in virtual reality, augmented reality, and robotics, where efficient and detailed 3D scene reconstruction is crucial. More qualitative results are available on the project's webpage.
    
    \item \textbf{BASED: Bundle-Adjusting Surgical Endoscopic Dynamic Video Reconstruction using Neural Radiance Fields} presents a groundbreaking approach to reconstructing deformable scenes from endoscopic videos, a critical advancement for intraoperative navigation, surgical visual perception, and robotic surgery. The authors, Shreya Saha, Sainan Liu, Shan Lin, Jingpei Lu, and Michael Yip, tackle the limitations of previous modular approaches confined to specific camera and scene settings. Their work leverages Neural Radiance Fields (NeRF) to learn 3D implicit representations of dynamic and deformable scenes, even with unknown camera poses. This method is particularly applicable to endoscopic surgical scenes in robotic surgery, removing constraints of known camera poses and overcoming drawbacks of current dynamic scene reconstruction techniques. The authors demonstrate the versatility of their model across diverse camera and scene settings, showing promise for both current and future robotic surgical systems. This advancement is foundational for autonomous robotic interventions in minimally invasive surgery, offering a more accurate and flexible approach to surgical scene reconstruction.
    
    \item \textbf{NeuRBF: A Neural Fields Representation with Adaptive Radial Basis Functions} is a type of neural fields that employs general radial bases for signal representation, addressing the limitations of state-of-the-art neural fields that rely on grid-based representations. The authors, Zhang Chen, Zhong Li, Liangchen Song, Lele Chen, Jingyi Yu, Junsong Yuan, and Yi Xu from OPPO US Research Center, University at Buffalo, and ShanghaiTech University, propose a method that builds upon general radial bases with flexible kernel position and shape, offering higher spatial adaptivity and a closer fit to target signals. To enhance the channel-wise capacity of radial basis functions, they compose them with multi-frequency sinusoidal functions, extending a radial basis to multiple Fourier radial bases without extra parameters. This technique facilitates the representation of details. By combining adaptive radial bases with grid-based ones, their hybrid approach inherits both adaptivity and interpolation smoothness. The method is designed to adapt to different types of signals effectively and demonstrates higher accuracy and compactness than prior arts in experiments on 2D image and 3D signed distance field representation. When applied to neural radiance field reconstruction, it achieves state-of-the-art rendering quality with a small model size and comparable training speed. This advancement is significant for applications in neural rendering, medical imaging, acoustic synthesis, and climate prediction, where accurate and compact neural fields representations are crucial.
    
    \item \textbf{SHACIRA: Scalable HAsh-grid Compression for Implicit Neural Representations} is a framework for compressing feature grids in Implicit Neural Representations (INR), addressing the challenge of large memory consumption. The authors, Sharath Girish, Abhinav Shrivastava, and Kamal Gupta from the University of Maryland, propose SHACIRA, a task-agnostic approach that reparameterizes feature grids with quantized latent weights and applies entropy regularization in the latent space. This method allows for significant compression across various domains, including images, videos, and radiance fields, without the need for post-hoc pruning or quantization stages. SHACIRA outperforms existing INR approaches in both quantitative and qualitative results, without requiring large datasets or domain-specific heuristics. This advancement is crucial for storage and streaming applications where memory consumption is a bottleneck. The approach represents a significant step in efficiently managing the memory requirements of INRs, making them more practical for diverse applications.
    
    \item \textbf{Learning Effective NeRFs and SDFs Representations with 3D Generative Adversarial Networks for 3D Object Generation: Technical Report for ICCV 2023 OmniObject3D Challenge} presents a solution for 3D object generation in the ICCV 2023 OmniObject3D Challenge. The authors, Zheyuan Yang, Yibo Liu, Guile Wu, Tongtong Cao, Yuan Ren, Yang Liu, and Bingbing Liu from Huawei Noah’s Ark Lab, focus on overcoming the challenges in generating complex, textured, and high-fidelity 3D objects. Their approach involves learning effective Neural Radiance Fields (NeRFs) and Signed Distance Functions (SDFs) representations using 3D Generative Adversarial Networks (GANs). They employ efficient geometry-aware 3D GANs as the backbone, incorporating label embedding and color mapping to train the model on different taxonomies simultaneously. The resulting features are aggregated through a decoder to generate NeRFs-based representations for rendering high-fidelity synthetic images, while optimizing SDFs to effectively represent objects with 3D meshes. This method can be effectively trained with only a few images of each object from a variety of classes, rather than requiring a large number of images per object or one model per class. This solution was one of the top-3-place solutions in the ICCV 2023 OmniObject3D Challenge, demonstrating its effectiveness in 3D object generation.
    
    \item \textbf{FG-NeRF: Flow-GAN based Probabilistic Neural Radiance Field for Independence-Assumption-Free Uncertainty Estimation} presents a novel approach to enhance Neural Radiance Fields (NeRFs) by incorporating probabilistic modeling without relying on the independence assumption of points in the radiance field. The authors, Songlin Wei, Jiazhao Zhang, Yang Wang, Fanbo Xiang, Hao Su, He Wang from Peking University, Beijing Academy of Artificial Intelligence, and UC San Diego, propose FG-NeRF, which combines the generative capability of adversarial learning with the expressivity of normalizing flow. This method explicitly models the density-radiance distribution of the entire scene, representing the probabilistic NeRF as a mean-shifted probabilistic residual neural model. FG-NeRF is trained without an explicit likelihood function, thereby avoiding the independence assumption. The training involves downsampling images to form fixed-size patches for patch-based adversarial learning. FG-NeRF demonstrates state-of-the-art performance by predicting lower rendering errors and more reliable uncertainty on both synthetic and real-world datasets. This advancement is crucial for applications in robotics, autonomous driving, and human-computer interaction, where incorporating uncertain information is essential. The authors' approach represents a significant step in enhancing NeRFs for more accurate and reliable scene modeling.
    
    \item \textbf{MatrixCity: A Large-scale City Dataset for City-scale Neural Rendering and Beyond} introduces MatrixCity, a comprehensive and high-quality synthetic dataset designed to advance neural rendering research at the city scale. The authors, Yixuan Li, Lihan Jiang, Linning Xu, Yuanbo Xiangli, Zhenzhi Wang, Dahua Lin, and Bo Dai from The Chinese University of Hong Kong and Shanghai AI Laboratory, address the lack of suitable datasets for city-scale neural rendering. MatrixCity, developed using Unreal Engine 5, provides a diverse collection of aerial and street city views, complete with ground-truth camera poses and various data modalities. The dataset's flexibility in environmental control allows for data collection under dynamic conditions like varying lighting, weather, and human and car crowds. MatrixCity comprises 67k aerial images and 452k street images from two city maps totaling 28km². The authors conduct a thorough benchmark on MatrixCity, revealing unique challenges in city-scale neural rendering and suggesting potential improvements for future research. This dataset is a significant contribution to the field, offering opportunities for research in neural rendering and related applications.
    
    \item \textbf{DreamGaussian: Generative Gaussian Splatting for Efficient 3D Content Creation} introduces a novel framework for accelerating the optimization process in image- and text-to-3D tasks. The authors, Jiaxiang Tang, Jiawei Ren, Hang Zhou, Ziwei Liu, and Gang Zeng from Peking University, Nanyang Technological University, and Baidu Inc., propose DreamGaussian, a generative 3D Gaussian Splatting model that significantly speeds up the creation of high-quality textured meshes. This method contrasts with the occupancy pruning used in Neural Radiance Fields by employing progressive densification of 3D Gaussians, resulting in faster convergence for 3D generative tasks. Additionally, DreamGaussian includes an efficient algorithm for converting 3D Gaussians into textured meshes and a fine-tuning stage to refine details. The framework demonstrates superior efficiency and competitive generation quality, producing high-quality textured meshes from a single-view image in just 2 minutes, approximately ten times faster than existing methods. This advancement is significant for applications in 3D content creation, offering a more efficient solution for generating detailed 3D models from images and text.
    
    \item \textbf{HAvatar: High-fidelity Head Avatar via Facial Model Conditioned Neural Radiance Field} presents a novel approach for creating high-resolution, photo-realistic, and view-consistent head avatars. The authors, Xiaochen Zhao, Lizhen Wang, Jingxiang Sun, Hongwen Zhang, Jinli Suo, and Yebin Liu from Tsinghua University, China, address the challenge of modeling an animatable 3D human head avatar under lightweight setups. They introduce a hybrid explicit-implicit 3D representation that combines the expressiveness of Neural Radiance Fields (NeRF) with the prior information from a parametric template. This method, Facial Model Conditioned Neural Radiance Field, integrates synthetic-renderings-based conditioning to fuse parametric model information into the implicit field without constraining its topological flexibility. The approach overcomes inconsistent shape issues and improves animation stability. Additionally, by adopting an overall GAN-based architecture using an image-to-image translation network, the method achieves high-resolution, realistic, and view-consistent synthesis of dynamic head appearance. The experiments demonstrate that this method outperforms existing methods in 3D head avatar animation, marking a significant advancement in the field of head avatar synthesis and animation.
    
    \item \textbf{Forward Flow for Novel View Synthesis of Dynamic Scenes} introduces a novel approach to novel view synthesis in dynamic scenes using a neural radiance field (NeRF) and forward warping. The authors, Xiang Guo, Jiadai Sun, Yuchao Dai, Guanying Chen, Xiaoqing Ye, Xiao Tan, Errui Ding, Yumeng Zhang, and Jingdong Wang from Northwestern Polytechnical University, CUHK-Shenzhen, and Baidu Inc., address the limitations of existing methods that use a static NeRF for the canonical space and render dynamic images by mapping sampled 3D points back to the canonical space with a learned backward flow field. This backward flow field is often non-smooth and discontinuous, making it difficult to fit with smooth motion models. To overcome this, the authors propose estimating a forward flow field to directly warp the canonical radiance field to other time steps. This forward flow field is smoother and more continuous within the object region, benefiting motion model learning. They represent the canonical radiance field with voxel grids for efficient forward warping and propose a differentiable warping process, including an average splatting operation and an inpaint network, to resolve mapping issues. Their method outperforms existing methods in both novel view rendering and motion modeling, demonstrating the effectiveness of forward flow motion modeling. This advancement is significant for applications in virtual reality, augmented reality, data augmentation, and image editing, where accurate and efficient rendering of dynamic scenes is crucial.
    
    \item \textbf{Multi-task View Synthesis with Neural Radiance Fields} introduces MuvieNeRF, a novel framework designed to address the challenges of multi-task visual learning in computer vision. Developed by Shuhong Zheng, Zhipeng Bao, Martial Hebert, and Yu-Xiong Wang from the University of Illinois Urbana-Champaign and Carnegie Mellon University, MuvieNeRF reinterprets multi-task prediction as a set of novel-view synthesis tasks for multiple scene properties, including RGB. The framework integrates two key modules: Cross-Task Attention (CTA) and Cross-View Attention (CVA), enabling efficient use of information across multiple views and tasks. This approach allows MuvieNeRF to simultaneously synthesize multiple scene properties with promising visual quality, even outperforming conventional discriminative models in various settings. The authors demonstrate MuvieNeRF's universal applicability across different NeRF backbones through extensive evaluation on both synthetic and realistic benchmarks. This advancement is significant for applications in robotics and other fields where handling multiple tasks and synthesizing different scene properties from novel viewpoints is crucial. The code for MuvieNeRF is available on GitHub, contributing further to the research community.
    
    \item \textbf{MMPI: a Flexible Radiance Field Representation by Multiple Multi-plane Images Blending} introduces a novel approach for synthesizing high-quality views of complex scenes using Neural Radiance Fields (NeRF) based on Multi-Plane Images (MPI). The authors, Yuze He, Peng Wang, Yubin Hu, Wang Zhao, Ran Yi, Yong-Jin Liu, and Wenping Wang, address the limitations of existing NeRF works that use MPI representation, which are typically confined to simple forward-facing scenes with limited camera movement. Their solution, Multiple Multi-plane Images (MMPI), encodes the scene with a set of multiple MPIs arranged to cover a wide, unbounded range of the scene. This approach includes a reliability field for each MPI, blending sampled colors and densities for effective volume rendering. The authors propose a two-stage reliability learning scheme, where each MPI is initially trained individually, followed by joint training using an adaptive blending technique. This method allows for larger weights to be given to MPIs with better local representation abilities. The MMPI framework demonstrates the potential to extend MPI to render more complex scenes, such as large-range or even 360-degree scenes, overcoming the limitations of predefined camera frustums. This advancement is significant for applications in virtual reality, autonomous driving, and other fields requiring high-quality image synthesis of complex scenes. The authors' approach represents a significant step in enhancing the capabilities of NeRF for more diverse and complex scene renderings.
    
    \item \textbf{Enabling Neural Radiance Fields (NeRF) for Large-scale Aerial Images -- A Multi-tiling Approach and the Geometry Assessment of NeRF} presents a novel approach to scale Neural Radiance Fields for large-scale aerial datasets and provides a thorough geometry assessment of NeRF. The authors, Ningli Xu, Rongjun Qin, Debao Huang, and Fabio Remondino, introduce a location-specific sampling technique and a multi-camera tiling (MCT) strategy to reduce memory consumption during image loading and representation training. MCT decomposes a large-frame image into multiple tiled images with different camera models, allowing these small-frame images to be fed into the training process as needed for specific locations without a loss of accuracy. The authors implement their method on Mip-NeRF and compare its geometry performance with three photogrammetric Multi-View Stereo (MVS) pipelines on two typical aerial datasets against LiDAR reference data. The results suggest that the proposed NeRF approach produces better completeness and object details than traditional approaches, although it still falls short in terms of accuracy. This advancement is significant for applications in 3D reconstruction and aerial photogrammetry, offering a more efficient and detailed approach to large-scale aerial data processing.
    
    \item \textbf{How Many Views Are Needed to Reconstruct an Unknown Object Using NeRF?} explores the optimization of view planning for online active object reconstruction using Neural Radiance Fields (NeRFs). The authors, Sicong Pan, Liren Jin, Hao Hu, Marija Popović, and Maren Bennewitz, address the computational inefficiency of previous NeRF-based methods, which rely on an iterative paradigm of retraining NeRF with each new image and planning paths to the next best view. Their approach, based on the Prediction of the Required number of Views (PRV), posits that the number of views needed for reconstruction depends on the object's complexity. They develop PRVNet, a deep neural network designed to predict the required number of views, allowing for tailored data acquisition and planning of globally shortest paths. The method is trained using supervision labels generated from the ShapeNet dataset. Simulated experiments demonstrate that their PRV-based view planning method outperforms baselines in reconstruction quality while significantly reducing movement cost and planning time. The approach's generalization ability is further validated in real-world experiments. This advancement is crucial for robotics applications where efficient and accurate 3D reconstruction is essential, offering a more effective method for view planning in NeRF-based object reconstruction.
    
    \item \textbf{PC-NeRF: Parent-Child Neural Radiance Fields under Partial Sensor Data Loss in Autonomous Driving Environments} introduces a novel 3D scene reconstruction framework designed to address the challenges faced by autonomous vehicles, especially when partial sensor data is lost. The authors, Xiuzhong Hu, Guangming Xiong, Zheng Zang, Peng Jia, Yuxuan Han, and Junyi Ma, propose the parent-child neural radiance field (PC-NeRF). This framework comprises two modules: the parent NeRF and the child NeRF, which simultaneously optimize scene-level, segment-level, and point-level scene representations. The unique aspect of PC-NeRF is its ability to efficiently utilize sensor data by leveraging the segment-level representation capabilities of child NeRFs, enabling quick acquisition of an approximate volumetric representation of the scene even with limited observations. Extensive experiments demonstrate that PC-NeRF achieves high-precision 3D reconstruction in large-scale scenes and effectively tackles situations where partial sensor data is lost. Additionally, it boasts high deployment efficiency with limited training time. This advancement is particularly significant for autonomous driving applications, where robust and efficient 3D scene reconstruction is crucial for safe and effective navigation. The authors plan to make their approach implementation and pre-trained models available on GitHub, contributing further to the research community.
    
    \item \textbf{Neural Processing of Tri-Plane Hybrid Neural Fields} presents an innovative approach to processing neural fields, particularly focusing on tasks such as classification and part segmentation. The authors, Adriano Cardace, Pierluigi Zama Ramirez, Francesco Ballerini, Allan Zhou, Samuele Salti, and Luigi Di Stefano from the University of Bologna and Stanford University, explore the use of neural fields parameterized by shared networks and individual neural fields parameterized as large Multi-Layer Perceptrons (MLPs). They address the challenges associated with processing these fields due to the high dimensionality of the weight space and sensitivity to random initialization. The paper introduces hybrid representations based on tri-planes, which have emerged as a more effective and efficient alternative to realize neural fields. The authors demonstrate that the tri-plane discrete data structure encodes rich information that can be effectively processed by standard deep-learning machinery. They establish an extensive benchmark covering various fields such as occupancy, signed/unsigned distance, and radiance fields. Their approach achieves task performance far superior to frameworks that process large MLPs and almost on par with architectures handling explicit representations, while maintaining the same reconstruction quality. This advancement is significant for applications in 3D world representation and various types of neural fields, offering a more efficient and effective method for processing neural fields in tasks like classification and segmentation.
    
    \item \textbf{LEAP: Liberate Sparse-view 3D Modeling from Camera Poses} is a groundbreaking approach to 3D modeling from sparse views without relying on camera pose information. The authors, Hanwen Jiang, Zhenyu Jiang, Yue Zhao, and Qixing Huang from the Department of Computer Science at The University of Texas at Austin, challenge the conventional necessity of accurate camera poses in multi-view 3D modeling. Their novel pose-free approach, LEAP, discards pose-based operations and instead learns geometric knowledge directly from data. LEAP employs a neural volume shared across scenes, parameterized to encode geometry and texture priors. This neural volume is updated for each scene by aggregating 2D image features in a feature-similarity-driven manner, enabling novel view synthesis from any viewpoint. The authors demonstrate that LEAP significantly outperforms prior methods using predicted poses from state-of-the-art pose estimators and performs comparably to approaches using ground-truth poses, while being 400 times faster than PixelNeRF. LEAP's ability to generalize to novel object categories and scenes and its learning of knowledge resembling epipolar geometry is particularly notable. This advancement is significant for applications in 3D vision where sparse and unposed views are common, such as in online product imaging, offering a more efficient and accurate alternative to traditional 3D modeling techniques.
    
    \item \textbf{MIMO-NeRF: Fast Neural Rendering with Multi-input Multi-output Neural Radiance Fields} presents a significant advancement in Neural Radiance Fields (NeRFs) by introducing the MIMO-NeRF, a multi-input multi-output approach for faster neural rendering. The author, Takuhiro Kaneko from NTT Corporation, addresses the slow rendering issue inherent in typical NeRFs, which use a single-input single-output (SISO) multilayer perceptron (MLP) for mapping 3D coordinates and view direction to color and volume density. MIMO-NeRF replaces the SISO MLP with a MIMO MLP, conducting mappings in a group-wise manner, thus reducing the number of MLPs running and improving rendering speed. However, this approach introduces ambiguity in color and volume density, as these values are determined in a non-unique manner by a set of input coordinates that vary by viewpoint, grouping, and sampling. To mitigate this challenge, the author proposes a self-supervised learning method that regularizes the MIMO MLP with multiple fast reformulated MLPs, alleviating ambiguity without using pretrained models. The comprehensive experimental evaluation, including comparative and ablation studies, shows that MIMO-NeRF achieves a good trade-off between speed and quality with reasonable training time. This advancement is crucial for applications in photo editing, content creation, virtual reality, and environmental understanding, where efficient and high-quality novel view synthesis is essential.
    
    \item \textbf{Adaptive Multi-NeRF: Exploit Efficient Parallelism in Adaptive Multiple Scale Neural Radiance Field Rendering} introduces a novel approach to enhance Neural Radiance Fields (NeRF) for real-time rendering applications. The authors, Tong Wang and Shuichi Kurabayashi from Cygames Research, Cygames Inc., Japan, address the challenge of lengthy training and rendering processes in NeRF, which hinder its widespread adoption for real-time applications. Their method, Adaptive Multi-NeRF, subdivides scenes into axis-aligned bounding boxes using a tree hierarchy approach, assigning smaller NeRFs to different-sized subspaces based on scene complexity. This ensures a specific neural representation for each scene part, optimizing scene subdivision with a guidance density grid. The method achieves a balanced workload suitable for small Multilayer Perceptrons (MLPs), allowing for parallel inference and higher GPU utilization. This advancement is significant for applications requiring real-time high-fidelity image synthesis, such as virtual reality and augmented reality. The approach represents a significant step in making NeRF a practical solution for real-time applications by accelerating the rendering process and increasing efficiency.
    
    \item \textbf{EvDNeRF: Reconstructing Event Data with Dynamic Neural Radiance Fields} is a pioneering pipeline for generating event data and training an event-based dynamic NeRF for reconstructing event streams in scenes with both rigid and non-rigid deformations. The authors, Anish Bhattacharya, Ratnesh Madaan, Fernando Cladera, Sai Vemprala, Rogerio Bonatti, Kostas Daniilidis, Ashish Kapoor, Vijay Kumar, Nikolai Matni, and Jayesh K. Gupta from various institutions including the University of Pennsylvania, Microsoft, and Scaled Foundations, focus on overcoming the limitations of standard cameras in capturing fast motion. Event cameras, which register asynchronous per-pixel brightness changes at MHz rates with high dynamic range, are ideal for observing fast motion with minimal motion blur. While NeRFs offer high-quality geometric-based learnable rendering, previous work with events has only considered static scenes. EvDNeRF can predict event streams of dynamic scenes from either a static or moving viewpoint between any desired timestamps, effectively serving as an event-based simulator for a given scene. This method outperforms standard dynamic NeRFs paired with event simulators, especially in test-time predictions of events at fine time resolutions. This advancement is significant for applications in computer graphics and robotics, particularly in simulating and reconstructing dynamic scenes that are challenging for traditional cameras. The authors have made their simulated and real datasets, as well as code for both event-based data generation and training of event-based dynamic NeRF models, publicly available.
    
    \item \textbf{USB-NeRF: Unrolling Shutter Bundle Adjusted Neural Radiance Fields} is a significant advancement in NeRF technology, specifically addressing the challenges posed by rolling shutter (RS) images. The authors, Moyang Li, Peng Wang, Lingzhe Zhao, Bangyan Liao, Peidong Liu, and their team from Westlake University, ETH Zürich, and Zhejiang University, propose USB-NeRF, a novel framework capable of correcting rolling shutter distortions and recovering accurate camera motion trajectories. This method models the physical image formation process of a RS camera within the NeRF framework. USB-NeRF stands out for its ability to remove RS effects, synthesize novel view images, and estimate camera motion more effectively than previous methods. Additionally, it can be used to recover high-fidelity, high frame-rate global shutter videos from a sequence of RS images. This advancement is crucial for applications in computer vision and 3D reconstruction, where the accuracy of camera pose estimation and the quality of image synthesis are paramount. The approach represents a significant step in enhancing NeRF algorithms to handle the complexities of RS images, improving their applicability in real-world scenarios.
    
    \item \textbf{GETAvatar: Generative Textured Meshes for Animatable Human Avatars} introduces a groundbreaking approach for creating animatable human avatars with high-quality textures and geometries. The authors, Xuanmeng Zhang, Jianfeng Zhang, Rohan Chacko, Hongyi Xu, Guoxian Song, Yi Yang, and Jiashi Feng from various universities and ByteDance, address two major challenges in 3D-aware full-body human generation: generating detailed geometries like garment wrinkles and achieving high-resolution rendering with volumetric radiance fields and neural renderers. GETAvatar, their proposed generative model, directly generates explicit textured 3D meshes for human avatars with photorealistic appearance and fine geometric details. The method includes an articulated 3D human representation with explicit surface modeling, enriched by learning from 2D normal maps of 3D scan data. It utilizes a rasterization-based renderer for efficient high-resolution image generation. Extensive experiments demonstrate that GETAvatar achieves state-of-the-art performance in both appearance and geometry quality, capable of generating images at high resolutions and frame rates. This advancement is significant for applications in video games, AR/VR, and movie production, where creating high-quality 3D human avatars with explicit control over camera and body poses is crucial. The authors plan to release their code and models, contributing further to the research community.
    
    \item \textbf{Efficient-3DiM: Learning a Generalizable Single-image Novel-view Synthesizer in One Day} presents a significant advancement in the field of novel view synthesis from a single image. The authors, Yifan Jiang, Hao Tang, Jen-Hao Rick Chang, Liangchen Song, Zhangyang Wang, and Liangliang Cao from Apple and the University of Texas at Austin, address the challenge of synthesizing photorealistic novel views, which has been a significant hurdle in computer vision. Their proposed framework, Efficient-3DiM, leverages a pre-trained diffusion model specifically designed for 2D image synthesis. This model, when optimized on a 3D finetuning task, demonstrates remarkable capabilities in producing photorealistic views. However, training such a model typically requires extensive data, parameters, and computational resources. To overcome these challenges, the authors introduce several pragmatic strategies, including a crafted timestep sampling strategy, a superior 3D feature extractor, and an enhanced training scheme. These innovations enable Efficient-3DiM to reduce the total training time from 10 days to less than one day, using the same computational platform. The method's efficiency and generalizability are demonstrated through comprehensive experiments. This advancement is significant for applications in immersive free-viewpoint interaction, particularly in the realms of computer vision and computer graphics, where creating realistic interactive imagery from a single image is crucial. The paper represents a significant step in achieving interactivity and realism in novel view synthesis.
    
    \item \textbf{Shielding the Unseen: Privacy Protection through Poisoning NeRF with Spatial Deformation} introduces an innovative method to safeguard user privacy against the generative capabilities of Neural Radiance Fields (NeRF) models. The authors, Yihan Wu, Brandon Y. Feng, and Heng Huang from the University of Maryland, have developed a novel poisoning attack method that subtly alters observed views. These changes are imperceptible to the human eye but significantly disrupt NeRF's ability to reconstruct a 3D scene accurately. The approach involves a bi-level optimization algorithm with a Projected Gradient Descent (PGD)-based spatial deformation. Tested extensively on two common NeRF benchmark datasets consisting of 29 real-world scenes, the results demonstrate that this privacy-preserving method notably impairs NeRF's performance. The method's adaptability across various perturbation strengths and NeRF architectures highlights its versatility. This work provides crucial insights into NeRF's vulnerabilities and underscores the importance of considering privacy risks in developing robust 3D scene reconstruction algorithms. Contributing to the broader conversation on responsible AI and generative machine learning, this study emphasizes the need to protect user privacy and respect creative ownership in the digital age.
    
    \item \textbf{Point-Based Radiance Fields for Controllable Human Motion Synthesis} introduces a novel method for synthesizing controllable human motion using static point-based radiance fields. The authors, Haitao Yu, Deheng Zhang, Peiyuan Xie, Tianyi Zhang, and their team from ETH Zurich, focus on overcoming the limitations of previous editable neural radiance field methods in achieving complex 3D human editing. Their approach utilizes an explicit point cloud to train the static 3D scene and applies deformation by encoding the point cloud translation using a deformation MLP. The method ensures rendering consistency with the canonical space training by estimating local rotation using SVD and interpolating per-point rotation to the query view direction of the pre-trained radiance field. Extensive experiments demonstrate that this approach significantly outperforms state-of-the-art methods in fine-level complex deformation, generalizable to other 3D characters besides humans. This advancement is significant for applications in visual effects, video games, and telepresence, where easy control over character poses and photorealistic rendering in dynamic scenes are crucial. The authors provide their code on GitHub, contributing further to the research community.
    
    \item \textbf{BID-NeRF: RGB-D image pose estimation with inverted Neural Radiance Fields} presents an innovative approach to improve the Inverted Neural Radiance Fields (iNeRF) algorithm, which is crucial for image pose estimation. The authors, Ágoston István Csehi and Csaba Máté Józsa from Nokia Bell Labs and Budapest University of Technology and Economics, focus on enhancing NeRFs, a novel neural space representation model capable of synthesizing photorealistic novel views of real-world scenes or objects. Their contributions include extending the localization optimization objective with a depth-based loss function, introducing a multi-image based loss function that uses a sequence of images with known relative poses without increasing computational complexity, omitting hierarchical sampling during volumetric rendering, and extending the sampling interval for better convergence. These modifications significantly improve the convergence speed and extend the basin of convergence. This advancement is particularly useful for extended reality, robotics, and Simultaneous Localization and Mapping (SLAM) applications, where robust and efficient image pose estimation is crucial. The paper represents a significant step in enhancing the capabilities of NeRFs for precise and efficient image pose estimation in various applications.
    
    \item \textbf{Targeted Adversarial Attacks on Generalizable Neural Radiance Fields} addresses the vulnerability of Neural Radiance Fields (NeRFs) to adversarial attacks, a critical issue as NeRFs gain prominence in applications like augmented reality, robotics, and virtual environments. The authors, András Horváth, Csaba M. Józsa, and their team, focus on demonstrating how generalizable NeRFs can be compromised by low-intensity adversarial attacks and adversarial patches. These attacks can generate a specific, predefined output scene, posing a significant threat to the reliability of NeRFs in critical applications. The paper highlights the importance of understanding and mitigating these vulnerabilities, as NeRFs represent 3D scenes as continuous functions and are increasingly used for complex and detailed scene reconstruction. The susceptibility of NeRFs to such attacks could lead to the production of unrealistic maps and representations, potentially causing erroneous outcomes in applications that rely on accurate scene reconstruction and navigation. This research is crucial in highlighting the need for robustness against adversarial attacks in the development and deployment of NeRF-based applications. The paper represents a significant step in addressing the security aspects of NeRFs, particularly in scenarios where accurate and reliable 3D scene representation is essential.
    
    \item \textbf{Improving Neural Radiance Field using Near-Surface Sampling with Point Cloud Generation} presents a novel approach to enhance Neural Radiance Fields (NeRF) by incorporating near-surface sampling. The authors, Hye Bin Yoo, Hyun Min Han, Sung Soo Hwang, and Il Yong Chun from various South Korean universities, focus on overcoming the limitations of NeRF, such as long training times and degraded rendering quality due to sampling points from occluded regions or spaces unlikely to contain objects. Their method estimates the geometry of a 3D scene and performs sampling around the surface of a 3D object using depth images from the training set. This approach significantly improves rendering quality compared to the original NeRF and state-of-the-art depth-based NeRF methods. Additionally, it accelerates the training time of a NeRF model. The paper's findings are crucial for applications in metaverse and virtual reality, where generating novel views accurately is important. The proposed near-surface sampling framework represents a significant advancement in the field of view synthesis, particularly in enhancing the efficiency and quality of NeRF models.
    
    \item \textbf{Geometry Aware Field-to-field Transformations for 3D Semantic Segmentation} presents a novel approach for 3D semantic segmentation using Neural Radiance Fields (NeRFs) with only 2D supervision. The authors, Dominik Hollidt, Clinton Wang, Polina Golland, and Marc Pollefeys, focus on extracting features along a surface point cloud to achieve a compact and sample-efficient representation conducive to 3D reasoning. This method is unique in its unsupervised learning via masked autoencoding, enabling few-shot segmentation. The approach is agnostic to scene parameterization, making it applicable to scenes fit with any type of NeRF. This advancement is significant for applications in robotics, augmented reality, autonomous driving, and other fields where 3D segmentations are crucial. The authors' method promises to advance many applications by enabling dense 3D scene representations and segmentations, overcoming challenges such as the lower amount of 3D annotated data and higher computational complexity associated with 3D methods. The paper represents a significant step in 3D semantic segmentation, particularly in leveraging NeRFs for efficient and effective scene understanding.
    
    \item \textbf{LocoNeRF: A NeRF-based Approach for Local Structure from Motion for Precise Localization} presents a novel approach to improve visual localization using Structure from Motion (SfM) techniques with Neural Radiance Fields (NeRF). The authors, Artem Nenashev, Mikhail Kurenkov, Andrei Potapov, Iana Zhura, Maksim Katerishich, and Dzmitry Tsetserukou, address the limitations of global SfM, which suffers from high latency, and the challenges of local SfM, which requires large image databases for accurate reconstruction. Their solution utilizes NeRF as opposed to image databases, significantly reducing the space required for storage. The proposed method achieves an accuracy of 0.068 meters compared to ground truth, slightly lower than the most advanced method COLMAP, but with a much smaller database size. This advancement is crucial for mobile robotics, where precise localization is necessary for navigation and control tasks. The approach represents a significant step in visual localization, particularly in reducing memory costs, shortening time, and improving accuracy and robustness for mobile robots.
    
    \item \textbf{Neural Impostor: Editing Neural Radiance Fields with Explicit Shape Manipulation} "Neural Impostor: Editing Neural Radiance Fields with Explicit Shape Manipulation" introduces a novel hybrid representation, Neural Impostor, for efficiently editing neural implicit fields, particularly in terms of geometry modification. Developed by Ruiyang Liu, Jinxu Xiang, Bowen Zhao, Ran Zhang, Jingyi Yu, and Changxi Zheng from Tencent Pixel Lab and ShanghaiTech University, this framework incorporates an explicit tetrahedral mesh alongside a multigrid implicit field designated for each tetrahedron within the explicit mesh. The key innovation lies in bridging explicit shape manipulation with geometric editing of implicit fields using multigrid barycentric coordinate encoding. This approach allows for a range of editing operations, including deformation, composition, and generation of neural implicit fields, while maintaining complex volumetric appearance. The authors demonstrate the robustness and adaptability of Neural Impostor through diverse examples and experiments, including editing both synthetic objects and real captured data. This advancement is significant for applications in 3D content creation and manipulation, offering a transformative solution for editing neural radiance fields. 
    
    \item \textbf{Bi-directional Deformation for Parameterization of Neural Implicit Surfaces} introduces a novel neural algorithm for intuitive editing of 3D objects represented as neural implicit surfaces. The authors, Baixin Xu, Jiangbei Hu, Fei Hou, Kwan-Yee Lin, Wayne Wu, Chen Qian, and Ying He from various universities and research institutes, focus on parameterizing neural implicit surfaces to simple parametric domains like spheres, cubes, or polycubes. This approach enables the representation of 3D radiance fields as 2D fields, facilitating visualization and editing tasks. The method computes a bi-directional deformation between 3D objects and their chosen parametric domains without requiring prior information. It employs a forward mapping of points on the zero level set of the 3D object to a parametric domain, followed by a backward mapping through inverse deformation. The process ensures bijectivity using a cycle loss while optimizing the smoothness of both deformations. The framework is compatible with existing neural rendering pipelines, taking multi-view images as input to reconstruct 3D geometry and compute texture maps. It also introduces a technique for intrinsic radiance decomposition, supporting view-independent material editing and view-dependent shading editing. The method allows for immediate rendering of edited textures through volume rendering without network re-training and supports co-parameterization of multiple objects and texture transfer between them. This advancement is significant for applications in virtual reality, augmented reality, and digital content creation, where intuitive and efficient editing of 3D objects is crucial. The source code for this research will be made publicly available.
    
    \item \textbf{A Real-time Method for Inserting Virtual Objects into Neural Radiance Fields} presents a groundbreaking approach for augmenting Neural Radiance Fields (NeRF) with virtual objects in real-time. The authors, Keyang Ye, Hongzhi Wu, Xin Tong, and Kun Zhou from Zhejiang University and Microsoft Research Asia, have developed a method that overcomes several challenges in object insertion for augmented reality applications. Their technique accurately estimates lighting, handles occlusions, and casts soft, detailed shadows onto 3D surfaces. By exploiting the rich information about lighting and geometry in a NeRF, the method produces realistic lighting and shadowing effects and allows interactive manipulation of the virtual object. The approach combines near-field lighting from NeRF and an environment lighting to account for sources not covered by the NeRF, integrates an opacity map from the NeRF for blending the virtual object with the background scene, and uses a precomputed field of spherical signed distance fields for shadow casting. This method significantly outperforms state-of-the-art techniques, offering superior fidelity in virtual object insertion and great potential for augmented reality systems. The paper represents a significant advancement in the field of neural radiance fields, particularly in enhancing AR experiences with realistic and interactive virtual object integration.
    
    \item \textbf{High-Fidelity 3D Head Avatars Reconstruction through Spatially-Varying Expression Conditioned Neural Radiance Field} introduces a groundbreaking method for reconstructing 3D head avatars with intricate facial expressions. The authors, Minghan Qin, Yifan Liu, Yuelang Xu, Xiaochen Zhao, Yebin Liu, and Haoqian Wang from Tsinghua University, address the challenge of retaining detailed facial expressions in NeRF-based 3D head avatar methods. Their approach, Spatially-Varying Expression (SVE) conditioning, integrates spatial positional features and global expression information, enabling the neural radiance field to capture complex facial expressions and achieve realistic rendering and geometry details. The method includes a coarse-to-fine training strategy, with geometry initialization at the coarse stage and adaptive importance sampling at the fine stage, enhancing both geometric and rendering quality. Extensive experiments demonstrate the superior performance of this method in rendering and geometry quality on various datasets. This advancement is significant for applications in VR/AR, gaming, and teleconferencing, where realistic 3D head avatars are crucial. The authors' approach represents a significant step in 3D head avatar reconstruction, particularly in achieving high-fidelity facial expressions.
    
    \item \textbf{Leveraging Neural Radiance Fields for Uncertainty-Aware Visual Localization} presents a novel approach to enhance scene coordinate regression (SCR) for visual localization using Neural Radiance Fields (NeRF). The authors, Le Chen, Weirong Chen, Rui Wang, and Marc Pollefeys, focus on generating training samples for SCR through NeRF, addressing challenges like artifacts in rendered data and redundancy. Their method, U-NeRF, predicts uncertainties for rendered color and depth images, revealing data reliability at the pixel level. SCR is formulated as deep evidential learning with epistemic uncertainty, which evaluates information gain and scene coordinate quality. A novel view selection policy based on uncertainties significantly improves data efficiency. The method is validated on public datasets, demonstrating its ability to select samples with the most information gain and enhance performance efficiently. This advancement is crucial for robotics applications where accurate visual localization is essential. The approach represents a significant step in visual localization, particularly in efficiently training SCR models with high-quality data.
    
    \item \textbf{rpcPRF: Generalizable MPI Neural Radiance Field for Satellite Camera} introduces a novel approach for novel view synthesis of satellite images using Neural Radiance Fields (NeRF). The authors, Tongtong Zhang, Yuanxiang Li, and their team, address the challenge of adapting NeRF for satellite cameras, which often require multiple input views. Their solution, rpcPRF, is a Multiplane Images (MPI) based Planar neural Radiance Field for Rational Polynomial Camera (RPC). This model is unique in its ability to work with single or few input images and perform well on images from unseen scenes. The key innovation lies in using reprojection supervision to induce the predicted MPI to learn the correct geometry between 3D coordinates and images. Additionally, rpcPRF overcomes the need for dense depth supervision by incorporating rendering techniques of radiance fields. The method combines the benefits of implicit representations with the advantages of the RPC model, capturing continuous altitude space while learning the 3D structure. The end-to-end model synthesizes novel views with a new RPC and reconstructs the altitude of the scene. On datasets like TLC from ZY-3 and SatMVS3D with urban scenes from WV-3, rpcPRF outperforms state-of-the-art NeRF-based methods in image fidelity, reconstruction accuracy, and efficiency. This advancement is significant for applications in remote sensing and computer vision, particularly in large-scale satellite photogrammetry, where novel view synthesis and dense altitude mapping are crucial.
    
    \item \textbf{Dynamic Appearance Particle Neural Radiance Field} introduces a novel approach, DAP-NeRF, for modeling dynamic 3D scenes using a particle-based representation. Developed by Ancheng Lin, Jun Li, and their team from the University of Technology Sydney, DAP-NeRF addresses the limitations of existing dynamic NeRFs, which typically use deformation fields and suffer from a close coupling of appearance and motion. DAP-NeRF consists of a static field and a dynamic field, with the latter quantized as a collection of appearance particles. These particles carry visual information of dynamic elements in the scene and are equipped with a motion model. The entire model, including the static field, visual features, and motion models of the particles, is learned from monocular videos without prior geometric knowledge of the scene. This method effectively captures not only the appearance but also the physically meaningful motions in a dynamic 3D scene. The paper represents a significant advancement in the field of 3D scene modeling, particularly for dynamic environments, and is expected to have applications in virtual reality, augmented reality, and computer graphics, where accurate and realistic rendering of dynamic scenes is essential.
    
    \item \textbf{Point-DynRF: Point-based Dynamic Radiance Fields from a Monocular Video} introduces a novel framework for generating novel views from monocular videos of dynamic scenes. The authors, Byeongjun Park and Changick Kim from the Korea Advanced Institute of Science and Technology (KAIST), propose Point-DynRF, which efficiently represents dynamic radiance fields using neural 3D points generated from the input video. This method addresses the limitations of previous approaches that only enforce geometric consistency between adjacent input frames, making it challenging to represent global scene geometry. Point-DynRF reconstructs neural point clouds directly from geometric proxies and optimizes both radiance fields and geometric proxies using proposed losses, allowing them to complement each other. The framework is validated with experiments on the NVIDIA Dynamic Scenes Dataset and several causally captured monocular video clips. This advancement is significant for applications in virtual reality, augmented reality, and video editing, where understanding and rendering dynamic scenes from monocular videos is crucial. The approach represents a significant step in novel view synthesis, particularly in scenarios involving dynamic objects and complex camera trajectories.

    \item \textbf{CBARF: Cascaded Bundle-Adjusting Neural Radiance Fields from Imperfect Camera Poses} presents a novel 3D reconstruction framework, CBARF, designed to optimize camera poses and reconstruct scenes using Neural Radiance Fields (NeRF) from imperfect camera poses. The authors, Hongyu Fu, Xin Yu, Lincheng Li, and Li Zhang, address the limitations of existing volumetric neural rendering techniques in synthesizing high-quality novel views when camera poses are inaccurate. CBARF optimizes camera poses in a coarse-to-fine manner and employs a density voxel grid to generate high-quality 3D reconstructed scenes and images in novel views. The framework introduces a neighbor-replacement strategy and a novel criterion to identify and rectify poorly estimated camera poses. This method significantly improves the performance of bundle-adjustment and novel view synthesis, especially in scenarios with large camera pose noise. The approach is significant for applications in 3D reconstruction and virtual reality, where accurate camera pose estimation is crucial for high-quality scene rendering. CBARF represents a significant advancement in handling imperfect camera poses in NeRF-based scene reconstruction.

    \item \textbf{Active Perception using Neural Radiance Fields} explores a novel approach to active perception in autonomous agents, focusing on maximizing mutual information between past and future observations. The authors, Siming He, Christopher D. Hsu, Dexter Ong, Yifei Simon Shao, Pratik Chaudhari, and their team, propose a framework that integrates a neural radiance field (NeRF)-like representation with planning and state estimation. This representation captures photometric, geometric, and semantic properties of the scene, making it suitable for synthesizing new observations from different viewpoints. The authors utilize a sampling-based planner to calculate predictive information from synthetic observations along dynamically-feasible trajectories. The approach is demonstrated in realistic 3D indoor environments for exploring cluttered spaces, employing semantic uncertainty to assess the completion of exploration tasks. This method is particularly relevant for search and rescue operations, where quick and comprehensive exploration of unknown environments is crucial. The paper represents a significant advancement in active perception, offering insights into various fields where autonomous exploration and information discovery are essential. The code for this research is available on GitHub, contributing to the broader research community.

    \item \textbf{ProteusNeRF: Fast Lightweight NeRF Editing using 3D-Aware Image Context} introduces a novel framework for interactive editing of Neural Radiance Fields (NeRF) assets. Developed by Binglun Wang, Niladri Shekhar Dutt, and Niloy J. Mitra from University College London, this framework, named ProteusNeRF, offers a fast and efficient solution for editing NeRF assets using existing image manipulation tools. The key innovation lies in its 3D-aware image context, which links edits across multiple views, enabling text-guided edits that are both quick and lightweight. ProteusNeRF stands out for its simple yet effective neural network architecture that maintains a low memory footprint while allowing incremental guidance through user-friendly image-based edits. The representation facilitates straightforward object selection via semantic feature distillation at the training stage. More importantly, it introduces a local 3D-aware image context for view-consistent image editing, which can then be distilled into fine-tuned NeRFs through geometric and appearance adjustments. The authors demonstrate the effectiveness of ProteusNeRF on various examples, showcasing appearance and geometric edits, and report a significant speedup over concurrent work focusing on text-guided NeRF editing. This advancement is significant for applications in virtual reality, augmented reality, and digital content creation, where interactive and efficient editing of photorealistic 3D objects is crucial. Further details and video results are available on the project webpage.

    \item \textbf{DYNVIDEO-E: Harnessing Dynamic NeRF for Large-Scale Motion-and View-Change Human-Centric Video Editing} introduces a groundbreaking approach to video editing, particularly for human-centric videos with large-scale motion and view changes. The authors, Jia-Wei Liu, Yan-Pei Cao, Jay Zhangjie Wu, Weijia Mao, Yuchao Gu, Rui Zhao, Jussi Keppo, Ying Shan, and Mike Zheng Shou, propose using dynamic Neural Radiance Fields (NeRF) as a human-centric video representation. This approach allows editing to be performed in 3D space and propagated to the entire video via a deformation field. Key features of their method include multi-view multi-pose Score Distillation Sampling (SDS) from both 2D personalized diffusion priors and 3D diffusion priors, reconstruction losses on the reference image, text-guided local parts super-resolution, and style transfer for 3D background space. Extensive experiments demonstrate that DynVideo-E significantly outperforms state-of-the-art approaches on challenging datasets in terms of human preference. This advancement is significant for applications in video production and editing, where handling complex human-centric videos with large-scale motion and view changes is crucial. The project page and code will be released to the community, contributing further to the research field.

    \item \textbf{TraM-NeRF: Tracing Mirror and Near-Perfect Specular Reflections through Neural Radiance Fields} introduces a novel reflection tracing method tailored for Neural Radiance Fields (NeRF) to handle mirror-like and near-perfect specular reflecting objects. The authors, Leif Van Holland, Ruben Bliersbach, Jan U. Müller, Patrick Stotko, and Reinhard Klein from the University of Bonn, address the challenge of representing such objects in NeRF, which often leads to severe artifacts in synthesized renderings. Their approach explicitly models the reflection behavior using physically plausible materials and estimates the reflected radiance with Monte-Carlo methods within the volume rendering formulation. This method involves efficient strategies for importance sampling and transmittance computation along rays from only a few samples. The authors demonstrate that their novel method enables the training of consistent representations of challenging scenes and achieves superior results compared to previous state-of-the-art approaches. This advancement is significant for applications in video gaming, movies, advertisement, education, and AR/VR scenarios, where photorealistic rendering of complex scenes with specular reflections is crucial. The paper represents a significant step in enhancing the capabilities of NeRF in handling complex reflection phenomena in various indoor and outdoor scenes.

    \item \textbf{Towards Abdominal 3-D Scene Rendering from Laparoscopy Surgical Videos using NeRFs} presents a novel approach to enhance laparoscopic surgery by reconstructing three-dimensional (3-D) anatomical structures of the abdomen from laparoscopic images and videos. The authors, Khoa Tuan Nguyen, Francesca Tozzi, Nikdokht Rashidian, Wouter Willaert, Joris Vankerschaver, and Wesley De Neve, explore the use of Neural Radiance Fields (NeRFs) for this purpose. NeRFs have gained attention for their ability to generate photorealistic images from a 3-D static scene, offering a more comprehensive exploration of the abdomen through the synthesis of new views. This method stands out from other techniques like Simultaneous Localization and Mapping (SLAM) and depth estimation. The paper presents a comprehensive examination of NeRFs in the context of laparoscopy surgical videos, aiming to render abdominal scenes in 3-D. While the experimental results are promising, the authors acknowledge substantial challenges that require further exploration in future research. This advancement is significant for medical applications, particularly in minimally invasive surgery, where enhanced visualization can aid in the detection and diagnosis of medical conditions.

    \item \textbf{SPEC-NERF: Multi-Spectral Neural Radiance Fields} introduces a groundbreaking approach for reconstructing multispectral radiance fields and camera spectral sensitivity functions (SSFs) from color images filtered by different filters. The authors, Jiabao Li, Yuqi Li, Ciliang Sun, Chong Wang, and Jinhui Xiang from Ningbo University, propose Spec-NeRF, which focuses on modeling the physical imaging process. This method requires only a low-cost trichromatic camera and several off-the-shelf color filters, making it more practical than using specialized 3D scanning and spectral imaging equipment. Spec-NeRF applies the estimated SSFs and radiance field to synthesize novel views of multispectral scenes. The experiments on both synthetic and real scenario datasets demonstrate that utilizing filtered RGB images with learnable NeRF and SSFs can achieve high fidelity and promising spectral reconstruction while retaining NeRF's inherent capability to comprehend geometric structures. This advancement is significant for applications in plant modeling, agriculture surveillance, preservation of digital cultural heritage, and material classification, where accurate spectral and geometric information is crucial. The code for Spec-NeRF is available at the provided GitHub link, contributing further to the research community.
    
    \item \textbf{SIRe-IR: Inverse Rendering for BRDF Reconstruction with Shadow and Illumination Removal in High-Illuminance Scenes} introduces a novel approach for inverse rendering in scenes with intense illumination and significant shadows. The authors, Ziyi Yang, Yanzhen Chen, Xinyu Gao, Yazhen Yuan, Yu Wu, Xiaowei Zhou, and Xiaogang Jin from Zhejiang University and Tencent, present SIRe-IR, an implicit inverse rendering method that effectively decomposes a scene into environment map, albedo, and roughness. This method uses non-linear mapping and regularized visibility estimation to accurately model the indirect radiance field, normal, visibility, and direct light simultaneously. SIRe-IR excels in removing both shadows and indirect illumination in materials, even under intense lighting conditions, leading to high-quality albedo and roughness reconstruction without shadow interference. The approach outperforms existing methods in both quantitative and qualitative evaluations. This advancement is significant for applications in computer graphics and computer vision, where precise factorization of scene geometry, materials, and lighting from 2D images is crucial, especially in challenging lighting conditions. The authors plan to release their code, contributing further to the research community.

    \item \textbf{UE4-NeRF: Neural Radiance Field for Real-Time Rendering of Large-Scale Scene} introduces a novel neural rendering system, UE4-NeRF, designed for real-time rendering of large-scale scenes. The authors, Jiaming Gu, Minchao Jiang, Hongsheng Li, Xiaoyuan Lu, Guangming Zhu, Syed Afaq Ali Shah, Liang Zhang, and Mohammed Bennamoun, propose a method that partitions large scenes into sub-NeRFs, represented by polygonal meshes. These meshes are initialized with multiple regular octahedra within the scene, and their vertices are continuously optimized during training. The system employs Level of Detail (LOD) techniques to train meshes with varying levels of detail for different observation levels. Integrating with the rasterization pipeline in Unreal Engine 4 (UE4), UE4-NeRF achieves real-time rendering of large-scale scenes at 4K resolution with frame rates of up to 43 FPS. The experimental results show that this method attains rendering quality comparable to state-of-the-art approaches, while also offering real-time performance and facilitating scene editing in UE4. This advancement is significant for applications in virtual reality, computer games, and the Metaverse, where real-time interactive rendering of large-scale 3D scenes is crucial. The project page provides further details and resources related to this research.

    \item \textbf{Sync-NeRF: Generalizing Dynamic NeRFs to Unsynchronized Videos} introduces a novel approach to address the challenge of reconstructing dynamic scenes from unsynchronized multi-view videos using Neural Radiance Fields (NeRF). The authors, Seoha Kim, Jeongmin Bae, Youngsik Yun, Hahyun Lee, Gun Bang, and Youngjung Uh, propose Sync-NeRF, which incorporates time offsets for individual unsynchronized videos and jointly optimizes these offsets with NeRF. This method significantly improves the performance of various baseline models and naturally works as a means of synchronizing videos without manual effort. The research is motivated by the common issue of inaccurate video synchronization in multi-view dynamic datasets. Sync-NeRF addresses this by assuming that the same frames in multi-view videos are captured at different times, contrary to existing dynamic NeRFs that assume simultaneous capture. The experiments conducted on the Plenoptic Video Dataset and a newly built Unsynchronized Dynamic Blender Dataset demonstrate the effectiveness of Sync-NeRF in handling unsynchronized videos. This advancement is significant for applications in virtual reality, augmented reality, and 3D modeling, where accurate reconstruction of dynamic scenes from multi-view videos is crucial, especially in scenarios with unsynchronized data capture. The project page provides further details and resources related to this research.

    \item \textbf{Manifold NeRF: View-Dependent Image Feature Supervision for Few-Shot Neural Radiance Fields} presents a novel method, ManifoldNeRF, to enhance novel view synthesis using Neural Radiance Fields (NeRF) from a limited number of images. The authors, Daiju Kanaoka, Motoharu Sonogashira, Hakaru Tamukoh, and Yasutomo Kawanishi, address the challenge of training NeRF with few images, particularly in complex scenes. ManifoldNeRF introduces a new approach for supervising feature vectors at unknown viewpoints by interpolating features from neighboring known viewpoints. This method provides more appropriate supervision for each unknown viewpoint, leading to better learning of the volume representation compared to DietNeRF, a previous extension of NeRF. The authors conducted experiments to demonstrate the superior performance of ManifoldNeRF and explored various subsets of viewpoints to identify effective viewpoint patterns for real-world applications. This advancement is significant for applications in virtual reality, augmented reality, and 3D modeling, where generating high-quality images from limited viewpoints is crucial. The proposed method represents a significant step in improving the practicality and effectiveness of NeRF in few-shot scenarios, particularly in complex environments. The code for ManifoldNeRF is available online, contributing further to the research community.

    \item \textbf{CAwa-NeRF: Instant Learning of Compression-Aware NeRF Features} presents a novel approach to address the storage size challenge in Neural Radiance Fields (NeRF) models. The authors, Omnia Mahmoud, Théo Ladune, and Matthieu Gendrin from Orange Innovation, France, focus on learning high-quality neural graphics primitives with reduced storage requirements. They introduce an entropy-aware training scheme with a rate-distortion trade-off, inspired by COOL-CHIC, to learn low entropy features throughout the training. This method allows for exporting zip-compressed feature grids at the end of model training with negligible extra time overhead, without altering the storage architecture or parameters used in the original Instant-NGP (INGP) paper. The proposed method is adaptable to any model and has been tested extensively, showing impressive results on different kinds of static scenes. For instance, in single object masked background scenes, CAwa-NeRF compresses the feature grids to a fraction of the original size without any loss in Peak Signal-to-Noise Ratio (PSNR). This advancement is significant for applications with storage constraints, particularly in streaming, network, and cloud applications, where high-quality NeRF models are desired but limited by storage capacity. The approach represents a significant step in making NeRFs more accessible and practical for a wider range of applications.

    \item \textbf{Cross-view Self-localization from Synthesized Scene-graphs} introduces a novel approach for enhancing visual place recognition in scenarios where database images are provided from sparse viewpoints. The authors, Ryogo Yamamoto and Kanji Tanaka, focus on synthesizing database images from unseen viewpoints using Neural Radiance Fields (NeRF) technology. They explore a hybrid scene model that combines view-invariant appearance features computed from raw images with view-dependent spatial-semantic features computed from synthesized images. These features are then fused into scene graphs, which are learned and recognized by a graph neural network. This method addresses the challenges of storing a large number of synthesized images and the assumption that database images are available from multiple viewpoints similar to the unseen viewpoint of the synthesized image. The effectiveness of the proposed method was verified using a novel cross-view self-localization dataset with many unseen views generated using a photorealistic Habitat simulator. This advancement is significant for applications in autonomous driving and robotics, where robust place classification at very different viewpoints is crucial. The approach represents a significant step in visual place recognition, particularly in scenarios with severe viewpoint changes.

    \item \textbf{Open-NeRF: Towards Open Vocabulary NeRF Decomposition} addresses the challenge of decomposing Neural Radiance Fields (NeRF) into objects from an open vocabulary, a critical task for object manipulation in 3D reconstruction and view synthesis. The authors, Hao Zhang, Fang Li, and Narendra Ahuja from the University of Illinois Urbana-Champaign, present Open-vocabulary Embedded Neural Radiance Fields (Open-NeRF). This method leverages large-scale, off-the-shelf segmentation models like the Segment Anything Model (SAM) and introduces an integrate-and-distill paradigm with hierarchical embeddings. Open-NeRF first utilizes these models to generate hierarchical 2D mask proposals from varying viewpoints, which are then aligned and integrated within the 3D space, subsequently distilled into the 3D field. This process ensures consistent recognition and granularity of objects from different viewpoints, even in challenging scenarios involving occlusion and indistinct features. The experimental results show that Open-NeRF outperforms state-of-the-art methods in open-vocabulary scenarios, offering a promising solution to NeRF decomposition guided by open-vocabulary queries. This advancement is significant for applications in robotics and vision-language interaction in open-world 3D scenes, enabling novel applications where flexible and accurate object manipulation in 3D space is crucial.

    \item \textbf{PERF: Panoramic Neural Radiance Field from a Single Panorama} presents a groundbreaking framework for 360-degree novel view synthesis, enabling 3D roaming in complex scenes from a single panorama. The authors, Guangcong Wang, Peng Wang, Zhaoxi Chen, Wenping Wang, Chen Change Loy, and Ziwei Liu, have developed PERF, which overcomes the limitations of previous works that focused on a limited field of view with few occlusions. PERF introduces a collaborative RGBD inpainting method and a progressive inpainting-and-erasing method to elevate a 360-degree 2D scene to a 3D scene. The process begins with predicting a panoramic depth map from a single panorama, followed by reconstructing visible 3D regions with volume rendering. A collaborative RGBD inpainting approach, derived from an RGB Stable Diffusion model and a monocular depth estimator, is integrated into a NeRF for completing RGB images and depth maps from random views. Finally, an inpainting-and-erasing strategy is introduced to ensure consistent geometry between newly-sampled and reference views. The integration of these components into the learning of NeRFs achieves promising results, as demonstrated by extensive experiments on Replica and a new dataset, PERF-in-the-wild. PERF's application potential spans panorama-to-3D, text-to-3D, and 3D scene stylization, offering a significant advancement for real-world applications requiring 360-degree scene understanding and interaction. The project page and code are available online, contributing further to the research community.

    \item \textbf{4D-Editor: Interactive Object-level Editing in Dynamic Neural Radiance Fields via Semantic Distillation} introduces a novel framework, 4D-Editor, for interactive object-level editing in dynamic scenes represented by Neural Radiance Fields (NeRF). The authors, Dadong Jiang, Zhihui Ke, Xiaobo Zhou, Tie Qiu, and Xidong Shi from Tianjin University, have developed a method that allows users to interactively edit multiple objects within a dynamic NeRF using simple user strokes on a single frame. The key innovation of 4D-Editor is the incorporation of a hybrid semantic feature distillation into the original dynamic NeRF, which maintains spatial-temporal consistency after editing. Additionally, the framework includes Recursive Selection Refinement for enhancing object segmentation accuracy and Multi-view Reprojection Inpainting to address holes caused by incomplete scene capture. The extensive experiments and real-world editing examples demonstrate that 4D-Editor achieves photorealistic editing results on dynamic NeRFs. This advancement is significant for applications in VR/AR, computer animation, and education, where flexible and realistic editing of dynamic scenes is crucial. The 4D-Editor represents a significant step forward in the field of interactive 3D scene editing, offering a practical solution for object-level manipulation in dynamic environments.

    \item \textbf{HyperFields: Towards Zero-Shot Generation of NeRFs from Text} introduces a groundbreaking method for generating text-conditioned Neural Radiance Fields (NeRFs) using a dynamic hypernetwork. The authors, Sudarshan Babu, Richard Liu, Avery Zhou, Michael Maire, Greg Shakhnarovich, and Rana Hanocka, have developed HyperFields, which maps text token embeddings to the space of NeRFs. This method involves a novel approach of NeRF distillation training, where scenes encoded in individual NeRFs are distilled into one dynamic hypernetwork, enabling the network to fit over a hundred unique scenes. HyperFields can predict novel in-distribution and out-of-distribution scenes either zero-shot or with a few fine-tuning steps. This fine-tuning benefits from accelerated convergence due to the learned general map, synthesizing novel scenes much faster than existing neural optimization-based methods. The authors' experiments demonstrate that both the dynamic architecture and NeRF distillation are crucial to the expressivity of HyperFields. This advancement is significant for applications in virtual reality, gaming, and digital content creation, where generating 3D scenes from textual descriptions enhances creativity and user experience. HyperFields represents a significant step in text-to-3D synthesis, offering a more efficient and expressive way to create complex 3D scenes from text prompts.

    \item \textbf{Reconstructive Latent-Space Neural Radiance Fields for Efficient 3D Scene Representations} introduces a novel approach to enhance Neural Radiance Fields (NeRFs) by integrating an autoencoder (AE) for rendering latent features instead of colors. The authors, Tristan Aumentado-Armstrong, Ashkan Mirzaei, Marcus A. Brubaker, Jonathan Kelly, Alex Levinshtein, Konstantinos G. Derpanis, and Igor Gilitschenski, propose a latent-space NeRF that can produce novel views with higher quality than standard color-space NeRFs. This method corrects certain visual artifacts and renders over three times faster. The approach is orthogonal to other techniques for improving NeRF efficiency and allows control over the tradeoff between efficiency and image quality by adjusting the AE architecture. The authors aim for their approach to form the basis of an efficient, yet high-fidelity, 3D scene representation for downstream tasks, especially in robotics scenarios requiring continual learning. This advancement is significant for applications in graphics, vision, and robotics, where problems with slow rendering speed and visual artifacts in NeRFs have hindered broader adoption. The proposed method represents a significant step in addressing these limitations, enhancing the potential of NeRFs for real-time and high-quality 3D scene representation.

    \item \textbf{INCODE: Implicit Neural Conditioning with Prior Knowledge Embeddings} introduces a novel approach to enhance Implicit Neural Representations (INRs) by incorporating deep prior knowledge. The authors, Amirhossein Kazerouni, Reza Azad, Alireza Hosseini, Dorit Merhof, Ulas Bagci, and their respective teams from various universities and institutes, have developed INCODE, which comprises a harmonizer network and a composer network. The harmonizer dynamically adjusts key parameters of the activation function in INRs, while the composer network optimizes the representation process using task-specific pre-trained models. This method excels in representing complex data and extends its capabilities to tackle intricate tasks such as audio, image, and 3D shape reconstructions, neural radiance fields (NeRFs), and inverse problems like denoising, super-resolution, inpainting, and CT reconstruction. INCODE demonstrates superior robustness, accuracy, quality, and convergence rate in comprehensive experiments, significantly broadening the scope of signal representation. This advancement is significant for applications in computer graphics, computer vision, virtual reality, and other fields where complex and high-dimensional data representation is crucial. The project's website provides further details and access to the code, marking a significant step in enhancing the potential of INRs in diverse signal-processing contexts.

    \item \textbf{TiV-NeRF: Tracking and Mapping via Time-Varying Representation with Dynamic Neural Radiance Fields} introduces a novel dynamic SLAM (Simultaneous Localization and Mapping) system capable of tracking camera poses and reconstructing moving objects in dynamic scenes. The authors, Chengyao Duan and Zhiliu Yang, propose a time-varying representation that extends the 3D positions of objects to 4D, incorporating time as an additional dimension. This system, named TiV-NeRF, processes RGB-D image sequences with timestamps and segmentation masks of dynamic objects. It employs a canonical field where points on dynamic objects are transformed from a deformation field, and colors and Signed Distance Function (SDF) of dynamic objects are regressed by a Multilayer Perceptron (MLP). The framework maintains two simultaneous processes: tracking and mapping, executed in turn. TiV-NeRF's novel approach addresses the limitations of traditional dynamic mapping frameworks, such as holes and ghost trail effects, by capturing position offsets through its time-varying representation. Additionally, it introduces an overlap-based keyframe selection strategy to reconstruct more complete dynamic objects. This advancement is significant for tasks like auto-cars navigation and mixed reality, where accurate dense mapping of dynamic environments is crucial. TiV-NeRF represents a significant step forward in the SLAM domain, particularly in dynamic scenarios, enhancing the performance of dynamic scenes reconstruction.

    \item \textbf{DynPoint: Dynamic Neural Point For View Synthesis} introduces a novel algorithm, DynPoint, designed to enhance the synthesis of novel views for unconstrained monocular videos. The authors, Kaichen Zhou, Jia-Xing Zhong, Sangyun Shin, Kai Lu, Yiyuan Yang, Andrew Markham, and Niki Trigoni from the University of Oxford, address the challenges faced by existing neural radiance field algorithms in dealing with uncontrolled or lengthy scenarios, which often require extensive training time specific to each new scenario. DynPoint focuses on predicting explicit 3D correspondence between neighboring frames, rather than encoding the entirety of the scenario information into a latent representation. This is achieved through the estimation of consistent depth and scene flow information across frames, which is then used to aggregate information from multiple reference frames to a target frame by constructing hierarchical neural point clouds. The result is a framework that enables swift and accurate view synthesis for desired views of target frames. Experimental results demonstrate that DynPoint significantly accelerates training time, typically by an order of magnitude, while yielding comparable outcomes to prior approaches. Additionally, it exhibits strong robustness in handling long-duration videos without the need for learning a canonical representation of video content. This advancement is significant for applications in computer vision, particularly in artificial reality and machine comprehension of appearance and geometric properties in dynamic real-world scenarios.

    \item \textbf{NeRF Revisited: Fixing Quadrature Instability in Volume Rendering} addresses a critical issue in Neural Radiance Fields (NeRF) related to volume rendering, specifically the instability caused by the choice of samples along a ray, termed quadrature instability. The authors, Mikaela Angelina Uy, George Kiyohiro Nakayama, Guandao Yang, Rahul Krishna Thomas, Leonidas Guibas, and Ke Li, propose a mathematically principled solution by reformulating the sample-based rendering equation to correspond to the exact integral under piecewise linear volume density. This approach resolves multiple issues inherent in classical sample-based rendering equations, such as conflicts between samples along different rays, imprecise hierarchical sampling, and non-differentiability of quantiles of ray termination distances with respect to model parameters. The proposed formulation offers several benefits, including sharper textures, better geometric reconstruction, and stronger depth supervision. It can also be used as a drop-in replacement for the volume rendering equation in existing methods like NeRFs. This advancement is significant for applications in neural rendering, enhancing the stability and quality of synthesized novel views in NeRF-based systems. The authors' work represents a crucial step in addressing the limitations of NeRF, particularly in scenarios requiring high-quality rendering and geometric reconstruction.

    \item \textbf{FPO++: Efficient Encoding and Rendering of Dynamic Neural Radiance Fields by Analyzing and Enhancing Fourier PlenOctrees} presents a significant advancement in the real-time rendering of dynamic Neural Radiance Fields (NeRF). The authors, Saskia Rabich, Patrick Stotko, Reinhard Klein, and their team from the University of Bonn, focus on addressing the artifacts introduced by compression in Fourier PlenOctrees, a method used for efficient representation in NeRF. Through an in-depth analysis of these artifacts, they propose an improved representation, particularly a novel density encoding that adapts the Fourier-based compression to the characteristics of the transfer function used in volume rendering. This leads to a substantial reduction of artifacts in the dynamic model. Additionally, they introduce a training data augmentation that relaxes the periodicity assumption of the compression. The effectiveness of these enhanced Fourier PlenOctrees is demonstrated through quantitative and qualitative evaluations on synthetic and real-world scenes. This work is significant for applications in augmented reality (AR) and virtual reality (VR), advertisement, and entertainment, where photorealistic rendering of dynamic real-world scenes is crucial. The proposed method represents a breakthrough towards synthesizing photorealistic novel views of complex static scenes from posed multi-view images recorded by commodity cameras, addressing limitations in training times and rendering performance.

    \item \textbf{Single-view 3D Scene Reconstruction with High-fidelity Shape and Texture} introduces a novel framework for reconstructing detailed 3D scenes from single-view images, addressing the challenges of accurately recovering both object shapes and textures. The authors, Yixin Chen, Junfeng Ni, Nan Jiang, Yaowei Zhang, Yixin Zhu, and Siyuan Huang, propose a method that utilizes Single-view neural implicit Shape and Radiance field (SSR) representations. This approach leverages explicit 3D shape supervision and volume rendering of color, depth, and surface normal images. To overcome the shape-appearance ambiguity inherent in single-view reconstructions, the authors introduce a two-stage learning curriculum incorporating both 3D and 2D supervisions. A key feature of their framework is its ability to generate fine-grained textured meshes and integrate rendering capabilities into the single-view 3D reconstruction model. This integration results in significantly improved textured 3D object reconstruction and supports rendering images from novel viewpoints. The method also enables composing object-level representations into flexible scene representations, facilitating applications in holistic scene understanding and 3D scene editing. Extensive experiments demonstrate the effectiveness of this method, which holds immense importance for applications in virtual reality, augmented reality, and robotics, where understanding and interacting with the real 3D world is crucial.

    \item \textbf{Novel View Synthesis from a Single RGBD Image for Indoor Scenes} presents an innovative approach for synthesizing novel view images from a single RGBD (Red Green Blue-Depth) input. The authors, Congrui Hetang and Yuping Wang, focus on the challenging task of Novel View Synthesis (NVS) in computer vision, which has extensive applications but is often limited by the need for multiple images or computationally intensive methods. Their method converts an RGBD image into a point cloud and renders it from a different viewpoint, formulating the NVS task into an image translation problem. They leverage generative adversarial networks, specifically using CycleGAN for unsupervised learning and Pix2Pix for supervised learning, to style-transfer the rendered image to achieve a result similar to a photograph taken from the new perspective. This method circumvents the limitations of traditional multi-image techniques and holds significant promise for practical, real-time applications in NVS. The approach is validated on the SUN3D dataset, demonstrating its effectiveness and potential for applications in virtual or augmented reality, scene understanding and editing, and simulation for robotics and autonomous vehicles. This work represents a significant advancement in NVS, offering a simpler and more streamlined approach for generating novel views from single-shot images.

    \item \textbf{INeAT: Iterative Neural Adaptive Tomography} introduces a novel neural rendering method for Computed Tomography (CT) reconstruction, addressing the challenges of posture perturbations and pose shifts encountered in CT scanning processes. The authors, Bo Xiong, Changqing Su, Zihan Lin, You Zhou, and Zhaofei Yu, propose Iterative Neural Adaptive Tomography (INeAT), which incorporates iterative posture optimization to counteract the influence of posture variations in data. INeAT refines the posture corresponding to input images iteratively based on the reconstructed 3D volume, using a posture feedback optimization strategy. This method demonstrates artifact-suppressed and resolution-enhanced reconstruction in scenarios with significant pose disturbances. INeAT maintains comparable reconstruction performance even with data from unstable-state acquisitions, reducing the time required for CT scanning and relaxing the requirements on imaging hardware systems. This advancement is significant for applications in short-time and low-cost CT technology, particularly in medical imaging, scientific observation, and industrial detection. INeAT's ability to handle substantial posture perturbations and enhance reconstruction quality underscores its potential for broadening the scope and efficiency of CT technology in various fields.

    \item \textbf{Efficient Cloud Pipelines for Neural Radiance Fields} explores the development of cloud pipelines for generating Neural Radiance Fields (NeRFs), a technology that has gained significant attention in the computer vision community for its multi-view representation of scenes or objects. The authors, Derek Jacoby, Donglin Xu, Weder Ribas, Minyi Xu, Ting Liu, Vishwanath Jeyaraman, Mengdi Wei, Emma De Blois, Yvonne Coady, and others, focus on the computational cost of these generative AI models and the necessity of constructing cloud pipelines for their potential realization in client applications. The paper documents the process of collecting aerial imagery data and processing it into point clouds and NeRFs on two platforms: Compute Canada (Digital Research Alliance of Canada) and Microsoft Azure. The research highlights the advantages of commercial cloud platforms like Azure over academic High-Performance Computing (HPC) clusters, such as on-demand scalability, cost-effectiveness, and global accessibility. The infrastructure targets large-scale datasets, like urban geometry acquired by remote sensing, and processing pipelines to utilize this data. NeRFs offer a perceptually-based storage mechanism supportive of photorealistic renderings in multi-level geospatial datasets, used for efficient representations of large urban scenes. This work is significant for applications in geospatial analytics, virtual production, and other fields where large-scale, photorealistic rendering of complex scenes is crucial. The technology lends itself to powerful experiences in virtual production through NeRF integrations in platforms like Unity and Unreal Engine, showcasing a shift towards more efficient and scalable methods of representing and processing complex visual data.

    \item \textbf{Estimating 3D Uncertainty Field: Quantifying Uncertainty for Neural Radiance Fields} introduces a novel approach to quantify uncertainty in Neural Radiance Fields (NeRF), particularly focusing on unseen space, including occluded and outside scene content. The authors, Jianxiong Shen, Ruijie Ren, Adria Ruiz, and Francesc Moreno-Noguer, propose a method to estimate a 3D Uncertainty Field based on learned incomplete scene geometry. This field explicitly identifies unseen regions by considering accumulated transmittance along each camera ray, inferring 2D pixel-wise uncertainty with high values for rays casting towards occluded or outside scene content. Additionally, the authors model a stochastic radiance field to quantify uncertainty on the learned surface. The experiments demonstrate that this approach is unique in explicitly reasoning about high uncertainty in both 3D unseen regions and the involved 2D rendered pixels, compared to recent methods. The designed uncertainty field is ideally suited for real-world robotics tasks, such as next-best-view selection. This advancement is significant for applications in robotics, where understanding the reliability of model predictions is crucial for tasks like exploration and planning in unknown environments. The method represents a crucial step in enhancing the safety and efficiency of robotic systems by providing a more comprehensive understanding of the uncertainty associated with NeRF-based predictions in unobserved areas.

    \item \textbf{A Neural Radiance Field-Based Architecture for Intelligent Multilayered View Synthesis} introduces an innovative approach for enhancing mobile ad hoc networks (MANETs) using a Neural Radiance Field-based architecture. The authors, D. Dhinakaran, S. M. Udhaya Sankar, G. Elumalai, and N. Jagadish Kumar, focus on improving on-demand source routing systems in MANETs, which consist of a large and dense community of mobile nodes that communicate solely through wireless interfaces. The proposed protocol, Optimized Route Selection via Red Imported Fire Ants (RIFA) Strategy, aims to identify the least-expensive nominal capacity acquisition that ensures the transportation of realistic transport and its durability in the event of any node failure. The method involves predicting route failure and energy utilization to select the path during the routing phase. The results of the proposed work are assessed based on performance parameters like energy usage, packet delivery rate (PDR), and end-to-end (E2E) delay. The outcome demonstrates that the proposed strategy is preferable, increasing network lifetime while reducing node energy consumption and typical E2E delay under most network performance measures and factors. This advancement is significant for applications in wireless communication and networking, where efficient and reliable routing is crucial for the stability and performance of mobile ad hoc networks. The approach represents a crucial step in enhancing the robustness and efficiency of MANETs, particularly in challenging environments where dynamic topology and limited resources are prevalent.
    
    \item \textbf{VR-NeRF: High-Fidelity Virtualized Walkable Spaces} presents an end-to-end system for capturing, reconstructing, and real-time rendering of walkable spaces in virtual reality using neural radiance fields. The authors, Linning Xu, Vasu Agrawal, William Laney, Tony Garcia, Aayush Bansal, Changil Kim, Samuel Rota Bulò, Lorenzo Porzi, Peter Kontschieder, Aljaž Božič, Dahua Lin, Michael Zollhöfer, and Christian Richardt, have developed a custom multi-camera rig for densely capturing walkable spaces with high image resolution and dynamic range. They extend instant neural graphics primitives with a novel perceptual color space for learning accurate HDR appearance and an efficient mip-mapping mechanism for level-of-detail rendering with anti-aliasing. The multi-GPU renderer enables high-fidelity volume rendering of the neural radiance field model at full VR resolution of dual 2K×2K at 36 Hz. The system demonstrates high-quality results on challenging high-fidelity datasets and compares favorably against existing baselines. This advancement is significant for virtual reality applications, offering an immersive VR experience with high-fidelity walkable spaces. The release of their dataset further contributes to the research community, providing a valuable resource for developing and testing advanced VR technologies. This work represents a substantial step in enhancing the realism and interactivity of virtual environments, crucial for applications in gaming, simulation, and virtual tours.
    
    \item \textbf{Instruct Pix2NeRF: Instructed 3D Portrait Editing from a Single Image} presents a groundbreaking end-to-end diffusion-based framework for 3D-aware portrait editing using natural language instructions. The authors, Jianhui Li, Shilong Liu, Zidong Liu, Yikai Wang, Kaiwen Zheng, Jinghui Xu, Jianmin Li, and Jun Zhu, have developed InstructPix2NeRF, which enables instructed 3D-aware portrait editing from a single open-world image. The core of this framework is a conditional latent 3D diffusion process that lifts 2D editing to 3D space by learning the correlation between the paired images’ difference and the instructions via triplet data. A novel token position randomization strategy is introduced, allowing multi-semantic editing in a single pass while preserving the portrait identity. Additionally, an identity consistency module is proposed to modulate the extracted identity signals into the diffusion process, enhancing multi-view 3D identity consistency. Extensive experiments validate the effectiveness of InstructPix2NeRF, demonstrating its superiority against strong baselines both quantitatively and qualitatively. This method represents a significant advancement in the field of human-instructed 3D-aware editing for open-world portraits, offering a new dimension of interactivity and creativity in applications such as virtual reality, gaming, and digital art, where natural language-driven 3D portrait manipulation is highly desirable.
    
    \item \textbf{CONSISTENT 4D: Consistent 360° Dynamic Object Generation from Monocular Video} introduces a novel approach, Consistent4D, for generating 4D dynamic objects from uncalibrated monocular videos. The authors, Yanqin Jiang, Li Zhang, Jin Gao, Weimin Hu, and Yao Yao, have developed a method that casts 360-degree dynamic object reconstruction as a 4D generation problem, thereby eliminating the need for multi-view data collection and camera calibration. This is achieved by leveraging an object-level 3D-aware image diffusion model as the primary supervision signal for training Dynamic Neural Radiance Fields (DyNeRF). The authors propose a Cascade DyNeRF to facilitate stable convergence and temporal continuity under the supervision signal, which is discrete along the time axis. To achieve spatial and temporal consistency, they introduce an Interpolation-driven Consistency Loss, optimized by minimizing the discrepancy between rendered frames from DyNeRF and interpolated frames from a pre-trained video interpolation model. Extensive experiments show that Consistent4D performs competitively with prior art alternatives, offering new possibilities for 4D dynamic object generation from monocular videos. This method also demonstrates advantages for conventional text-to-3D generation tasks. This advancement is significant for applications in virtual content creation, autonomous driving simulation, and medical image analysis, where dynamic 3D information recovery from monocular video observations is crucial.
    
    \item \textbf{Animating NeRFs from Texture Space: A Framework for Pose-Dependent Rendering of Human Performances} introduces a novel NeRF-based framework for pose-dependent rendering of human performances. The authors, Paul Knoll, Wieland Morgenstern, Anna Hilsmann, and Peter Eisert, have developed a method where the radiance field is warped around an SMPL body mesh, creating a new surface-aligned representation. This representation can be animated through skeletal joint parameters provided to the NeRF in addition to the viewpoint for pose-dependent appearances. The method includes the corresponding 2D UV coordinates on the mesh texture map and the distance between the query point and the mesh. To address mapping ambiguities and random visual variations, the authors introduce a novel remapping process that refines the mapped coordinates. The experiments demonstrate that this approach results in high-quality renderings for novel-view and novel-pose synthesis. This advancement is significant for applications in virtual reality, gaming, and animation, where creating high-quality controllable 3D human models from multi-view RGB videos is crucial. The framework represents a significant step in extending the capabilities of NeRFs to dynamic and controllable synthesis of human performances, enhancing the realism and interactivity of digital human representations.
    
    \item \textbf{Long-Term Invariant Local Features via Implicit Cross-Domain Correspondences} introduces a novel approach to improve visual localization in the presence of long-term visual domain variations such as different seasons and daytimes. The authors, Zador Pataki, Mohammad Altillawi, Menelaos Kanakis, Rèmi Pautrat, Fengyi Shen, Ziyuan Liu, Luc Van Gool, and Marc Pollefeys, propose Implicit Cross-Domain Correspondences (iCDC). This method represents the same environment with multiple Neural Radiance Fields, each fitting the scene under individual visual domains. iCDC utilizes the underlying 3D representations to generate accurate correspondences across different long-term visual conditions, enhancing cross-domain localization performance and significantly reducing the performance gap. When evaluated on popular long-term localization benchmarks, the trained networks consistently outperform existing methods. This work marks a substantial stride toward more robust visual localization pipelines for long-term deployments, opening up research avenues in the development of long-term invariant descriptors. The method is particularly valuable for applications in autonomous vehicles, augmented reality, and other fields where accurate visual localization across varying environmental conditions is crucial.
    
    \item \textbf{UP-NeRF: Unconstrained Pose-Prior-Free Neural Radiance Fields} introduces a novel approach to optimize Neural Radiance Field (NeRF) with unconstrained image collections without camera pose prior. The authors, Injae Kim, Minhyuk Choi, and Hyunwoo J. Kim, tackle challenges in handling unconstrained images with varying illumination and transient occluders. UP-NeRF employs surrogate tasks that optimize color-insensitive feature fields and a separate module for transient occluders to minimize their influence on pose estimation. Additionally, the method introduces a candidate head for robust pose estimation and transient-aware depth supervision to reduce the impact of incorrect prior. The experiments demonstrate superior performance of UP-NeRF compared to baselines like BARF and its variants in challenging internet photo collections, such as the Phototourism dataset. This advancement is significant for applications in virtual/augmented reality, autonomous systems, and robotics, where generating high-quality novel view images from diverse and unconstrained image sets is crucial. UP-NeRF represents a significant step in overcoming the limitations of NeRF, particularly in scenarios with varying illuminations and transient occluders, enhancing the practicality and applicability of NeRF in real-world settings.
    
    \item \textbf{Fast Sun-aligned Outdoor Scene Relighting based on TensoRF} introduces Sun-aligned Relighting TensoRF (SR-TensoRF), a method for outdoor scene relighting in Neural Radiance Fields (NeRF). The authors, Yeonjin Chang, Yearim Kim, Seunghyeon Seo, Jung Yi, and Nojun Kwak, have developed a lightweight and rapid pipeline aligned with the sun, simplifying the workflow by eliminating the need for environment maps. This approach is based on the insight that shadows, unlike viewpoint-dependent albedo, are determined by light direction. By directly using the sun direction as an input during shadow generation, the inference process is significantly simplified. Additionally, SR-TensoRF leverages the training efficiency of TensoRF, incorporating a proposed cubemap concept, resulting in notable acceleration in both training and rendering processes compared to existing methods. This advancement is significant for applications like virtual environments, entertainment media, and architectural visualization, where relighting enhances realism and versatility. SR-TensoRF addresses the unique challenges of outdoor scenes, such as the dominance of sunlight as a light source and the unbounded nature of outdoor settings, providing a more efficient and simplified approach to relighting in these contexts.
    
    \item \textbf{High-fidelity 3D Reconstruction of Plants using Neural Radiance Field} presents a novel application of Neural Radiance Field (NeRF) technology in the field of Precision Agriculture (PA) for accurate reconstruction of plant phenotypes. The authors, Kewei Hu, Ying Wei, Yaoqiang Pan, Hanwen Kang, and Chao Chen, focus on two fundamental tasks within plant phenotyping: the synthesis of 2D novel-view images and the 3D reconstruction of crop and plant models. They explore two state-of-the-art methods, Instant-NGP and Instant-NSR, the former excelling in generating high-quality images with impressive training and inference speed, and the latter improving reconstructed geometry by incorporating the Signed Distance Function (SDF) during training. The study introduces a novel plant phenotype dataset comprising real plant images from production environments, aiming to explore the advantages and limitations of NeRF in agricultural contexts. The results show that NeRF performs commendably in synthesizing novel-view images and achieves reconstruction results competitive with leading commercial software for 3D Multi-View Stereo (MVS)-based reconstruction. However, the study also highlights NeRF's drawbacks, such as slow training speeds, performance limitations in cases of insufficient sampling, and challenges in obtaining geometry quality in complex setups. This research introduces a new paradigm in plant phenotyping, offering a powerful tool capable of generating multiple representations, such as multi-view images, point clouds, and meshes, from a single process. This advancement is significant for modern sensor technologies in PA, providing non-invasive and high-throughput plant phenotyping solutions.
    
    \item \textbf{LRM: Large Reconstruction Model for Single Image to 3D} introduces the first Large Reconstruction Model (LRM) that predicts the 3D model of an object from a single input image in just 5 seconds. The authors, Yicong Hong, Kai Zhang, Jiuxiang Gu, Sai Bi, Yang Zhou, Difan Liu, Feng Liu, Kalyan Sunkavalli, Trung Bui, and Hao Tan, have developed a transformer-based architecture with 500 million learnable parameters to directly predict a neural radiance field (NeRF) from the input image. Unlike previous methods trained on small-scale datasets in a category-specific fashion, LRM is trained end-to-end on massive multi-view data containing around 1 million objects, including both synthetic renderings and real captures. This combination of a high-capacity model and large-scale training data enables LRM to be highly generalizable and produce high-quality 3D reconstructions from various testing inputs, including real-world in-the-wild captures and images from generative models. LRM represents a significant advancement in industrial design, animation, gaming, and AR/VR applications, offering a generic and efficient approach to instantly create 3D shapes from single images. This model addresses the underlying ambiguity of 3D geometry in a single view and leverages the generalization capability of 2D diffusion models for multi-view supervision, overcoming the limitations of pre-trained 2D generative models and the impracticality of per-shape optimization processes.

    \item \textbf{Learning Robust Multi-Scale Representation for Neural Radiance Fields from Unposed Images} introduces an improved solution for neural image-based rendering in computer vision, particularly for reconstructing 3D objects from unposed images. The authors, Nishant Jain, Suryansh Kumar, and Luc Van Gool, focus on synthesizing realistic images of a scene from novel viewpoints using images taken from a freely moving camera. The key ideas include recovering accurate camera parameters from unposed day-to-day images and modeling object content at different resolutions due to likely dramatic camera motion in such images. The approach leverages scene rigidity, multi-scale neural scene representation, and single-image depth prediction. It makes camera parameters learnable within a neural fields-based modeling framework, constraining relative pose between successive frames based on per view depth prediction. Absolute camera pose estimation is modeled via a graph-neural network-based multiple motion averaging within the multi-scale neural-fields network. Optimizing the introduced loss function provides camera intrinsic, extrinsic, and image rendering from unposed images. The authors demonstrate that precise camera-pose estimates are essential within the scene representation framework for accurately modeling multi-scale neural scene representation from day-to-day acquired unposed multi-view images. The approach is validated with extensive experiments on several benchmark datasets, highlighting its suitability for realistic indoor and outdoor scenes with dramatic camera motion. This advancement is significant for applications in computer vision, graphics, and robotics, where novel view synthesis from unposed images is crucial.
    
    \item \textbf{ConRad: Image Constrained Radiance Fields for 3D Generation from a Single Image} introduces a novel method for reconstructing 3D objects from a single RGB image. The authors, Senthil Purushwalkam and Nikhil Naik, have developed Image Constrained Radiance Fields (ConRad), a variant of neural radiance fields that efficiently captures the appearance of an input image from one viewpoint. ConRad addresses the challenge of aligning the appearance between the input image and 3D reconstructions, a common issue in naively extending existing methods. The training algorithm leverages a single RGB image in conjunction with pretrained Diffusion Models to optimize the parameters of the ConRad representation. Extensive experiments demonstrate that ConRad simplifies the preservation of image details while producing realistic 3D reconstructions. Compared to existing state-of-the-art baselines, ConRad shows superior fidelity to the input and produces more consistent 3D models, with significantly improved quantitative performance on a ShapeNet object benchmark. This advancement is significant for applications in robotic manipulation and navigation systems, video games, augmented reality, and e-commerce, where generating the full 3D structure from a single image is crucial. ConRad represents a significant step in capturing the capability of generating full 3D structures from limited visual information, akin to human perception.
    
    \item \textbf{VoxNeRF: Bridging Voxel Representation and Neural Radiance Fields for Enhanced Indoor View Synthesis} introduces VoxNeRF, a novel approach that combines volumetric representations with Neural Radiance Fields (NeRFs) to enhance the quality and efficiency of indoor view synthesis. The authors, Sen Wang, Wei Zhang, Stefano Gasperini, Shun-Cheng Wu, and Nassir Navab, address the challenges of creating high-quality view synthesis in indoor environments, particularly for real-time deployment. VoxNeRF constructs structured scene geometry and converts it into a voxel-based representation, employing multi-resolution hash grids to adaptively capture spatial features and manage occlusions and intricate geometry of indoor scenes. A unique voxel-guided efficient sampling technique is proposed, focusing computational resources on the most relevant portions of ray segments, thus substantially reducing optimization time. VoxNeRF outperforms state-of-the-art methods in three public indoor datasets, achieving gains in both training and rendering times, and even surpassing Instant-NGP in speed. This advancement is significant for immersive applications, such as VR/AR, where creating realistic indoor environments in real-time is crucial. VoxNeRF represents a significant step towards real-time, high-quality indoor view synthesis, combining the strengths of voxel representation and NeRFs.
    
    \item \textbf{BakedAvatar: Baking Neural Fields for Real-Time Head Avatar Synthesis} introduces a novel representation for real-time neural head avatar synthesis, deployable in a standard polygon rasterization pipeline. The authors, Hao-Bin Duan, Miao Wang, Jin-Chuan Shi, Xu-Chuan Chen, and Yan-Pei Cao, have developed a method that extracts deformable multi-layer meshes from learned isosurfaces of the head and computes expression-, pose-, and view-dependent appearances that can be baked into static textures for efficient rasterization. This approach includes a three-stage pipeline for neural head avatar synthesis, which involves learning continuous deformation, manifold, and radiance fields, extracting layered meshes and textures, and fine-tuning texture details with differential rasterization. The experimental results demonstrate that this representation generates synthesis results of comparable quality to other state-of-the-art methods while significantly reducing the inference time required. The authors showcase various head avatar synthesis results from monocular videos, including view synthesis, face reenactment, expression editing, and pose editing, all at interactive frame rates on commodity devices. This advancement is significant for applications in VR/AR, telepresence, and video games, where synthesizing photorealistic 4D human head avatars in real-time is essential for enhancing user experience and realism. "BakedAvatar" represents a crucial step in overcoming the computational limitations of existing Neural Radiance Fields (NeRF)-based methods in real-time applications.
    
    \item \textbf{UMedNeRF: Uncertainty-Aware Single View Volumetric Rendering for Medical Neural Radiance Fields} introduces a novel network for enhancing medical imaging, particularly in computed tomography (CT). The authors, Jing Hu, Qinrui Fan, Shu Hu, Siwei Lyu, Xi Wu, and Xin Wang, propose the Uncertainty-aware MedNeRF (UMedNeRF) network, which learns a continuous representation of CT projections from 2D X-ray images. This network is designed to obtain internal structure and depth information from X-ray images and uses multi-task adaptive loss weights to ensure the quality of the generated images. UMedNeRF is trained on publicly available knee and chest datasets, demonstrating its capability in rendering CT projections with a single X-ray and outperforming other methods based on generated radiation fields. The network addresses the challenge of reducing patients' exposure to ionizing radiation during CT imaging by efficiently reconstructing 3D medical images from 2D X-rays. This advancement is significant for clinical medicine, offering a safer and less invasive alternative for diagnosing various pathologies. UMedNeRF represents a crucial step in medical imaging, combining deep learning models with neural radiance fields to improve the quality and safety of diagnostic imaging.

    \item \textbf{ASSIST: Interactive Scene Nodes for Scalable and Realistic Indoor Simulation} introduces a novel object-wise neural radiance field as a panoptic representation for compositional and realistic simulation. The authors, Zhide Zhong, Jiakai Cao, Songen Gu, Sirui Xie, Weibo Gao, Liyi Luo, Zike Yan, Hao Zhao, and Guyue Zhou, have developed a unique scene node data structure that stores information of each object in a unified manner, enabling online interaction in both intra- and cross-scene settings. This structure incorporates a differentiable neural network along with bounding boxes and semantic features, allowing user-friendly interaction on independent objects to scale up novel view simulation. Users can interactively edit scenes, such as querying, adding, duplicating, deleting, transforming, or swapping objects through mouse/keyboard controls or language instructions. The experiments demonstrate the efficacy of the proposed method, where scaled realistic simulation is achieved through interactive editing and compositional rendering. The system generates color images, depth images, and panoptic segmentation masks in a 3D consistent manner. This advancement is significant for applications in robotics, where scalable and realistic data generation is crucial for robot learning. ASSIST represents a versatile approach to generating realistic data, reducing the sim-to-real gap, and enhancing the capabilities of synthetic scene simulators.
    
    \item \textbf{Aria-NeRF: Multimodal Egocentric View Synthesis} presents a novel approach to accelerate research in developing rich, multimodal scene models trained from egocentric data, inspired by Neural Radiance Fields (NeRFs). The authors, Jiankai Sun, Jianing Qiu, Chuanyang Zheng, John Tucker, Javier Yu, and Mac Schwager, focus on constructing NeRF-like models from egocentric image sequences, pivotal in understanding human behavior and applicable in VR/AR. They envision egocentric NeRF-like models as realistic simulations, contributing to the advancement of intelligent agents for real-world tasks. The future of egocentric view synthesis may include novel environment representations augmented with multimodal sensors like IMU for egomotion tracking, audio sensors, eye-gaze trackers, and more. To support this development, the authors present a comprehensive multimodal egocentric video dataset featuring RGB images, eye-tracking camera footage, audio recordings, atmospheric pressure readings, positional coordinates from GPS, connectivity details, and information from dual-frequency IMU datasets. This dataset, collected with the Meta Aria Glasses wearable device platform, serves as a robust foundation for furthering the understanding of human behavior and enabling more immersive experiences in VR, AR, and robotics. The authors evaluated two baseline NeRF-based models, Nerfacto and NeuralDiff, on this dataset, highlighting opportunities for improvement using various sensing modalities beyond vision. This work represents a significant step in creating immersive virtual environments and developing intelligent agents for real-world applications.
    
    \item \textbf{L$_{0}$-Sampler: An L$_{0}$ Model Guided Volume Sampling for NeRF} introduces an innovative upgrade to the Hierarchical Volume Sampling (HVS) strategy of Neural Radiance Fields (NeRF), aiming to enhance rendering and reconstruction tasks. The authors, Liangchen Li and Juyong Zhang, propose the L0-Sampler, which incorporates the L0 model into the weight function w(t) to guide the sampling process. This method uses piecewise exponential functions instead of piecewise constant functions for interpolation, resulting in a quasi-L0 weight distribution along rays. This approach not only approximates the L0 weight distributions more accurately but also can be easily implemented with minimal code modifications and without additional computational burden. The L0-Sampler, when applied to NeRF and its related tasks like 3D reconstruction, achieves stable performance improvements. The authors demonstrate that their method can produce more concentrated and precise sampling, leading to better results in rendering and reconstruction compared to the traditional HVS method. This advancement is significant for applications in novel view synthesis, 3D surface reconstruction, and dynamic deformation, where efficient and accurate volume rendering is crucial. The L0-Sampler represents a key step forward in improving the sampling efficiency and concentrating sampling points closer to the surface for better rendering outcomes.
    
    \item \textbf{Drivable 3D Gaussian Avatars (D3GA)} presents the first 3D controllable model for human bodies rendered with Gaussian splats, offering a significant advancement in creating photorealistic, drivable human avatars. The authors, Wojciech Zielonka, Timur Bagautdinov, Shunsuke Saito, Michael Zollhöfer, Justus Thies, and Javier Romero, utilize the 3D Gaussian Splatting (3DGS) technique to render realistic humans at real-time framerates using dense calibrated multi-view videos. D3GA departs from the traditional linear blend skinning (LBS) method for point deformation and instead employs cage deformations, a classic volumetric deformation method. These deformations are driven by joint angles and keypoints, making them more suitable for communication applications. The experiments conducted on nine subjects with varied body shapes, clothes, and motions show that D3GA achieves higher-quality results than state-of-the-art methods using the same training and test data. This work represents a significant step forward in creating photorealistic representations of humans in real-time, suitable for applications in telepresence, virtual reality, and animation, where high-quality, real-time rendering of human avatars is essential.
    
    \item \textbf{Reconstructing Continuous Light Field from Single Coded Image} proposes a method for reconstructing a continuous light field of a target scene from a single observed image. The authors, Yuya Ishikawa, Keita Takahashi, Chihiro Tsutake, and Toshiaki Fujii, integrate joint aperture-exposure coding for compressive light-field acquisition with a neural radiance field (NeRF) for view synthesis. This integration allows effective embedding of 3-D scene information into an observed image, overcoming the limitations of previous works that used joint aperture-exposure coding only for reconstructing discretized light-field views. NeRF-based neural rendering enables high-quality view synthesis of a 3-D scene from continuous viewpoints, but it struggles with quality when only a single image is given as input. The authors' method addresses this by combining these two techniques into an efficient and end-to-end trainable pipeline. Trained on a wide variety of scenes, their method can reconstruct continuous light fields accurately and efficiently without any test time optimization. This work is the first to bridge the gap between camera design for efficiently acquiring 3-D information and neural rendering. The method's ability to reconstruct continuous light fields from a single coded image has significant implications for various applications, including depth estimation, object/material recognition, view synthesis, and 3-D display, where efficient and high-quality reconstruction of 3-D scenes from limited data is crucial.
    
    \item \textbf{EvaSurf: Efficient View-Aware Implicit Textured Surface Reconstruction on Mobile Devices} is a novel method for reconstructing real-world 3D objects, particularly suited for applications in virtual reality, video games, and animations. The authors, Jingnan Gao, Zhuo Chen, Yichao Yan, Bowen Pan, Zhe Wang, Jiangjing Lyu, and Xiaokang Yang, introduce EvaSurf, an efficient view-aware implicit textured surface reconstruction method optimized for mobile devices. Unlike traditional methods that rely on photo-consistency constraints or learned features, or differentiable rendering methods like Neural Radiance Fields (NeRF) that require excessive runtime, EvaSurf employs an efficient surface-based model with a multi-view supervision module for accurate mesh reconstruction. It also learns an implicit texture embedded with Gaussian lobes to capture view-dependent information. The explicit geometry and implicit texture enable the use of a lightweight neural shader, reducing computation expense and supporting real-time rendering on common mobile devices. EvaSurf demonstrates the capability to reconstruct high-quality appearance and accurate mesh on both synthetic and real-world datasets. Remarkably, it can be trained in just 1-2 hours using a single GPU and run on mobile devices at over 40 FPS, making it practical for daily applications that require real-time, high-fidelity 3D reconstruction.
    
    \item \textbf{Adaptive Shells for Efficient Neural Radiance Field Rendering} introduces a method for efficiently rendering neural radiance fields by focusing volumetric rendering on a narrow band around the object. The authors, Zian Wang, Tianchang Shen, Merlin Nimier-David, Nicholas Sharp, Jun Gao, Alexander Keller, Sanja Fidler, Thomas Müller, and Zan Gojcic, propose a neural radiance formulation that transitions between volumetric- and surface-based rendering. This approach constructs an explicit mesh envelope that spatially bounds a neural volumetric representation. In solid regions, the envelope nearly converges to a surface and can often be rendered with a single sample. The method generalizes the NeuS formulation with a learned spatially-varying kernel size, fitting a wide kernel to volume-like regions and a tight kernel to surface-like regions. An explicit mesh of a narrow band around the surface is extracted, with the width determined by the kernel size, and the radiance field within this band is fine-tuned. At inference time, rays are cast against the mesh, evaluating the radiance field only within the enclosed region, significantly reducing the number of samples required. Experiments show that this approach enables efficient rendering at very high fidelity and the extracted envelope facilitates downstream applications such as animation and simulation. This advancement is significant for applications requiring real-time rendering of complex scenes, such as virtual reality, gaming, and simulation, where balancing rendering speed and visual fidelity is crucial.

    \item \textbf{Removing Adverse Volumetric Effects From Trained Neural Radiance Fields} presents a method for synthesizing clear views from Neural Radiance Fields (NeRFs) trained in foggy environments. The authors, Andreas L. Teigen, Mauhing Yip, Victor P. Hamran, Vegard Skui, Annette Stahl, and Rudolf Mester, argue that traditional NeRF models can replicate scenes filled with fog and propose a technique to remove the fog when synthesizing novel views. By calculating the global contrast of a scene, they estimate a density threshold that effectively removes all visible fog, allowing NeRF to render clear views of objects in fog-filled environments. To benchmark performance in such scenarios, they introduce a new dataset expanding original synthetic NeRF scenes with the addition of fog and natural environments. Their approach extends the post-training rendering pipeline and can be used with any generic, pre-trained NeRF model. The method involves simple density thresholding to extract fog-free images from the radiance field, preserving finer details of the model. An algorithm is also created to automatically estimate the threshold after a model has been trained to convergence. This advancement is significant for applications in robotics, such as localization, mapping, and object manipulation, particularly in adverse conditions like fog and haze, where clear vision is crucial.

    \item \textbf{SAX-NeRF (Structure-Aware Sparse-View X-ray 3D Reconstruction)} introduces a novel framework for sparse-view X-ray 3D reconstruction, addressing the limitations of existing NeRF algorithms in capturing structural contents of imaged objects. The authors, Yuanhao Cai, Jiahao Wang, Alan Yuille, Zongwei Zhou, and Angtian Wang, propose Structure-Aware X-ray Neural Radiodensity Fields (SAX-NeRF). SAX-NeRF features a Line Segment-based Transformer (Lineformer) as its backbone, designed to capture internal structures of objects in 3D space by modeling dependencies within each line segment of an X-ray. Additionally, a Masked Local-Global (MLG) ray sampling strategy is presented to extract contextual and geometric information in 2D projection. The authors also collect a larger-scale dataset, X3D, covering a wider range of X-ray applications. Experiments on X3D demonstrate that SAX-NeRF surpasses previous NeRF-based methods in novel view synthesis and CT reconstruction. This advancement is significant for applications in medicine, biology, security, and industry, where X-ray imaging is crucial for revealing internal structures of objects. SAX-NeRF's ability to reconstruct 3D representations from sparse-view X-rays with reduced radiation exposure is particularly beneficial in healthcare and industrial inspection.

    \item \textbf{EN-SLAM (Implicit Event-RGBD Neural SLAM)} introduces a novel framework to address challenges in non-ideal scenarios often encountered in simultaneous localization and mapping (SLAM), such as motion blur or lighting variation. The authors, Delin Qu, Chi Yan, Dong Wang, Jie Yin, Dan Xu, Bin Zhao, and Xuelong Li, propose the first event-RGBD implicit neural SLAM framework, EN-SLAM, which leverages the high rate and high dynamic range advantages of event data for tracking and mapping. EN-SLAM introduces a differentiable Camera Response Function (CRF) rendering technique to generate distinct RGB and event camera data via a shared radiance field. This field is optimized by learning a unified implicit representation with captured event and RGBD supervision. Additionally, EN-SLAM employs a temporal aggregating optimization strategy for event joint tracking and global bundle adjustment, enhancing tracking accuracy and robustness. The authors construct simulated and real captured datasets, DEV-Indoors and DEV-Reals, for evaluation. Experimental results demonstrate that EN-SLAM outperforms state-of-the-art methods in tracking and mapping accuracy, achieving real-time performance in challenging environments. This advancement is significant for applications in virtual and augmented reality, robot navigation, and autonomous driving, particularly in extreme environments where traditional visual SLAM systems struggle.
    
    \item \textbf{NePF: Neural Photon Field for Single-Stage Inverse Rendering} introduces a novel single-stage framework to address the ill-posed problem of inverse rendering from multi-view images. The authors, Tuen-Yue Tsui and Qin Zou, present Neural Photon Field (NePF), which uniquely recovers geometry, material, and illumination properties uniformly, unlike previous methods that operate in multiple stages and extract properties from various multi-layer perceptrons across different neural fields. NePF utilizes the physical implication behind the weight function of neural implicit surfaces and view-dependent radiance, introducing a coordinate-based illumination model for rapid volume physically-based rendering and implementing a subsurface scattering model for diffuse estimation. Evaluated on both real and synthetic datasets, NePF demonstrates superiority in recovering high-fidelity geometry and visually plausible material attributes. This approach challenges the necessity of multi-stage solutions in inverse rendering, offering a unified, efficient alternative that synchronizes the recovery of desired properties with the outgoing radiance on rays. This advancement is significant for applications in computer vision and graphics, particularly in scenarios where recovering detailed properties of an object from multi-view images is crucial.
    
    \item \textbf{EVE-NeRF: Entangled View-Epipolar Information Aggregation for Generalizable Neural Radiance Fields} introduces a novel approach to enhance the quality of generalizable 3D representation in Neural Radiance Fields (NeRF). The authors, Zhiyuan Min, Yawei Luo, Wei Yang, Yuesong Wang, and Yi Yang, propose EVE-NeRF, which conducts view-epipolar feature aggregation in an entangled manner by injecting scene-invariant appearance continuity and geometry consistency priors into the aggregation process. This method effectively addresses the lack of inherent geometric and appearance constraints resulting from one-dimensional interactions in existing methods. EVE-NeRF includes two key components: the View-Epipolar Interaction Module (VEI) and the Epipolar-View Interaction Module (EVI), both adopting a dual-branch structure to integrate view and epipolar information concurrently. VEI engages with the features of sampling points re-projected on all source views, while EVI aggregates features along the epipolar line in each reference view. Extensive experiments demonstrate that EVE-NeRF attains state-of-the-art performance across various evaluation scenarios, excelling in the accuracy of 3D scene geometry and appearance reconstruction. This advancement is significant for applications in novel view synthesis, particularly in scenarios where synthesizing new scenes without scene-specific re-training is essential.
    
    \item \textbf{Towards Function Space Mesh Watermarking: Protecting the Copyright of Signed Distance Fields} proposes FuncMark, a robust and invisible watermarking method designed to protect the copyright of signed distance fields (SDFs). The authors, Xingyu Zhu, Guanhui Ye, Chengdong Dong, Xiapu Luo, and Xuetao Wei, address the urgent need to identify the intellectual property of SDFs, which represent 3D geometries in continuous function space and can be used to extract explicit 3D models at arbitrary resolution. FuncMark leverages analytic on-surface deformations to embed binary watermark messages in SDFs. These deformations can survive isosurfacing, thus being inherited by the extracted meshes for further watermark message decoding. The method is capable of recovering the message with high-resolution meshes extracted from SDFs and detecting the watermark even with extremely sparse mesh vertices. Moreover, FuncMark remains robust against various distortions, including remeshing. Extensive experiments demonstrate that FuncMark significantly outperforms state-of-the-art approaches, with the message still detectable even when only 50 vertex samples are given. This advancement is crucial for protecting the intellectual property rights of SDFs, especially given their increasing use in enhancing neural radiance fields (NeRF) for geometry representation and the potential to extract original meshes from any SDF-enhanced implicit neural field.
    
    \item \textbf{Hyb-NeRF (A Multiresolution Hybrid Encoding for Neural Radiance Fields)} is a novel neural radiance field with a multi-resolution hybrid encoding that achieves efficient neural modeling and fast rendering while allowing for high-quality novel view synthesis. The authors, Yifan Wang, Yi Gong, and Yuan Zeng, introduce Hyb-NeRF, which represents the scene using different encoding strategies from coarse-to-fine resolution levels. This method combines memory-efficient learnable positional features at coarse resolutions with the fast optimization speed and local details of hash-based feature grids at fine resolutions. Additionally, Hyb-NeRF incorporates cone tracing-based features in its learnable positional encoding to eliminate encoding ambiguity and reduce aliasing artifacts. Extensive experiments on both synthetic and real-world datasets show that Hyb-NeRF achieves faster rendering speed, better rendering quality, and even a lower memory footprint compared to previous state-of-the-art methods. This advancement is significant for applications in computer vision and computer graphics, particularly in synthesizing novel views in real-time at photorealistic quality, a long-standing problem in these fields.

    \item \textbf{3D Face Style Transfer with a Hybrid Solution of NeRF and Mesh Rasterization} addresses the challenge of 3D face style transfer, aiming to generate stylized novel views of a 3D human face with multi-view consistency. The authors, Jianwei Feng and Prateek Singhal, propose a hybrid framework combining Neural Radiance Field (NeRF) and mesh rasterization. This approach overcomes the 3D inconsistency issues and blurriness encountered when applying traditional 2D style transfer methods to different viewpoints of the same face. The framework involves three stages: training a NeRF model on input face images to learn the 3D geometry, extracting a mesh from the trained NeRF, and applying 2D style transfer to the mesh using mesh rasterization for fast rendering. This method effectively combines the high-fidelity geometry reconstruction of NeRF with the rapid rendering speed of mesh rasterization. The proposed solution is significant for applications in digital art, entertainment, and virtual reality, where creating consistent and high-quality stylized 3D representations of human faces from various viewpoints is crucial.
    
    \item \textbf{"Retargeting Visual Data with Deformation Fields} introduces a novel image editing method that extends the concept of seam carving for content-aware resizing, including object removal, to a broader range of visual data formats. The authors, Tim Elsner, Julia Berger, Tong Wu, Victor Czech, Lin Gao, and Leif Kobbelt, propose using a neural network to learn a deformation that maintains output plausibility while deforming areas with low information content. This technique is applicable to various types of visual data, including images, 3D scenes represented as neural radiance fields, and polygon meshes. The method is based on describing the editing and retargeting of images more generally by a displacement field, yielding a generalization of content-aware deformations. Their experiments across different visual data types show that this method achieves better content-aware retargeting compared to previous methods. This advancement is significant for applications in media editing, where retargeting images, videos, 3D objects, or entire 3D scenes is essential for fitting them into allocated spaces or changing aspect ratios while preserving relevant content, details, and features.
    
    \item \textbf{Compact 3D Gaussian Representation for Radiance Field} introduces a novel approach to address the computational bottleneck in Neural Radiance Fields (NeRFs) due to volumetric rendering. The authors, Joo Chan Lee, Daniel Rho, Xiangyu Sun, Jong Hwan Ko, and Eunbyung Park, focus on reducing the number of 3D Gaussians in 3D Gaussian splatting (3DGS) without sacrificing performance and compressing Gaussian attributes like view-dependent color and covariance. They propose a learnable mask strategy that significantly reduces the number of Gaussians while preserving high performance. Additionally, they introduce a compact representation of view-dependent color using a grid-based neural field instead of relying on spherical harmonics and employ codebooks for vector quantization to compactly represent the geometric attributes of Gaussian. Their extensive experiments show over 10× reduced storage and enhanced rendering speed while maintaining the quality of the scene representation compared to 3DGS. This work provides a comprehensive framework for 3D scene representation, achieving high performance, fast training, compactness, and real-time rendering. This advancement is significant for applications in neural rendering, where rendering photorealistic 3D scenes from limited input data is crucial for diverse applications, including virtual reality and digital content creation.
    
    \item \textbf{Towards Transferable Multi-modal Perception Representation Learning for Autonomy: NeRF-Supervised Masked AutoEncoder} proposes a unified self-supervised pre-training framework for transferable multi-modal perception representation learning. The author, Xiaohao Xu, introduces NeRF-Supervised Masked AutoEncoder (NS-MAE), which leverages Neural Radiance Fields (NeRF) for masked multi-modal reconstruction. This method uses multi-modal embeddings extracted from corrupted multi-modal input signals, such as Lidar point clouds and images, and renders them into projected multi-modal feature maps via neural rendering. The original multi-modal signals then serve as reconstruction targets for the rendered feature maps, enabling self-supervised representation learning. Extensive experiments show that the representation learned via NS-MAE demonstrates promising transferability for diverse multi-modal and single-modal perception models on various 3D perception downstream tasks. NS-MAE's approach is particularly effective in balancing the need for unified optimization formulation and learning transferable multi-modal representation. This advancement is significant for autonomous driving, where multi-modal perception is crucial for sensing the surrounding scene and extracting and fusing representations from diverse modalities.
    
    \item \textbf{Posterior Distillation Sampling} is a novel optimization method for parametric image editing based on diffusion models. The authors, Juil Koo, Chanho Park, and Minhyuk Sung, have developed Posterior Distillation Sampling (PDS), which reformulates the 2D image editing method into an optimization form. Unlike generation, editing requires a balance between conforming to the target attribute and preserving the identity of the source content. PDS achieves this by matching the stochastic latents of the source and the target, enabling the sampling of targets in diverse parameter spaces that align with a desired attribute while maintaining the source’s identity. This optimization resembles running a generative process with the target attribute but aligning this process with the trajectory of the source’s generative process. Extensive editing results in Neural Radiance Fields and Scalable Vector Graphics representations demonstrate PDS's capability to sample targets across various parameter spaces, fulfilling the balance between target conformity and source identity preservation. This advancement is significant for applications in digital art, advertising, and entertainment, where parametric image editing is crucial for creating visually appealing and contextually aligned content.

    \item \textbf{Tube-NeRF (Efficient Imitation Learning of Visuomotor Policies from MPC using Tube-Guided Data Augmentation and NeRFs)} is a novel data augmentation framework that efficiently learns robust visuomotor policies from Model Predictive Control (MPC). The authors, Andrea Tagliabue and Jonathan P. How, combine imitation learning (IL) with a variant of robust MPC that accounts for process and sensing uncertainties. They design a data augmentation strategy using Neural Radiance Fields (NeRFs) to generate novel synthetic images and employ the properties of robust MPC (the tube) to select relevant views and efficiently compute corresponding actions. Tailored to the task of localization and trajectory tracking on a multirotor, Tube-NeRF learns a visuomotor policy that generates control actions using images from the onboard camera as the only source of horizontal position. The evaluations numerically demonstrate learning of a robust visuomotor policy with significant improvements in demonstration efficiency and training time over current IL methods. Additionally, the policies successfully transfer to a real multirotor, achieving accurate localization and low tracking errors despite large disturbances, with an onboard inference time of only 1.5ms. This advancement is significant for autonomous systems, particularly in mobile robotics, where efficient and robust sensorimotor policy training is crucial for onboard sensing, planning, and control.

    \item \textbf{ECRF: Entropy-Constrained Neural Radiance Fields Compression with Frequency Domain Optimization} introduces a novel compression model for Neural Radiance Fields (NeRF) that minimizes entropy in the frequency domain to effectively reduce data size. The authors, Soonbin Lee, Fangwen Shu, Yago Sánchez, Thomas Schierl, and Cornelius Hellge, propose using the discrete cosine transform (DCT) on tensorial radiance fields to compress the feature-grid. This grid is transformed into coefficients, which are then quantized and entropy encoded, following a process similar to traditional video coding pipelines. To achieve higher sparsity, they introduce an entropy parameterization technique specifically for DCT coefficients of the feature-grid. Optimized during the training phase, this model does not require fine-tuning or additional information. The lightweight compression pipeline for encoding and decoding makes it easier to apply volumetric radiance field methods in real-world applications. Experimental results demonstrate superior compression performance across various datasets. This advancement is significant for 3D scene modeling and novel-view synthesis, where reducing computational overhead and data size is crucial for efficient training, inference, and real-world applicability.
    
    \item \textbf{GaussianEditor: Swift and Controllable 3D Editing with Gaussian Splatting} presents an innovative and efficient 3D editing algorithm that addresses the limitations of traditional 3D editing methods and implicit 3D representations like Neural Radiance Field (NeRF). The authors, Yiwen Chen, Zilong Chen, Chi Zhang, Feng Wang, Xiaofeng Yang, Yikai Wang, Zhongang Cai, Lei Yang, Huaping Liu, and Guosheng Lin, have developed GaussianEditor, which is based on Gaussian Splatting (GS), a novel 3D representation. This method enhances precision and control in editing through Gaussian semantic tracing, which traces the editing target throughout the training process. Additionally, Hierarchical Gaussian splatting (HGS) is proposed to achieve stabilized and fine results under stochastic generative guidance from 2D diffusion models. The authors also develop editing strategies for efficient object removal and integration, a challenging task for existing methods. Comprehensive experiments demonstrate GaussianEditor's superior control, efficacy, and rapid performance, marking a significant advancement in 3D editing. This development is particularly significant for applications in gaming and virtual reality, where swift, controllable, and versatile 3D editing is crucial for creating realistic and complex scenes.
    
    \item \textbf{Animate124 (Animating One Image to 4D Dynamic Scene)} is the first framework to animate a single in-the-wild image into 3D video through textual motion descriptions. The authors, Yuyang Zhao, Zhiwen Yan, Enze Xie, Lanqing Hong, Zhenguo Li, and Gim Hee Lee, have developed a method that leverages an advanced 4D grid dynamic Neural Radiance Field (NeRF) model, optimized in three distinct stages using multiple diffusion priors. Initially, a static model is optimized using the reference image, guided by 2D and 3D diffusion priors. Subsequently, a video diffusion model is employed to learn the motion specific to the subject. However, the object in the 3D videos tends to drift away from the reference image over time, mainly due to misalignment between the text prompt and the reference image. To address this, a personalized diffusion prior is utilized in the final stage to correct the semantic drift. As a pioneering image-text-to-4D generation framework, Animate124 demonstrates significant advancements over existing baselines, as evidenced by comprehensive quantitative and qualitative assessments. This innovation holds immense potential for applications in video games, augmented reality, and virtual reality, where dynamic 3D scene generation from static images can enhance user experience and creative possibilities.
    
    \item \textbf{Neural Style Transfer for Computer Games} explores the relatively uncharted application of Neural Style Transfer (NST) in the realm of 3D computer games. The authors, Eleftherios Ioannou and Steve Maddock, present a novel approach for integrating depth-aware NST into the 3D rendering pipeline of computer games. While NST has been applied to images, videos, 3D meshes, and radiance fields, its use in 3D computer games has been limited, often resulting in undesired artifacts and diminished post-processing effects when used as a post-processing effect. The authors' framework demonstrates temporally consistent results of artistically stylized game scenes, outperforming state-of-the-art image and video NST methods. This advancement is validated through qualitative and quantitative experiments, showcasing the potential of NST to dynamically alter the visual aesthetics of a game in real-time. This innovation holds immense potential for game developers and players, offering the ability to dynamically change the visual style of a game's world and characters according to user preferences, thereby enhancing the creative and artistic aspects of gaming experiences.
    
    \item \textbf{NeuRAD: Neural Rendering for Autonomous Driving} is a robust novel view synthesis method tailored to dynamic autonomous driving (AD) data. The authors, Adam Tonderski, Carl Lindström, Georg Hess, William Ljungbergh, Lennart Svensson, and Christoffer Petersson, have developed this method to address the limitations of existing Neural Radiance Fields (NeRFs) in the AD community, such as long training times, dense semantic supervision, and lack of generalizability. NeuRAD features a simple network design and extensive sensor modeling for both camera and lidar, including rolling shutter, beam divergence, and ray dropping. It is applicable to multiple datasets out of the box and has been verified on five popular AD datasets, achieving state-of-the-art performance across the board. NeuRAD's capabilities include altering the pose of the ego vehicle and other road users, as well as adding or removing actors, making it suitable for use in sensor-realistic closed-loop simulators or as a powerful data augmentation engine. This advancement is significant for the field of autonomous driving, where creating editable digital clones of traffic scenes can explore safety-critical scenarios in a scalable manner without risking physical damage, and for generating corner-case training data.

    \item \textbf{Obj-NeRF: Extract Object NeRFs from Multi-view Images} presents a comprehensive pipeline for recovering the 3D geometry of a specific object from multi-view images using a single prompt. The authors of this paper have developed Obj-NeRF, which combines the 2D segmentation capabilities of the Segment Anything Model (SAM) with the 3D reconstruction ability of Neural Radiance Fields (NeRF). The process involves obtaining multi-view segmentation for the indicated object using SAM with a single prompt, followed by using the segmentation images to supervise NeRF construction. This method integrates several effective techniques and also includes the construction of a large object-level NeRF dataset containing diverse objects. Obj-NeRF is applied to various applications, including object removal, rotation, replacement, and recoloring, demonstrating its practicality. This advancement is significant for applications in 3D editing, novel view synthesis, and other downstream tasks where extracting specific objects from multi-view images is crucial for creating detailed and accurate 3D representations.
    
    \item \textbf{"Efficient Encoding of Graphics Primitives with Simplex-Based Structures} presents a novel approach to encoding graphics primitives, addressing the computational inefficiencies of traditional grid-based structures. The authors, Yibo Wen and Yunfan Yang, propose a simplex-based approach for encoding, which is more efficient and generalizable compared to grid-based representations. In n-dimensional space, their method requires interpolating the values of linearly increasing vertices in a simplex-based structure, as opposed to the exponential scaling with dimension in grid-based structures. They derive and prove a coordinate transformation, simplicial subdivision, and barycentric interpolation scheme for efficient sampling, similar to the simplex noise algorithm. The approach utilizes hash tables to store multiresolution features of interest points in the simplicial grid, which are then passed into a tiny fully connected neural network to parameterize graphics primitives. In their implementation, the proposed method showed a 9.4\% reduction in time compared to the baseline method in a 2D image fitting task and up to a 41.2\% speedup in volumetric rendering setups. This advancement is significant for applications in computer graphics and visualization, where efficient and high-quality representation of graphics primitives is crucial for creating complex visual scenes.

    \item \textbf{CaesarNeRF: Calibrated Semantic Representation for Few-Shot Generalizable Neural Rendering} introduces an end-to-end approach to address the challenges of generalizability and few-shot learning in Neural Radiance Fields (NeRF). The authors, Haidong Zhu, Tianyu Ding, Tianyi Chen, Ilya Zharkov, Ram Nevatia, and Luming Liang, have developed CaesarNeRF, which leverages scene-level CAlibratEdSemAntic Representation along with pixel-level representations. This method explicitly models pose differences of reference views to combine scene-level semantic representations, providing a calibrated holistic understanding. The calibration process aligns various viewpoints with precise location and is further enhanced by sequential refinement to capture varying details. Extensive experiments on public datasets, including LLFF, Shiny, mip-NeRF 360, and MVImgNet, demonstrate that CaesarNeRF delivers state-of-the-art performance across varying numbers of reference views, proving effective even with a single reference image. This advancement is significant for applications in view synthesis, virtual reality, and augmented reality, where rendering a scene from novel camera positions with limited reference views is crucial for creating immersive and realistic experiences.
    
    \item \textbf{PaintNeSF: Artistic Creation of Stylized Scenes with Vectorized 3D Strokes} innovates in the field of digital art creation by introducing a novel technique to generate stylized images of a 3D scene from multi-view 2D images. The authors, Hao-Bin Duan, Miao Wang, Yan-Xun Li, and Yong-Liang Yang, have developed Paint Neural Stroke Field (PaintNeSF), drawing inspiration from image-to-painting methods. Unlike existing methods that apply stylization at the voxel level, PaintNeSF simulates the progressive painting process of human artwork using vector strokes. The authors develop a palette of stylized 3D strokes from basic primitives and splines, treating the 3D scene stylization task as a multi-view reconstruction process based on these 3D stroke primitives. They introduce a differentiable renderer that allows optimizing stroke parameters using gradient descent and propose a training scheme to alleviate the vanishing gradient issue. Extensive evaluations show that PaintNeSF effectively synthesizes 3D scenes with significant geometric and aesthetic stylization while maintaining consistent appearance across different views. The method can be integrated with style loss and image-text contrastive models, extending its applications to color transfer and text-driven 3D scene drawing. This advancement is particularly significant for digital art creation, where it offers a new avenue for artists and designers to create stylized 3D scenes with a high degree of artistic control and creativity.
    
    \item \textbf{SOAC: Spatio-Temporal Overlap-Aware Multi-Sensor Calibration using Neural Radiance Fields} underscores a novel approach to achieve robust and accurate spatio-temporal sensor calibration, particularly crucial in autonomous driving. The authors, Quentin Herau, Nathan Piasco, Moussab Bennehar, Luis Roldão, Dzmitry Tsishkou, Cyrille Migniot, and Pascal Vasseur, leverage the ability of Neural Radiance Fields (NeRF) to represent different sensors modalities in a common volumetric representation. They design a partitioning approach based on the visible part of the scene for each sensor, formulating the calibration problem using only the overlapping areas. This strategy results in a more robust and accurate calibration that is less prone to failure. Validated on multiple established driving datasets, the method demonstrates better accuracy and robustness compared to existing methods, particularly in outdoor urban scenes. This advancement is significant for applications in autonomous systems, where precise multi-sensor calibration is essential for tasks like localization and perception, ensuring accuracy, reliability, and robustness in safety-critical operations.
    
    \item \textbf{Animatable Gaussians: Learning Pose-dependent Gaussian Maps for High-fidelity Human Avatar Modeling} introduces a novel approach for modeling animatable human avatars from RGB videos. The authors, Zhe Li, Zerong Zheng, Lizhen Wang, and Yebin Liu, have developed a new avatar representation that leverages 2D CNNs and 3D Gaussian splatting to create high-fidelity avatars. This method involves learning a parametric template from input videos and parameterizing it on two front and back canonical Gaussian maps, where each pixel represents a 3D Gaussian. The learned template adapts to wearing garments for modeling looser clothes like dresses. This template-guided 2D parameterization enables the use of a powerful StyleGAN-based CNN to learn pose-dependent Gaussian maps for modeling detailed dynamic appearances. Additionally, they introduce a pose projection strategy for better generalization with novel poses. The method creates lifelike avatars with dynamic, realistic, and generalized appearances, outperforming other state-of-the-art approaches. This advancement is particularly significant for applications in holoportation, the Metaverse, and the gaming and movie industries, where high-quality and animatable human avatar modeling is crucial.
    
    \item \textbf{SeamlessNeRF: Stitching Part NeRFs with Gradient Propagation} introduces a novel approach for seamless appearance blending of multiple Neural Radiance Fields (NeRFs), akin to the concept of "Poisson blending" in 2D image editing. The authors, Bingchen Gong, Yuehao Wang, Xiaoguang Han, and Qi Dou, propose SeamlessNeRF, aiming to optimize the appearance of a target radiance field for harmonious merging with a source field. This method involves a well-tailored optimization procedure constrained by pinning the radiance color at the intersecting boundary between the source and target fields and maintaining the original gradient of the target. Extensive experiments demonstrate the effectiveness of this approach in propagating the source appearance from the boundary area to the entire target field through gradients. SeamlessNeRF is the first work to introduce gradient-guided appearance editing to radiance fields, offering solutions for seamless stitching of 3D objects represented in NeRFs. This advancement is particularly significant for applications in digital art creation, virtual reality, and augmented reality, where seamless editing and merging of 3D objects and scenes are crucial for creating immersive and realistic experiences.
    
    \item \textbf{Animatable 3D Gaussian: Fast and High-Quality Reconstruction of Multiple Human Avatars} presents a novel approach for reconstructing high-quality drivable human avatars efficiently. The authors, Yang Liu, Xiang Huang, Minghan Qin, Qinwei Lin, and Haoqian Wang, propose the Animatable 3D Gaussian method, which learns human avatars from input images and poses. This method extends 3D Gaussians to dynamic human scenes by modeling a set of skinned 3D Gaussians and a corresponding skeleton in canonical space, deforming 3D Gaussians to posed space according to input poses. Key innovations include the introduction of hash-encoded shape and appearance to speed up training and the proposal of time-dependent ambient occlusion for high-quality reconstructions in scenes with complex motions and dynamic shadows. The method outperforms existing methods in training time, rendering speed, and reconstruction quality for both novel view synthesis and novel pose synthesis tasks. Notably, it can be easily extended to multi-human scenes, achieving comparable novel view synthesis results in a scene with ten people in only 25 seconds of training. This advancement is particularly significant for applications in virtual reality, gaming, sports broadcasting, and telepresence, where real-time rendering and fast reconstruction of high-quality digital humans are crucial.
    
    \item \textbf{Rethinking Directional Integration in Neural Radiance Fields} introduces a significant modification to the Neural Radiance Field (NeRF) rendering equation, aimed at enhancing the rendering quality of view-dependent effects. The authors, Congyue Deng, Jiawei Yang, Leonidas Guibas, and Yue Wang, have proposed a simple yet effective change applicable to any NeRF variation. By swapping the integration operator and the direction decoder network, their method integrates only the positional features along the ray, moving the directional terms out of the integration. This results in a disentanglement of view-dependent and independent components. The authors demonstrate that their modified equation is equivalent to classical volumetric rendering on object surfaces with Dirac densities in ideal cases. They also prove that, considering errors caused by network approximation and numerical integration, their rendering equation exhibits better convergence properties with lower error accumulations compared to classical NeRF. Furthermore, they interpret the modified equation as light field rendering with learned ray embeddings. Experiments across different NeRF variations show consistent improvements in the quality of view-dependent effects with this modification. This advancement is particularly significant for applications in virtual reality, augmented reality, and photorealistic rendering, where accurate and enhanced view-dependent rendering is crucial for creating immersive and realistic experiences.
    
    \item \textbf{RGBGrasp: Image-based Object Grasping by Capturing Multiple Views during Robot Arm Movement with Neural Radiance Fields} underscores a pioneering approach for robotic object grasping in diverse scenarios, particularly effective in handling objects with varying shapes, materials, and textures. The authors, Chang Liu, Kejian Shi, Kaichen Zhou, Haoxiao Wang, Jiyao Zhang, and Hao Dong, have developed a method that relies on a limited set of RGB views to perceive 3D surroundings containing transparent and specular objects for accurate grasping. This method utilizes pre-trained depth prediction models to establish geometry constraints, enabling precise 3D structure estimation under limited view conditions. Furthermore, the integration of hash encoding and a proposal sampler strategy significantly accelerates the 3D reconstruction process. These innovations enhance the adaptability and effectiveness of the algorithm in real-world scenarios. Comprehensive experimental validation demonstrates that RGBGrasp achieves remarkable success across a wide spectrum of object-grasping scenarios, establishing it as a promising solution for real-world robotic manipulation tasks. This advancement is particularly significant for applications in automated warehousing, manufacturing, and service robotics, where efficient and accurate object grasping is crucial.

    \item \textbf{SCALAR-NeRF: SCAlable LARge-scale Neural Radiance Fields for Scene Reconstruction} introduces a novel framework designed for scalable large-scale neural scene reconstruction. The authors, Yu Chen and Gim Hee Lee, have developed an encoder-decoder architecture where the encoder processes 3D point coordinates to produce encoded features, and the decoder generates geometric values including volume densities of signed distances and colors. Their approach begins with training a coarse global model on the entire image dataset, followed by partitioning the images into smaller blocks using KMeans, with each block modeled by a dedicated local model. They enhance the overlapping regions across different blocks by scaling up the bounding boxes of each local block. Notably, the decoder from the global model is shared across distinct blocks, promoting alignment in the feature space of local encoders. The authors propose an efficient methodology to fuse the outputs from these local models to attain the final reconstruction. This refined coarse-to-fine strategy enables SCALAR-NeRF to outperform state-of-the-art NeRF methods and demonstrates scalability for large-scale scene reconstruction. This advancement is particularly significant for applications in augmented reality, real-world simulation, and localization, where large-scale and detailed scene reconstruction is crucial.

    \item \textbf{DGNR: Density-Guided Neural Point Rendering of Large Driving Scenes} presents a novel framework for rendering large-scale driving scenes, addressing the challenges faced by Neural Radiance Field (NeRF) in such contexts. The authors of this paper have developed DGNR (Density-Guided Neural Rendering), which eliminates the need for geometric priors and learns a density space from scenes to guide the construction of a point-based renderer. This method uses a differentiable renderer to synthesize images from neural density features obtained from the learned density space. Key innovations include a density-based fusion module and geometric regularization to optimize the density space. DGNR has been validated on an autonomous driving dataset, demonstrating its effectiveness in synthesizing photorealistic driving scenes and achieving real-time capable rendering. This approach is particularly significant for autonomous driving simulation, sensor simulation, and traffic simulation, where it offers a robust platform for comprehensive testing and assessment of autonomous driving systems in a controlled virtual environment.
    
    \item \textbf{SplitNeRF: Split Sum Approximation Neural Field for Joint Geometry, Illumination, and Material Estimation} introduces a novel approach for digitizing real-world objects by estimating their geometry, material properties, and environmental lighting from a set of posed images with fixed lighting. The authors, Jesus Zarzar and Bernard Ghanem, have innovatively incorporated the split sum approximation used in image-based lighting for real-time physical-based rendering into Neural Radiance Field (NeRF) pipelines. Their method models the scene’s lighting with a single scene-specific MLP representing pre-integrated image-based lighting at arbitrary resolutions. A key aspect of their approach is the accurate modeling of pre-integrated lighting, achieved by exploiting a novel regularizer based on efficient Monte Carlo sampling. Additionally, they propose a new method for supervising self-occlusion predictions using a similar regularizer. The experimental results demonstrate the efficiency and effectiveness of their approach in estimating scene geometry, material properties, and lighting. Remarkably, their method attains state-of-the-art relighting quality after only about one hour of training on a single NVIDIA A100 GPU. This advancement has significant implications for the fields of computer graphics and computer vision, particularly in creating digital twins of objects that can be seamlessly integrated within photorealistic environments, a crucial aspect for applications in virtual reality, augmented reality, and digital content creation.
    
    \item \textbf{A Unified Approach for Text- and Image-guided 4D Scene Generation} introduces Dream-in-4D, a novel method for text-to-4D synthesis, addressing the complex challenge of generating dynamic 3D scenes guided by text and images. The authors, Yufeng Zheng, Xueting Li, Koki Nagano, Sifei Liu, Karsten Kreis, Otmar Hilliges, and Shalini De Mello, have developed a two-stage approach leveraging 3D and 2D diffusion guidance. The first stage involves learning a high-quality static 3D asset, while the second stage focuses on learning motion with video diffusion guidance using a deformable neural radiance field that disentangles the static asset from its deformation. This method also incorporates a multi-resolution feature grid for the deformation field with a displacement total variation loss. Dream-in-4D significantly advances image and motion quality, 3D consistency, and text fidelity for text-to-4D generation, as demonstrated in a user preference study. Additionally, its motion-disentangled representation allows for easy adaptation to controllable generation where appearance is defined by one or multiple images. This unified approach offers groundbreaking possibilities in gaming, AR/VR, and advertising, where synthesizing and animating 3D assets using intuitive text prompts can save significant time and expertise for animators.

    \item \textbf{UC-NERF: Neural Radiance Fields for Under-calibrated Multi-View Cameras in Autonomous Driving} introduces a novel method designed to address the challenges of applying Neural Radiance Field (NeRF) techniques to under-calibrated multi-camera systems, particularly in autonomous driving. The authors, Kai Cheng, Xiaoxiao Long, Wei Yin, Jin Wang, Zhiqiang Wu, Yuexin Ma, Kaixuan Wang, Xiaozhi Chen, and Xuejin Chen, tackled the inherent issues of under-calibration in multi-camera setups, such as inconsistent imaging effects and system errors from mechanical vibrations. Their solution, UC-NeRF, introduces a layer-based color correction to rectify color inconsistencies, virtual warping to generate diverse but color-consistent virtual views for color correction and 3D recovery, and a spatiotemporally constrained pose refinement for robust and accurate pose calibration. This method not only achieves state-of-the-art performance in novel view synthesis for multi-camera setups but also significantly aids in depth estimation in large-scale outdoor scenes. The applications of UC-NeRF in autonomous driving are profound, offering enhanced capabilities for generating synthetic data from diverse viewpoints for robust training of perception models and providing effective 3D scene representations for comprehensive environmental understanding.
    
    \item \textbf{Human Gaussian Splatting: Real-time Rendering of Animatable Avatars} presents a groundbreaking approach for real-time rendering of photorealistic human body avatars from multi-view videos. The authors, Arthur Moreau, Jifei Song, Helisa Dhamo, Richard Shaw, Yiren Zhou, and Eduardo Pérez-Pellitero, have developed a novel animatable human model based on 3D Gaussian Splatting, a recent and efficient alternative to neural radiance fields. This method represents the human body with a set of Gaussian primitives in a canonical space, which are deformed using a combination of forward skinning and local non-rigid refinement. The Human Gaussian Splatting (HuGS) model is learned end-to-end from multi-view observations and evaluated against state-of-the-art approaches for novel pose synthesis of clothed bodies. The authors' method outperforms existing techniques, achieving a PSNR 1.5dB better than the state-of-the-art on the THuman4 dataset while rendering at 20fps or more. This advancement is particularly significant for applications in virtual reality and video games, where the creation of realistic and animatable virtual human avatars is essential for immersive user experiences and content creation.
    
    \item \textbf{REF$^2$-NeRF: Reflection and Refraction aware Neural Radiance Field} discusses a novel approach to 3D reconstruction from multiple images, particularly in scenes with multiple glass objects, a challenge due to the presence of multiple reflection and refraction effects. The authors, Wooseok Kim, Taiki Fukiage, and Takeshi Oishi, proposed a NeRF-based modeling method that uniquely handles scenes containing glass cases. Their method models refraction and reflection by incorporating elements that are both dependent and independent of the viewer’s perspective. This innovative approach allows for the estimation of surfaces where refraction occurs, such as glass surfaces, and enables the separation and modeling of both direct and reflected light components. The REF2-NeRF method represents a significant advancement over existing methods, offering more accurate modeling of glass refraction and the overall scene. This development is particularly impactful in fields like virtual reality, augmented reality, and photorealistic rendering, where accurate representation of complex scenes with transparent objects is crucial for enhancing realism and user experience.
    
    \item \textbf{NeRFTAP: Enhancing Transferability of Adversarial Patches on Face Recognition using Neural Radiance Fields} has delved into the critical issue of adversarial attacks on face recognition (FR) technology, a domain with significant security implications. The authors, Xiaoliang Liu, Furao Shen, Feng Han, Jian Zhao, and Changhai Nie, focused on the challenge of enhancing the transferability of adversarial patches not only to different FR models but also directly to the victim's face images, a more practical threat in real-world scenarios. They proposed NeRFTAP, an innovative adversarial attack method that leverages NeRF-based 3D-GAN to generate new view face images for both source and target subjects, thereby enhancing the transferability of adversarial patches. A key feature of their approach is the introduction of a style consistency loss, which ensures the visual similarity between the adversarial UV map and the target UV map under a 0-1 mask. This addition significantly enhances the effectiveness and naturalness of the generated adversarial face images. Through extensive experiments and evaluations on various FR models, the authors demonstrated the superiority of NeRFTAP over existing attack techniques. Their work offers valuable insights into enhancing the robustness of FR systems against practical adversarial threats, a crucial consideration for applications involving phone unlocking, payment verification, and individual identification.
    
    \item \textbf{SyncTalk: The Devil is in the Synchronization for Talking Head Synthesis} has tackled the significant challenge of achieving high synchronization in the synthesis of realistic, speech-driven talking head videos. The authors, Ziqiao Peng, Wentao Hu, Yue Shi, Xiangyu Zhu, Xiaomei Zhang, Hao Zhao, Jun He, Hongyan Liu, and Zhaoxin Fan, identified the key issues with traditional Generative Adversarial Networks (GANs) and Neural Radiance Fields (NeRF) methods in maintaining consistent facial identity and synchronizing lip movements, facial expressions, and head poses. To overcome these challenges, they introduced SyncTalk, a NeRF-based method that effectively maintains subject identity and enhances synchronization and realism in talking head synthesis. SyncTalk employs a Face-Sync Controller for aligning lip movements with speech, a 3D facial blendshape model for capturing accurate facial expressions, and a Head-Sync Stabilizer for optimizing head poses. Additionally, the Portrait-Sync Generator restores hair details and blends the generated head with the torso for a seamless visual experience. The authors' extensive experiments and user studies demonstrated that SyncTalk outperforms state-of-the-art methods in synchronization and realism. This advancement is particularly significant in fields like digital assistants, virtual reality, and filmmaking, where the creation of lifelike, synchronized talking heads is essential for enhancing user experience and realism.
    
    \item \textbf{FisherRF: Active View Selection and Uncertainty Quantification for Radiance Fields using Fisher Information} has addressed the complex challenges of active view selection and uncertainty quantification in the realm of Radiance Fields, particularly Neural Radiance Fields (NeRF). The authors, Wen Jiang, Boshu Lei, and Kostas Daniilidis, tackled the inherent uncertainties in NeRF caused by limited 2D image availability, occlusions, depth ambiguities, and imaging errors. Their novel approach leverages Fisher Information to efficiently quantify observed information within Radiance Fields without needing ground truth data. This method significantly enhances the process of selecting the most informative views and quantifying pixel-wise uncertainty in NeRF models. Overcoming the limitations of existing methods that depend heavily on model architecture or specific density distribution assumptions, FisherRF achieves state-of-the-art results in both view selection and uncertainty quantification. This advancement is particularly impactful in applications like 3D reconstruction, virtual reality, and augmented reality, where accurate and efficient view selection and uncertainty estimation are crucial for rendering and reconstructing high-quality images. The authors also demonstrated the practicality of their method with the 3D Gaussian Splatting backend, enabling view selections at a remarkable speed of 70 frames per second.
    
    \item \textbf{Anisotropic Neural Representation Learning for High-Quality Neural Rendering} discusses a novel approach to enhance the quality of neural radiance fields (NeRFs) for view synthesis. The authors, Yifan Wang, Jun Xu, Yuan Zeng, and Yi Gong, identified a key limitation in traditional NeRFs: the neglect of directional information in point intervals during volume rendering, leading to ambiguous features and limited reconstruction quality. To address this, they proposed an anisotropic neural representation learning method, utilizing learnable view-dependent features. This method models the volumetric function with spherical harmonic (SH)-guided anisotropic features, parameterized by multilayer perceptrons (MLPs). This approach not only eliminates ambiguity but also maintains rendering efficiency. To prevent overfitting to anisotropy, the authors introduced a regularization of the energy of anisotropic features during training. Their method, compatible with existing NeRF-based frameworks, showed significant improvements in rendering quality across both synthetic and real-world scenes. This advancement has profound implications for applications in virtual reality, augmented reality, and photorealistic rendering, where high-quality visual representation is crucial. 
    
    \item \textbf{Contrastive Denoising Score for Text-guided Latent Diffusion Image Editing} introduces an innovative method for text-guided latent diffusion image editing, termed Contrastive Denoising Score (CDS). Building upon the Delta Denoising Score (DDS) framework, which itself is an advancement in image editing using text-to-image diffusion models, the authors proposed a significant enhancement. The key innovation of CDS lies in its ability to maintain structural integrity of the original image while allowing for controlled content transformation. This is achieved by incorporating a Contrastive Unpaired Translation (CUT) loss within the DDS framework, utilizing intermediate features from the self-attention layers of Latent Diffusion Models (LDM). These layers are rich in spatial information, enabling the model to enforce structural consistency between input and output images. The practical applications of this technique are vast, ranging from zero-shot image-to-image translation to neural radiance field editing. The method demonstrates a well-balanced interplay between preserving structural details and transforming content, as evidenced by qualitative results and comparisons. This approach has significant implications for fields requiring precise image editing, such as digital art creation, media and entertainment, and even medical imaging, where maintaining structural accuracy is crucial.

     \item \textbf{SparseGS: Real-Time 360° Sparse View Synthesis using Gaussian Splatting} is an innovative technique for few-shot view synthesis that significantly outperforms prior methods in both quality and runtime. The authors, Haolin Xiong, Sairisheek Muttukuru, Rishi Upadhyay, Pradyumna Chari, and Achuta Kadambi from the University of California, Los Angeles, address the challenge of novel view synthesis, which has gained popularity with the introduction of Neural Radiance Fields (NeRFs) and other implicit scene representation methods. Their approach is built on top of 3D Gaussian Splatting (3DGS), an explicit representation method that achieves real-time rendering with high-quality results. However, 3DGS, similar to NeRF, tends to overfit to training views in few-shot settings, leading to issues like background collapse and excessive floaters. The authors propose a method to enable training coherent 3DGS-based radiance fields of 360° scenes from sparse training views, integrating depth priors with generative and explicit constraints. This method significantly reduces background collapse, removes floaters, and enhances consistency from unseen viewpoints. Experiments show that their method outperforms base 3DGS by up to 30.5\% and NeRF-based methods by up to 15.6\% in LPIPS on the MipNeRF-360 dataset with substantially less training and inference cost. This advancement is crucial for applications requiring efficient and high-quality novel view synthesis, such as virtual reality and augmented reality.
     
     \item \textbf{PyNeRF: Pyramidal Neural Radiance Fields} introduces an innovative approach to enhance Neural Radiance Fields (NeRFs) by addressing the issue of aliasing artifacts in scenes captured at different camera distances. The authors, Haithem Turki, Michael Zollhöfer, Christian Richardt, and Deva Ramanan from Carnegie Mellon University and Meta Reality Labs Research, propose a modification to grid-based NeRF models by training model heads at different spatial grid resolutions. This method, inspired by divide-and-conquer NeRF extensions and classical approaches like Gaussian pyramids and mipmaps, allows for the use of coarser grids to render samples covering larger volumes, significantly improving rendering quality while maintaining minimal performance overhead. The approach is easily applicable to existing accelerated NeRF methods and has shown to reduce error rates by 20-90\% across synthetic and unbounded real-world scenes, with minimal performance impact. Compared to Mip-NeRF, their method reduces error rates by 20\% while training over 60 times faster. This advancement is significant for applications in photorealistic novel view synthesis, offering a more efficient and high-quality solution for rendering scenes with varying camera distances. The paper was presented at the 37th Conference on Neural Information Processing Systems (NeurIPS 2023).
     
     \item \textbf{Self-Evolving Neural Radiance Fields} introduces a novel framework to enhance Neural Radiance Fields (NeRF) for novel view synthesis and 3D reconstruction, especially in scenarios with sparse viewpoints. The authors, Jaewoo Jung, Jisang Han, Jiwon Kang, Seongchan Kim, Min-Seop Kwak, and Seungryong Kim from Korea University, Seoul, Korea, address the limitations of few-shot NeRF, which often overfits to sparse viewpoints due to the under-constrained nature of the task. SE-NeRF applies a self-training framework, formulating few-shot NeRF into a teacher-student framework. This approach guides the network to learn a more robust representation of the scene by training the student with additional pseudo labels generated from the teacher. By distilling ray-level pseudo labels using distinct distillation schemes for reliable and unreliable rays, obtained with a novel reliability estimation method, SE-NeRF enables NeRF to learn a more accurate and robust geometry of the 3D scene. This method improves the quality of rendered images and achieves state-of-the-art performance in multiple settings. The project page for SE-NeRF is available at https://ku-cvlab.github.io/SE-NeRF/.
     
     \item \textbf{Neural Parametric Gaussians for Monocular Non-Rigid Object Reconstruction} introduces Neural Parametric Gaussians (NPGs), a novel method for reconstructing dynamic objects from monocular videos. The authors, Devikalyan Das, Christopher Wewer, Raza Yunus, Eddy Ilg, and Jan Eric Lenssen from Saarland University and the Max Planck Institute for Informatics in Germany, tackle the severely underconstrained and challenging problem of monocular non-rigid reconstruction. NPGs employ a two-stage approach: initially fitting a low-rank neural deformation model for consistency in novel views, followed by optimizing 3D Gaussians driven by the coarse model for high reconstruction quality. This method introduces a local 3D Gaussian representation, where temporally shared Gaussians are anchored in and deformed by local oriented volumes. The combined model can be rendered as radiance fields, resulting in high-quality photorealistic reconstructions of non-rigidly deforming objects, maintaining 3D consistency across novel views. NPGs demonstrate superior results, especially in challenging scenarios with few multi-view cues, and have applications in various industries like movies, games, AR/VR, and robotics.
     
     \item \textbf{VideoRF: Rendering Dynamic Radiance Fields as 2D Feature Video Streams} introduces a groundbreaking approach for real-time streaming and rendering of dynamic radiance fields on mobile platforms, addressing the challenges of data storage and computational constraints. The authors, Liao Wang, Kaixin Yao, Chengcheng Guo, Zhirui Zhang, Qiang Hu, Jingyi Yu, Lan Xu, and Minye Wu from ShanghaiTech University, KU Leuven, and NeuDim, propose VideoRF, a novel technique that views dynamic radiance fields as serialized 2D feature image streams. This method leverages the temporal and spatial redundancy of the feature image stream, enabling efficient compression by 2D video codecs and real-time decoding using video hardware accelerators. VideoRF's rendering pipeline, paired with a deferred shading model, allows for efficient querying of radiance properties and real-time rendering on mobile devices. The team developed an interactive player for online streaming and rendering of dynamic scenes, offering a seamless free-viewpoint experience across various devices, from desktops to mobile phones. This advancement significantly enhances the accessibility and practicality of photorealistic Free-Viewpoint Video (FVV) in virtual reality and telepresence.

     \item \textbf{SANeRF-HQ (Segment Anything for NeRF in High Quality)} introduces an innovative framework for high-quality 3D segmentation of any object in a given scene using Neural Radiance Fields (NeRF). The authors, Yichen Liu, Benran Hu, Chi-Keung Tang, and Yu-Wing Tai from The Hong Kong University of Science and Technology, Carnegie Mellon University, and Dartmouth College, address the challenge of accurately and consistently segmenting objects in complex scenarios. SANeRF-HQ utilizes the Segment Anything Model (SAM) for open-world object segmentation guided by user-supplied prompts, while leveraging NeRF to aggregate information from different viewpoints. The method employs density field and RGB similarity to enhance the accuracy of segmentation boundaries during aggregation. SANeRF-HQ demonstrates a significant quality improvement over previous state-of-the-art methods in NeRF object segmentation, offering higher flexibility for object localization and more consistent object segmentation across multiple views.
     
     \item \textbf{GaussianHead: Impressive 3D Gaussian-based Head Avatars with Dynamic Hybrid Neural Field} introduces a novel approach for creating high-fidelity 3D head avatars using anisotropic 3D Gaussian primitives. The authors, Jie Wang, Xianyan Li, Jiucheng Xie, Feng Xu, and Hao Gao from Nanjing University of Posts and Telecommunications and Tsinghua University, address the challenge of balancing high fidelity, training speed, and resource consumption in head avatar generation. GaussianHead leverages canonical Gaussians to represent dynamic scenes and uses a 'dynamic' tri-plane as an efficient container for parameterized head geometry. This alignment with the underlying geometry and tri-plane factors results in aligned canonical factors for the canonical Gaussians. A tiny MLP decodes these factors into opacity and spherical harmonic coefficients of 3D Gaussian primitives, which are then rendered using an efficient differentiable Gaussian rasterizer. This method achieves optimal visual results in tasks like self-reconstruction, novel view synthesis, and cross-identity reenactment, while maintaining high rendering efficiency. The approach significantly benefits from the novel representation based on 3D Gaussians and the proper alignment transformation, eliminating biases introduced by fixed mappings.
     
     \item \textbf{Fast and accurate sparse-view CBCT reconstruction using meta-learned neural attenuation field and hash-encoding regularization} presents an innovative method for Cone Beam Computed Tomography (CBCT) reconstruction, addressing the challenge of reducing the number of projections in a CBCT scan while preserving image quality. The authors, Heejun Shin, Taehee Kim, Jongho Lee, Seyoung Chun, Seungryong Cho, and Dongmyung Shin from various institutions in South Korea, propose a fast and accurate sparse-view CBCT reconstruction (FACT) method. This method utilizes a meta-trained neural network and a hash-encoder, trained with a few scans, and a new regularization technique to reconstruct detailed anatomical structures. The FACT method is designed to provide better reconstruction quality and faster optimization speed with fewer view acquisitions (less than 50 views). This approach is particularly beneficial for reducing potential radiation exposure and considering typical scan times in medical imaging. The method demonstrates superior results over conventional algorithms in CBCT scans of different body parts and CT vendors, marking a significant advancement in medical imaging technology.

     \item \textbf{ColonNeRF: Neural Radiance Fields for High-Fidelity Long-Sequence Colonoscopy Reconstruction} presents a novel framework for reconstructing high-fidelity 3D models of the colon from long-sequence colonoscopy videos. The authors, Yufei Shi, Beijia Lu, Jia-Wei Liu, Ming Li, and Mike Zheng Shou from the National University of Singapore, address three major challenges in colonoscopy reconstruction: the dissimilarity among colon segments due to its convoluted shape, the co-existence of simple and intricately folded geometry structures, and sparse viewpoints due to constrained camera trajectories. ColonNeRF leverages neural rendering for novel view synthesis and introduces a region division and integration module to reconstruct the entire colon in a piecewise manner, reducing shape dissimilarity and ensuring geometric consistency. It also incorporates a multi-level fusion module to model both simple and complex geometry and a DensiNet module to densify camera poses under semantic consistency guidance. The framework outperforms existing methods on synthetic and real-world datasets, demonstrating superior performance in reconstructing clearer textures and more accurate geometric details.
     
     \item \textbf{Fast View Synthesis of Casual Videos} introduces an innovative approach for efficient novel view synthesis from in-the-wild videos, addressing challenges like scene dynamics and lack of parallax. The authors, Yao-Chih Lee, Zhoutong Zhang, Kevin Blackburn-Matzen, Simon Niklaus, Jianming Zhang, Jia-Bin Huang, and Feng Liu from the University of Maryland, College Park, and Adobe Research, revisit explicit video representations to synthesize high-quality novel views from a monocular video efficiently. They separate static and dynamic video content, building a global static scene model using an extended plane-based scene representation augmented with spherical harmonics and displacement maps. This model captures view-dependent effects and models non-planar complex surface geometry. For dynamic content, they represent it as per-frame point clouds for efficiency, accepting minor temporal inconsistencies due to motion. Their method quickly estimates this hybrid video representation and renders novel views in real time. The experiments demonstrate that their method can render high-quality novel views with comparable quality to state-of-the-art methods while being 100 times faster in training and enabling real-time rendering.
     
     \item \textbf{Mesh-Guided Neural Implicit Field Editing} introduces a groundbreaking approach for editing both the geometry and color of neural implicit fields using differentiable colored meshes. The authors, Can Wang, Mingming He, Menglei Chai, Dongdong Chen, and Jing Liao from City University of Hong Kong, Netflix Eyeline Studios, Google AR Perception, and Microsoft Cloud + AI, propose a method that combines the strengths of neural implicit fields and explicit 3D representations like polygonal meshes. Their approach involves extracting a polygonal mesh from the neural implicit field using a differentiable method with marching tetrahedra and assigning colors from volume renderings to this mesh. This differentiable colored mesh allows for gradient back-propagation from the explicit mesh to the implicit fields, enabling easy manipulation of the geometry and color of neural implicit fields. The authors introduce an octree-based structure in the optimization process to enhance user control from coarse to fine levels, focusing on edited regions and the surface part for fine-grained edits. This method accommodates various modifications, including object additions, component removals, area deformations, and color adjustments. Extensive experiments demonstrate the capabilities and effectiveness of this method in diverse scenes and editing operations. The project page for this research is available at https://cassiepython.github.io/MNeuEdit/.

     \item \textbf{StableDreamer: Taming Noisy Score Distillation Sampling for Text-to-3D} presents a significant advancement in text-to-3D generation, addressing the challenges of blurred appearances and multi-faced geometry commonly encountered in 2D diffusion models through score distillation sampling (SDS). The authors, Pengsheng Guo, Hans Hao, Adam Caccavale, Zhongzheng Ren, Edward Zhang, Qi Shan, Aditya Sankar, Alexander G. Schwing, Alex Colburn, and Fangchang Ma from Apple, identify the core challenges as the interaction among noise levels in the 2D diffusion process, the architecture of the diffusion network, and the 3D model representation. Their solution, StableDreamer, incorporates three key advances: formalizing the equivalence of the SDS generative prior and a supervised L2 reconstruction loss, a two-stage training strategy that combines image-space and latent-space diffusion for geometric precision and vivid color rendition, and adopting an anisotropic 3D Gaussians representation to enhance overall quality, reduce memory usage, and accelerate rendering speeds. StableDreamer effectively reduces multi-face geometries, generates fine details, and converges stably, representing a major step forward in 3D generation from text prompts.
     
     \item \textbf{Volumetric Rendering with Baked Quadrature Fields} introduces a novel Neural Radiance Field (NeRF) representation for non-opaque scenes, enabling fast inference by utilizing textured polygons. The authors, Gopal Sharma, Daniel Rebain, Kwang Moo Yi, Andrea Tagliasacchi from the University of British Columbia, Google DeepMind, Simon Fraser University, and the University of Toronto, address the computational expense and hardware limitations of traditional NeRF volume rendering. Their approach models the scene with polygons to obtain quadrature points required in volumetric rendering, along with their opacity and color from the texture. This method involves training a specialized field whose zero-crossings correspond to quadrature points, followed by marching cubes to obtain a polygonal mesh. The polygons are rasterized, and fragment shaders are used to obtain the final color image. This technique allows rendering on various devices and easy integration with existing graphics frameworks while retaining the benefits of volume rendering. The method is particularly suited for interactive visualization without the need for powerful GPUs, overcoming the challenge of representing scene geometry in neural volumetric rendering.
     
     \item \textbf{WavePlanes: A compact Wavelet representation for Dynamic Neural Radiance Fields} introduces a novel approach to enhance Dynamic Neural Radiance Fields (Dynamic NeRF) by addressing their resource-intensive nature and challenges in compression. The authors, Adrian Azzarelli, Nantheera Anantrasirichai, and David R Bull from the Visual Information Laboratory at the University of Bristol, propose WavePlanes, a fast and more compact explicit model. This model utilizes a multi-scale space and space-time feature plane representation using N-level 2-D wavelet coefficients. The inverse discrete wavelet transform reconstructs N feature signals at varying detail levels, which are linearly decoded to approximate the color and density of volumes in a 4-D grid. By exploiting the sparsity of wavelet coefficients, WavePlanes compresses a Hash Map containing only non-zero coefficients and their locations on each plane, resulting in a significantly reduced model size of approximately 12 MB. Compared to state-of-the-art plane-based models, WavePlanes is up to 15 times smaller, less computationally demanding, and achieves comparable results in as little as one hour of training without requiring custom CUDA code or high-performance computing resources. Additionally, the authors propose new feature fusion schemes that offer greater interpretability while maintaining performance. This advancement is significant for applications requiring dynamic scene modeling, offering a more efficient and interpretable solution for real-time rendering and compression. 

     \item \textbf{Re-Nerfing: Enforcing Geometric Constraints on Neural Radiance Fields through Novel Views Synthesis} presents an innovative multi-stage approach to improve Neural Radiance Fields (NeRFs), particularly in scenarios with limited visual overlap. The authors, Felix Tristram, Stefano Gasperini, Federico Tombari, Nassir Navab, and Benjamin Busam from the Technical University of Munich, VisualAIs, and Google, address the challenge of shape-radiance ambiguities in NeRF optimization, which often results in erroneous scene geometry and artifacts. Their method, Re-Nerfing, enhances geometric consistency and scene coverage by first training a NeRF with available views, then synthesizing pseudo-views adjacent to the original ones to simulate stereo or trifocal setups, and finally retraining a second NeRF with both original and pseudo views. This process enforces structural and epipolar constraints using the newly synthesized images. Extensive experiments on the mip-NeRF 360 dataset demonstrate the effectiveness of Re-Nerfing in both denser and sparser input scenarios, showing improvements over the state-of-the-art Zip-NeRF, even when trained with all views. This advancement is significant for applications in 3D scene representation and rendering, offering a more robust solution for content creation, rendering, and reconstruction workflows in sparse-view environments.
     
     \item \textbf{Calibrated Uncertainties for Neural Radiance Fields} introduces a groundbreaking method for obtaining calibrated uncertainties from Neural Radiance Fields (NeRF) models, a crucial component for enhancing the reliability and applicability of NeRF in various fields. The authors, Niki Amini-Naieni, Tomas Jakab, Andrea Vedaldi, and Ronald Clark from the University of Oxford, tackle the challenge of accurately measuring uncertainty in NeRF predictions. Traditional probabilistic NeRF methods, such as FlipNeRF, provide probabilistic outputs that represent uncertainty in color predictions but lack calibration, meaning they do not necessarily correspond to true confidence levels. The authors propose a robust and efficient metric to calculate per-pixel uncertainties from the predictive posterior distribution. They introduce two techniques that eliminate the need for held-out data: patch sampling, which involves training two NeRF models for each scene, and a novel meta-calibrator that requires training only one NeRF model. This approach achieves state-of-the-art uncertainty in the sparse-view setting while maintaining image quality, and is effective in applications like view enhancement and next-best view selection. This advancement is significant for autonomous systems, robotics, and augmented reality, where understanding the confidence level of predictions is crucial.
     
     \item \textbf{PointNeRF++: A multi-scale, point-based Neural Radiance Field} introduces a novel volume-rendering framework that effectively leverages point clouds for Neural Radiance Fields (NeRFs), particularly in challenging real-world situations with sparse or incomplete point clouds. The authors, Weiwei Sun, Eduard Trulls, Yang-Che Tseng, Sneha Sambandam, Gopal Sharma, Andrea Tagliasacchi, and Kwang Moo Yi from the University of British Columbia, Google Research, Google DeepMind, Simon Fraser University, and the University of Toronto, propose a simple representation that aggregates point clouds at multiple scale levels with sparse voxel grids at different resolutions. This approach addresses the limitations of existing point cloud-based neural rendering methods, which struggle with low-quality point clouds. By averaging across multiple scale levels among valid points and adding a global voxel at the coarsest scale, PointNeRF++ unifies classical and point-based NeRF formulations. The method significantly outperforms the state of the art on the NeRF Synthetic, ScanNet, and KITTI-360 datasets, demonstrating its effectiveness in novel-view synthesis from a limited number of images. This advancement is crucial for applications in uncontrolled, real-world scenarios where only a few views of a scene are available.

     \item \textbf{FINER: Flexible spectral-bias tuning in Implicit NEural Representation by Variable-periodic Activation Functions} introduces a novel approach to enhance Implicit Neural Representations (INR) by incorporating variable-periodic activation functions. The authors, Zhen Liu, Hao Zhu, Qi Zhang, Jingde Fu, Weibing Deng, Zhan Ma, Yanwen Guo, and Xun Cao from Nanjing University and Tencent Company, address the challenge of INR's restricted capability to tune their supported frequency set, which often results in imperfect performance when representing complex signals with multiple frequencies. Their solution, FINER, allows for flexible tuning of the supported frequency set by initializing the bias of the neural network within different ranges, activating sub-functions with various frequencies in the variable-periodic function. This innovation leads to improved performance in signal representation, as demonstrated in contexts such as 2D image fitting, 3D signed distance field representation, and 5D neural radiance fields optimization. FINER outperforms existing INRs, showcasing its effectiveness in a range of applications, including neural rendering, inverse imaging, and simulations.
     
     \item \textbf{C-NERF: Representing Scene Changes as Directional Consistency Difference-based NeRF} introduces a novel approach for detecting changes in a scene represented by Neural Radiance Fields (NeRFs). The authors, Rui Huang, Binbin Jiang, Qingyi Zhao, William Wang, Yuxiang Zhang, and Qing Guo from the Civil Aviation University of China, the University of South Carolina, and the Agency for Science, Technology and Research in Singapore, aim to address the challenge of detecting object variations in a scene. Their method, C-N ERF, can predict scene changes from multiple viewpoints using images captured at different times. This approach overcomes the limitations of existing NeRFs and 2D change detection methods, which often result in false or missing detections. C-N ERF performs spatial alignment of two NeRFs captured before and after changes, identifies change points based on a direction-consistent constraint, and designs a change map rendering process. This method is validated on a new dataset covering diverse scenarios with different changing objects, showing significant improvement over state-of-the-art 2D change detection and NeRF-based methods. The paper has applications in scene monitoring, measuring, remote sensing, surveillance, and robot navigation.
     
     \item \textbf{HeadGaS: Real-Time Animatable Head Avatars via 3D Gaussian Splatting} introduces a groundbreaking approach for creating real-time animatable 3D head avatars. The authors, Helisa Dhamo, Yinyu Nie, Arthur Moreau, Jifei Song, Richard Shaw, Yiren Zhou, and Eduardo Pérez-Pellitero from Huawei Noah’s Ark Lab, propose a hybrid model that combines the explicit representation of 3D Gaussian Splats (3DGS) with a base of learnable latent features. These features can be linearly blended with low-dimensional parameters from parametric head models to obtain expression-dependent final color and opacity values. HeadGaS, the first model to use 3DGS for 3D head reconstruction and animation, demonstrates state-of-the-art results in real-time inference frame rates, surpassing baselines by up to 2dB while accelerating rendering speed by over ten times. This method is essential for building digital avatars that look and behave like real humans, with applications in AR/VR, teleconferencing, and gaming. The model achieves high fidelity in appearance, is easy to capture, and enables expressive control, addressing the challenges in animatable 3D head reconstruction.

     \item \textbf{ReconFusion: 3D Reconstruction with Diffusion Priors} presents a novel approach to 3D scene reconstruction from a limited number of images using Neural Radiance Fields (NeRFs). The authors, Rundi Wu, Ben Mildenhall, Philipp Henzler, Keunhong Park, Ruiqi Gao, Daniel Watson, Pratul P. Srinivasan, Dor Verbin, Jonathan T. Barron, Ben Poole, and Aleksander Hołyński from Google Research, Columbia University, and Google DeepMind, address the challenge of NeRF's requirement for a large number of images to avoid artifacts in under-observed views. ReconFusion leverages a diffusion model trained for novel view synthesis to regularize NeRF optimization, significantly improving robustness and reducing artifacts even when the number of input views is as low as three. This method synthesizes realistic geometry and texture in underconstrained regions while preserving the appearance of observed regions. The authors demonstrate significant performance improvements over previous few-view NeRF reconstruction approaches across various real-world datasets, including forward-facing and 360-degree scenes. ReconFusion represents a major advancement in 3D reconstruction, enabling high-quality NeRFs from a minimal number of images, thus reducing the time-consuming capture process. The project page for ReconFusion is available at reconfusion.github.io.

     \item \textbf{HybridNeRF: Efficient Neural Rendering via Adaptive Volumetric Surfaces} introduces an innovative approach to neural rendering that combines the strengths of surface and volumetric representations. The authors, Haithem Turki, Vasu Agrawal, Samuel Rota Bulò, Lorenzo Porzi, Peter Kontschieder, Deva Ramanan, Michael Zollhöfer, and Christian Richardt from Meta Reality Labs and Carnegie Mellon University, address the inefficiencies in rendering Neural Radiance Fields (NeRFs), which typically require many samples per ray due to their volumetric nature. Their solution, HybridNeRF, efficiently models most real-world objects as surfaces, significantly reducing the number of samples per ray, while still using volumetric rendering for challenging regions like semi-opaque and thin structures. This hybrid approach leads to a substantial improvement in rendering speed without sacrificing quality, achieving real-time frame rates at virtual reality resolutions. HybridNeRF outperforms state-of-the-art baselines, including rasterization-based methods, by 15-30
     
     \item \textbf{Feature 3DGS: Supercharging 3D Gaussian Splatting to Enable Distilled Feature Fields} presents a groundbreaking method that enhances 3D Gaussian Splatting by integrating large 2D foundation models through feature field distillation. The authors, Shijie Zhou, Haoran Chang, Sicheng Jiang, Zhiwen Fan, Zehao Zhu, Dejia Xu, Pradyumna Chari, Suya You, Zhangyang Wang, and Achuta Kadambi from the University of California, Los Angeles, the University of Texas at Austin, and DEVCOM Army Research Laboratory, address the limitations of Neural Radiance Fields (NeRF) in rendering speed and implicitly represented feature fields. Their approach extends the capabilities of 3D Gaussian Splatting beyond novel view synthesis to include semantic segmentation, language-guided editing, and promptable segmentations. This method overcomes the continuity artifacts that reduce feature quality in NeRF pipelines and achieves state-of-the-art performance in real-time radiance field rendering. By enabling 3D Gaussian splatting on arbitrary-dimension semantic features via 2D foundation model distillation, the authors significantly advance the field of 3D scene representation. This development is crucial for applications requiring real-time rendering and semantically aware tasks like editing and segmentation.
     
     \item \textbf{SO-NeRF: Active View Planning for NeRF using Surrogate Objectives} introduces a novel approach to optimize the data-gathering process for Neural Radiance Fields (NeRF) by actively planning a sequence of views that yield maximal reconstruction quality. The authors, Keifer Lee, Shubham Gupta, Sunglyoung Kim, Bhargav Makwana, Chao Chen, and Chen Feng from New York University, propose Surrogate Objectives for Active Radiance Fields (SOAR), a set of interpretable functions that evaluate the goodness of views using geometric and photometric visual cues. These cues include surface coverage, geometric complexity, textural complexity, and ray diversity. The key innovation lies in the development of SOARNet, a deep network that infers SOAR scores, enabling effective view selection in seconds instead of hours, without prior visits to candidate views or training any radiance field during planning. The experiments demonstrate that SOARNet significantly outperforms baselines with approximately 80x speed-up while achieving better or comparable reconstruction qualities. SOAR is model-agnostic, making it applicable across various neural-implicit to explicit approaches. This advancement is crucial for applications in robotics and intelligent embodied agents, where efficient and accurate environmental representation is essential.

     \item \textbf{Evaluating the point cloud of individual trees generated from images based on Neural Radiance fields (NeRF) method} explores the application of Neural Radiance Fields (NeRF) in the 3D reconstruction of individual trees, a task of significant importance in precision forestry management and research. The authors, Hongyu Huang, Guoji Tian, Chongcheng Chen, and their team from Fuzhou University and the National Engineering Research Center of Geospatial Information Technology in China, address the challenges posed by the complex branch structures of trees and occlusions from stems, branches, and foliage. They utilize NeRF for individual tree reconstruction based on images collected by various cameras, comparing the resulting point cloud models with those derived from traditional photogrammetric reconstruction and laser scanning methods. The study finds that NeRF offers higher successful reconstruction rates, better canopy area reconstruction, and requires fewer input images compared to photogrammetric methods. While NeRF demonstrates significant advantages in reconstruction efficiency and adaptability to complex scenes, the generated point cloud tends to be noisy and of lower resolution. However, the accuracy of tree structural parameters (tree height and diameter at breast height) extracted from photogrammetric point clouds remains higher than those derived from NeRF point clouds. The results highlight the potential of NeRF for individual tree reconstruction, offering new perspectives and methods for precision forestry management.

     \item \textbf{Artist-Friendly Relightable and Animatable Neural Heads} introduces a groundbreaking method for creating neural head avatars that are both animatable and relightable, addressing the need for flexible and artist-friendly digital avatars in various applications. The authors, Yingyan Xu, Prashanth Chandran, Sebastian Weiss, Markus Gross, Gaspard Zoss, and Derek Bradley from ETH Zürich and DisneyResearch | Studios, build upon a dynamic avatar representation using a Mixture of Volumetric Primitives (MVP). This approach allows for efficient rendering of animatable neural heads with high visual quality. The key innovation lies in the ability to relight these avatars to match any distant environment map or nearfield light sources while providing full dynamic control over facial shape. This enables the playback of captured performances and the generation of novel artistically-created performances with full control over scene illumination and viewpoint. The method represents a significant advancement in the field of neural avatars, offering a solution that combines motion and illumination requirements in a single architecture. This development is crucial for applications in video games, films, VR experiences, and telepresence, where realistic and flexible digital avatars are essential.
     
     \item \textbf{Identity-Obscured Neural Radiance Fields: Privacy-Preserving 3D Facial Reconstruction} introduces a groundbreaking method for creating neural head avatars that are both animatable and relightable, addressing the need for flexible and artist-friendly digital avatars in various applications. The authors, Yingyan Xu, Prashanth Chandran, Sebastian Weiss, Markus Gross, Gaspard Zoss, and Derek Bradley from ETH Zürich and DisneyResearch | Studios, build upon a dynamic avatar representation using a Mixture of Volumetric Primitives (MVP). This approach allows for efficient rendering of animatable neural heads with high visual quality. The key innovation lies in the ability to relight these avatars to match any distant environment map or nearfield light sources while providing full dynamic control over facial shape. This enables the playback of captured performances and the generation of novel artistically-created performances with full control over scene illumination and viewpoint. The method represents a significant advancement in the field of neural avatars, offering a solution that combines motion and illumination requirements in a single architecture. This development is crucial for applications in video games, films, VR experiences, and telepresence, where realistic and flexible digital avatars are essential.
     
     \item \textbf{Towards 4D Human Video Stylization} introduces a pioneering approach for 4D (3D and time) human video stylization, integrating style transfer, novel view synthesis, and human animation within a unified framework. The authors, Tiantian Wang, Xinxin Zuo, Fangzhou Mu, Jian Wang, and Ming-Hsuan Yang from the University of California, Merced, University of Alberta, University of Wisconsin-Madison, and Snap Inc., address the limitations of existing video stylization methods that are restricted to specific viewpoints and lack the capability to generalize to novel views and poses in dynamic scenes. They leverage Neural Radiance Fields (NeRFs) to represent videos, conducting stylization in the rendered feature space. The method involves a dual representation of both the human subject and the surrounding scene using two NeRFs, facilitated by a novel geometry-guided tri-plane representation. This approach enhances feature representation robustness and enables animation across various poses and viewpoints. The stylization is performed within the NeRFs' rendered feature space, striking a balance between stylized textures and temporal coherence. The framework's unique capability to accommodate novel poses and viewpoints makes it a versatile tool for creative human video stylization. The source code and trained models are available on their GitHub page.

     \item \textbf{Multi-View Unsupervised Image Generation with Cross Attention Guidance} introduces MIRAGE, a novel pipeline for unsupervised training of a pose-conditioned diffusion model on single-category datasets. The authors, Llukman Cerkezi, Aram Davtyan, Sepehr Sameni, and Paolo Favaro from the University of Bern, Switzerland, address the challenges in novel view synthesis, particularly the scalability issues associated with Neural Radiance Field (NeRF) models that rely on precisely annotated multi-view images. MIRAGE leverages pretrained self-supervised Vision Transformers (DINOv2) to identify object poses by clustering datasets based on visibility and locations of specific object parts. The pose-conditioned diffusion model, trained on pose labels and equipped with cross-frame attention at inference time, ensures cross-view consistency. This is further enhanced by a novel hard-attention guidance mechanism. MIRAGE demonstrates superior performance in novel view synthesis on real images and robustness to diverse textures and geometries, as shown in experiments on synthetic images generated with pretrained Stable Diffusion. This advancement is significant for applications requiring novel view synthesis of real images, offering a more scalable and unsupervised approach compared to traditional NeRF models.

     \item \textbf{EAGLES: Efficient Accelerated 3D Gaussians with Lightweight EncodingS} presents a significant advancement in the field of novel-view scene synthesis using 3D Gaussian splatting (3D-GS). The authors, Sharath Girish, Kamal Gupta, and Abhinav Shrivastava from the University of Maryland, College Park, address the challenges of lengthy training times and slow rendering speeds associated with Neural Radiance Fields (NeRFs). Their method, EAGLES, utilizes quantized embeddings to drastically reduce memory storage requirements and employs a coarse-to-fine training strategy for faster and more stable optimization of Gaussian point clouds. This approach results in scene representations with fewer Gaussians and quantized representations, leading to faster training times and rendering speeds for real-time rendering of high-resolution scenes. EAGLES significantly reduces memory consumption by more than an order of magnitude while maintaining reconstruction quality. The effectiveness of this method is validated on various datasets and scenes, demonstrating its ability to preserve visual quality while consuming 10-20 times less memory and offering faster training and inference speeds. This advancement is crucial for applications requiring real-time rendering of high-resolution scenes, offering a more efficient solution compared to traditional NeRF methods. The project page and code for EAGLES are available online.
     
     \item \textbf{VOODOO 3D: Volumetric Portrait Disentanglement for One-Shot 3D Head Reenactment}  by Phong Tran, Egor Zakharov, Long-Nhat Ho, Anh Tuan Tran, Liwen Hu, and Hao Li from MBZUAI, ETH Zurich, VinAI Research, and Pinscreen, introduces a novel 3D-aware one-shot head reenactment method based on a fully volumetric neural disentanglement framework. This method is designed for real-time, high-fidelity, and view-consistent output, making it suitable for 3D teleconferencing systems based on holographic displays. Traditional 3D-aware reenactment methods often struggle with identity leakage from the driver or unnatural expressions due to their reliance on linear face models like 3DMM for disentangling appearance and expressions. VOODOO 3D addresses these issues by proposing a neural self-supervised disentanglement approach that lifts both the source image and driver video frame into a shared 3D volumetric representation based on tri-planes. This allows for manipulation with expression tri-planes extracted from driving images and rendering from any view using neural radiance fields. The paper showcases state-of-the-art performance across various datasets and high-quality 3D-aware head reenactment for challenging subjects, including non-frontal head poses and complex expressions.
     
     \item \textbf{NeuSD: Surface Completion with Multi-View Text-to-Image Diffusion} by Savva Ignatyev, Daniil Selikhanovych, Oleg Voynov, Yiqun Wang, Peter Wonka, Stamatios Lefkimmiatis, and Evgeny Burnaev from Skoltech, Russia, AIRI, Russia, Chongqing University, China, KAUST, Saudi Arabia, and AI Foundation and Algorithm Lab, Russia, presents a novel method for 3D surface reconstruction from multiple images where only part of the object is captured. The approach, NeuSD, leverages neural radiance fields for reconstructing visible parts of a surface and employs pre-trained 2D diffusion models for completing unobserved regions. Key innovations include using normal maps for Score Distillation Sampling (SDS) instead of color renderings, freezing SDS noise during training for coherent gradients and better convergence, and introducing Multi-View SDS to condition generation of non-observable surface parts without altering the underlying 2D Stable Diffusion model. The paper demonstrates significant qualitative and quantitative improvements over existing methods on the BlendedMVS dataset, offering a robust solution for realistic scenarios where camera locations around an object are not evenly sampled.

     \item \textbf{TriHuman : A Real-time and Controllable Tri-plane Representation for Detailed Human Geometry and Appearance Synthesis} by Heming Zhu, Fangneng Zhan, Christian Theobalt, and Marc Habermann from the Max Planck Institute for Informatics and Saarland Informatics Campus, Germany, presents a novel approach to creating digital doubles of real humans. The paper addresses the challenge of generating photorealistic and geometrically detailed virtual humans in real-time, a task traditionally requiring extensive work by experienced artists. TriHuman introduces a human-tailored, deformable, and efficient tri-plane representation, which non-rigidly warps global ray samples into an undeformed tri-plane texture space. This method effectively handles the mapping of global points to the same tri-plane locations and conditions the tri-plane feature representation on skeletal motion for dynamic appearance and geometry changes. The paper demonstrates TriHuman's capability in achieving real-time performance, state-of-the-art pose-controllable geometry synthesis, and photorealistic rendering quality. This advancement in neural human rendering and human modeling is significant for applications in the movie industry, gaming, telecommunication, and VR/AR. The project page is available at \url{https://vcai.mpi-inf.mpg.de/projects/trihuman}.

     \item \textbf{SwiftBrush: One-Step Text-to-Image Diffusion Model with Variational Score Distillation} by Thuan Hoang and Nguyen Anh Tran from VinAI Research, Hanoi, Vietnam, introduces a novel approach to accelerate text-to-image diffusion models. Despite the impressive capabilities of these models in generating high-resolution and diverse images from text prompts, they are often hindered by slow iterative sampling processes. SwiftBrush addresses this limitation through an image-free distillation scheme, inspired by text-to-3D synthesis. This method allows for the creation of a 3D neural radiance field that aligns with the input prompt, significantly speeding up the image generation process. The paper showcases SwiftBrush's ability to produce high-fidelity images roughly 20 times faster than Stable Diffusion, a notable advancement in the field of AI-driven image generation. SwiftBrush represents a significant step forward in making text-to-image diffusion models more efficient and practical for real-world applications.
     
     \item \textbf{Nuvo: Neural UV Mapping for Unruly 3D Representations} by Pratul P. Srinivasan, Stephan J. Garbin, Dor Verbin, Jonathan T. Barron, and Ben Mildenhall from Google Research, introduces a groundbreaking UV mapping method tailored for complex 3D geometries produced by state-of-the-art 3D reconstruction and generation models. Traditional UV mapping algorithms struggle with the fragmented texture atlases resulting from the application to volume densities recovered by Neural Radiance Fields (NeRFs) and related techniques. Nuvo, the proposed method, employs a neural field to represent a continuous UV mapping, optimizing it for only the visible points that affect the scene's appearance. This approach ensures valid and well-behaved mappings for challenging geometries, facilitating detailed appearance editing. The paper demonstrates Nuvo's robustness and its ability to produce editable UV mappings for intricate 3D models. This advancement in UV mapping technology has significant implications for 3D content creation, particularly in view synthesis and appearance editing for complex real-world meshes.
     
     \item \textbf{360° Volumetric Portrait Avatar} by Jalees Nehvi, Berna Kabadayi, Julien Valentin, and Justus Thies from the Max Planck Institute for Intelligent Systems, Technical University of Darmstadt, and Microsoft, introduces a groundbreaking method for reconstructing 360° photo-realistic portrait avatars of human subjects from monocular video inputs. The paper addresses the limitations of state-of-the-art monocular avatar reconstruction methods, which struggle with capturing side and back views due to reliance on 3DMM-based facial tracking. The authors propose a template-based tracking approach for the torso, head, and facial expressions, enabling complete coverage of a human subject's appearance from all angles. This method involves training a neural volumetric representation based on neural radiance fields, with a focus on modeling appearance changes, particularly in the mouth region. The proposed deformation-field-based blend basis allows for interpolating between different appearance states. The paper showcases the first monocular technique that reconstructs an entire 360° avatar, marking a significant advancement in the field of digital human avatar creation. The project page is available at \url{https://jalees018.github.io/3VP-Avatar}.

     \item \textbf{CoGS: Controllable Gaussian Splatting} by Heng Yu, Joel Julin, Zoltán Á. Milacski, Koichiro Niinuma, and László A. Jeni from the Robotics Institute at Carnegie Mellon University and Fujitsu Research of America, introduces a novel method for capturing and re-animating the 3D structure of articulated objects. The paper addresses the challenges of complex, resource-intensive multi-view setups and the high training and rendering costs of single-camera Neural Radiance Fields (NeRFs). CoGS, or Controllable Gaussian Splatting, offers a streamlined alternative, enabling direct manipulation of scene elements and real-time control of dynamic scenes without pre-computing control signals. The method uses time-varying 3D Gaussians to learn a dynamic 3D representation from monocular images, allowing for fine-scale control at the per-Gaussian level. CoGS outperforms existing dynamic and controllable neural representations in visual fidelity, as demonstrated in evaluations using both synthetic and real-world datasets. The paper represents a significant advancement in 3D scene reconstruction and manipulation. The project page is available at \url{https://cogs2023.github.io}.
     
     \item \textbf{IL-NeRF: Incremental Learning for Neural Radiance Fields with Camera Pose Alignment} by Letian Zhang, Ming Li, Chen Chen, and Jie Xu from Middle Tennessee State University, University of Central Florida, and University of Miami, introduces a novel framework for incremental training of Neural Radiance Fields (NeRF) in scenarios where camera poses are unknown. The paper addresses the challenge of catastrophic forgetting in NeRF when processing data sequentially, a common issue in practical applications like automotive and remote sensing. IL-NeRF's key innovation lies in selecting a set of past camera poses as references to initialize and align the camera poses of incoming image data, followed by a joint optimization of camera poses and replay-based NeRF distillation. This approach enables efficient and effective incremental learning of 3D scenes, significantly improving rendering quality by up to 54.04\% compared to baselines. The paper demonstrates IL-NeRF's capability to handle incremental NeRF training in real-world indoor and outdoor scenes, offering a practical solution for dynamic 3D scene reconstruction. The project page is available at \url{https://ilnerf.github.io/}.

     \item \textbf{NeVRF: Neural Video-based Radiance Fields for Long-duration Sequences} by Minye Wu and Tinne Tuytelaars from KU Leuven, introduces a novel approach to creating dynamic radiance fields using Neural Radiance Fields (NeRF) for long-duration dynamic sequences. The paper addresses the challenges of balancing quality and storage size and handling complex scene changes in dynamic sequences. The proposed Neural Video-based Radiance Field (NeVRF) representation combines neural radiance field with image-based rendering, enabling photo-realistic novel view synthesis for long-duration inward-looking scenes. NeVRF introduces a multi-view radiance blending approach to predict radiance directly from multi-view videos and incorporates continual learning techniques for efficient frame reconstruction from sequential data without revisiting previous frames. This allows for long-duration free-viewpoint video and, with a tailored compression approach, compact representation of dynamic scenes. The paper demonstrates the effectiveness of NeVRF in rendering long-duration sequences, reconstructing sequential data, and compact data storage, making dynamic radiance fields more practical for real-world applications.

     \item \textbf{Learning for CasADi: Data-driven Models in Numerical Optimization} by Tim Salzmann, Jon Arrizabalaga, Joel Andersson, Marco Pavone, and Markus Ryll from the Munich Institute of Robotics and Machine Intelligence, Technical University of Munich, Germany, and other institutions, introduces the Learning for CasADi (L4CasADi) framework. This framework bridges the gap between deep learning and numerical optimization by enabling the seamless integration of PyTorch-learned models with CasADi, a tool for efficient and potentially hardware-accelerated numerical optimization. L4CasADi is demonstrated through two tutorial examples: optimizing a fish’s trajectory in a turbulent river using a PyTorch model to represent the turbulent flow, and leveraging an implicit Neural Radiance Field for optimal control. The paper highlights the challenges of integrating learned process models into numerical optimizations and presents L4CasADi as a solution that combines the strengths of deep learning and numerical optimization tools. This work represents a significant advancement in data-driven control systems, offering a novel approach to solving complex real-world problems. The L4CasADi framework is available under the MIT license at \url{https://github.com/Tim-Salzmann/l4casadi}.
     
     \item \textbf{NVFi: Neural Velocity Fields for 3D Physics Learning from Dynamic Videos} by Jinxi Li, Ziyang Song, and Bo Yang from the vLAR Group at The Hong Kong Polytechnic University, introduces a novel method to model 3D scene dynamics from multi-view videos. Unlike existing works that focus primarily on novel view synthesis within a training time period, NVFi aims to simultaneously learn the geometry, appearance, and physical velocity of 3D scenes solely from video frames. This approach enables multiple applications, including future frame extrapolation, unsupervised 3D semantic scene decomposition, and dynamic motion transfer. The method comprises three major components: a keyframe dynamic radiance field, an interframe velocity field, and a joint keyframe and interframe optimization module, which is crucial for effectively training both networks. To validate their method, the authors introduce two dynamic 3D datasets: the Dynamic Object dataset and the Dynamic Indoor Scene dataset. Extensive experiments demonstrate NVFi's superior performance over baselines, particularly in future frame extrapolation and unsupervised 3D semantic scene decomposition. The paper represents a significant advancement in understanding and modeling dynamic 3D scenes. The code and data are available at \url{https://github.com/vLAR-group/NVFi}.
     
     \item \textbf{CorresNeRF: Image Correspondence Priors for Neural Radiance Fields}by Yixing Lao, Xiaogang Xu, Zhipeng Cai, Xihui Liu, and Hengshuang Zhao from The University of Hong Kong, Zhejiang Lab, Zhejiang University, and Intel Labs, introduces a novel method to enhance Neural Radiance Fields (NeRFs) for novel view synthesis and surface reconstruction, particularly in scenarios with sparse input views. The paper presents CorresNeRF, which leverages image correspondence priors computed by off-the-shelf methods to supervise NeRF training. The authors design adaptive processes for augmentation and filtering to generate dense and high-quality correspondences, which are then used to regularize NeRF training through correspondence pixel reprojection and depth loss terms. This approach significantly improves both photometric and geometric metrics in NeRF models, demonstrating its effectiveness across different datasets and NeRF variants. The paper showcases how this simple yet effective technique of using correspondence priors can be applied as a plug-and-play module, enhancing the performance of NeRFs in real-world applications like 3D portrait reconstruction and city digital reconstruction. The project page is available at \url{https://yxlao.github.io/corres-nerf/}.

     \item \textbf{Learning Naturally Aggregated Appearance for Efficient 3D Editing} by Ka Leong Cheng, Qiuyu Wang, Zifan Shi, Kecheng Zheng, Yinghao Xu, Hao Ouyang, Qifeng Chen, and Yujun Shen from HKUST, Ant Group, CAD\&CG ZJU, and Stanford, introduces AGAP, a novel approach for efficient 3D editing using Neural Radiance Fields (NeRF). The paper addresses the challenge of editing 3D scenes represented by NeRF, which are typically unfavorable for editing due to their implicit nature. AGAP replaces the color field in NeRF with an explicit 2D appearance aggregation, also known as a canonical image. This allows users to easily customize their 3D editing using 2D image processing tools. To facilitate convenient editing and avoid distortion, AGAP includes a projection field that maps 3D points onto 2D pixels for texture lookup. This field is initialized with a pseudo canonical camera model and optimized with offset regularity to ensure the naturalness of the aggregated appearance. The method supports various 3D editing tasks, such as stylization, interactive drawing, and content extraction, without the need for re-optimization for each case. The paper demonstrates the generalizability and efficiency of AGAP in 3D editing. The project page is available at \url{https://felixcheng97.github.io/AGAP/}.
     
     \item \textbf{TeTriRF: Temporal Tri-Plane Radiance Fields for Efficient Free-Viewpoint Video} by Minye Wu, Zehao Wang, Georgios Kouros, and Tinne Tuytelaars from KU Leuven, introduces a novel technology, Temporal Tri-Plane Radiance Fields (TeTriRF), to address the significant storage and computational complexity challenges in Neural Radiance Fields (NeRF) for Free-Viewpoint Video (FVV). TeTriRF employs a hybrid representation combining tri-planes and voxel grids, enabling scaling to long-duration sequences and scenes with complex motions or rapid changes. The authors propose a group training scheme for high training efficiency and temporally consistent, low-entropy scene representations. This approach, coupled with a compression pipeline using off-the-shelf video codecs, achieves an order of magnitude less storage size compared to state-of-the-art methods. TeTriRF demonstrates competitive quality with a higher compression rate, making it a significant advancement in FVV technology. The project page is available at \url{https://wuminye.github.io/projects/TeTriRF/}. 

     \item \textbf{WaterHE-NeRF: Water-ray Tracing Neural Radiance Fields for Underwater Scene Reconstruction} by Jingchun Zhou, Tianyu Liang, Zongxin He, Dehuan Zhang, Weishi Zhang, Xianping Fu, and Chongyi Li from Dalian Maritime University and Nankai University, introduces a novel approach to underwater scene reconstruction using Neural Radiance Fields (NeRF). The paper addresses the challenges posed by light attenuation in water, which significantly affects the quality of underwater images. The authors propose WaterHE-NeRF, a method that incorporates water-ray tracing based on Retinex theory to accurately encode color, density, and illuminance attenuation in three-dimensional space. This approach allows for the generation of both degraded and clear multi-view images, optimizing image restoration by combining reconstruction loss with Wasserstein distance. Additionally, the use of histogram equalization (HE) as pseudo-ground truth enhances the network's accuracy in preserving original details and color distribution. The effectiveness of WaterHE-NeRF is validated through extensive experiments on real underwater datasets and synthetic datasets. The paper discusses advancements in underwater imaging and scene reconstruction, offering a solution that effectively handles the unique challenges of underwater light transport.

     \item \textbf{COLMAP-Free 3D Gaussian Splatting} by Yang Fu, Sifei Liu, Amey Kulkarni, Jan Kautz, Alexei A. Efros, and Xiaolong Wang from UC San Diego, NVIDIA, and UC Berkeley, presents an innovative approach for novel view synthesis and camera pose estimation without known camera parameters. This method, named CF-3DGS, leverages the explicit geometric representation of 3D Gaussian Splatting and the continuity of input video streams. It processes input frames sequentially, progressively growing the 3D Gaussians set by incorporating one input frame at a time, thus eliminating the need for pre-computed camera poses. This approach significantly improves view synthesis and camera pose estimation, especially under large motion changes, compared to previous methods. The paper addresses the challenges in scene reconstruction and novel view synthesis that arise from the reliance on Structure-from-Motion (SfM) preprocessing, which can be time-consuming and error-prone. By integrating pose estimation directly within the framework, CF-3DGS offers a solution to the long-standing chicken-and-egg problem in 3D scene reconstruction and camera registration. The project page is available at \url{https://oasisyang.github.io/colmap-free-3dgs}.
     
     \item \textbf{SMERF: Streamable Memory Efficient Radiance Fields for Real-Time Large-Scene Exploration} by Daniel Duckworth, Peter Hedman, Christian Reiser, Peter Zhizhin, Jean-François Thibert, Mario Lučić, Richard Szeliski, and Jon Barron from Google DeepMind, Google Research, Google Inc., Tübingen AI Center, and the University of Tübingen, introduces an innovative approach for real-time view synthesis of large scenes. The paper addresses the challenge of balancing high-quality rendering with the computational and memory constraints of real-time applications. SMERF achieves state-of-the-art accuracy among real-time methods for large scenes up to 300 m² at a volumetric resolution of 3.5 mm³. The method is based on two primary contributions: a hierarchical model partitioning scheme that increases model capacity while controlling compute and memory usage, and a distillation training strategy that ensures high fidelity and internal consistency. SMERF enables full six degrees of freedom (6DOF) navigation within a web browser and renders in real-time on various devices, including smartphones and laptops. The authors demonstrate that SMERF surpasses current real-time novel view synthesis methods in standard benchmarks and large scenes, offering a significant speed advantage over state-of-the-art radiance field models. The paper is a major advancement in real-time rendering technology, particularly for large-scale scenes. The project website is available at \url{https://smerf-3d.github.io}.
     
     \item \textbf{3DGEN: A GAN-based approach for generating novel 3D models from image data} by Antoine Schnepf, Flavian Vasile, and Ugo Tanielian from Criteo AI Lab, introduces a novel approach for generating 3D models from images using Generative Adversarial Networks (GANs). The paper addresses the relatively unexplored area of 3D model generation, which has significant potential in fields like game design, video production, and physical product design. The authors present 3DGEN, a model that combines the strengths of Neural Radiance Fields (NeRF) for object reconstruction and GAN-based image generation. This architecture is capable of generating plausible meshes for objects within the same category as the training images, showing a notable improvement in generation quality compared to state-of-the-art baselines. The paper highlights the rapid advancements in machine-assisted creativity, particularly in image synthesis, and proposes a solution to bridge the gap between reconstruction and generation in 3D modeling. This work represents a significant step forward in the application of GANs and NeRF for creative and design purposes in 3D environments.

     \item \textbf{Neural Radiance Fields for Transparent Object Using Visual Hull} by Heechan Yoon and Seungkyu Lee from Kyunghee University, South Korea, addresses the complex challenge of novel view synthesis for transparent objects, a task that is particularly difficult due to the refraction of light on transparent surfaces causing visual distortions. The authors propose a method based on Neural Radiance Fields (NeRF), which has shown remarkable performance in view synthesis. However, NeRF's limitation lies in its inability to consider refracted light rays on transparent object surfaces. To overcome this, the authors introduce a three-step process: reconstructing the 3D shape of a transparent object using visual hull, simulating the refraction of rays inside the transparent object according to Snell’s law, and sampling points through refracted rays for input into NeRF. This method effectively addresses the limitations of conventional NeRF with transparent objects, offering a more accurate and realistic rendering of scenes containing transparent materials. The paper represents a significant advancement in the field of computer vision and graphics, particularly in rendering and synthesizing views of transparent objects.
     
     \item \textbf{ProNeRF: Learning Efficient Projection-Aware Ray Sampling for Fine-Grained Implicit Neural Radiance Fields} by Juan Luis Gonzalez Bello, Minh-Quan Viet Bui, and Munchurl Kim from Korea Advanced Institute of Science and Technology (KAIST), South Korea, introduces a novel method to enhance Neural Radiance Fields (NeRF) for 3D scene reconstruction and view synthesis. The paper addresses the challenge of achieving a balance between memory footprint, speed, and quality in NeRF-based methods. ProNeRF is equipped with a Projection-Aware Sampling (PAS) network and a new training strategy for ray exploration and exploitation, enabling efficient fine-grained particle sampling. This approach allows ProNeRF to significantly reduce inference times while maintaining superior image quality and detail compared to existing high-quality methods. The method achieves state-of-the-art metrics, being significantly faster and yielding higher PSNR than NeRF and other sampler-based methods. ProNeRF's effectiveness is demonstrated through extensive experimental results on forward-facing and 360 datasets, LLFF and Blender, respectively. The paper represents a significant advancement in the field of neural rendering, offering an optimal trade-off between memory, speed, and quality.

     \item \textbf{SpectralNeRF: Physically Based Spectral Rendering with Neural Radiance Field} introduces a groundbreaking approach to spectral rendering in computer graphics using Neural Radiance Fields (NeRF). The authors, Ru Li, Jia Liu, Guanghui Liu, Shengping Zhang, Bing Zeng, and Shuaicheng Liu from Harbin Institute of Technology, Weihai, China, and University of Electronic Science and Technology of China, Chengdu, China, propose SpectralNeRF, an end-to-end NeRF-based architecture that generates high-quality physically based renderings from a novel spectral perspective. This method involves generating a series of spectrum maps across different wavelengths and combining them to produce RGB output. SpectralNeRF's architecture, comprising SpectralMLP and Spectrum Attention UNet (SAUNet), constructs spectral radiance fields to obtain spectrum maps of novel views, which are then processed to produce RGB images under white-light illumination. This approach represents a more physically-based method of ray-tracing and improves the performance of NeRF-based methods in complex scenes. The experimental results demonstrate SpectralNeRF's superiority over recent NeRF-based methods in synthesizing new views on both synthetic and real datasets. The paper offers significant advancements in spectral rendering, providing a novel solution for generating photorealistic images with detailed spectral information.

     \item \textbf{CF-NeRF: Camera Parameter Free Neural Radiance Fields with Incremental Learning} introduces a groundbreaking approach to 3D reconstruction and novel view synthesis using Neural Radiance Fields (NeRF) without the need for traditional camera parameters. The authors, Qingsong Yan, Qiang Wang, Kaiyong Zhao, Jie Chen, Bo Li, Xiaowen Chu, and Fei Deng from Wuhan University, Harbin Institute of Technology (Shenzhen), XGRIDS, Hong Kong Baptist University, The Hong Kong University of Science and Technology, and Hubei Luojia Laboratory, propose CF-NeRF, a method that incrementally reconstructs 3D representations and recovers camera parameters inspired by incremental structure from motion (SfM). This approach addresses the limitations of existing NeRF variants that rely on complex pipelines for camera parameter provision and struggle with rotation scenarios in practice. CF-NeRF estimates camera parameters image by image, reconstructing the scene through initialization, implicit localization, and optimization. Tested on the challenging real-world dataset NeRFBuster, CF-NeRF demonstrates robustness to camera rotation and achieves state-of-the-art results without requiring prior information or constraints. This method significantly simplifies the process of 3D reconstruction and novel view synthesis, making it more accessible and efficient for practical applications.
     
     \item \textbf{VaLID: Variable-Length Input Diffusion for Novel View Synthesis} introduces a groundbreaking approach to enhance Novel View Synthesis (NVS) using generative modeling, specifically addressing the limitations of existing methods in terms of flexibility and input variability. The authors, Shijie Li, Farhad G. Zanjani, Haitam Ben Yahia, Yuki M. Asano, Juergen Gall, and Amirhossein Habibian from the University of Bonn, Qualcomm AI Research, and the University of Amsterdam, propose VaLID, a method that processes each pose-image pair separately and fuses them into a unified visual representation. This representation guides image synthesis at target views, overcoming the challenge of inconsistency and computational costs as the number of input source-view images increases. The Multi-view Cross Former module in VaLID maps variable-length input data to fixed-size output data, and a two-stage training strategy is introduced for improved efficiency. The method demonstrates effectiveness over multiple datasets, outperforming previous approaches in NVS. The paper promises significant advancements in AR/VR applications and content creation, offering a robust and efficient solution for synthesizing high-quality and visually consistent views from variable input sources.

     \item \textbf{Scene 3-D Reconstruction System in Scattering Medium} presents an innovative approach to underwater 3D reconstruction using Neural Radiance Fields (NeRF). The authors, Zhuoyifan Zhang, Lu Zhang, Liang Wang, and Haoming Wu from Beijing University of Posts and Telecommunications and Hainan University, tackle the challenges of poor image quality and low rendering efficiency in underwater scenes. Their system begins by enhancing underwater videos captured by a monocular camera to correct image quality issues caused by water's physical properties, ensuring consistency across frames. Keyframe selection is then performed to optimize resource utilization and mitigate the impact of dynamic objects on reconstruction. The selected keyframes, after pose estimation using COLMAP, are processed through a 3D reconstruction improvement using NeRF based on multi-resolution hash coding. This method allows for rapid, high-quality 3D reconstruction of underwater scenes, addressing key issues in existing underwater 3D reconstruction systems. The paper contributes significantly to the field of underwater scene reconstruction, offering a novel solution that combines image enhancement and advanced neural radiance fields techniques.

     \item \textbf{iComMa: Inverting 3D Gaussians Splatting for Camera Pose Estimation via Comparing and Matching} introduces a novel method for 6D pose estimation in computer vision, particularly addressing the challenges in scenarios with significant initial angular deviations. The authors, Yuan Sun, Xuan Wang, Yunfan Zhang, Jie Zhang, Caigui Jiang, Yu Guo, and Fei Wang from Xi’an Jiaotong University and Ant Group, propose iComMa, which models pose estimation as the problem of inverting 3D Gaussian Splatting (3DGS) with both comparing and matching loss. Unlike conventional methods that rely on a target's CAD model or specific network training for particular object classes, iComMa employs a render-and-compare strategy for precise pose estimation and a matching module to enhance robustness against adverse initializations by minimizing distances between 2D keypoints. This approach systematically incorporates the characteristics of render-and-compare and matching-based methods, effectively addressing intricate and challenging scenarios, including those with substantial angular deviations. The experimental results on synthetic and complex real-world data demonstrate the superior precision and robustness of iComMa in challenging scenarios, making it a significant contribution to fields like robotics, SLAM, augmented reality, and virtual reality.
     
     \item \textbf{Aleth-NeRF: Illumination Adaptive NeRF with Concealing Field Assumption} introduces a novel approach to enhance Neural Radiance Fields (NeRF) for processing images captured under challenging lighting conditions, such as low light or over-exposure. The authors, Ziteng Cui, Lin Gu, Xiao Sun, Xianzheng Ma, Yu Qiao, and Tatsuya Harada from The University of Tokyo, Shanghai AI Laboratory, RIKEN AIP, and the University of Oxford, propose the concept of a "Concealing Field." This field assigns transmittance values to the surrounding air to account for illumination effects, enabling NeRF to learn reasonable density and color estimations for objects even in dimly lit situations. The Concealing Field also mitigates over-exposed emissions during the rendering stage. This innovative approach allows Aleth-NeRF to generate novel views with natural illumination from multi-view images taken in adverse lighting conditions. The authors also present a comprehensive multi-view dataset captured under challenging illumination conditions for evaluation.

     \item \textbf{ColNeRF: Collaboration for Generalizable Sparse Input Neural Radiance Field} presents a novel approach to enhance Neural Radiance Fields (NeRF) for synthesizing novel views from sparse input. The authors, Zhangkai Ni, Peiqi Yang, Wenhan Yang, Hanli Wang, Lin Ma, and Sam Kwong from Tongji University, Peng Cheng Laboratory, Meituan, and City University of Hong Kong, address the challenges posed by sparse input in NeRF, which typically requires dense supervision for high-quality results. ColNeRF introduces a collaborative module that aligns information from various views and imposes self-supervised constraints to ensure multi-view consistency in both geometry and appearance. The proposed Collaborative Cross-View Volume Integration module (CCVI) captures complex occlusions and infers the spatial location of objects. Additionally, the method uses self-supervision of target rays projected in multiple directions to ensure geometric and color consistency. This collaboration at both input and output ends enables ColNeRF to capture richer and more generalized scene representations, leading to higher-quality novel view synthesis. The extensive experimental results demonstrate that ColNeRF outperforms state-of-the-art sparse input generalizable NeRF methods and exhibits superiority in fine-tuning for new scenes, reducing computational costs significantly.

     \item \textbf{3DGS-Avatar: Animatable Avatars via Deformable 3D Gaussian Splatting} introduces an efficient method for creating animatable human avatars from monocular videos. The authors, Zhiyin Qian, Shaofei Wang, Marko Mihajlovic, Andreas Geiger, and Siyu Tang from ETH Zürich, University of Tübingen, and Tübingen AI Center, leverage 3D Gaussian Splatting to overcome the limitations of existing methods based on Neural Radiance Fields (NeRFs). These traditional methods, while achieving high-quality novel-view/novel-pose image synthesis, require extensive training time and are slow at inference. The proposed approach uses 3D Gaussian Splatting and a non-rigid deformation network to reconstruct animatable clothed human avatars that can be trained within 30 minutes and rendered at real-time frame rates (50+ FPS). The explicit nature of their representation allows for as-isometric-as-possible regularizations on both the Gaussian mean vectors and the covariance matrices, enhancing the model's generalization on highly articulated unseen poses. The experimental results show that their method achieves comparable or even better performance than state-of-the-art approaches in animatable avatar creation from a monocular input, while being significantly faster in training and inference. 
     
     \item \textbf{OccNeRF: Self-Supervised Multi-Camera Occupancy Prediction with Neural Radiance Fields} introduces a novel method for reconstructing 3D structures of surrounding environments, particularly beneficial for autonomous driving planning and navigation. The authors, Chubin Zhang, Juncheng Yan, Yi Wei, Jiaxin Li, Li Liu, Yansong Tang, Yueqi Duan, and Jiwen Lu from Tsinghua University, Beijing National Research Center for Information Science and Technology, Gaussian Robotics, and Xiaomi Car, address the challenge of generating occupancy ground truth without relying on LiDAR point clouds, which are not available in vision-based systems. Their approach, OccNeRF, employs neural rendering to convert occupancy fields into multi-camera depth maps, supervised by multi-frame photometric consistency. For semantic occupancy prediction, they utilize strategies to refine prompts and filter outputs from a pretrained open-vocabulary 2D segmentation model. Extensive experiments on the nuScenes dataset for self-supervised depth estimation and semantic occupancy prediction tasks demonstrate the effectiveness of OccNeRF. The method offers a significant advancement in vision-based perception, particularly in the context of autonomous driving, where understanding the 3D world is crucial.
     
     \item \textbf{ZeroRF: Fast Sparse View 360° Reconstruction with Zero Pretraining} introduces a novel per-scene optimization method for efficiently reconstructing 360° views from sparse input using neural field representations. The authors, Ruoxi Shi, Xinyue Wei, Cheng Wang, and Hao Su from UC San Diego, address the challenges faced by Neural Radiance Fields (NeRF) and similar approaches when dealing with sparse input views. ZeroRF integrates a tailored Deep Image Prior into a factorized NeRF representation, uniquely parametrizing feature grids with a neural network generator. This approach enables efficient sparse view 360° reconstruction without any pretraining or additional regularization. ZeroRF demonstrates versatility and superiority in both quality and speed, achieving state-of-the-art results on benchmark datasets. Its significance extends to applications in 3D content generation and editing, offering a fast and efficient solution for high-fidelity image synthesis from limited views.
     
     \item In \textbf{SLS4D (Sparse Latent Space for 4D Novel View Synthesis)}, , the authors introduce a novel 4D representation technique for dynamic neural radiance fields. This approach, termed SLS4D, focuses on reducing the number of network parameters required for dynamic NeRF (Neural Radiance Fields) by utilizing a sparse latent space. The method abandons the direct use of coordinate relationships, instead relying on relative relationships in a more compact and informative latent feature space. This shift allows the model to capture high-frequency temporal information more accurately, leading to improved dynamic alignment and rendering quality. The paper highlights the use of time slots to effectively fit dynamics and characterize radiance fields in a latent space. The spatial latent feature space learns adaptive weights with an attention mechanism, integrating more global priors and achieving superior rendering quality. The authors' extensive experiments on public datasets demonstrate that SLS4D achieves state-of-the-art results in dynamic novel view synthesis, offering higher quality rendering with significantly fewer parameters compared to previous works. In summary, the paper contributes a sparse latent space for 4D representation, a spatial latent feature space with attention-based adaptive weights, and a time slot encoding for enhanced temporal information processing. These innovations collectively improve the precision of dynamic scene representation and the quality of novel-view synthesis.
     
     \item In \textbf{RANRAC (Robust Neural Scene Representations via Random Ray Consensus}, the authors focus on enhancing the robustness of neural scene representation and rendering techniques against inconsistencies and occlusions in observations. They propose a novel combination of these techniques with dedicated outlier removal methods, such as RANSAC. This approach aims to improve robustness by removing the influence of outliers, distinguishing between inliers and outliers in the data, and optimizing neural fields based on inliers. The method exhibits robustness and versatility, accommodating a wide range of neural fields-based reconstruction methods. It inherits the strengths of RANSAC in handling various classes of outliers without relying on semantics. The paper validates the approach using synthetic data for multi-class single-shot reconstruction with LFNs and demonstrates significant quality improvements over the baseline in the presence of occlusions. The authors also showcase robust photorealistic reconstructions of 3D objects using unconditioned Neural Radiance Fields (NeRFs) from real-world image sequences with distractors. The key contributions include a robust RANSAC-based reconstruction method for multi-class single-shot reconstruction via LFNs, an analysis of RANSAC’s hyperparameters and convergence expectations, and a qualitative/quantitative evaluation of the algorithm.
     
     \item \textbf{PLGSLAM (Progressive Neural Scene Represenation with Local to Global Bundle Adjustment)} by Tianchen Deng et al. presents PLGSLAM, a novel neural visual SLAM system designed for high-fidelity surface reconstruction and robust camera tracking in real-time. Addressing the challenges of large indoor scenes and long sequences, PLGSLAM introduces a progressive scene representation method that dynamically allocates new local scene representations trained with frames within a local sliding window. This approach enhances scalability and robustness, even under pose drifts. The system utilizes tri-planes for local high-frequency features and incorporates multi-layer perceptron (MLP) networks for low-frequency features, smoothness, and scene completion in unobserved areas. Additionally, the authors propose a local-to-global bundle adjustment method with a global keyframe database to mitigate increased pose drifts in long sequences. The experimental results demonstrate PLGSLAM's state-of-the-art performance in scene reconstruction and tracking across various datasets and scenarios, including both small and large-scale indoor environments. The authors plan to open-source the code upon paper acceptance, indicating potential widespread applications in fields like autonomous driving, robotics, and virtual/augmented reality.
     
     \item In \textbf{LAENeRF: Local Appearance Editing for Neural Radiance Fields}, the authors address the limitations of existing methods in editing Neural Radiance Fields (NeRFs) by introducing LAENeRF, a novel approach for local appearance editing of pre-trained NeRFs. This method is built upon NeRFShop and Instant-NGP (iNGP), utilizing a 3-dimensional grid as the primitive for selecting scene content. LAENeRF's unique feature is its ability to model smooth transitions to adjacent content, resulting in more visually appealing edits. It introduces a NeRF-like module that learns a palette-based decomposition of colors within a selected region, significantly reducing memory requirements and increasing performance. The method ensures multi-view consistency and allows for interactive recoloring and style transfer of the selected region. The authors demonstrate that LAENeRF is not only the first interactive approach for NeRF appearance editing but also outperforms previous methods in local recoloring and stylization, both qualitatively and quantitatively. This advancement has practical implications in fields requiring detailed and localized editing of 3D scenes, such as virtual reality, game development, and film production.
     
     \item In \textbf{SlimmeRF: Slimmable Radiance Fields} by Shiran Yuan, Hao Zhao, and others, the authors introduce SlimmeRF, a novel approach to Neural Radiance Field (NeRF) models. NeRF and its variants have been successful in novel view synthesis and 3D scene reconstruction, but they often face a trade-off between model size and accuracy. SlimmeRF addresses this by allowing for test-time trade-offs between model size and accuracy, making it suitable for various computing budgets. This is achieved through a newly proposed algorithm named Tensorial Rank Incrementation (TRaIn), which incrementally increases the rank of the model's tensorial representation during training. The authors observe that SlimmeRF is particularly effective in sparse-view scenarios, sometimes even improving accuracy after slimming. This is attributed to the model's ability to discard erroneous information stored in components of higher ranks. The paper's findings have significant implications for applications requiring flexible, high-quality 3D scene reconstruction and view synthesis across different hardware capabilities.
     
     \item In \textbf{MVHuman (Tailoring 2D Diffusion with Multi-view Sampling For Realistic 3D Human Generation)}, the authors present MVHuman, a novel approach for generating human radiance fields from text guidance. This method leverages pre-trained Stable Diffusions to directly sample consistent multi-view images without requiring fine-tuning or distilling. The core of MVHuman is a multi-view sampling strategy that tailors the denoising processes of the pre-trained network. This includes view-consistent conditioning, using "consistency-guided noises," optimizing latent codes, and employing cross-view attention layers. The generated multi-view images are then used for geometry refinement and 3D radiance field generation, followed by a neural blending scheme for free-view rendering. The authors demonstrate the effectiveness of MVHuman through extensive experiments, highlighting its superiority over existing 3D human generation methods. This technique holds potential for applications in telepresence and immersive experiences in VR/AR, offering a more accessible way to create realistic human avatars.
     
     \item In \textbf{Learning Dense Correspondence for NeRF-Based Face Reenactment} by Songlin Yang et al., the authors address the challenge of face reenactment, particularly focusing on establishing dense correspondence between various face representations for motion transfer. They utilize Neural Radiance Field (NeRF) as the fundamental representation, enhancing the performance of multi-view face reenactment in terms of photo-realism and 3D consistency. The key contribution of this work is the proposal of a novel framework that adopts tri-planes as the fundamental NeRF representation and decomposes face tri-planes into three components: canonical tri-planes, identity deformations, and motion. A significant aspect of their method is the Plane Dictionary (PlaneDict) module, which efficiently maps motion conditions to a linear weighted addition of learnable orthogonal plane bases. This framework is notable for being the first to achieve one-shot multi-view face reenactment without relying on a 3D parametric model prior, demonstrating superior results in fine-grained motion control and identity preservation compared to previous methods.
     
     \item In \textbf{PNeRFLoc (Visual Localization with Point-based Neural Radiance Fields)} by Boming Zhao, Luwei Yang, Mao Mao, Hujun Bao, and Zhaopeng Cui, the authors introduce a novel framework for visual localization, named PNeRFLoc. This framework integrates structure-based methods and rendering-based optimization with Neural Radiance Fields (NeRF) representation. The key innovation lies in its dual approach: initially estimating pose by matching 2D and 3D feature points, akin to traditional methods, and subsequently refining the pose using novel view synthesis through rendering-based optimization. The authors developed a feature adaption module to bridge the gap between features used for visual localization and neural rendering, and an efficient rendering-based framework with a warping loss function. PNeRFLoc demonstrated superior performance on synthetic datasets where the 3D NeRF model is well-learned and outperformed existing NeRF-boosted localization methods on real-world benchmark datasets. This approach holds significant promise for applications in robot navigation, augmented reality, and virtual reality, where precise visual localization is crucial.
     
     \item In \textbf{AE-NeRF (Audio Enhanced Neural Radiance Field for Few Shot Talking Head Synthesis)} by Dongze Li et al., the authors address the challenge of synthesizing realistic talking heads with limited data. They introduce Audio Enhanced Neural Radiance Field (AE-NeRF), a novel approach that significantly improves the quality and fidelity of generated portraits, even with a few-shot dataset. The key innovation lies in the Audio Aware Aggregation module, which fuses features based on audio similarity between reference and target images. Additionally, the Audio-Aligned Face Generation strategy employs a dual-NeRF framework to separately model audio-related and audio-independent facial regions. This method shows superior performance in image fidelity, audio-lip synchronization, and generalization ability, outperforming state-of-the-art methods, especially in scenarios with limited training data or iterations. The paper's contributions are particularly relevant for applications in digital human creation, filmmaking, and virtual reality, where high-quality, audio-driven talking head generation is essential.
     
     \item \textbf{GauFRe (Gaussian Deformation Fields for Real-time Dynamic Novel View Synthesis)} introduces a method for dynamic scene reconstruction using deformable 3D Gaussians, tailored for monocular video. The authors, Yiqing Liang, Numair Khan, Zhengqin Li, Thu Nguyen-Phuoc, Douglas Lanman, James Tompkin, and Lei Xiao, build upon the efficiency of Gaussian splatting and extend it to accommodate dynamic elements. This is achieved through a deformable set of Gaussians in a canonical space and a time-dependent deformation field defined by a multi-layer perceptron (MLP). The method also includes a static Gaussian point cloud to focus the MLP's representational power on largely static natural scenes. The concatenated dynamic and static point clouds form the input for the Gaussian Splatting rasterizer, enabling real-time rendering. The pipeline is optimized end-to-end with a self-supervised rendering loss. This approach allows for faster optimization and rendering compared to state-of-the-art dynamic neural radiance field methods, making it a significant contribution to the field of 3D reconstruction from RGB images, particularly in scenarios involving continual deformation, like human subjects.
     
     \item In \textbf{FastSR-NeRF (Improving NeRF Efficiency on Consumer Devices with A Simple Super-Resolution Pipeline)} by Chien-Yu Lin and colleagues from the University of Washington and Apple, Inc., the authors address the challenge of enhancing the efficiency of Neural Radiance Field (NeRF) models on consumer devices. They propose a straightforward pipeline that combines existing NeRF models with super-resolution (SR) techniques. This approach, termed FastSR-NeRF, aims to upscale NeRF outputs while maintaining high-quality image generation and significantly boosting inference speeds. A key innovation in their method is the use of a lightweight augmentation technique called random patch sampling, which reduces the SR computing overhead and allows for faster training—up to 23 times faster than existing methods. This makes it feasible to run these models on consumer devices like Apple MacBooks. The authors demonstrate that their pipeline can upscale NeRF outputs by 2-4 times, increasing inference speeds by up to 18 times on an NVIDIA V100 GPU and 12.8 times on an M1 Pro chip. The paper concludes that SR is a simple yet effective technique for improving the efficiency of NeRF models for consumer-level applications.
     
     \item \textbf{Text-Image Conditioned Diffusion for Consistent Text-to-3D Generation} by Yuze He et al. addresses the challenge of generating consistent 3D models from textual descriptions. The authors propose a novel approach that leverages pre-trained 2D diffusion models, specifically Neural Radiance Fields (NeRFs), to optimize text-to-3D generation. Traditional methods often struggle with maintaining multi-view consistency due to the stochastic nature of diffusion models, leading to issues like the generation of multiple copies of content or drifted content from the text prompt. To overcome this, the authors introduce multi-view image conditions into the NeRF optimization process, ensuring fine-grained view consistency. This method effectively reduces the occurrence of floaters (caused by excessive densities) and empty spaces (due to insufficient densities) in the generated 3D models. The paper demonstrates that this approach achieves state-of-the-art performance on the T3Bench dataset. This advancement has significant implications for digital asset generation in fields like gaming, where it can enhance creativity and efficiency in the design process.
     
     \item \textbf{MixRT (Mixed Neural Representations For Real-Time NeRF Rendering)} by Chaojian Li et al. introduces a novel approach to Neural Radiance Field (NeRF) rendering, particularly focusing on real-time applications in large-scale scenes. Traditional methods often rely on complex mesh representations or resource-intensive ray marching, but the authors challenge this by demonstrating that high-quality geometry is not essential for photorealistic rendering. Their proposed solution, MixRT, combines a low-quality mesh, a view-dependent displacement map, and a compressed NeRF model. This approach effectively utilizes existing graphics hardware, enabling real-time NeRF rendering on edge devices. The MixRT framework, optimized for WebGL, achieves over 30 frames per second at 1280x720 resolution on a MacBook M1 Pro laptop, offering better rendering quality and smaller storage size compared to state-of-the-art methods. This advancement in NeRF rendering technology has significant implications for immersive interactions on edge devices, providing a balance between rendering speed, quality, and resource efficiency.
     
     \item \textbf{ZS-SRT (An Efficient Zero-Shot Super-Resolution Training Method for Neural Radiance Fields)} by Xiang Feng et al. introduces a novel framework for enhancing the efficiency of reconstructing high-resolution novel views using Neural Radiance Fields (NeRF) without requiring high-resolution training data. The authors propose a two-stage zero-shot super-resolution training framework that leverages internal learning within a single scene. Initially, the framework learns a scene-specific degradation mapping on a pretrained low-resolution coarse NeRF. Subsequently, it optimizes a super-resolution fine NeRF through inverse rendering with the degradation mapping, enabling gradient backpropagation from low-resolution 2D space to super-resolution 3D sampling space. Additionally, the authors incorporate a temporal ensemble strategy during inference to address scene estimation errors. This method is significant as it avoids the need for high-resolution inputs or additional scene data during training and ensures cross-view consistency, potentially benefiting applications in virtual reality, augmented reality, and telepresence.
     
     \item \textbf{Reducing Shape-Radiance Ambiguity in Radiance Fields with a Closed-Form Color Estimation Method} by Qihang et al, addresses a significant challenge in the Neural Radiance Field (NeRF) domain. NeRF, a technique in computer graphics and vision, synthesizes realistic novel view images of a 3D scene from a collection of posed 2D images. However, it suffers from the shape-radiance ambiguity problem, where it can recover training views accurately but may produce incorrect shapes, leading to poor novel views. The authors propose a novel regularization method to disentangle the density and color fields in NeRF. Their key innovation is a closed-form color estimation method that recovers the color field of a scene given a density field and a set of posed training images. This approach addresses the challenges of occlusion and non-uniformly distributed views in estimating the color fields. The method then uses the photometric loss derived from the estimated color field to provide independent supervision for the density field, improving the performance of explicit NeRFs like Plenoxels and DVGO.
     
     In practical applications, the authors demonstrate that their method enhances the quality of rendered images and corrects geometric errors more effectively than existing volume and ray-based losses. They also discuss the computational efficiency of their approach, noting that it significantly reduces computation overhead. Despite its effectiveness, the method has limitations in recovering highly reflective objects perfectly, as illustrated with the example of a reflective surface on scissors. The paper concludes with acknowledgments and references, highlighting the support from various research grants and foundations. The authors suggest that future work could focus on systematically solving the challenge with highly reflective objects and potentially eliminating the need for a parameterized color field in NeRF, thereby only requiring training and storage for the density field.

     \item \textbf{SpecNeRF (Gaussian Directional Encoding for Specular Reflections)}, a novel approach to modeling specular reflections in neural radiance fields (NeRFs), was introduced by Li Ma, Vasu Agrawal, Haithem Turki, Changil Kim, Chen Gao, Pedro Sander, Michael Zollhöfer, and Christian Richardt. The authors identified a gap in existing NeRF methodologies, which struggled with accurately rendering glossy surfaces under complex indoor lighting. To address this, they proposed a learnable Gaussian directional encoding that captures the spatially-varying nature of near-field lighting, enabling efficient evaluation of preconvolved specular color at any 3D location with varying roughness coefficients. This method significantly improved the modeling of challenging specular reflections in NeRFs, leading to more physically meaningful component decomposition.
     
     The authors also introduced a data-driven geometry prior to alleviate the shape-radiance ambiguity in reflection modeling. This involved deploying a monocular normal estimation network to supervise the normal of the geometry at the beginning of the training stage, enhancing the reconstruction of normals and specular reflections. The effectiveness of SpecNeRF was demonstrated through experiments on several public datasets, where it outperformed existing methods in photorealistic rendering of reflective scenes and provided more accurate color component decomposition. This technique has broad applications, particularly in enhancing the photorealism of reconstructed NeRF scenes with glossy surfaces, and can be applied in various real-world scenarios where accurate modeling of specular reflections is crucial.

     \item \textbf{Deep Learning on 3D Neural Fields} by Pierluigi Zama Ramirez et al. introduces 'nf2vec', a novel framework for embedding 3D objects represented by Neural Fields (NFs) into compact, meaningful latent codes. This approach addresses the challenge of integrating NFs, which are essentially neural networks, into deep learning pipelines for various downstream tasks. The authors demonstrate nf2vec's effectiveness in embedding 3D objects and its application in tasks like classification, retrieval, part segmentation, unconditioned generation, completion, and surface reconstruction. They also show that nf2vec can learn a smooth latent space, enabling interpolation of NFs representing unseen 3D objects.
     
     The paper explores the use of nf2vec in various scenarios, including the processing of NFs capturing both the geometry and appearance of 3D objects, like neural radiance fields (NeRFs). It demonstrates that nf2vec embeddings can be used as input for deep learning tasks, and also as output of generative frameworks, highlighting the versatility of nf2vec. The authors also investigate the potential of learning a mapping between two distinct nf2vec latent spaces, showing that this approach can effectively handle tasks like point cloud completion and surface reconstruction.
     
     In terms of real-world applications, nf2vec presents a significant advancement in handling and processing 3D data. Its ability to embed complex 3D shapes into compact representations and the ease of integrating these embeddings into existing deep learning frameworks make it a promising tool for various applications in computer vision, graphics, and beyond. The framework's efficiency in processing NFs and its adaptability to different 3D data representations suggest its potential in areas like digital twin technology, 3D modeling, and virtual reality.

     \item \textbf{Compact 3D Scene Representation via Self-Organizing Gaussian Grids} by Wieland Morgenstern et al. introduces a novel approach for modeling static 3D scenes using 3D Gaussian Splatting (3DGS). This technique, in contrast to Neural Radiance Fields, employs efficient rasterization for fast, high-quality rendering, but faced challenges with large storage sizes. The authors addressed this by organizing 3DGS parameters into a structured 2D grid, significantly reducing storage requirements without compromising visual quality. This was achieved through a novel algorithm that arranges high-dimensional Gaussian parameters into a 2D grid while preserving their neighborhood structure and enforcing local smoothness during training. The method integrates smoothly with established renderers and achieves a substantial reduction in storage size (8x to 26x) for complex scenes without increasing training time. This advancement marks a significant leap in 3D scene distribution and consumption, particularly for applications on resource-constrained devices or fast web applications. The paper demonstrates the method's effectiveness through various experiments and comparisons with existing techniques, highlighting its potential for real-world applications in efficient 3D scene rendering and storage.
     
     \item \textbf{ShowRoom3D (Text to High-Quality 3D Room Generation Using 3D Priors)} is a groundbreaking method for creating 3D room-scale scenes from text descriptions. The technique, developed by the authors, employs a three-stage process using a 3D diffusion prior, MVDiffusion, to optimize Neural Radiance Fields (NeRF). Initially, the camera is centered to form the basic room structure, followed by a novel pose transformation in the second stage for varied camera positions, enhancing room geometry and viewpoint diversity. The final stage refines the model for rendering from any position and rotation. This method surpasses existing approaches in structural integrity, clarity, and perspective consistency, marking a significant advancement in VR/AR and Metaverse applications, despite challenges like time consumption and oversaturation in results.

     \item \textbf{NeRF-VO (Real-Time Sparse Visual Odometry with Neural Radiance Fields)}, a novel monocular visual odometry system, was introduced by Jens Naumann, Binbin Xu, Stefan Leutenegger, and Xingxing Zuo. This system integrates learning-based sparse visual odometry for low-latency camera tracking with a neural radiance scene representation for advanced dense reconstruction and novel view synthesis. The authors employed a transformer-based neural network for inferring dense geometry cues, including monocular dense depth and normals. They optimized a neural radiance field that implicitly represents the 3D scene by minimizing the disparity between captured images and predicted dense geometric cues relative to the renderings generated from the neural radiance field. The system architecture comprised three main components: a sparse visual tracking front-end, a dense geometry enhancement module, and a NeRF-based dense mapping back-end. NeRF-VO demonstrated state-of-the-art performance in pose estimation accuracy, novel view synthesis fidelity, and dense reconstruction quality across various synthetic and real-world datasets. It achieved this while maintaining a high camera tracking frequency and consuming less GPU memory. The authors' work presents a significant advancement in the field of 3D computer vision, offering promising applications in robotics and mixed reality.

     \item \textbf{DyBluRF (Dynamic Deblurring Neural Radiance Fields for Blurry Monocular Video)} introduces a novel framework for enhancing video view synthesis by addressing the challenge of motion blur in monocular videos. The authors, Minh-Quan Viet Bui, Jongmin Park, Jihyong Oh, and Munchurl Kim, propose DyBluRF, a dynamic deblurring Neural Radiance Field (NeRF) framework, which significantly outperforms previous state-of-the-art methods in synthesizing sharp spatio-temporal views from blurry videos. DyBluRF comprises two main stages: the Interleave Ray Refinement (IRR) stage and the Motion Decomposition-based Deblurring (MDD) stage. The IRR stage focuses on reconstructing dynamic 3D scenes and refining inaccurate camera poses, while the MDD stage introduces an innovative incremental latent sharp-rays prediction (ILSP) approach for handling blurriness due to camera and object motion. This framework is particularly effective in scenarios where camera poses extracted from blurry videos are imprecise, a common issue in real-world applications. The authors demonstrate DyBluRF's superior performance through extensive experiments, showcasing its potential in various applications like immersive video experiences and 3D scene reconstruction from monocular video sources.

     \item \textbf{Gaussian Splatting with NeRF-based Color and Opacity} by Dawid Malarz et al. introduces a novel hybrid model that combines Gaussian Splatting (GS) and Neural Radiance Fields (NeRF) for rendering 3D objects. This approach, known as Viewing Direction Gaussian Splatting (VDGS), leverages the strengths of both GS and NeRF, offering faster training and inference times than NeRF alone, while maintaining high-quality renders. GS, which does not require neural networks, encodes 3D objects in Gaussian distributions, allowing for rapid development in dynamic scene modeling. However, it struggles with conditioning due to the need for a large number of Gaussian components. NeRF, on the other hand, excels at generating sharp renders from new viewpoints but is computationally expensive. The authors' hybrid model uses GS to represent the shape of 3D objects and a NeRF-based neural network to encode color and opacity changes based on viewing direction. This method results in improved rendering of shadows, light reflections, and transparency in 3D objects. The paper demonstrates the effectiveness of VDGS through quantitative evaluations and comparisons with existing methods, highlighting its potential applications in areas requiring efficient and high-quality 3D rendering.
     
     \item \textbf{Neural Point Cloud Diffusion for Disentangled 3D Shape and Appearance Generation} presents a novel method for generating 3D assets, crucial in fields like movie content creation, game design, and AR/VR applications. The authors introduced a hybrid approach combining a neural point cloud with a neural radiance field, enabling the disentanglement of coarse object shape from local appearance. This method allows for independent control and sampling of shape and appearance, a significant advancement over existing models that lacked this capability. The authors trained a generalizable Point-NeRF renderer across various instances, using the resulting neural point clouds to train a diffusion model. This model operates on high-dimensional latent spaces and is capable of denoising point positions and features simultaneously. The paper demonstrates that this approach significantly outperforms previous methods in generation quality, achieving reduced FID scores by 30-90\% and matching the performance of state-of-the-art methods that do not offer disentanglement capabilities. The method's application in generating 3D shapes and appearances separately has profound implications for enhancing control and flexibility in 3D content creation across various industries.

     \item \textbf{PlatoNeRF (3D Reconstruction in Plato's Cave via Single-View Two-Bounce Lidar)} is useful for reconstructing 3D scene geometry from a single view using two-bounce signals captured by a single-photon lidar. The authors, affiliated with the Massachusetts Institute of Technology and Meta, address the challenge of single-view 3D reconstruction, which is critical for applications like autonomous vehicles and extended reality. Traditional methods either rely on data priors, which may not be physically accurate, or struggle with detecting shadows in ambient light or low albedo backgrounds. PlatoNeRF overcomes these limitations by modeling two-bounce optical paths with NeRF (Neural Radiance Fields), using lidar transient data for supervision. This approach allows for the reconstruction of both visible and occluded geometry without data priors or reliance on controlled ambient lighting or scene albedo. The method consists of rendering primary rays from the camera to the scene and modeling rays that scatter to the virtual light, supervised with transients measured by a single-photon lidar. The authors demonstrate that their method can accurately reconstruct scenes from a single view without hallucinating scene details and is robust to ambient light, scene albedo, and spatial and temporal resolution. The paper also includes implementation details and experiments validating the method on simulated datasets, showcasing its potential for real-world applications in various fields requiring accurate 3D reconstructions from limited viewpoints.
 
     \item \textbf{Density Uncertainty Quantification with NeRF-Ensembles: Impact of Data and Scene Constraints} by Miriam Jäger, Steven Landgraf, and Boris Jutzi, explores the application of Neural Radiance Fields (NeRFs) in 3D scene reconstructions. The authors propose the use of NeRF-Ensembles to provide a density uncertainty estimate alongside the mean density. They demonstrate that data constraints, such as low-quality images and poses, lead to increased density uncertainty and decreased predicted density. The paper also investigates the impact of scene constraints like acquisition constellations, occlusions, and material properties on density uncertainty. The authors highlight the advantages of NeRF-Ensembles in enhancing robustness and removing artifacts. They conduct their methodology on synthetic and real datasets, including real data under realistic recording conditions. The paper also reviews related work in NeRFs, uncertainty estimation in deep learning, and uncertainty estimation in neural radiance fields, providing a comprehensive background for their research. The methodology involves training an ensemble of several NeRFs on the same input data to build a NeRF-Ensemble, which then provides an average network output with mean density and a corresponding quantification of the density uncertainty in 3D space. This approach is shown to be effective in improving the quality of 3D scene reconstructions and in the removal of artifacts in post-processing.

     \item \textbf{PoseGen (Learning to Generate 3D Human Pose Dataset with NeRF)}, a novel framework introduced by Mohsen Gholami, Rabab Ward, and Z. Jane Wang, leverages Neural Radiance Fields (NeRF) to generate 3D human pose datasets. This approach addresses the limitations of public datasets, which often lack diversity in human poses and camera viewpoints. The authors proposed a method where PoseGen learns to generate data (human 3D poses and images) with feedback from a pre-trained pose estimator. This data is optimized to enhance the robustness of the pre-trained model, focusing on out-of-distribution (OOD) samples to improve model generalizability. The paper details how PoseGen uses a generator to output 3D human poses and camera viewpoints, which are then rendered into images using NeRF. These images are used to estimate the generated 3D poses, with the estimation error providing feedback to the generator. This setup allows for the generation of both in-distribution and OOD data. The authors experimented with different latent space distributions for the pose generator, concluding that a normal distribution is more effective for their objectives. The PoseGen framework was tested extensively, showing significant improvements in two baseline models (SPIN and HybrIK) across four datasets. The authors demonstrated PoseGen's effectiveness in generating human synthetic datasets and its application in enhancing the accuracy of pre-trained models, particularly in medical applications where accurate pose estimation is crucial. This research opens new avenues for convenient, user-specific data generation without the need for 3D human scans, marking a significant advancement in the field of 3D human pose estimation.
    
     \item \textbf{Deformable 3D Gaussian Splatting for Animatable Human Avatars} talks about ParDy-Human, a novel method introduced by Jung et al., revolutionizes the creation of animatable human avatars using a technique called 3D Gaussian Splatting. This approach, distinct from traditional implicit methods, requires significantly fewer camera views and human poses for training, and does not rely on ground truth masks. The authors developed a fully explicit model that can generate dynamic human avatars with realistic details. The core of ParDy-Human involves parameter-driven dynamics integrated into 3D Gaussian Splatting, where 3D Gaussians are deformed by a human pose model to animate the avatar. This method consists of two main modules: one that deforms canonical 3D Gaussians according to SMPL vertices, and another that predicts per Gaussian deformations to handle dynamics beyond SMPL vertex deformations. The images are synthesized through a rasterizer. This technique, which allows for mask-free training and efficient full-resolution image synthesis on consumer hardware, has shown superior performance in rendering avatars in novel poses, as demonstrated on the ZJU-MoCap and THUman4.0 datasets. ParDy-Human's application extends to animation, virtual reality, and gaming, offering a more efficient and realistic approach to human avatar generation.

     \item \textbf{Efficient Deformable Tissue Reconstruction via Orthogonal Neural Plane} by Chen Yang, Kailing Wang, Yuehao Wang, Qi Dou, Xiaokang Yang, and Wei Shen, introduces a novel framework called Forplane, designed to efficiently reconstruct deformable tissues in surgical procedures using neural radiance fields (NeRF). The authors conceptualized surgical procedures as 4D volumes, breaking them down into static and dynamic fields composed of orthogonal neural planes. This factorization led to decreased memory usage and faster optimization. They introduced a spatiotemporal importance sampling scheme to improve performance in regions with tool occlusion and large motions, accelerating training. An efficient ray marching method was applied to skip sampling among empty regions, significantly improving inference speed. Forplane accommodates both binocular and monocular endoscopy videos, demonstrating extensive applicability and flexibility. The experiments on two in vivo datasets, the EndoNeRF and Hamlyn datasets, showed that Forplane substantially accelerates both the optimization process (by over 100 times) and the inference process (by over 15 times) while maintaining or improving quality across various non-rigid deformations. This significant performance improvement promises to be a valuable asset for future intraoperative surgical applications. The authors' approach addresses the challenges in high-quality real-time reconstruction of deformable tissues in both monocular and binocular endoscopy videos, providing valuable insights for future intraoperative applications.

     \item \textbf{Neural BSSRDF (Object Appearance Representation Including Heterogeneous Subsurface Scattering)} by Thomson TG et al. introduces a groundbreaking neural method for rendering heterogeneous translucent objects. This method, diverging from traditional BSSRDF models, incorporates full object geometry and heterogeneities, akin to a neural radiance field adaptable to various lighting environments. Utilizing a multi-layer perceptron with skip connections, the authors effectively represent an object's appearance based on spatial position, observation direction, and incidence direction. This approach excels in storing complex materials and accurately rendering them in previously unseen lighting conditions. The paper's significance lies in its potential applications in realistic rendering for computer graphics, offering enhanced capabilities for depicting intricate materials and lighting scenarios in industries like movie production, gaming, and virtual reality.

     \item \textbf{SERF (Fine-Grained Interactive 3D Segmentation and Editing with Radiance Fields)} by Kaichen et al, addresses the challenges in fine-grained 3D-based interactive editing. The authors introduced SERF, a novel algorithm that integrates multi-view 3D reconstruction algorithms with pre-trained 2D models to create a neural mesh representation. This representation is used for accurate and interactive 3D segmentation without 3D supervision and supports a range of 3D editing operations, including geometry editing and texture painting. The method involves a two-step process: first, constructing the mesh using multi-view approaches and projecting color and geometry features onto it, and second, employing a novel surface rendering technique that leverages local information for robustness against deformation. The authors also presented a new 3D segmentation method that uses interactive 2D instructions and a pre-trained 2D segmentation model, enhancing the intuitiveness of the editing process. Their approach demonstrated superior results in representation quality and editing ability on both real and synthetic data. The paper also delves into related works in novel view synthesis, NeRF segmentation, and 3D editing, highlighting the advancements and limitations in these areas. The methodology section details the process of constructing the neural mesh, surface rendering techniques, and the 3D mesh segmentation approach, emphasizing the efficiency and accuracy of SERF in fine-grained interactive 3D editing.
     
     \item \textbf{Pano-NeRF (Synthesizing High Dynamic Range Novel Views with Geometry from Sparse Low Dynamic Range Panoramic Images)}Pano-NeRF, a novel approach presented by Zhan Lu et al, addressed the challenges in synthesizing High Dynamic Range (HDR) novel views from sparse Low Dynamic Range (LDR) panoramic images. The authors observed that each pixel in panoramic images could act as both a signal for scene lighting information and a light source for illuminating other pixels. This led to the development of irradiance fields from sparse LDR panoramic images, enhancing observation counts for accurate geometry recovery and leveraging irradiance-radiance attenuation for HDR reconstruction. Their method outperformed state-of-the-art techniques in both geometry recovery and HDR reconstruction, also offering a byproduct of spatially-varying lighting estimation. The paper detailed the formulation of irradiance fields, focusing on geometry recovery with sparse inputs and HDR reconstruction with LDR inputs. They integrated irradiance fields into radiance fields for joint optimization, modifying the MLP in a NeRF-based method to output an additional variable for albedo estimation. The experiments conducted on synthetic and real captured data demonstrated Pano-NeRF's superiority in geometry recovery and novel view synthesis, especially in sparse input scenarios. This technique holds significant potential for applications in extended reality (XR), offering immersive experiences like virtual walks in 360° scenes and inserting virtual objects with accurate lighting information.

     \item \textbf{2D-Guided 3D Gaussian Segmentation} by Kun Lan et al, introduces a novel method for segmenting 3D Gaussian representations. The authors, affiliated with the University of Science and Technology of China and the AI Thrust at HKUST(GZ), address the limitations of existing 3D Gaussian segmentation techniques, which are often cumbersome and inefficient for segmenting multiple objects simultaneously. Their approach leverages 2D segmentation maps as supervision, guiding the learning of added 3D Gaussian semantic information. This method involves nearest neighbor clustering and statistical filtering to refine segmentation results, demonstrating its effectiveness in experiments with object-centric and 360° scenes. The technique shows promise in applications of 3D understanding and editing, offering a more efficient and accurate alternative to previous methods. The paper's contributions include the development of an efficient 3D Gaussian segmentation method supervised by 2D segmentation, capable of learning the semantic information of a 3D scene quickly and segmenting multiple objects in a short time. The authors' experiments on various datasets demonstrate the method's effectiveness, achieving significant improvements in metrics like mIOU.

     \item \textbf{INFAMOUS-NeRF (ImproviNg FAce MOdeling Using Semantically-Aligned Hypernetworks with Neural Radiance Fields)} by Andrew Hou et al, presents a groundbreaking approach in 3D face modeling. This method ingeniously integrates hypernetworks with Neural Radiance Fields (NeRF) to significantly enhance representation power for multiple subjects, a notable challenge in traditional NeRF applications. A key innovation is the learning of semantically-aligned latent spaces, enabling coherent and meaningful face editing across different subjects. The authors further refine the technique with a photometric surface constraint for improved boundary rendering and an adaptive sampling algorithm to optimize training efficiency. Demonstrating superior performance in single-image novel view synthesis and 3DMM fitting across various datasets, INFAMOUS-NeRF marks a significant leap forward in applications such as augmented reality, face recognition, and digital entertainment, offering both high-quality representation and editable 3D face models.

     \item \textbf{SUNDIAL (3D Satellite Understanding through Direct, Ambient, and Complex Lighting Decomposition)}, a novel approach for 3D satellite scene reconstruction using neural radiance fields, was introduced by Nikhil Behari et al from Harvard University and the Massachusetts Institute of Technology. The paper, titled addresses the challenges of traditional 3D modeling in remote sensing, such as limited multi-view baselines, varying illumination conditions, and transient scene changes. The authors developed a comprehensive method that jointly learns satellite scene geometry, illumination components, and sun direction in a single-model approach. They proposed a secondary shadow ray casting technique to improve scene geometry, enable physically-based disentanglement of scene albedo and illumination, and determine the components of illumination from direct, ambient, and complex sources. The technique incorporates lighting cues and geometric priors from remote sensing literature, modeling physical properties like shadows and scattered sky illumination. The authors evaluated SUNDIAL against existing NeRF-based techniques, demonstrating improved scene and lighting disentanglement, novel view and lighting rendering, and geometry and sun direction estimation in challenging scenes. This work has significant implications for fields like environmental science, urban planning, agriculture, and disaster response, where accurate 3D modeling from satellite imagery is crucial.
     
     \item \textbf{DL3DV-10K (A Large-Scale Scene Dataset for Deep Learning-based 3D Vision)}, a large-scale scene dataset for deep learning-based 3D vision, was introduced by Lu Ling and colleagues. This dataset, containing 10,510 videos at 4K resolution across 65 types of locations, aims to address the limitations of existing datasets in deep learning-based 3D analysis. It provides a comprehensive benchmark for novel view synthesis (NVS) and supports learning-based 3D representation techniques. The authors highlighted the progress in deep learning-based 3D vision, particularly Neural Radiance Field (NeRF) based 3D representation learning and its applications in NVS. However, they identified a critical gap in existing datasets, which are either synthetic or cover a narrow selection of real-world scenes, limiting the comprehensive benchmarking of NVS methods and the generalizability of deep 3D representation learning methods.
     
     To fill this gap, DL3DV-10K was created, featuring a diverse range of real-world scenarios, including both bounded and unbounded scenes with varying levels of reflection, transparency, and lighting. This dataset enables a more robust evaluation of NVS methods in complex real-world scenarios. The authors conducted a comprehensive benchmark of recent NVS methods on DL3DV-10K, revealing valuable insights for future research. They also demonstrated the potential of DL3DV-10K in enhancing the generalizability of NeRF, confirming the importance of diversity and scale in learning a universal scene prior. The dataset's design and collection process, including guidelines for video capture and quality assurance, were meticulously detailed, ensuring the representation of a wide range of real-world complexities. This dataset is poised to significantly impact the field of deep learning-based 3D vision, offering a robust platform for developing and testing new methods in NVS and 3D representation learning.

     \item \textbf{KeyNeRF (Informative Rays Selection for Few-Shot Neural Radiance Fields)}, a novel approach for training NeRF in few-shot scenarios, was introduced by Marco Orsingher et al. The paper, titled addresses the challenge of lengthy per-scene optimization in NeRF, which limits its practical usage, especially in resource-constrained settings. The authors proposed a method that focuses on key informative rays, selected at both the camera and pixel levels. At the camera level, a view selection algorithm promotes baseline diversity while ensuring scene coverage. At the pixel level, sampling is based on a probability distribution derived from local image entropy. This approach, requiring minimal changes to existing NeRF codebases, outperforms state-of-the-art methods in efficiency and simplicity.
     
     The authors also discussed related work, highlighting the limitations of existing few-shot NeRF methods that rely on complex loss functions or additional inputs. They emphasized the importance of view selection and rays sampling, noting that these aspects are often overlooked in typical use cases like object scanning from videos. The proposed method, KeyNeRF, improves upon these aspects by introducing a view selection algorithm that ensures scene coverage and promotes baseline diversity, and an entropy-based rays sampling technique that focuses on high-frequency details for faster convergence.
     
     In their experiments, the authors demonstrated KeyNeRF's superior performance on standard benchmarks compared to existing methods, without requiring additional inputs or complex loss functions. They also conducted ablation studies to analyze the impact of the number of poses and training iterations on image quality metrics, further validating the effectiveness of their view selection and rays sampling techniques. Overall, KeyNeRF presents a significant advancement in the field of image-based 3D reconstruction, with potential applications in robotics, virtual reality, and autonomous driving.

     \item \textbf{Inpaint4DNeRF (Promptable Spatio-Temporal NeRF Inpainting with Generative Diffusion Models)} by Han Jiang et al introduces a novel approach for editing 3D and 4D scenes using Neural Radiance Fields (NeRF) and generative diffusion models. The authors developed a method that leverages the capabilities of state-of-the-art stable diffusion models, such as ControlNet, for direct generation of completed background content in both static and dynamic scenes. This technique involves generating a subset of completed images, termed 'seed images', from which simple 3D geometry proxies are derived. These seed images guide the generation of other views, ensuring multiview consistency among all completed images. The method, Inpaint4DNeRF, is designed to handle the generative inpainting of NeRFs, allowing for the removal of static foreground objects specified by a user-supplied prompt while maintaining multiview consistency. The authors' approach is distinguished by its ability to manipulate specific objects within a given background NeRF, ensuring consistency with the unmasked background and partially masked foreground objects. The paper also discusses related work in NeRF editing, inpainting techniques, and text-guided visual content generation, highlighting the novelty and potential applications of their method in fields like digital art creation and virtual/augmented reality. The proposed framework consists of three main stages: training view pre-processing, progressive training, and a 4D extension for dynamic cases. This method addresses the challenge of inpainting within 3D and 4D scenes while maintaining view consistency, even under perspective changes, and is promptable in the sense that the inpainted content matches the text description while being consistent with the underlying 3D or 4D spatiotemporal scene.

     \item \textbf{GD$^2$-NeRF (Generative Detail Compensation via GAN and Diffusion for One-shot Generalizable Neural Radiance Fields)} by Xiao Pan et al, introduces a novel framework for enhancing one-shot novel view synthesis (O-NVS) using Neural Radiance Fields (NeRF). The authors identified that existing One-shot Generalizable NeRF (OG-NeRF) methods, while efficient, often produce blurry results due to their reliance on limited reference images. To address this, they proposed GD2-NeRF, a framework that integrates Generative Adversarial Networks (GANs) and pre-trained diffusion models into OG-NeRF, following a coarse-to-fine strategy. In the coarse stage, the One-stage Parallel Pipeline (OPP) efficiently incorporates a GAN model into the OG-NeRF pipeline, using in-distribution detail priors from the training dataset to alleviate blurriness and balance sharpness and fidelity. The fine stage introduces the Diffusion-based 3D-consistent Enhancer (Diff3DE), which leverages pre-trained image diffusion models to add rich, plausible out-distribution details while maintaining 3D consistency. This approach significantly improves detail rendering in novel view synthesis without requiring per-scene finetuning. The authors' experiments on synthetic and real-world datasets demonstrated that GD2-NeRF markedly enhances detail rendering in novel views, achieving a good balance between sharpness and fidelity, and providing rich, plausible details with decent 3D consistency. This framework opens up new possibilities for realistic and detailed 3D reconstructions in various applications, ranging from virtual reality to 3D modeling, where high-quality novel view synthesis is crucial.
     
     \item \textbf{Sharp-NeRF (Grid-based Fast Deblurring Neural Radiance Fields Using Sharpness Prior)}, a novel technique introduced by Park etl al, addresses the challenge of rendering sharp images from blurry inputs using NeRF. The authors identified that traditional NeRF models struggle with visual quality degradation under imperfect conditions like defocus blurring, a common issue in images captured by cameras. To tackle this, Sharp-NeRF employs grid-based kernels to model the sharpness or blurriness of a scene, calculating the sharpness level of pixels to learn spatially varying blur kernels. This approach significantly accelerates training time, taking only half an hour to train, compared to previous methods that required extensive computational time due to the use of MLP. The paper also delves into the application of discrete learnable kernels for blurring convolution operations, a departure from the MLP-based methods prevalent in earlier studies. This strategy not only reduces computational costs but also enhances the ability to handle complex real-world scenarios. The authors introduced a sharpness prior to the framework, grouping pixels based on similar sharpness levels, which ensures consistent learning of spatially varying sharpness with minimal increase in learnable parameters. Additionally, they proposed random patch sampling during training to further optimize training time without compromising image quality. The effectiveness of Sharp-NeRF was demonstrated through experiments on real-world datasets, showing considerable improvements in training speed and visual quality compared to existing methods. The technique's potential applications span various fields, including visual effects, e-commerce, AR/VR, and robotics, where high-quality image synthesis from 3D models is crucial. The authors' contributions, particularly in developing grid-based kernels and incorporating sharpness priors, mark a significant advancement in the field of neural rendering and image deblurring.

     \item \textbf{Deblurring 3D Gaussian Splatting} by Byeonghyeon Lee et al, presents a novel framework for real-time deblurring in 3D scenes. The authors addressed the challenge of rendering high-quality images from blurry inputs, a common issue in practical scenarios like augmented/virtual reality and robotics. Traditional NeRF methods, while effective in synthesizing photorealistic images, struggle with blurriness and are computationally intensive. The authors proposed an innovative approach using 3D Gaussian Splatting (3D-GS), a method that combines numerous colored 3D Gaussians to represent scenes and employs a differentiable splatting-based rasterization for efficient rendering. This technique significantly outperforms NeRF in rendering speed, achieving over 200 frames per second. To tackle the issue of blurriness, particularly defocus blur, the authors introduced a small MLP that adjusts the covariance of each 3D Gaussian, simulating the intermingling of neighboring pixels during training. This process allows for the reconstruction of sharp details from blurry images. Additionally, they developed a method to compensate for the sparsity in point clouds generated from blurry images, enhancing scene reconstruction. The paper's experiments demonstrated the effectiveness of this approach, achieving state-of-the-art rendering quality at a remarkably faster speed compared to existing models. This work represents a significant advancement in real-time 3D scene rendering and deblurring, offering practical applications in various fields requiring rapid and high-quality image processing.

     \item \textbf{PlanarNeRF (Online Learning of Planar Primitives with Neural Radiance Fields)} by Zheng Chen et al introduced a novel framework, PlanarNeRF, for detecting dense 3D planar primitives using online learning. This framework uniquely integrates scene appearance and geometry, enhancing 3D plane detection. The authors developed a lightweight plane fitting module to estimate plane parameters and a global memory bank for consistent cross-frame correspondence. PlanarNeRF operates in two modes: supervised (PlanarNeRF-S) using sparse 2D plane annotations and self-supervised (PlanarNeRF-SS) without annotations. The framework's design allows for efficient learning from sparse training signals, significantly improving training efficiency. The authors demonstrated PlanarNeRF's effectiveness through extensive experiments, showing remarkable improvement over existing methods. This approach holds great potential for applications in Virtual Reality, Augmented Reality, robotic manipulation, and various data processing tasks like object detection, registration, pose estimation, and SLAM. PlanarNeRF's ability to capture planar primitives in structured environments like indoor rooms and urban buildings makes it a significant advancement in the field of computer vision.

     \item \textbf{3D Visibility-aware Generalizable Neural Radiance Fields for Interacting Hands} by Xuan Huang et al. introduces a novel framework, Visibility-aware Neural Radiance Field (V A-NeRF), specifically designed for rendering 3D images of interacting hands. The authors identified that existing Neural Radiance Fields (NeRFs) struggled with scenes involving severe inter-hand occlusions and view variations, particularly when limited to single-view inputs. To address these challenges, they developed V A-NeRF, which first constructs a mesh-based representation of hands and extracts geometric and textural features. A key innovation is the feature fusion module that adaptively merges features from both hands based on the visibility of query points and mesh vertices, thus enhancing the recovery of features in unseen areas. Additionally, the authors introduced an adversarial learning paradigm with a novel discriminator that generates pixel-wise visibility maps, offering fine-grained supervision for unseen areas and improving the visual quality of synthesized images. This approach significantly outperformed conventional NeRFs in experiments conducted on the Interhand2.6M dataset. The applications of this technique are vast, including hand pose estimation and improving the realism in virtual and augmented reality environments, particularly in scenarios involving complex hand interactions.
     
     \item \textbf{Noise-NeRF (Hide Information in Neural Radiance Fields using Trainable Noise)}, a novel method for embedding steganographic information in NeRF, was introduced by Qinglong Huang et al. Addressing the limitations of previous NeRF steganography methods, which often compromised model integrity and steganographic quality, Noise-NeRF innovatively utilizes trainable noise. This approach allows for embedding hidden information without altering the model's weights, thereby preserving the rendering quality of the NeRF model. The technique involves adding noise under specific viewpoints, causing the NeRF's neural network to produce intentional error outputs, rendering the hidden information. To enhance the steganography quality and efficiency, the authors proposed two strategies: Adaptive Pixel Selection and Pixel Perturbation. These strategies optimize the noise update process, ensuring high-quality steganographic content recovery while maintaining the original rendering quality of the NeRF model. The method demonstrated state-of-the-art performance in both steganography and rendering quality, proving effective in super-resolution image steganography. This advancement holds significant potential for applications in digital media, virtual reality, and augmented reality, where information confidentiality and data security are paramount.

     \item \textbf{SIGNeRF (Scene Integrated Generation for Neural Radiance Fields)}, a novel approach for NeRF scene editing and object generation, was introduced by Jan-Niklas Dihlmann et al in their paper. This method, distinct from existing generative 3D approaches, allows for fast and controllable editing of photorealistic scenes. The authors leveraged depth-conditioned diffusion models to generate 3D consistent views, introducing a multi-view reference sheet of modified images. This reference sheet is key to updating an image collection consistently, refining the original NeRF with newly generated images in one go. The technique provides fine control over spatial location and shape guidance, either through selected regions or external meshes. The paper also discusses related work in text-to-image and text-to-3D generation, highlighting the advancements in diffusion probabilistic models and the combination of image diffusion models with NeRFs. The authors note the challenges in editing NeRFs and the limitations of existing methods, which often require artistic proficiency and manual labor. SIGNeRF addresses these challenges by offering a more controlled and efficient approach to generative NeRF editing. The method involves a two-step process: generating a reference sheet and using it to iteratively update the images in the NeRF dataset. This process ensures multi-view consistency and integrates the edits seamlessly into the existing scene. The authors demonstrate the effectiveness of SIGNeRF in various applications, including object insertion, modification, and scene editing, showcasing its potential in creating photorealistic and consistent 3D scenes.
     \item \textbf{Progress and Prospects in 3D Generative AI: A Technical Overview including 3D human} by Song Bai and Jie Li provides a comprehensive survey of advancements in 3D generative AI, particularly focusing on 3D object, character, and motion generation. The authors highlight the significant role of stable diffusion techniques, multi-view consistency control methods, and realistic human models like SMPL-X in creating consistent and near-realistic 3D models. They delve into neural network-based 3D storing and rendering models, such as NeRF and 3DGS, which have revolutionized the efficiency and realism of neural rendered models. The paper also discusses the multimodality capabilities of large language models in converting language inputs into human motion outputs. In the realm of single 3D object generation, various AI techniques are explored, including point cloud diffusion models and wavelet domain-based diffusion models. The authors describe methods for generating 3D models from multi-angle images, categorizing them into iterative refinement for detailed models and neural networks for rapid generation. They address the challenges in generating unified scenes and maintaining consistency in 3D models, citing examples like ZeroNVS which uses Denoising Diffusion Implicit Models (DDIM) for better scene consistency. The paper also covers 3D human model generation, noting the complexity due to facial expressions and clothing textures. Techniques like SMPL and SMPL-X are widely used for human modeling, providing detailed and animate 3D human models. Iterative methods like DreamWaltz and HumanNorm are discussed for their ability to generate high-precision human models, while non-iterative methods like GTA and Chupa are noted for their rapid texture modeling capabilities. 
     \item \textbf{Hi-Map (Hierarchical Factorized Radiance Field for High-Fidelity Monocular Dense Mapping)} introduces Hi-Map, a novel approach for monocular dense mapping using NeRF without relying on external depth priors. The authors developed a method that simplifies and generalizes scene representation for efficient computation and fast convergence on new observations. This is achieved by representing the scene as a hierarchical feature grid, factorizing it into feature planes and vectors, and using the SDF as a proxy for rendering to infer volume density. They also introduced a dual-path encoding strategy to enhance mapping quality, particularly in distant and textureless regions. The paper demonstrates Hi-Map's superiority in geometric and textural accuracy over existing NeRF-based monocular mapping methods through extensive experiments. This technique has significant implications for embodied intelligent systems like robots, enabling them to perform scene-understanding tasks and navigate complex environments with high fidelity 3D maps, providing timely feedback for human interaction. The method's efficiency and fidelity in dense mapping using only posed RGB inputs mark a significant advancement in the field, with potential applications in robotics, autonomous navigation, and virtual reality.
     
     \item \textbf{RustNeRF (Robust Neural Radiance Field with Low-Quality Images)}, a novel framework presented by Mengfei Li et al, addresses the limitations of traditional NeRF in handling low-quality, degraded images. The authors identified that existing NeRF models falter when trained with degraded images, leading to artifacts and suboptimal results. To overcome this, RustNeRF introduces a 3D-aware preprocessing network that incorporates real-world degradation modeling, effectively enhancing image restoration while maintaining multi-view consistency. This approach allows for the restoration of high-fidelity images from degraded inputs, a significant advancement in the field of 3D scene modeling and novel-view synthesis. The authors also propose an implicit multi-view guidance method to excavate redundant information across multiple views, further refining the restoration process. Extensive experiments demonstrated RustNeRF's superiority over existing methods in dealing with real-world image degradation. This innovation opens up new possibilities for applying NeRF in practical scenarios where high-quality image datasets are not readily available, such as in custom real-world datasets with various types of degradation. RustNeRF's ability to robustly handle low-quality images marks a significant step forward in the practical application of NeRF technology.

     \item \textbf{A Survey on 3D Gaussian Splatting} by Guikun Chen and Wenguan Wang presents a comprehensive overview of 3D GS, a novel technique in explicit radiance field and computer graphics. The authors describe 3D GS as a significant evolution from NeRF methodologies, offering real-time rendering capabilities with enhanced control and editability. This method uses millions of 3D Gaussians for explicit scene representations, differing from NeRF's implicit, coordinate-based models. The authors elaborate on the practical applications of 3D GS, highlighting its potential in virtual reality, interactive media, and beyond. They provide a comparative analysis of leading 3D GS models, emphasizing their performance and utility in various applications. The paper also discusses the challenges and future research directions in this field, aiming to foster further exploration and advancement in explicit radiance field representation. 
     The authors delve into the technical aspects of 3D GS, explaining its advantages over traditional NeRF methods in terms of computational efficiency and control. They describe the process of novel-view synthesis with learned 3D Gaussians, emphasizing the method's ability to handle complex scenes and high-resolution outputs efficiently. The paper also covers the optimization process of 3D GS, detailing the steps involved in constructing and refining the 3D Gaussian representations for accurate scene capture. This includes discussions on parameter optimization, loss function, and density control of the Gaussians. The authors highlight the importance of balancing computational efficiency with image synthesis quality, showcasing 3D GS's potential to revolutionize scene representation and rendering in computer graphics.

     \item \textbf{CTNeRF (Cross-Time Transformer for Dynamic Neural Radiance Field from Monocular Video)} by Xingyu Miao et al. presents a novel approach to generate high-quality novel views from monocular videos of complex and dynamic scenes. The authors identified limitations in previous methods like DynamicNeRF, which struggled with accurately modeling the motion of complex objects, leading to blurry renderings. To overcome this, they introduced a module that operates in both time and frequency domains, aggregating features of object motion to learn relationships between frames and generate higher-quality images. Their experiments showed significant improvements over state-of-the-art methods in dynamic scene datasets.
     The authors' method involves two main components: one focusing on the static background and the other on the dynamic foreground, which are then blended to obtain the final novel view. They introduced a Ray-based cross-time transformer to address the potential loss of intricate details during feature aggregation and a Global Spatio-Temporal Filter to mitigate blurring. The approach also includes multi-view aggregation, where static and dynamic regions are reconstructed separately and then blended. The method's effectiveness is further enhanced by a ray-based cross-time aggregation module, which handles temporal variations in input feature vectors of adjacent frames using a cross-time converter. This approach, as demonstrated in their ablation studies, effectively enhances the quality of synthesized images. The technique has applications in fields like film production and virtual reality, offering a promising approach for scene reconstruction and novel view synthesis in dynamic scenes.

     \item \textbf{FPRF (Feed-Forward Photorealistic Style Transfer of Large-Scale 3D Neural Radiance Fields)} introduced a novel method for stylizing large-scale 3D scenes using arbitrary style reference images, while maintaining multi-view appearance consistency. The authors developed a style-decomposed 3D neural radiance field, leveraging AdaIN's feed-forward stylization machinery. This approach enabled efficient stylization of large-scale 3D scenes, overcoming the limitations of previous methods that required tedious per-style or per-scene optimization and were restricted to small-scale scenes. FPRF's key innovation was the introduction of a style dictionary module and style attention, which allowed for efficient retrieval of style matches for each local part of the 3D scene from a set of diverse style references. The method also preserved multi-view consistency by applying semantic matching and style transfer processes directly onto queried features in 3D space. The authors demonstrated FPRF's capability to achieve high-quality photorealistic stylization for large-scale scenes with diverse reference images. This technique has potential applications in enriching virtual 3D spaces for XR applications and realistically augmenting real-world autonomous driving datasets. The authors' approach represents a significant advancement in the field of photorealistic style transfer for large-scale 3D scenes.
     
     \item \textbf{Diffusion Priors for Dynamic View Synthesis from Monocular Videos} presents a novel approach to dynamic novel view synthesis, a technique crucial for capturing the temporal evolution of visual content in videos. The authors identified the challenge in existing methods, which struggle to distinguish between motion and structure, especially in scenarios with complex dynamic motions and limited camera poses. To address this, they introduced a pipeline that integrates knowledge from a pre-trained RGB-D diffusion model, fine-tuned on video frames using customization techniques. This pipeline involves representing a 4D scene with separate NeRFs for dynamic and static components, ensuring geometric consistency while preserving scene identity. The authors conducted extensive experiments on challenging datasets, demonstrating the robustness and utility of their approach in complex motion scenarios. Their method, DpDy, outperformed baseline methods in qualitative and quantitative evaluations, including user studies. The technique's application extends to various domains, offering significant potential for understanding and interacting with the real world through enhanced video content analysis and synthesis. This advancement in dynamic novel view synthesis marks a significant step in the field, showcasing the effective use of diffusion priors in resolving ambiguities between motion and structure and hallucinating unseen or occluded regions in dynamic scenes.

     \item \textbf{GO-NeRF (Generating Virtual Objects in Neural Radiance Fields)}, a novel method introduced by Peng Dai, Feitong Tan, Xin Yu, Yinda Zhang, and Xiaojuan Qi, addresses the challenge of generating high-quality 3D objects within existing 3D scenes represented as NeRF. The authors developed a compositional rendering formulation and context-aware learning objectives to seamlessly integrate generated 3D objects into existing scenes without unintended modifications. Their user-friendly interface allows users to specify object locations in a 3D scene using a simple point-and-click method. The method employs a separate NeRF for the object, optimized using a variety of loss functions, including Inpainting SDS loss, geometry loss, saturation loss, and style loss, to ensure the object's harmonious integration into the scene. The authors demonstrated the effectiveness of GO-NeRF through extensive experiments on various datasets, showcasing its superiority in generating context-compatible 3D virtual objects and seamlessly compositing them into pre-trained neural radiance fields. This technique holds significant potential for applications in virtual scene creation and editing, offering a highly immersive experience in various real-world scenarios.
     
     \item \textbf{TRIPS (Trilinear Point Splatting for Real-Time Radiance Field Rendering)}, presented by Linus Franke et al, introduces a novel approach to radiance field rendering. The authors aimed to overcome the limitations of existing methods like 3D Gaussian Splatting and ADOP, which either suffered from blurriness and cloudy artifacts or struggled with temporal instability and large gaps in point clouds. TRIPS combines the strengths of both by rasterizing points into a screen-space image pyramid, allowing for rendering large points using a single trilinear write. A lightweight neural network is then employed for reconstructing a hole-free image, maintaining high detail levels. This method is entirely differentiable, enabling automatic optimization of point sizes and positions. The authors demonstrated that TRIPS surpasses current state-of-the-art methods in rendering quality while maintaining real-time frame rates on standard hardware. The technique is particularly effective in rendering intricate geometries and expansive landscapes. Additionally, the paper delves into the optimization strategy, implementation details, and evaluations against other methods, showcasing TRIPS' superiority in various scenarios. This approach holds significant potential for applications in computer graphics and vision, particularly in scenarios demanding high-quality, real-time rendering of complex scenes.

     \item \textbf{Fast High Dynamic Range Radiance Fields for Dynamic Scenes} by Guanjun Wu et al. presents HDR-HexPlane, a novel framework for synthesizing HDR images in dynamic scenes. The authors identified a gap in existing NeRF methods, which primarily handle LDR images and struggle with nonuniform illumination in dynamic scenes. To address this, they developed HDR-HexPlane, which extends HDR techniques to dynamic scenes by learning 3D scenes from 2D images captured at various exposures. This method employs a learnable exposure mapping function for adaptive exposure values and a camera response function for stable learning. The authors also constructed a dataset containing multiple dynamic scenes with diverse exposures for evaluation. The paper's primary contribution lies in its end-to-end NeRF framework for dynamic scene representation in HDR, capable of efficiently learning scene exposures and rendering balanced images considering over/under-exposed regions. This technique has significant implications for real-world applications, particularly in scenarios with strong light sources or high contrast, such as indoor lighting, outdoor sunlight, and scenes with flames or reflections. The HDR-HexPlane framework effectively addresses the challenges of color inconsistency in multi-view dynamic scenes and offers a robust solution for HDR image synthesis in dynamic environments.
     
     \item \textbf{}

     \item \textbf{}
     
\end{itemize}

\end{itemize}
(Work in Progress...)

\section{Supporting Datasets}
\label{present:datasets}
This section discusses the datasets that have been generated or directly used for different types of tasks and evaluation purposes for a comprehensive review of Radiance Fields and similar research ideas that are discussed later in this survey. These comprise of the following.
\begin{itemize}
\item \textbf{LLFF (Light Field for the Language of Light and Form) dataset} provides the real-world scenes at 4K ultra-high resolution of training views~\cite{mildenhall2019local}. This dataset comprises of 8 forward-facing scenes with each scene having views between 20 views and 60 views. These scenes were ($4032 \times 3024$) while NeRF-based methods use scenes with resolution ($1008 \times 756$) for training the MLPs and rendering (inference) from them. Further, NeRF-based methods also use COLMAP~\cite{schoenberger2016sfm,schoenberger2016mvs} for estimating the corresponding camera poses.

This dataset has been used for the following research articles:
\begin{itemize}
    \item Original NeRF (Neural Radiance Fields)~\cite{mildenhall2020nerf}
    \item NeRF-W (Neural Radiance Fields in the Wild)~\cite{martin2021nerf}
\end{itemize}

\item \textbf{Tanks \& Temples dataset} was introduced by Knapitsch et al~\cite{knapitsch2017tanks} in 2017. It is widely used benchmark for evaluating 3D reconstruction algorithms. The dataset consists of a diverse set of High-Resolution RGB-D (color and depth) scans of light-scale indoor and outdoor scenes, captured using a high-quality structured light scanner. The scenes in the dataset are categorized based on ``Tanks'' (for indoor scenes), and ``Temples'' (for outdoor scenes). The dataset provides ground truth 3D models of the scenes, which are reconstructed from the RGB-D scans using a state-of-the-art 3D reconstruction pipeline. These ground truth models can be used to quantitatively evaluate the performance of 3D reconstruction algorithms by comparing their output against the reference models. It also comprises of evaluation metrics, including ``mean per-point distance'' between the reconstructed and reference models, and ``completeness'', which measures the percentage of the reference model that is covered by the reconstructed model. This dataset is used in research to evaluate and compare performance of various 3D reconstruction algorithms, including MVS, Volumetric Fusion, and Learning-based approaches such as NeRF and its variants. By providing a challenging and diverse set of scenes with ground truth data, the Tanks and Temples dataset enables the development and assessment of new 3D reconstruction methods in a standardized and reproducible manner.

This dataset has been used for the following research articles:
\begin{itemize}
    \item IDR (Implicit Differentiable Renderer)~\cite{yariv2020multiview}
    \item MVSNeRF (Multi-View Stereo Neural Radiance Fields)~\cite{chen2021mvsnerf}
\end{itemize}

\item \textbf{ScanNet} or \textbf{`Studio'} dataset~\cite{dai2017scannet}is a large-scale dataset for 2.5D and 3D understanding tasks, introduced in 2017. The dataset contains RGB-D scans of more than 1500 indoor scenes, captured using a consumer-grade depth sensor (Microsoft Kinect). The scenes include various residential and office environments, covering a wide range of object categories and room layouts. It provides rich annotations for each scan, including instance-level semantic segmentation, object bounding boxes, and object-level 3D CAD model alignments. Additionally, camera poses and camera calibration parameters are provided, enabling the reconstruction of the 3D scene structure from the RGB-D scans. It has been widely used for various 3D understanding tasks, such as semantic segmentation, 3D object detection, and scene classification. It has also been used to train and evaluate 3D reconstruction algorithms, including learning-based methods like Neural Radiance Fields (NeRF) and its variants. For neural rendering-based applications, ScanNet serves as an important source of data to train models that can understand and reconstruct indoor scenes. By leveraging the rich annotations provided by ScanNet, researchers can develop and evaluate methods that perform tasks such as view synthesis, relighting, and object insertion or removal. The availability of camera poses and calibration parameters allows neural rendering models to incorporate geometric constraints and multi-view consistency during training, leading to improved performance and generalization.

For neural rendering-based applications, ScanNet serves as an important source of data to train models that can understand and reconstruct indoor scenes. By leveraging the rich annotations provided by ScanNet, researchers can develop and evaluate methods that perform tasks such as view synthesis, relighting, and object insertion or removal. The availability of camera poses and calibration parameters allows neural rendering models to incorporate geometric constraints and multi-view consistency during training, leading to improved performance and generalization.

This dataset has been used for the following research articles:
\begin{itemize}
    \item GIRAFFE (Compositional Generative Radiance Fields)~\cite{niemeyer2021giraffe}
    \item DeRF (Decomposed Radiance Fields)~\cite{rebain2021derf}
\end{itemize}

\item \textbf{Synthetic-NeRF dataset}~\cite{martin2021nerf}, introduced by Martin-Brualla et al. in the paper "NeRF in the Wild: Neural Radiance Fields for Unconstrained Photo Collections," is a collection of synthetic scenes created to train and evaluate Neural Radiance Field (NeRF) models and their variants. The dataset consists of several synthetic scenes, each containing a 3D object rendered with different camera viewpoints, lighting conditions, and backgrounds. By using synthetic data, the dataset provides precise ground truth geometry, camera poses, and lighting parameters, which are difficult to obtain for real-world scenes. It is valuable for training and evaluating NeRF models, as it allows researchers to assess the performance of their methods under controlled conditions with known ground truth. This helps in understanding the limitations and strengths of the models, as well as identifying areas for improvement. The dataset is also useful for studying the impact of various factors, such as lighting, camera viewpoint, and object complexity, on the performance of NeRF-based methods.

For 3D reconstruction and neural rendering applications, Synthetic NeRF dataset serves as a benchmark to evaluate the quality of the generated reconstructions and renderings. The ground truth geometry and camera parameters enable quantitative evaluation of the reconstructed 3D models and synthesized views, allowing for a fair comparison between different methods. This helps in advancing the state-of-the-art in 3D reconstruction and neural rendering by identifying the best-performing models and techniques.

This dataset has been used for the following research articles:
\begin{itemize}
    \item GIRAFFE (Compositional Generative Radiance Fields)~\cite{niemeyer2021giraffe}
    \item PixelNeRF~\cite{yu2021pixelnerf}
\end{itemize}

\item \textbf{Blender Dataset} is a collection of synthetic scenes created using the open-source 3D software Blender. The dataset is commonly used to train and evaluate neural rendering models, including NeRF and its variants, for tasks such as 3D reconstruction and view synthesis. It typically consists of a variety of scenes, each containing 3D objects rendered with different camera viewpoints, lighting conditions, and backgrounds. By using synthetic data, the dataset provides precise ground truth geometry, camera poses, and lighting parameters that are difficult to obtain for real-world scenes.

The Blender dataset is valuable for training and evaluating neural rendering models, as it allows researchers to assess the performance of their methods under controlled conditions with known ground truth. This helps in understanding the limitations and strengths of the models, as well as identifying areas for improvement. The dataset is also useful for studying the impact of various factors, such as lighting, camera viewpoint, and object complexity, on the performance of neural rendering-based methods.

For 3D reconstruction and neural rendering applications, the Blender dataset serves as a benchmark to evaluate the quality of the generated reconstructions and renderings. The ground truth geometry and camera parameters enable quantitative evaluation of the reconstructed 3D models and synthesized views, allowing for a fair comparison between different methods. This helps in advancing the state-of-the-art in 3D reconstruction and neural rendering by identifying the best-performing models and techniques.

\item \textbf{ShapeNet dataset}~\cite{chang2015shapenet} is a large-scale collection of 3D object models, introduced in 2015. The dataset contains over 3 million 3D models across 55 categories, spanning a wide range of object types and complexity levels. The models in ShapeNet are represented as watertight 3D meshes, and many of them are annotated with additional information, such as semantic labels, part hierarchies, and alignment with 2D images. It is widely used for various 3D understanding tasks, including 3D shape classification, part segmentation, and 3D object detection. It has also been utilized for 3D reconstruction tasks, where researchers train models to predict the 3D shape of objects from input data such as 2D images, depth maps, or point clouds.

In the context of neural rendering applications such as Neural Radiance Fields (NeRF) and its variants, ShapeNet can be used to create synthetic datasets for training and evaluation. By rendering the 3D models in ShapeNet with different camera viewpoints, lighting conditions, and backgrounds, researchers can generate ground truth data for tasks like view synthesis and 3D reconstruction. This allows for the development and evaluation of NeRF-based methods under controlled conditions with known ground truth geometry and camera poses. This dataset enables the study of various factors that affect the performance of NeRF-based methods, such as object complexity, lighting, and camera viewpoint. Additionally, the dataset allows for the evaluation of NeRF models' ability to generalize across different object categories and shape variations.


\item \textbf{Kubric Synthetic Dataset:} The authors for the FORGE (Few-view Object Reconstruction that GEneralizes)~\cite{jiang2022few} constructed this dataset for computer vision community. However, Kubric is an open-source software package that enables the creation of custom synthetic datasets by rendering 3D scenes. By leveraging Kubric, researchers and practitioners can generate synthetic datasets tailored to their specific needs, including datasets for 3D reconstruction and neural rendering applications such as Neural Radiance Fields (NeRF) and its variants. This dataset allows users to define 3D scenes with various objects, materials, lighting conditions, and camera poses. The software package is built on top of the Blender rendering engine, which supports both real-time and offline, high-quality rendering. Kubric provides a Python API to programmatically create and manipulate 3D scenes, making it easy to generate large amounts of data with diverse variations.

For 3D reconstruction and neural rendering applications, custom synthetic datasets created using Kubric can be used for training and evaluation purposes. Researchers can render the defined 3D scenes with different camera viewpoints, lighting conditions, and backgrounds to generate ground truth data for tasks like view synthesis and 3D reconstruction. This allows for the development and evaluation of NeRF-based methods under controlled conditions with known ground truth geometry and camera poses. Using Kubric-generated synthetic datasets enables the study of various factors that affect the performance of NeRF-based methods, such as object complexity, lighting, and camera viewpoint. Additionally, the custom datasets allow for the evaluation of NeRF models' ability to generalize across different object categories and shape variations.

This dataset has been used for the following research articles:
\begin{itemize}
    \item GIRAFFE (Compositional Generative Neural Feature Fields)~\cite{niemeyer2021giraffe}
\end{itemize}

\item \textbf{ETH-XGaze}, introduced by ~\cite{zhang2020eth}, consists of high-resolution images with various head poses and gaze directions, and has been acquired by leveraging a multi-view camera system under different lighting conditions. It contains more than 1 million annotated images of 80 subjects, captured using a custom multi-camera rig in a variety of head poses and gaze directions. The training set consists of 80 subjects with a total of 756K frames. Each of these frames is composes of 18 different view camera view images. The test-set (also person-specific) contains 15 subjects with total of two hundred images for each subject with ground truth gaze labels. The dataset provides detailed annotations, including 3D gaze vectors, 3D head poses, 2D eye landmarks, and facial landmark locations.

While the primary purpose of the ETH-XGAZE dataset is for gaze estimation, it can also be used for related computer vision tasks, such as head pose estimation and facial landmark detection. However, the dataset is not specifically designed for 3D reconstruction or neural rendering applications like Neural Radiance Fields (NeRF) and its variants.

Nevertheless, ETH-XGAZE could potentially be used as a source of data for training and evaluating NeRF-based models on tasks involving facial geometry and appearance. The multi-view nature of the dataset and the availability of head pose information may allow researchers to develop models that can reconstruct 3D facial geometry or perform view synthesis of faces under various head poses and gaze directions. Additionally, the dataset could be used to study the effects of different head poses and gaze directions on the performance of NeRF-based methods.

\item \textbf{MPII Gaze dataset}~\cite{zhang2017s} is a dataset primarily designed for gaze estimation. It contains more than 213,000 images of 337 subjects, captured in natural, everyday settings using a variety of devices such as webcams, laptops, and mobile phones. The dataset provides annotations for 3D gaze vectors and head poses, as well as 2D eye landmarks and facial landmark locations. While the primary purpose of the MPII Gaze dataset is for gaze estimation, it can also be used for related computer vision tasks, such as head pose estimation and facial landmark detection. However, the dataset is not specifically designed for 3D reconstruction or neural rendering applications like Neural Radiance Fields (NeRF) and its variants.

Nevertheless, MPII Gaze could potentially be used as a source of data for training and evaluating NeRF-based models on tasks involving facial geometry and appearance. The availability of head pose information and the diverse set of images captured in natural settings may allow researchers to develop models that can reconstruct 3D facial geometry or perform view synthesis of faces under various head poses and gaze directions. Additionally, the dataset could be used to study the effects of different head poses and gaze directions on the performance of NeRF-based methods.

\item \textbf{MPIIFaceGaze}, curated by Zhang et al~\cite{zhang2017s} is based on gaze estimation dataset from MPIIGaze~\cite{zhang2015appearance}. MPIIFaceGaze contains 3000 face images with 2-D gaze labels for every 15 subjects. While the primary purpose of the MPIIFaceGaze dataset is for gaze estimation, it can also be used for related computer vision tasks, such as head pose estimation and facial landmark detection. However, the dataset is not specifically designed for 3D reconstruction or neural rendering applications like Neural Radiance Fields (NeRF) and its variants.

Nevertheless, MPIIFaceGaze could potentially be used as a source of data for training and evaluating NeRF-based models on tasks involving facial geometry and appearance. The availability of head pose information and the diverse set of images captured in natural settings may allow researchers to develop models that can reconstruct 3D facial geometry or perform view synthesis of faces under various head poses and gaze directions. Additionally, the dataset could be used to study the effects of different head poses and gaze directions on the performance of NeRF-based methods.
\item \textbf{ColumbiaGaze} (curated by Smith et al~\cite{smith2013gaze}) is a dataset composed of 5,880 high-resolution images from 56 people, captured in controlled laboratory conditions using a single high-resolution camera. The dataset provides annotations for head pose and gaze direction, as well as eye and facial landmark locations. The images were acquired with 5-fixed head poses and 21 fixed gaze directions per pose. While the primary purpose of the Columbia Gaze dataset is for gaze estimation, it can also be used for related computer vision tasks, such as head pose estimation and facial landmark detection. However, the dataset is not specifically designed for 3D reconstruction or neural rendering applications like Neural Radiance Fields (NeRF) and its variants.

Nevertheless, Columbia Gaze could potentially be used as a source of data for training and evaluating NeRF-based models on tasks involving facial geometry and appearance. The availability of head pose information and the high-resolution images captured under controlled conditions may allow researchers to develop models that can reconstruct 3D facial geometry or perform view synthesis of faces under various head poses and gaze directions. Additionally, the dataset could be used to study the effects of different head poses and gaze directions on the performance of NeRF-based methods.
\item \textbf{GazeCapture dataset} is a dataset with different poses for different amounts of illumination conditions and obtained through a crowd-sourcing endeavor~\cite{krafka2016eye}. Its text set only contains 150 people. The dataset contains more than 2.4 million images of over 1,500 subjects, captured using the front-facing cameras of various iOS devices in natural, everyday settings. The dataset provides annotations for 2D gaze coordinates on the device screen, along with facial landmark locations and head pose information. While the primary purpose of the GazeCapture dataset is for gaze estimation, it can also be used for related computer vision tasks, such as head pose estimation and facial landmark detection. However, the dataset is not specifically designed for 3D reconstruction or neural rendering applications like Neural Radiance Fields (NeRF) and its variants.

Nevertheless, GazeCapture could potentially be used as a source of data for training and evaluating NeRF-based models on tasks involving facial geometry and appearance. The availability of head pose information and the diverse set of images captured in natural settings may allow researchers to develop models that can reconstruct 3D facial geometry or perform view synthesis of faces under various head poses and gaze directions. Additionally, the dataset could be used to study the effects of different head poses and gaze directions on the performance of NeRF-based methods.

\item \textbf{DTU Robot Image Dataset}, created by the Technical University of Denmark (DTU), is a dataset designed for multi-view stereo (MVS) and 3D reconstruction tasks. It consists of 60 scenes captured using a robotic arm and a high-resolution camera. The dataset provides images, camera parameters (intrinsics and extrinsics), and dense ground-truth point clouds for each scene, making it an excellent resource for evaluating 3D reconstruction and neural rendering methods.

The usage of the DTU Robot Image Dataset for 3D reconstruction and different neural rendering applications such as Neural Radiance Fields (NeRF) includes:
\begin{enumerate}
    \item \textbf{Training:} Researchers can use the DTU dataset to train NeRF-based models on a diverse set of scenes with varying complexities and geometries. The high-resolution images and accurate camera parameters facilitate learning and generalization of the models across different scenes.
    \item \textbf{Evaluation:} The DTU dataset can be used to evaluate the performance of NeRF-based models on 3D reconstruction and view synthesis tasks. With dense ground-truth point clouds, researchers can quantitatively measure the accuracy of the reconstructed geometry and compare it with other methods.
    \item \textbf{Benchmarking: }The DTU Robot Image Dataset serves as a valuable benchmark for comparing different 3D reconstruction and neural rendering methods. It has been widely used in the research community, allowing for fair comparisons between different methods in terms of reconstruction quality, rendering speed, and other performance metrics.
\end{enumerate}
In summary, the DTU Robot Image Dataset is a useful resource for training, evaluating, and benchmarking 3D reconstruction and neural rendering methods such as NeRF. Its diverse set of scenes, high-resolution images, and accurate camera parameters provide a challenging testbed for developing and assessing the performance of various methods in reconstructing complex real-world scenes.

This dataset has been used for the following research articles:
\begin{itemize}
    \item Original NeRF (Neural Radiance Fields)~\cite{mildenhall2020nerf}
    \item MVSNeRF (Multi-View Stereo Neural Radiance Fields)~\cite{chen2021mvsnerf}
\end{itemize}

\item \textbf{CLEVR (Compositional Language and Elementary Visual Reasoning) dataset}, introduced by Johnson et al. in their paper "CLEVR: A Diagnostic Dataset for Compositional Language and Elementary Visual Reasoning," is designed for visual question answering (VQA) and scene understanding tasks. It contains images of 3D-rendered objects with varying shapes, colors, materials, and sizes, along with their corresponding scene graphs and semantic information.

Though CLEVR is not explicitly designed for 3D reconstruction or neural rendering, it can still be used for training and evaluating NeRF-based models and related applications. The potential usage of CLEVR for 3D reconstruction and different neural rendering applications such as Neural Radiance Fields includes:
\begin{itemize}
    \item \textbf{Training:} Researchers can use the CLEVR dataset to train NeRF-based models on synthetic scenes with known ground truth geometry and appearance. The dataset provides diverse and complex object configurations, which can help models learn to represent and render various 3D structures.
    \item \textbf{Evaluation:} The CLEVR dataset can be used to evaluate the performance of NeRF-based models on tasks such as view synthesis, 3D reconstruction, and object-centric rendering. Given the ground truth scene graphs and semantic information, quantitative evaluation of the model performance is possible.
    \item \textbf{Object-Centric Neural Rendering:} Since the CLEVR dataset focuses on object-centric scenes, it can be particularly useful for NeRF-based methods that aim to learn object-centric radiance fields. Researchers can leverage the dataset's scene graphs and semantic information to develop models that can synthesize novel views of individual objects or object compositions.
    \item \textbf{Scene Understanding:} The CLEVR dataset can also be used to study the role of scene understanding in neural rendering applications. Researchers can explore how incorporating semantic information and scene graphs into NeRF-based models may improve their performance in rendering and view synthesis tasks.
\end{itemize}
While the CLEVR dataset is primarily designed for visual question answering and scene understanding tasks, it can be repurposed for 3D reconstruction and neural rendering applications involving NeRF. Its diverse and complex object configurations, along with the availability of ground truth geometry and semantic information, make it a valuable resource for training and evaluating NeRF-based models in object-centric rendering and scene understanding scenarios.

\item \textbf{Matterport3D dataset}, introduced by Chang et al. in their paper "Matterport3D: Learning from RGB-D Data in Indoor Environments," is a large-scale dataset designed for various computer vision tasks, including 3D reconstruction, semantic segmentation, and scene understanding. The dataset contains 10,800 panoramic RGB-D images captured from 90 different indoor scenes, such as homes, offices, and public spaces. It provides high-resolution images, depth data, camera poses, and a 3D mesh reconstruction for each scene, making it a valuable resource for 3D reconstruction and neural rendering applications.

The usage of the Matterport3D dataset for 3D reconstruction and different neural rendering applications such as Neural Radiance Fields (NeRF) includes:
\begin{enumerate}
    \item Training: Researchers can use the Matterport3D dataset to train NeRF-based models on a diverse set of indoor scenes with complex geometries and lighting conditions. The high-resolution images, depth data, and accurate camera poses facilitate learning and generalization across various scenes and environments.
    \item Evaluation: The Matterport3D dataset can be used to evaluate the performance of NeRF-based models on tasks such as view synthesis, 3D reconstruction, and handling occlusions in indoor environments. The dataset's 3D mesh reconstructions enable quantitative evaluation of the model's performance in reconstructing complex scenes.
    \item Benchmarking: The Matterport3D dataset serves as a valuable benchmark for comparing different 3D reconstruction and neural rendering methods in indoor environments. By testing on the same dataset, researchers can compare their methods against others in terms of reconstruction quality, rendering speed, and other performance metrics.
    \item Scene Understanding: The Matterport3D dataset can also be used to study the role of scene understanding in neural rendering applications. Researchers can explore how incorporating semantic information from the dataset's semantic segmentation labels into NeRF-based models may improve their performance in rendering and view synthesis tasks.
\end{enumerate}
The Matterport3D dataset is a useful resource for training, evaluating, and benchmarking 3D reconstruction and neural rendering methods such as NeRF, particularly in indoor environments. Its diverse set of scenes, high-resolution images, depth data, and camera poses provide a challenging testbed for developing and assessing the performance of various methods in reconstructing complex real-world indoor scenes.

This dataset has been used for the following research articles:
\begin{itemize}
    \item GIRAFFE (Compositional Generative Neural Feature Fields)~\cite{niemeyer2021giraffe}
    \item DeRF (Decomposed Radiance Fields)~\cite{rebain2021derf}
\end{itemize}

\end{itemize}
(Work in Progress...)

\section{Objective Function}
\label{present:obj_funct}
Since objective functions (or mostly known as Cost or Loss function) in Machine learning play an integral part for training the model for different applications. Here, we discuss the
\begin{itemize}
\item \textbf{Functional Loss} is a generic term in the context of neural rendering and NeRF models. However, it can refer to any loss function that captures the difference between the target functionality or behavior of the generated data and the ground truth data. The equation for a functional loss would depend on the specific application and the desired behavior or property that needs to be preserved. Functional loss could potentially be used to ensure that the generated images or 3D representations have certain desired properties or behavior, such as smoothness, continuity, or consistency across different views. It has different applications in the domain of Computer Vision and Neural Rendering such as the following.
\begin{itemize}
    \item \textbf{Neural Rendering:} Functional loss could be used in NeRF models to enforce specific properties, such as temporal consistency in dynamic scenes or smoothness in the generated images.
    \item \textbf{Image-to-Image Translation:} In tasks like style transfer, functional loss could be used to ensure that the generated images maintain certain properties, such as preserving the structure of the input image or generating images with the desired style or texture.
    \item \textbf{3D Reconstruction:} Functional loss could be used in 3D reconstruction tasks to enforce geometric or topological properties, such as smoothness, compactness, or the preservation of sharp features.
\end{itemize}

As functional loss is a broad term and its definition varies depending on the application, it's difficult to pinpoint specific NeRF models, architectures, or extensions that use this loss function. However, it's worth noting that many NeRF-based models incorporate additional loss functions alongside reconstruction loss to enforce specific properties or behavior. For example, NeRF-W~\cite{martin2021nerf} uses perceptual loss in addition to reconstruction loss to improve the quality of the generated images. Other NeRF models may incorporate similar loss functions to enforce the desired properties or behavior.

\item \textbf{Reconstruction Loss (or Photometric Loss)}, is a measure of the discrepancy between the predicted image and the ground truth image. It is commonly used in image synthesis, neural rendering, and other computer vision tasks. In the context of NeRF, reconstruction loss is used to penalize the difference between the predicted colors and the ground truth colors for a given set of rays.
The reconstruction loss is usually calculated as the MSE between the predicted and ground truth colors, averaged over all pixels in the image:
\begin{equation}
    \label{eqn:101}
    L_{recons} = 
    \frac{1}{N} \sum ||C_{pred} - C_{gt}||^2
\end{equation}

where $N$ is the number of pixels, $C_{pred}$ is the predicted color, and $C_{gt}$ is the ground truth color.

Applications of this function in the Neural Rendering and other tasks include:
\begin{itemize}
\item \textbf{Neural Rendering:} Reconstruction loss is widely used to train NeRF models and their variants. It ensures that the predicted colors align well with the ground truth colors, leading to accurate and realistic view synthesis.
\item \textbf{Image-to-Image Translation:} In tasks such as style transfer, reconstruction loss is used to preserve the content of the input image while applying the desired style.
\item \textbf{Depth Estimation:} Reconstruction loss can be used in combination with depth consistency loss to ensure the predicted depth maps align with the ground truth depth maps.
\end{itemize}

Almost all NeRF models, architectures, and extensions use reconstruction loss as their primary loss function, including but not limited to:

\begin{itemize}
\item The original NeRF (Neural Radiance Fields)~\cite{mildenhall2020nerf}
\item NeRF-W (NeRF in the Wild)~\cite{martin2021nerf}
\item MVSNeRF (Multi-View Stereo Neural Radiance Fields)~\cite{chen2021mvsnerf}
\item GIRAFFE (Compositional Generative Neural Feature Fields)~\cite{niemeyer2021giraffe}
\item DeRF (Decomposed Radiance Fields)~\cite{rebain2021derf}
\item NeRFies~\cite{park2021nerfies}
\item PixelNeRF~\cite{yu2021pixelnerf}
\end{itemize}
Reconstruction loss is a fundamental component of the training process for NeRF-based models, as it helps the models learn to generate accurate and realistic images that match the ground truth data.

\item \textbf{Perceptual Loss}, also known as feature loss or content loss, is a measure of the discrepancy between the high-level features of the predicted image and the ground truth image. It is commonly used in image synthesis, neural rendering, and other computer vision tasks where preserving high-level features is important. Perceptual loss is often computed using pre-trained deep neural networks, such as VGG or ResNet, as feature extractors.

The perceptual loss is usually calculated as the MSE between the high-level features of the predicted and ground truth images, extracted from the pre-trained network as shown in the eqn~\ref{eqn:102}
\begin{equation}
    \label{eqn:102}
    L_{percp} = 
    \frac{1}{N} \sum ||F(C_{pred}) - F(C_{gt})||^2
\end{equation}
where $N$ is the number of pixels, $F(\cdot)$ is the feature extraction function, $C_{pred}$ is the predicted color, and $C_{gt}$ is the ground truth color.

Applications of this function in the Neural Rendering and other tasks include:
\begin{itemize}
\item \textbf{Neural Rendering:} Perceptual loss can be used in NeRF models and their variants to ensure that the predicted images preserve high-level features and structures, leading to more perceptually pleasing view synthesis.
\item \textbf{Image-to-Image Translation:} In tasks such as style transfer or image super-resolution, perceptual loss is used to ensure that the translated images maintain the content and structure of the input image while applying the desired style or enhancing the resolution.
\item \textbf{GANs:} Perceptual loss is often used in GANs to help generate images that are more visually pleasing and have more realistic high-level features.
\end{itemize}

While perceptual loss is not as commonly used in NeRF-based models as reconstruction loss, there are some NeRF models and extensions that leverage perceptual loss to improve the quality of the generated images:

\begin{itemize}
\item NeRF-W (NeRF in the Wild)~\cite{martin2021nerf} uses a combination of reconstruction loss and perceptual loss to improve the quality of the generated images in unconstrained settings.
\item Non Rigid NeRF~\cite{tretschk2021non} uses perceptual loss along with reconstruction loss to ensure that the generated images maintain the high-level structures and appearance of the ground truth data.
\end{itemize}
These are a few instances of NeRF models and extensions that use perceptual loss. The use of perceptual loss can help improve the visual quality of the generated images by focusing on high-level features, which are often more important for human perception.

\item \textbf{Disentanglement Loss} aims to encourage a learned representation to capture distinct factors of variation in the data, such as geometry, appearance, illumination, or motion. This loss function is often used in unsupervised and self-supervised learning tasks, where ground truth labels for the factors of variation are not available.

The equation for disentanglement loss depends on the specific implementation and the factors of variation that need to be disentangled. One common approach for disentanglement loss is the $\beta$-VAE (Variational Autoencoder) framework, which modifies the standard VAE objective with an additional weighting factor $\beta$ as shown in the eqn~\ref{eqn:103}
\begin{equation}
    \label{eqn:103}
L_{disent} = L_{reconst} + \beta * L_{KL}
\end{equation}

Here, $L_{reconst}$ is the reconstruction loss, $L_{KL}$ is the Kullback-Leibler divergence term that enforces the latent distribution to be close to a prior distribution (e.g., a standard Gaussian), and $\beta$ is a hyperparameter that controls the trade-off between reconstruction and disentanglement.

Usage for Neural Rendering and other tasks are as follows.

\begin{itemize}
\item \textbf{Neural Rendering:} Disentanglement loss can be used in NeRF models and their variants to separate distinct factors of variation, such as geometry and appearance, enabling more controllable editing and manipulation of the generated scenes.
\item \textbf{Unsupervised Representation Learning:} Disentanglement loss is often used in unsupervised learning tasks, such as autoencoders and VAEs, to learn meaningful and interpretable representations of the data.
\item \textbf{Image-to-Image Translation:} Disentanglement loss can be used in style transfer tasks to separate content and style factors, enabling the transfer of one factor while preserving the other.
\end{itemize}

In the context of NeRF models, architectures, and extensions that use disentanglement loss or a similar concept:

\begin{itemize}
    \item DeRF~\cite{rebain2021derf} explicitly disentangles the geometry and appearance factors in the learned radiance field representation, enabling more controllable editing and manipulation of the generated scenes.
    \item GIRAFFE~\cite{niemeyer2021giraffe} uses a compositional approach to separate object geometry and appearance, which allows for more flexible scene manipulation and editing.
    \item Object-Centric Neural Scene Rendering~\cite{guo2020object}: This work disentangles object geometry and appearance by learning separate latent spaces for each factor, enabling more controllable and intuitive scene editing.

\end{itemize}
These are a few examples of NeRF models and extensions that use disentanglement loss or a similar concept to separate distinct factors of variation in the learned representations. Disentangling factors of variation can be beneficial for a variety of tasks, as it leads to more interpretable, controllable, and flexible representations.

\item \textbf{Color Loss}, also known as color consistency loss, measures the difference between the predicted colors and the ground truth colors in a rendered image. In the context of Neural Radiance Fields (NeRF), color loss is essentially a part of the reconstruction loss, which measures the difference between the rendered radiance and the observed radiance from the input images. The color loss can be formulated using various metrics, with Mean Squared Error (MSE) being the most common.

For a given input ray r and a set of input images, NeRF models predict the radiance for each point along the ray. The color loss can be calculated as the mean squared error between the predicted color $C_{pred}$ and the ground truth color $C_{gt}$ as  shown in the eqn~\ref{eqn:104}

\begin{equation}
    \label{eqn:104}
L_{color} = ||C_{pred} - C_{gt}||^2
\end{equation}

Applications of this function in the Neural Rendering and other tasks include:
\begin{itemize}
\item \textbf{Neural Rendering:} In NeRF models and other neural rendering approaches, color loss is used to ensure that the generated images or 3D representations have accurate and consistent colors compared to the input images.
\item \textbf{Image-to-Image Translation:} Color loss can be used in tasks like style transfer or image synthesis to ensure that the generated images maintain certain color properties from the input image or the target style.
\item \textbf{3D Reconstruction:} Color loss can be used in 3D reconstruction tasks to measure the difference between the colors of the reconstructed 3D model and the ground truth model, encouraging accurate and detailed 3D reconstructions.
\end{itemize}

As mentioned earlier, color loss is essentially a part of the reconstruction loss in the context of NeRF models. Almost all NeRF models, including the original NeRF~\cite{mildenhall2020nerf}, NeRF-W~\cite{martin2021nerf}, and many other NeRF extensions, employ some form of color loss as a part of the reconstruction loss to ensure that the generated images or 3D representations have accurate and consistent colors compared to the input images.

\end{itemize}
(Work in Progress...)

\section{Evaluation Metrics}
\label{present:evalmetrics}
Since, we are discussing a comprehensive survey of paper, it is vital for readers to understand what are the metrics used to evaluate Radiance Fields. The main focus of this discussion is about PSNR (Peak Signal to Noise Ratio), SSIM (structural similarity index measure), LPIPS (Learned Perceptual Image Patch Similarity)~\cite{zhang2018unreasonable}, Chamfer distance, etc.

\begin{itemize}
\item \textbf{PSNR (Peak-Signal-to-Noise Ratio)} is a widely-used metric to evaluate image quality, particularly for image compression and restoration tasks. It measures the ratio between the maximum possible signal power (i.e., the maximum value of pixel intensity) and the power of the corrupting noise (i.e., the mean squared error between the original and reconstructed images). Higher PSNR values indicate better image quality. However, PSNR can be a poor indicator of perceptual similarity, as it does not capture human perception of image quality accurately.
\item \textbf{SSIM (Structural Similarity Index)} is a more perceptually relevant metric that compares two images based on their luminance, contrast, and structure. It takes into account changes in pixel intensities, spatial dependencies, and contrasts in texture. SSIM ranges from -1 to 1, where 1 indicates a perfect match between the original and reconstructed images. SSIM provides a better assessment of image quality concerning human perception compared to PSNR.
\item \textbf{LPIPS (Learned Perceptual Image Patch Similarity)}~\cite{zhang2018unreasonable} is a  metric to evaluate the perceptual similarity of the rendered views/poses when compared to the ground truth for that particular image from that particular viewing direction. It is a perceptual metric that uses deep learning to measure the similarity between two images. It is based on the features extracted from a pre-trained convolutional neural network (CNN) and is designed to align better with human judgments of image similarity. Lower LPIPS values indicate higher perceptual similarity between the images being compared. LPIPS is more robust to small geometric and textural differences and is better suited for evaluating generative models and image synthesis tasks.
\item \textbf{NIQE (Natural Image Quality Evaluator)} is a metric for evaluating the quality of natural images. It measures the degree of naturalness of an image by comparing its statistics to those of a large database of natural images. The formula for NIQE is not publicly available, but it involves computing various image statistics such as the mean, standard deviation, and higher-order moments.

\item \textbf{Chamfer Distance} is a geometric metric used to evaluate the similarity between two point sets, typically employed in 3D reconstruction tasks. It measures the average distance between each point in one set to its nearest neighbor in the other set. Lower Chamfer Distance values indicate better geometric alignment between the point sets. While it is not directly used to evaluate image quality in Radiance Fields, Chamfer Distance can be used to assess the quality of the underlying 3D geometry reconstructed by the model.

\item \textbf{KID (Kernel Inception Distance)} is a metric used to evaluate the quality and diversity of generated images in generative models, such as GANs and VAEs. KID compares the feature distributions of real images and generated images by measuring the distance between the embeddings of these images in a feature space, typically obtained from an Inception network (a deep convolutional neural network pretrained on a large-scale image classification task, such as ImageNet).

The main idea behind KID is to measure the similarity between the distributions of real and generated images by comparing the statistics of their feature embeddings. The distance between these distributions is computed using the MMD (Maximum Mean Discrepancy) metric with a RBF (radial basis function) kernel.

To compute the KID, the following steps are typically taken:

\begin{enumerate}
    \item \textbf{Feature Extraction:} Extract feature embeddings for both the real images and generated images using an Inception network. This involves passing the images through the Inception network and extracting the activations from one of the intermediate layers, which serve as the feature embeddings.
    \item \textbf{Compute MMD:} Calculate the Maximum Mean Discrepancy (MMD) between the feature embeddings of the real and generated images. MMD is a metric that measures the distance between two probability distributions by comparing the means of samples drawn from these distributions. The MMD can be computed as shown in the eqn.~\ref{eqn:n-4}.
    \begin{equation}
    \label{eqn:n-4}
        MMD^2 = ||\mu_p - \mu_q||^2 + 2 * \sum (k(p_i, q_j) - k(p_i, p_j) - k(q_i, q_j))
    \end{equation}
    
    where $\mu_p$ and $\mu_q$ are the means of the feature embeddings for the real images ($p$) and generated images ($q$), respectively, and $k(., .)$ is the RBF kernel function applied to pairs of feature embeddings.
    
    \item \textbf{Compute KID:} The KID is simply the MMD value computed in step 2.
\end{enumerate}
KID has several advantages over other metrics for evaluating the quality and diversity of generated images, such as the Inception Score (IS) and Frechet Inception Distance (FID). For example, KID is unbiased, which means it provides a more reliable estimate of the true distance between the distributions of real and generated images, and it is less sensitive to the choice of the number of samples used for the computation. Additionally, KID is computationally efficient and can be calculated using fewer samples than other metrics, making it a popular choice for evaluating generative models in various research studies and applications.

\item \textbf{FID (Frechet Inception Distance)}~\cite{heusel2017gans} is another metric used to assess generative models, comparing the feature distributions of real and generated images using the Inception-v3 network. FID captures both the quality and diversity of generated images and is considered a more reliable metric compared to Inception Score. Like Inception Score, FID can be applied to 2D projections of Radiance Fields-generated scenes to assess their visual quality and diversity.

\item \textbf{ATE (Absolute Trajectory Error)} is a metric used to evaluate the accuracy of an estimated trajectory in comparison to a ground truth trajectory. ATE is commonly used in the evaluation of SLAM (Simultaneous Localization and Mapping) algorithms, visual odometry, and other robotics or computer vision tasks where a system's position and orientation are estimated over time.

ATE is computed by comparing the estimated trajectory and the ground truth trajectory, taking into account both the position and the orientation of the system at each time step. To compute the ATE, the following steps are typically taken:
\begin{enumerate}
    \item Alignment: Since the estimated trajectory and the ground truth trajectory might have different scales, rotations, or translations, they need to be aligned before computing the error. This alignment is usually achieved by finding a transformation matrix (a combination of rotation, translation, and scale) that minimizes the difference between the two trajectories.
    \item Error Calculation: After the alignment, the ATE is calculated as the RMSE (root mean squared error) between the aligned estimated trajectory and the ground truth trajectory. This involves computing the Euclidean distance between the corresponding positions in the two trajectories and averaging the squared distances over all time steps. Then, the square root of the average is taken to obtain the final ATE value.
\end{enumerate}
Mathematically, given an estimated trajectory $P = {p_1, p_2, ..., p_n}$ and a ground truth trajectory $Q = {q_1, q_2, ..., q_n}$ (after alignment), the ATE can be calculated as shown in the eqn.~\ref{eqn:n-3}.

\begin{equation}
    \label{eqn:n-3}
    ATE = sqrt\left(\frac{1}{n} * \sum ||p_i - q_i||^2\right)
\end{equation}
where $n$ is the number of time steps, $p_i$ and $q_i$ are the positions at time step i in the estimated and ground truth trajectories, respectively, and $||p_i - q_i||$ is the Euclidean distance between $p_i$ and $q_i$.

ATE is a useful metric for evaluating the performance of various algorithms that estimate a system's trajectory over time, as it provides an overall measure of the accuracy and consistency of the estimated positions and orientations compared to the ground truth.

\item \textbf{RPE (Relative Pose Error)} ($\textrm{RPE}_{r}$ (Relative Rotation Error), $\textrm{RPE}_{t}$ (Relative Translation Error)) is a metric used to evaluate the accuracy of an estimated trajectory in comparison to a ground truth trajectory, taking into account both the position and orientation of a system over a specific time interval. RPE is commonly used in the evaluation of SLAM (Simultaneous Localization and Mapping) algorithms, visual odometry, and other robotics or computer vision tasks where a system's position and orientation are estimated over time.

Unlike Absolute Trajectory Error (ATE), which compares the estimated and ground truth trajectories directly, RPE focuses on the relative changes in pose between pairs of time steps. This makes it a more suitable metric for evaluating systems where the global consistency of the trajectory is less important than the local consistency.

To compute the RPE, the following steps are typically taken:
\begin{enumerate}
    \item Select Time Intervals: Choose a specific time interval ($\Delta$) over which the relative pose changes will be evaluated. This can be a fixed interval or a variable one, depending on the application.
    \item Compute Relative Poses: For both the estimated trajectory and the ground truth trajectory, compute the relative pose changes between pairs of poses separated by the chosen time interval.
    \item Error Calculation: After computing the relative poses, calculate the error between the estimated relative poses and the ground truth relative poses. This can be done using various metrics, such as the Euclidean distance for position errors, and the angular distance for orientation errors. The overall RPE can then be calculated as the RMSE between the corresponding relative pose errors.
\end{enumerate}

Mathematically, given an estimated trajectory $P = {p_1, p_2, ..., p_n}$ and a ground truth trajectory $Q = {q_1, q_2, ..., q_n}$, and a time interval delta, the RPE can be calculated as shown in the eqn~\ref{eqn:n-2}

\begin{equation}
    \label{eqn:n-2}
    RPE = sqrt\left(\frac{1}{n-\Delta} * \sum (||\Delta p_i - \Delta q_i||^2) \right)
\end{equation}

where n is the number of time steps, $\Delta p_i$ and $\Delta q_i$ are the relative pose changes between time steps $i$ and $i+\Delta$ in the estimated and ground truth trajectories, respectively, and $||\Delta p_i - \Delta q_i||$ is the error metric used to measure the difference between the relative poses (e.g., Euclidean distance for position errors, angular distance for orientation errors).

RPE is a useful metric for evaluating the performance of various algorithms that estimate a system's trajectory over time, as it provides a measure of the local consistency of the estimated poses compared to the ground truth, which can be more informative than global consistency measures like ATE in certain applications.

\item \textbf{AVI (Across View-direction Inconsistency)} was introduced in Bahat et al~\cite{bahat2022neural} to demonstrate the advantages of performing operations in the volumetric domain to produce renderings that are geometrically consistent for different viewing directions. AVI quantifies the pairwise-geometrical inconsistency between test frames adjacent to each other  (in terms of viewing direction). This is done by leveraging Ground Truth Optical Flow extracted from every adjacent frame pairs in the synthetically rendered scene. The authors warp the first frame onto the second and calculate the induced per-pixel RMS difference across different channels to get per-pixel error map. Finally, the Optical Flow is used to mask-out map regions corresponding to the objects waning and waxing between two frames to compute an error map which can then be averaged to calculate AVI. The AVI can be denoted using the eq.~\ref{eq:n-1} given below.

\begin{equation}\label{eq:n-1}
\textrm{AVI} = \frac{1}{n} \sum_{r \in \hat{i}_{sk}^{j}(r)}{\textrm{Mask}_{j, j+1} \cdot || \textrm{Warp}(\hat{i}_{sk}^{j})(r) - \hat{i}_{sk}^{j+1}(r)||_{2}}
\end{equation}
where $\textrm{Warp}(\cdot)$ represents the warping operation in accordance with the Optical Flow, $\textrm{Mask}_{j, j+1}$ is binary mask to filter out inconsistent image regions, $n$ is the number of pixels, and $\hat{i}_{sk}^{j}(r)$ and $\hat{i}_{sk}^{j}(r)$ depict the two adjacent frames during Super Resolution.
This metric was used to determine the consistency for the approach by Bahat et al~\cite{bahat2022neural} discussed later in this survey.\\

Both the optical flow for forward flow $f_{j \rightarrow j+1}$ and backward flow $f_{j+1 \rightarrow j}$  amongst all adjacent view-direction pair j, j+1. The warping function uses the backward optical flow $f_{j+1 \rightarrow j}$ between adjacent viewing directions by the eq.~\ref{eq:n}.

\begin{equation}\label{eq:n}
\textrm{Warp}(\hat{i}_{sk}^{j})(r) = \hat{i}_{sk}^{j}(r + f_{j+1 \rightarrow j} (r))
\end{equation}

where $r$ denotes the 2D pixel location.

The warping can produce ghosting artifacts due to occlusions (pixels in frame $j$ with no correspondence in $j+1$). Masking prevents  ghosting artifacts from dominating the AVI metric and hence an error mask was introduced. This is done by first counting the number of pixels matched using forward flow for the same pixel $r$ as given by the eq.~\ref{eq:n+1}:

\begin{equation}\label{eq:n+1}
\textrm{Count}_{j,j+1}(r) = \#\{q | \lfloor q + f_{j \rightarrow j+1} (q) \rfloor = r\}
\end{equation}

The pixels without any correspondence from viewing direction $j+1$ to viewing direction $j$ is defined according to mask equation in eq.~\ref{eq:n+2}.

\begin{equation}\label{eq:n+2}
    \tilde{\textrm{Mask}}_{j, j+1} = \begin{cases}
    0, & \text{if $\textrm{Count}_{j, j+1}(r) = 0$}\\
    1, & \text{otherwise}
\end{cases}
\end{equation}

The floor operation while computing $\textrm{Count}$ can lead to ghosting operations due to sub-pixel occlusions, the authors apply morphological closing followed by erosion to the mask by leveraging cross structure element of unit radius. Hence, we get the final error mask as given in the eq.~\ref{eq:n+3}.

\begin{equation}\label{eq:n+3}
\textrm{Mask}_{j, j+1} = \textrm{erosion}(\textrm{closing}(\tilde{\textrm{Mask}}_{j, j+1}))
\end{equation}
Hence, during the computation of AVI score, only unmasked elements are considered during computing mean. So, only the values for $n$ as depicted in eq.~\ref{eq:n+4} are considered in eq.~\ref{eq:n-1}
\begin{equation}\label{eq:n+4}
n = \#\{r|\textrm{Mask}_{j,j+1}(r) = 1\}
\end{equation}
Figure~\ref{fig:n} displays the effects of the raw masks and the different morphological and it can be seen that the same motion vectors and hence, same masks were used for computing AVI score for each method in the paper.

\begin{figure}[hbt!]
\begin{subfigure}{.24\textwidth}
  \centering
  \centerline{\includegraphics[width=20mm]{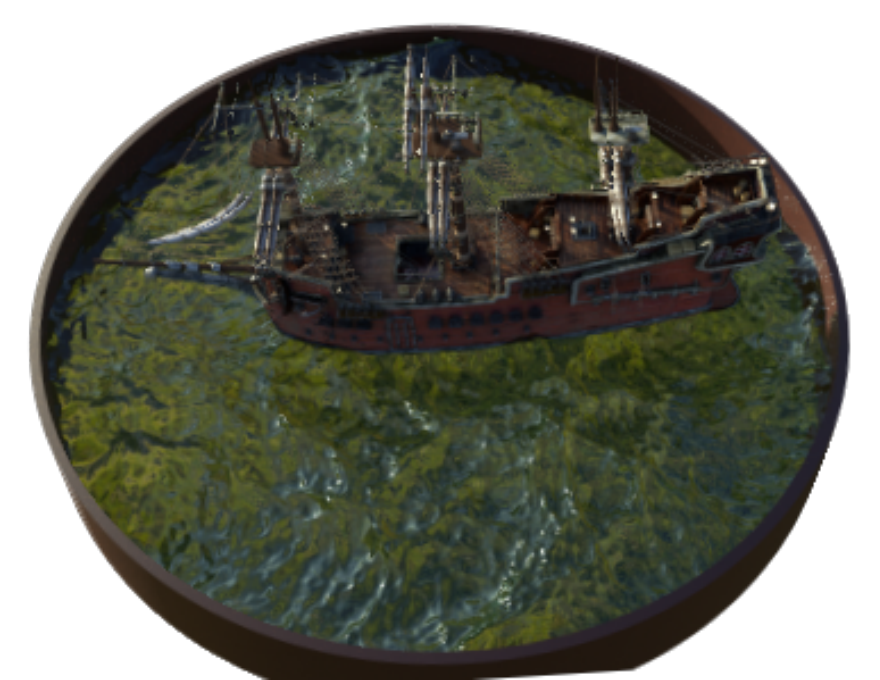}}  
  \caption{No Masking}
  \label{fig:subn(1)}
\end{subfigure}
\begin{subfigure}{.24\textwidth}
  \centering
  \centerline{\includegraphics[width=20mm]{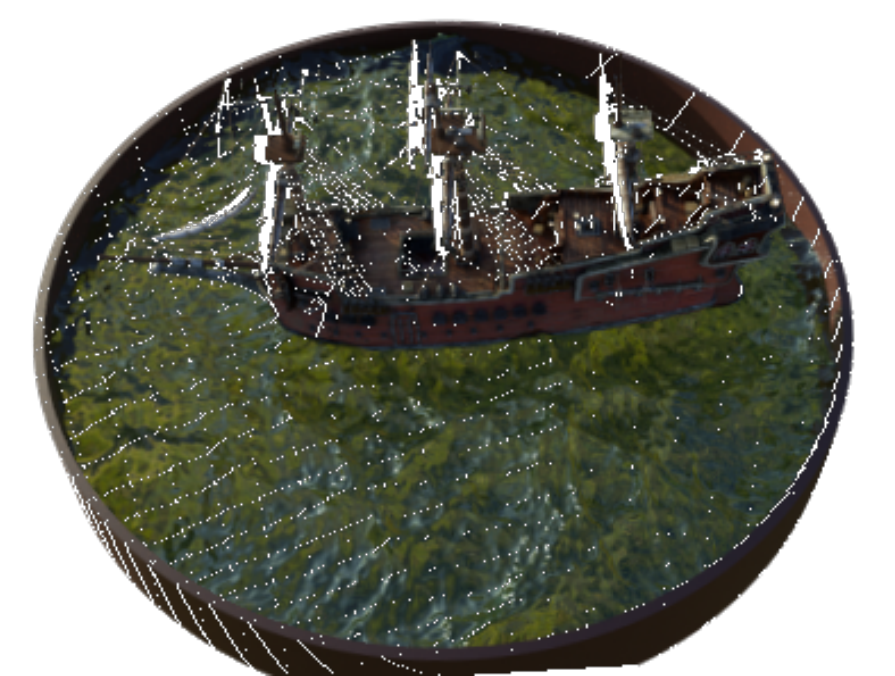}}  
  \caption{With $\tilde{Mask_{j, j+1}}$}
  \label{fig:subn(2)}
\end{subfigure}
\begin{subfigure}{.24\textwidth}
  \centering
  \centerline{\includegraphics[width=20mm]{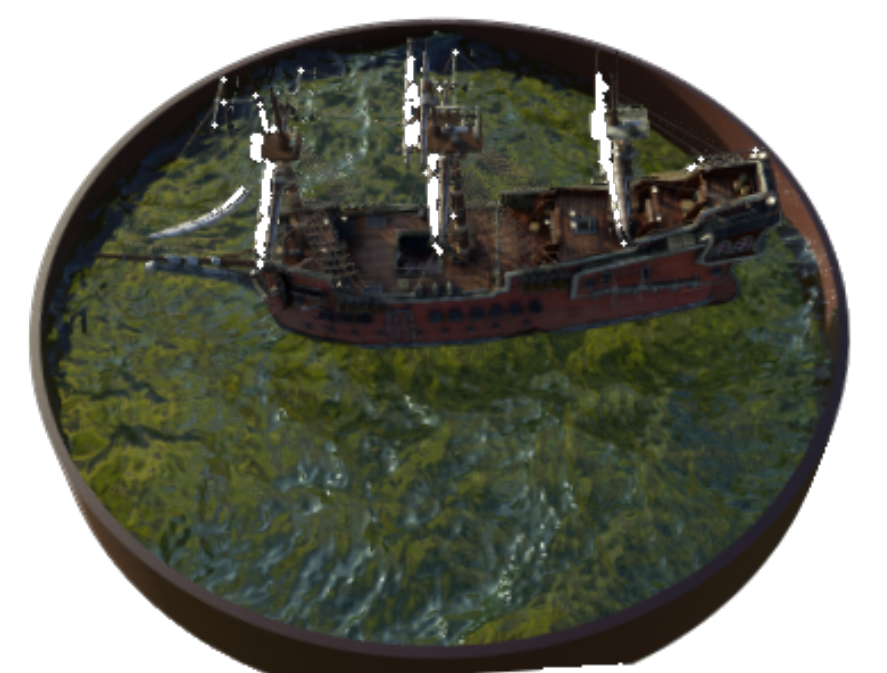}}  
  \caption{$\textrm{closing}(\tilde{\textrm{M}}_{j, j+1})$}
  \label{fig:subn(3)}
\end{subfigure}
\begin{subfigure}{.24\textwidth}
  \centering
  \centerline{\includegraphics[width=20mm]{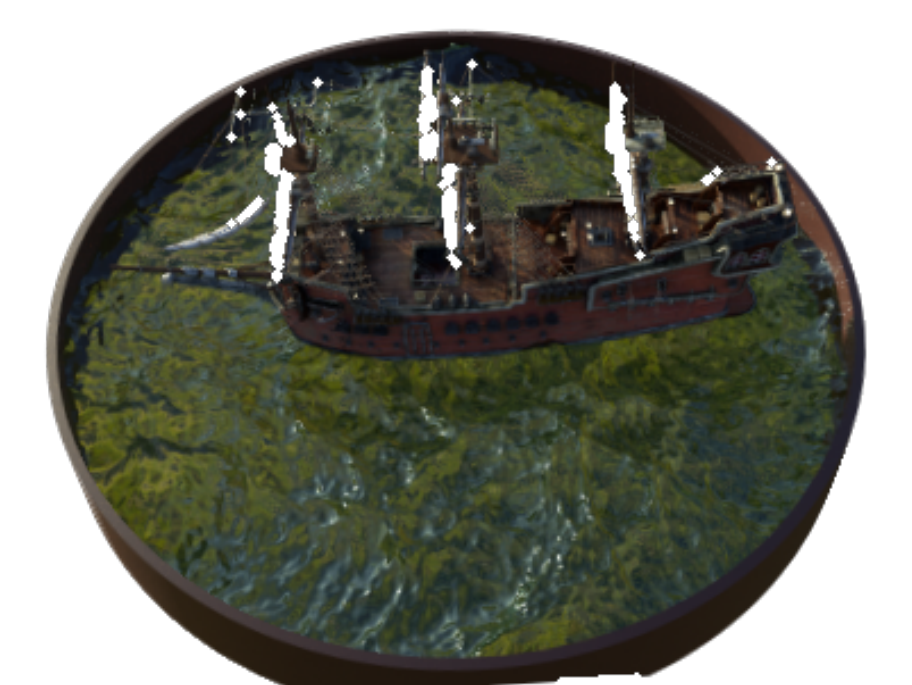}}  
  \caption{$\textrm{closing+diln}$}
  \label{fig:subn(4)}
\end{subfigure}
\caption{\textbf{Effects of Mask ($\tilde{\textrm{M}}_{j, j+1}$ or $\tilde{\textrm{Mask}}_{j, j+1}$).} To remove ghosting artifacts from the ship mast's (warped image), masking was employed before the error calculation. (b-d) depicts the masking effect with/without processing and morphological image operators; (d) depicts the fully processed mask that yields significantly reduced ghosting artifacts}
\label{fig:n}
\end{figure}
\end{itemize}
(Work in Progress...)

\section{Different Explored Applications}
\label{present:diffexploredappns}

This literature tries to list the different applications for NeRFs. Further, a rough statistics for the applications of Neural Radiance Fields have been shown in the fig~\ref{fig:figx+1}. Further, we discuss these applications briefly. This can help build the basis for how the future of NeRFs and Neural Rendering.

\afterpage{
\begin{figure*}[!htp]
    \centering
    \includegraphics[width=0.6\textwidth]{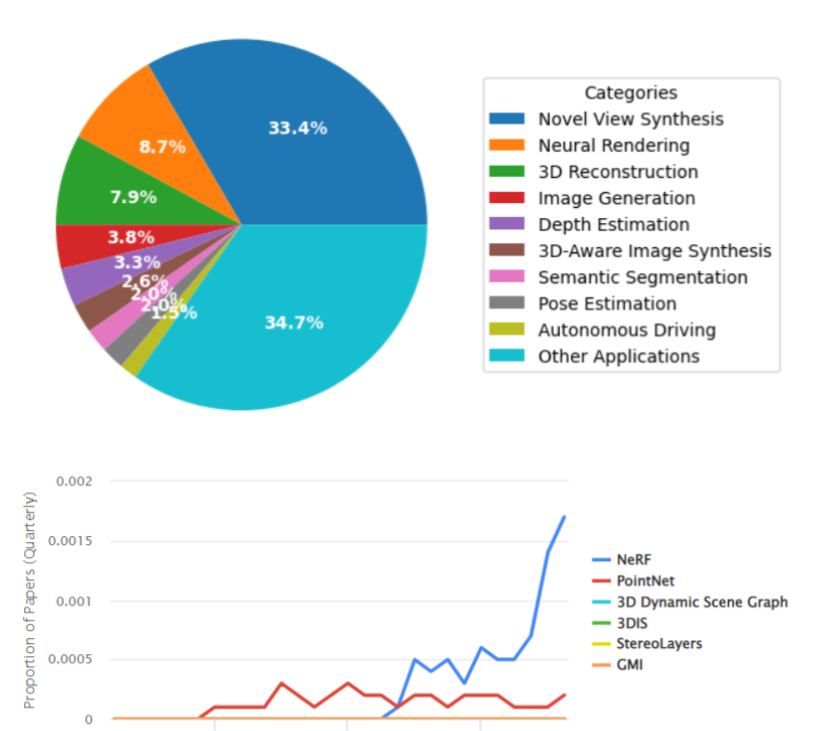}
    \caption[Caption for LOF]{The statistics related to NeRF-based models, extensions and improvements based on applications\footnotemark}
    \label{fig:figx+1}
\end{figure*}
\footnotetext{These statistics and graphs have been taken from~\url{https://paperswithcode.com/method/nerf}}
}

\begin{itemize}
\item \textbf{Shape Estimation} involves recovering the 3D structure of an object or scene from 2D images or point clouds. Applications include 3D modeling, reverse engineering, medical imaging, and robotics. Techniques such as stereo vision, photometric stereo, and depth sensing are used to estimate the shape of objects.
\item \textbf{Pose Reconstruction} aims to recover the 3D position and orientation of an object or a camera in a coordinate system. This task is crucial for applications like robotics, autonomous navigation, augmented reality, and motion capture. Methods such as bundle adjustment, visual odometry, and SLAM can be used for pose reconstruction.
\item \textbf{Pose Estimation} deals with estimating the position and orientation of a specific object or human body parts in 2D or 3D space. Applications include human-computer interaction, animation, sports analysis, and surveillance. Techniques used for pose estimation include keypoint detection, part-based models, and deep learning-based approaches.
\item \textbf{View Reconstruction} aims to generate novel views of a scene or object from a set of input images. This technique is used in virtual reality, image-based rendering, and video synthesis. Methods such as light field rendering, image-based rendering, and neural rendering (e.g., NeRF) can be employed for view reconstruction.
\item \textbf{Multi-View Reconstruction} is a 3D reconstruction technique that uses multiple images of a scene or object captured from different viewpoints. Applications include 3D modeling, computer-aided design, and cultural heritage preservation. Techniques such as SfM, MVS, and volumetric fusion are used for multi-view reconstruction.
\item \textbf{Super-Resolution} is the process of enhancing the resolution and quality of an image by combining information from multiple low-resolution images or by learning a mapping from low-resolution to high-resolution images. Applications include satellite imaging, medical imaging, and video upscaling. Techniques such as sparse representation, deep learning-based methods (e.g., SRGAN), and multi-frame super-resolution can be used for this task.
\item \textbf{Gaze Redirection} refers to the process of manipulating the appearance of a person's eyes in an image or video to change the perceived gaze direction. This technique is useful in video conferencing, eye-tracking studies, and virtual reality. Approaches for gaze redirection include geometric transformations, image warping, and generative models (e.g., GANs).
\end{itemize}
(Work in Progress...)

\section{Neural Radiance Fields in Future}
(Work in Progress...)
\label{future}

\section{Conclusion}
\label{conclusion}
(Work in Progress...)

\bibliographystyle{splncs04}
\bibliography{Mybibliography}

\end{document}